\documentclass[10pt,twocolumn,letterpaper]{article}

\usepackage{iccv}
\usepackage{times}
\usepackage{epsfig}
\usepackage{graphicx}
\usepackage{amsmath}
\usepackage{amssymb}
\usepackage{comment}
\usepackage{capt-of}

\usepackage[pagebackref=true,breaklinks=true,colorlinks,bookmarks=false]{hyperref}

\iccvfinalcopy 

\def\iccvPaperID{8652} 

\usepackage{multirow}
\usepackage{hhline}
\usepackage{array}
\usepackage{tabu}
\usepackage{booktabs}
\usepackage{amsmath}
\newcolumntype{C}[1]{>{\centering\arraybackslash}p{#1}}
\newcommand\norm[1]{\left\lVert#1\right\rVert}

\DeclareMathOperator*{\argmin}{arg\,min}

\makeatletter
\newcommand{\thickhline}{%
    \noalign {\ifnum 0=`}\fi \hrule height 1pt
    \futurelet \reserved@a \@xhline
}
\makeatletter
\newcommand{\thickvline}{%
    \noalign {\ifnum 0=`}\fi \vrule height 1pt
    \futurelet \reserved@a \@xvline
}
\newcolumntype{"}{@{\hskip\tabcolsep\vrule width 1pt\hskip\tabcolsep}}
\makeatother

\usepackage{xcolor}
\definecolor{darkg}{rgb}{0,0.6,0}
\definecolor{lgreen}{rgb}{0,0.5,0}
\definecolor{egreen}{rgb}{0.1,0.6,0.5}
\definecolor{amethyst}{rgb}{0.6, 0.4, 0.8}
\definecolor{orange}{rgb}{0.93,0.48,0.03}
\definecolor{blueviolet}{rgb}{0.54,0.16,0.88}

\newcommand*\ruleline[2]{\par\noindent\raisebox{.8ex}{\makebox[{#1}]{\hrulefill\hspace{1ex}\raisebox{-.8ex}{#2}\hspace{1ex}\hrulefill}}}

\begin{document}

\title{StyleCLIP: Text-Driven Manipulation of StyleGAN Imagery}

\author{
Or Patashnik$^\dagger$\thanks{Equal contribution, ordered alphabetically} \hspace{6mm}  Zongze Wu$^{\ddagger*}$ \hspace{6mm}  Eli Shechtman$^\mathsection $ \hspace{6mm}  Daniel Cohen-Or$^\dagger$ \hspace{6mm}  Dani Lischinski$^\ddagger$ \\
$^\ddagger$Hebrew University of Jerusalem \hspace{10mm}  $^\dagger$Tel-Aviv University \hspace{10mm} $^\mathsection$Adobe Research
}

\twocolumn[{%
	\renewcommand\twocolumn[1][]{#1}%
	\maketitle
	\begin{center}
		\vspace{-8mm}
        \setlength{\tabcolsep}{2pt}
        \begin{tabular}{cccccc}
        \includegraphics[width=0.32\columnwidth]{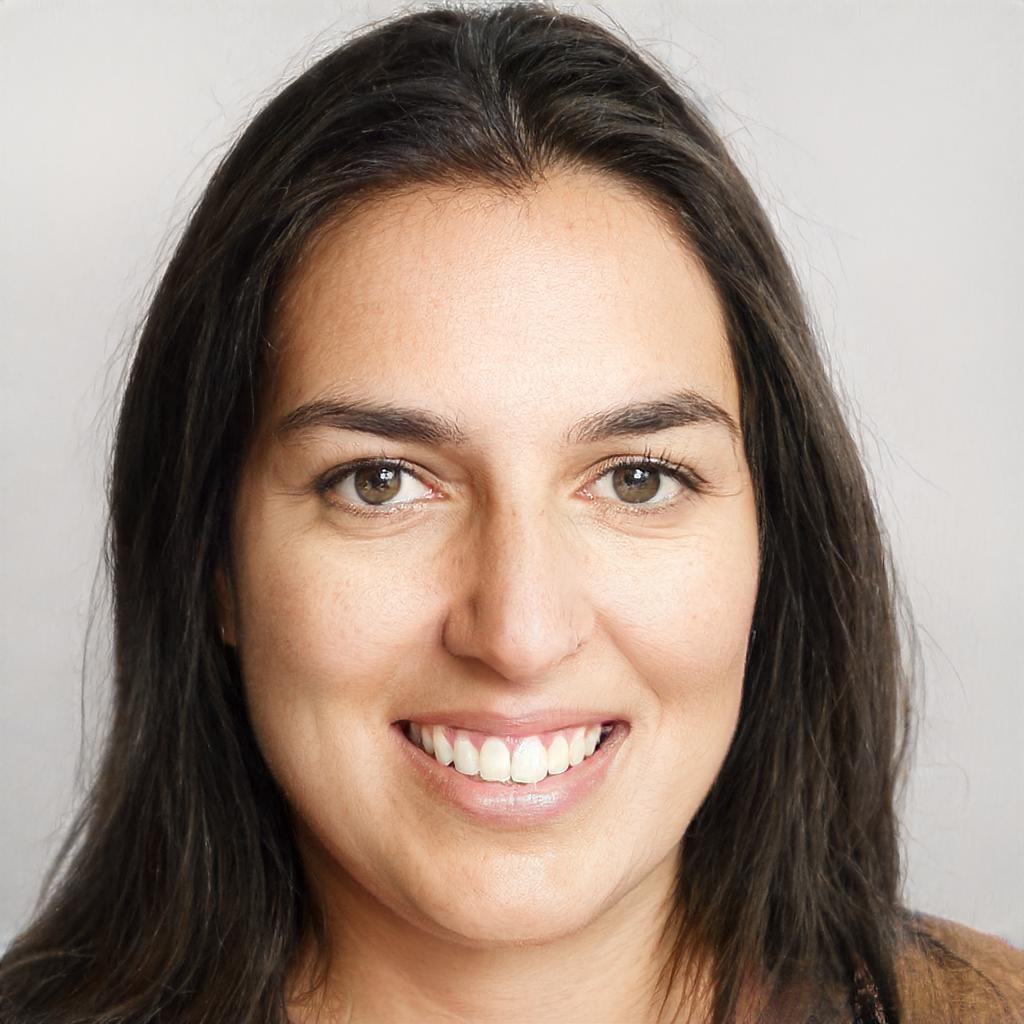} &
        \includegraphics[width=0.32\columnwidth]{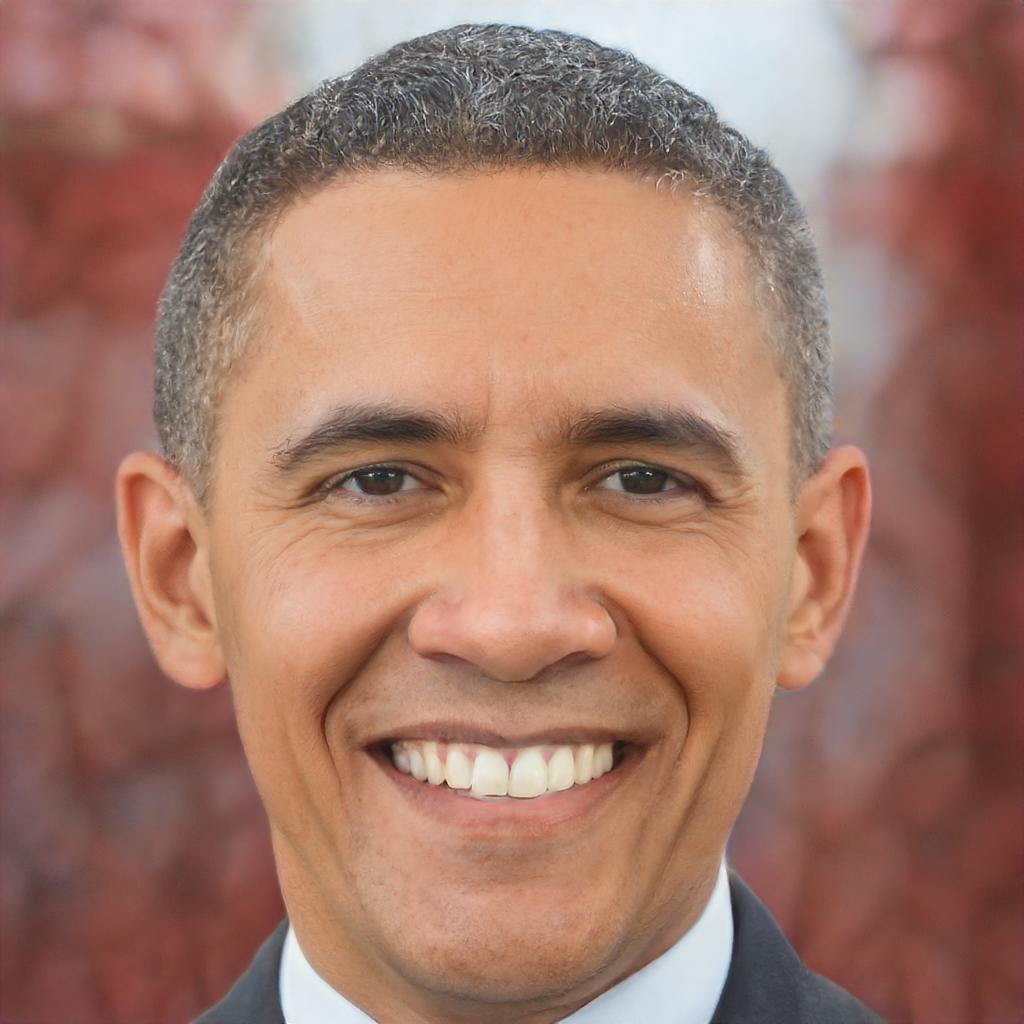}&
        \includegraphics[width=0.32\columnwidth]{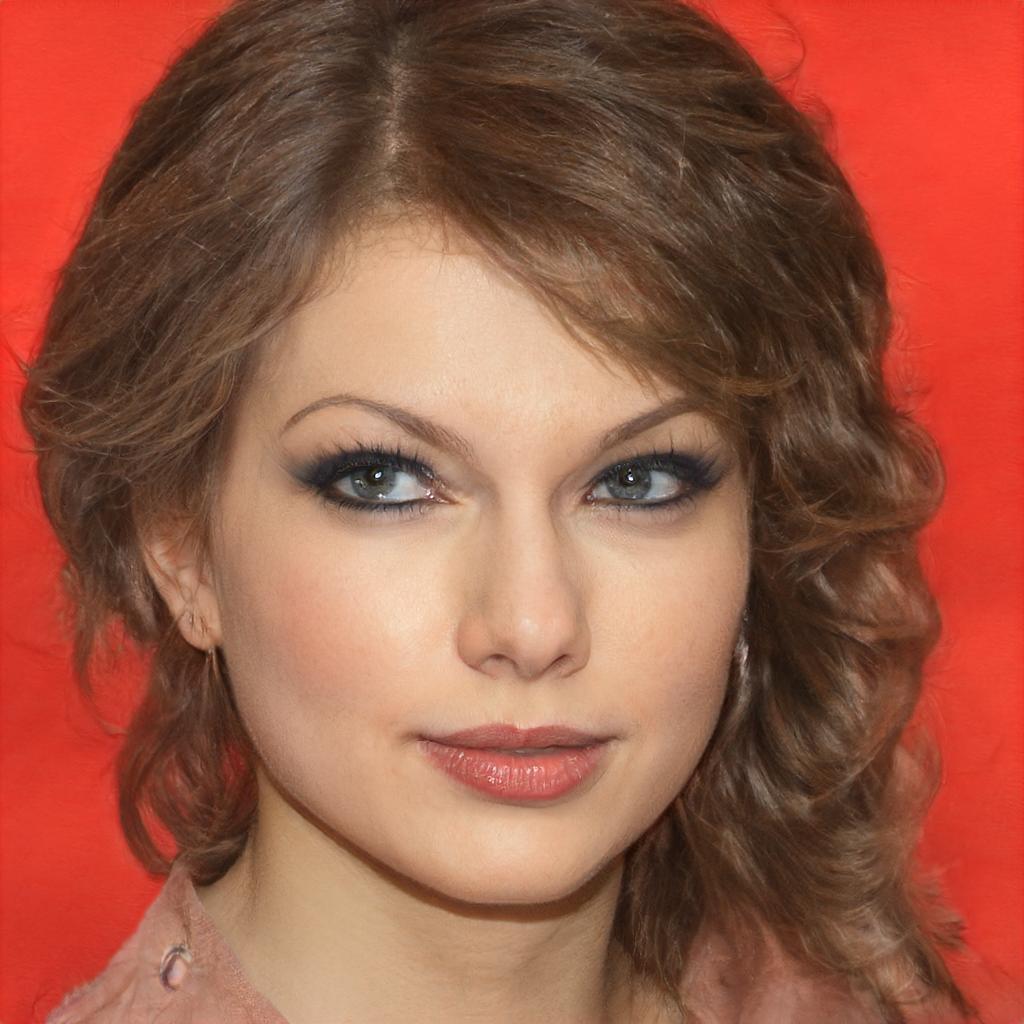} &
        \includegraphics[width=0.32\columnwidth]{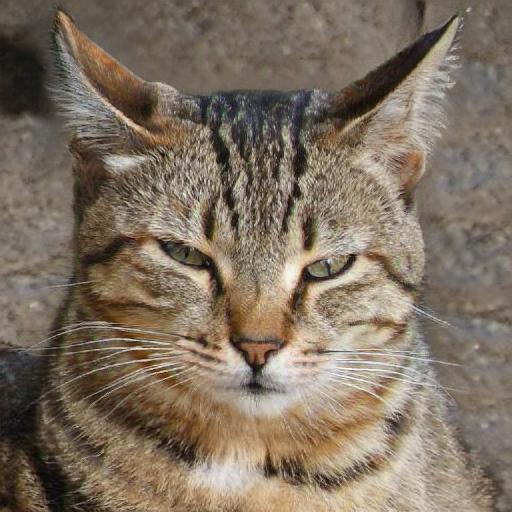} &
        \includegraphics[width=0.32\columnwidth]{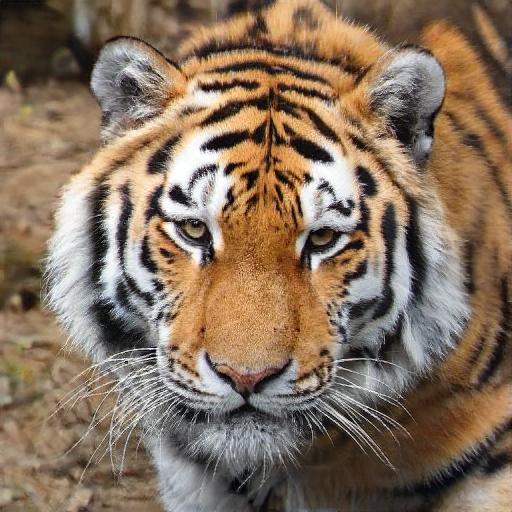} &
        \includegraphics[width=0.32\columnwidth]{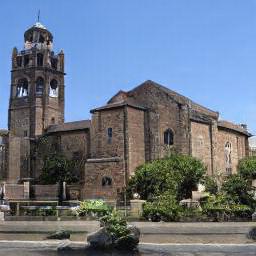}
        \\
        \includegraphics[width=0.32\columnwidth]{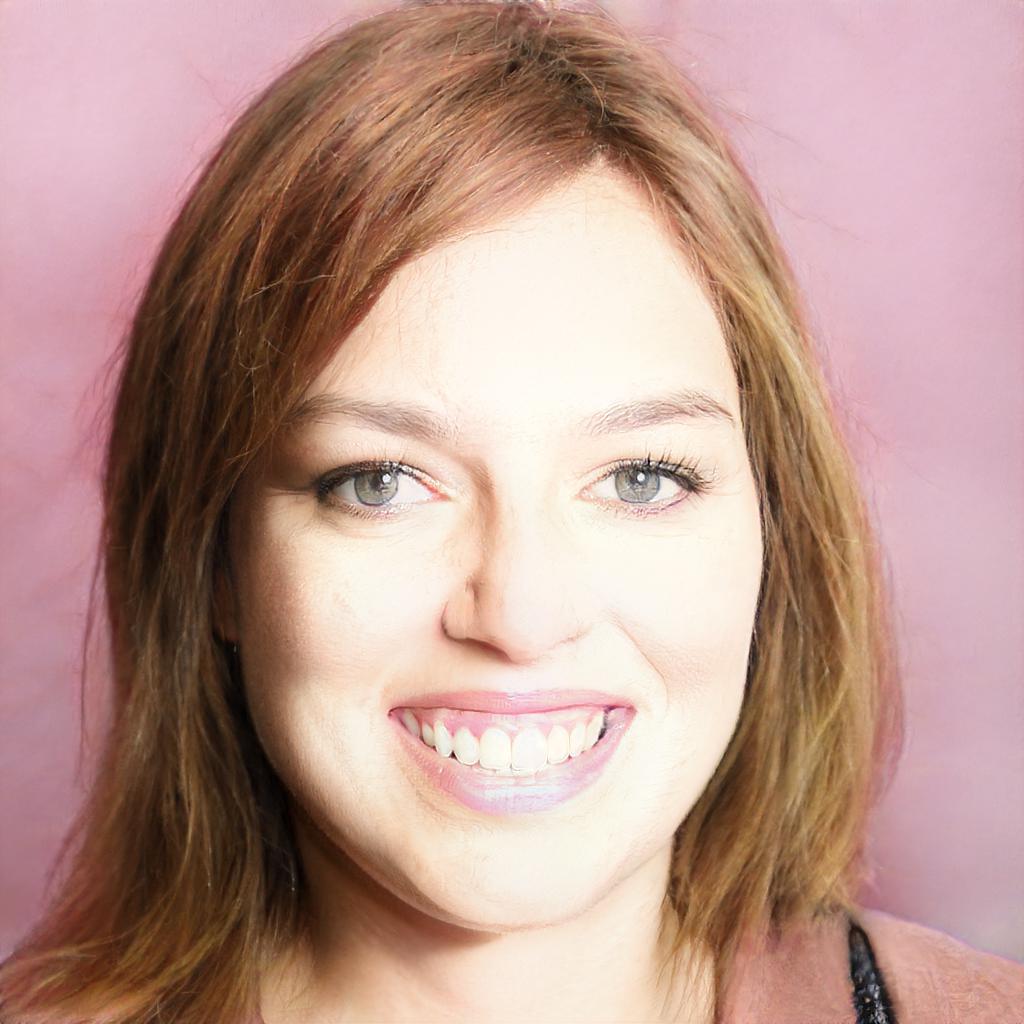} &
        \includegraphics[width=0.32\columnwidth]{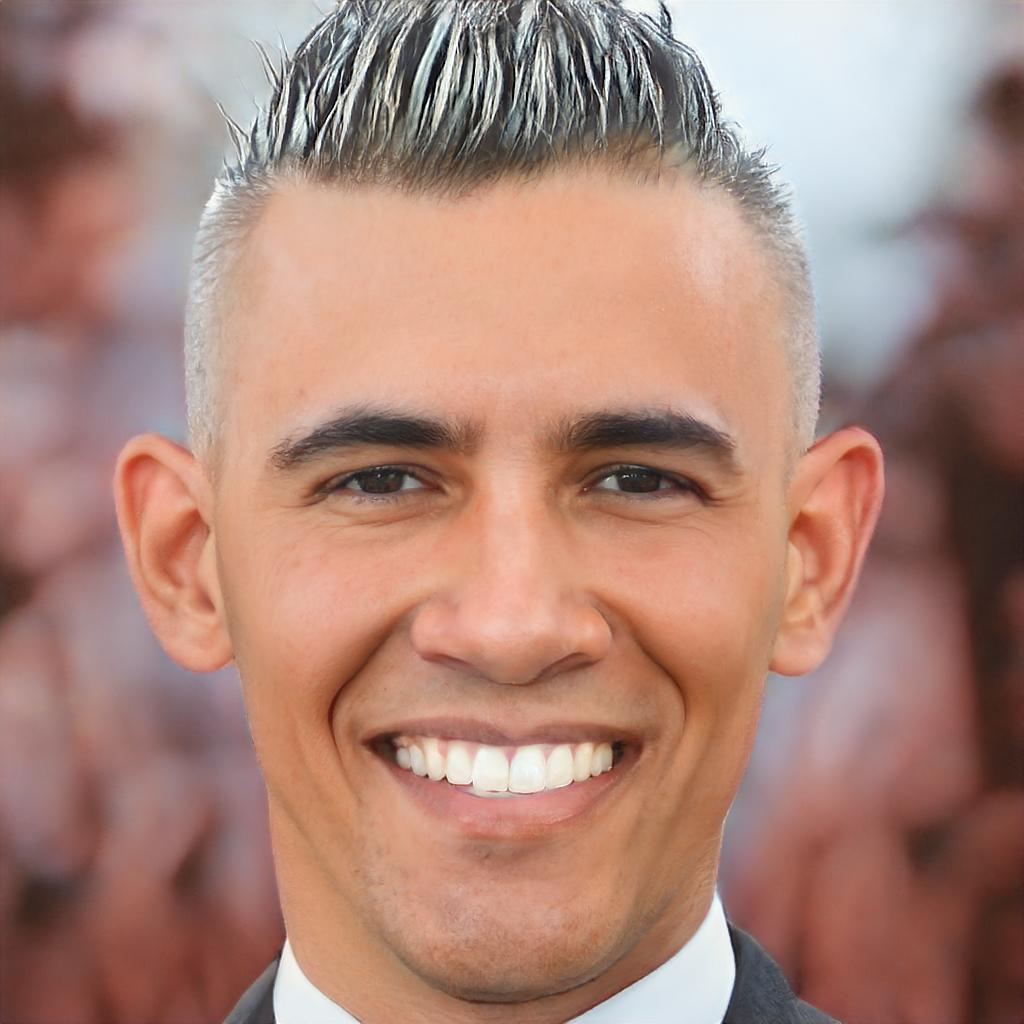}&
        \includegraphics[width=0.32\columnwidth]{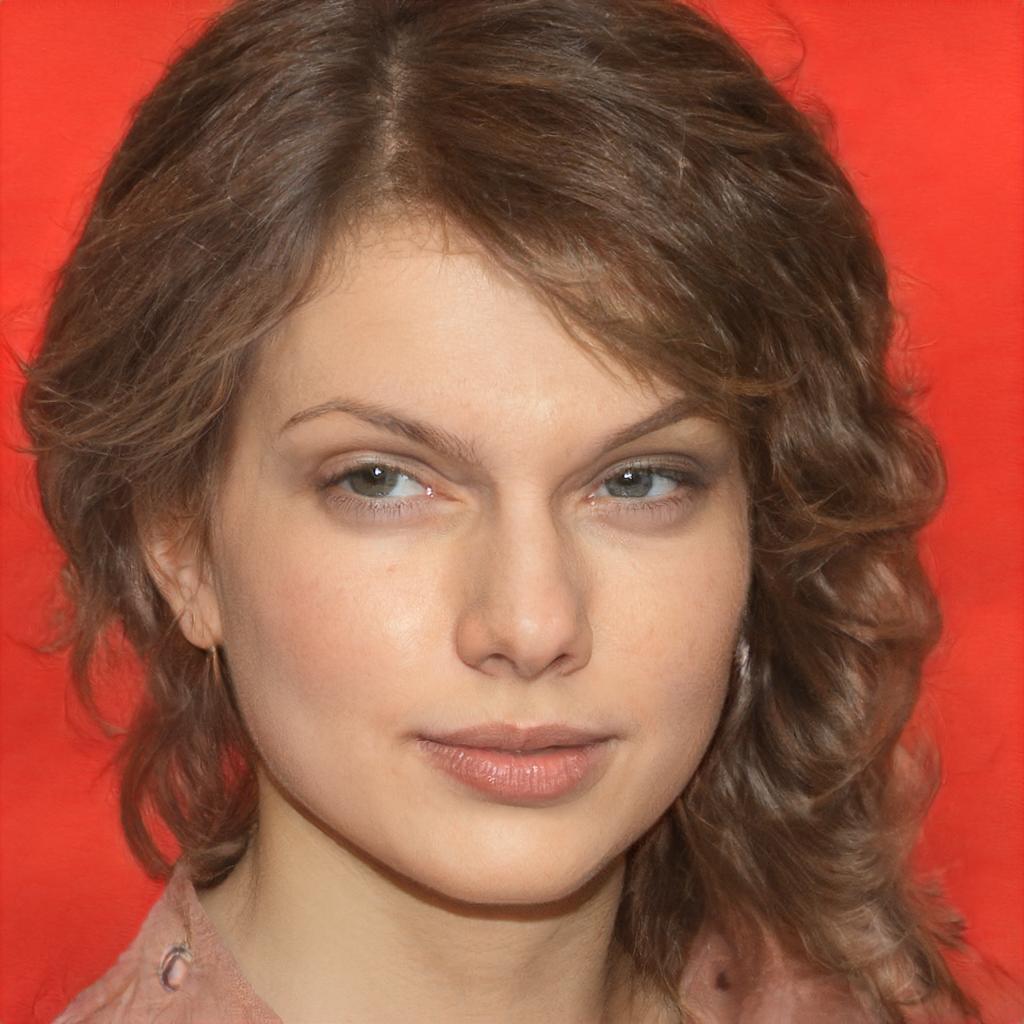} &
		\includegraphics[width=0.32\columnwidth]{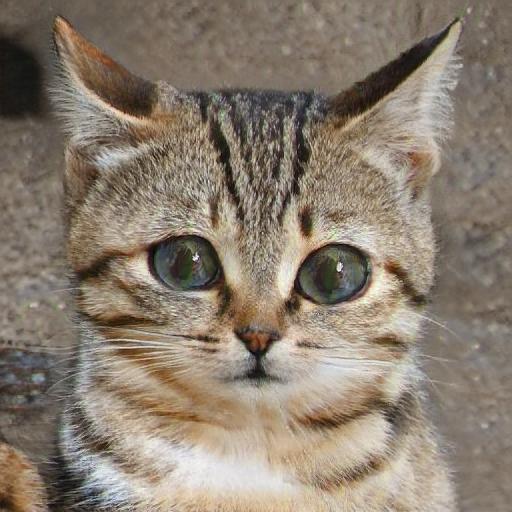} &
		\includegraphics[width=0.32\columnwidth]{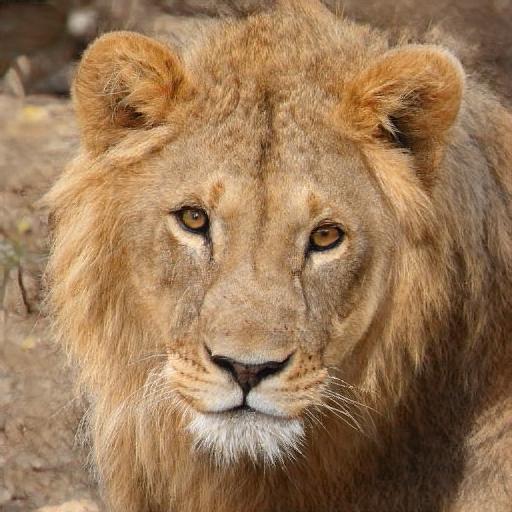} &
		\includegraphics[width=0.32\columnwidth]{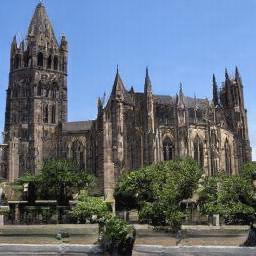}        
        \\
        {\footnotesize ``Emma Stone''} & {\footnotesize ``Mohawk hairstyle''} & {\footnotesize ``Without makeup''} & {\footnotesize ``Cute cat''} & {\footnotesize ``Lion''} & {\footnotesize ``Gothic church''}
        \end{tabular}
    \vspace{1mm}
    \captionof{figure}{Examples of text-driven manipulations using StyleCLIP. Top row: input images; Bottom row: our manipulated results. The text prompt used to drive each manipulation appears under each column.}
    \label{fig:teaser}
	\end{center}
    \vspace{2mm}
}]


\begin{abstract}
Inspired by the ability of StyleGAN to generate highly realistic images in a variety of domains, much recent work has focused on understanding how to use the latent spaces of StyleGAN to manipulate generated and real images.
However, discovering semantically meaningful latent manipulations typically involves painstaking human examination of the many degrees of freedom, or an annotated collection of images for each desired manipulation. 
In this work, we explore leveraging the power of recently introduced Contrastive Language-Image Pre-training (CLIP) models in order to develop a text-based interface for StyleGAN image manipulation that does not require such manual effort.
We first introduce an optimization scheme that utilizes a CLIP-based loss to modify an input latent vector in response to a user-provided text prompt.
Next, we describe a latent mapper that infers a text-guided latent manipulation step for a given input image, allowing faster and more stable text-based manipulation.
Finally, we present a method for mapping a text prompts to input-agnostic directions in StyleGAN's style space, enabling interactive text-driven image manipulation. Extensive results and comparisons demonstrate the effectiveness of our approaches.

\vfill
\end{abstract}

\vspace{-8mm}\noindent\rule{0.4\columnwidth}{0.4pt}\\
{\footnotesize $^*$ Equal contribution, ordered alphabetically. Code and video are available on \url{https://github.com/orpatashnik/StyleCLIP}}

\section{Introduction}
Generative Adversarial Networks (GANs)~\cite{goodfellow2014generative} have revolutionized image synthesis,
with recent style-based generative models~\cite{karras2019style,karras2020analyzing,karras2020ada} boasting some of the most realistic synthetic imagery to date.
Furthermore, the learnt intermediate latent spaces of StyleGAN have been shown to possess disentanglement properties~\cite{collins2020editing, shen2020interpreting, harkonen2020ganspace, tewari2020stylerig, wu2020stylespace}, which enable utilizing pretrained models to perform a wide variety of image manipulations on synthetic, as well as real, images.

Harnessing StyleGAN's expressive power requires developing simple and intuitive interfaces for users to easily carry out their intent.
Existing methods for semantic control discovery either involve manual examination (e.g.,~\cite{harkonen2020ganspace,shen2020interpreting,wu2020stylespace}), a large amount of annotated data, or pretrained classifiers \cite{shen2020interfacegan,abdal2020styleflow}.
Furthermore, subsequent manipulations are typically carried out by moving along a direction in one of the latent spaces, using a parametric model, such as a 3DMM in StyleRig \cite{tewari2020stylerig}, or a trained normalized flow in StyleFlow \cite{abdal2020styleflow}. Specific edits, such as virtual try-on~\cite{lewis2021vogue} and aging~\cite{alaluf2021matter} have also been explored.

Thus, existing controls enable image manipulations only along preset semantic directions, severely limiting the user's creativity and imagination. Whenever an additional, unmapped, direction is desired, further manual effort and/or large quantities of annotated data are necessary.

In this work, we explore leveraging the power of recently introduced Contrastive Language-Image Pre-training (CLIP) models in order to enable intuitive text-based semantic image manipulation that is neither limited to preset manipulation directions, nor requires additional manual effort to discover new controls.
The CLIP model is pretrained on 400 million image-text pairs harvested from the Web, and since natural language is able to express a much wider set of visual concepts, combining CLIP with the generative power of StyleGAN opens fascinating avenues for image manipulation.
Figures~\ref{fig:teaser} shows several examples of unique manipulations produced using our approach.
Specifically, in this paper we investigate three techniques that combine CLIP with StyleGAN:
\begin{enumerate}
	\item Text-guided latent optimization, where a CLIP model is used as a loss network \cite{johnson2016perceptual}. This is the most versatile approach, but it requires a few minutes of optimization to apply a manipulation to an image.
	\item A latent residual mapper, trained for a specific text prompt. Given a starting point in latent space (the input image to be manipulated), the mapper yields a local step in latent space.
	\item A method for mapping a text prompt into an input-agnostic (global) direction in StyleGAN's style space, providing control over the manipulation strength as well as the degree of disentanglement.
\end{enumerate}

The results in this paper and the supplementary material demonstrate a wide range of semantic manipulations on images of human faces, animals, cars, and churches. These manipulations range from abstract to specific, and from extensive to fine-grained. Many of them have not been demonstrated by any of the previous StyleGAN manipulation works, and all of them were easily obtained using a combination of pretrained StyleGAN and CLIP models.

%

\section{Related Work}
\label{sec:related}

\subsection{Vision and Language}

\paragraph{Joint representations}
Multiple works learn cross-modal Vision and language (VL) representations~\cite{Desai2020VirTexLV, sariyildiz2020learning, Tan2019LXMERTLC, Lu2019ViLBERTPT, Li2019VisualBERTAS, Su2020VL-BERT:, Li2020UnicoderVLAU, Chen2020UNITERUI, Li2020OscarOA} for a variety of tasks, such as language-based image retrieval, image captioning, and visual question answering. Following the success of BERT~\cite{Devlin2019BERTPO} in various language tasks, recent VL methods typically use Transformers~\cite{NIPS2017_3f5ee243} to learn the joint representations.
A recent model, based on Contrastive Language-Image Pre-training (CLIP)~\cite{radford2021learning}, learns a multi-modal embedding space, which may be used to estimate the semantic similarity between a given text and an image. CLIP was trained on 400 million text-image pairs, collected from a variety of publicly available sources on the Internet. The representations learned by CLIP have been shown to be extremely powerful, enabling state-of-the-art zero-shot image classification on a variety of datasets. We refer the reader to OpenAI's Distill article \cite{distill2021multimodal} for an extensive exposition and discussion of the visual concepts learned by CLIP.

\vspace{-3mm}
\paragraph{Text-guided image generation and manipulation} 
The pioneering work of Reed \etal~\cite{Reed2016GenerativeAT} approached text-guided image generation by training a conditional GAN~\cite{mirza2014conditional}, conditioned by text embeddings obtained from a pretrained encoder. 
Zhang \etal~\cite{zhang2017stackgan, Zhang2019StackGANRI} improved image quality by using multi-scale GANs. AttnGAN~\cite{Xu2018AttnGANFT} incorporated an attention mechanism between the text and image features.
Additional supervision was used in other works~\cite{Reed2016GenerativeAT, Li2019ObjectDrivenTS, Koh2020TexttoImageGG} to further improve the image quality.

A few studies focus on text-guided image manipulation.
Some methods~\cite{Dong2017SemanticIS, Nam2018TextAdaptiveGA, Liu2020DescribeWT} use a GAN-based encoder-decoder architecture, to disentangle the semantics of both input images and text descriptions.
ManiGAN~\cite{li2020manigan} introduces a novel text-image combination module, which produces high-quality images. 
Differently from the aforementioned works, we propose a single framework that combines the high-quality images generated by StyleGAN, with the rich multi-domain semantics learned by CLIP.

Recently, DALL·E~\cite{unpublished2021dalle,ramesh2021zeroshot}, a 12-billion parameter version of GPT-3~\cite{Brown2020LanguageMA}, which at 16-bit precision requires over 24GB of GPU memory, has shown a diverse set of capabilities in generating and applying transformations to images guided by text. In contrast, our approach is deployable even on a single commodity GPU.

A concurrent work to ours, TediGAN~\cite{xia2020tedigan}, also uses StyleGAN for text-guided image generation and manipulation. By training an encoder to map text into the StyleGAN latent space, one can generate an image corresponding to a given text. To perform text-guided image manipulation, TediGAN encodes both the image and the text into the latent space, and then performs style-mixing to generate a corresponding image.
In Section~\ref{sec:experiments} we demonstrate that the manipulations achieved using our approach reflect better the semantics of the driving text. 

In a recent online post, Perez~\cite{perez2021imagesfromprompts} describes a text-to-image approach that combines StyleGAN and CLIP in a manner similar to our latent optimizer in Section~\ref{sec:opt}.
Rather than synthesizing an image from scratch, our optimization scheme, as well as the other two approaches described in this work, focus on image manipulation. While text-to-image generation is an intriguing and challenging problem, we believe that the image manipulation abilities we provide constitute a more useful tool for the typical workflow of creative artists.    

\subsection{Latent Space Image Manipulation}
Many works explore how to utilize the latent space of a pretrained generator for image manipulation~\cite{collins2020editing, tewari2020stylerig, wu2020stylespace}. Specifically, the intermediate latent spaces in StyleGAN have been shown to enable many disentangled and meaningful image manipulations.
Some methods learn to perform image manipulation in an end-to-end fashion, by training a network that encodes a given image into a latent representation of the manipulated image~\cite{nitzan2020dis, richardson2020encoding, alaluf2021matter}. 
Other methods aim to find latent paths, such that traversing along them result in the desired manipulation.
Such methods can be categorized into: (i) methods that use image annotations to find meaningful latent paths~\cite{shen2020interpreting, abdal2020styleflow}, and (ii) methods that find meaningful directions without supervision, and require manual annotation for each direction~\cite{harkonen2020ganspace, shen2020closedform, voynov2020unsupervised, wang2021a}. 

While most works perform image manipulations in the $\mathcal{W}$ or $\mathcal{W+}$ spaces, Wu \etal~\cite{wu2020stylespace} proposed to use the \emph{StyleSpace} $\mathcal{S}$, and showed that it is better disentangled than $\mathcal{W}$ and $\mathcal{W+}$.
Our latent optimizer and mapper work in the $\mathcal{W+}$ space, while the input-agnostic directions that we detect are in $\mathcal{S}$.
In all three, the manipulations are derived directly from text input, and
our only source of supervision is a pretrained CLIP model.
As CLIP was trained on hundreds of millions of text-image pairs, our approach is generic and can be used in a multitude of domains without the need for domain- or manipulation-specific data annotation.

\newcommand{\clipI}{\mathcal{I}}
\newcommand{\clipT}{\mathcal{T}}
\newcommand{\Sspace}{\mathcal{S}}
\newcommand{\Wspace}{\mathcal{W}}
\newcommand{\Wplus}{\mathcal{W}+}
\newcommand{\Gs}{G(s)}
\newcommand{\Ds}{{\Delta s}}
\newcommand{\Di}{{\Delta i}}
\newcommand{\Dt}{{\Delta t}}

\section{StyleCLIP Text-Driven Manipulation}

In this work we explore three ways for text-driven image manipulation, all of which combine the generative power of StyleGAN with the rich joint vision-language representation learned by CLIP.

We begin in Section \ref{sec:opt} with a simple latent optimization scheme, where a given latent code of an image in StyleGAN's $\mathcal{W}+$ space is optimized by minimizing a loss computed in CLIP space.
The optimization is performed for each (source image, text prompt) pair.
Thus, despite it's versatility, several minutes are required to perform a single manipulation, and the method can be difficult to control.
A more stable approach is described in Section \ref{sec:mapper}, where a mapping network is trained to infer a manipulation step in latent space, in a single forward pass. The training takes a few hours, but it must only be done once per text prompt. 
The direction of the manipulation step may vary depending on the starting position in $\mathcal{W}+$, which corresponds to the input image, and thus we refer to this mapper as \emph{local}. 

Our experiments with the local mapper reveal that, for a wide variety of manipulations, the directions of the manipulation step are often similar to each other, despite different starting points. Also, since the manipulation step is performed in $\mathcal{W}+$, it is difficult to achieve fine-grained visual effects in a disentangled manner. Thus, in Section~\ref{sec:global} we explore a third text-driven manipulation scheme, which transforms a given text prompt into an input agnostic (i.e., \emph{global} in latent space) mapping direction.
The global direction is computed in StyleGAN's style space $\Sspace$ \cite{wu2020stylespace}, which is better suited for fine-grained and disentangled visual manipulation, compared to $\Wplus$.

\begin{table}
	\centering
	\resizebox{\columnwidth}{!}{%
	\begin{tabular}{|c|c|c|c|c|c|}
		\hline
		& \begin{tabular}[c]{@{}c@{}}pre-\\proc.\end{tabular} & 
		  \begin{tabular}[c]{@{}c@{}}train\\time\end{tabular} &
		  \begin{tabular}[c]{@{}c@{}}infer.\\time\end{tabular} &
		  \begin{tabular}[c]{@{}c@{}}input image\\dependent\end{tabular} &
		  \begin{tabular}[c]{@{}c@{}}latent\\space\end{tabular} \\ \hline
		optimizer & -- & -- & 98 sec & yes & $\Wplus$ \\ \hline
		mapper & -- & 10 -- 12h & 75 ms & yes & $\Wplus$ \\ \hline
		global dir. & 4h & -- & 72 ms & no & $\Sspace$ \\ \hline
	\end{tabular}}
		\vspace{1mm}
		\caption{\label{tab:methods}
		Our three methods for combining StyleGAN and CLIP. The latent step inferred by the optimizer and the mapper depends on the input image, but the training is only done once per text prompt. The global direction method requires a one-time pre-processing, after which it may be applied to different (image, text prompt) pairs. Times are for a single NVIDIA GTX 1080Ti GPU.
		}
\end{table}

Table \ref{tab:methods} summarizes the differences between the three methods outlined above, while visual results and comparisons are presented in the following sections.

\section{Latent Optimization}
\label{sec:opt}
\newcommand{\Dclip}{D_{\text{CLIP}}}


A simple approach for leveraging CLIP to guide image manipulation is through direct latent code optimization. Specifically, given a source latent code $w_s \in \mathcal{W}+$, and a directive in natural language, or a \emph{text prompt} $t$, we solve the following optimization problem:
\begin{equation}
	\label{eq:opt}
    \argmin_{w \in \mathcal{W}+} {\Dclip(G(w), t) + \lambda_{\text{L2}} \norm{w - w_s}_2 + \lambda_{\text{ID}} \mathcal{L}_\text{ID}(w)},
\end{equation}
where $G$ is a pretrained StyleGAN\footnote{We use StyleGAN2~\cite{karras2020analyzing} in all our experiments.} generator and $\Dclip$ is the cosine distance between the CLIP embeddings of its two arguments. 
Similarity to the input image is controlled by the $L_2$ distance in latent space,
and by the identity loss~\cite{richardson2020encoding}: 
\begin{equation}
\label{eq:id-loss}
    \mathcal{L}_{\text{ID}}\left (w \right ) = 1-\left \langle R(G(w_s)),R(G(w)) \right \rangle ,
\end{equation} 
where $R$ is a pretrained ArcFace~\cite{deng2019arcface} network for face recognition, and $\langle \cdot, \cdot \rangle$ computes the cosine similarity between it's arguments.
We solve this optimization problem through gradient descent, by back-propagating the gradient of the objective in \eqref{eq:opt} through the pretrained and fixed StyleGAN generator $G$ and the CLIP image encoder.

In Figure \ref{fig:opt_results} we provide several edits that were obtained using this optimization approach after 200-300 iterations. The input images were inverted by e4e \cite{tov2021designing}. 
Note that visual characteristics may be controlled explicitly (beard, blonde) or implicitly, by indicating a real or a fictional person (Beyonce, Trump, Elsa). 
The values of $\lambda_{\text{L2}}$ and $\lambda_{\text{ID}}$ depend on the nature of the desired edit. For changes that shift towards another identity, $\lambda_{\text{ID}}$ is set to a lower value.

\begin{figure*}[t]
    \centering
    \includegraphics[width=\textwidth]{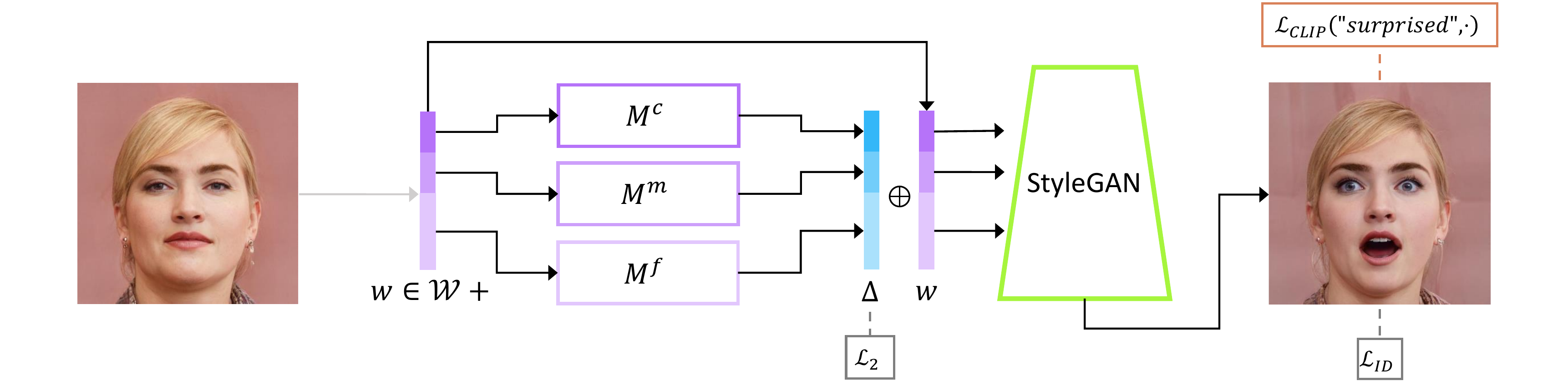}
    \caption{The architecture of our text-guided mapper (using the text prompt ``surprised'', in this example). The source image (left) is inverted into a latent code $w$. Three separate mapping functions are trained to generate residuals (in blue) that are added to $w$ to yield the target code, from which a pretrained StyleGAN (in green) generates an image (right), assessed by the CLIP and identity losses.
	}
    \label{fig:mapper_arch}
\end{figure*}

\begin{figure}[tb]
	\setlength{\tabcolsep}{1pt}
	\centering
	{\footnotesize
		\begin{tabular}{C{0.24\linewidth} C{0.24\linewidth} C{0.24\linewidth} C{0.24\linewidth}}
			\includegraphics[width=\linewidth]{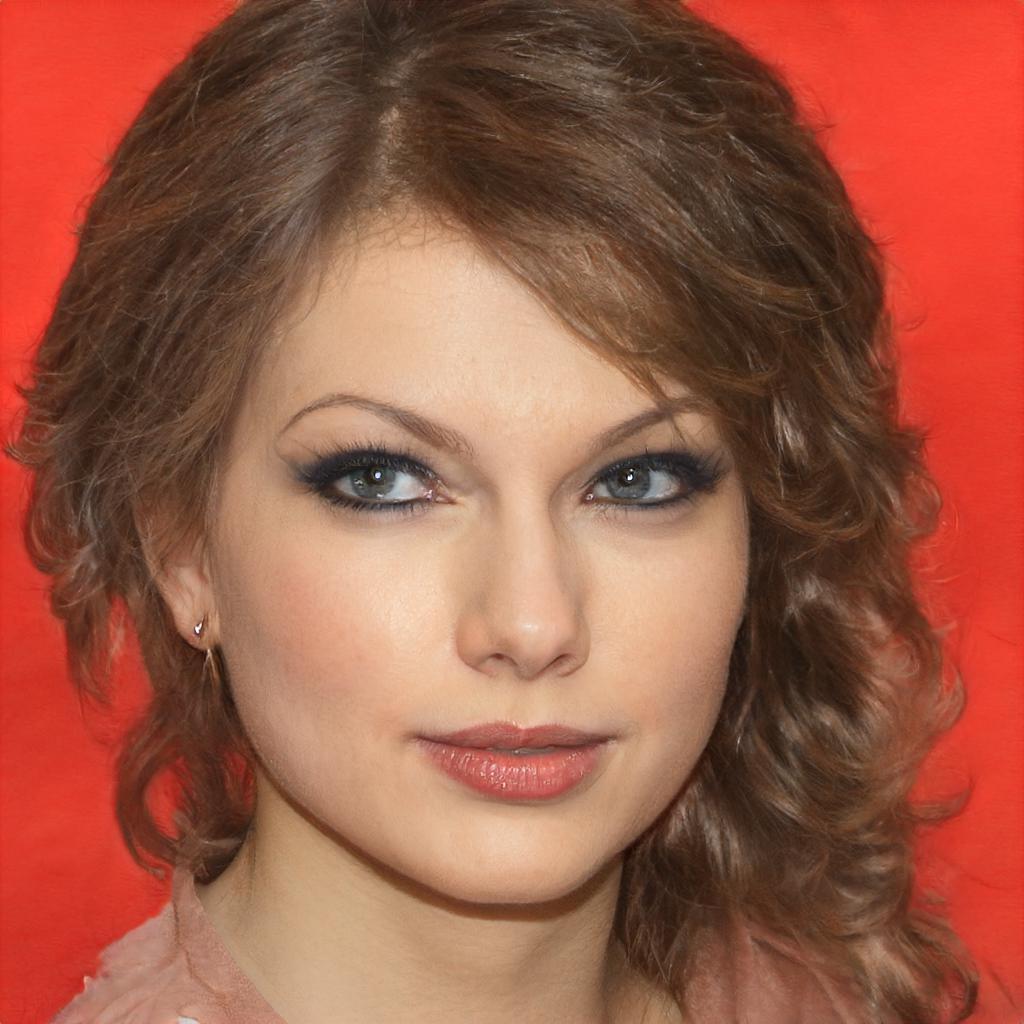} &
			\includegraphics[width=\linewidth]{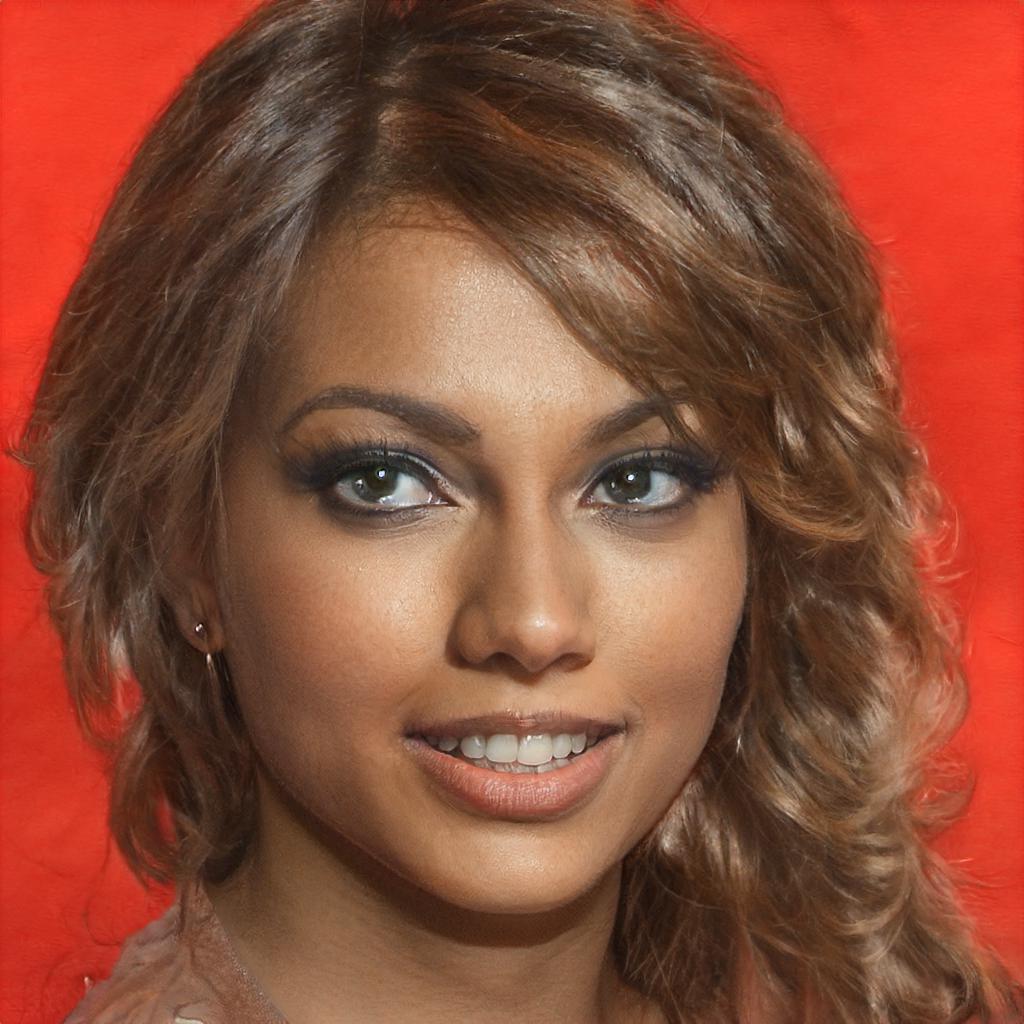} &
			\includegraphics[width=\linewidth]{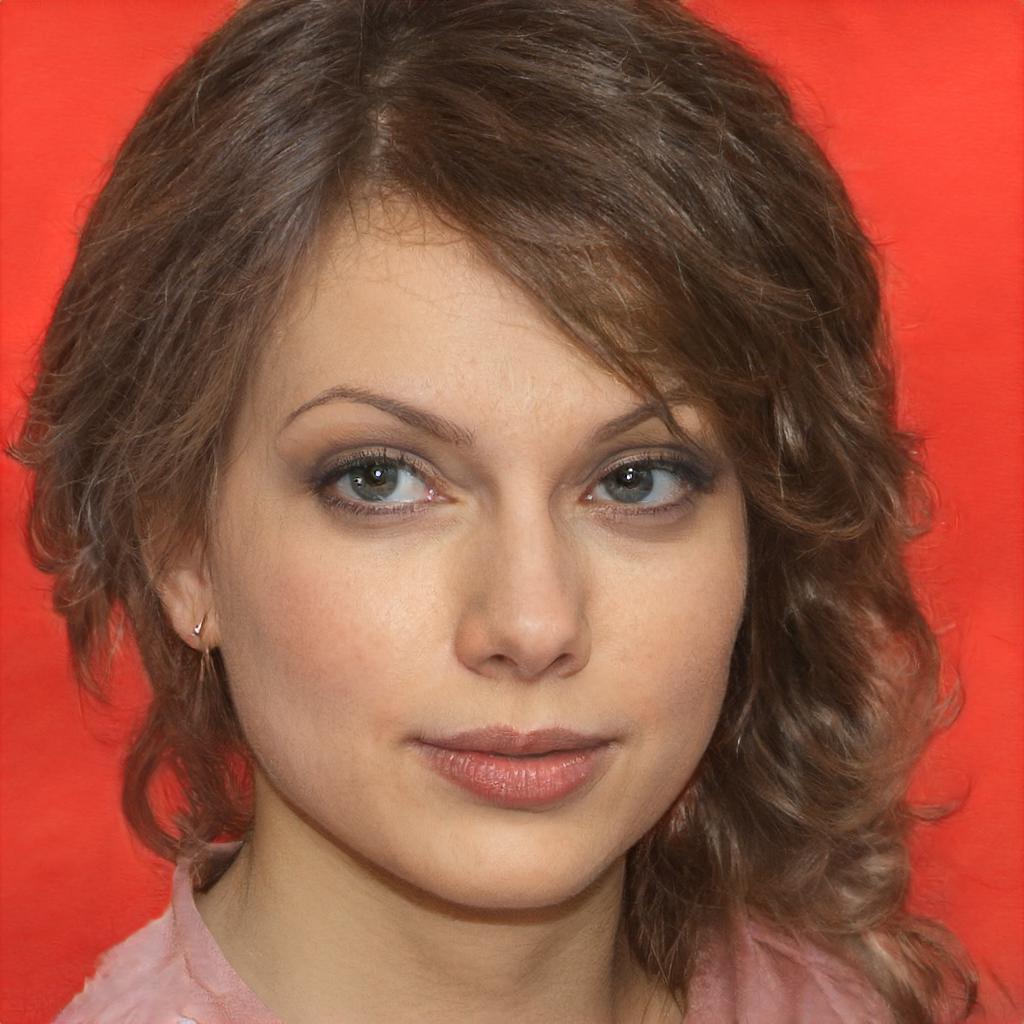} &
			\includegraphics[width=\linewidth]{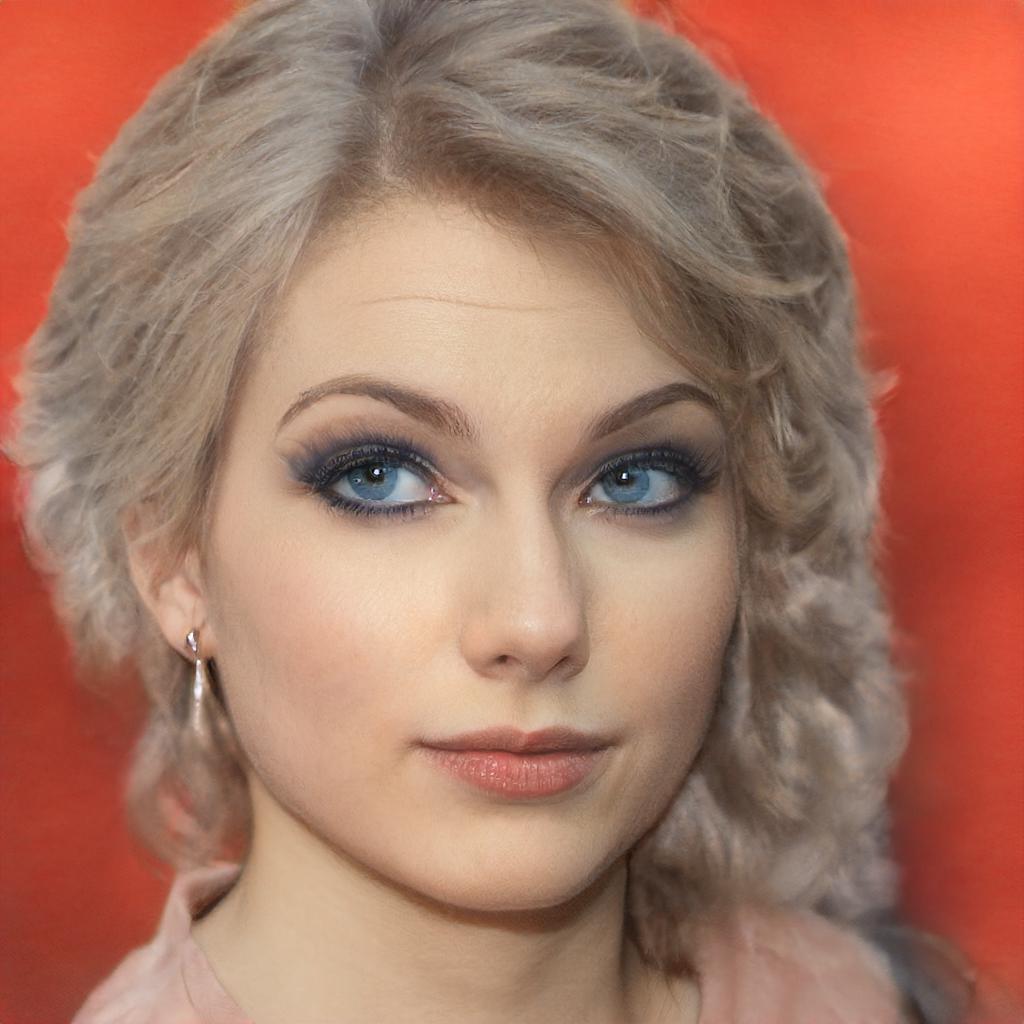}
			\\
			Input & ``Beyonce'' (0.004, 0) & ``A woman without makeup'' (0.008, 0.005) & ``Elsa from \phantom{m} Frozen'' \phantom{m} (0.004, 0) \\
			\includegraphics[width=\linewidth]{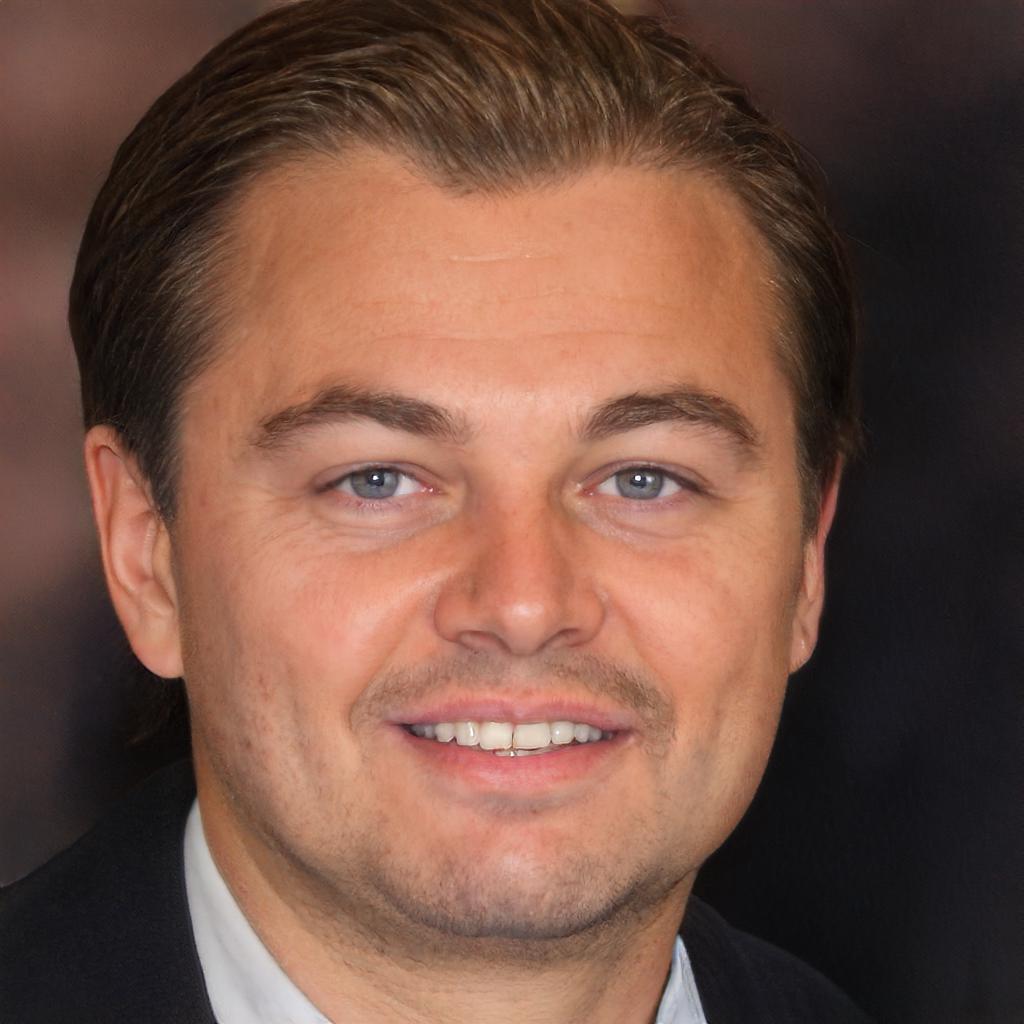} &
			\includegraphics[width=\linewidth]{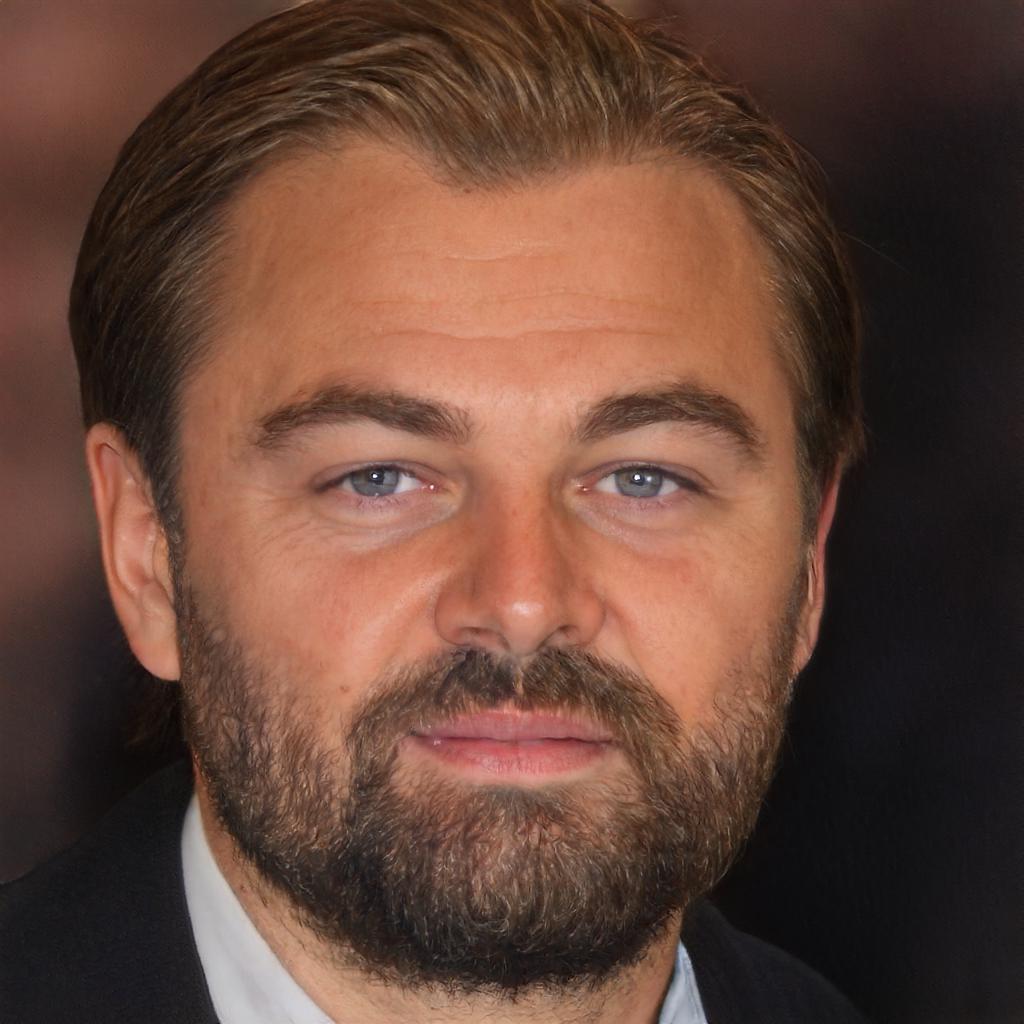} &
			\includegraphics[width=\linewidth]{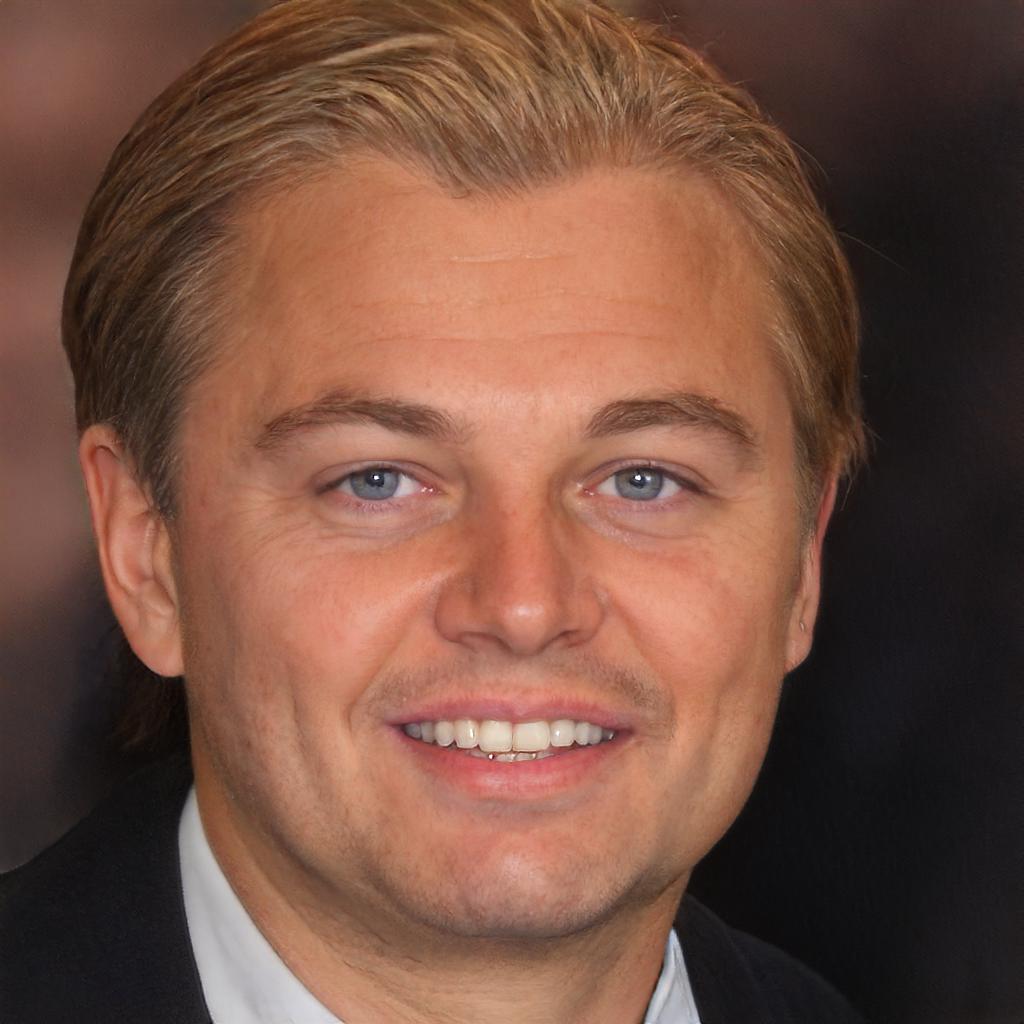} &
			\includegraphics[width=\linewidth]{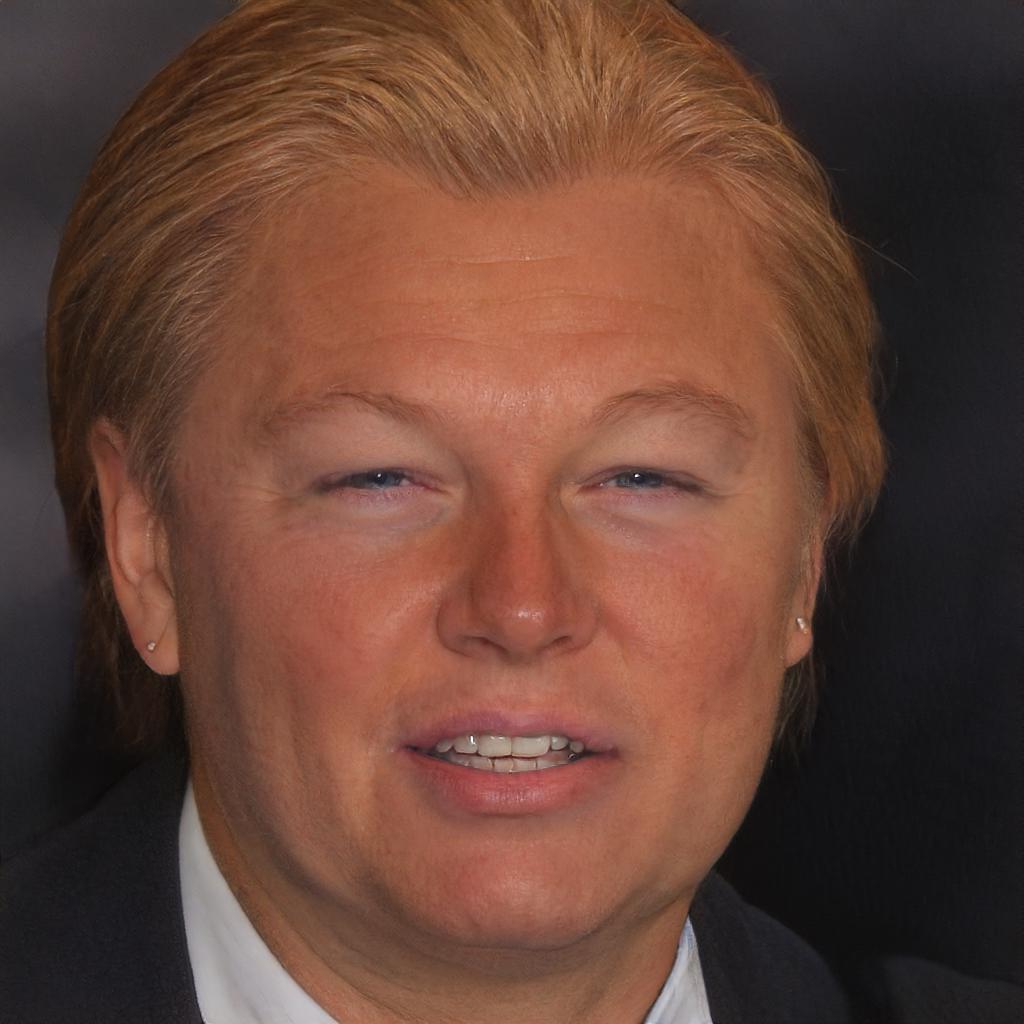}
			\\
			Input & ``A man with a \phantom{mm} beard'' \phantom{mm} (0.008, 0.005) & ``A blonde man'' (0.008, 0.005) & ``Donald Trump'' (0.0025, 0) \\
		\end{tabular}
	}
	\caption{Edits of real celebrity portraits obtained by latent optimization. The driving text prompt and the $(\lambda_{\text{L2}},\lambda_{\text{ID}})$ parameters for each edit are indicated under the corresponding result.}
	\label{fig:opt_results}
\end{figure}

\section{Latent Mapper}
\label{sec:mapper}

\begin{table*}[]
	\centering
	\begin{tabular}{|c|c|c|c|c|c|c|c|c|}
		\hline
		& \textbf{Mohawk} & \textbf{Afro} & \textbf{Bob-cut} & \textbf{Curly} & \textbf{Beyonce} & \textbf{Taylor Swift} & \textbf{Surprised} & \textbf{Purple hair}\\
		\hline
		\textbf{Mean} & 0.82 & 0.84 & 0.82 & 0.84 & 0.83 & 0.77 & 0.79 & 0.73\\
		\hline 
		\textbf{Std} & 0.096 & 0.085 & 0.095 & 0.088 & 0.081 & 0.107 & 0.893 & 0.145\\
		\hline
	\end{tabular}
	\vspace{1mm}
	\caption{Average cosine similarity between manipulation directions obtained from mappers trained using differnt text prompts.
	}
	\label{tab:mapper-directions}
\end{table*}

The latent optimization described above is versatile, as it performs a dedicated optimization for each (source image, text prompt) pair. On the downside, several minutes of optimization are required to edit a single image, and the method is somewhat sensitive to the values of its parameters.
Below, we describe a more efficient process, where a mapping network is trained, for a specific text prompt $t$, to infer a manipulation step $M_t(w)$ in the $\mathcal{W+}$ space, for any given latent image embedding $w \in \mathcal{W}+$.

\paragraph{Architecture}
The architecture of our text-guided mapper is depicted in Figure~\ref{fig:mapper_arch}.
It has been shown that different StyleGAN layers are responsible for different levels of detail in the generated image~\cite{karras2019style}. Consequently, it is common to split the layers into three groups (coarse, medium, and fine), and feed each group with a different part of the (extended) latent vector. We design our mapper accordingly, with three fully-connected networks, one for each group/part. The architecture of each of these networks is the same as that of the StyleGAN mapping network, but with fewer layers (4 rather than 8, in our implementation). Denoting the latent code of the input image as $w = (w_{c}, w_{m}, w_{f})$, the mapper is defined by 
\begin{equation}
	\label{eq:mapper3}
    M_t(w) = (M^c_t(w_c), M^m_t(w_m), M^f_t(w_f)).
\end{equation}
Note that one can choose to train only a subset of the three mappers. There are cases where it is useful to preserve some attribute level and keep the style codes in the corresponding entries fixed.

\begin{figure}[tb]
	\setlength{\tabcolsep}{1pt}
	\centering
	{\footnotesize
		\begin{tabular}{C{0.19\linewidth} C{0.19\linewidth} C{0.19\linewidth} C{0.19\linewidth} C{0.19\linewidth}}
			\includegraphics[width=\linewidth]{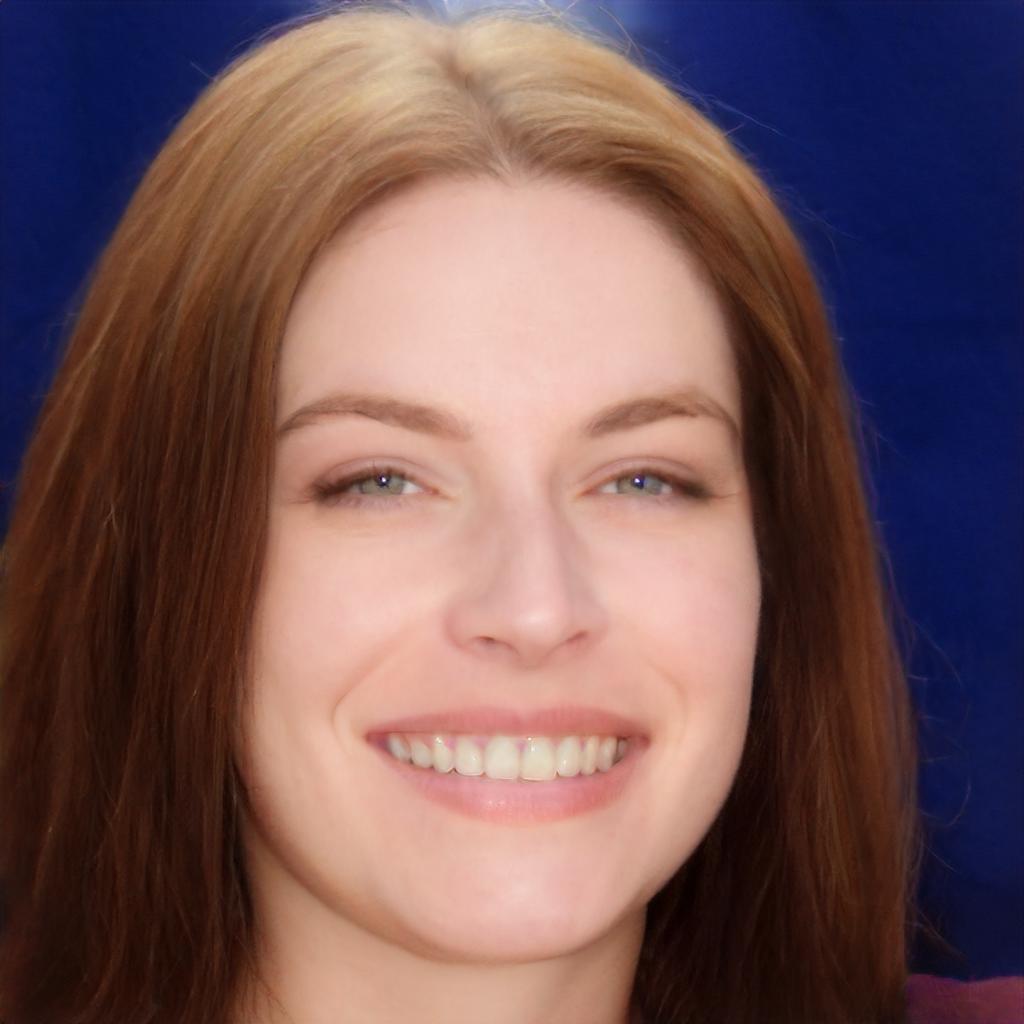} &
            \includegraphics[width=\linewidth]{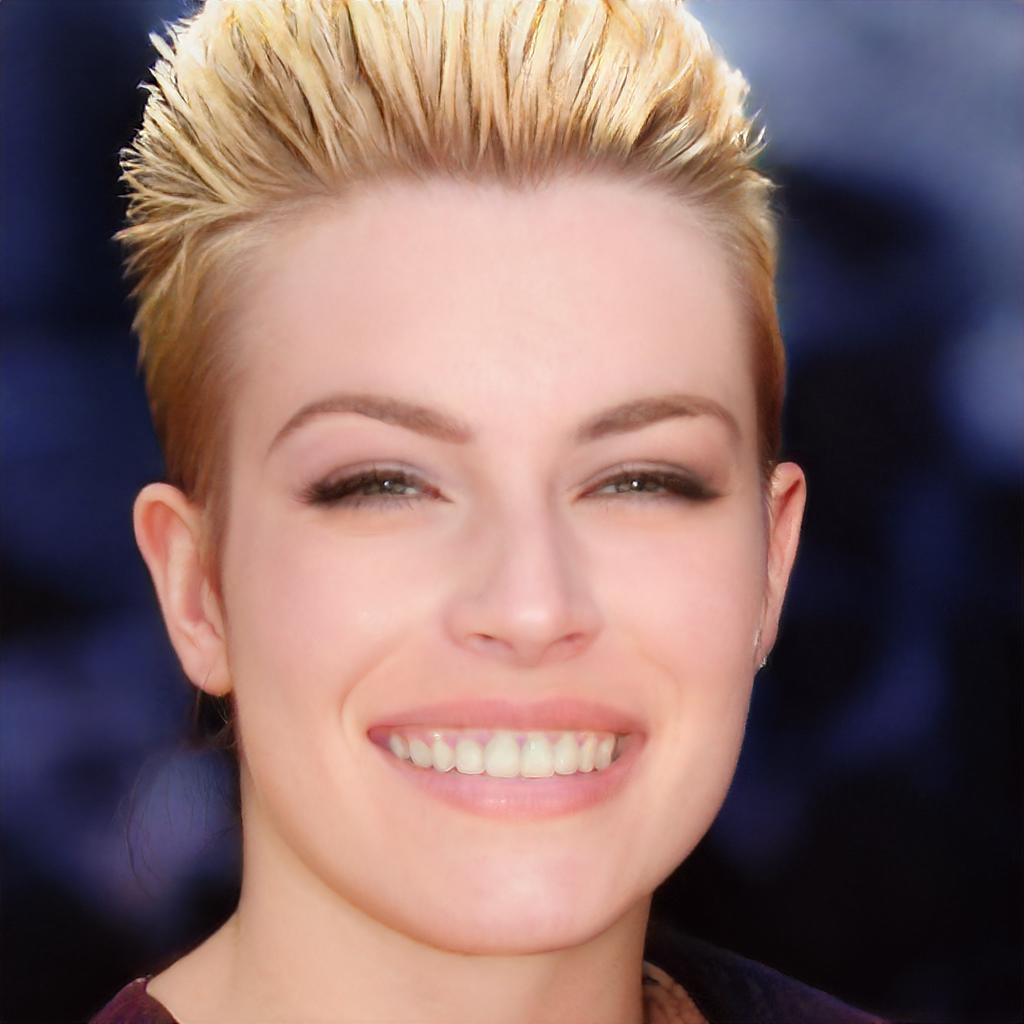} &
            \includegraphics[width=\linewidth]{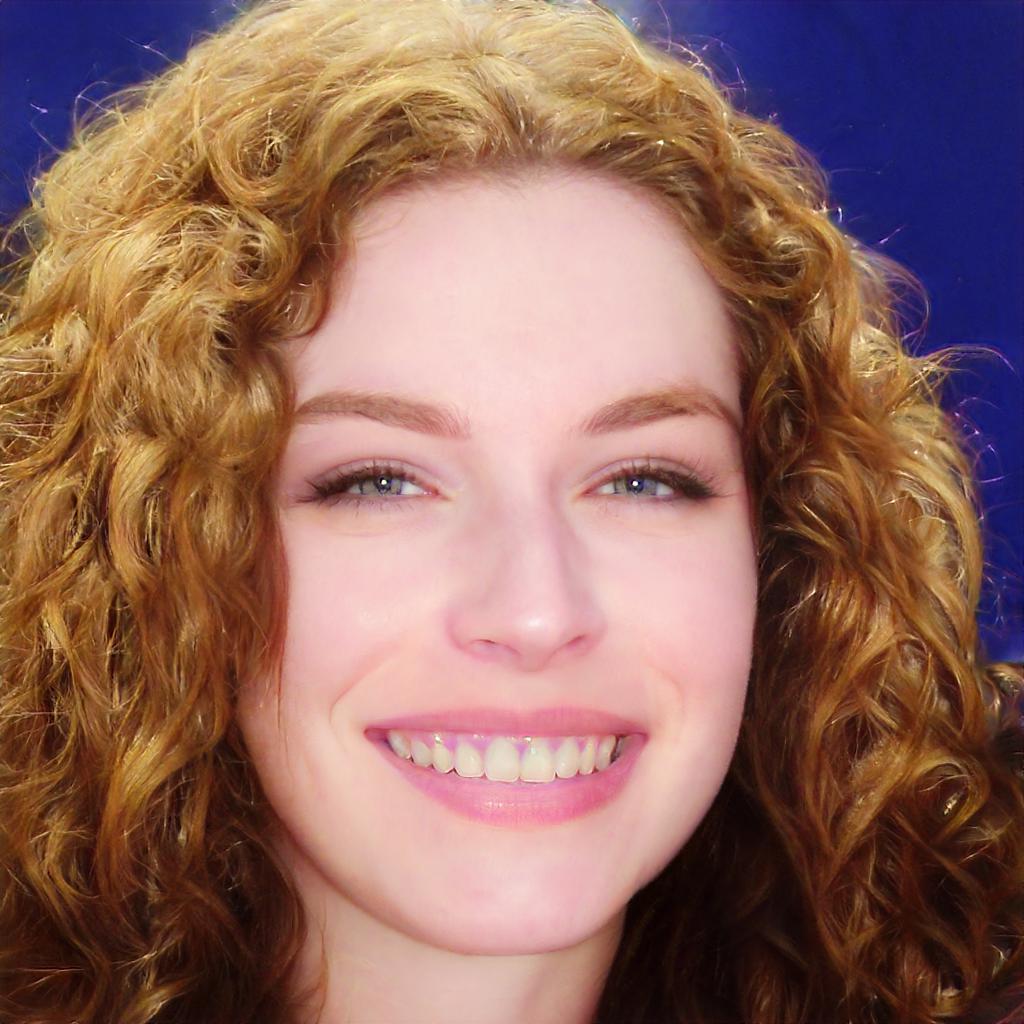} &
            \includegraphics[width=\linewidth]{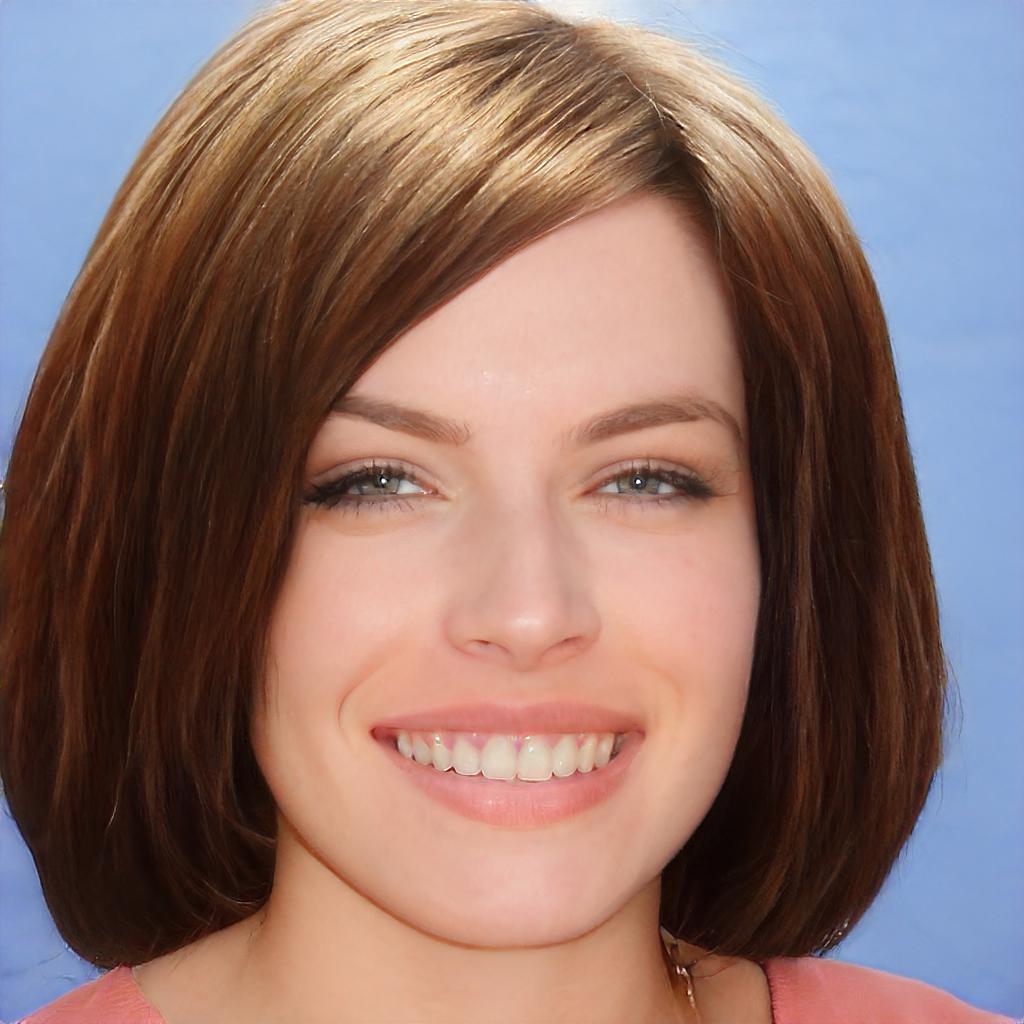} &
            \includegraphics[width=\linewidth]{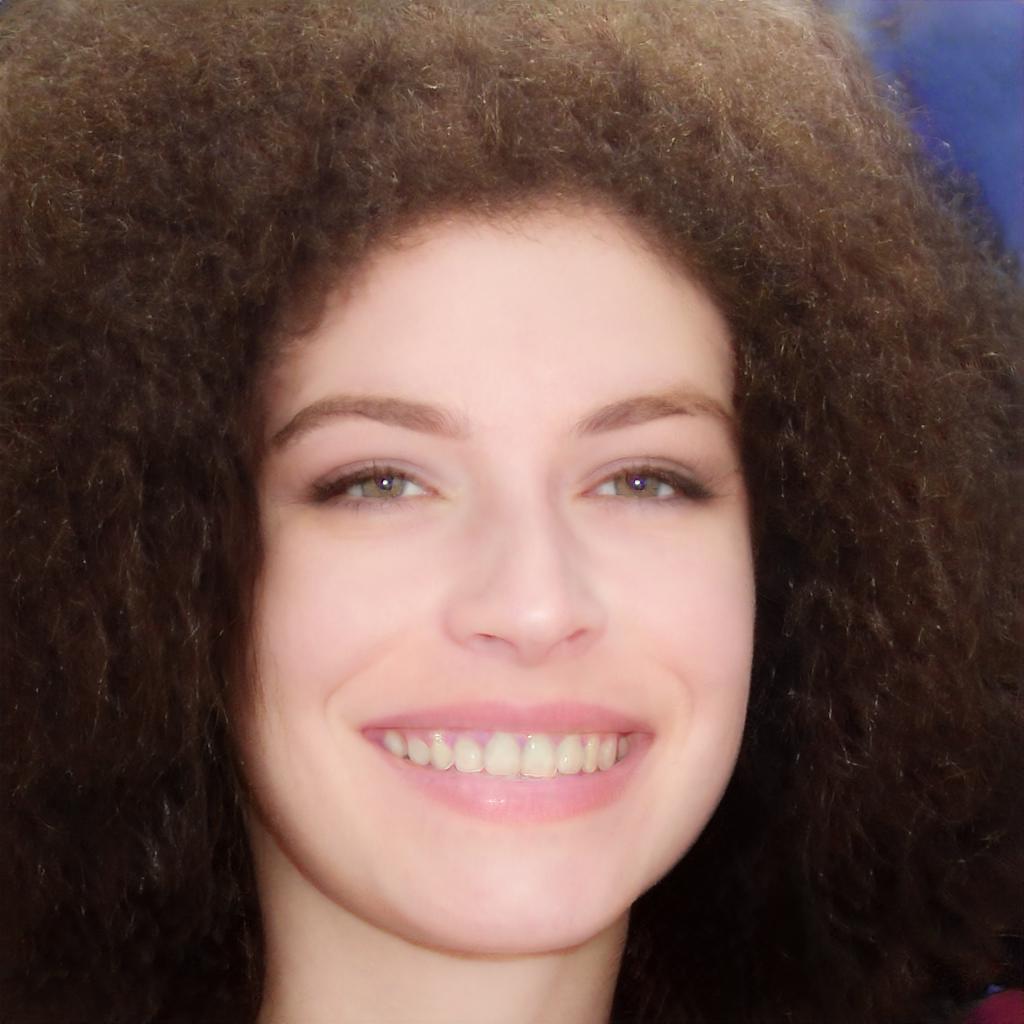} 
			\\
			\includegraphics[width=\linewidth]{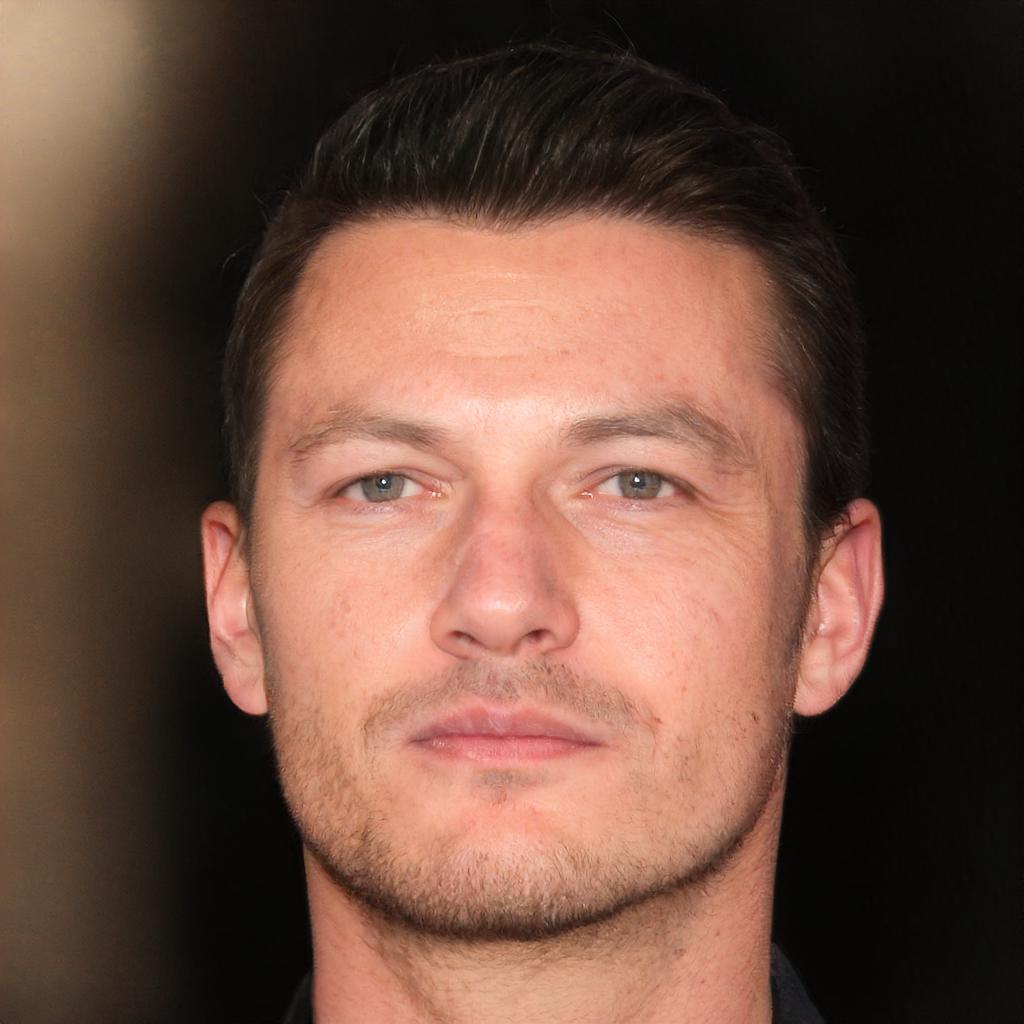} &
            \includegraphics[width=\linewidth]{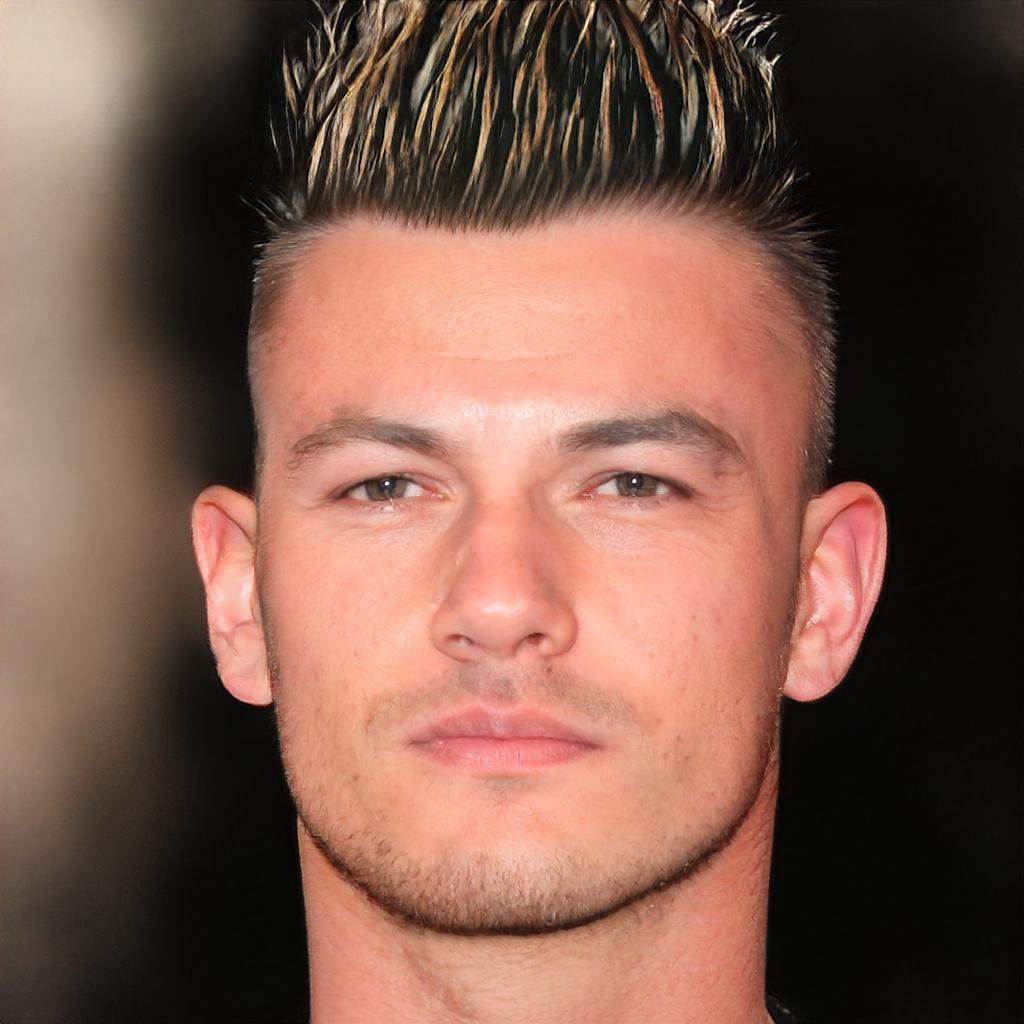} &
            \includegraphics[width=\linewidth]{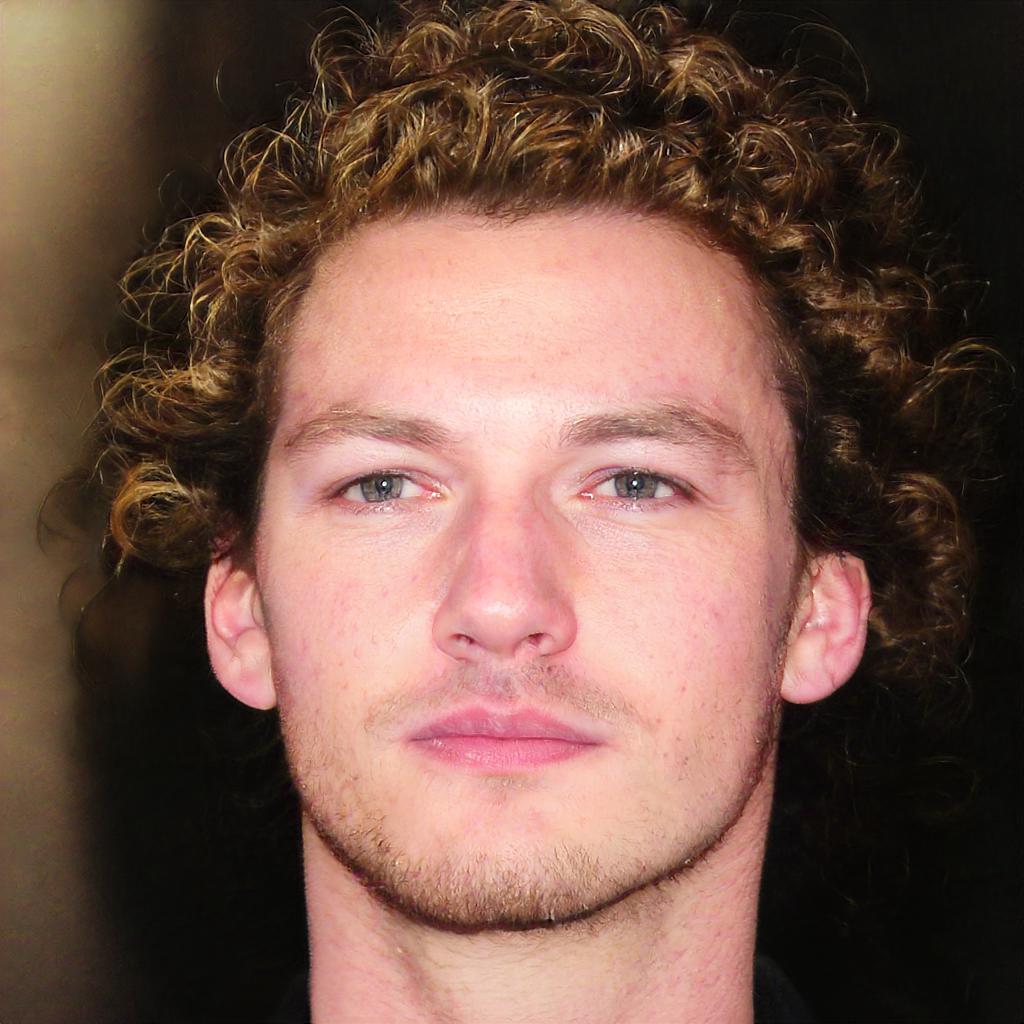} &
            \includegraphics[width=\linewidth]{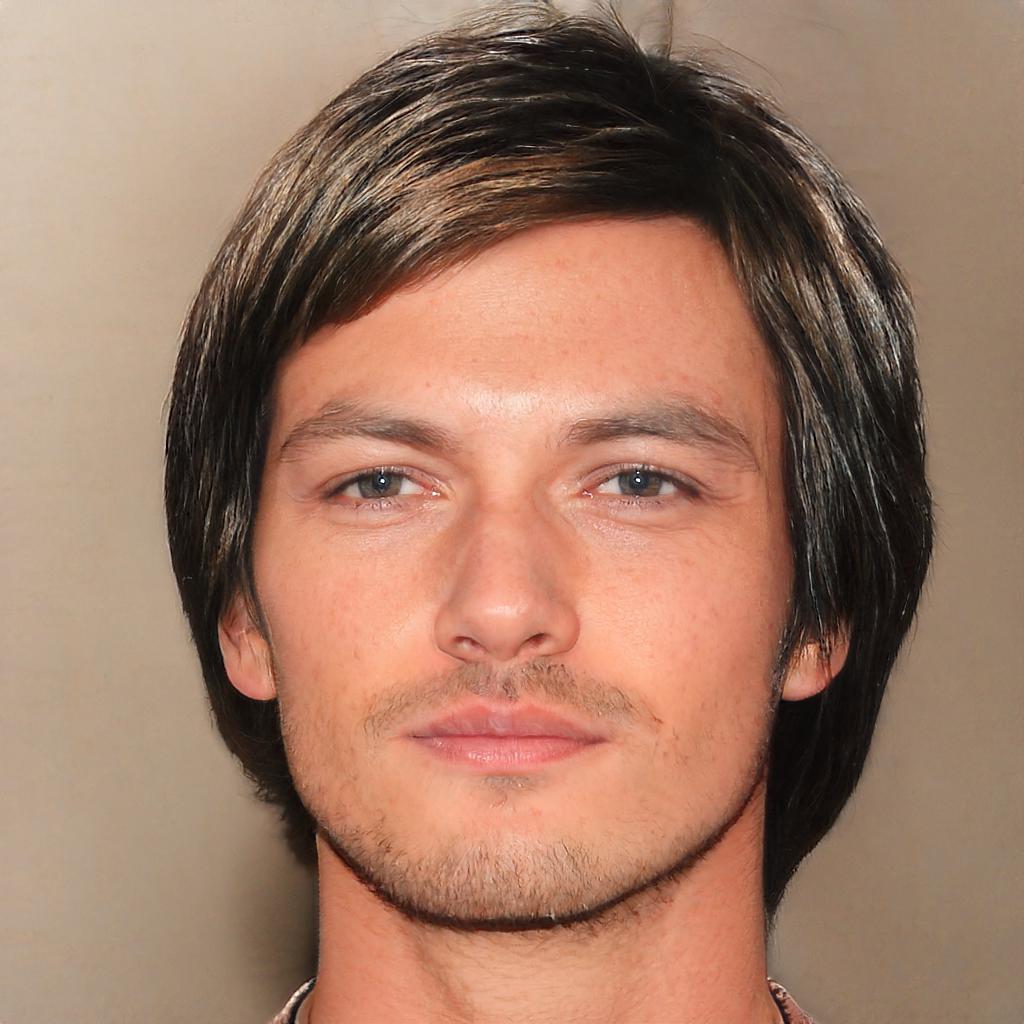} &
            \includegraphics[width=\linewidth]{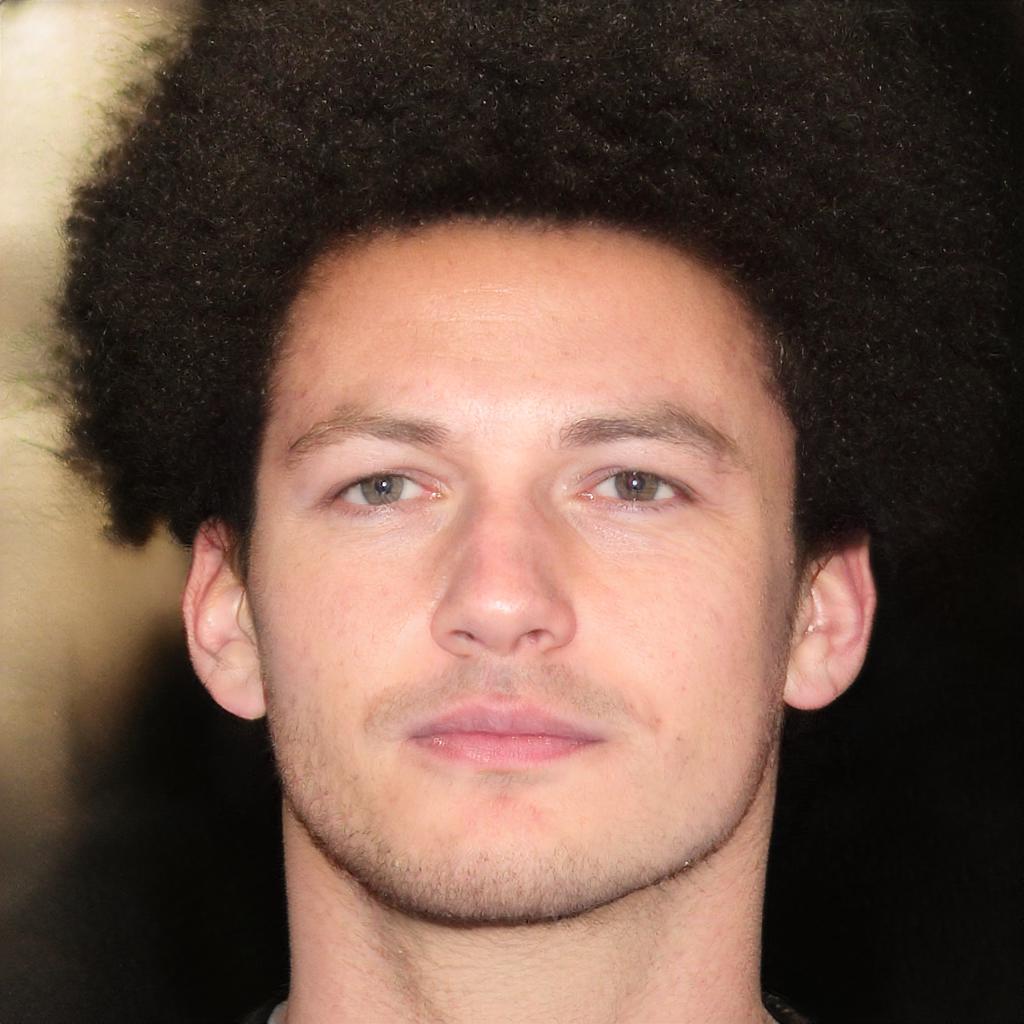} 
			\\
			\includegraphics[width=\linewidth]{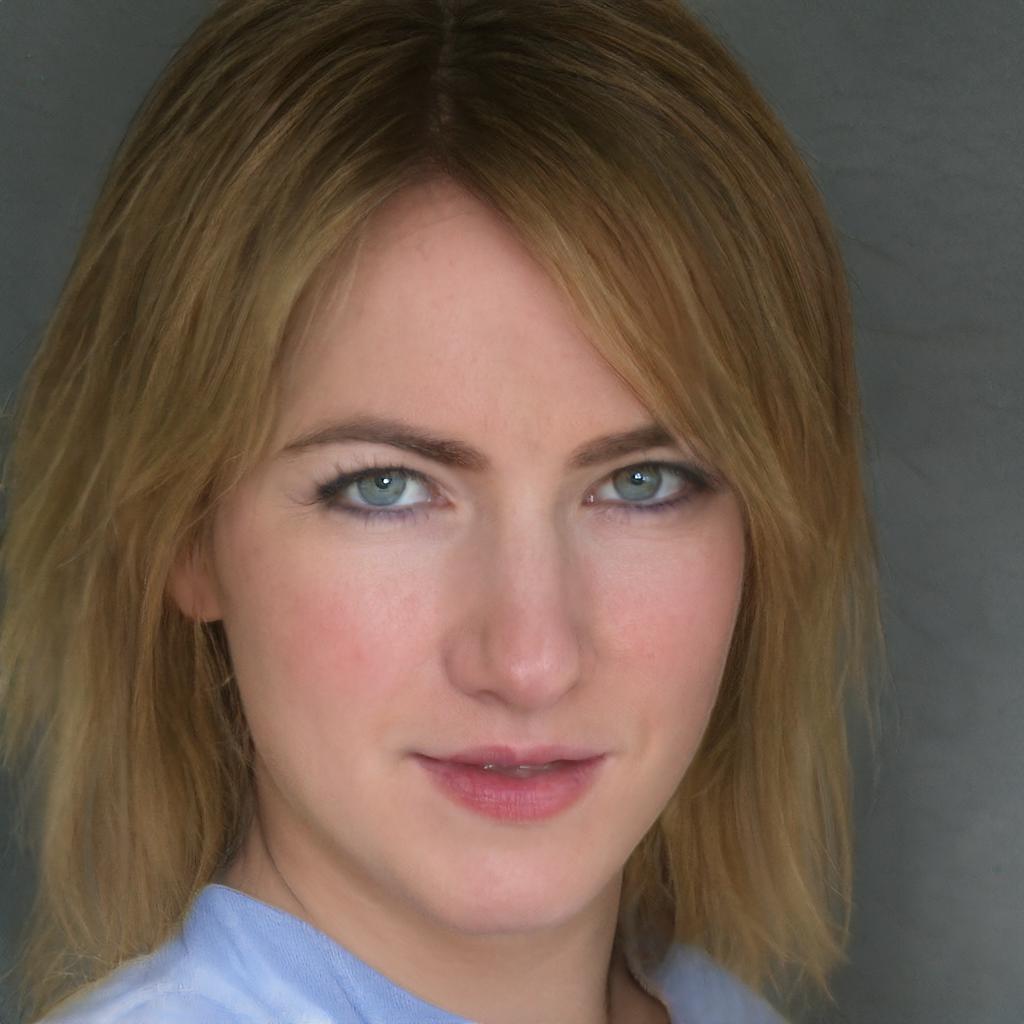} &
            \includegraphics[width=\linewidth]{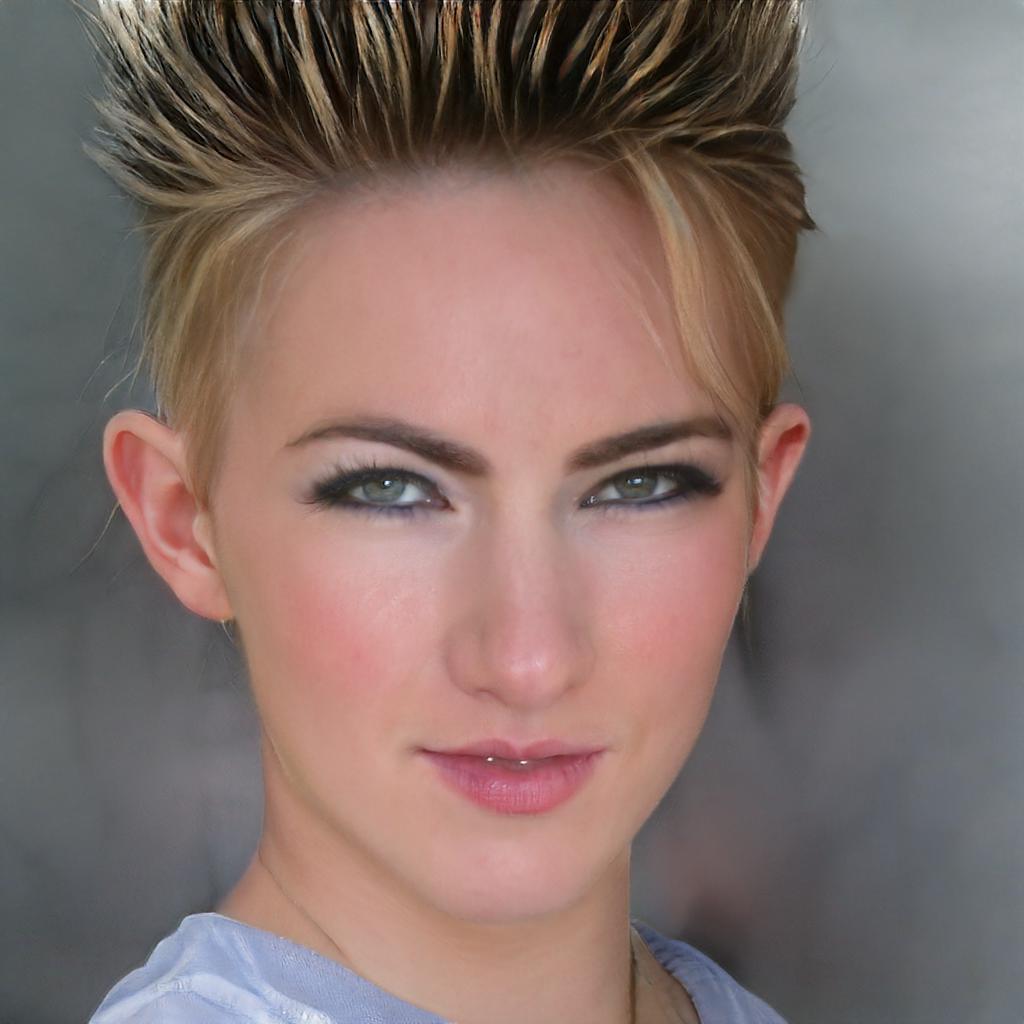} &
            \includegraphics[width=\linewidth]{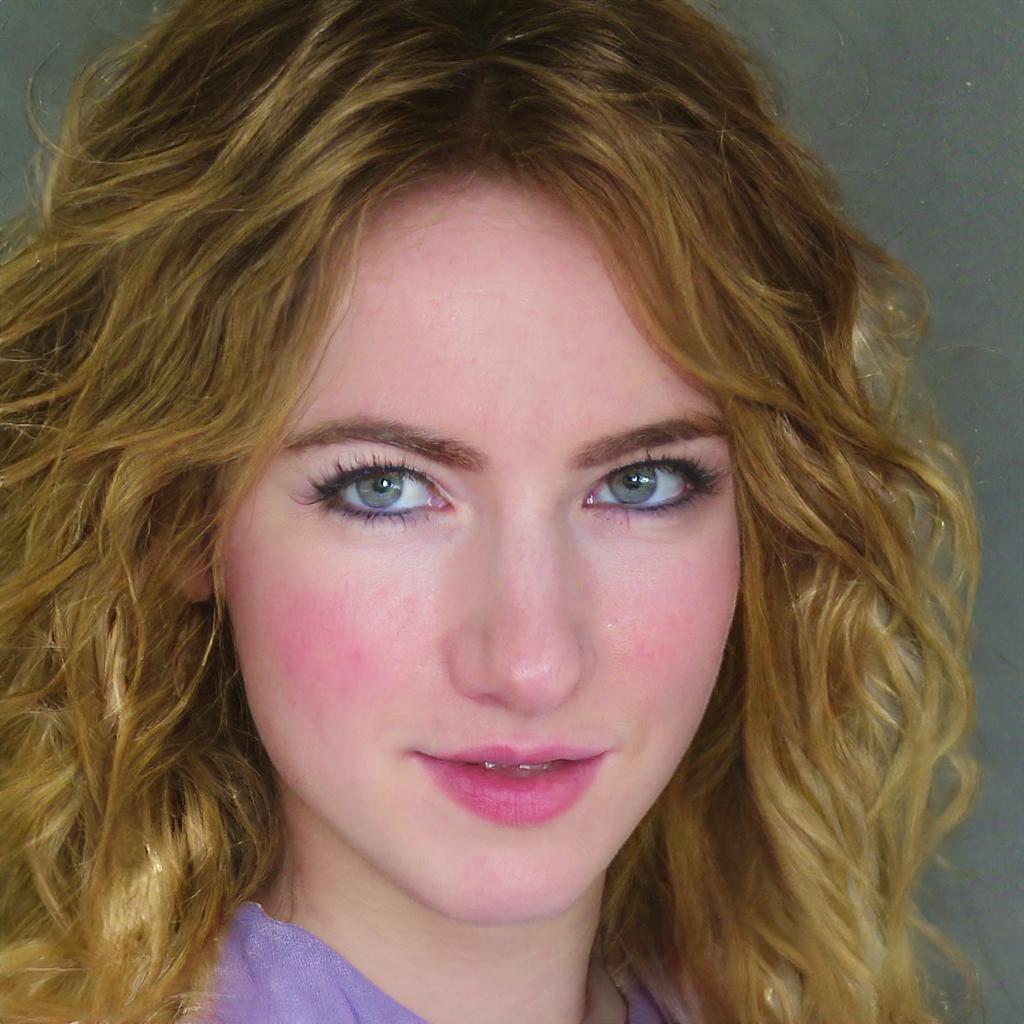} &
            \includegraphics[width=\linewidth]{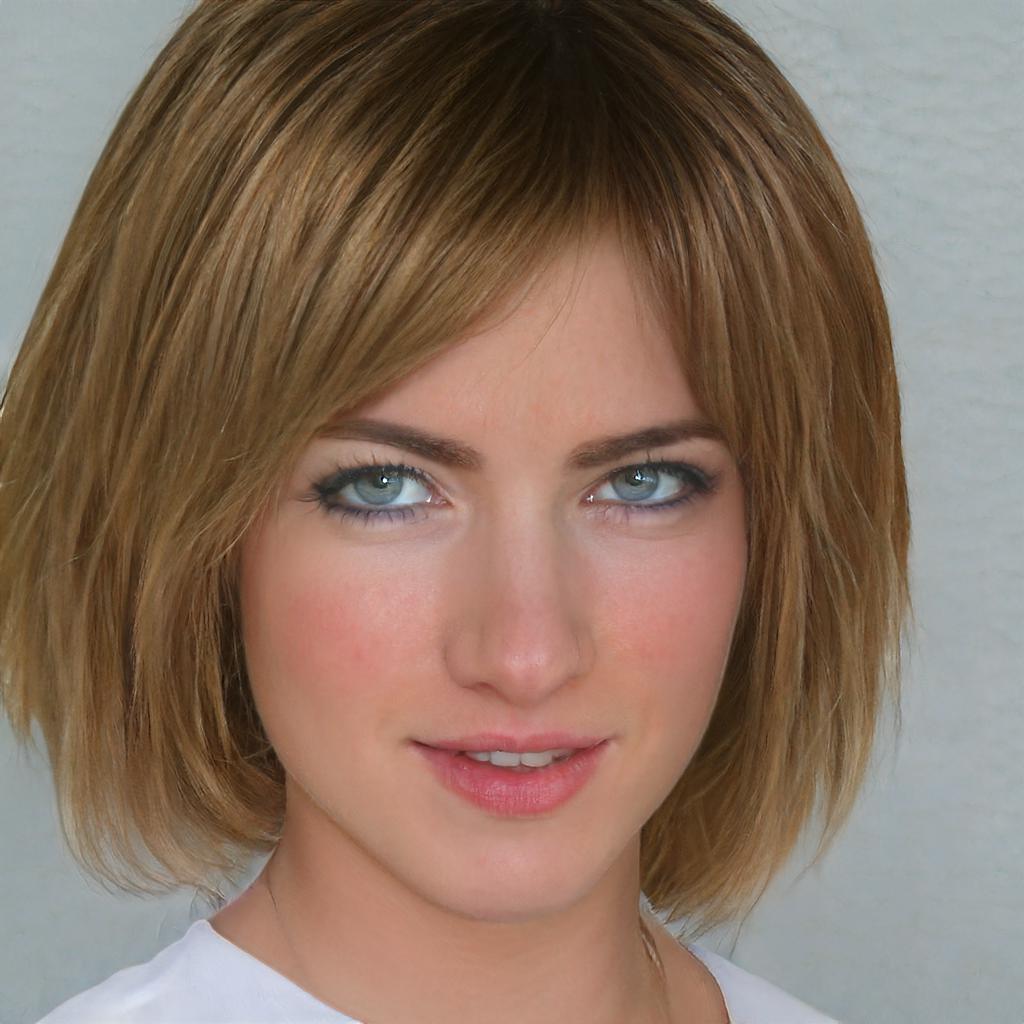} &
            \includegraphics[width=\linewidth]{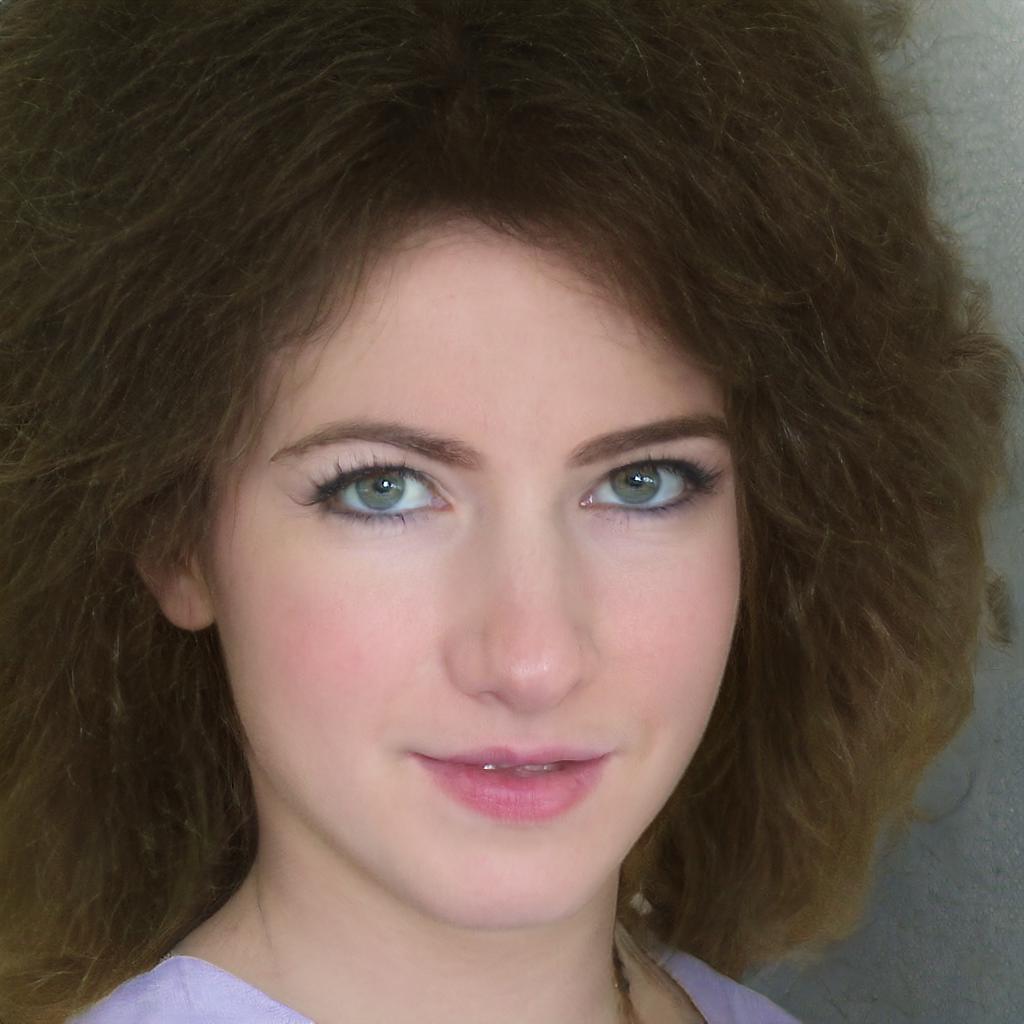} 
			\\
			\includegraphics[width=\linewidth]{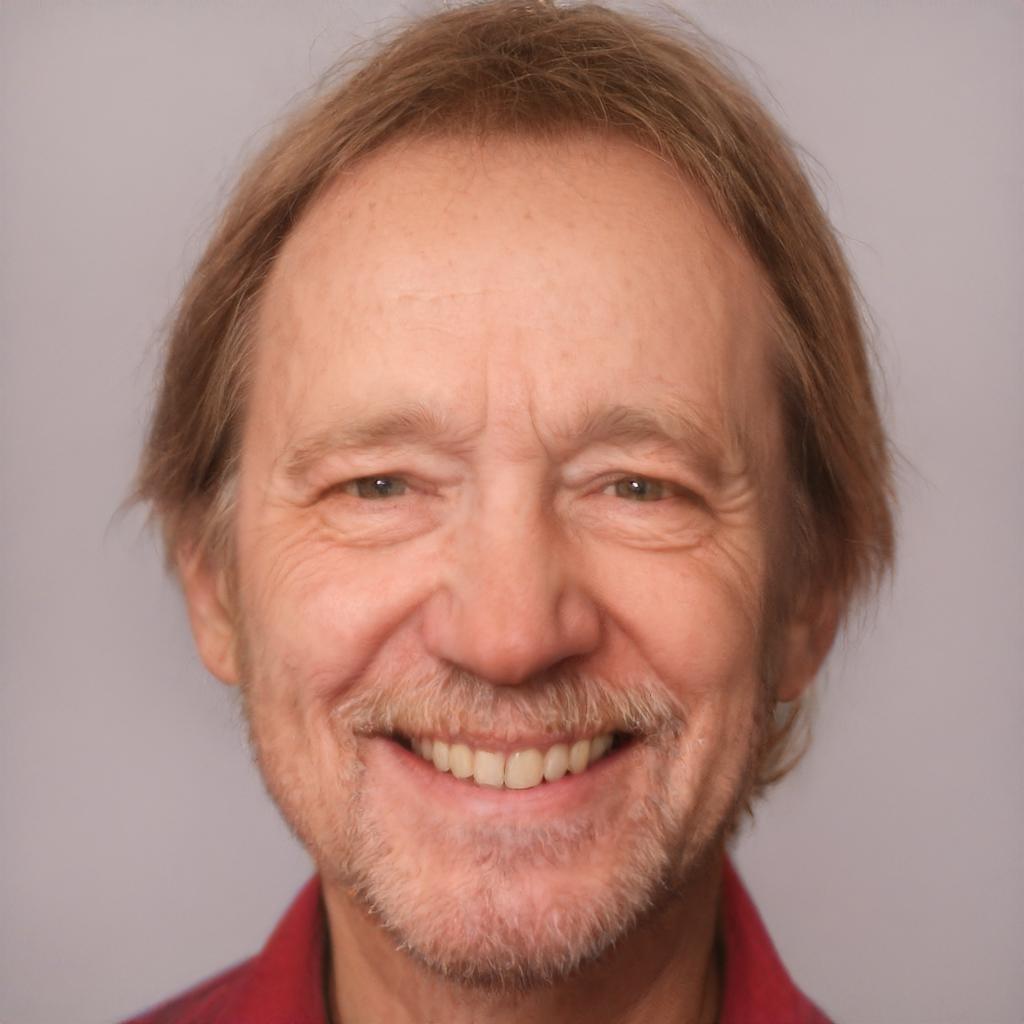} &
            \includegraphics[width=\linewidth]{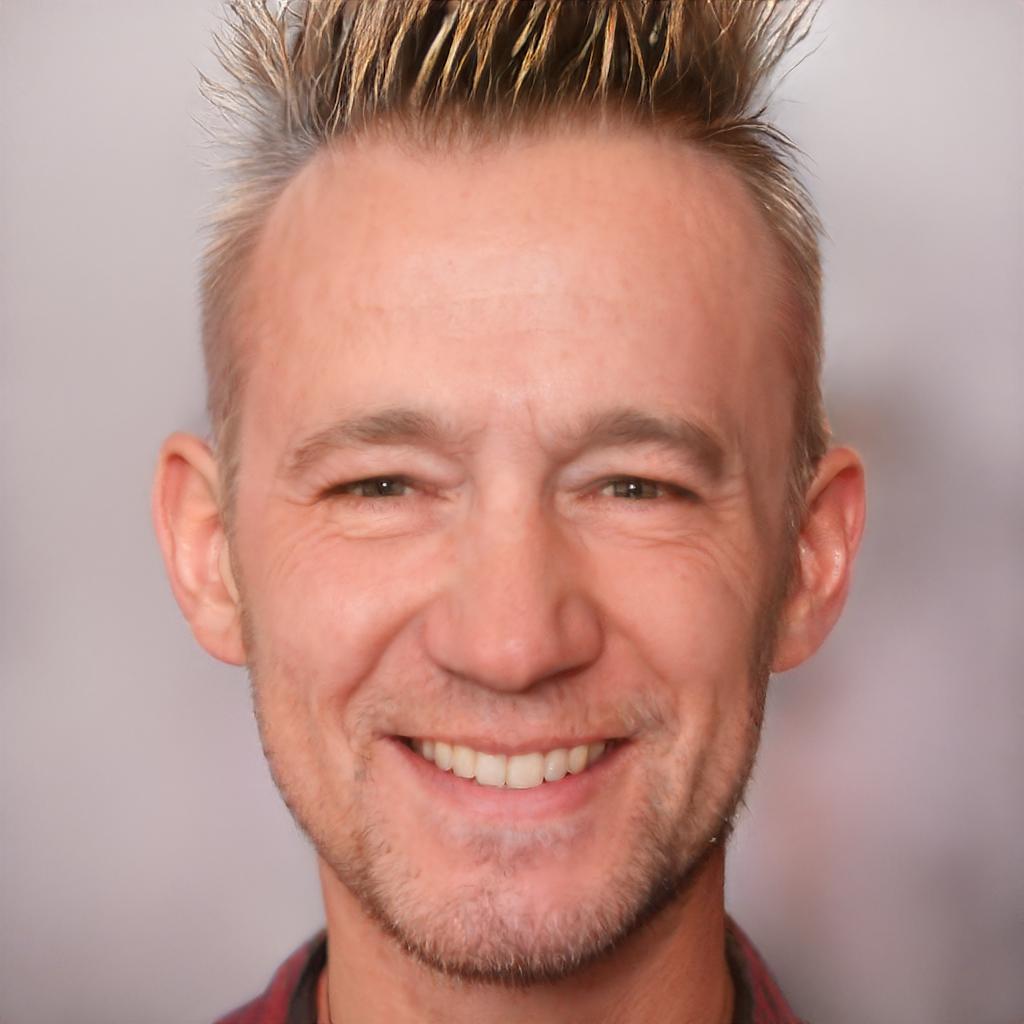} &
            \includegraphics[width=\linewidth]{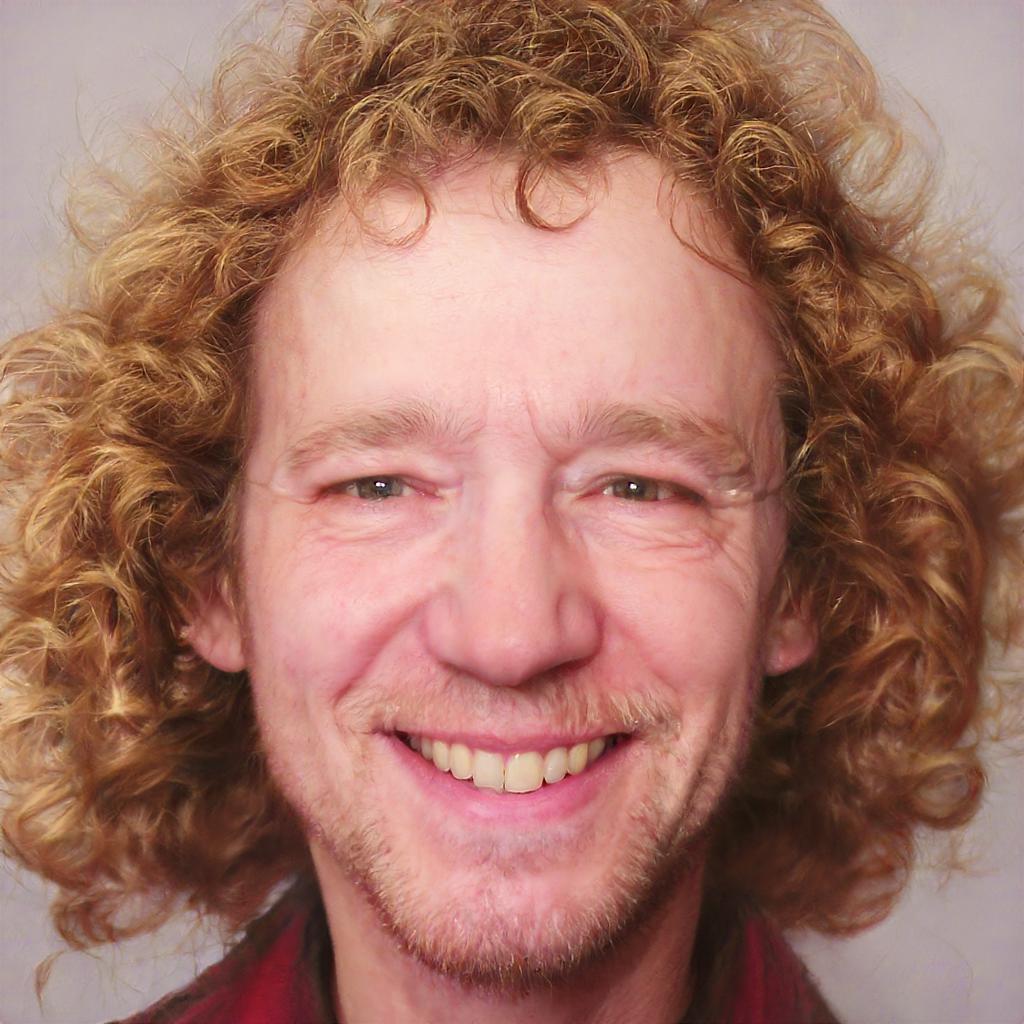} &
            \includegraphics[width=\linewidth]{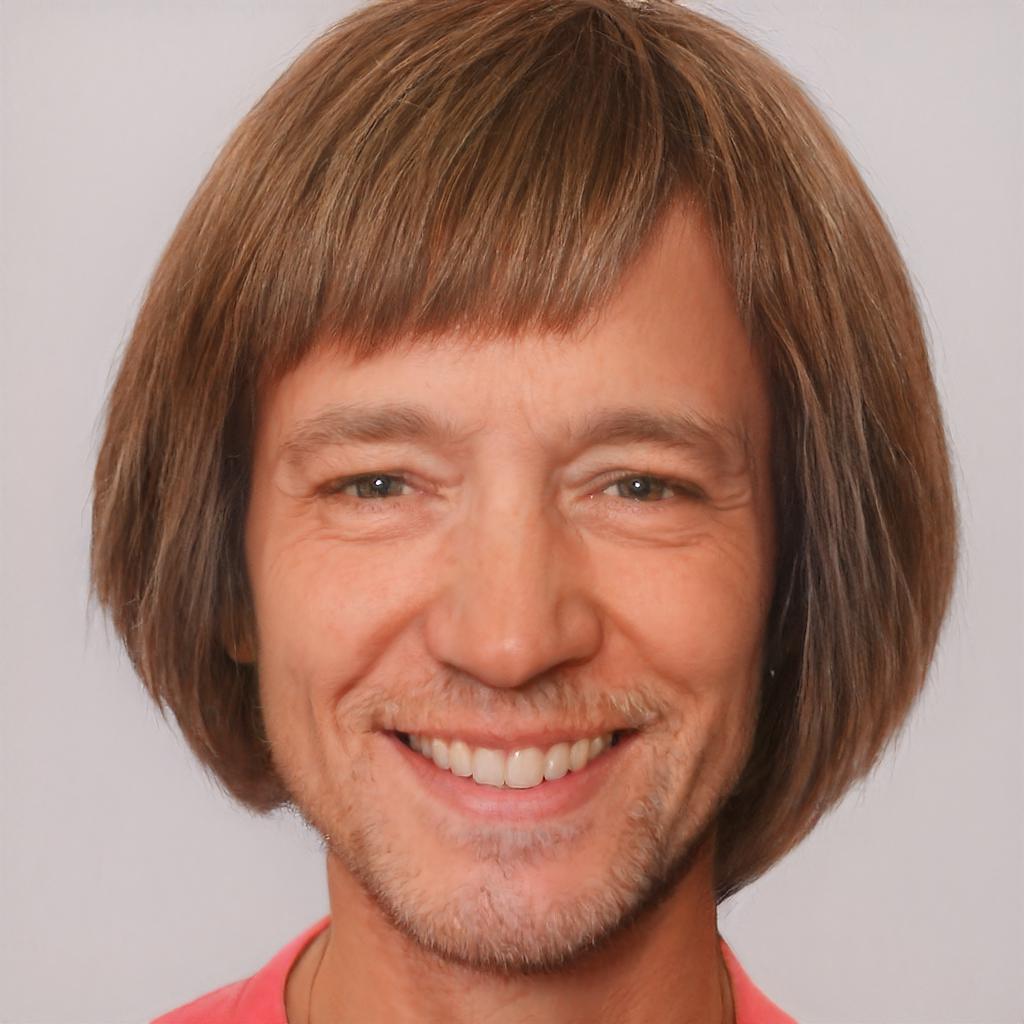} &
            \includegraphics[width=\linewidth]{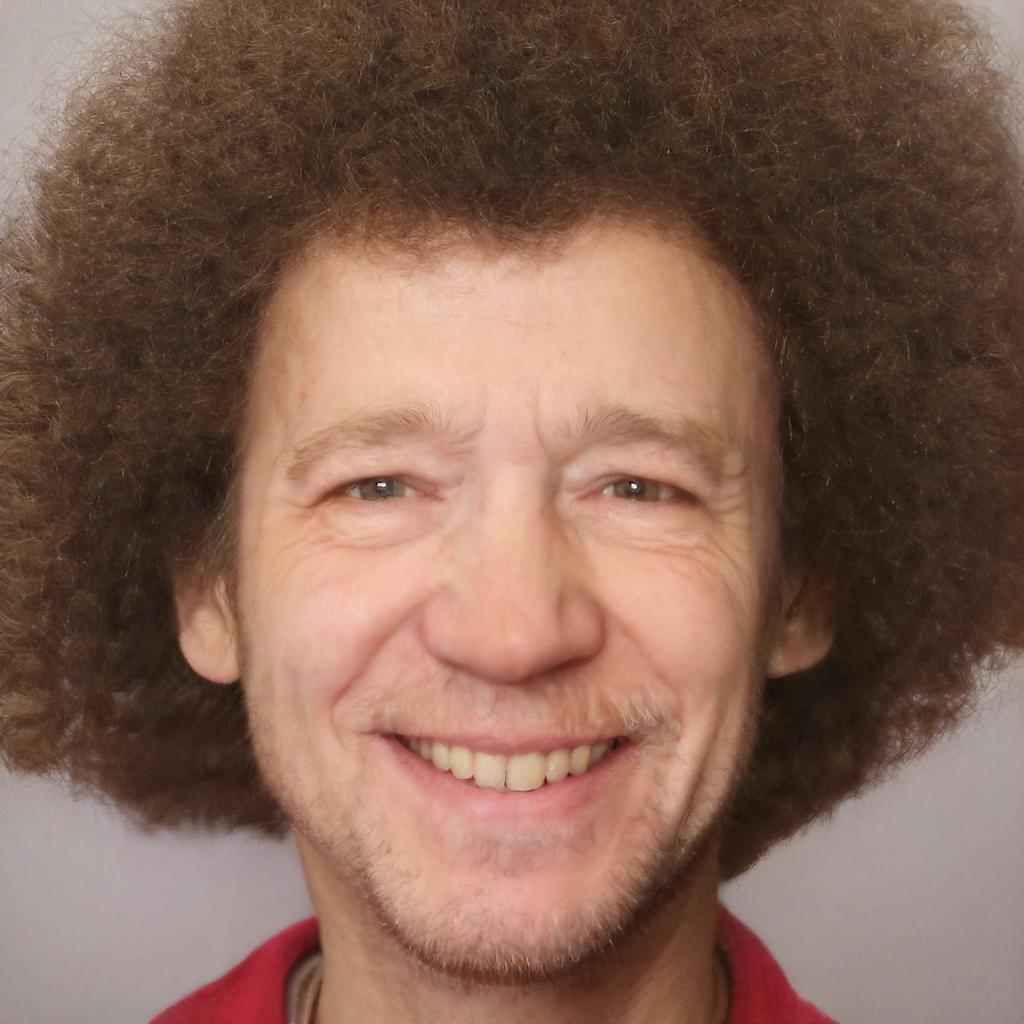} 
			\\
			Input & ``Mohawk hairstyle'' & ``Curly hair'' & ``Bob-cut hairstyle'' &  ``Afro hairstyle'' 
		\end{tabular}
	}
	\caption{Hair style edits using our mapper. The driving text prompts are indicated below each column. All input images are inversions of real images.}
	\label{fig:mapper-hair}
\end{figure}

\paragraph{Losses}
Our mapper is trained to manipulate the desired attributes of the image as indicated by the text prompt $t$, while preserving the other visual attributes of the input image. The CLIP loss, $\mathcal{L}_\text{CLIP}(w)$ guides the mapper to minimize the cosine distance in the CLIP latent space:
\begin{equation}
    \mathcal{L}_\text{CLIP}(w) = \Dclip(G(w + M_t(w)), t),
\end{equation}
where $G$ denotes again the pretrained StyleGAN generator. To preserve the visual attributes of the original input image, we minimize the $L_2$ norm of the manipulation step in the latent space.
Finally, for edits that require identity preservation, we use the identity loss defined in eq.~(\ref{eq:id-loss}).
Our total loss function is a weighted combination of these losses:
\begin{equation}
    \mathcal{L}(w) = \mathcal{L}_\text{CLIP}(w) + \lambda_{L2}\norm{M_t(w)}_2  + \lambda_\text{ID} \mathcal{L}_\text{ID}(w).
\end{equation}
As before, when the edit is expected to change the identity, we do not use the identity loss. 
The parameter values we use for the examples in this paper are $\lambda_\text{L2} = 0.8, \lambda_\text{ID} = 0.1$, 
except for the ``Trump'' manipulation in Figure~\ref{fig:global-vs-mapper}, where the parameter values
we use are $\lambda_\text{L2} = 2, \lambda_\text{ID} = 0$.

In Figure~\ref{fig:mapper-hair} we provide several examples for hair style edits, where a different mapper used in each column. 
In all of these examples, the mapper succeeds in preserving the identity and most of the other visual attributes that are not related to hair. Note, that the resulting hair appearance is adapted to the individual; this is particularly apparent in the ``Curly hair'' and ``Bob-cut hairstyle'' edits.

\begin{figure}[tb]
	\setlength{\tabcolsep}{1pt}
	\centering
	{\footnotesize
		\begin{tabular}{C{0.24\linewidth} C{0.24\linewidth} C{0.24\linewidth} C{0.24\linewidth}}
			\includegraphics[width=\linewidth]{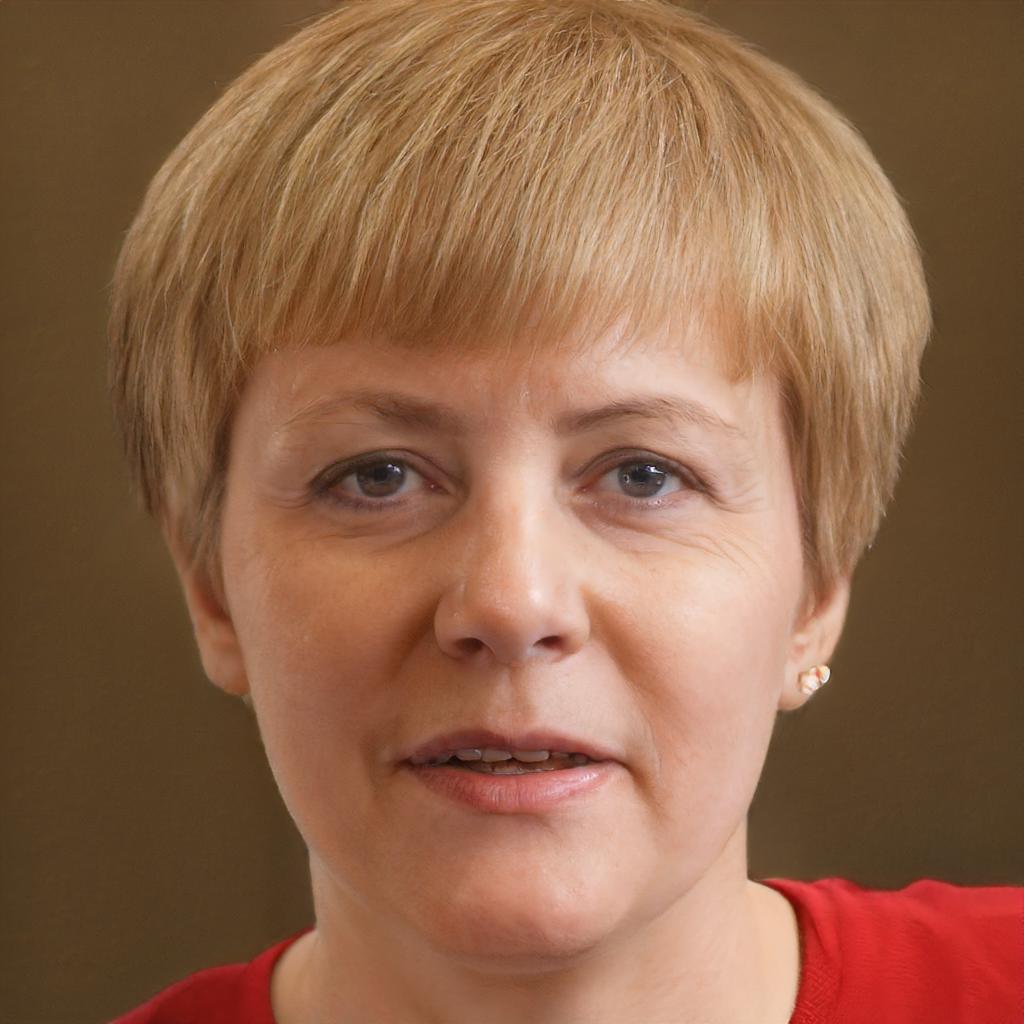} &
            \includegraphics[width=\linewidth]{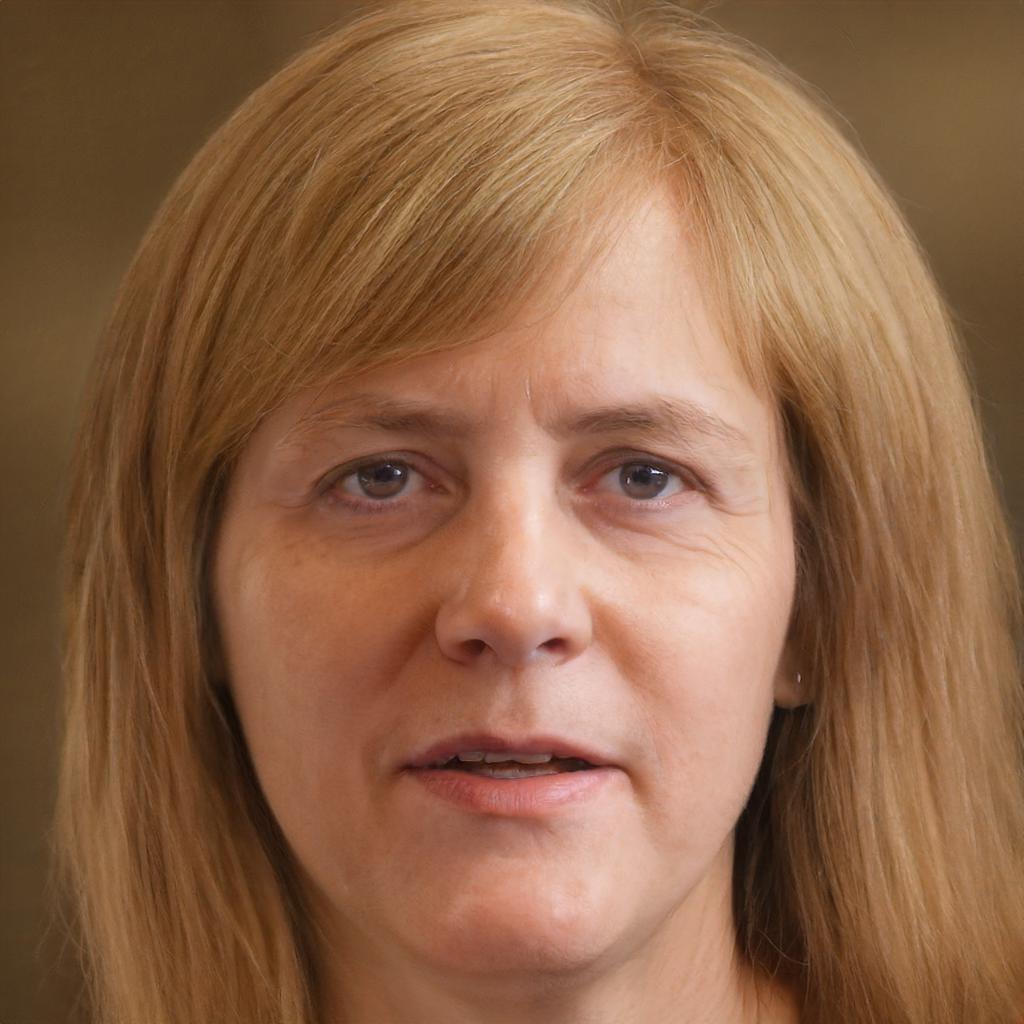} &
			\includegraphics[width=\linewidth]{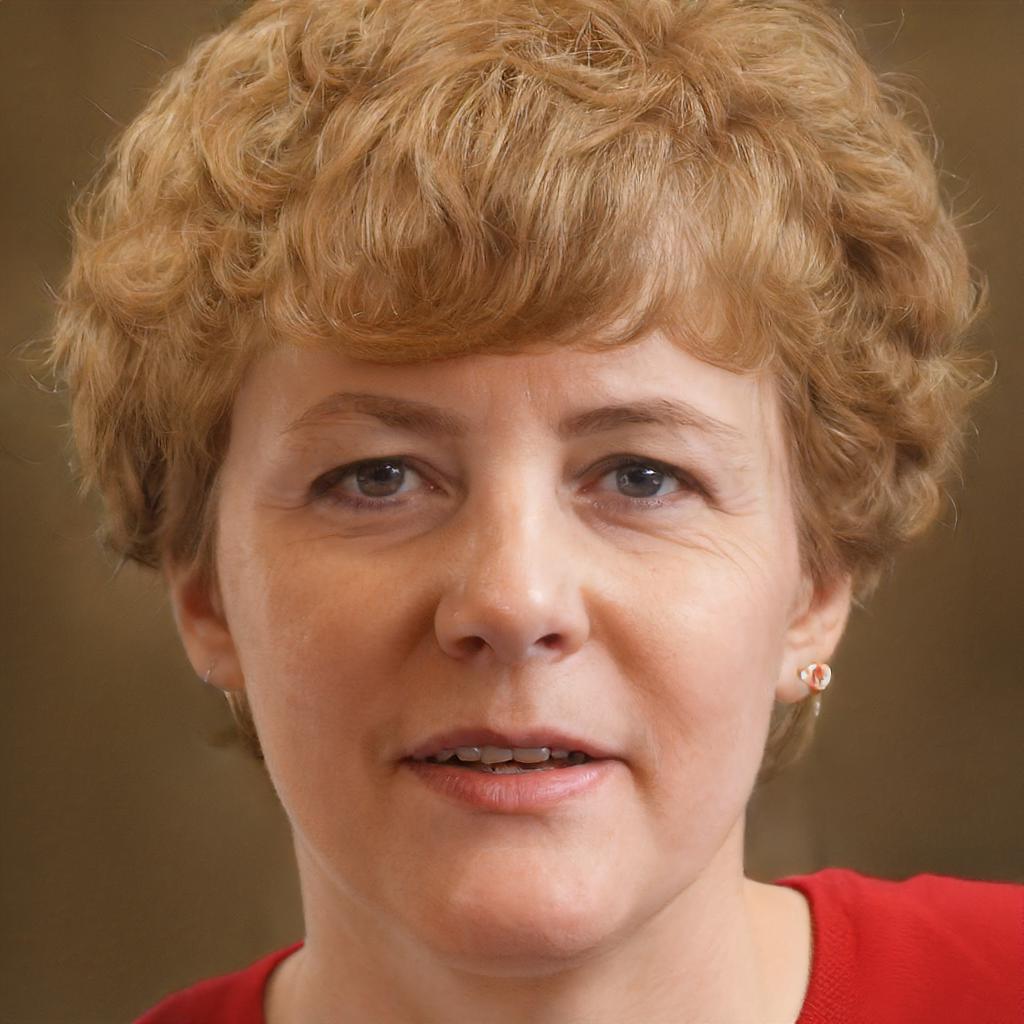} &
            \includegraphics[width=\linewidth]{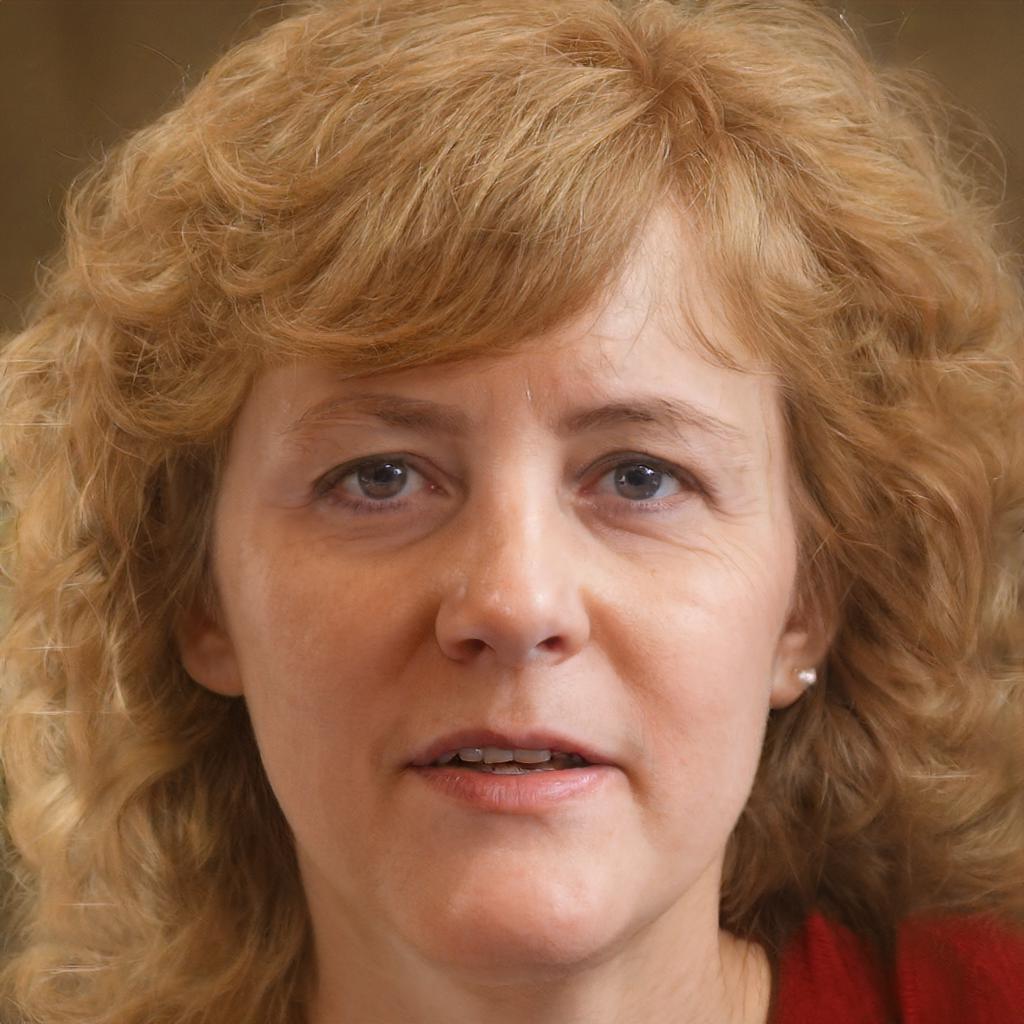} \\
            ``Straight short hair'' & ``Straight long hair'' & ``Curly short hair'' & ``Curly long hair''

		\end{tabular}
	}
	\caption{Controlling more than one attribute with a single mapper. The driving text for each mapper is indicated below each column.}
	\label{fig:mapper-multi}
\end{figure}

It should be noted that the text prompts are not limited to a single attribute at a time. Figure~\ref{fig:mapper-multi}
shows four different combinations of hair attributes, straight/curly and short/long, each yielding the expected outcome. This degree of control has not been demonstrated by any previous method we're aware of.

Since the latent mapper infers a custom-tailored manipulation step for each input image, it is interesting to examine the extent to which the direction of the step in latent space varies over different inputs.
To test this, we first invert the test set of CelebA-HQ~\cite{liu2015deep, karras2017progressive} using e4e \cite{tov2021designing}. Next, we feed the inverted latent codes into several trained mappers and compute the cosine similarity between all pairs of the resulting manipulation directions. The mean and the standard deviation of the cosine similarity for each mapper is reported in Table~\ref{tab:mapper-directions}.
The table shows that even though the mapper infers manipulation steps that are adapted to the input image, in practice, the cosine similarity of these steps for a given text prompt is high, implying that their directions are not as different as one might expect.

\section{Global Directions}
\label{sec:global}
\vspace{-1mm}
While the latent mapper allows fast inference time, we find that it sometimes falls short when a fine-grained disentangled manipulation is desired. 
Furthermore, as we have seen, the directions of different manipulation steps for a given text prompt tend to be similar. 
Motivated by these observations, in this section we propose a method for mapping a text prompt into a single, global direction in StyleGAN's style space $\Sspace$, which has been shown to be more disentangled than other latent spaces~\cite{wu2020stylespace}.

Let $s \in \Sspace$ denote a style code, and $G(s)$ the corresponding generated image.
Given a text prompt indicating a desired attribute, we seek a manipulation direction $\Ds$, such that $G(s + \alpha\Ds)$ yields an image where that attribute is introduced or amplified, without significantly affecting other attributes. 
The manipulation strength is controlled by $\alpha$.
Our high-level idea is to first use the CLIP text encoder to obtain a vector $\Dt$ in CLIP's joint language-image embedding and then map this vector into a manipulation direction $\Ds$ in $\Sspace$.
A stable $\Dt$ is obtained from natural language, using prompt engineering, as described below.
The corresponding direction $\Ds$ is then determined by assessing the relevance of each style channel to the target attribute.

More formally, denote by $\clipI$ the manifold of image embeddings in CLIP's joint embedding space, and by $\clipT$ the manifold of its text embeddings. We distinguish between these two manifolds, because there is no one-to-one mapping between them: an image may contain a large number of visual attributes, which can hardly be comprehensively described by a single text sentence; conversely, a given sentence may describe many different images.
During CLIP training, all embeddings are normalized to a unit norm, and therefore only the direction of embedding contains semantic information, while the norm may be ignored. Thus, in well trained areas of the CLIP space, we expect directions on the $\clipT$ and $\clipI$ manifolds that correspond to the same semantic changes to be roughly collinear (i.e., have large cosine similarity), and nearly identical after normalization. 

Given a pair of images, $G(s)$ and $G(s+\alpha\Ds)$, we denote their $\clipI$ embeddings by $i$ and $i + \Di$, respectively. Thus, the difference between the two images in CLIP space is given by $\Di$.
Given a natural language instruction encoded as $\Dt$, and assuming collinearity between $\Dt$ and $\Di$, we can determine a manipulation direction $\Ds$ by assessing the relevance of each channel in $\Sspace$ to the direction $\Di$.


\paragraph{From natural language to $\Dt$}

In order to reduce text embedding noise, Radford \etal~\cite{radford2021learning} utilize a technique called prompt engineering that feeds several sentences with the same meaning to the text encoder, and averages their embeddings. For example, for ImageNet zero-shot classification, a bank of 80 different sentence templates is used, such as ``a bad photo of a \{\}'', ``a cropped photo of the \{\}'', ``a black and white photo of a \{\}'', and ``a painting of a \{\}''. At inference time, the target class is automatically substituted into these templates to build a bank of sentences with similar semantics, whose embeddings are then averaged. This process improves zero-shot classification accuracy by an additional $3.5\%$ over using a single text prompt.


Similarly, we also employ prompt engineering (using the same ImageNet prompt bank) in order to compute stable directions in $\clipT$. Specifically, our method should be provided with text description of a target attribute and a corresponding neutral class. For example, when manipulating images of cars, the target attribute might be specified as ``a sports car'', in which case the corresponding neutral class might be ``a car''. Prompt engineering is then applied to produce the average embeddings for the target and the neutral class, and the normalized difference between the two embeddings is used as the target direction $\Dt$.


\paragraph{Channelwise relevance}

Next, our goal is to construct a style space manipulation direction $\Ds$ that would yield a change $\Di$, collinear with the target direction $\Dt$.
For this purpose, we need to assess the relevance of each channel $c$ of $\Sspace$ to a given direction $\Di$ in CLIP's joint embedding space.
We generate a collection of style codes $s \in \Sspace$, and perturb only the $c$ channel of each style code by adding a negative and a positive value.
Denoting by $\Di_c$ the CLIP space direction between the resulting pair of images, the relevance of channel $c$ to the target manipulation is estimated as the mean projection of $\Di_c$ onto $\Di$:
\begin{equation}
	R_c(\Di) = \mathbb{E}_{s \in \Sspace} \{ \Di_c \cdot \Di \}
\end{equation}

In practice, we use 100 image pairs to estimate the mean. 
The pairs of images that we generate are given by $G(s \pm \alpha \Ds_c)$, where $\Ds_c$ is a zero 
vector, except its $c$ coordinate, which is set to the standard deviation of the channel. The magnitude of the perturbation is set to $\alpha=5$. 

\begin{figure}[tb]
	\centering
	\setlength{\tabcolsep}{1pt}	
	\begin{tabular}{cccccc}
		& {\footnotesize $\alpha=-6$} & {\footnotesize $\alpha=-2$} & {\footnotesize Original} &{\footnotesize $\alpha=2$} &{\footnotesize $\alpha=6$} \\
		\rotatebox{90}{\footnotesize \phantom{k} $\beta=0.16$} &
		\includegraphics[width=0.18\columnwidth]{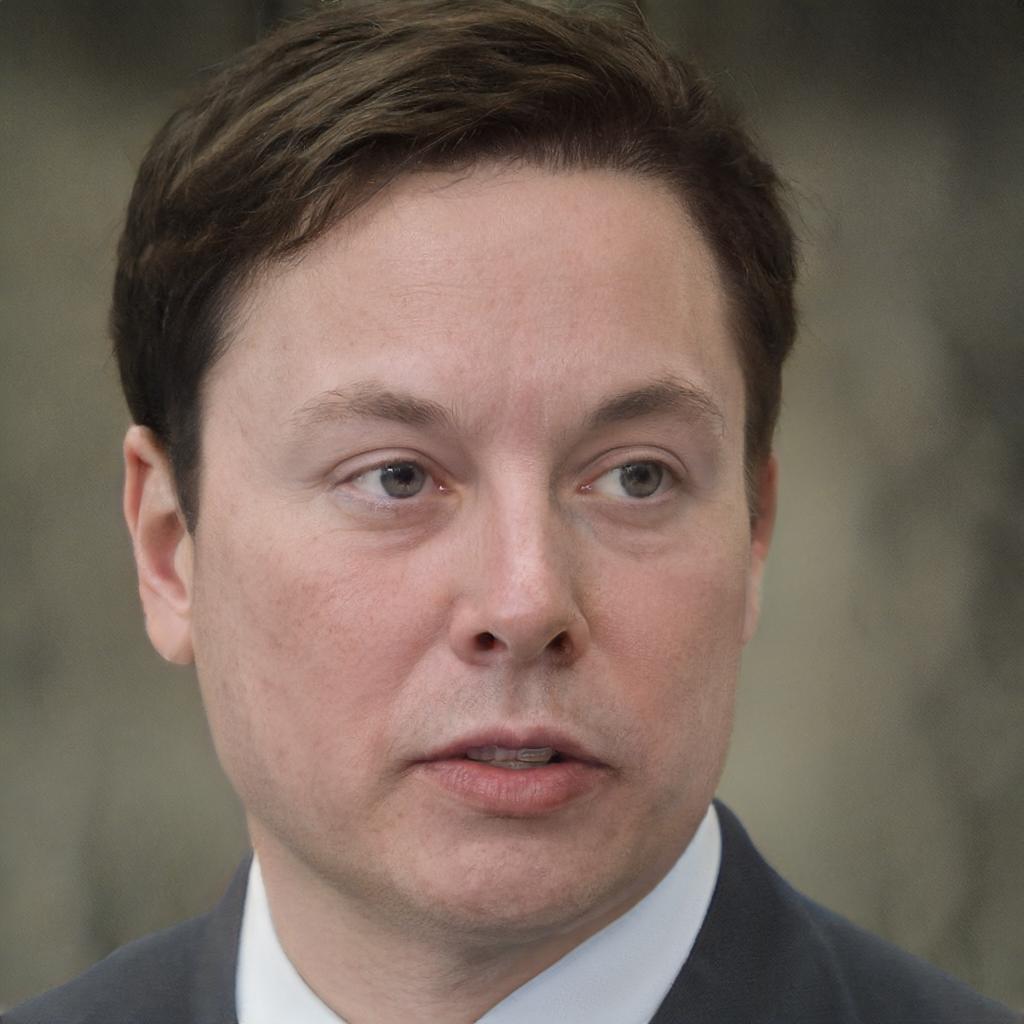} &
		\includegraphics[width=0.18\columnwidth]{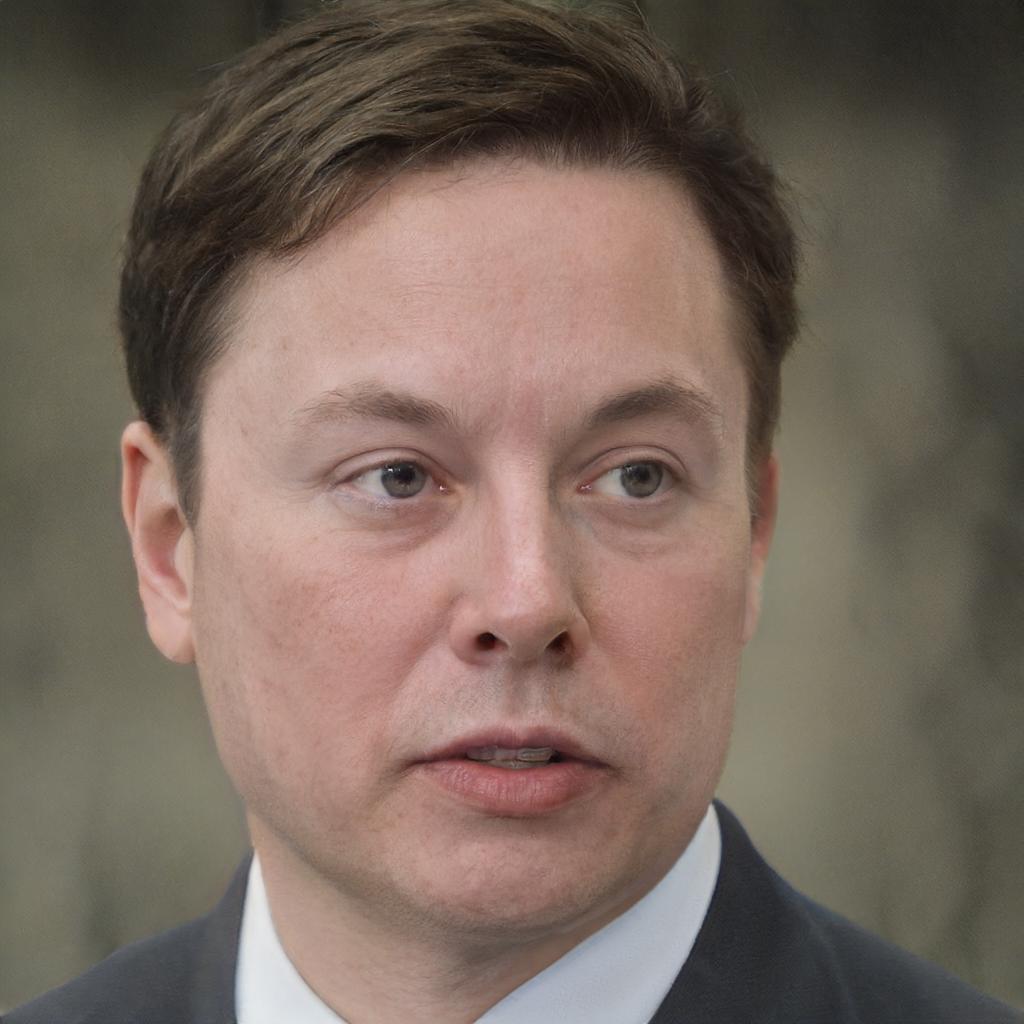} &
		\includegraphics[width=0.18\columnwidth]{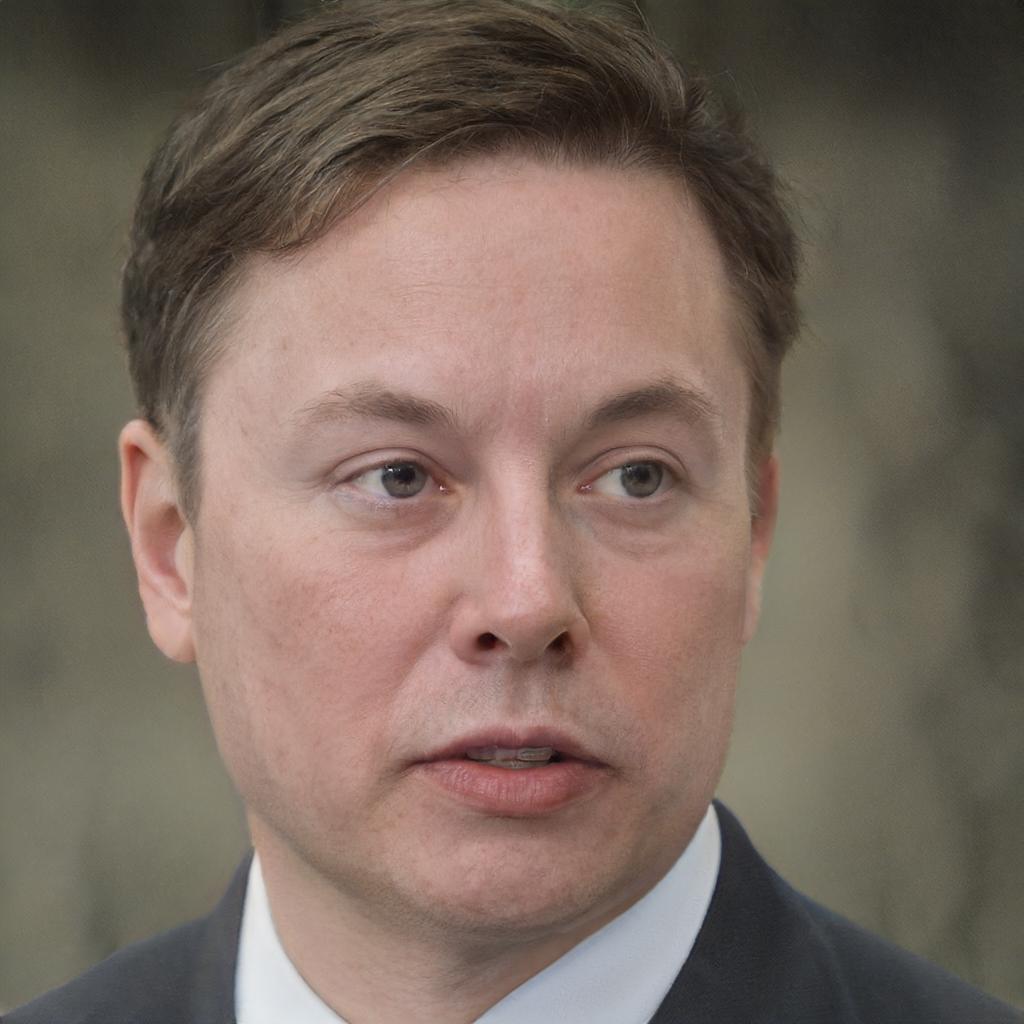} &
		\includegraphics[width=0.18\columnwidth]{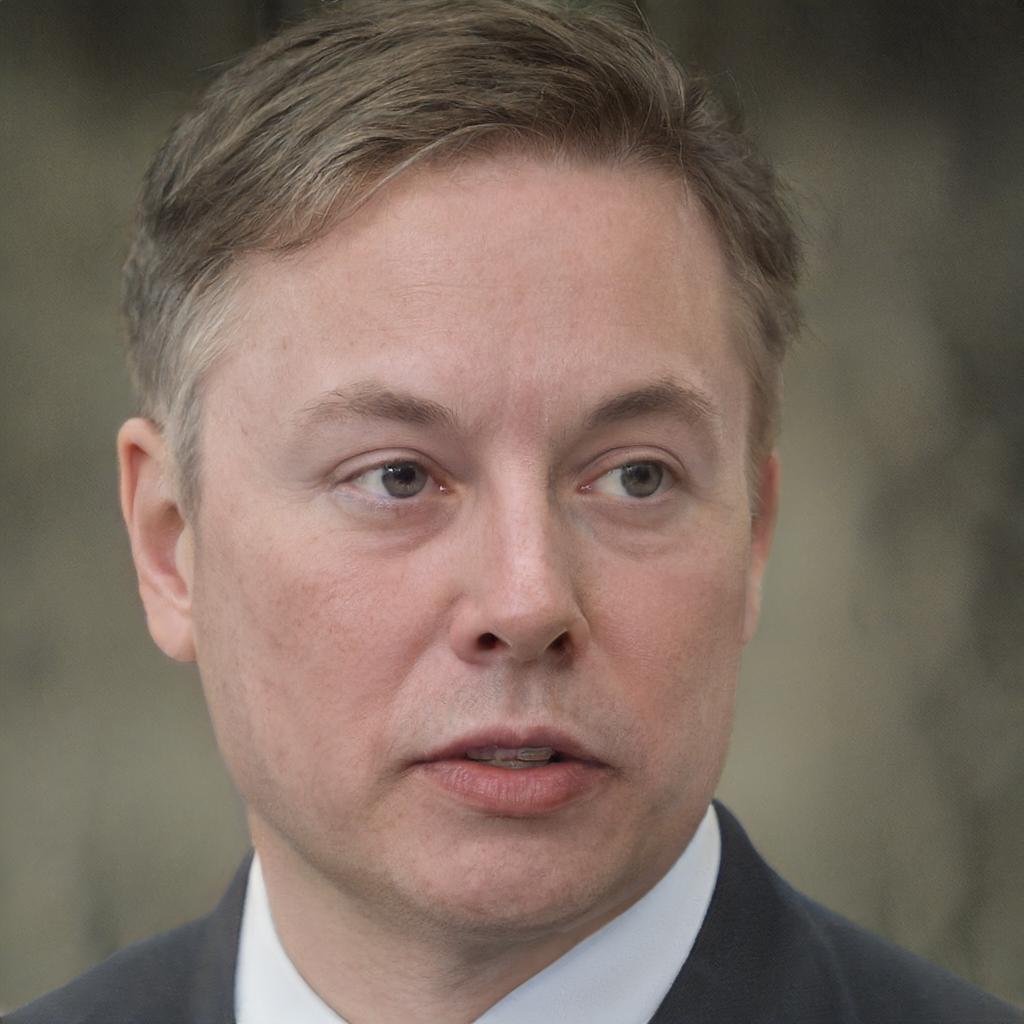} &
		\includegraphics[width=0.18\columnwidth]{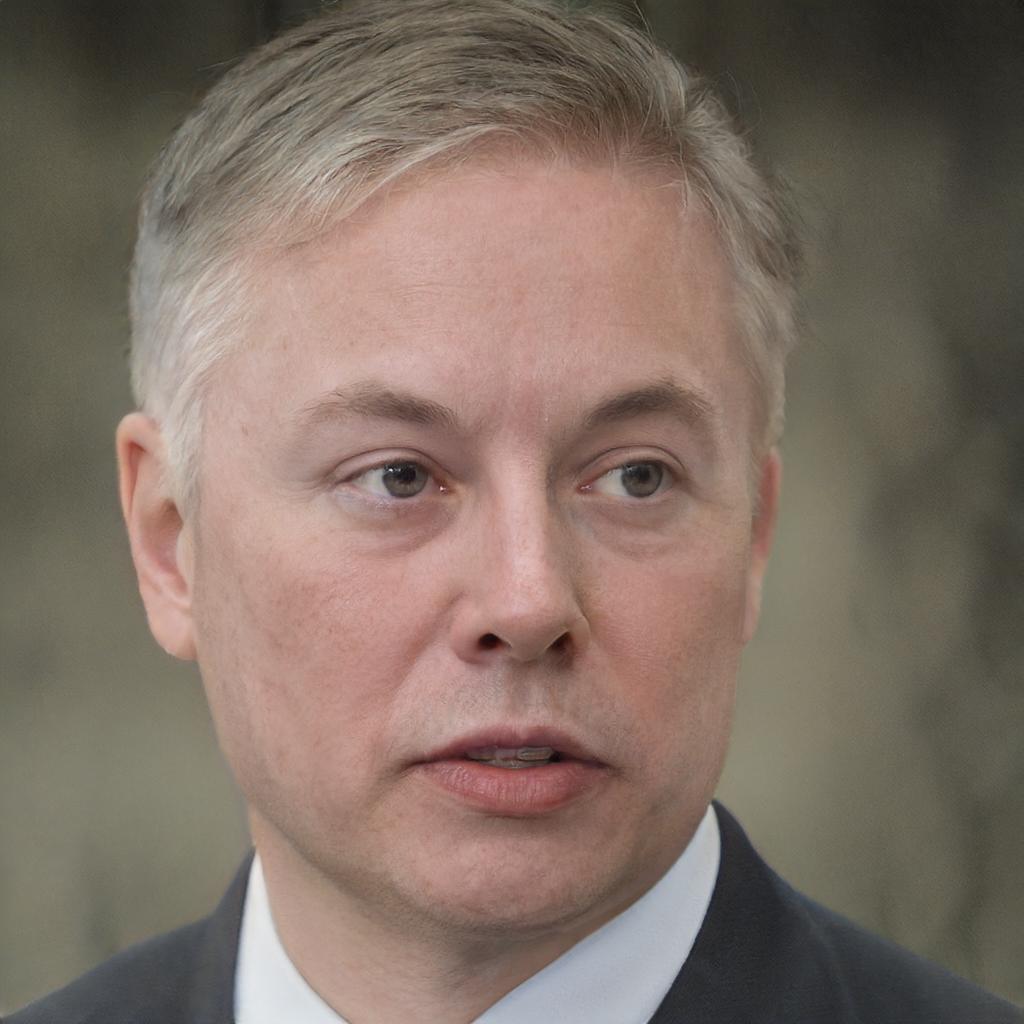} 
		\\
		\rotatebox{90}{\footnotesize \phantom{k} $\beta=0.14$} &
		\includegraphics[width=0.18\columnwidth]{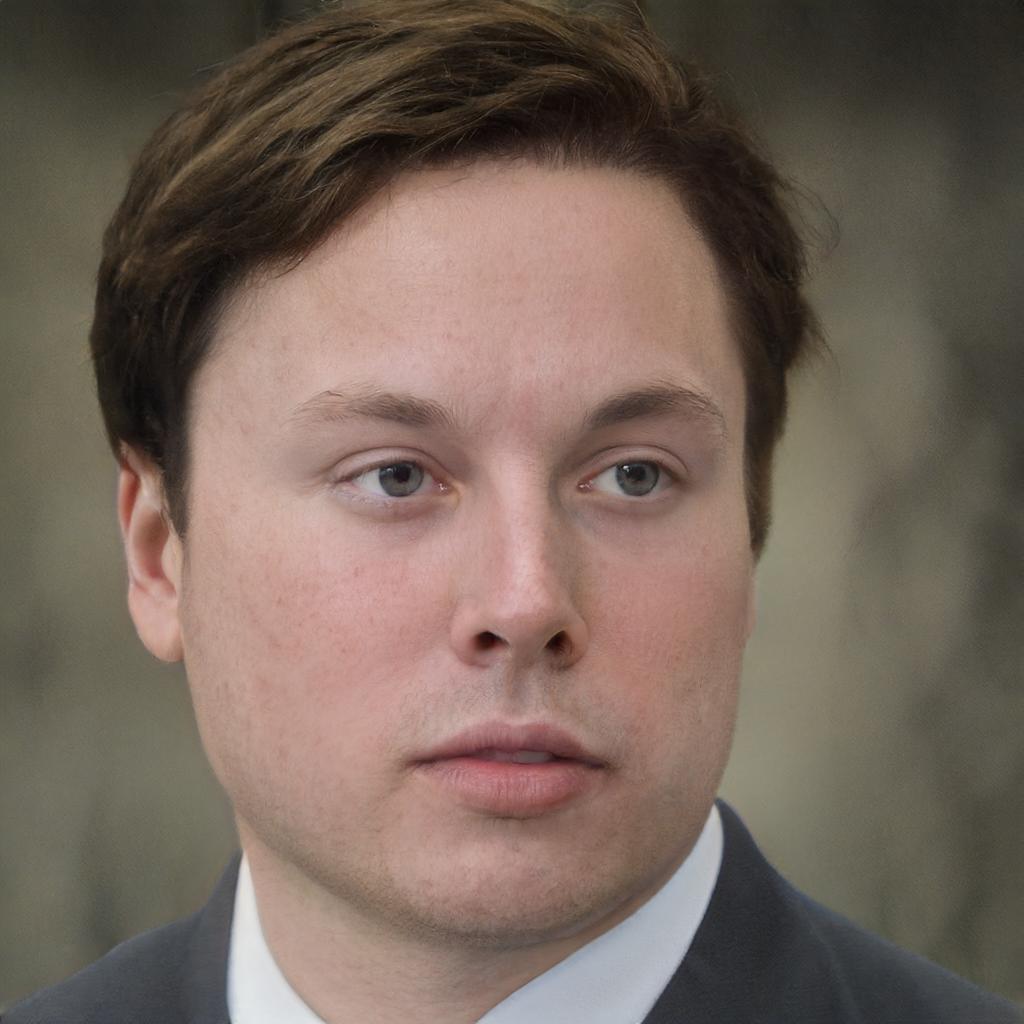} &
		\includegraphics[width=0.18\columnwidth]{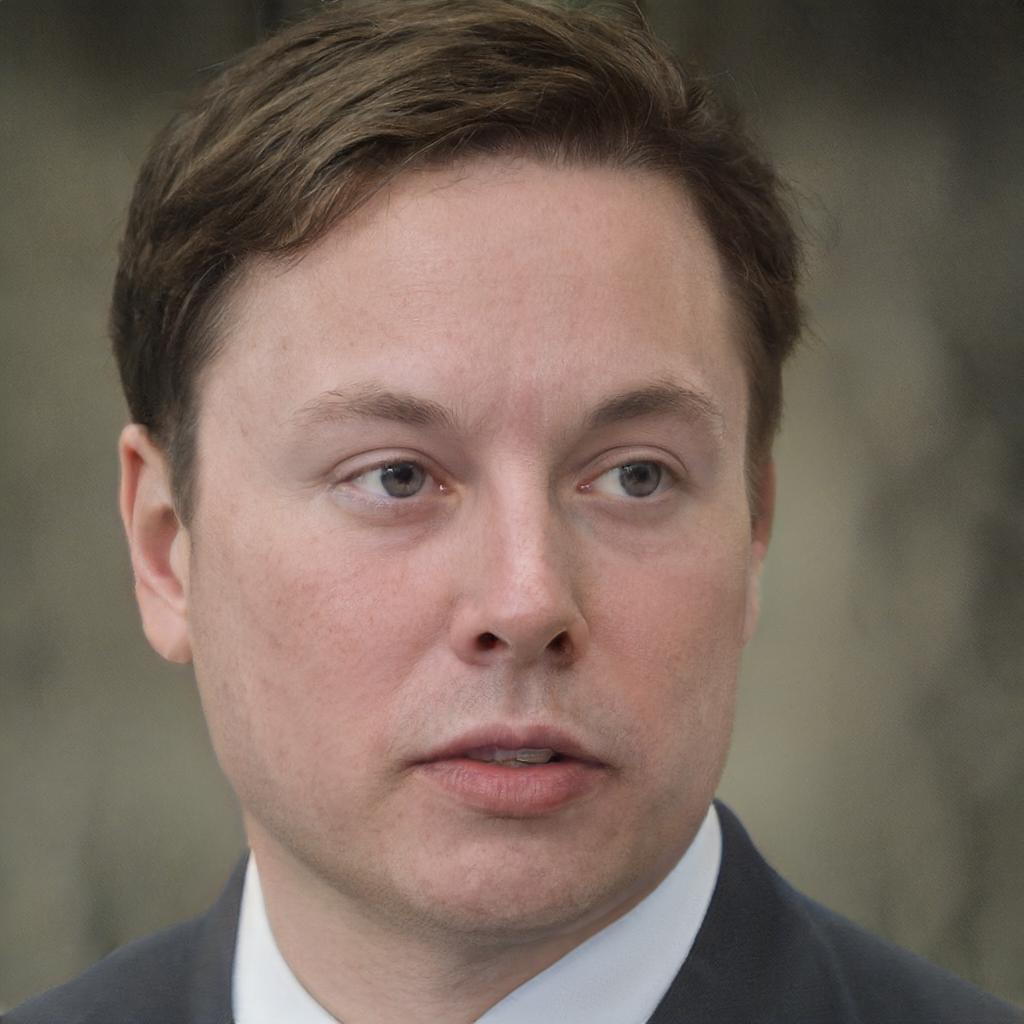} &
		\includegraphics[width=0.18\columnwidth]{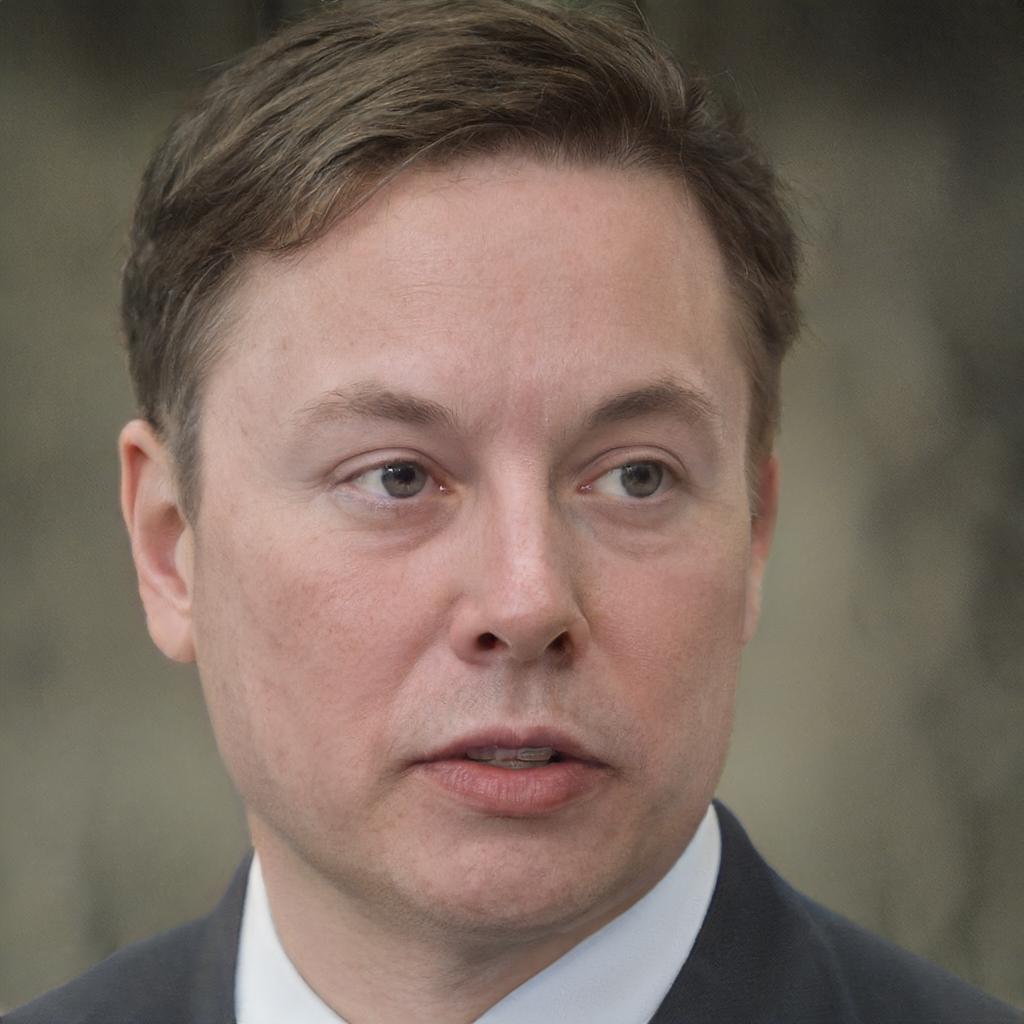} &
		\includegraphics[width=0.18\columnwidth]{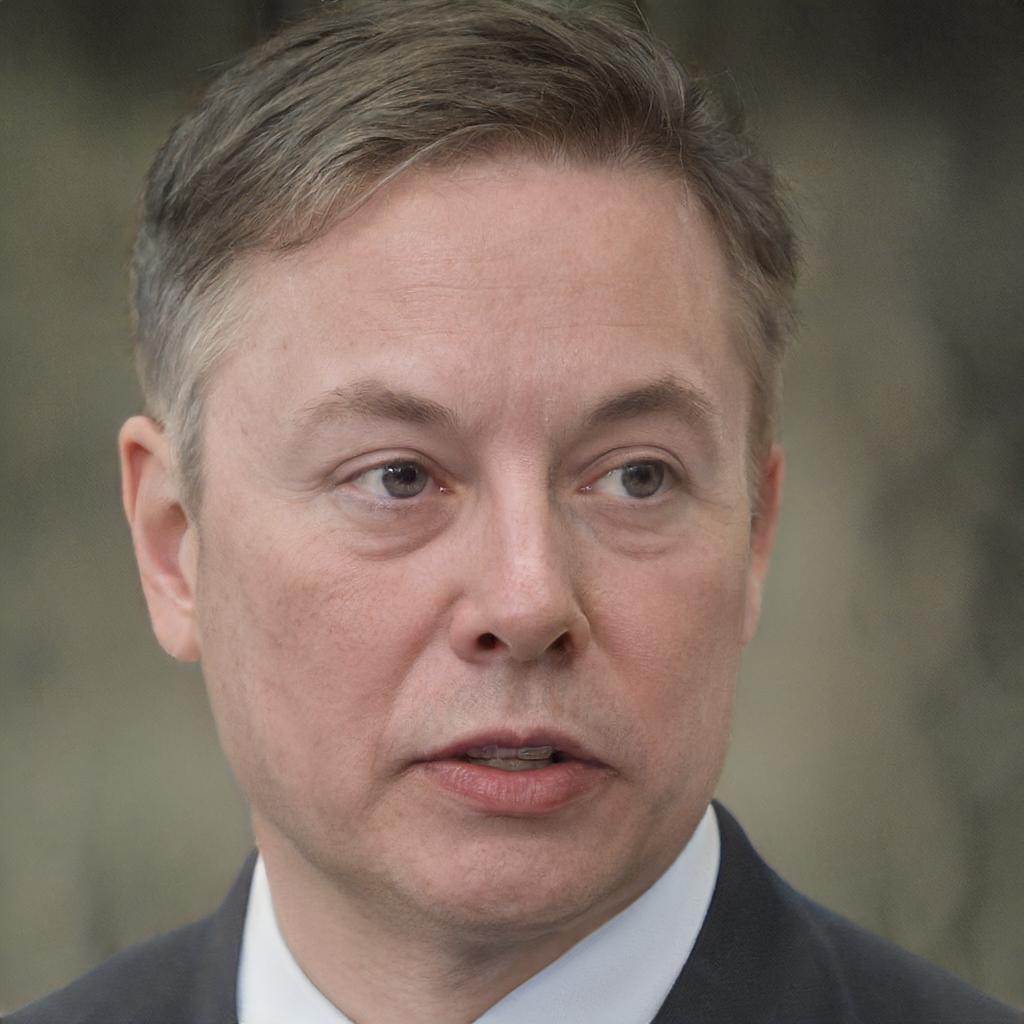} &
		\includegraphics[width=0.18\columnwidth]{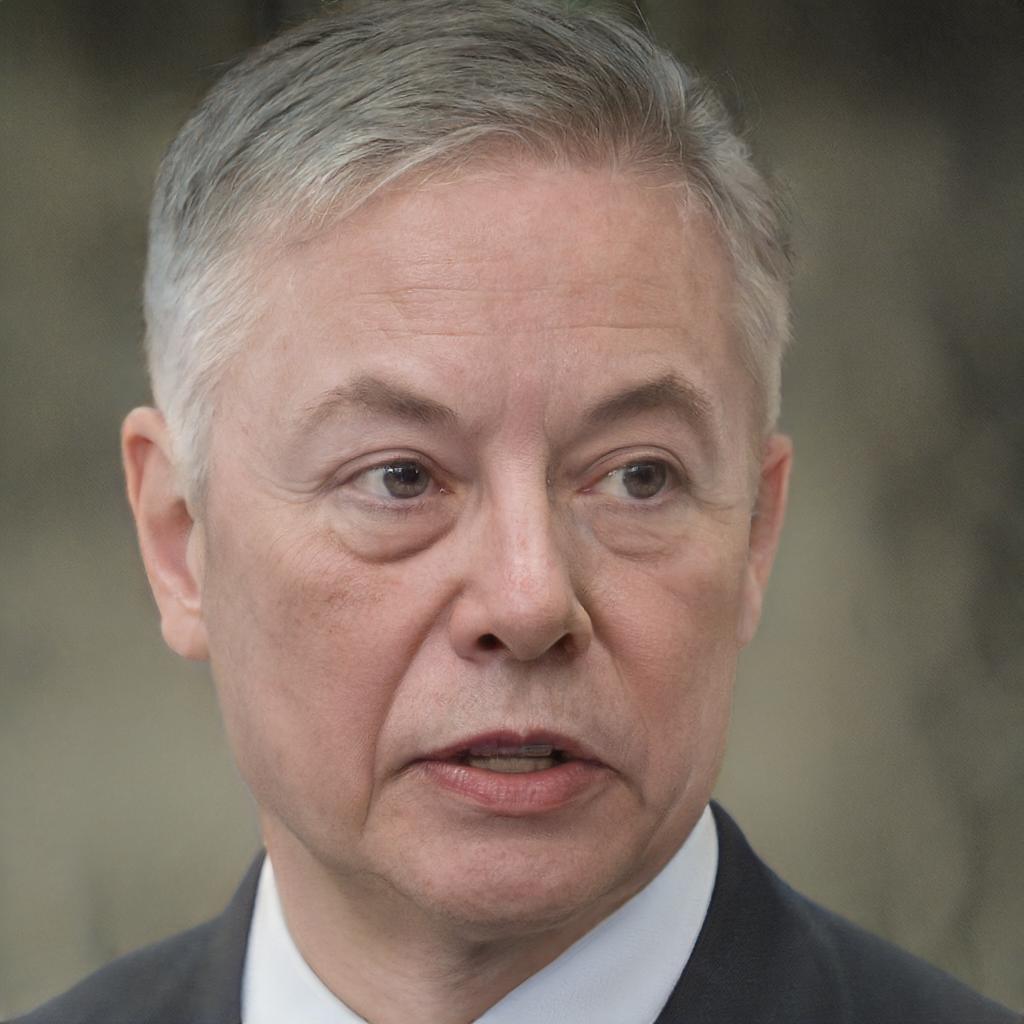} 
		\\
		\rotatebox{90}{\footnotesize \phantom{k}$\beta=0.11$} &
		\includegraphics[width=0.18\columnwidth]{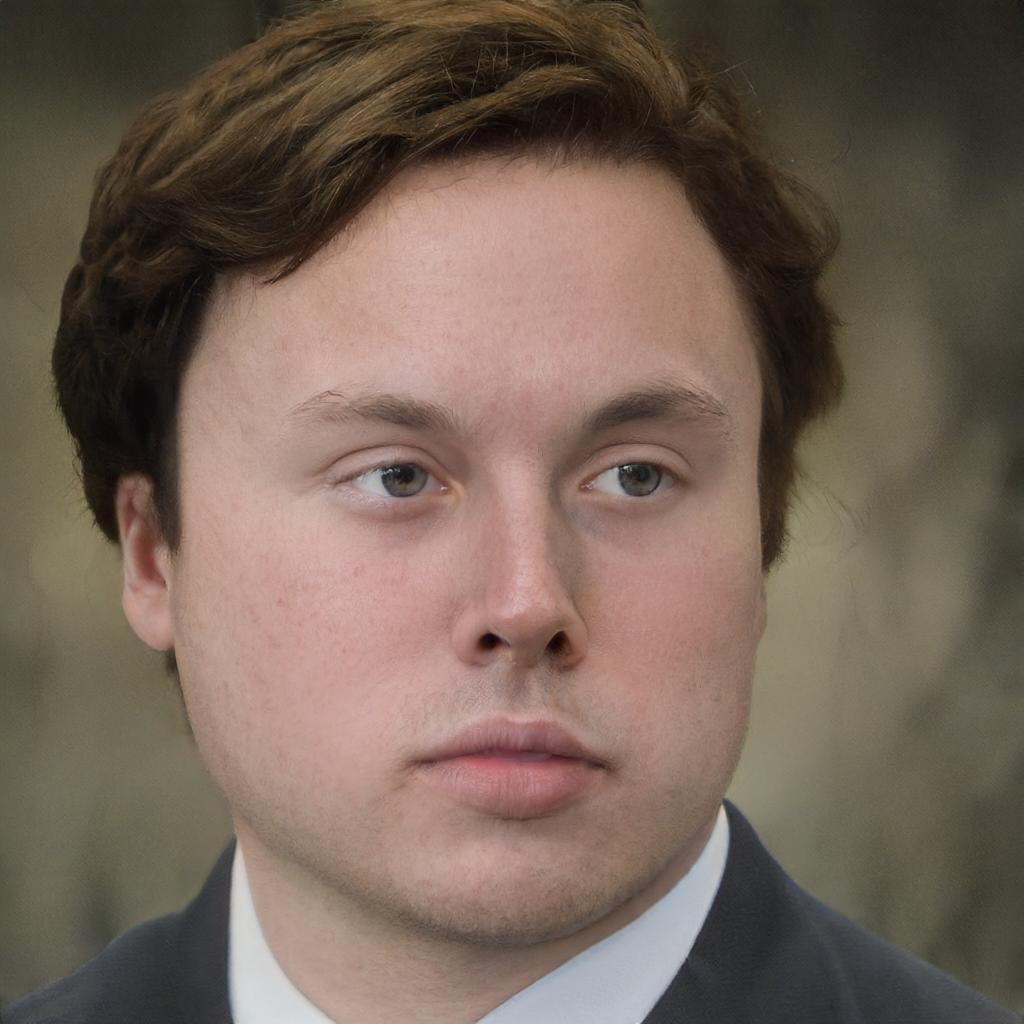} &
		\includegraphics[width=0.18\columnwidth]{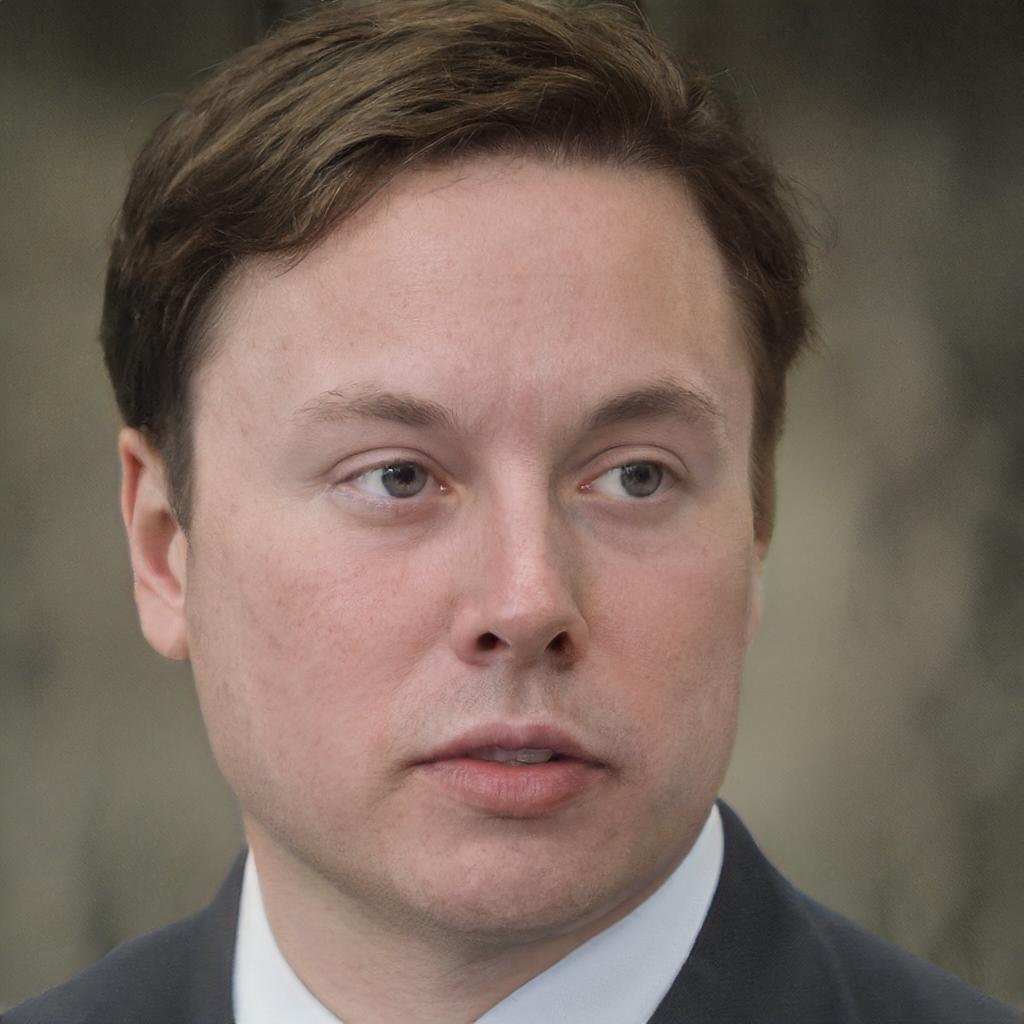} &
		\includegraphics[width=0.18\columnwidth]{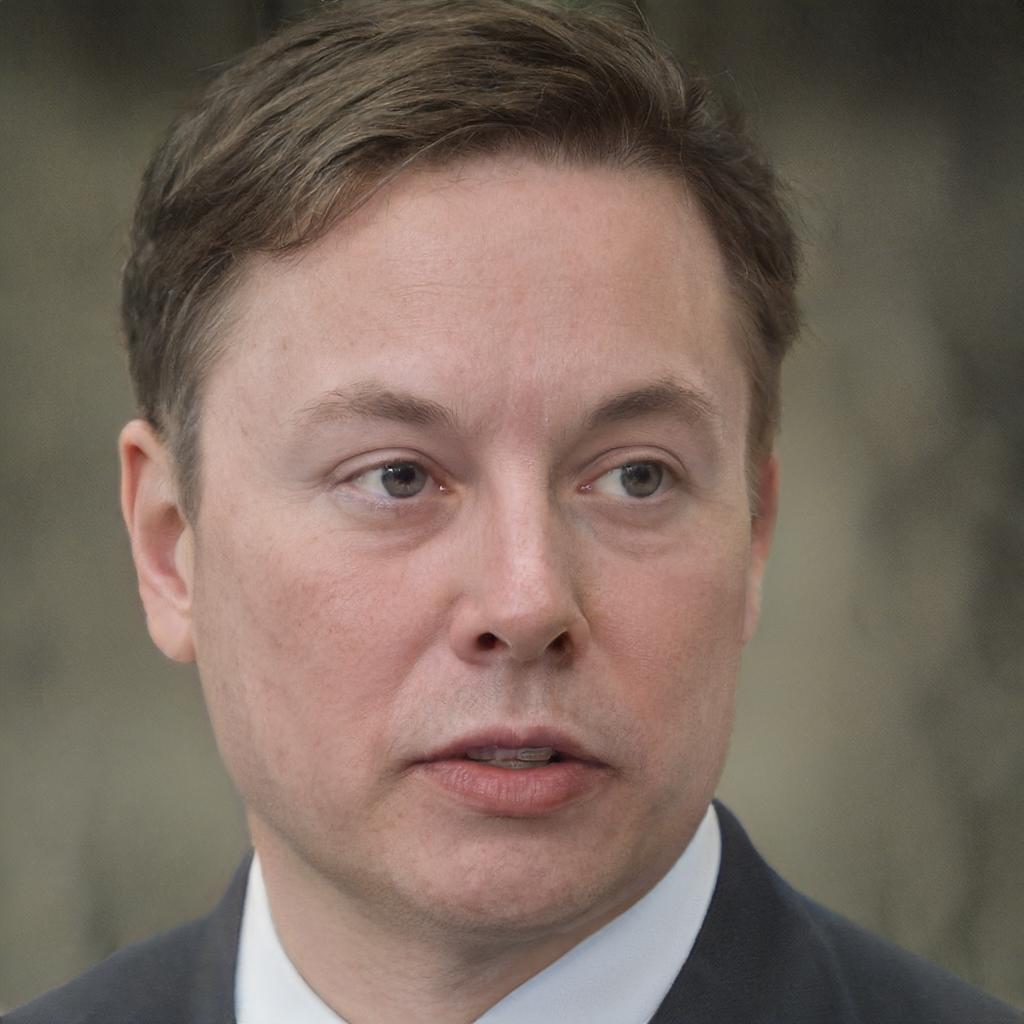} &
		\includegraphics[width=0.18\columnwidth]{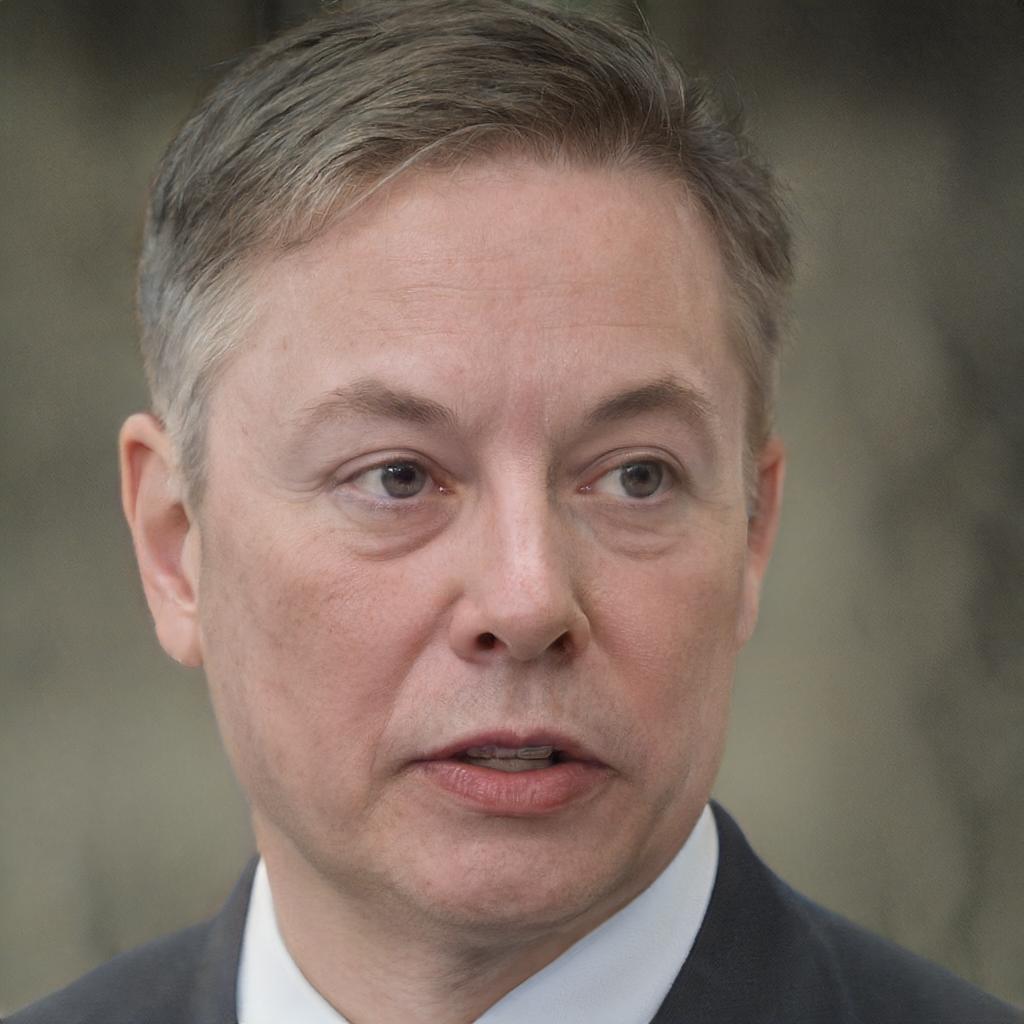} &
		\includegraphics[width=0.18\columnwidth]{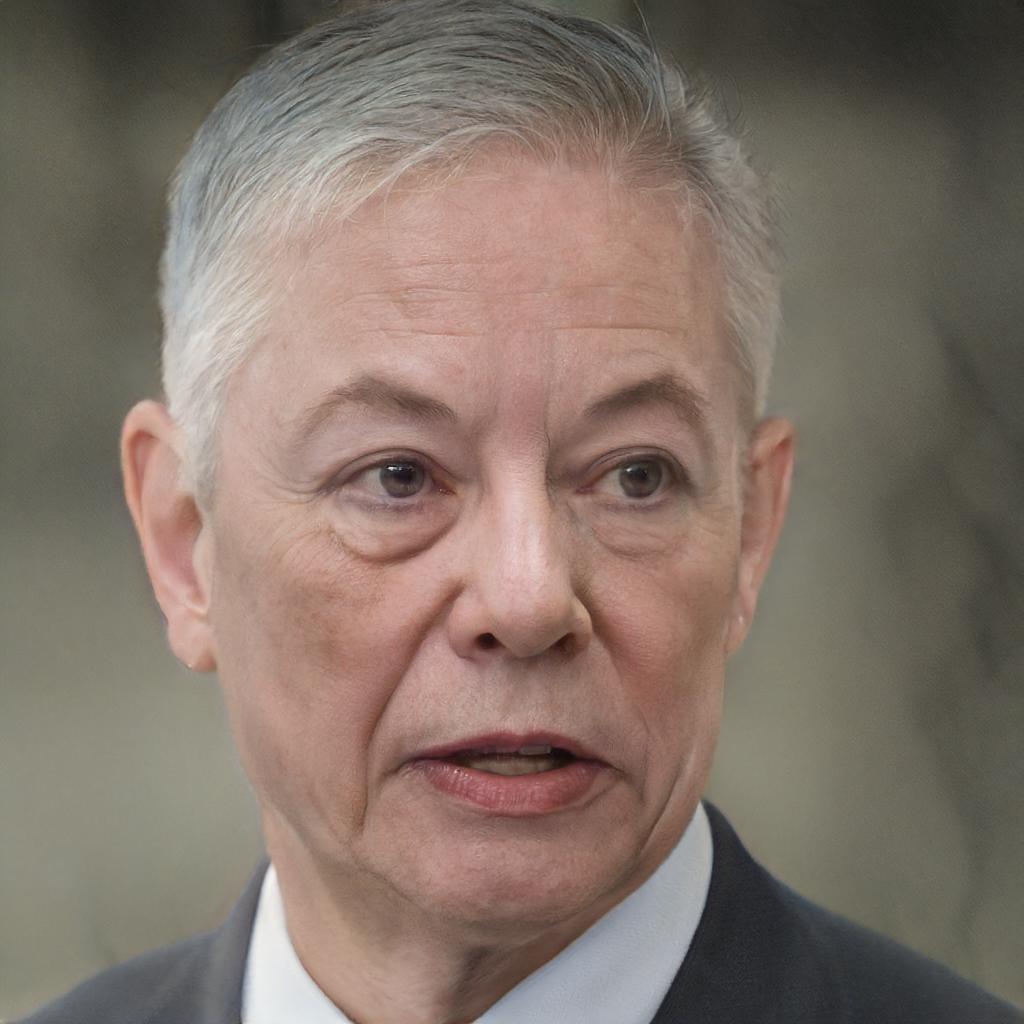} 
		\\
	\end{tabular}
	
	\caption{Image manipulation driven by the prompt ``grey hair'' for different manipulation strengths and disentanglement thresholds. Moving along the $\Ds$ direction, causes the hair color to become more grey, while steps in the $-\Ds$ direction yields darker hair. The effect becomes stronger as the strength $\alpha$ increases. When the disentanglement threshold $\beta$ is high, only the hair color is affected, and as $\beta$ is lowered, additional correlated attributes, such as wrinkles and the shape of the face are affected as well. 
	}
	\label{fig:disentanglement_strength}
\end{figure}

Having estimated the relevance $R_c$ of each channel, we ignore channels whose $R_c$ falls below a threshold $\beta$. This parameter may be used to control the degree of disentanglement in the manipulation: using higher threshold values results in more disentangled manipulations, but at the same time the visual effect of the manipulation is reduced.
Since various high-level attributes, such as age, involve a combination of several lower level attributes (for example, grey hair, wrinkles, and skin color), multiple channels are relevant, and in such cases lowering
the threshold value may be preferable, as demonstrated in Figure~\ref{fig:disentanglement_strength}.
To our knowledge, the ability to control the degree of disentanglement in this manner is unique to our approach.

In summary, given a target direction $\Di$ in CLIP space, we set
\begin{equation}
	\label{eq:direction}
	\Ds = \left\{ 
	\begin{array}{ccl}
		\Di_c \cdot \Di & \quad & \mbox{if } |\Di_c \cdot \Di| \geq \beta \\
		0 & \quad & \mbox{otherwise}
	\end{array} 	
	\right.
\end{equation}

\begin{figure*}[tb]
	\centering
	\setlength{\tabcolsep}{1.1pt}
	
	{\footnotesize
		\begin{tabular}{ccccccccc}
			Input & Pale & Tanned & Makeup & Curly Hair &  Straight Hair & Bob Cut & Hi-top Fade & Fringe Hair \\
			\includegraphics[width=0.10\textwidth]{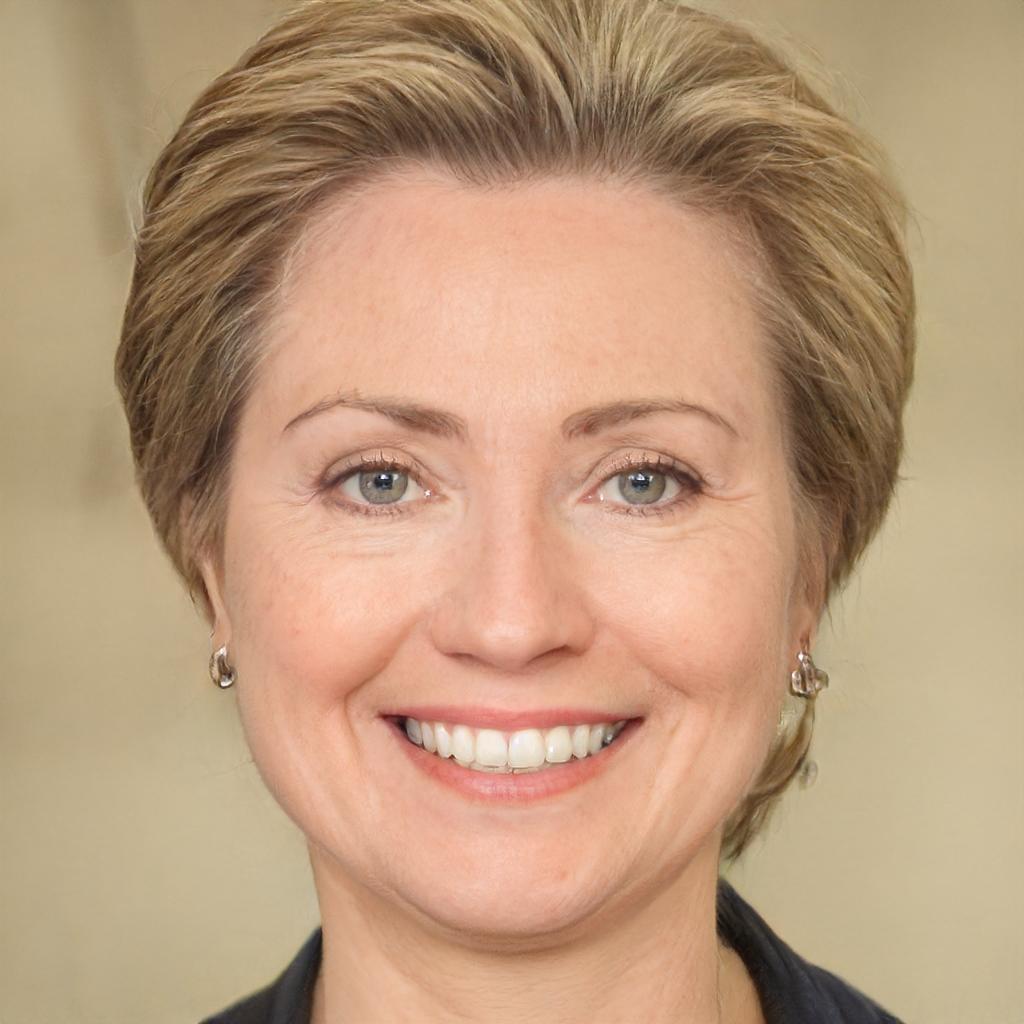} &
			\includegraphics[width=0.10\textwidth]{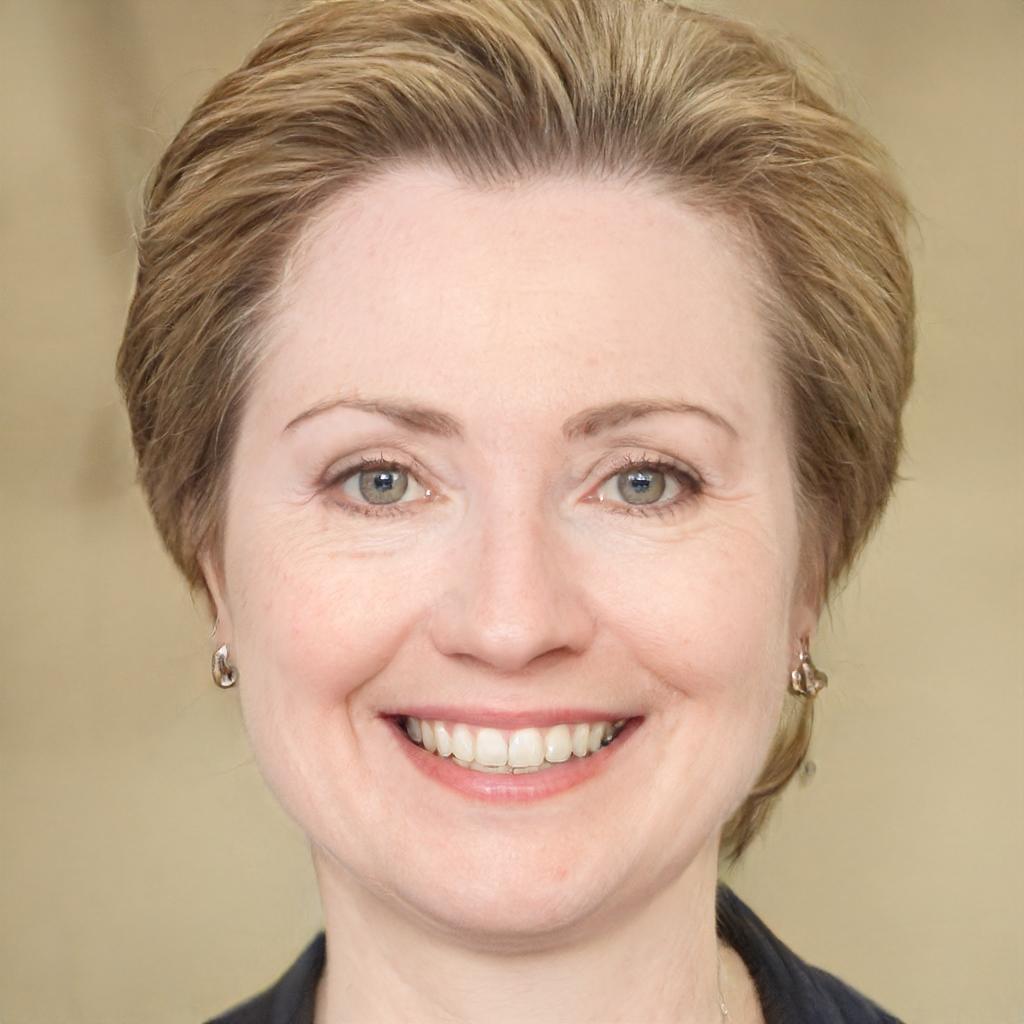} &
			\includegraphics[width=0.10\textwidth]{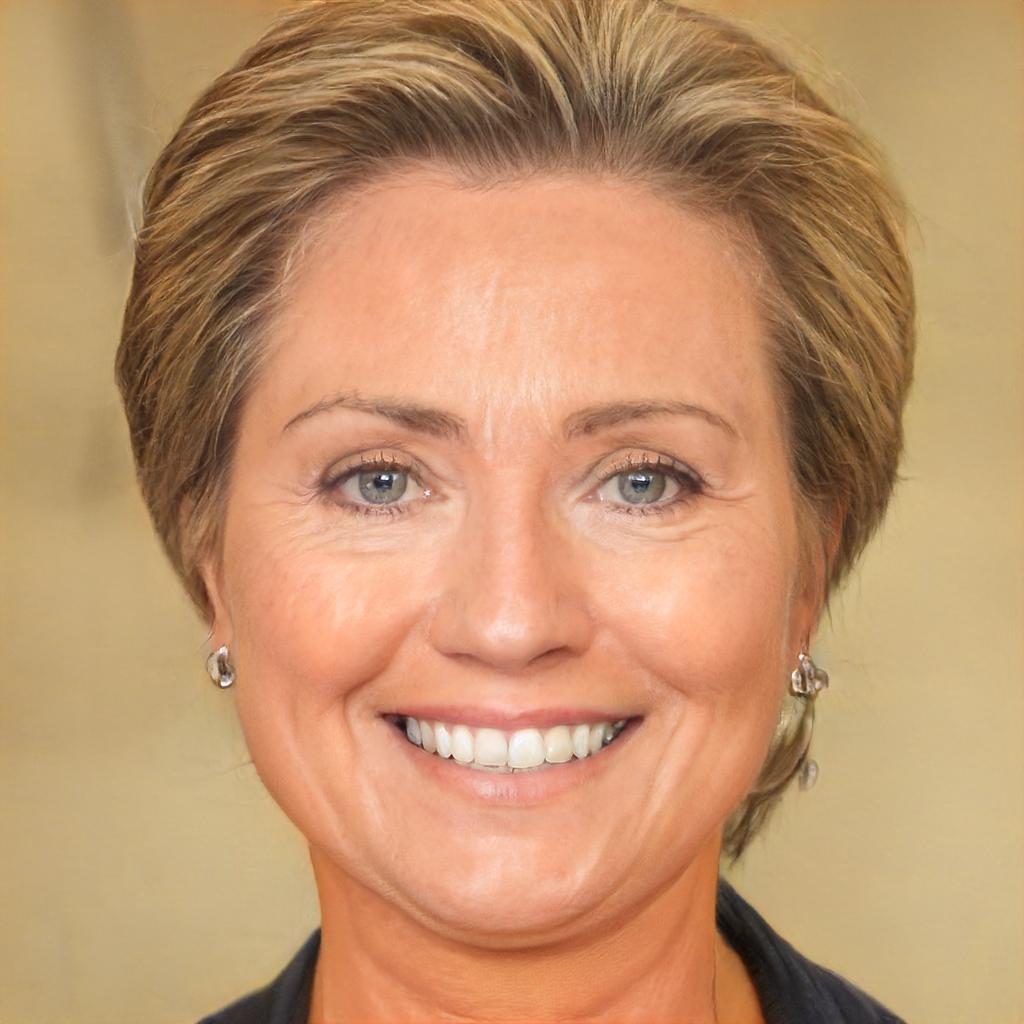} &
			\includegraphics[width=0.10\textwidth]{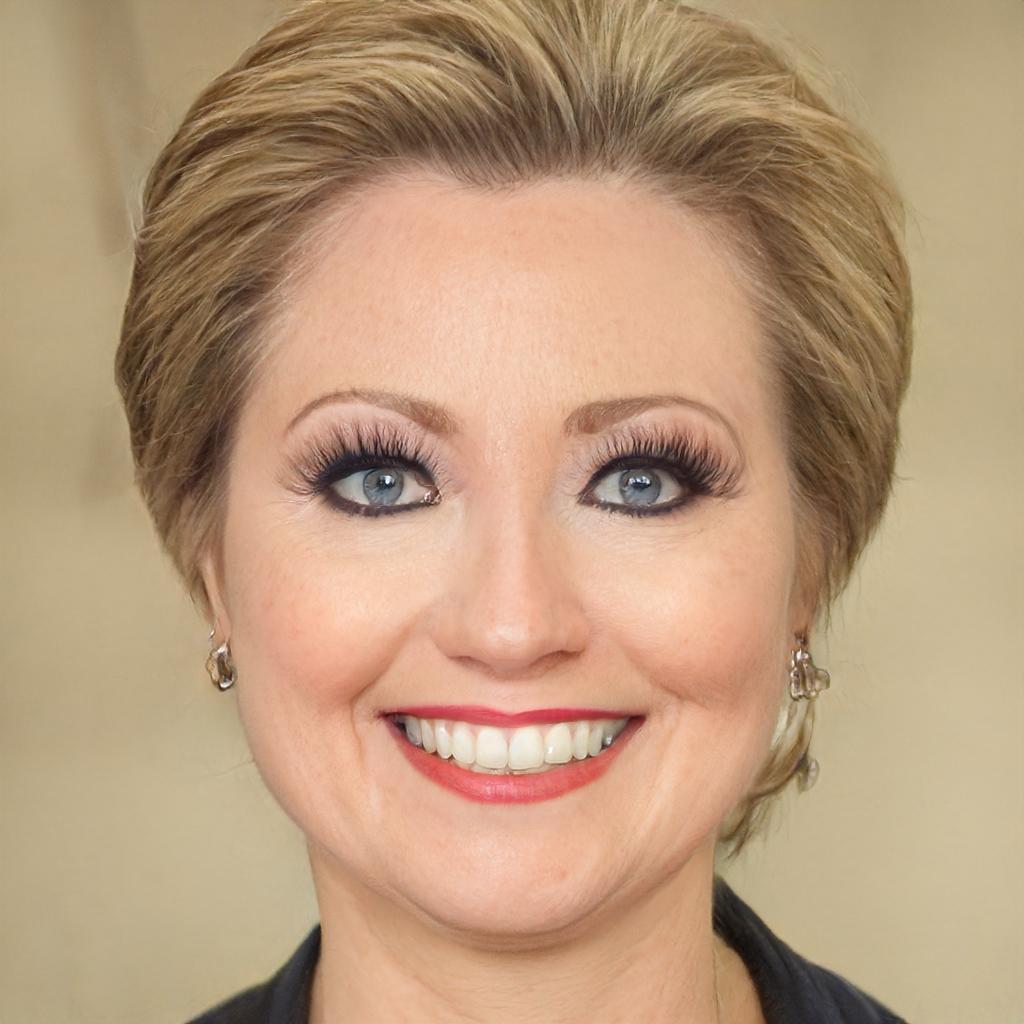} &
			\includegraphics[width=0.10\textwidth]{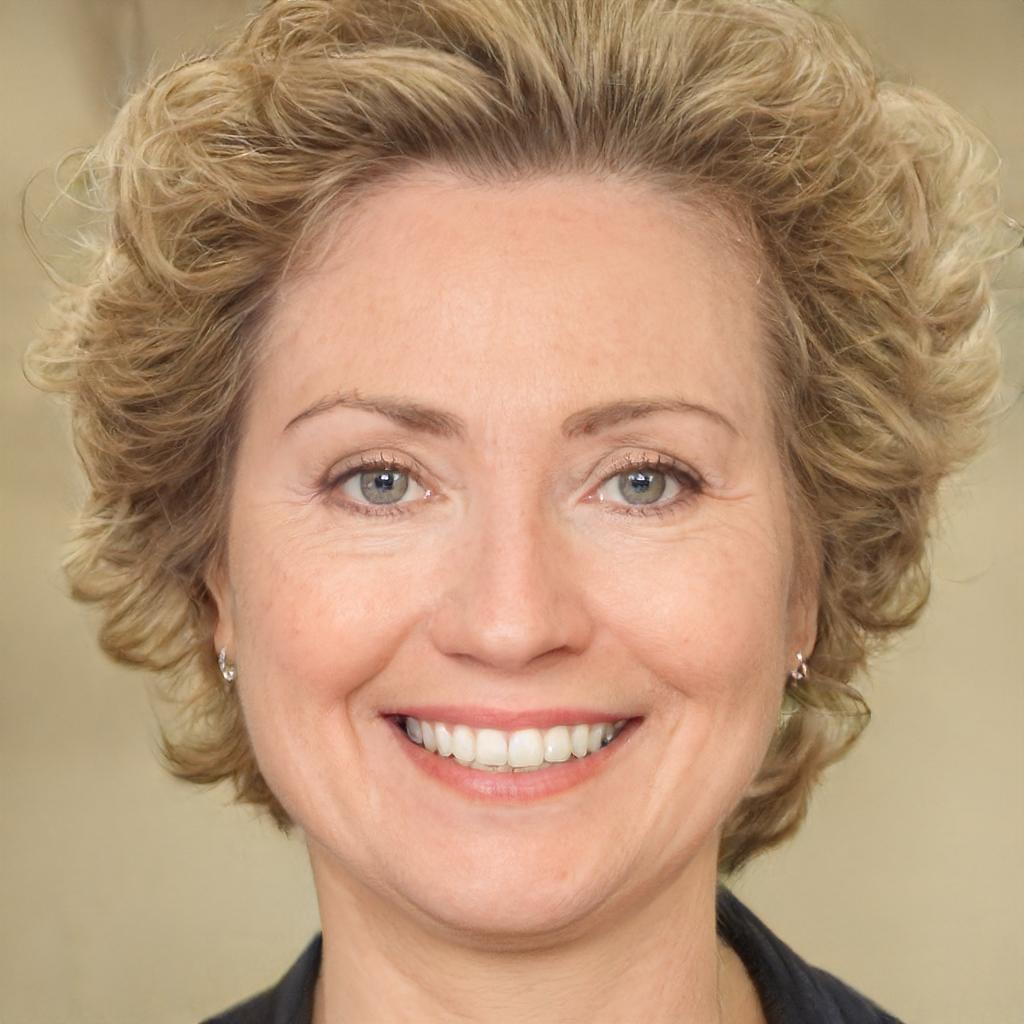} &
			\includegraphics[width=0.10\textwidth]{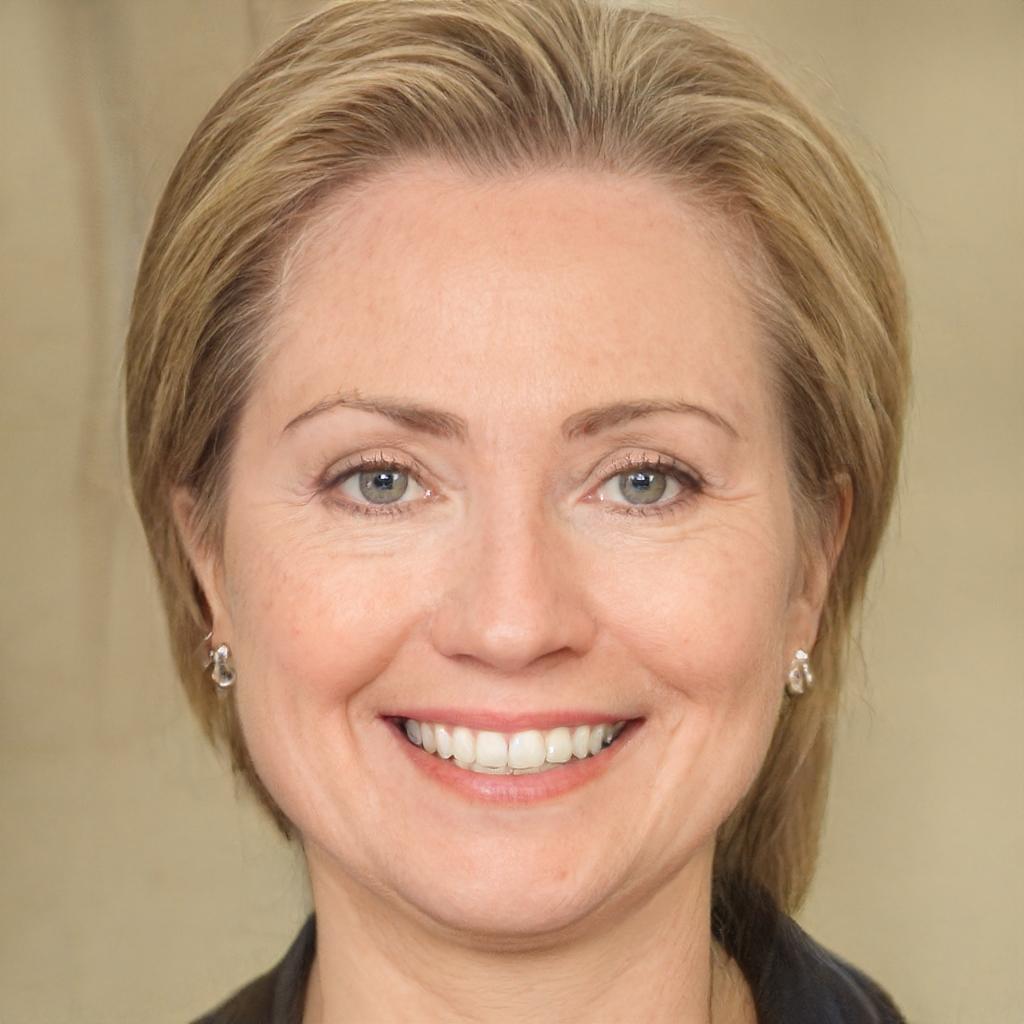} &
			\includegraphics[width=0.10\textwidth]{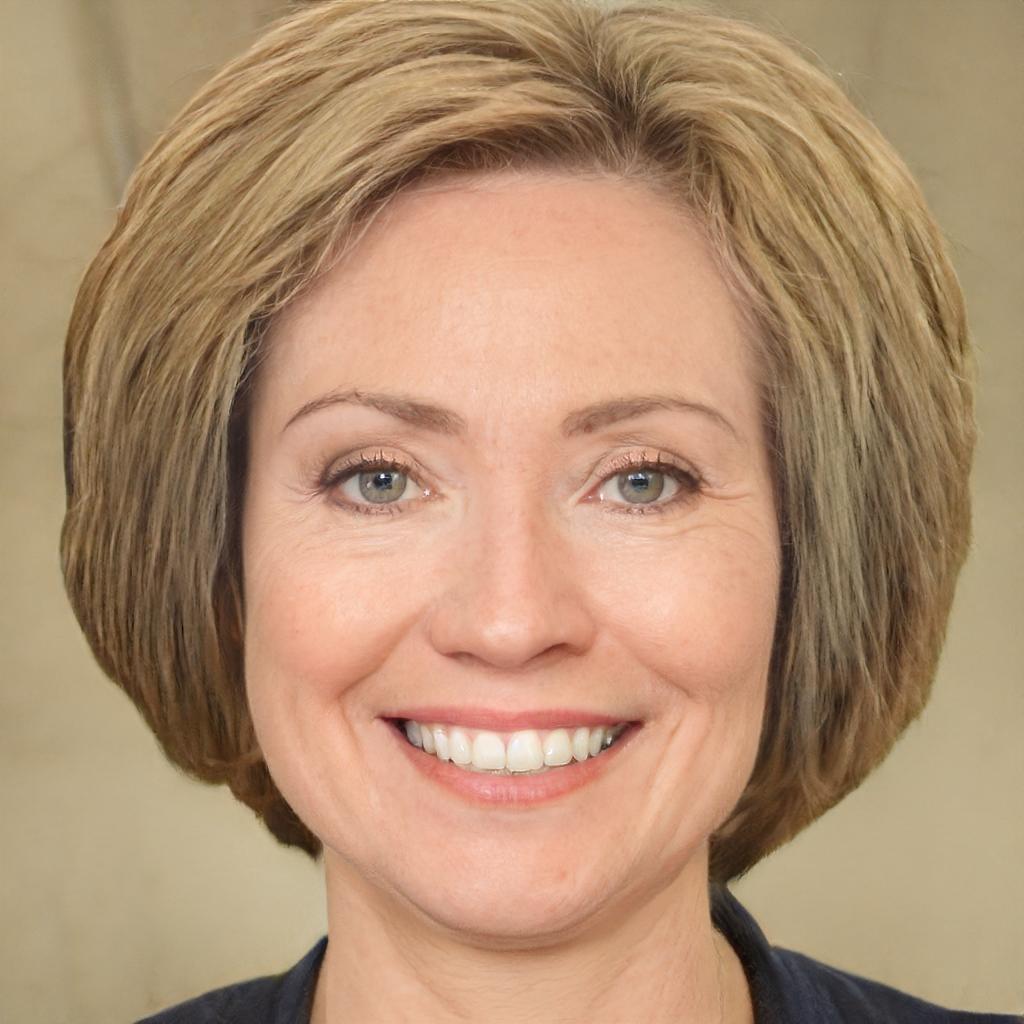} &
			\includegraphics[width=0.10\textwidth]{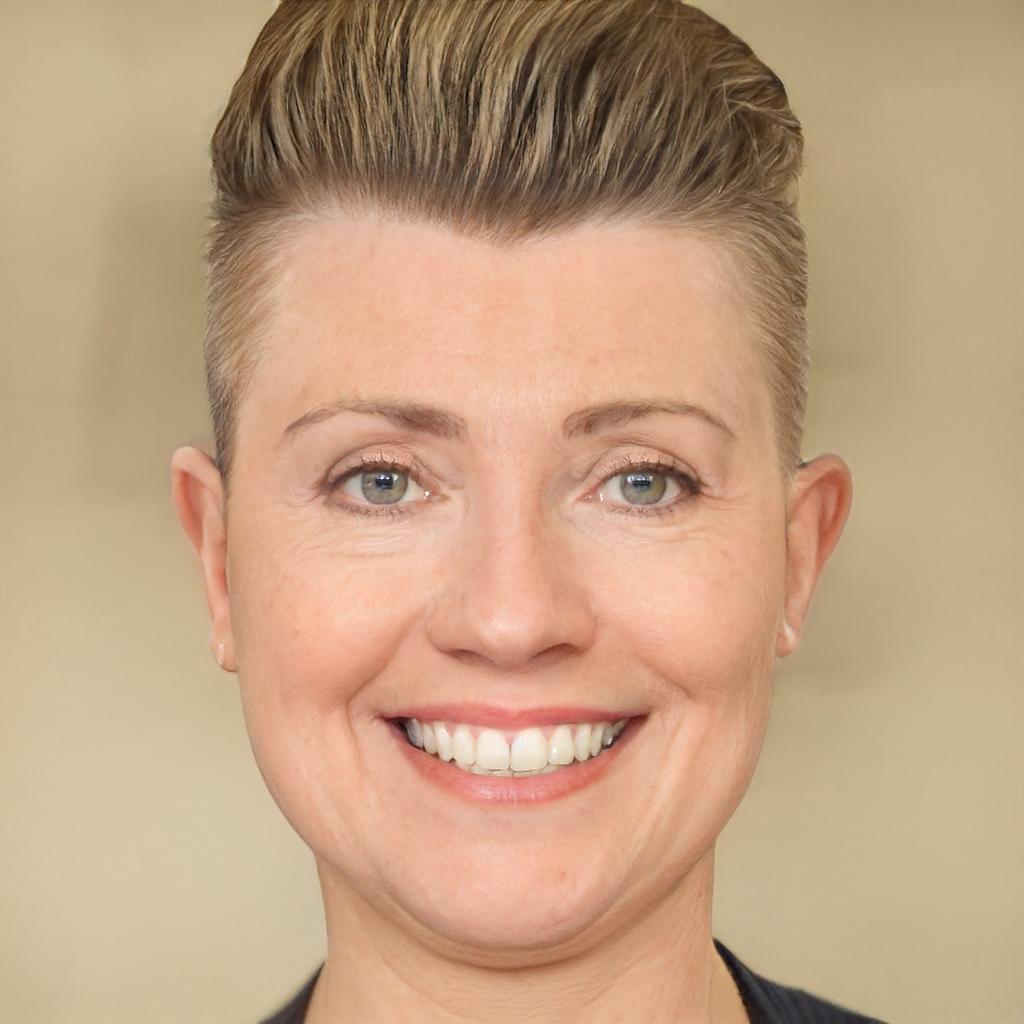} &
			\includegraphics[width=0.10\textwidth]{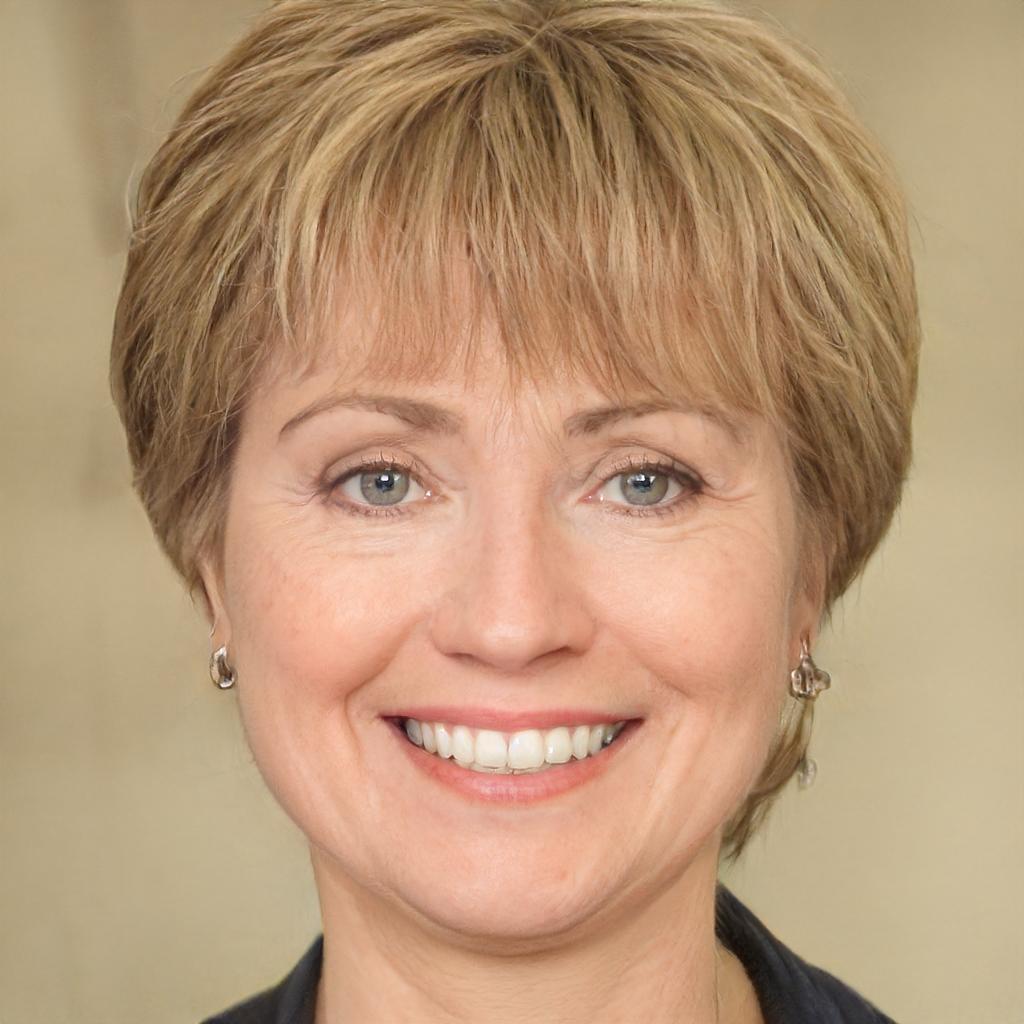}
			\\
			\includegraphics[width=0.10\textwidth]{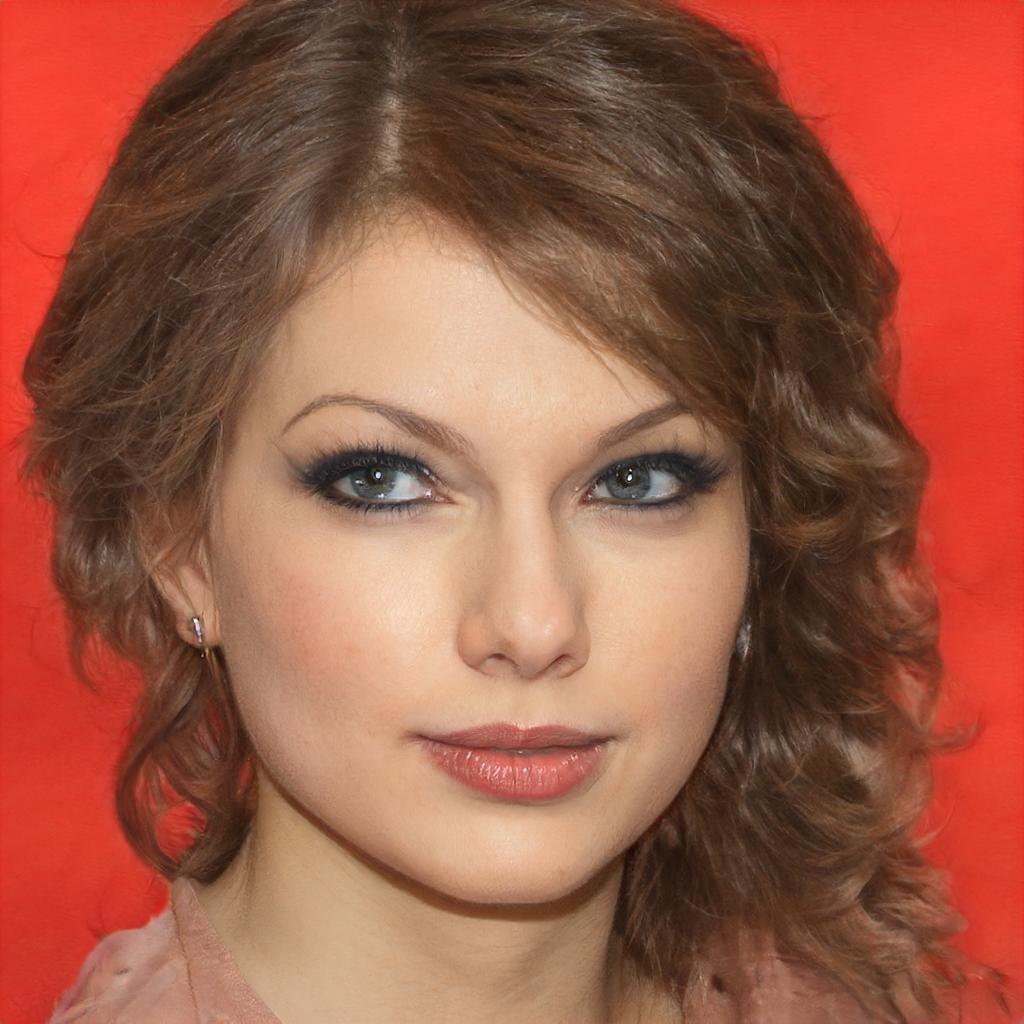} &
			\includegraphics[width=0.10\textwidth]{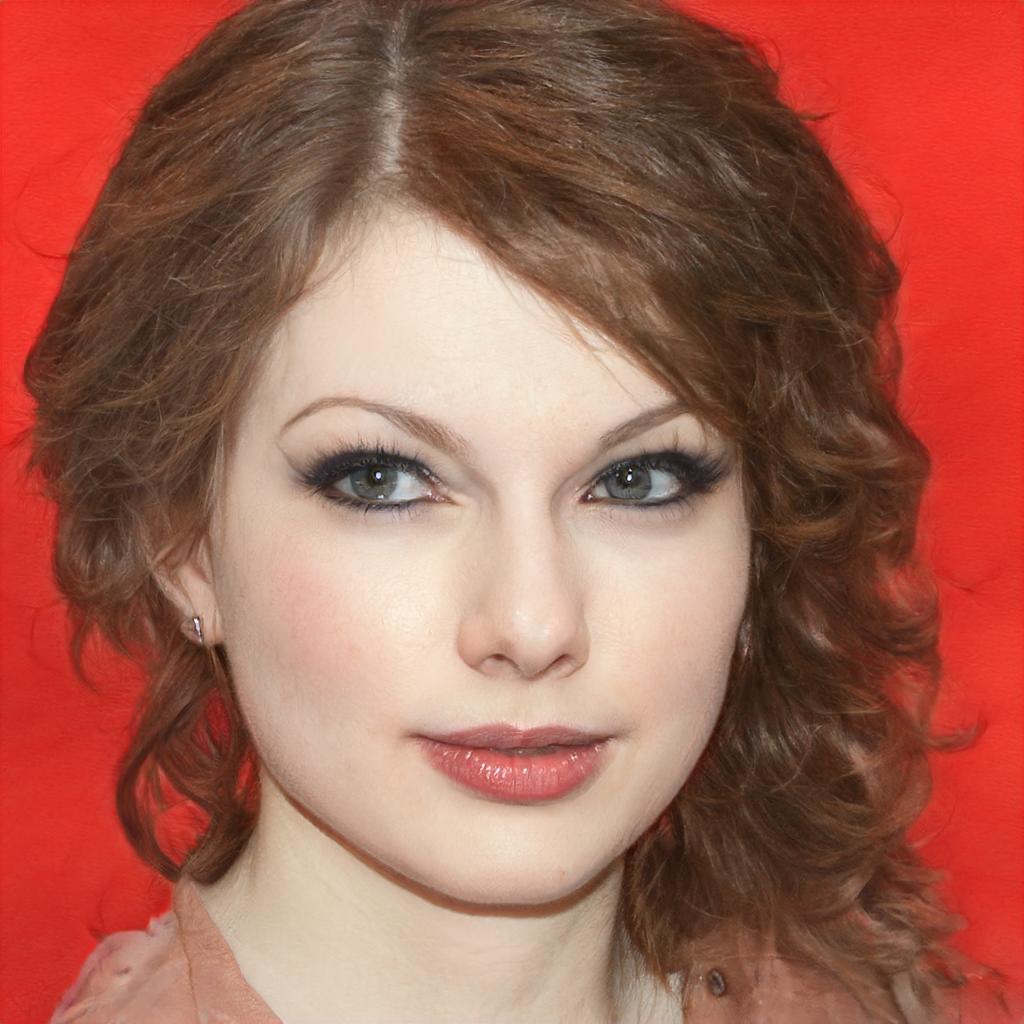} &
			\includegraphics[width=0.10\textwidth]{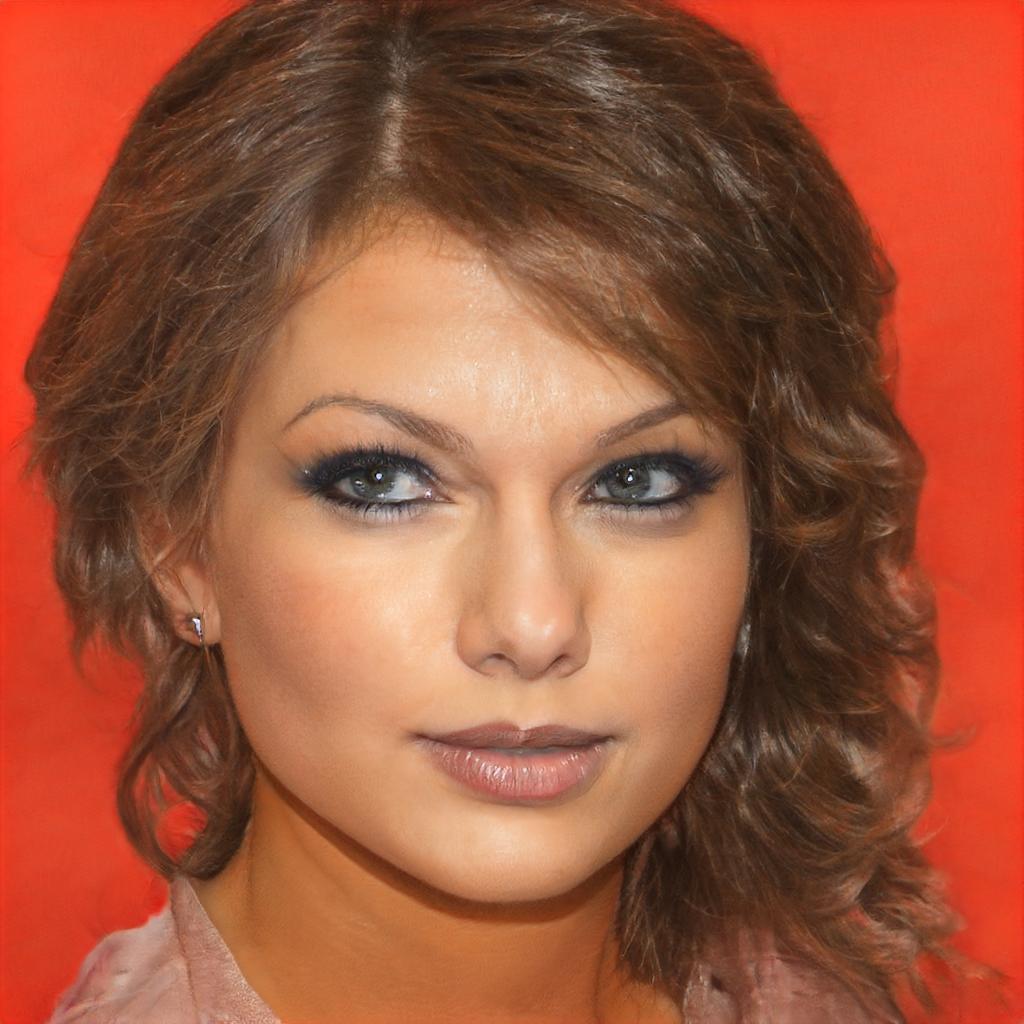} &
			\includegraphics[width=0.10\textwidth]{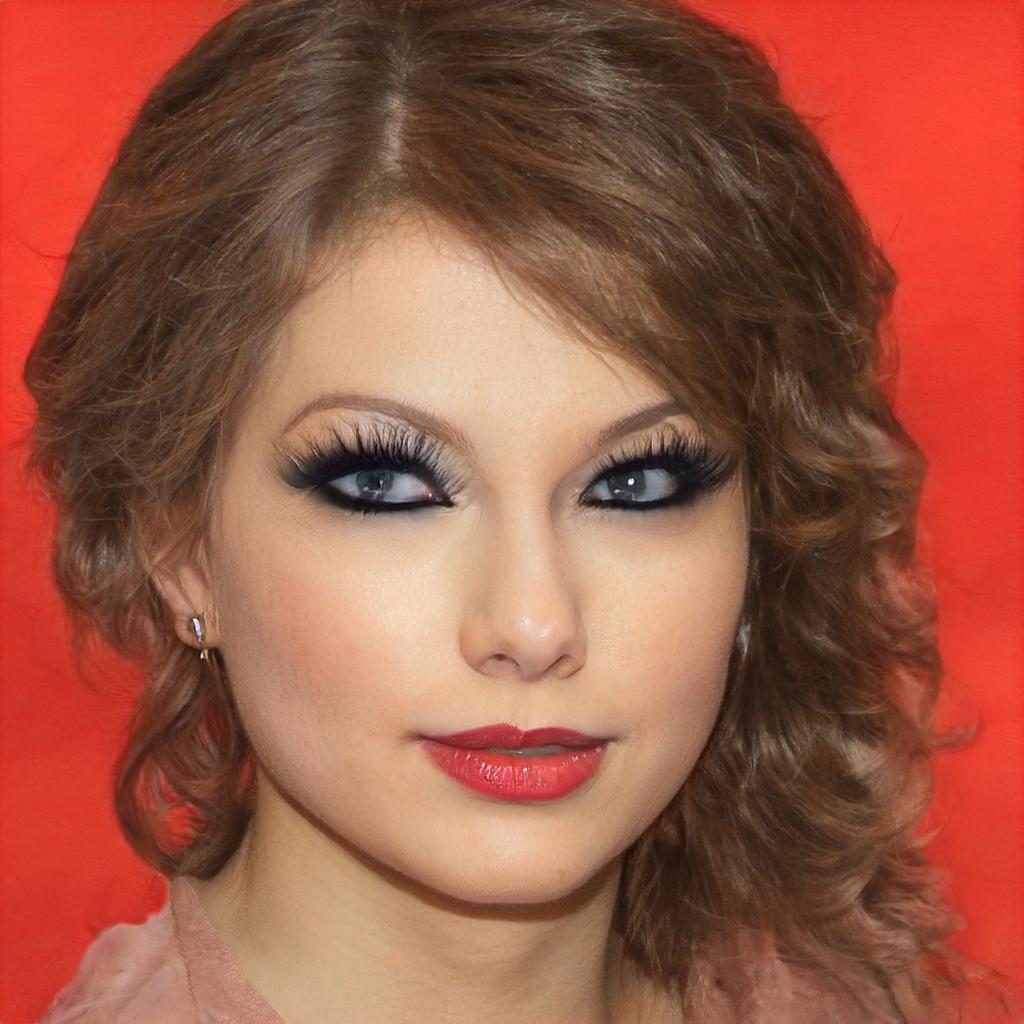} &
			\includegraphics[width=0.10\textwidth]{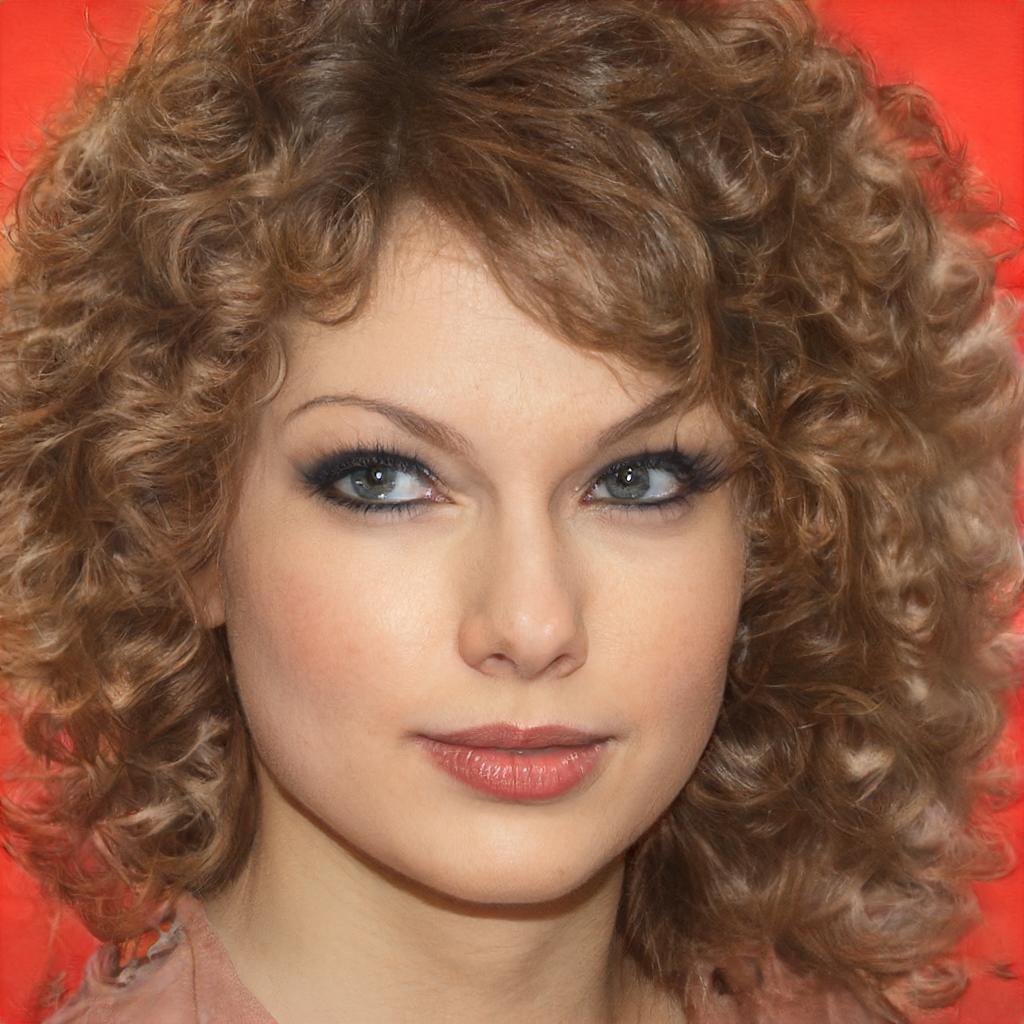} &
			\includegraphics[width=0.10\textwidth]{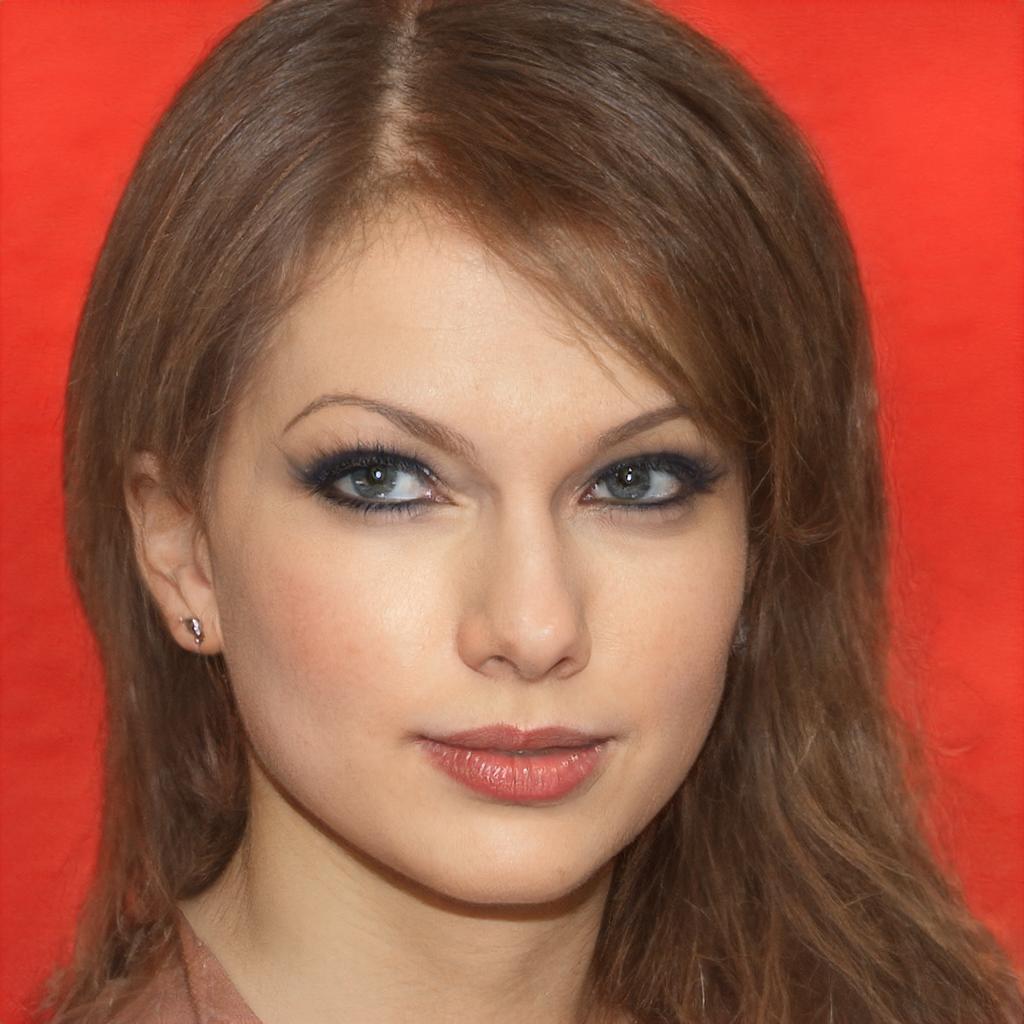} &
			\includegraphics[width=0.10\textwidth]{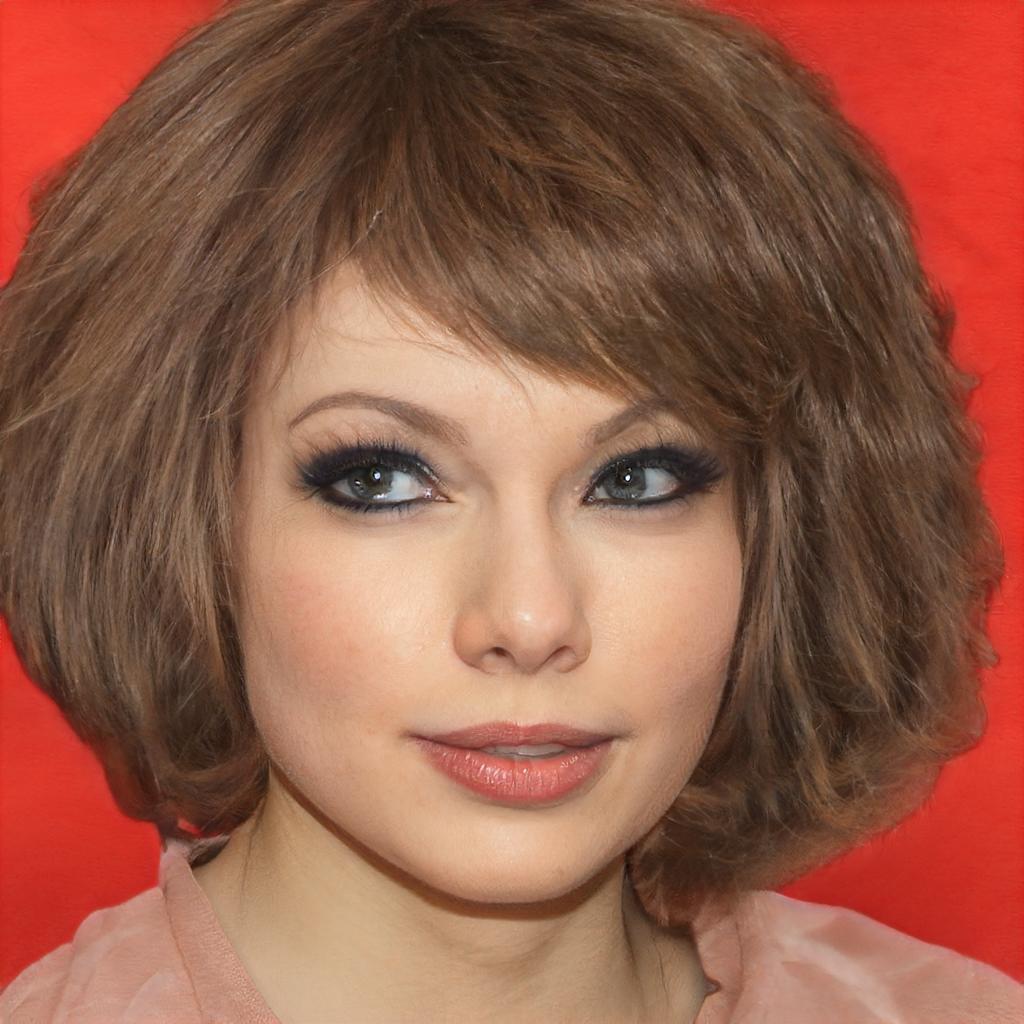} &
			\includegraphics[width=0.10\textwidth]{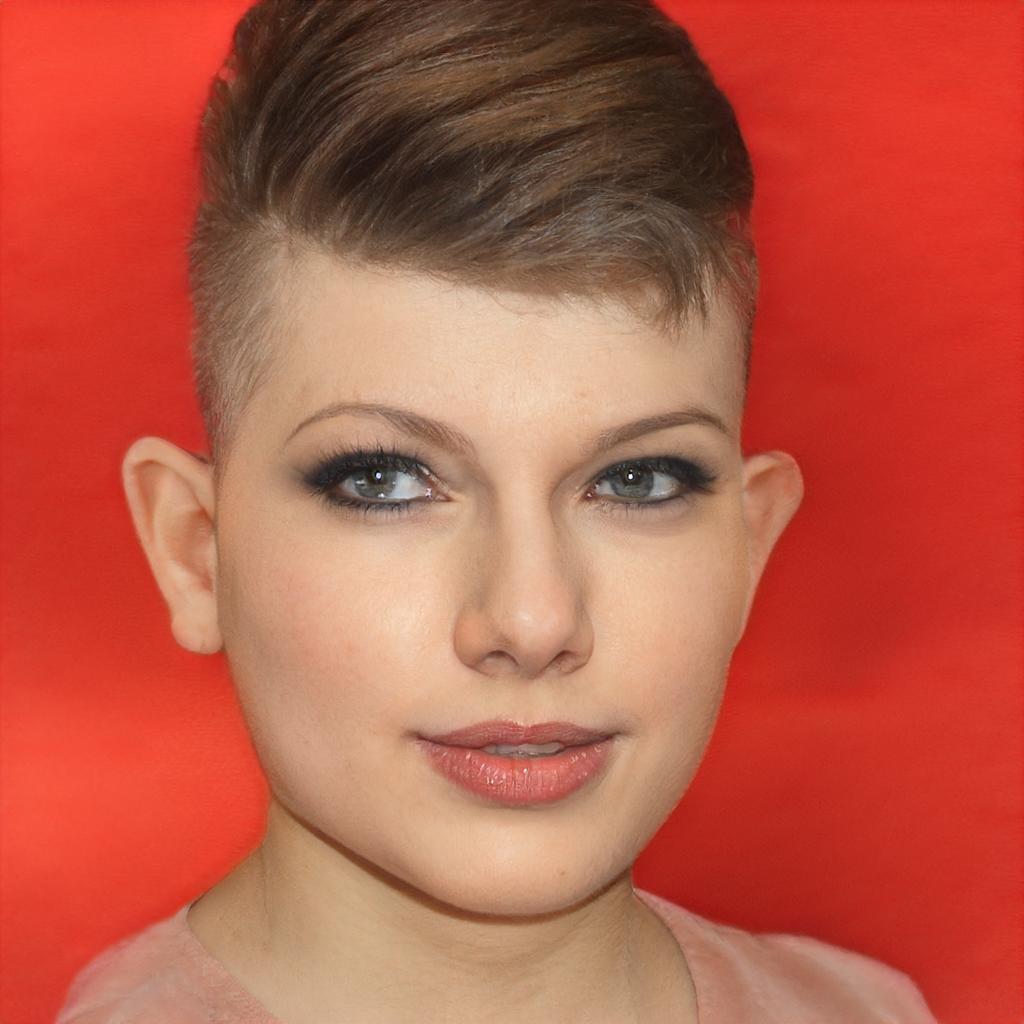} &
			\includegraphics[width=0.10\textwidth]{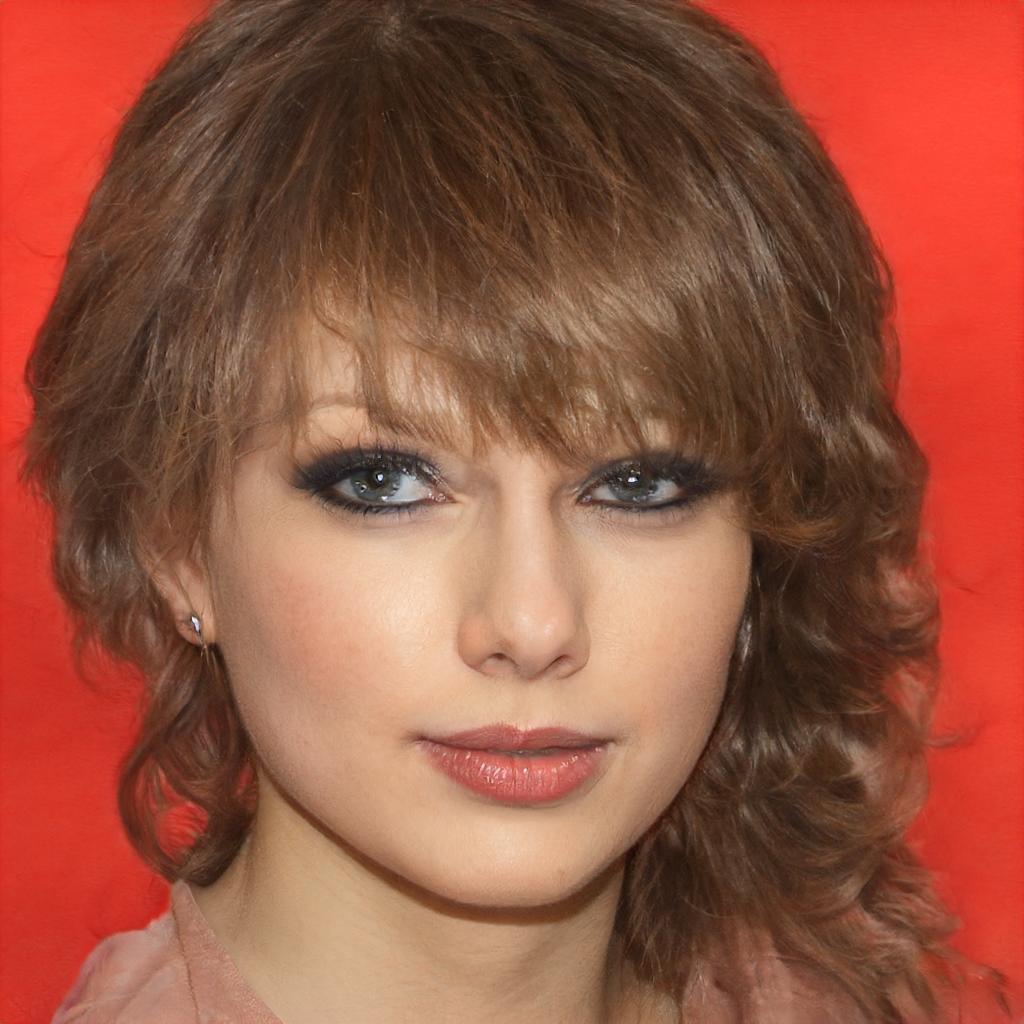}
			\\
			\includegraphics[width=0.10\textwidth]{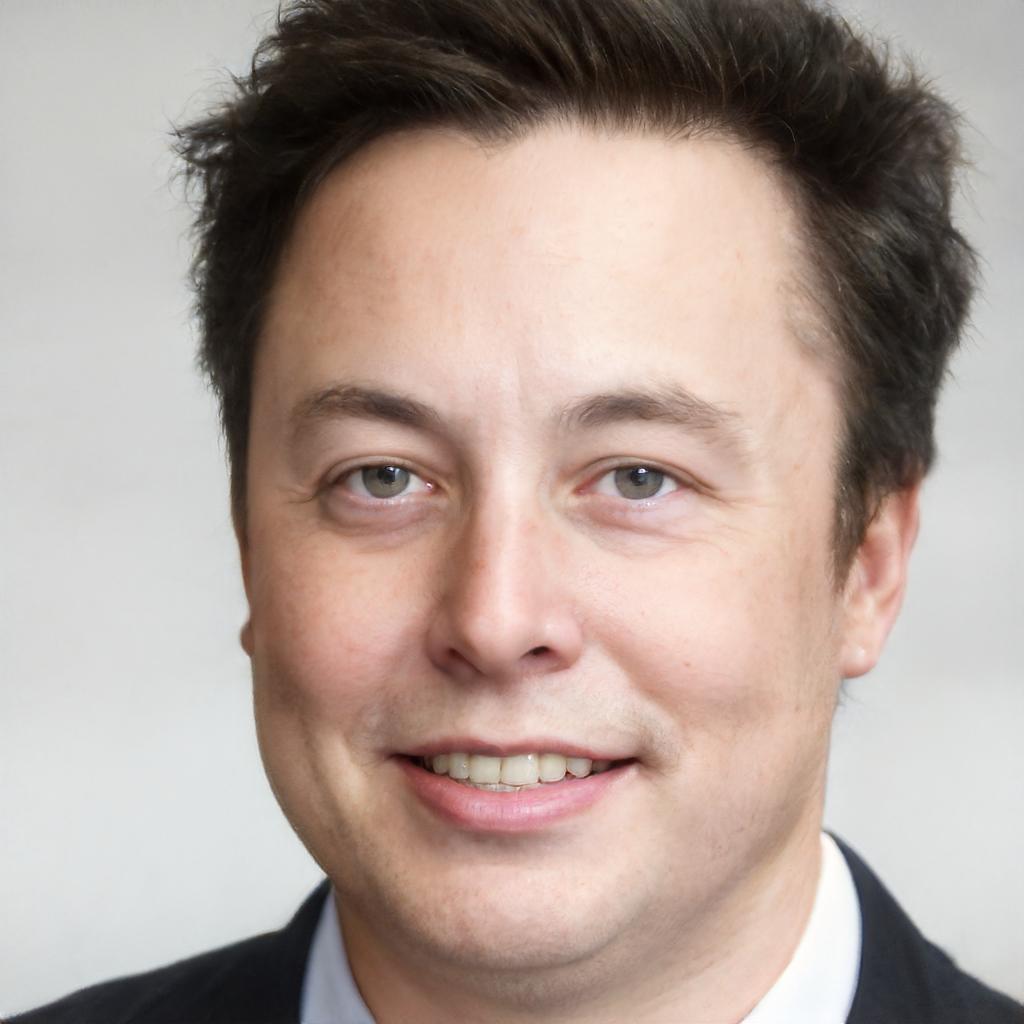} &
			\includegraphics[width=0.10\textwidth]{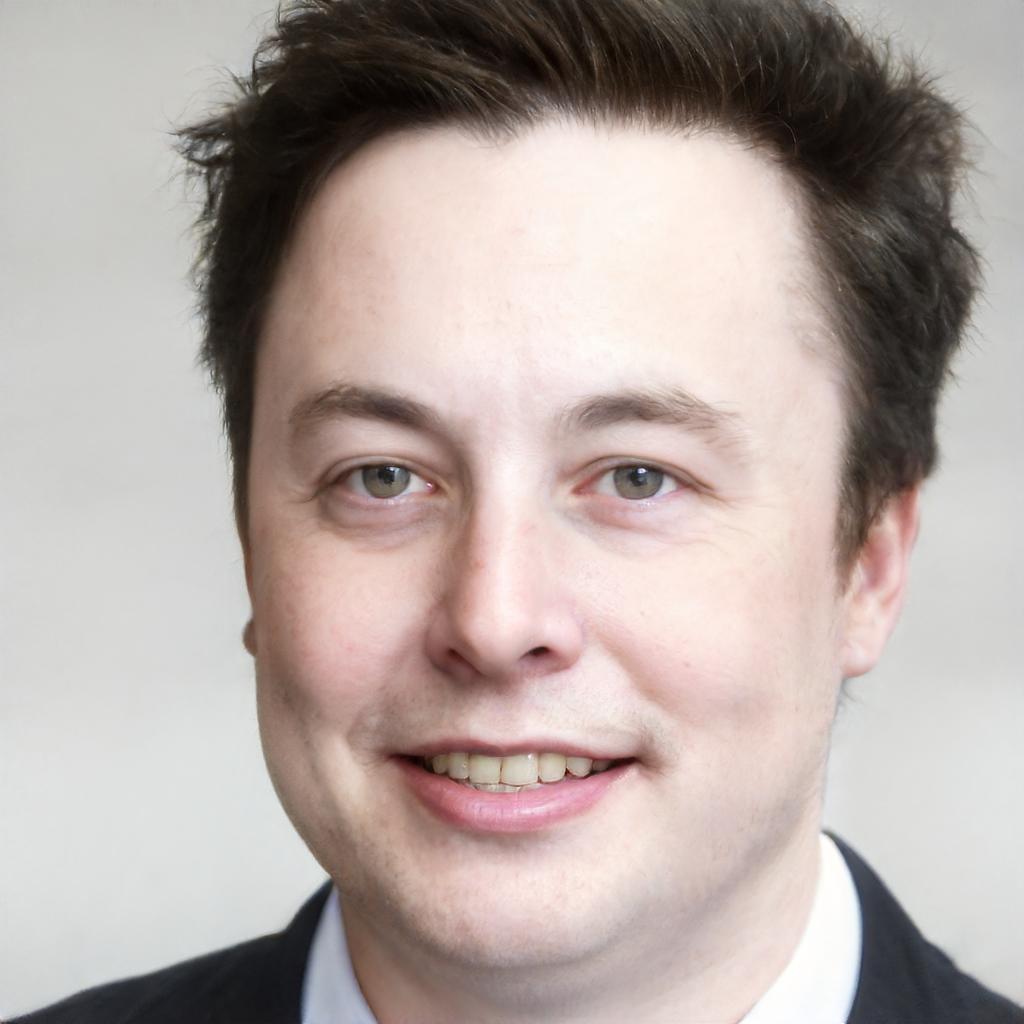} &
			\includegraphics[width=0.10\textwidth]{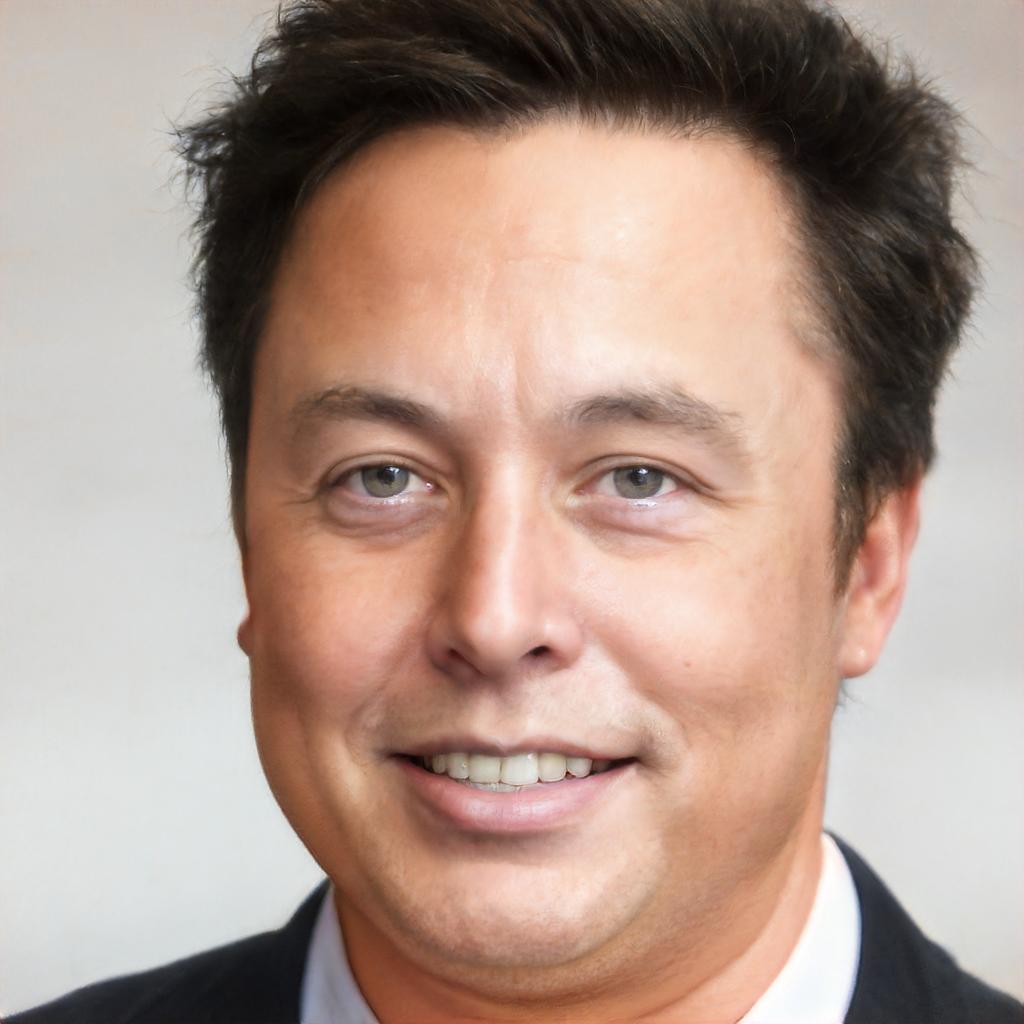} &
			\includegraphics[width=0.10\textwidth]{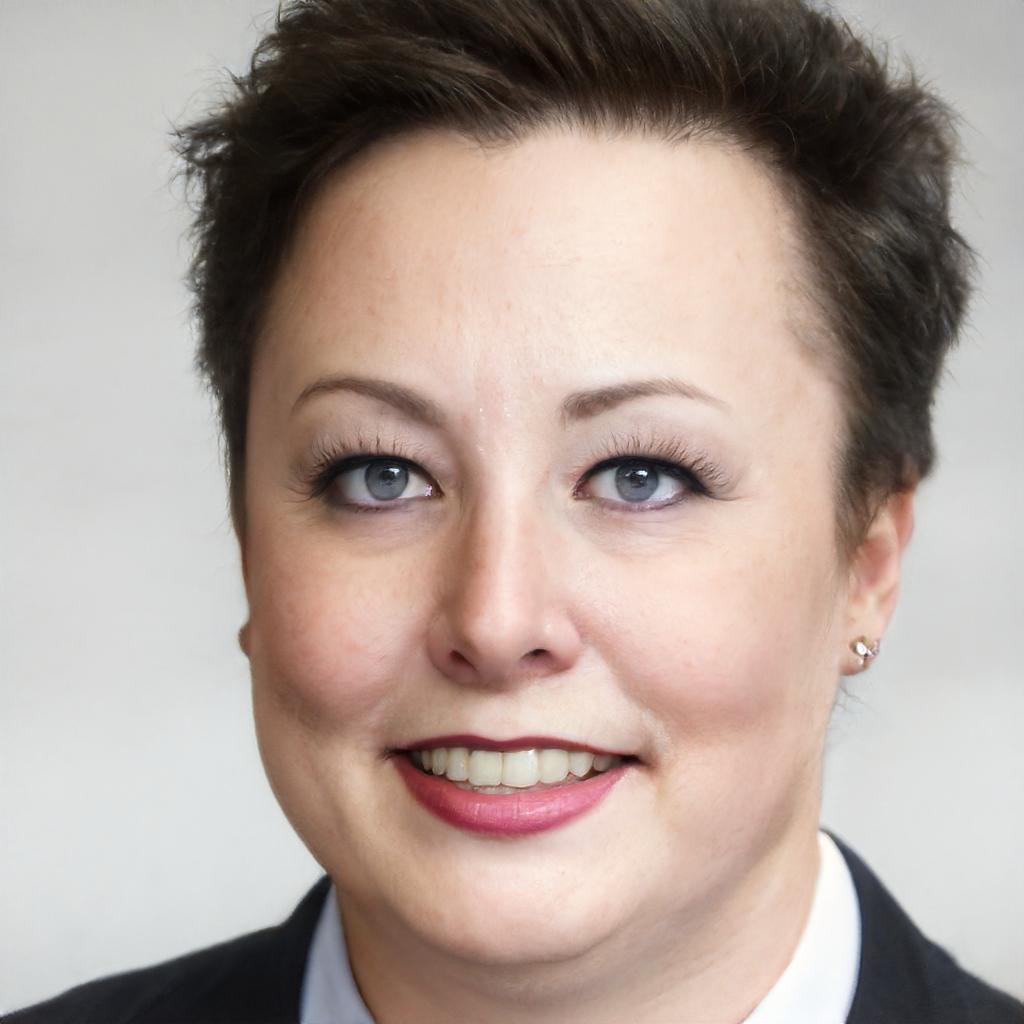} &
			\includegraphics[width=0.10\textwidth]{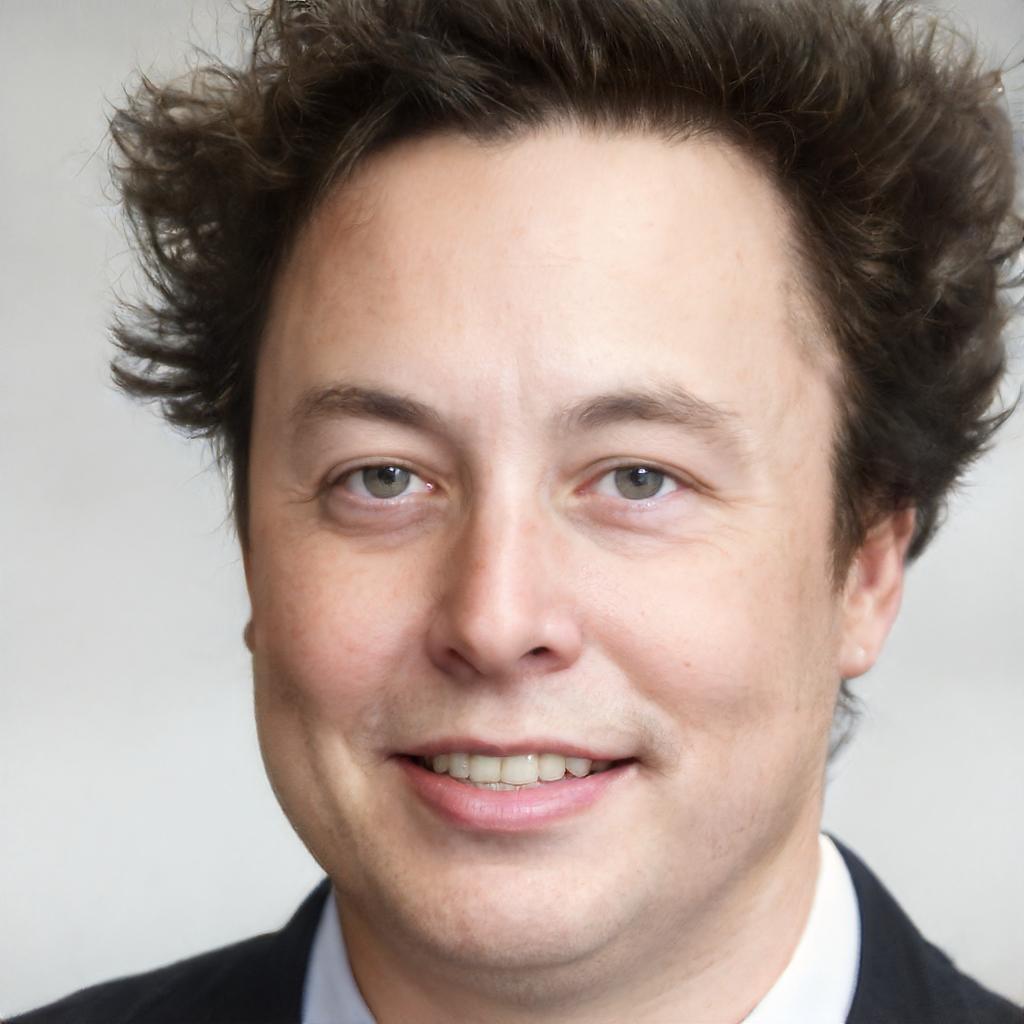} &
			\includegraphics[width=0.10\textwidth]{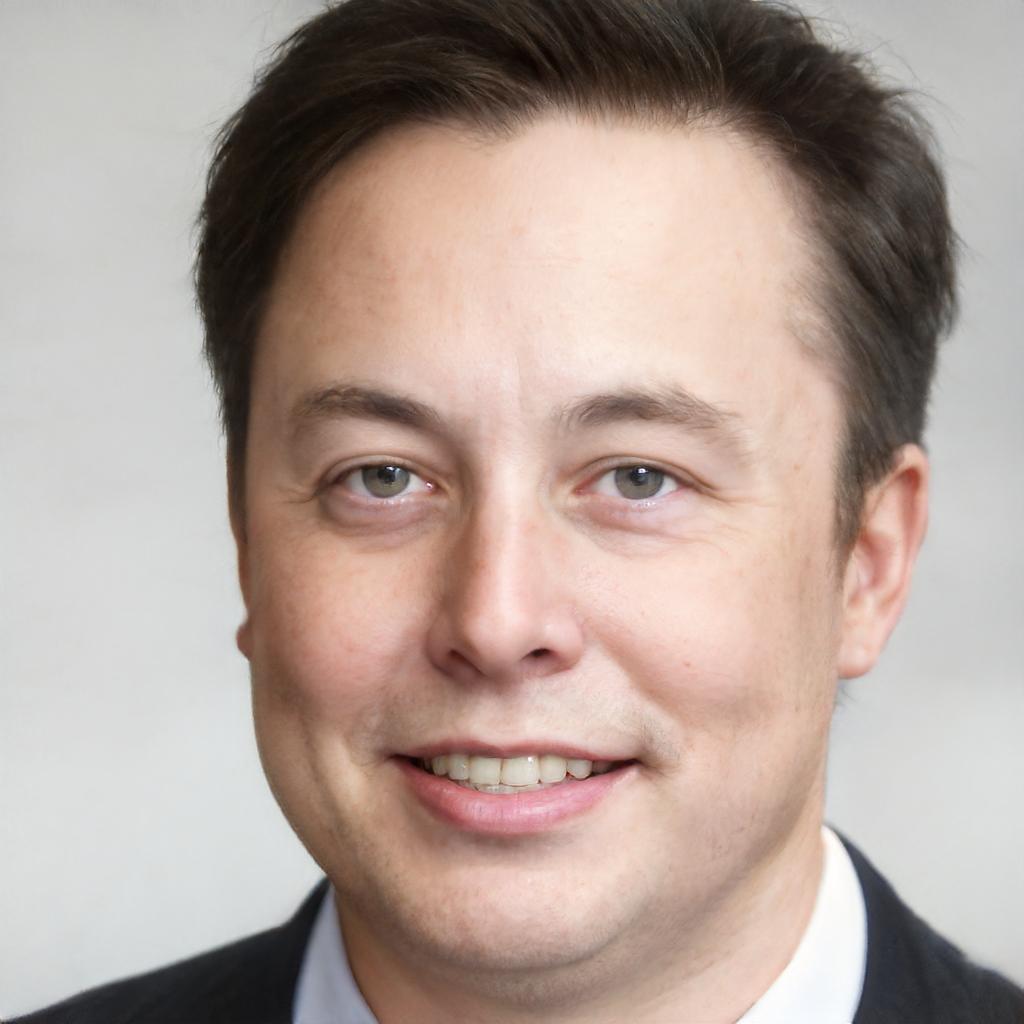} &
			\includegraphics[width=0.10\textwidth]{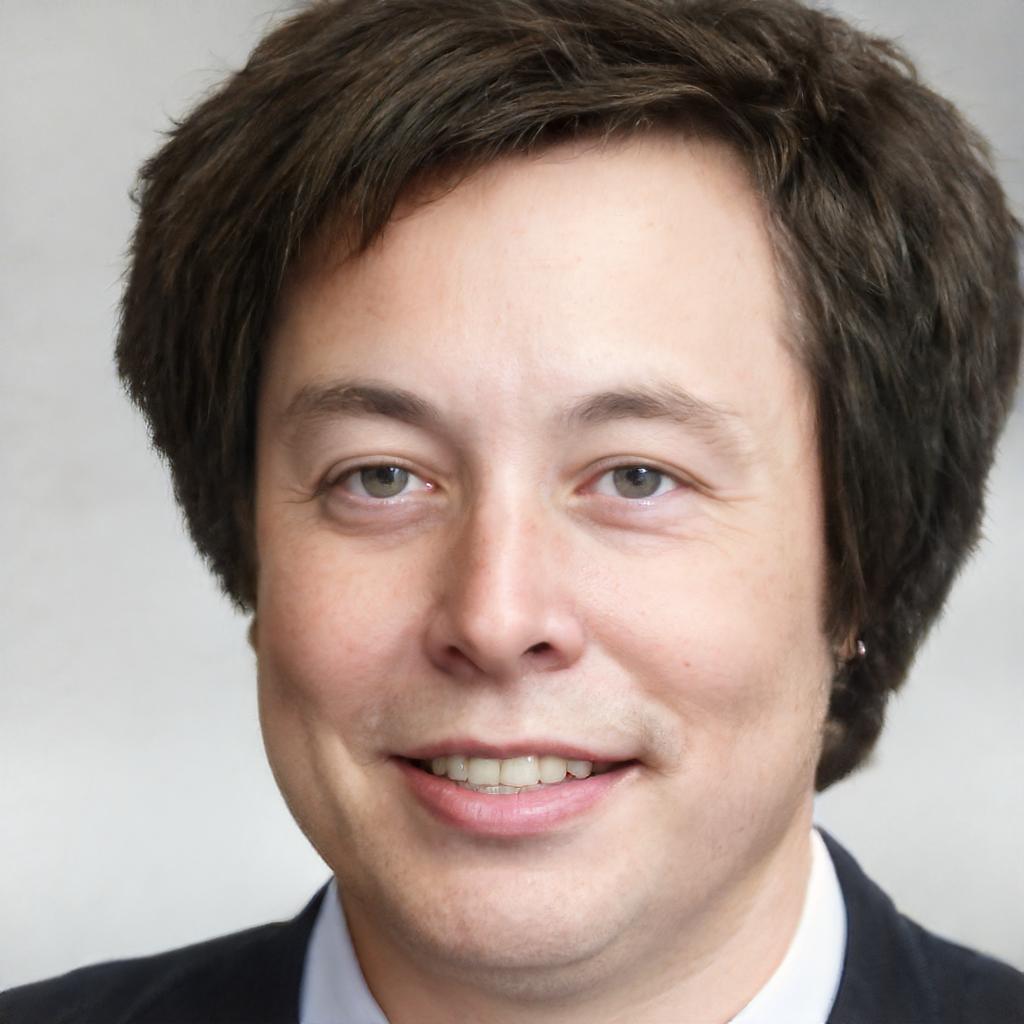} &
			\includegraphics[width=0.10\textwidth]{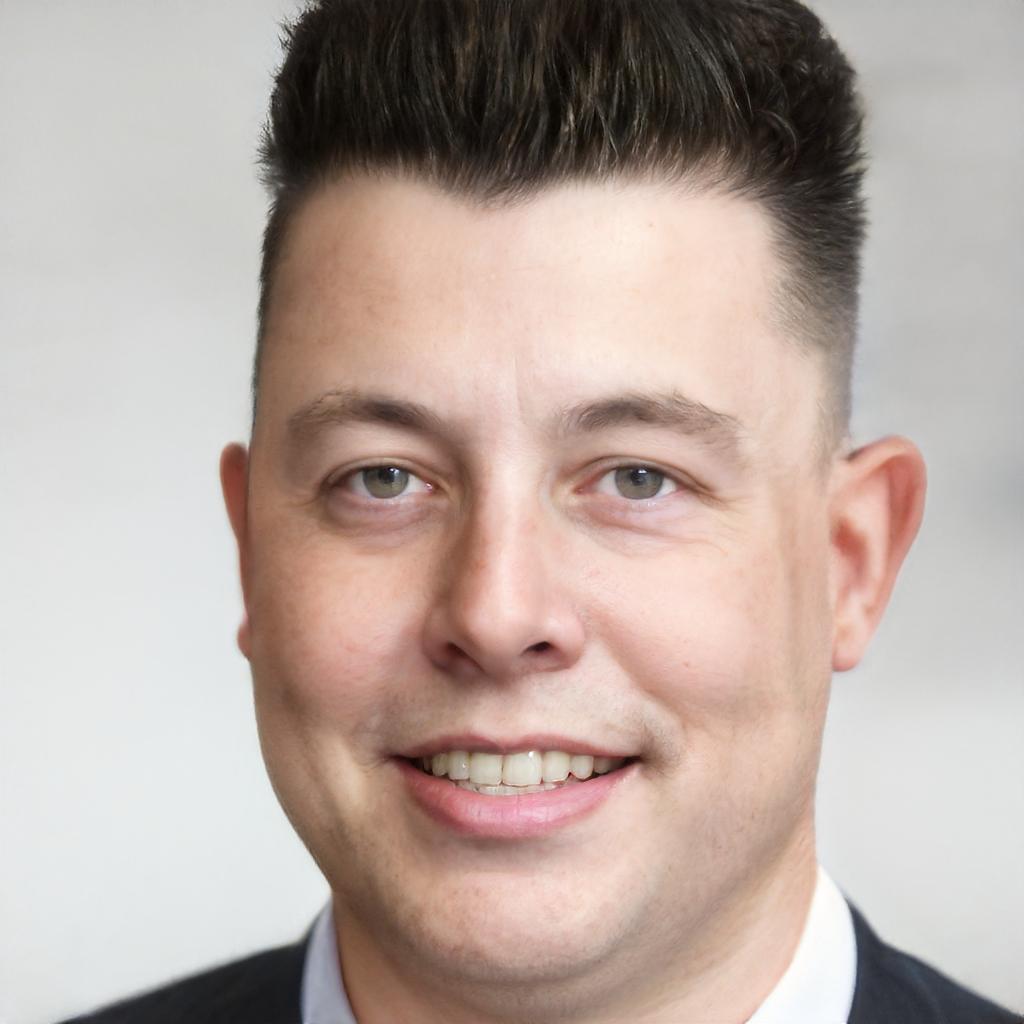} &
			\includegraphics[width=0.10\textwidth]{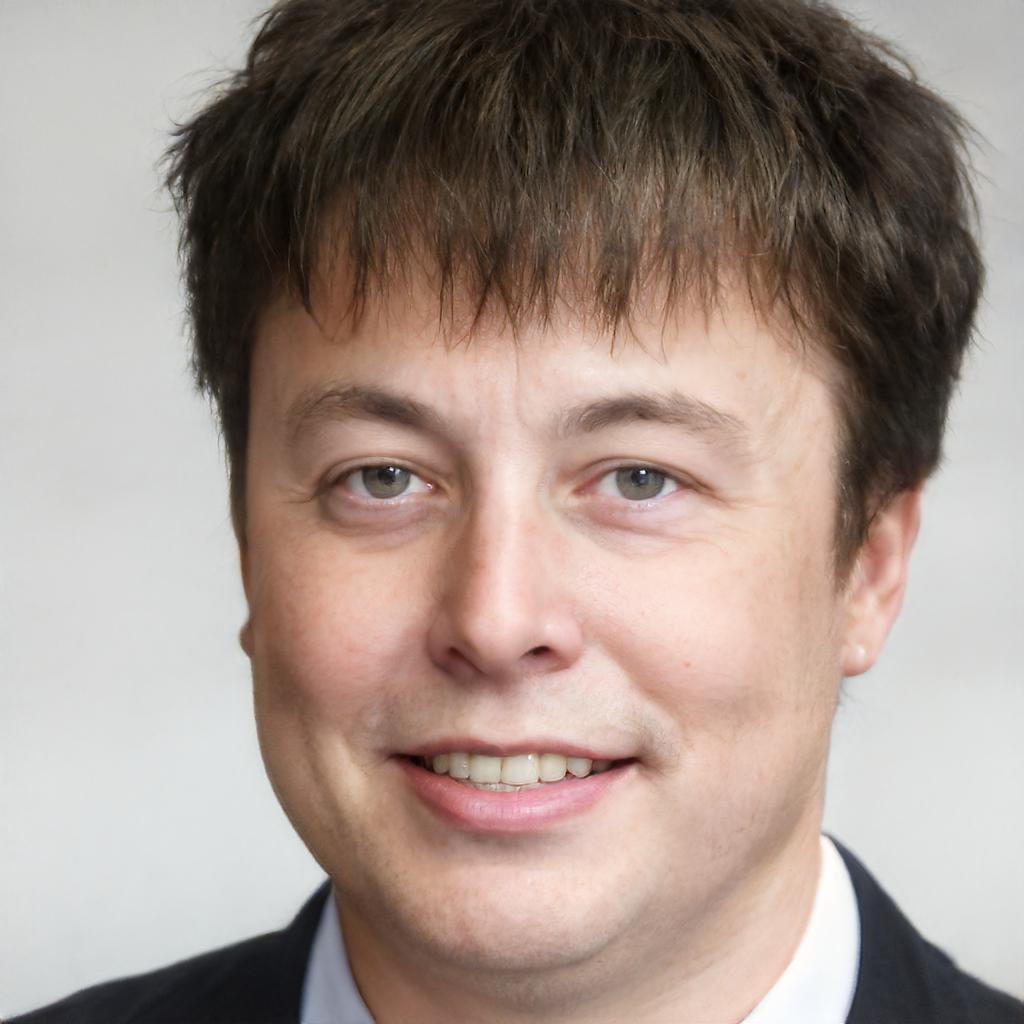}
			\\
			
			Input & Wrinkle & Sad & Angry & Surprised &  Beard & Bald & Grey Hair & Black Hair \\
			\includegraphics[width=0.10\textwidth]{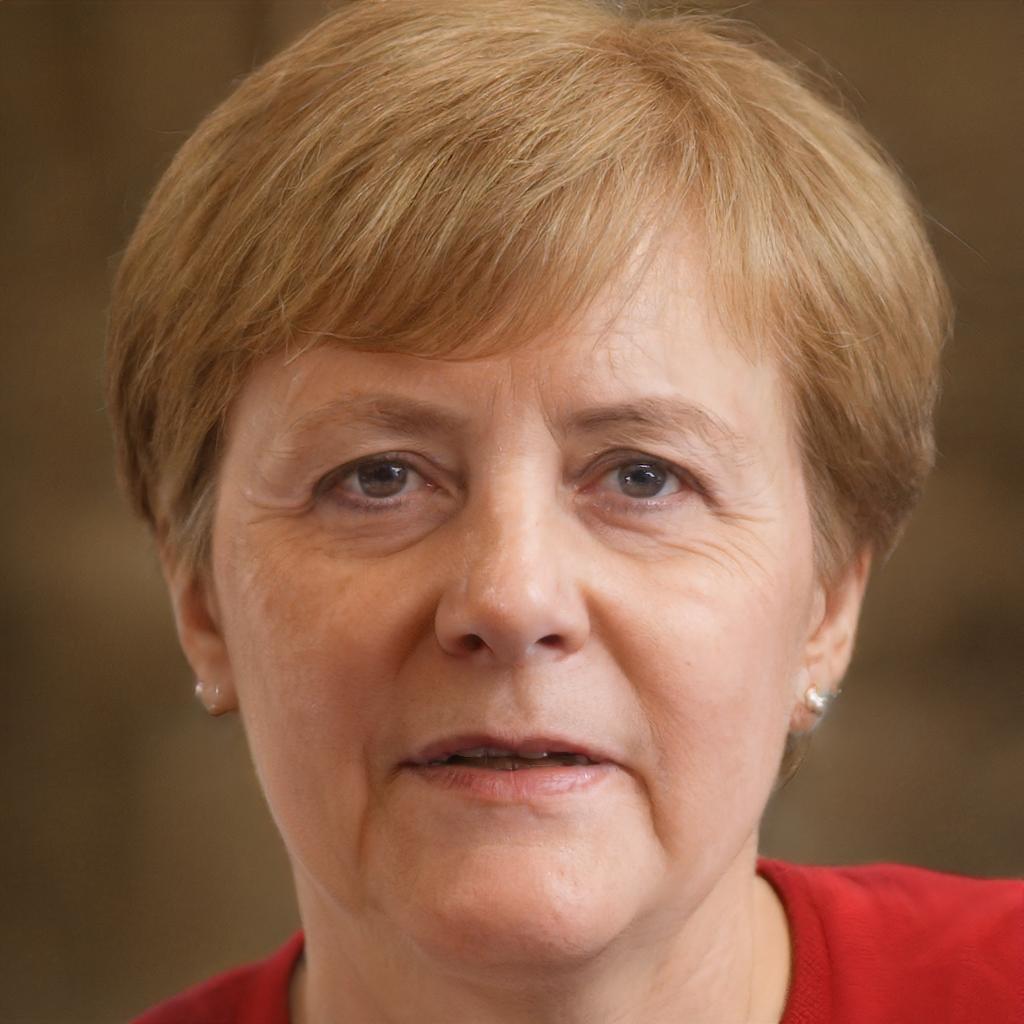} &
			\includegraphics[width=0.10\textwidth]{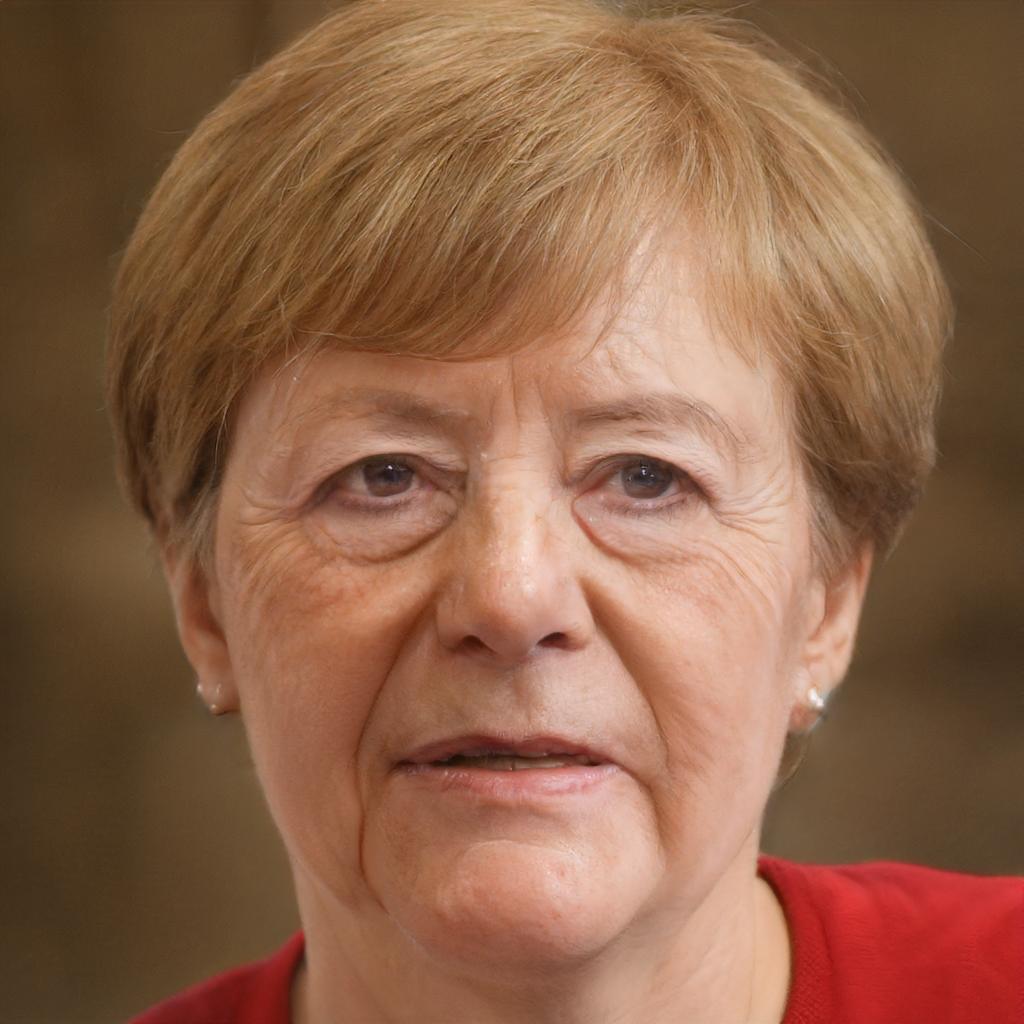} &
			\includegraphics[width=0.10\textwidth]{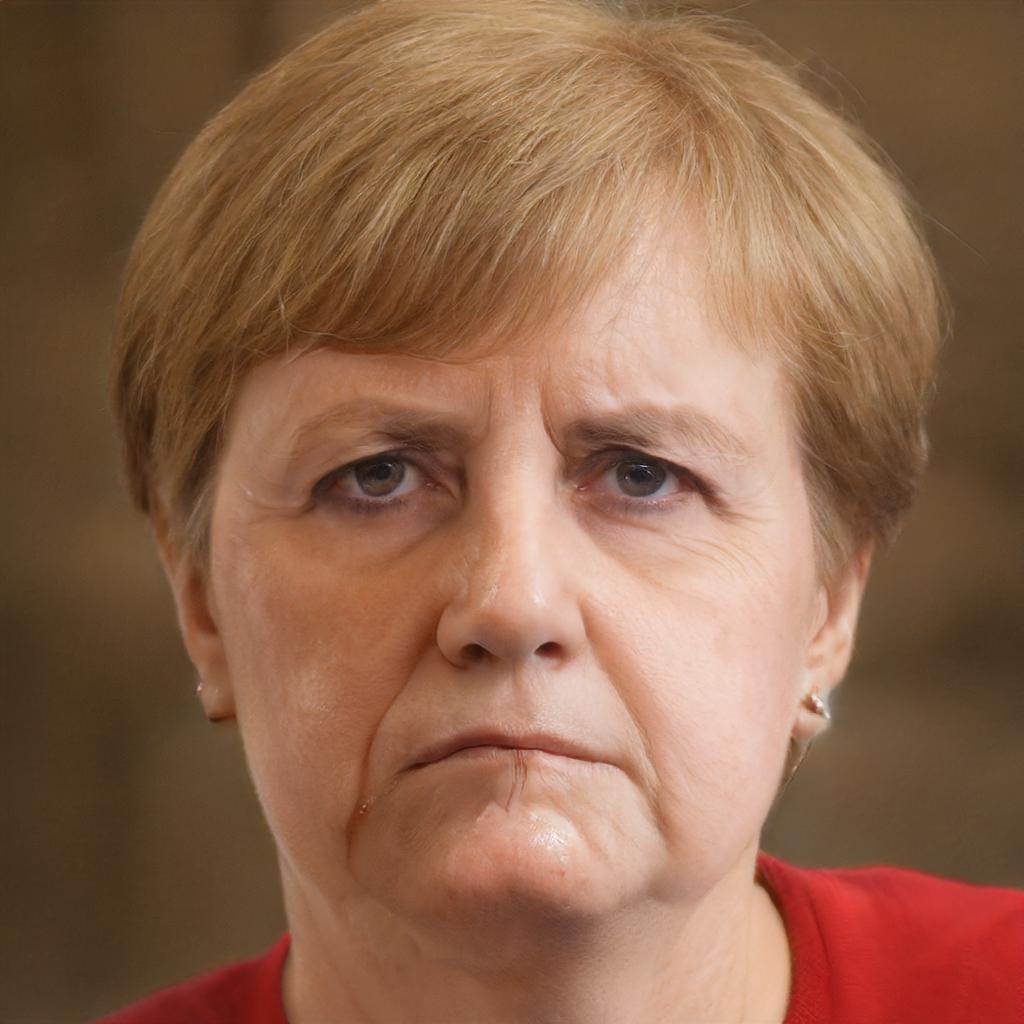} &
			\includegraphics[width=0.10\textwidth]{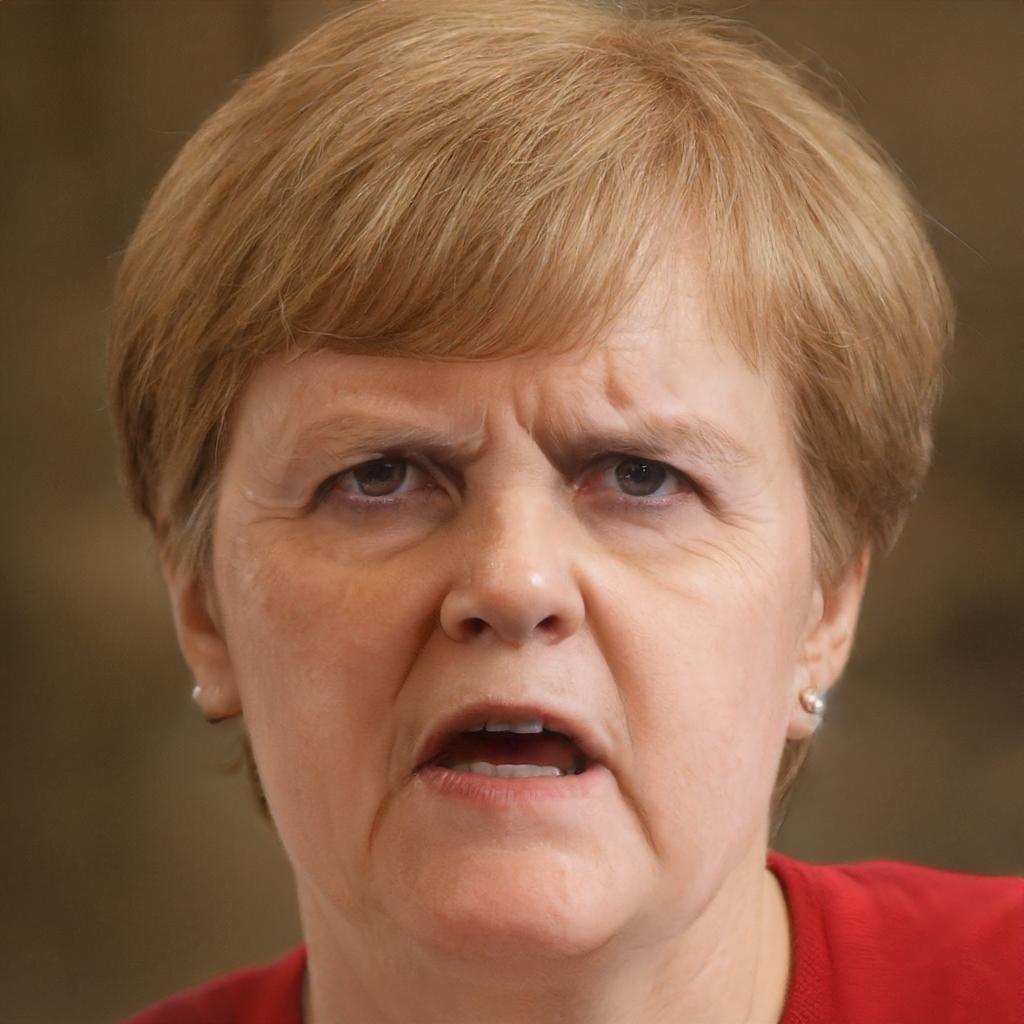} &
			\includegraphics[width=0.10\textwidth]{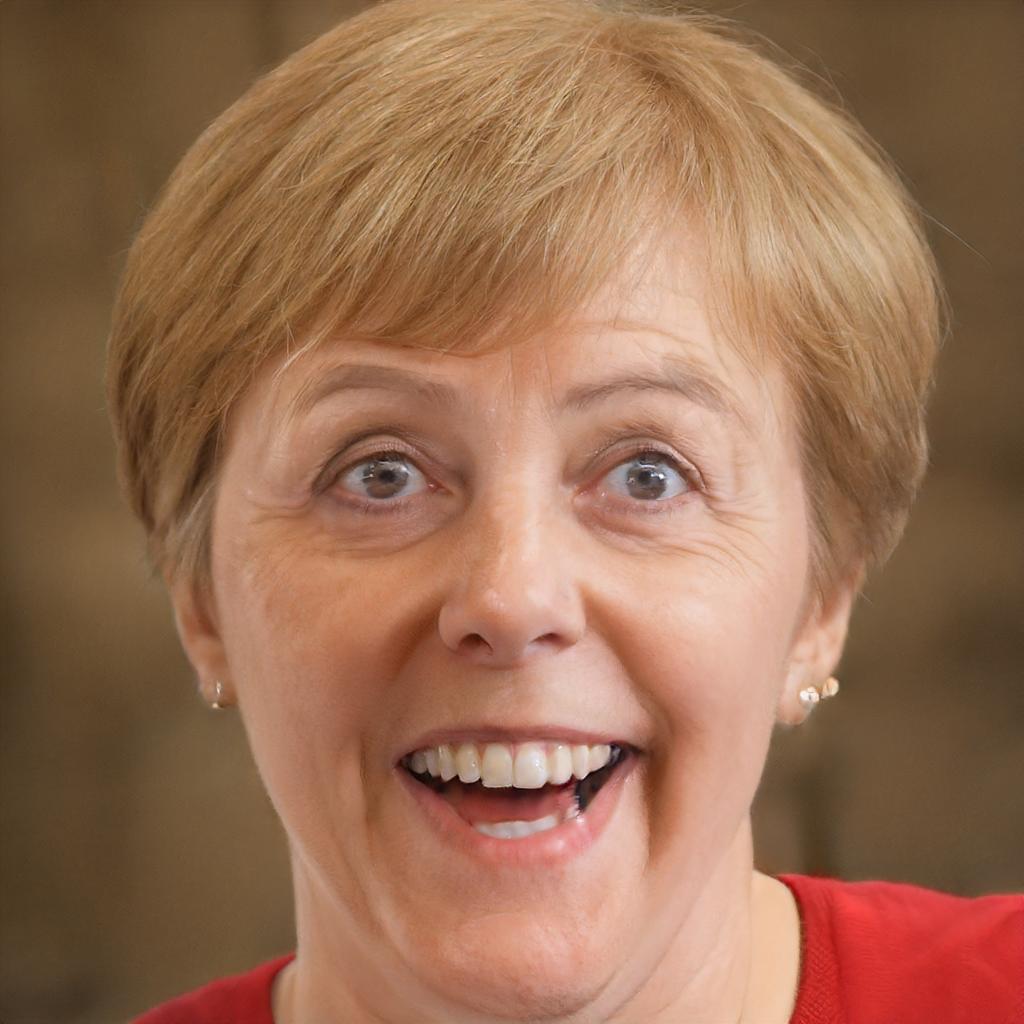} &
			\includegraphics[width=0.10\textwidth]{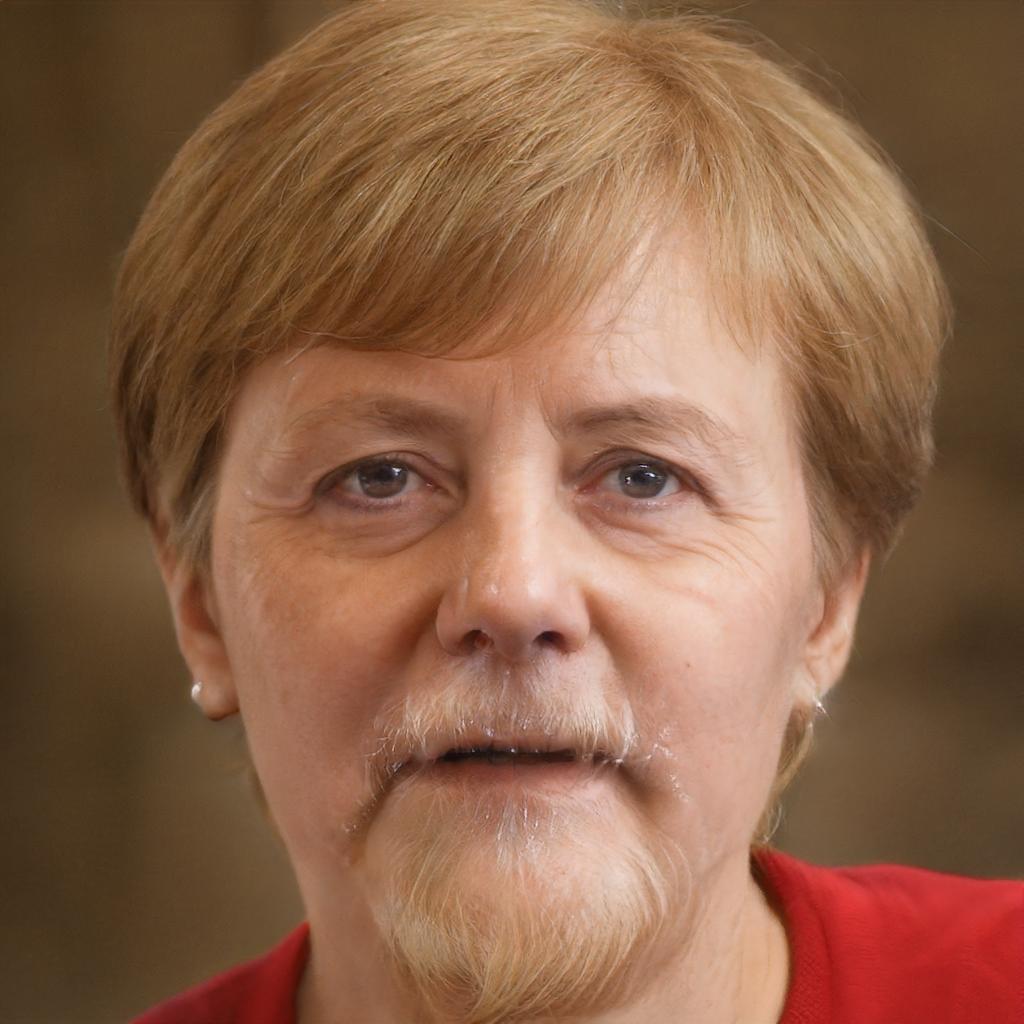} &
			\includegraphics[width=0.10\textwidth]{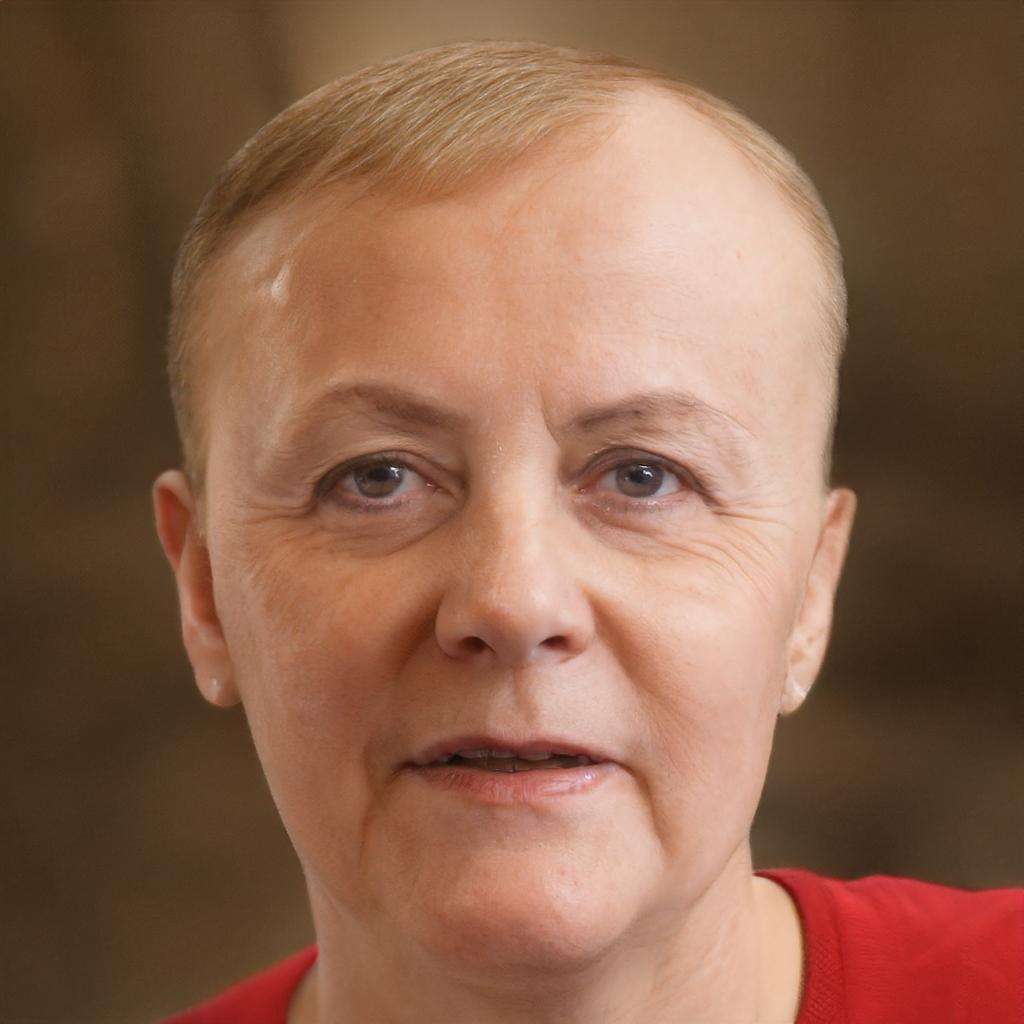} &
			\includegraphics[width=0.10\textwidth]{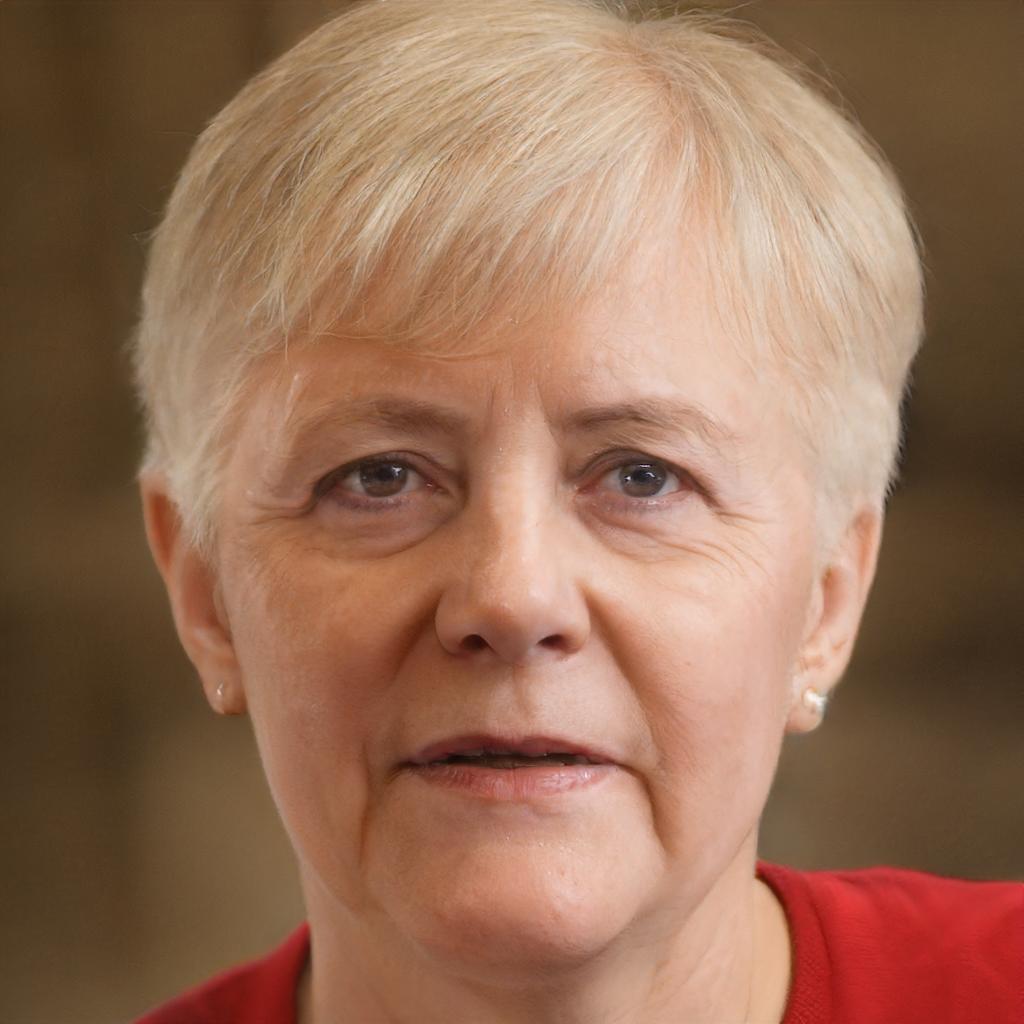} &
			\includegraphics[width=0.10\textwidth]{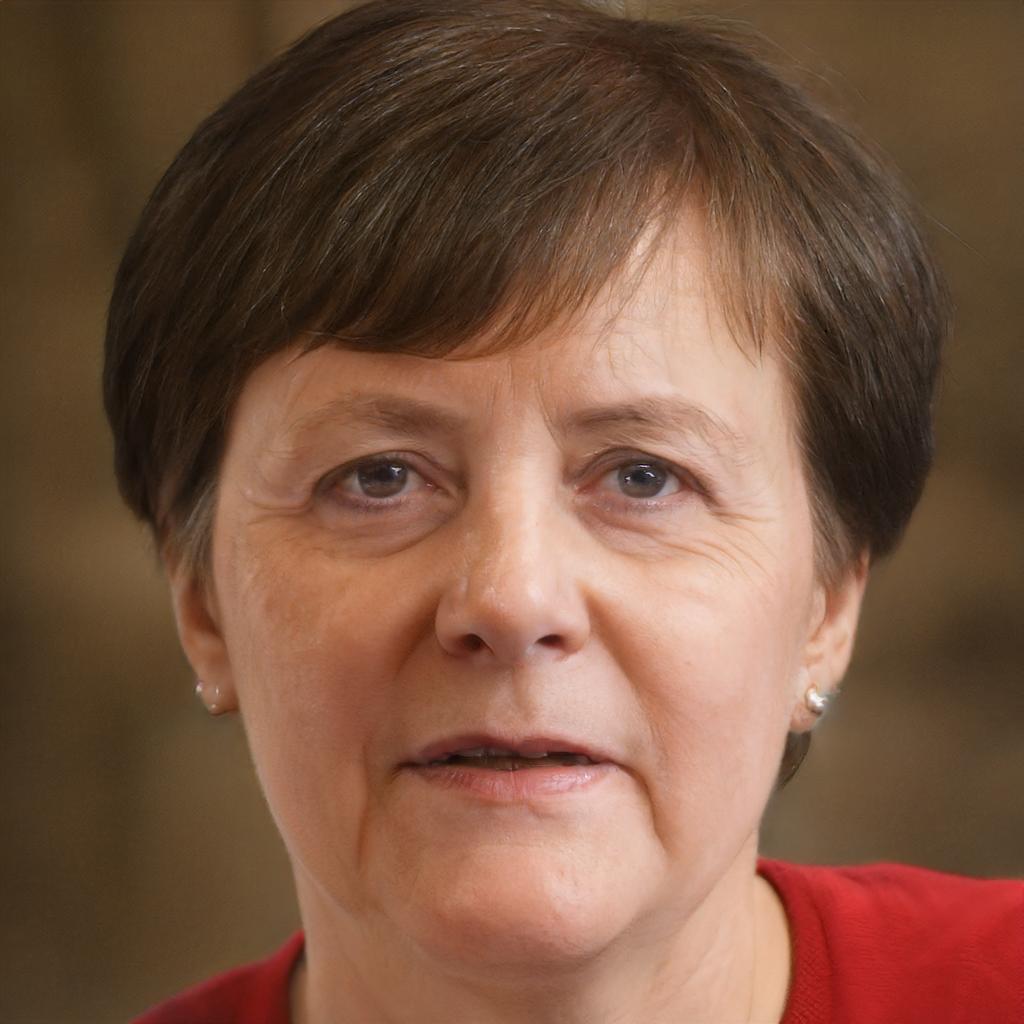}
			\\
			\includegraphics[width=0.10\textwidth]{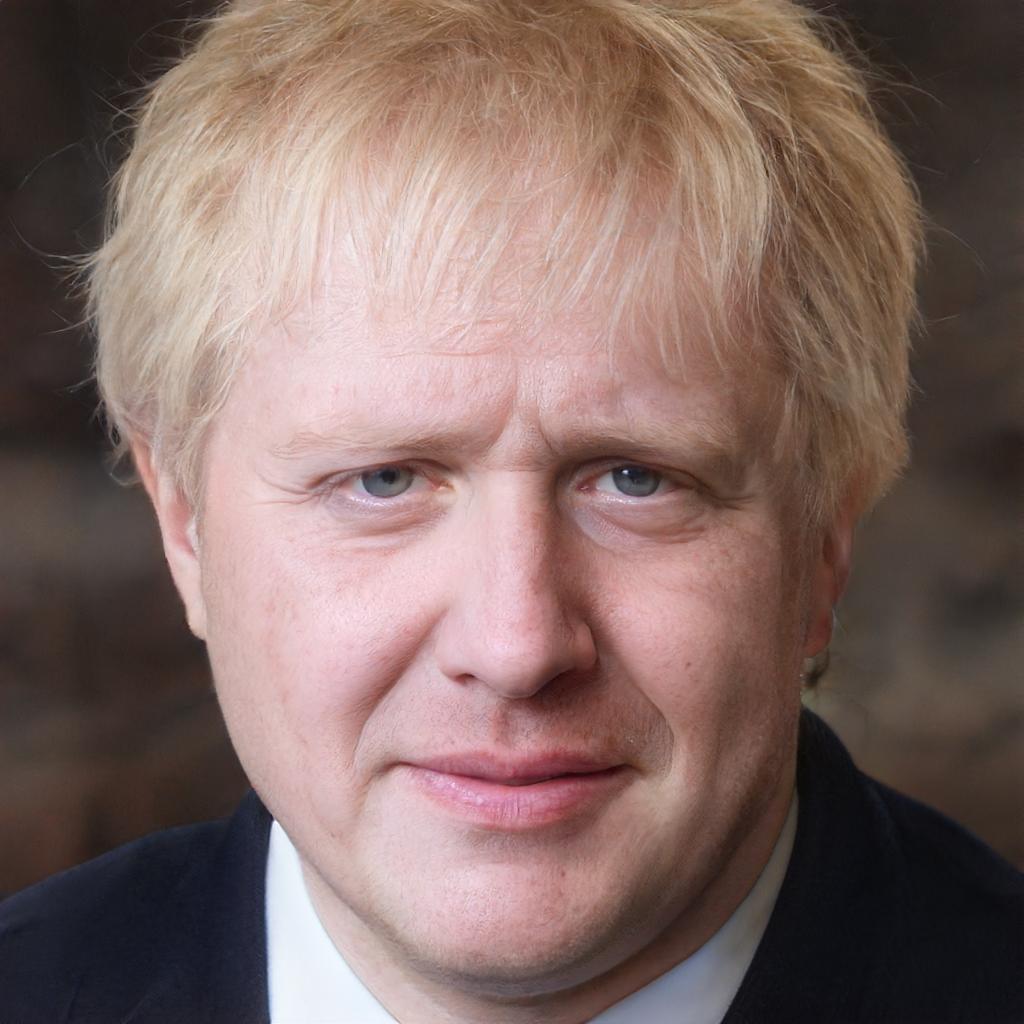} &
			\includegraphics[width=0.10\textwidth]{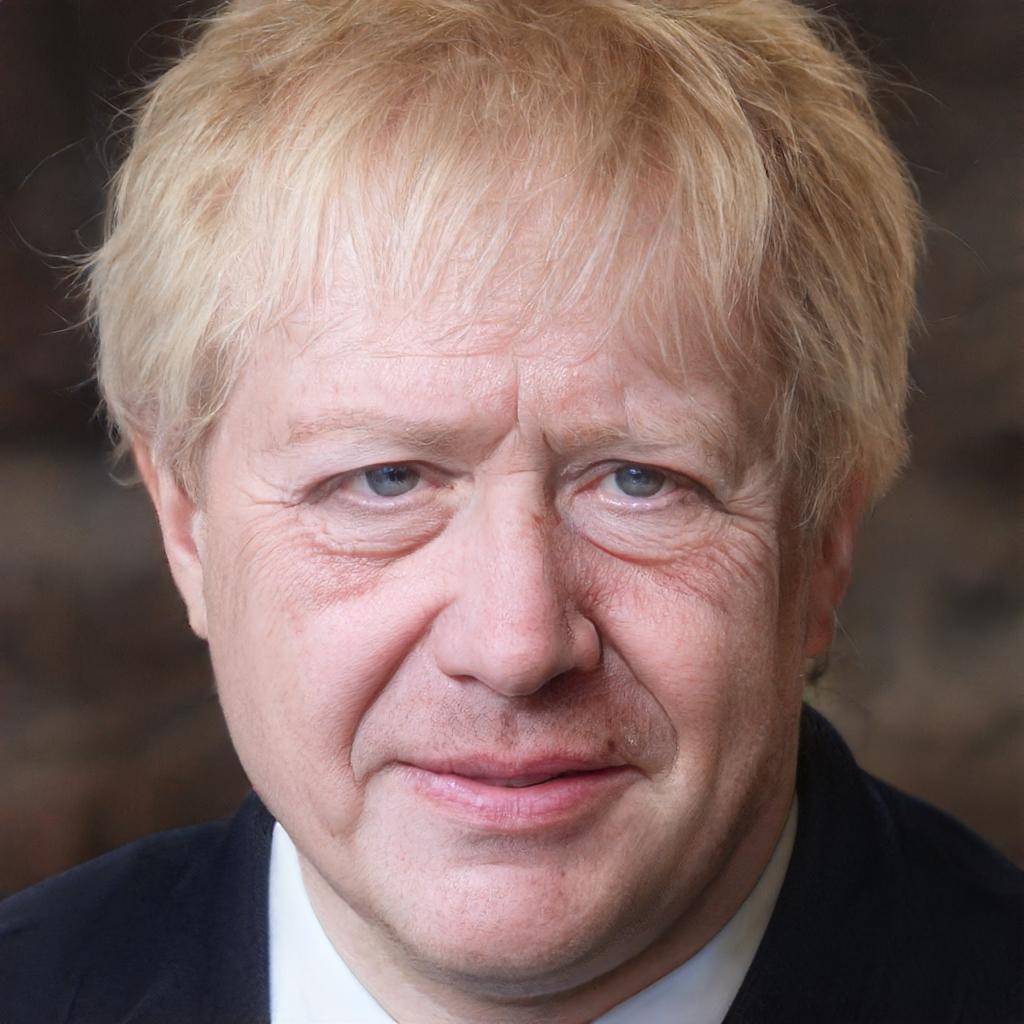} &
			\includegraphics[width=0.10\textwidth]{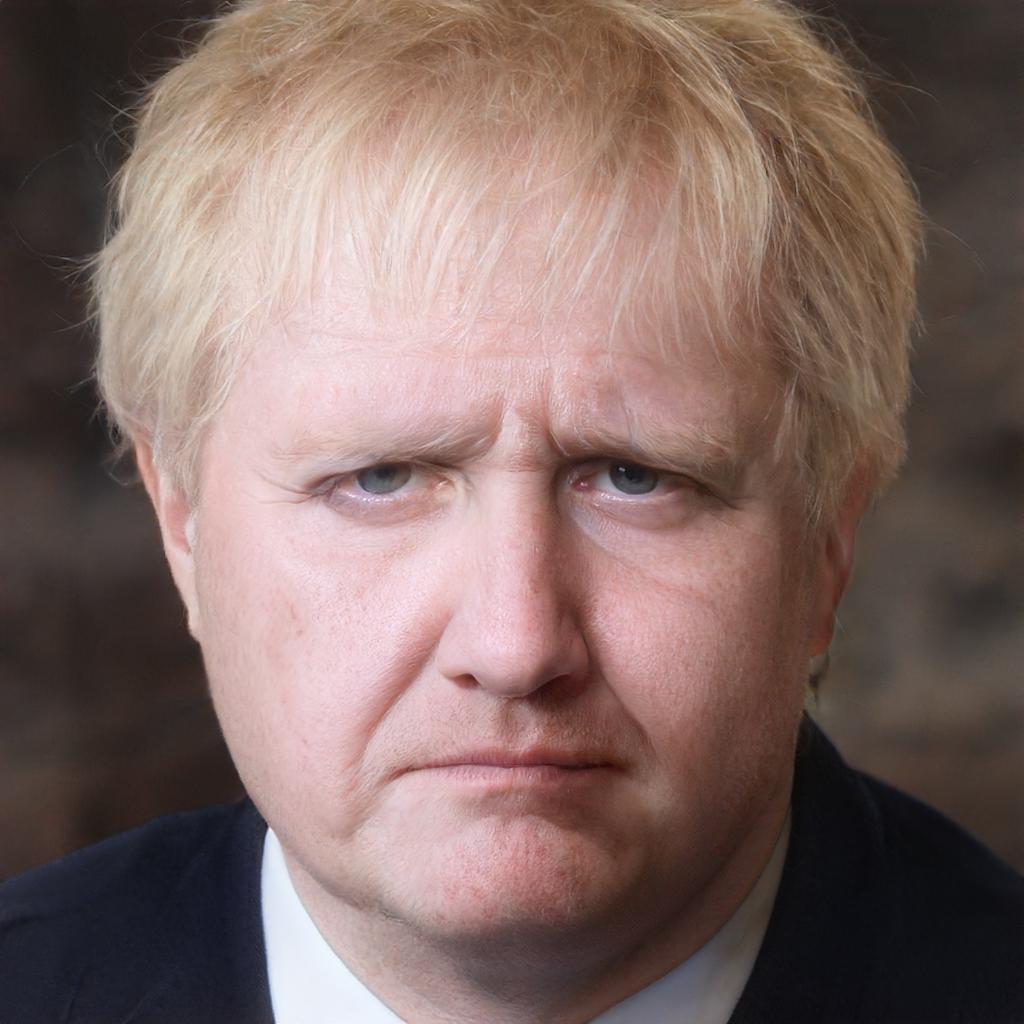} &
			\includegraphics[width=0.10\textwidth]{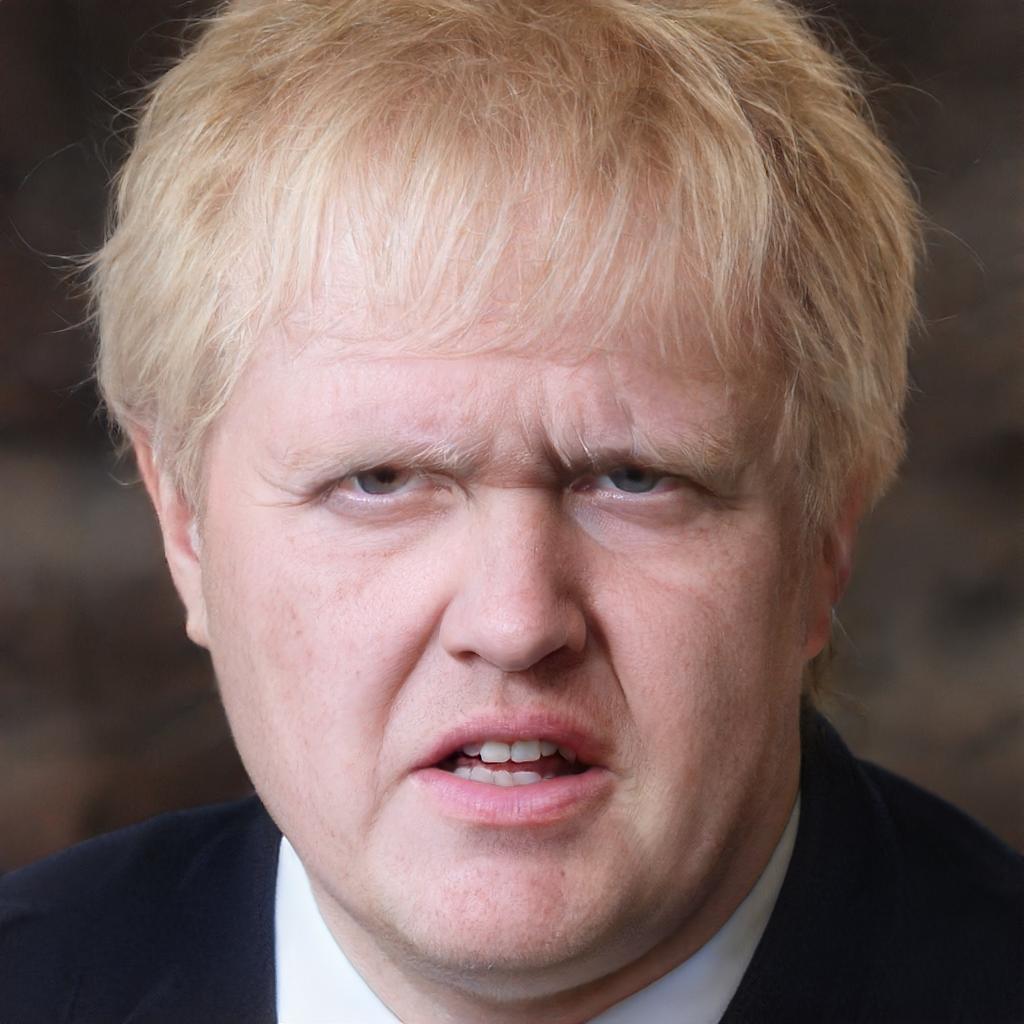} &
			\includegraphics[width=0.10\textwidth]{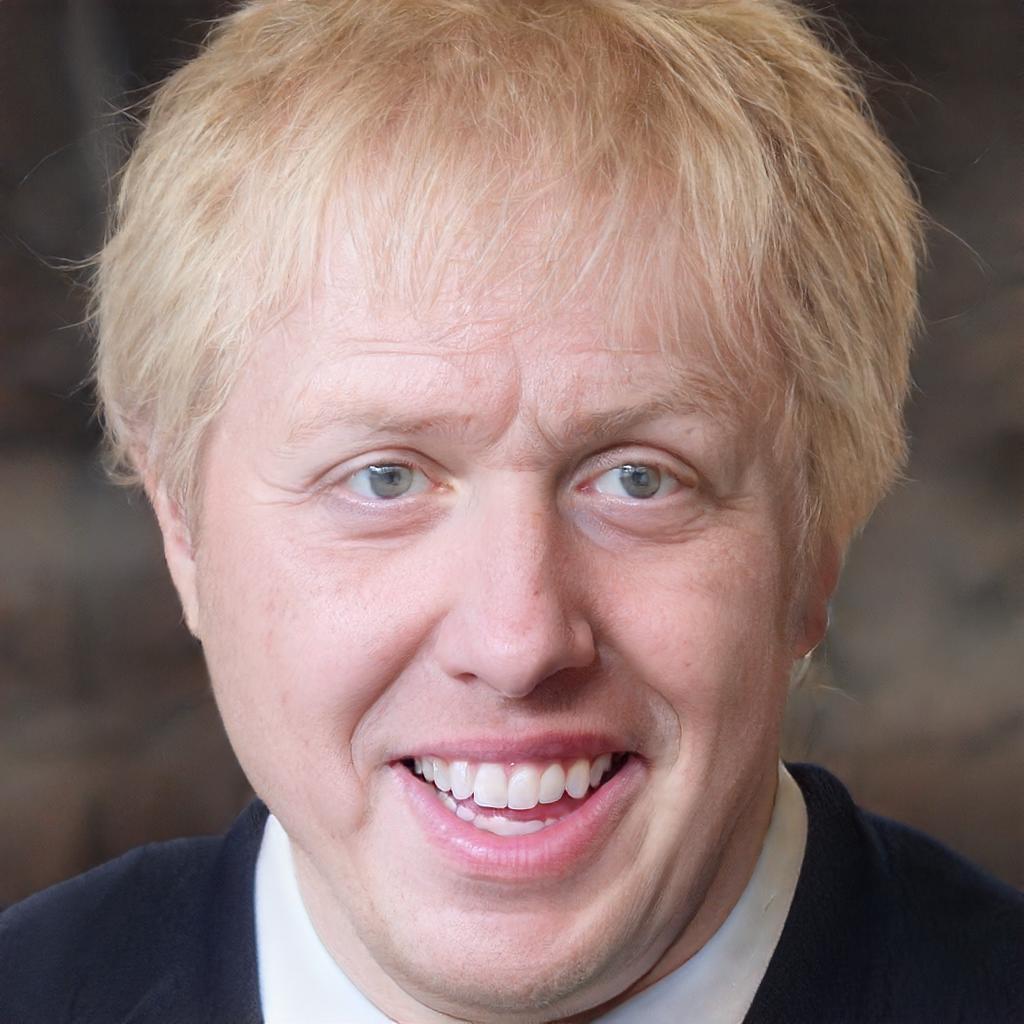} &
			\includegraphics[width=0.10\textwidth]{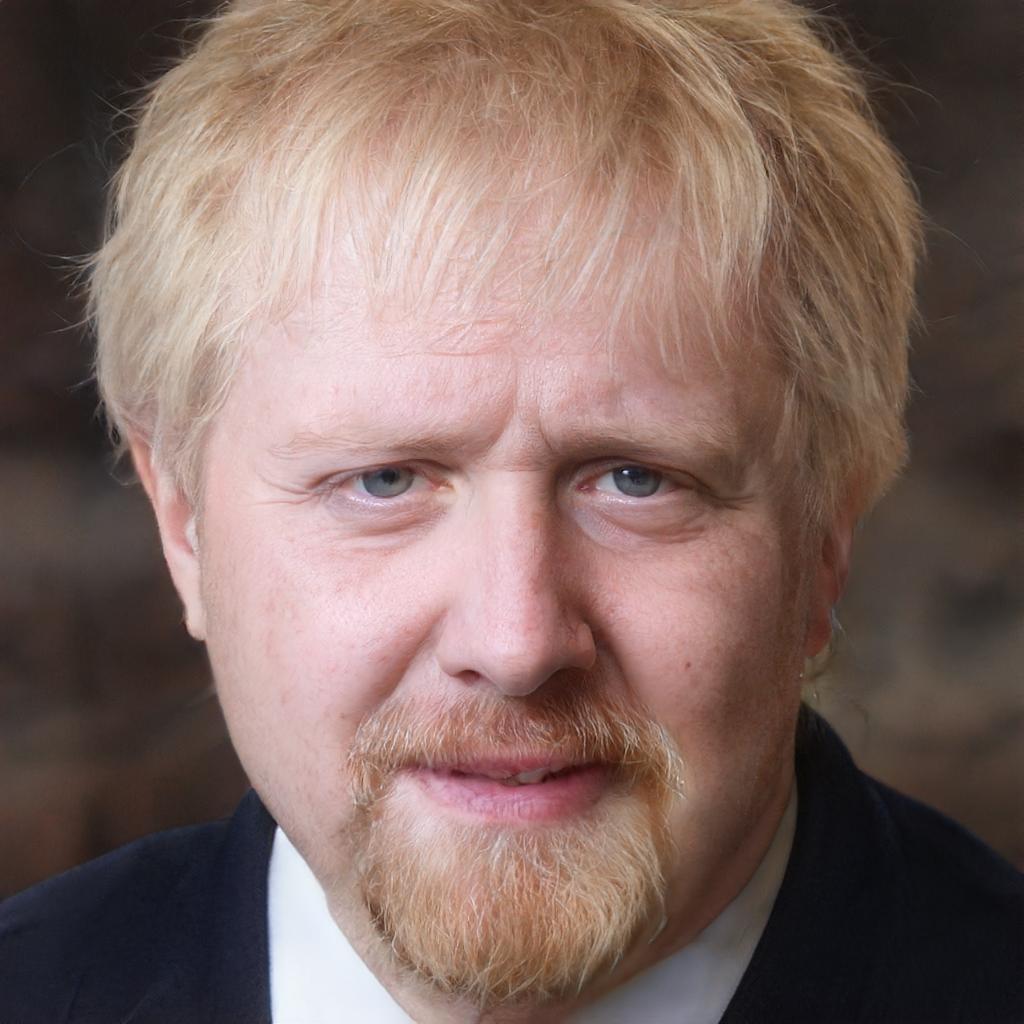} &
			\includegraphics[width=0.10\textwidth]{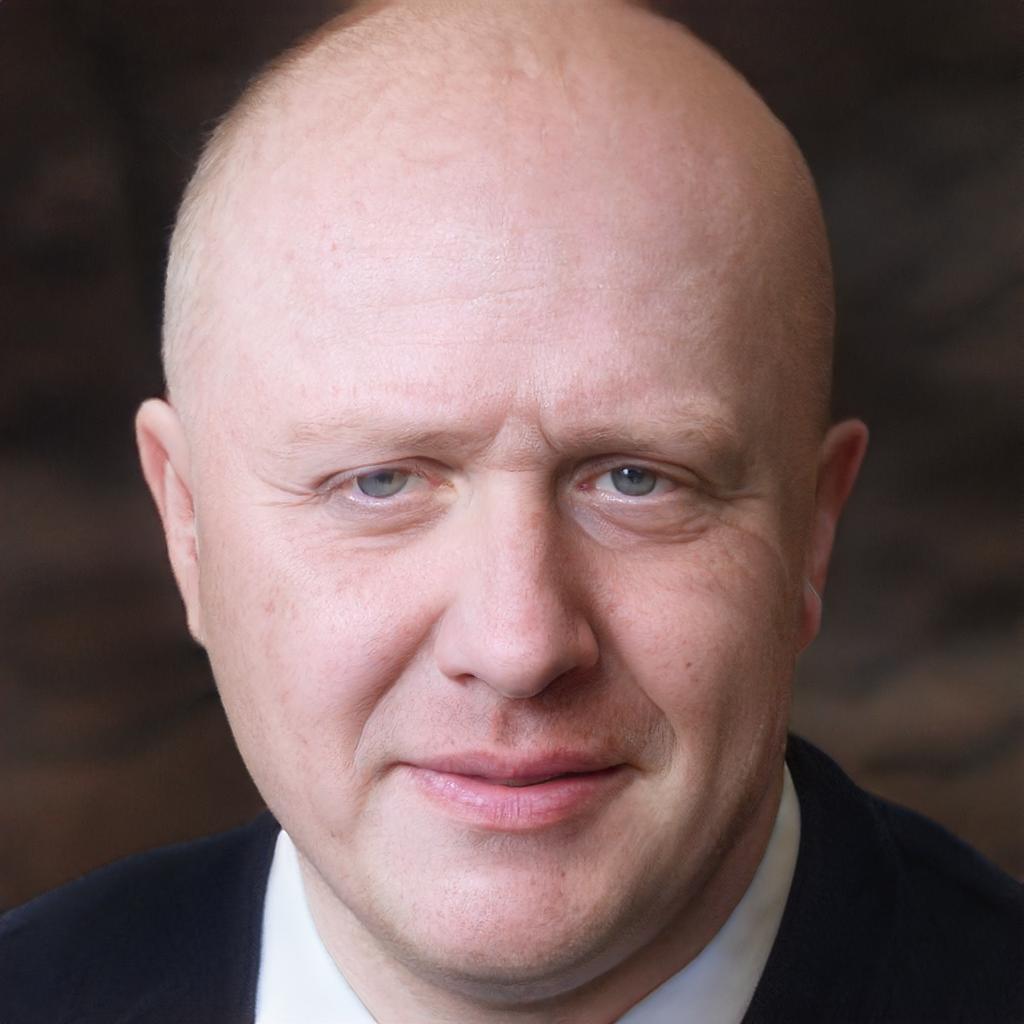} &
			\includegraphics[width=0.10\textwidth]{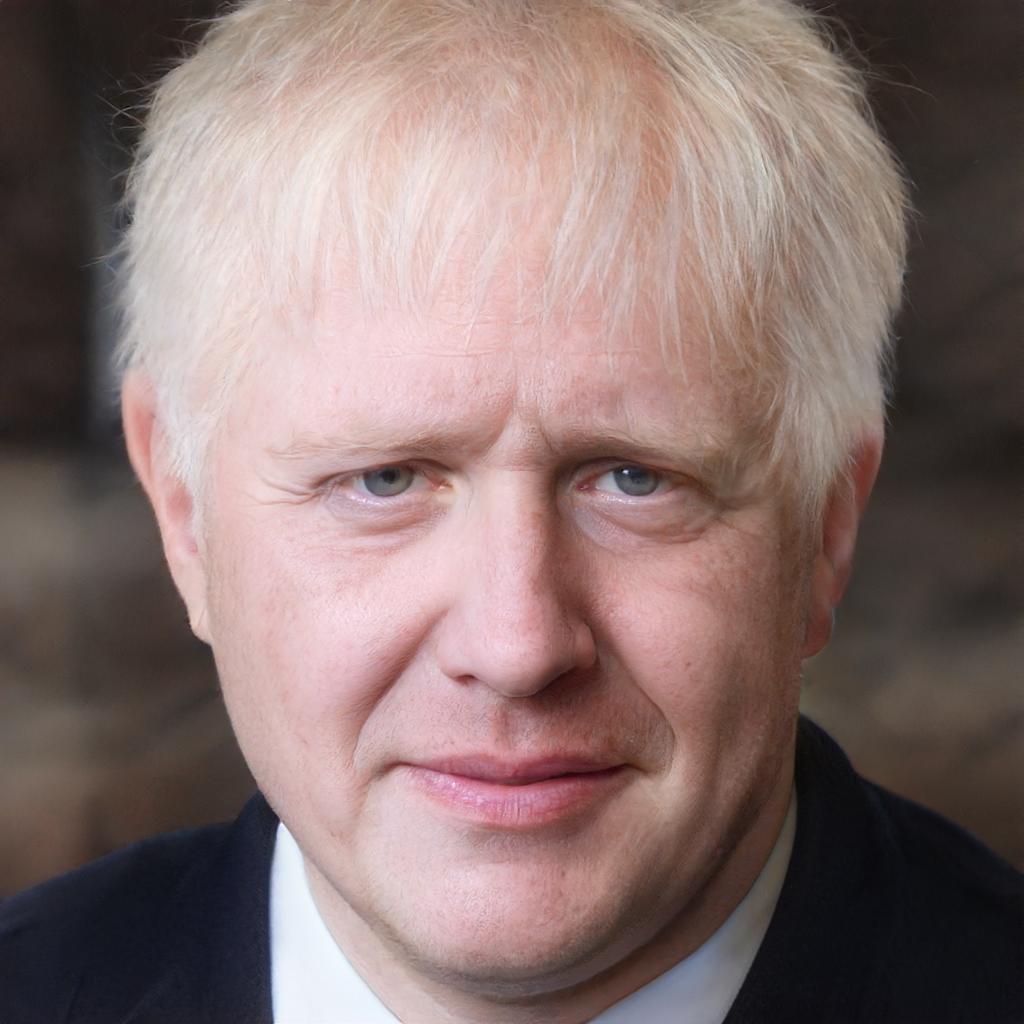} &
			\includegraphics[width=0.10\textwidth]{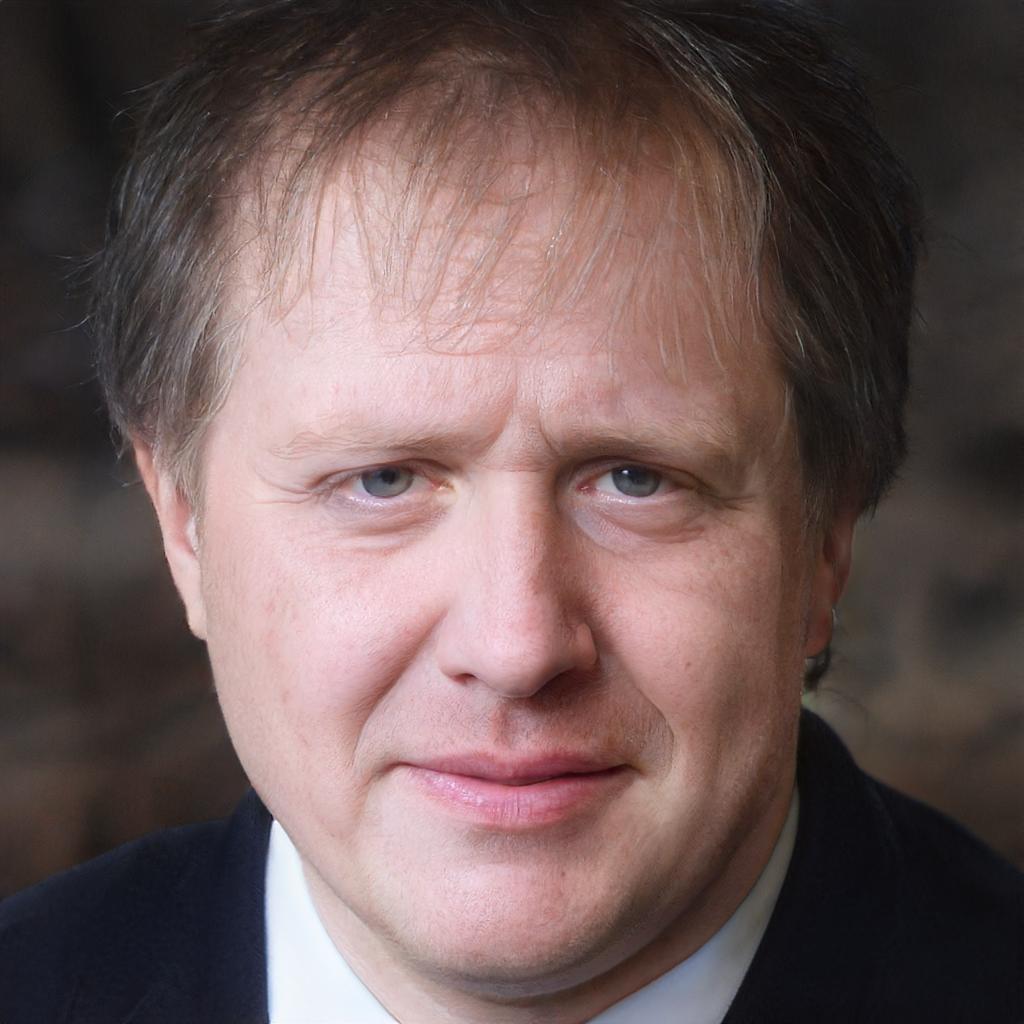}  
			\\
			\includegraphics[width=0.10\textwidth]{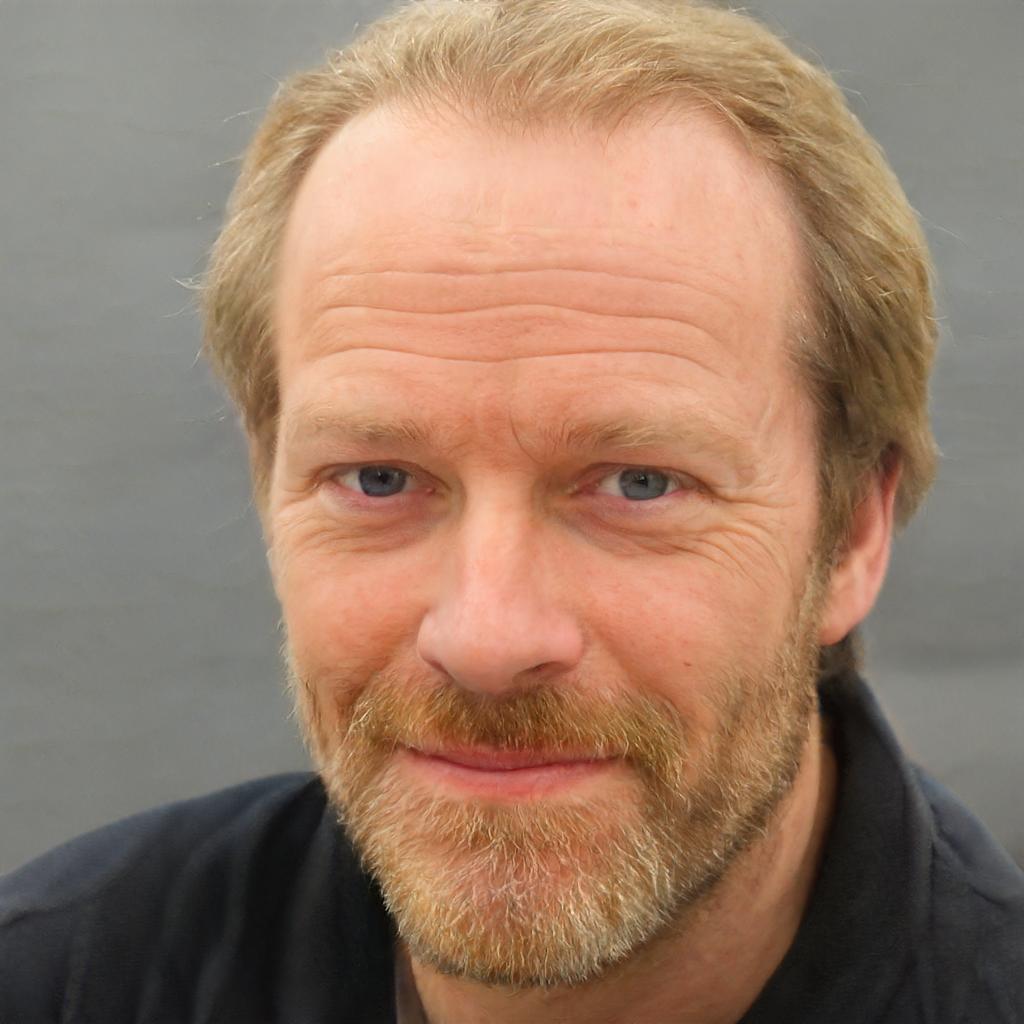} &
			\includegraphics[width=0.10\textwidth]{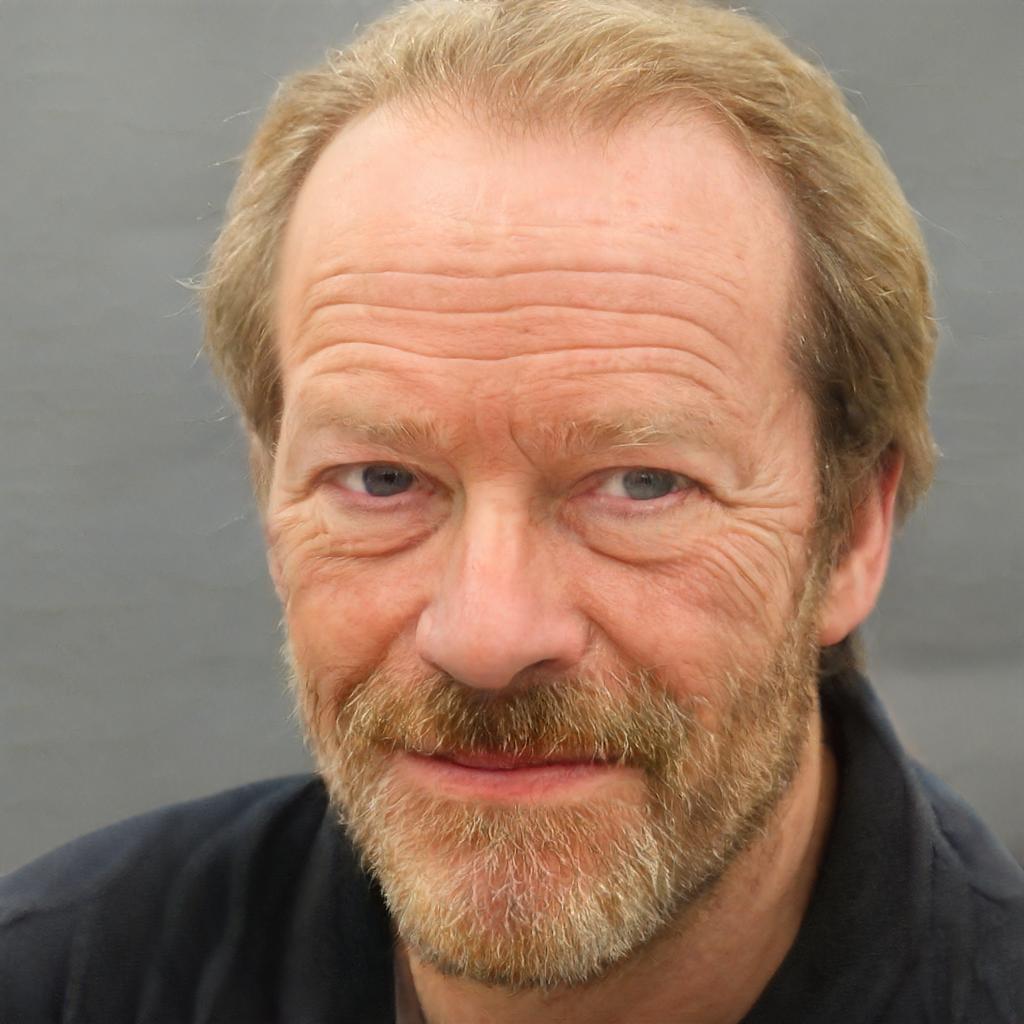} &
			\includegraphics[width=0.10\textwidth]{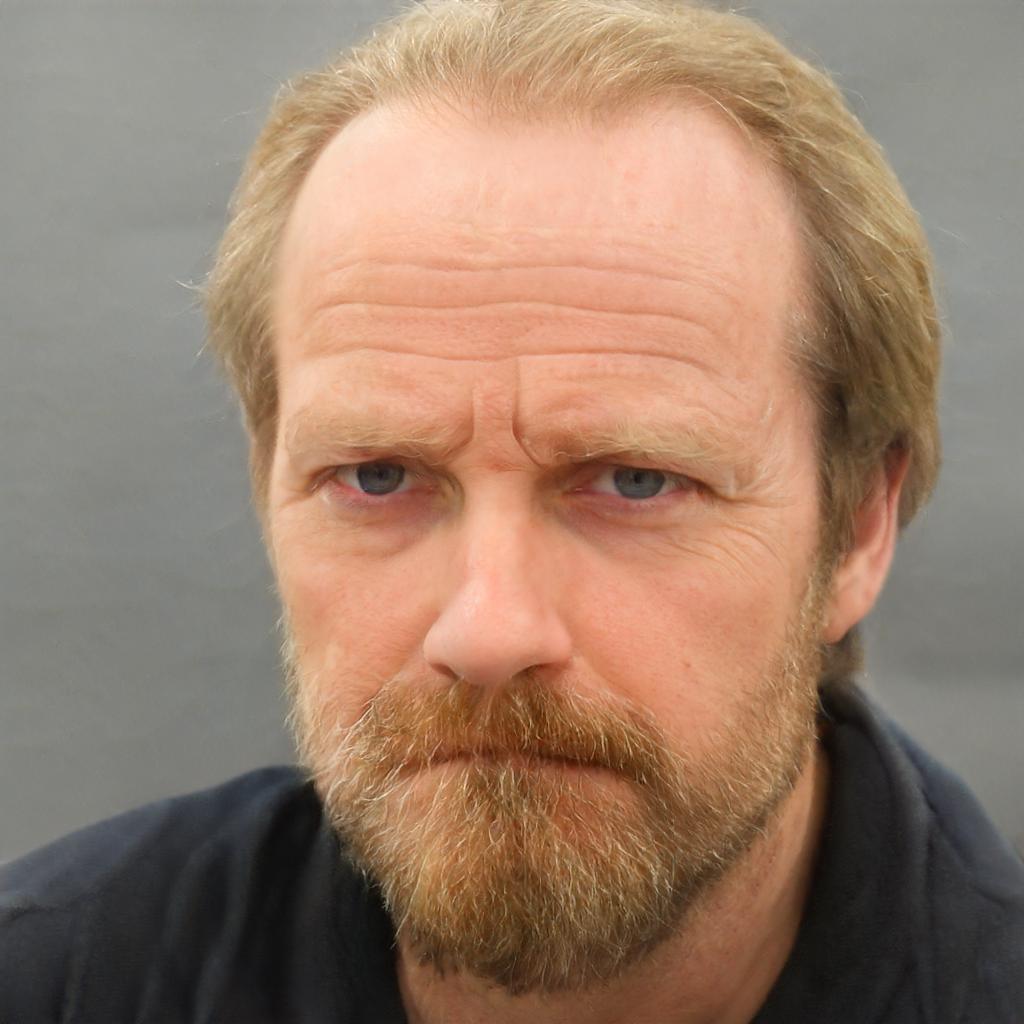} &
			\includegraphics[width=0.10\textwidth]{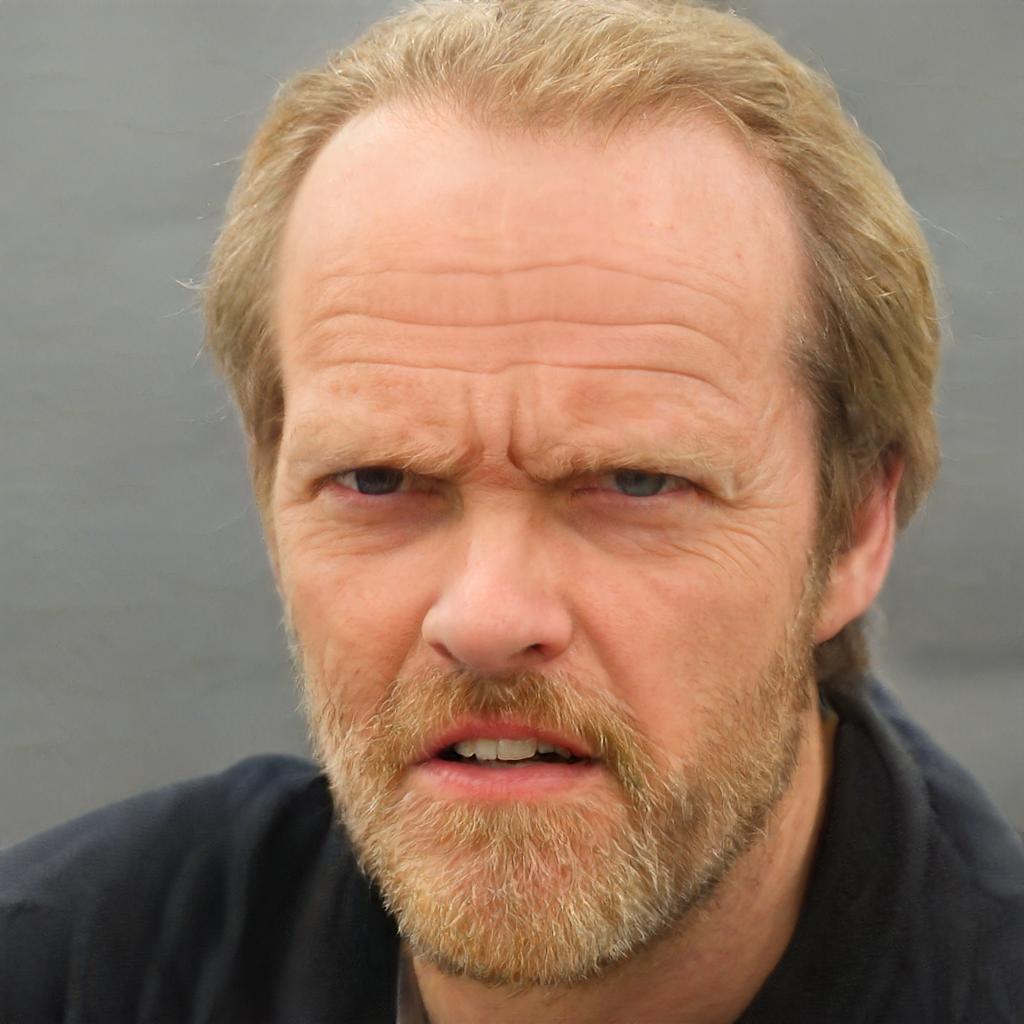} &
			\includegraphics[width=0.10\textwidth]{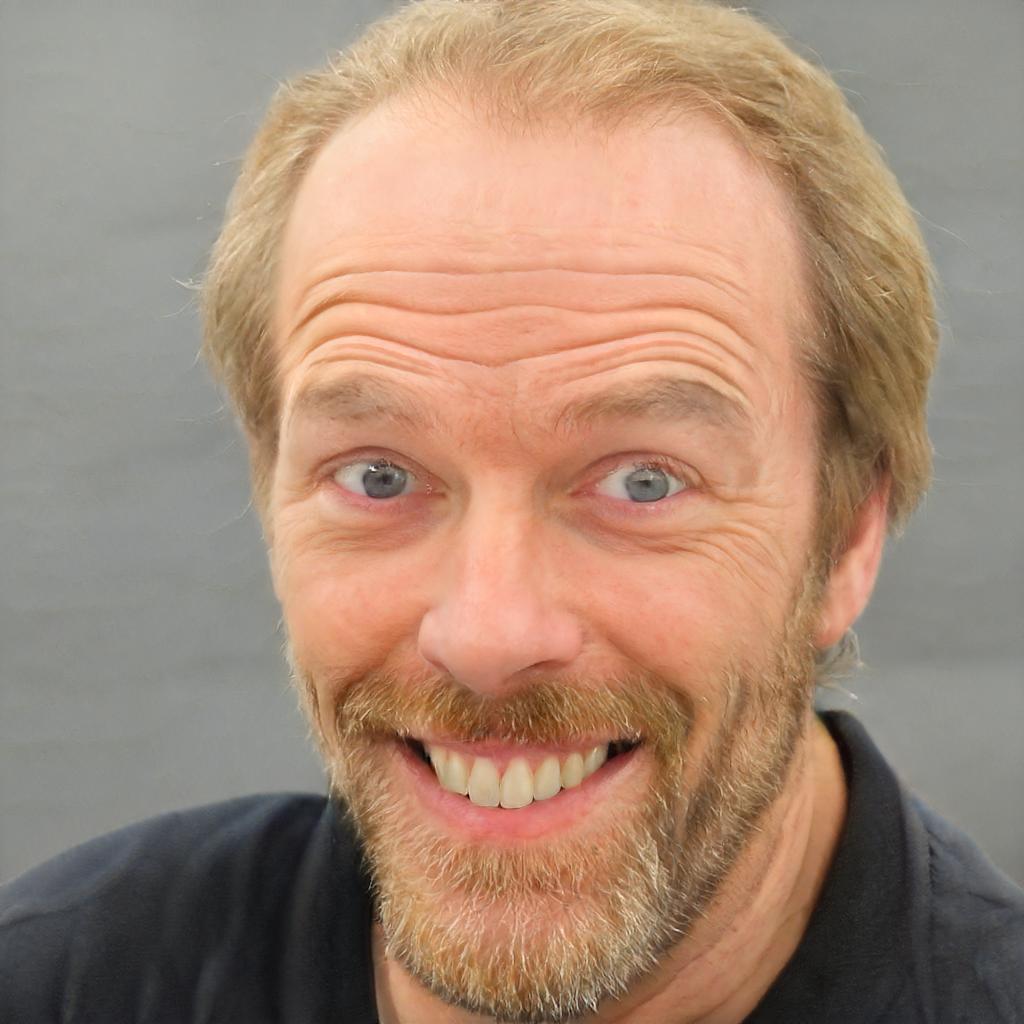} &
			\includegraphics[width=0.10\textwidth]{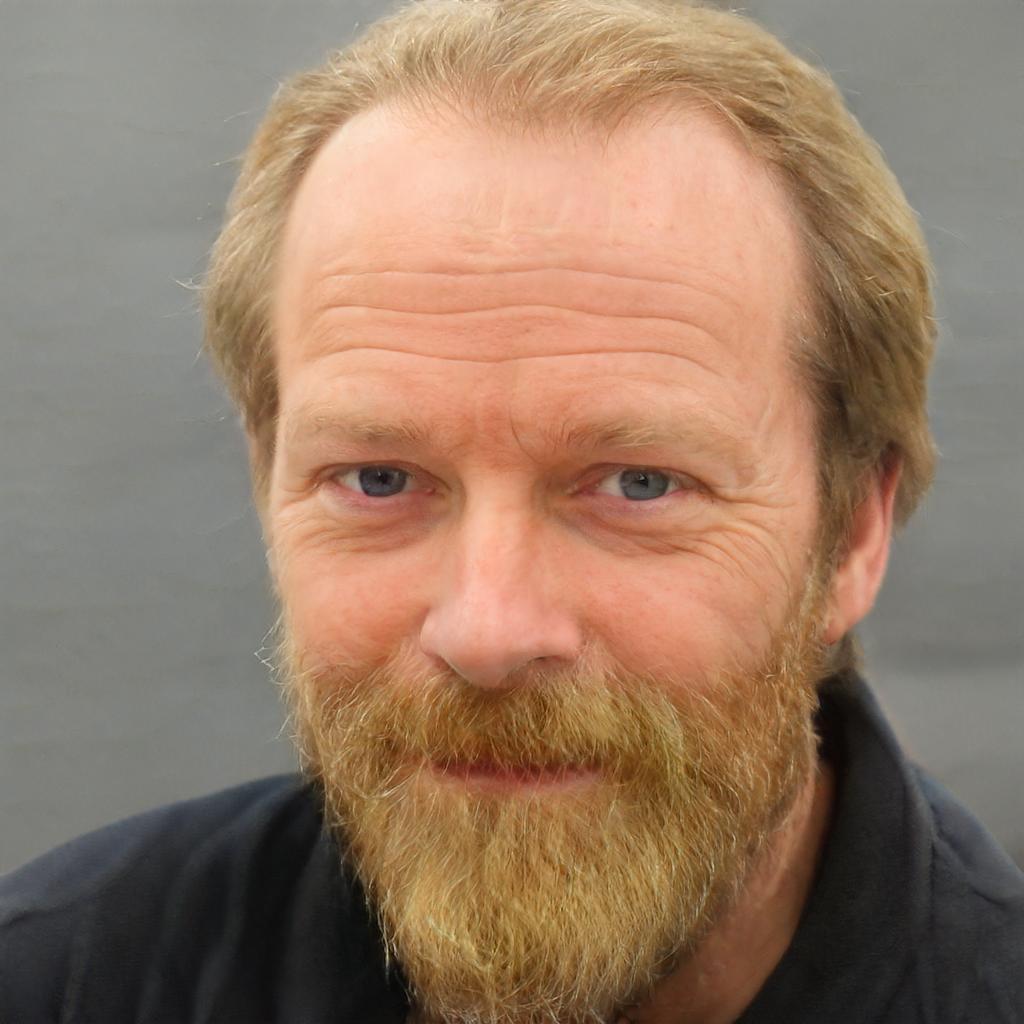} &
			\includegraphics[width=0.10\textwidth]{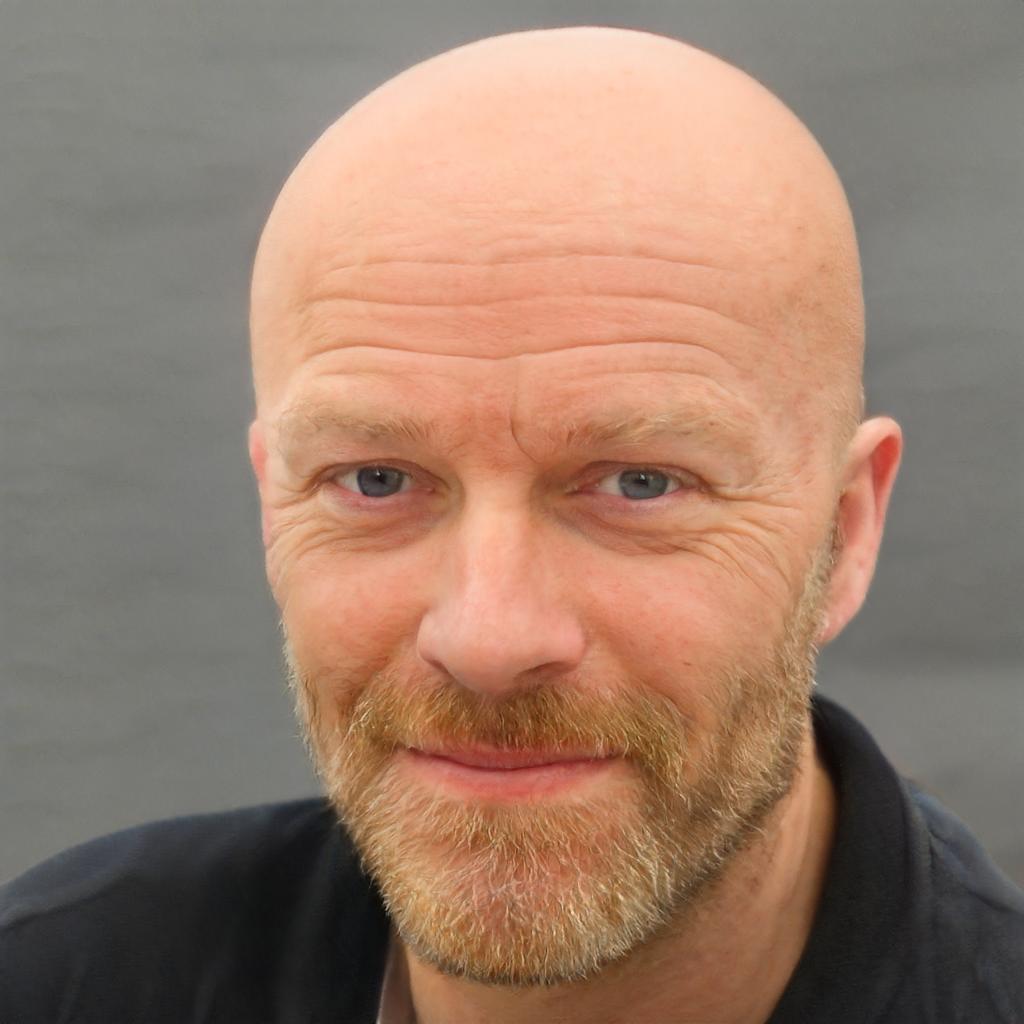} &
			\includegraphics[width=0.10\textwidth]{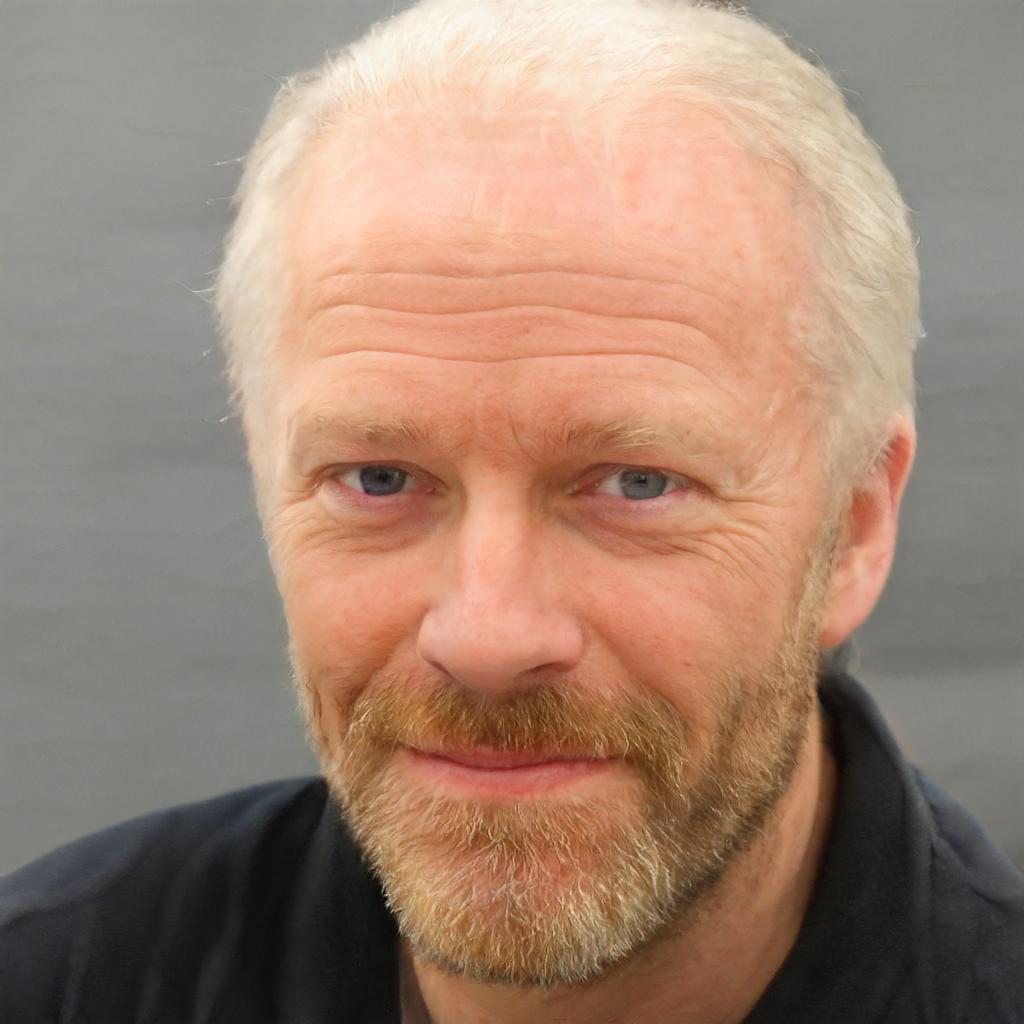} &
			\includegraphics[width=0.10\textwidth]{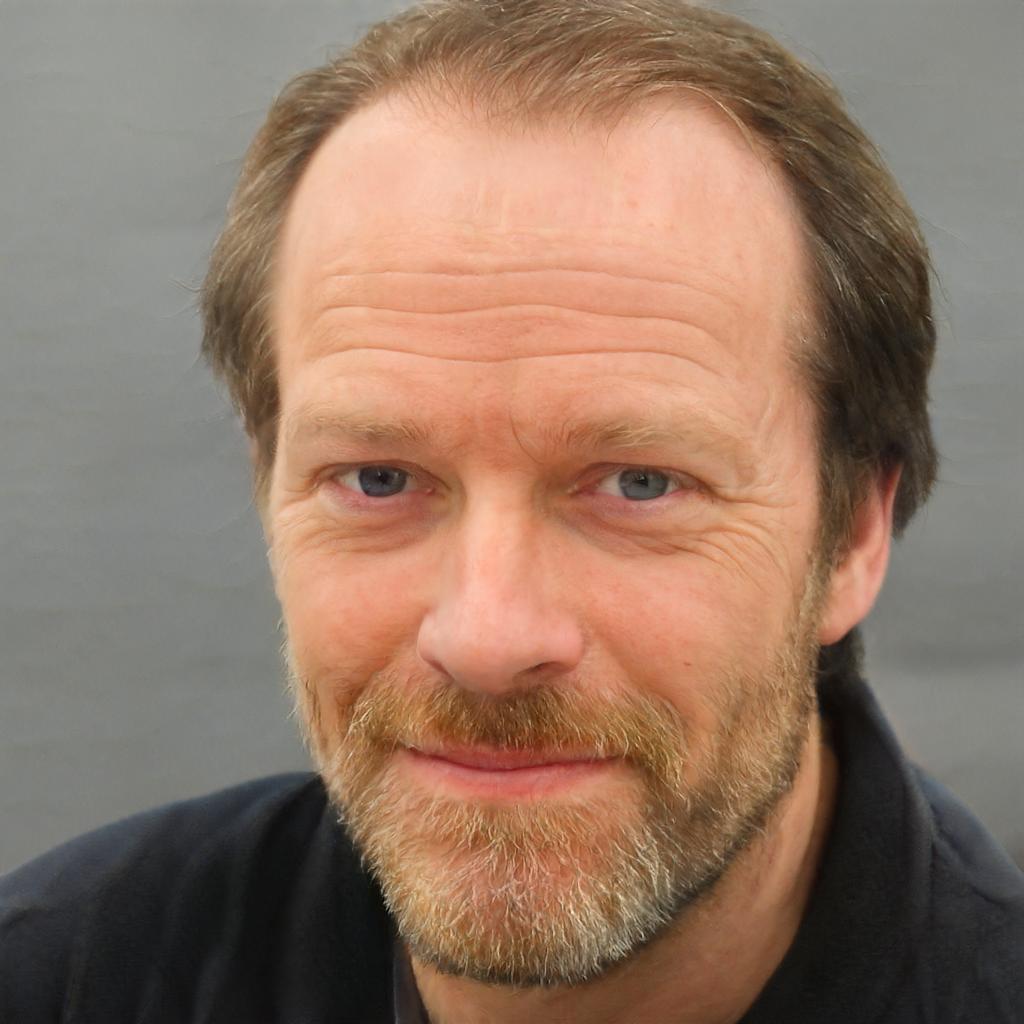}  
		\end{tabular}
	}
	\caption{A variety of edits along global text-driven manipulation directions, demonstrated on portraits of celebrities. Edits are performed using StyleGAN2 pretrained on FFHQ \cite{karras2019style}. The inputs are real images, embedded in $\mathcal{W+}$ space using the e4e encoder~\cite{tov2021designing}.
	The target attribute used in the text prompt is indicated above each column.}
	\label{fig:alex-faces}
\end{figure*}

\begin{figure*}[tb]
	\centering
	\setlength{\tabcolsep}{1.1pt}
	
	{\footnotesize
		\begin{tabular}{ccccccccccc}
			Input & Jeep & Sports & From Sixties & Classic &  & Input & Happy & Big Eyes & Golden Fur & Bulldog \\
			\includegraphics[width=0.09\textwidth]{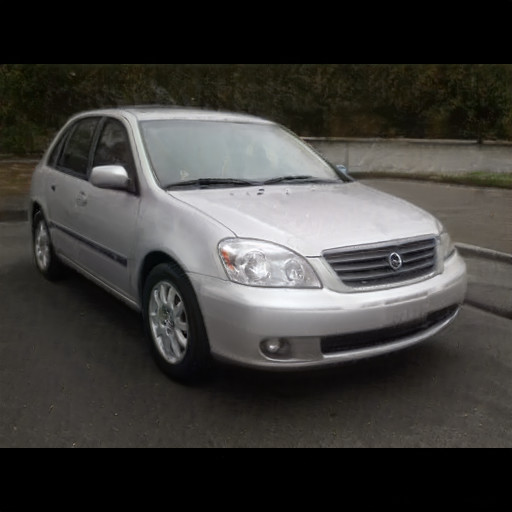} &
			\includegraphics[width=0.09\textwidth]{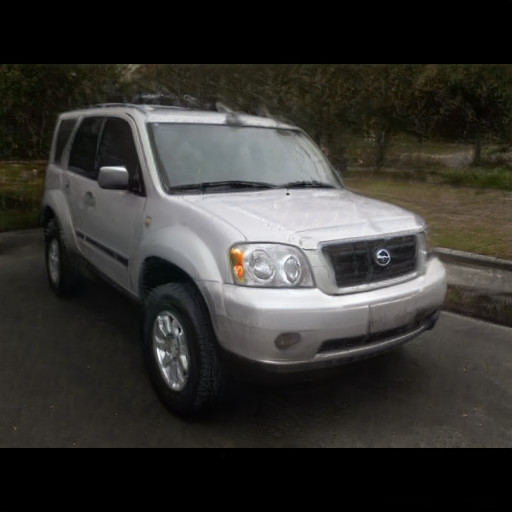} &
			\includegraphics[width=0.09\textwidth]{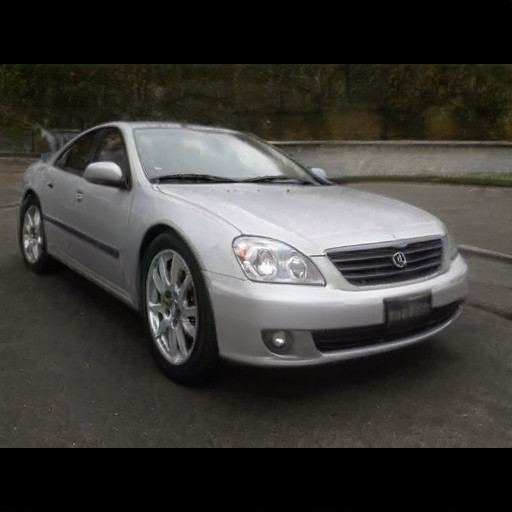} &
			\includegraphics[width=0.09\textwidth]{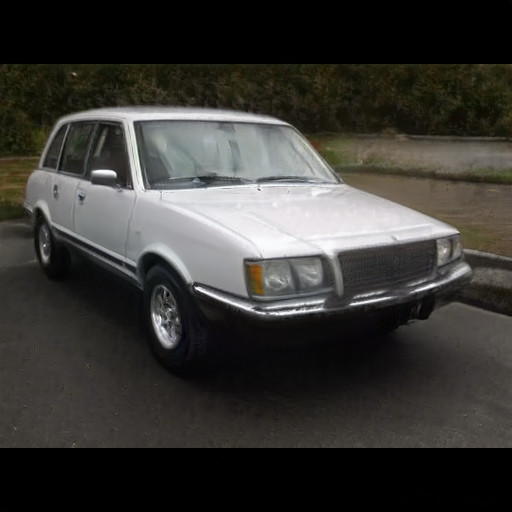} &
			\includegraphics[width=0.09\textwidth]{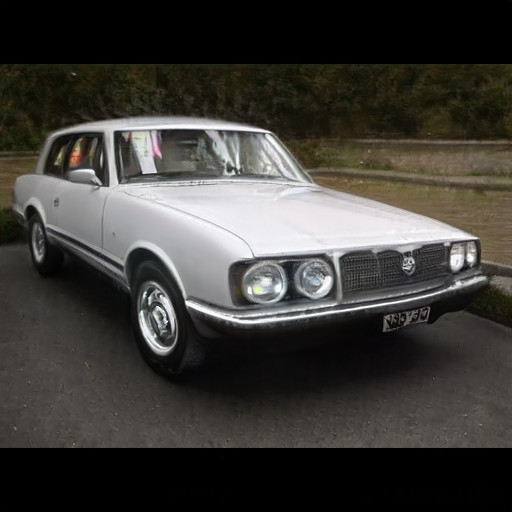} &
			\hspace{0.5cm} &
			\includegraphics[width=0.09\textwidth]{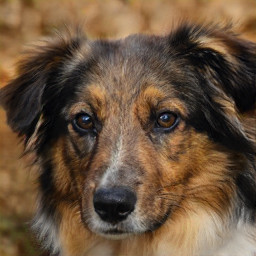} &
			\includegraphics[width=0.09\textwidth]{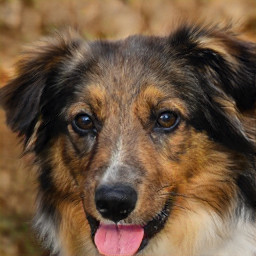} &
			\includegraphics[width=0.09\textwidth]{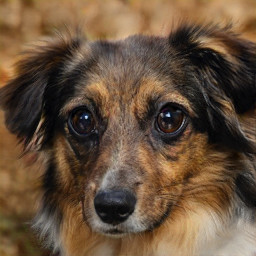} &
			\includegraphics[width=0.09\textwidth]{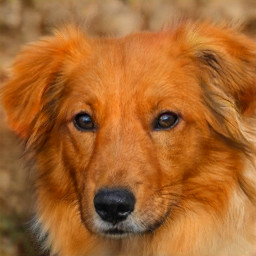} &
			\includegraphics[width=0.09\textwidth]{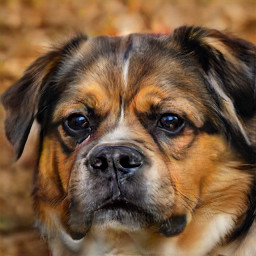} 
			\\
			\includegraphics[width=0.09\textwidth]{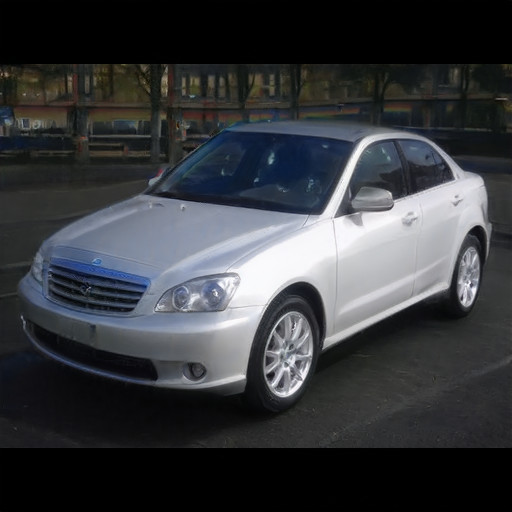} &
			\includegraphics[width=0.09\textwidth]{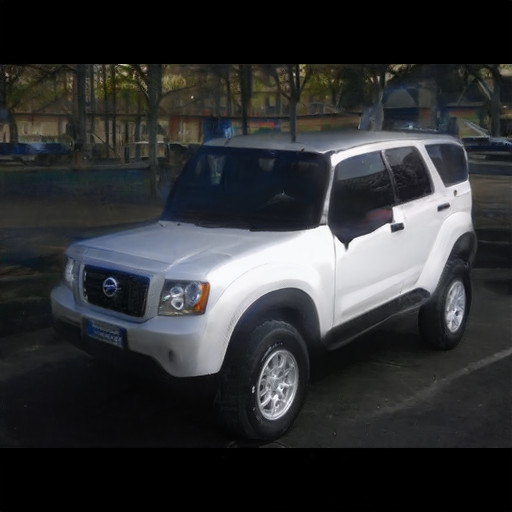} &
			\includegraphics[width=0.09\textwidth]{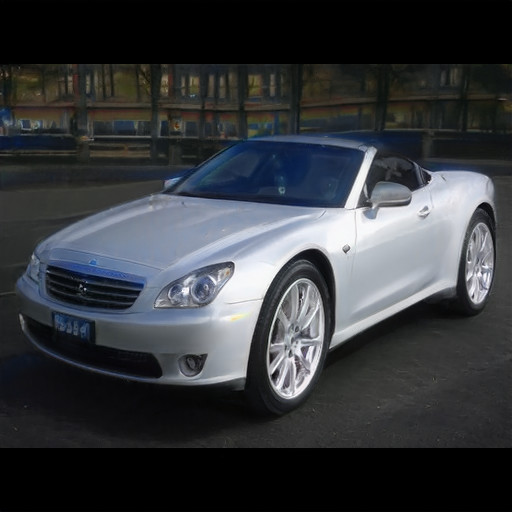} &
			\includegraphics[width=0.09\textwidth]{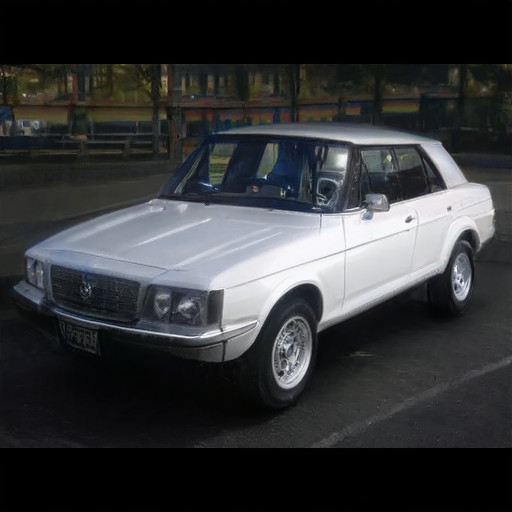} &
			\includegraphics[width=0.09\textwidth]{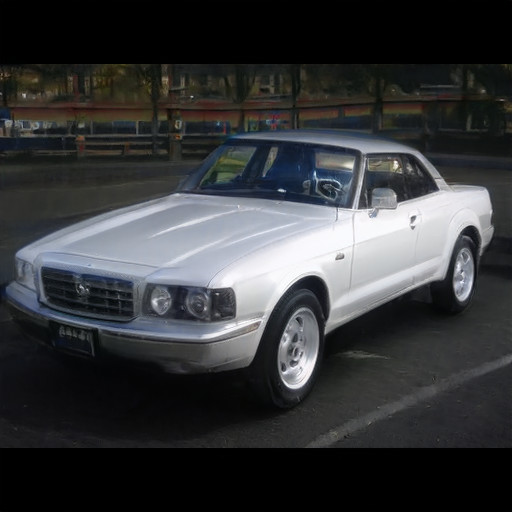} &
			\hspace{0.3cm} &
			\includegraphics[width=0.09\textwidth]{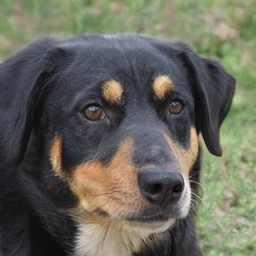} &
			\includegraphics[width=0.09\textwidth]{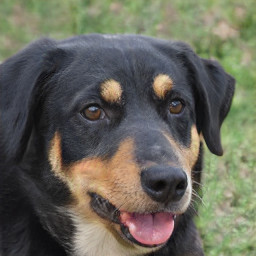} &
			\includegraphics[width=0.09\textwidth]{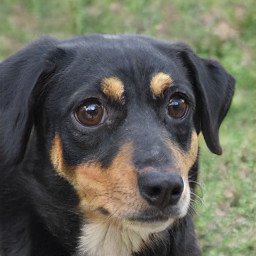} &
			\includegraphics[width=0.09\textwidth]{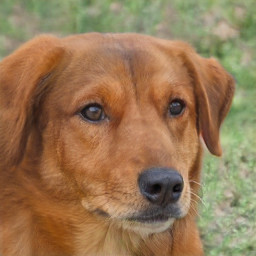} &
			\includegraphics[width=0.09\textwidth]{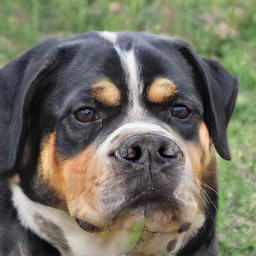} 
			\\
			\includegraphics[width=0.09\textwidth]{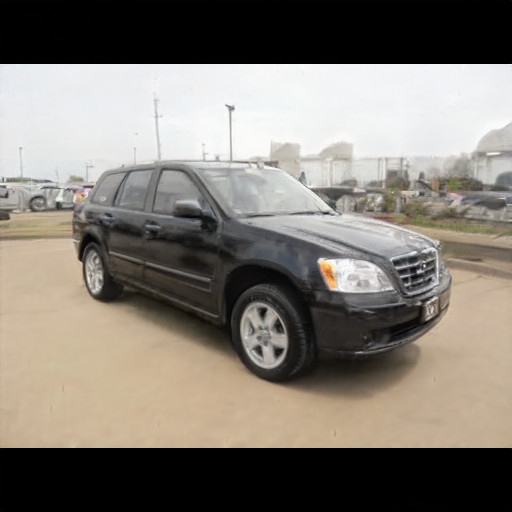} &
			\includegraphics[width=0.09\textwidth]{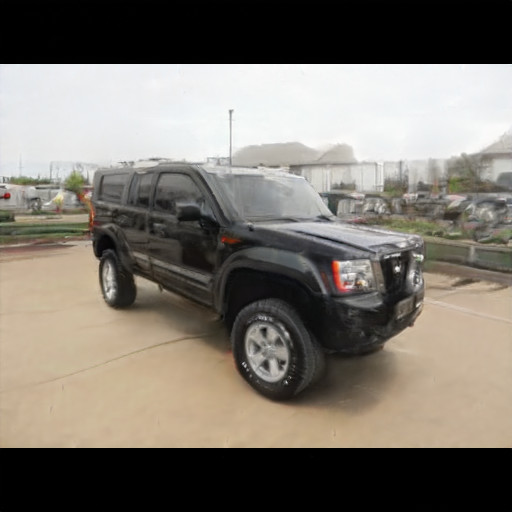} &
			\includegraphics[width=0.09\textwidth]{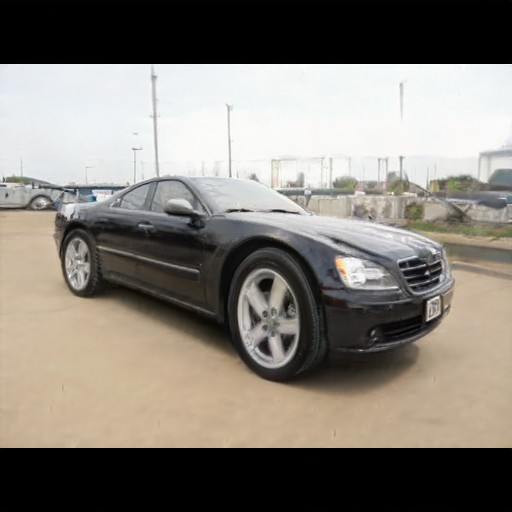} &
			\includegraphics[width=0.09\textwidth]{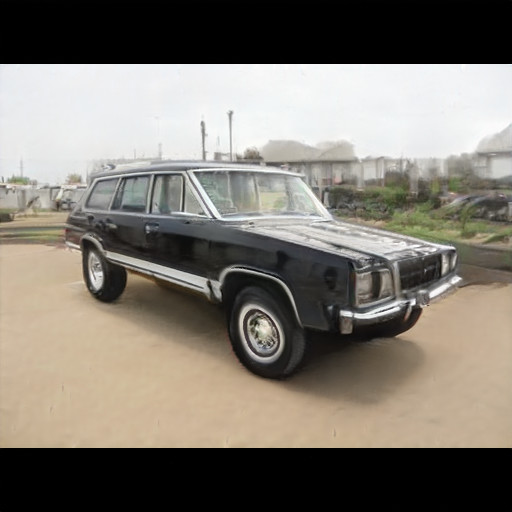} &
			\includegraphics[width=0.09\textwidth]{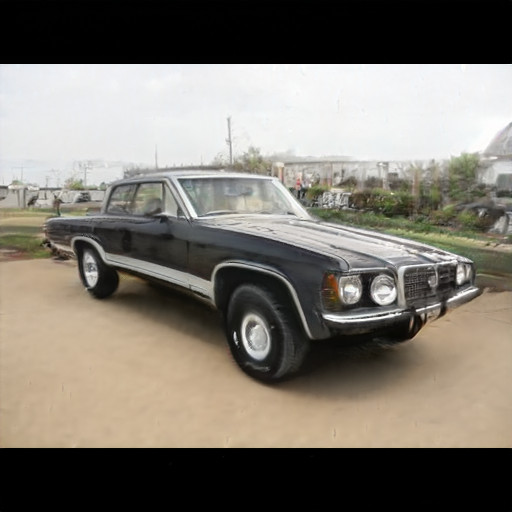} &
			\hspace{0.5cm} &
			\includegraphics[width=0.09\textwidth]{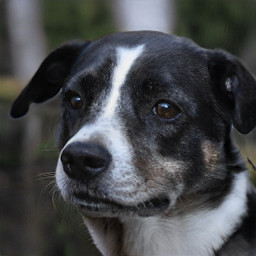} &
			\includegraphics[width=0.09\textwidth]{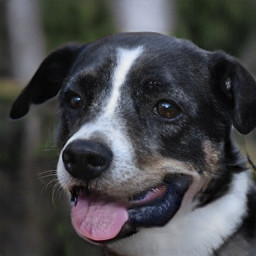} &
			\includegraphics[width=0.09\textwidth]{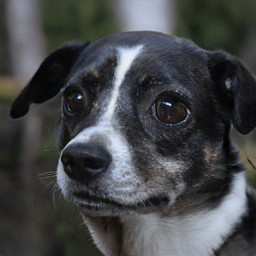} &
			\includegraphics[width=0.09\textwidth]{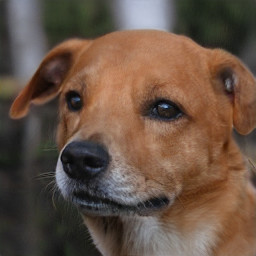} &
			\includegraphics[width=0.09\textwidth]{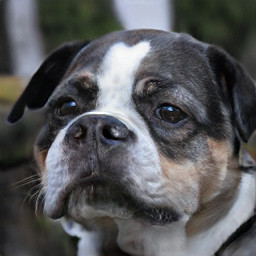} 
			\\

		\end{tabular}
	}
	\caption{A variety of edits along global text-driven manipulation directions. Left: using StyleGAN2 pretrained on LSUN cars~\cite{yu2015lsun}. Right: using StyleGAN2-ada~\cite{karras2020training} pretrained on AFHQ dogs~\cite{choi2020stargan}. The target attribute used in the text prompt is indicated above each column.}
	\label{fig:alex-nonfaces}
\end{figure*}

Figures~\ref{fig:alex-faces} and~\ref{fig:alex-nonfaces} show a variety of edits along text-driven manipulation directions determined as described above on images of faces, cars, and dogs.
The manipulations in Figure~\ref{fig:alex-faces} are performed using StyleGAN2 pretrained on FFHQ \cite{karras2019style}. The inputs are real images, embedded in $\mathcal{W+}$ space using the e4e encoder~\cite{tov2021designing}. The figure demonstrates text-driven manipulations of 18 attributes, including complex concepts, such as facial expressions and hair styles.
The manipulations in Figure~\ref{fig:alex-nonfaces} use StyleGAN2 pretrained on LSUN cars~\cite{yu2015lsun} (on real images) and on generated images from StyleGAN2-ada~\cite{karras2020training} pretrained on AFHQ dogs~\cite{choi2020stargan}.

\begin{figure*}[]
	\setlength{\tabcolsep}{1pt}
	\begin{center}
	{\footnotesize
	\begin{tabular}{cccccccccccc}
		  Input & TediGAN & Global & Mapper  & Input & TediGAN & Global & Mapper  & Input & TediGAN & Global & Mapper \\
	 
        
		\includegraphics[width=0.079\textwidth]{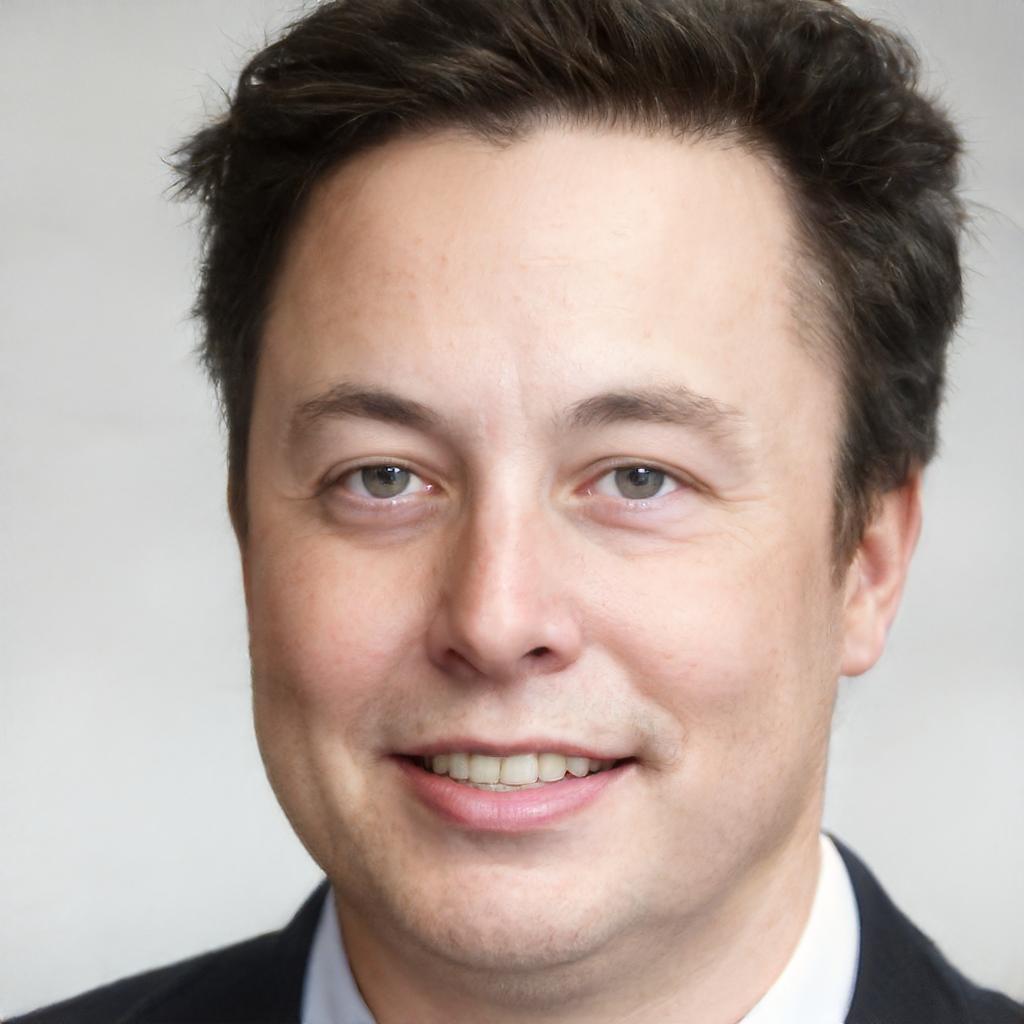} &
		\includegraphics[width=0.079\textwidth]{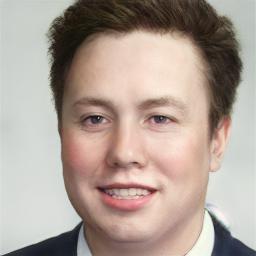} &
		\includegraphics[width=0.079\textwidth]{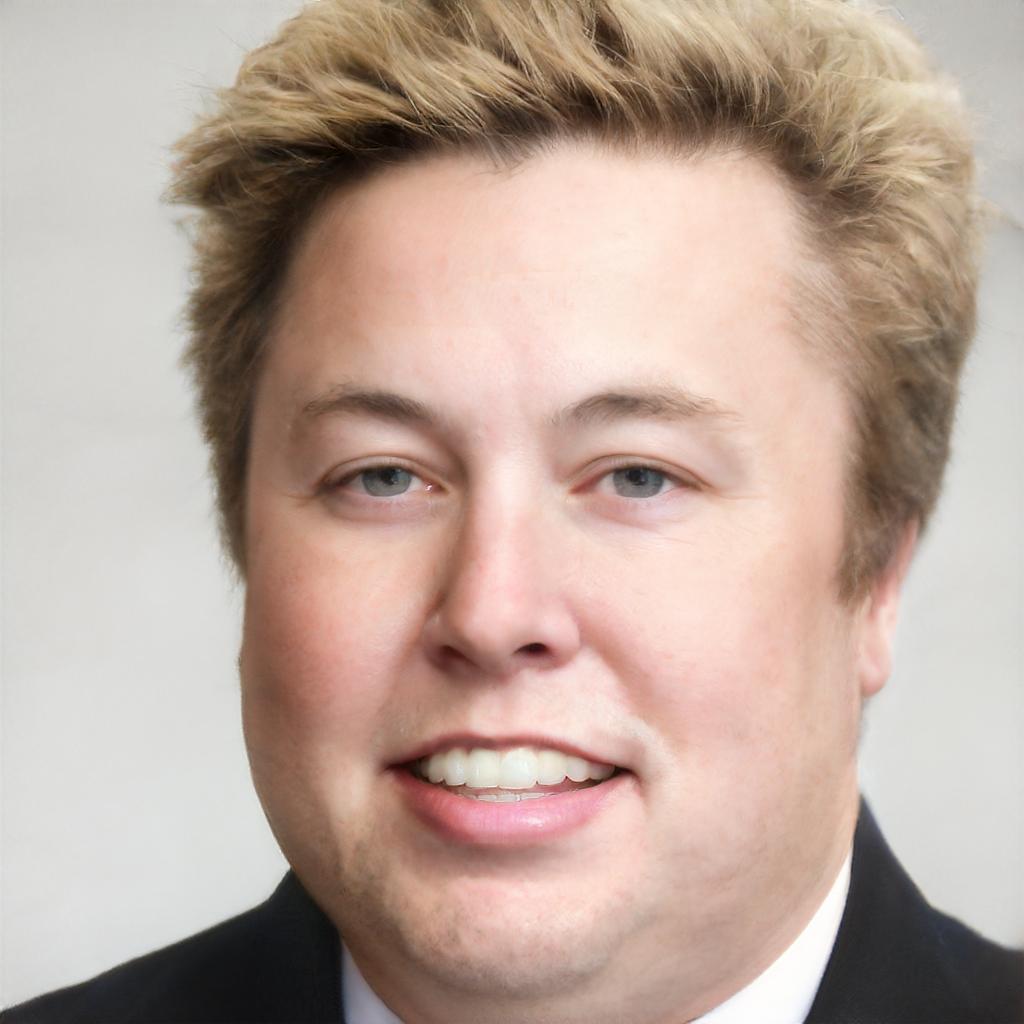} &
		\includegraphics[width=0.079\textwidth]{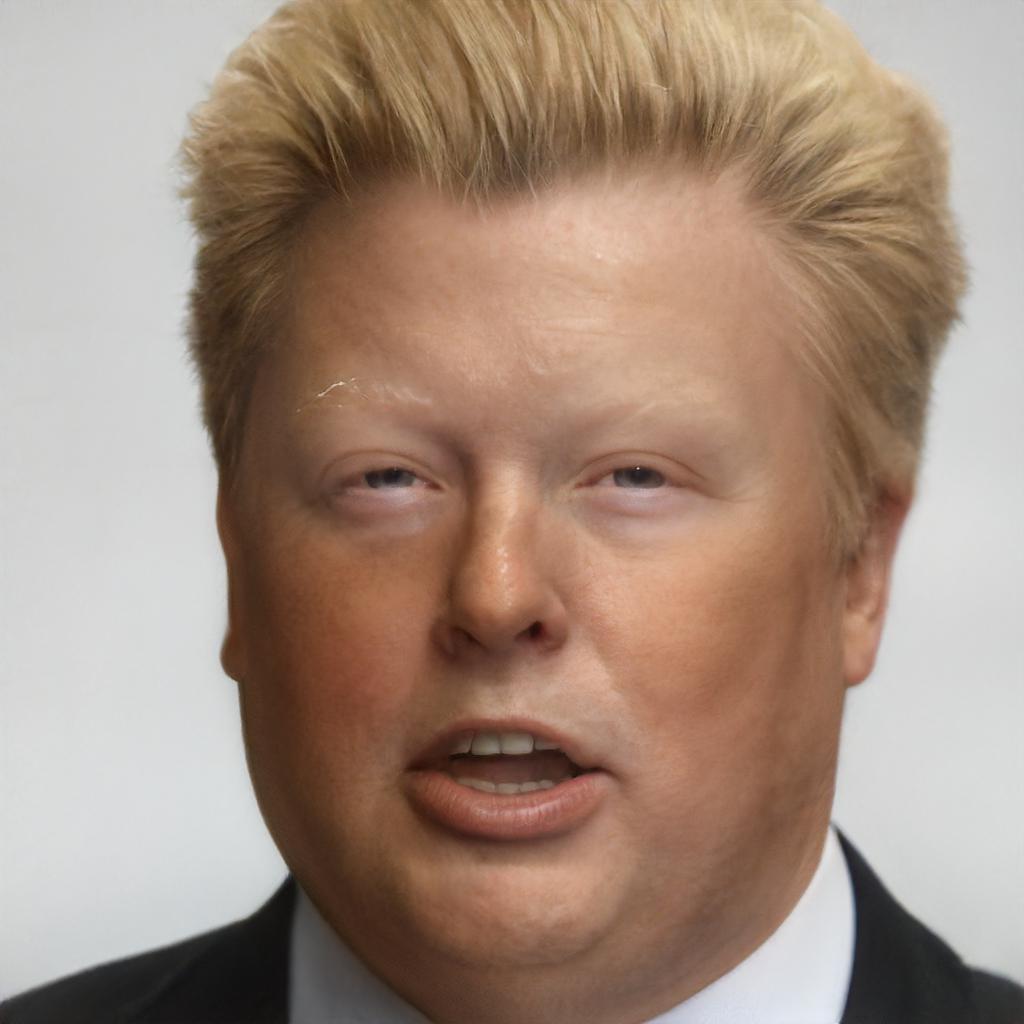} &
		
        
		\includegraphics[width=0.079\textwidth]{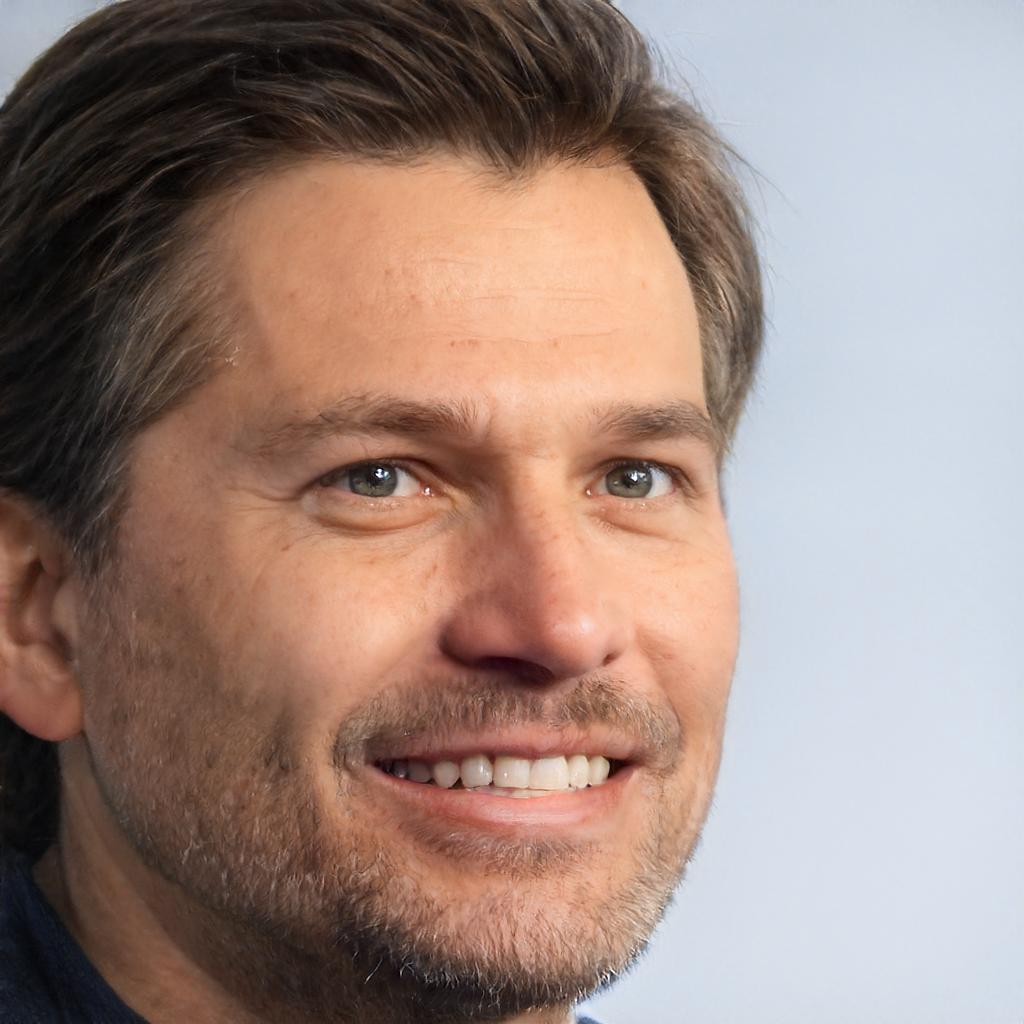} &
		\includegraphics[width=0.079\textwidth]{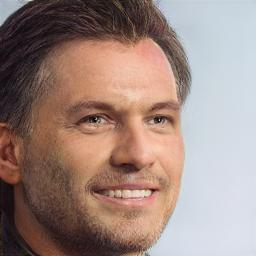} &
		\includegraphics[width=0.079\textwidth]{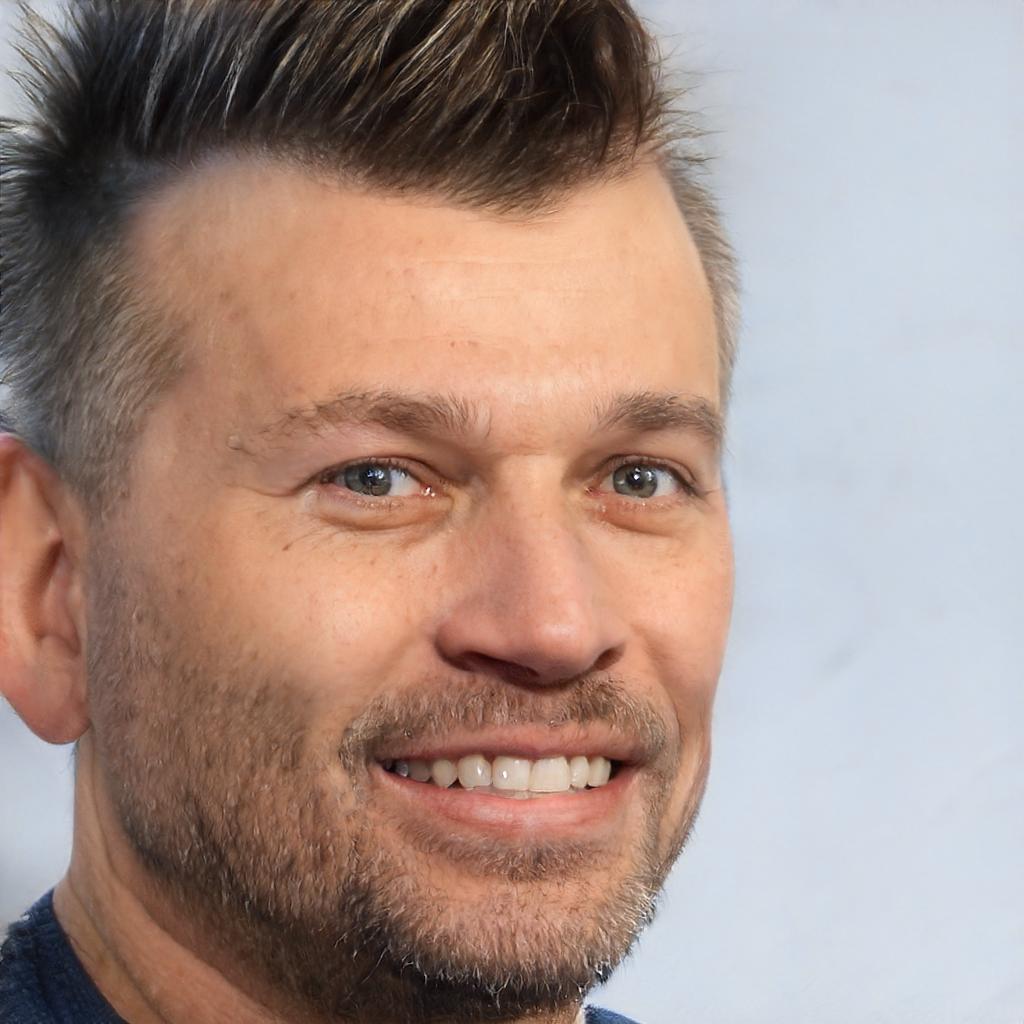} &
		\includegraphics[width=0.079\textwidth]{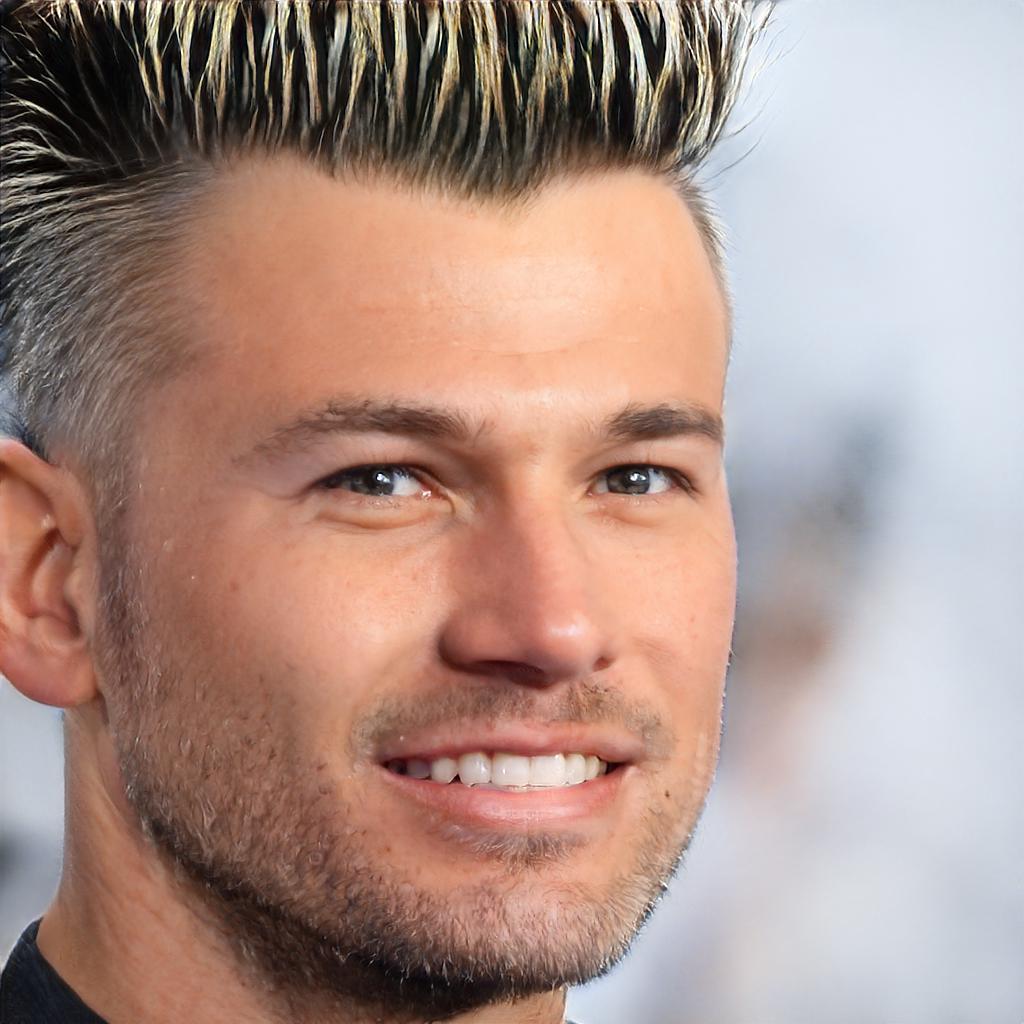} &
		
		
		\includegraphics[width=0.079\textwidth]{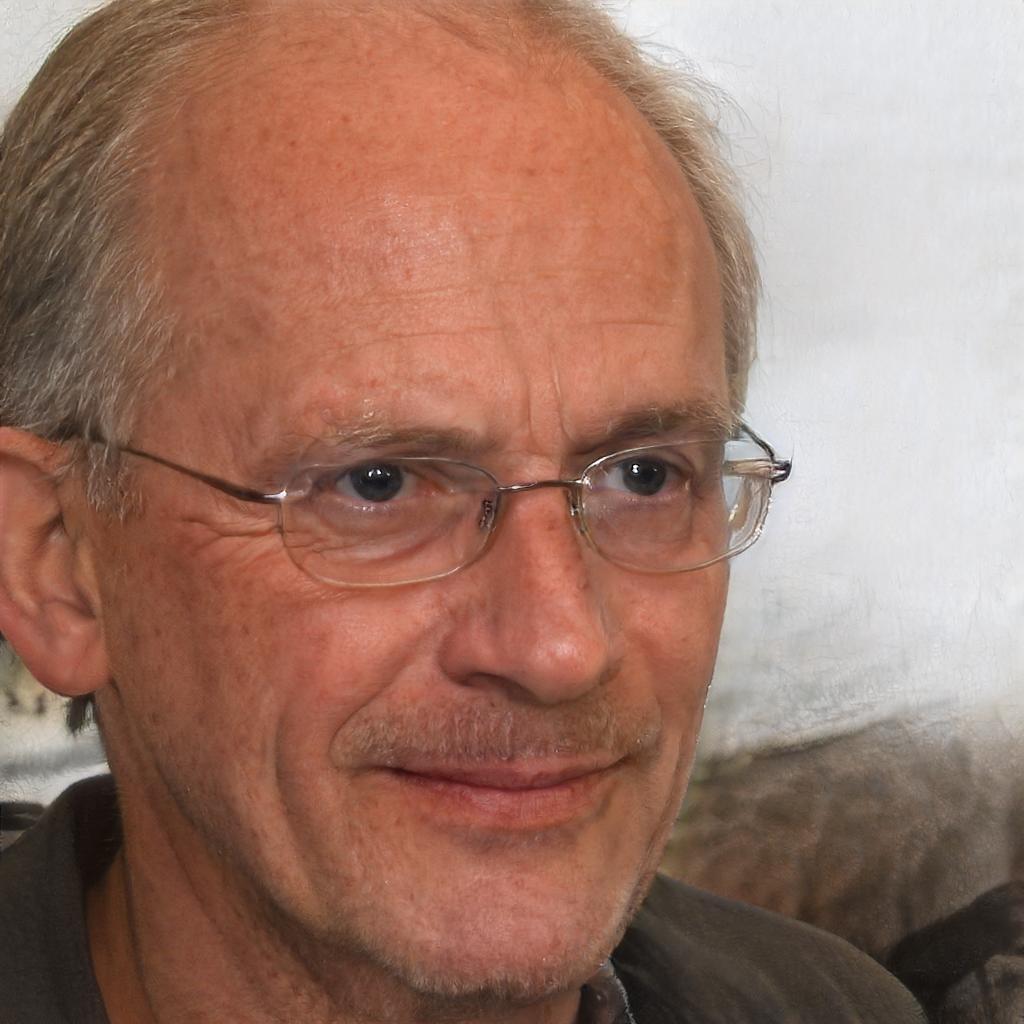} &
		\includegraphics[width=0.079\textwidth]{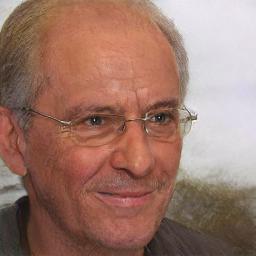} &
		\includegraphics[width=0.079\textwidth]{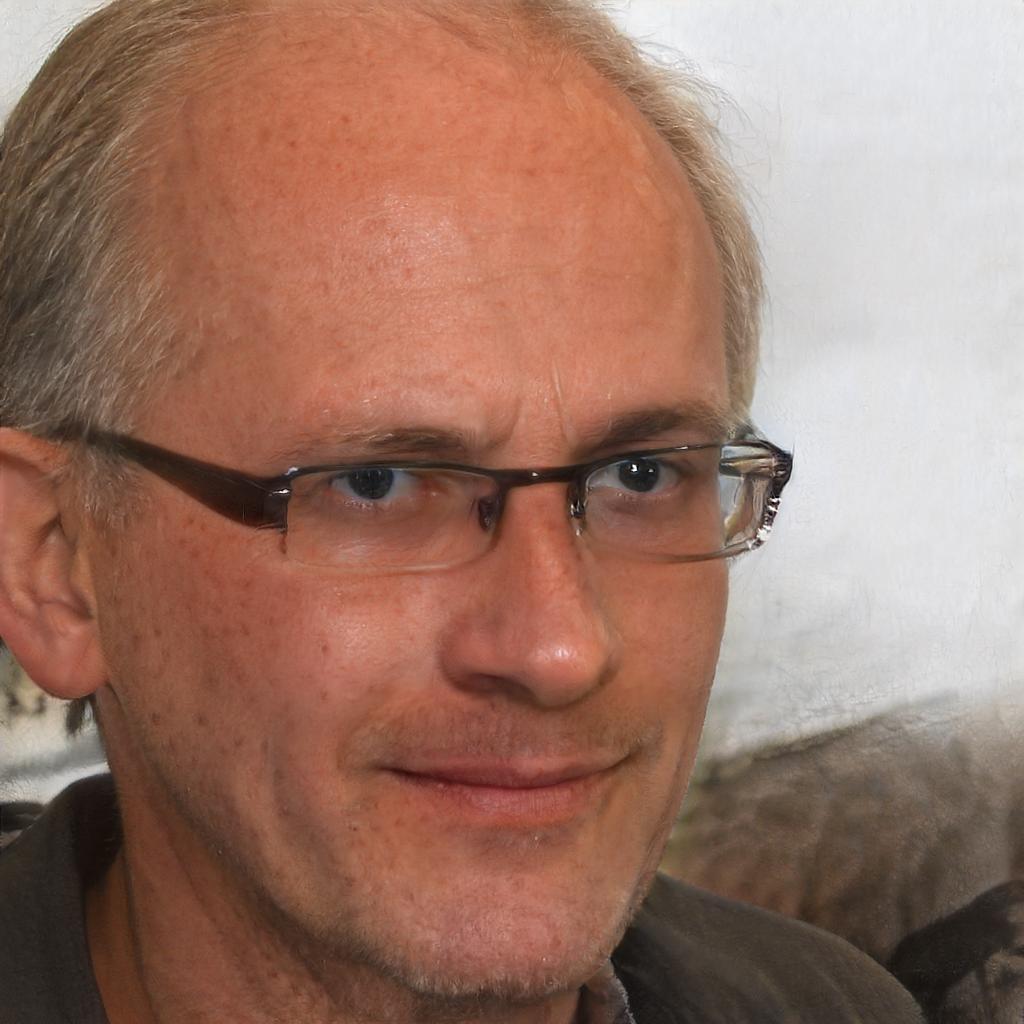} &
		\includegraphics[width=0.079\textwidth]{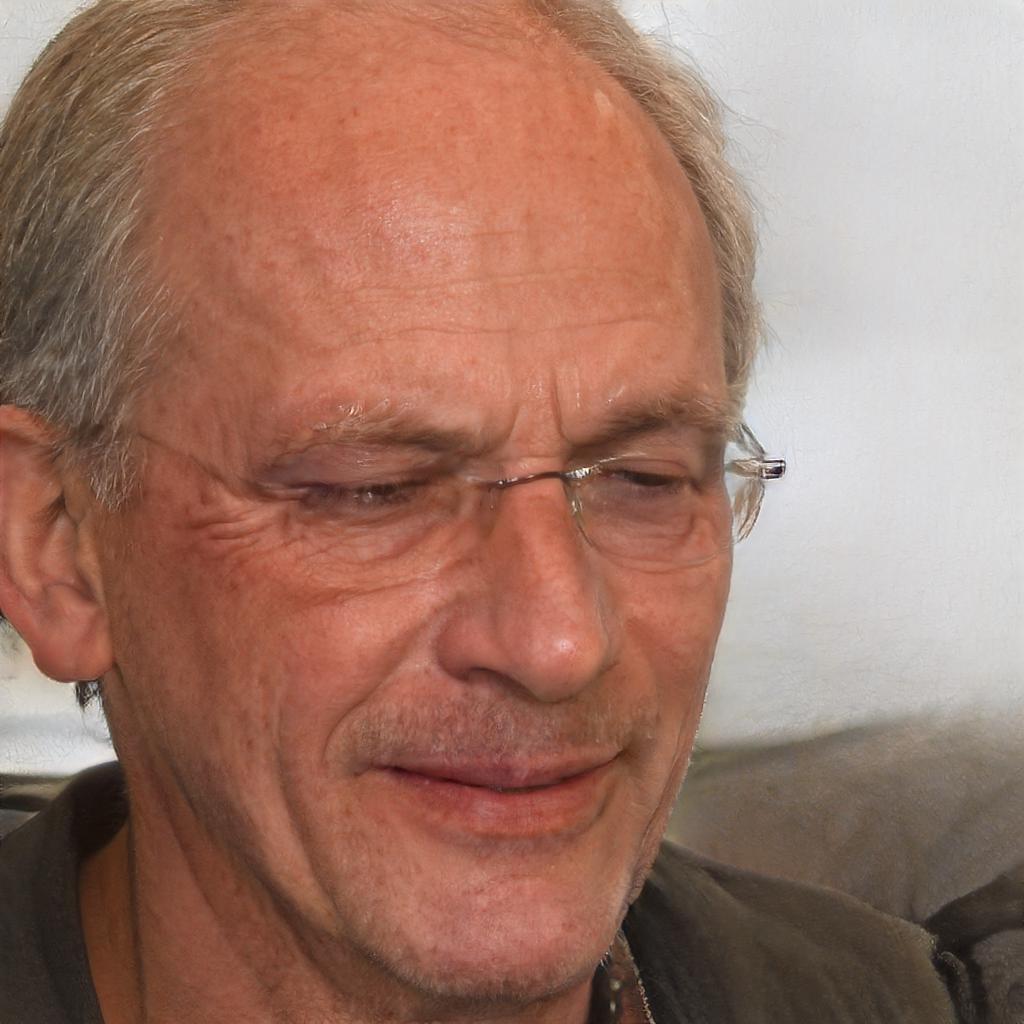} \\

		\includegraphics[width=0.079\textwidth]{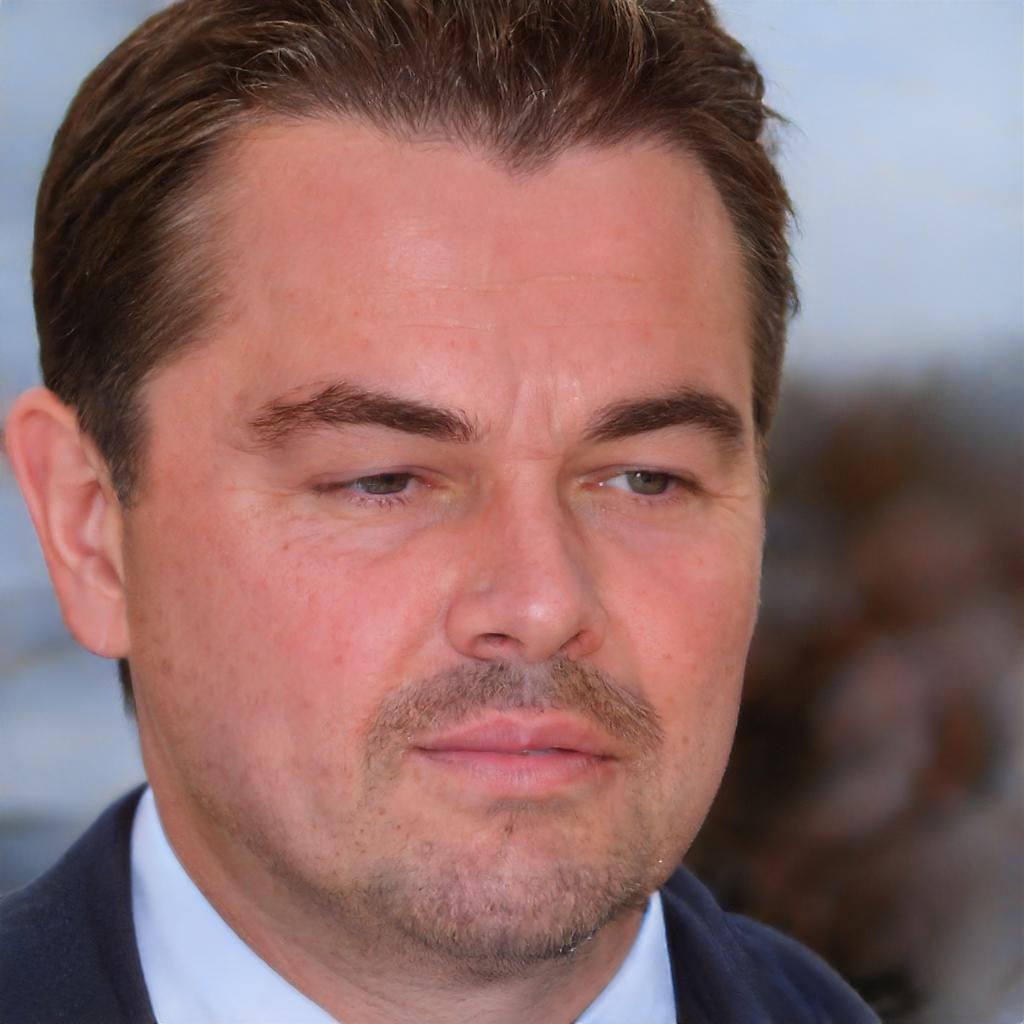} &
		\includegraphics[width=0.079\textwidth]{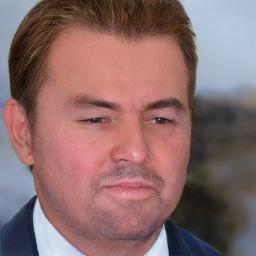} &
		\includegraphics[width=0.079\textwidth]{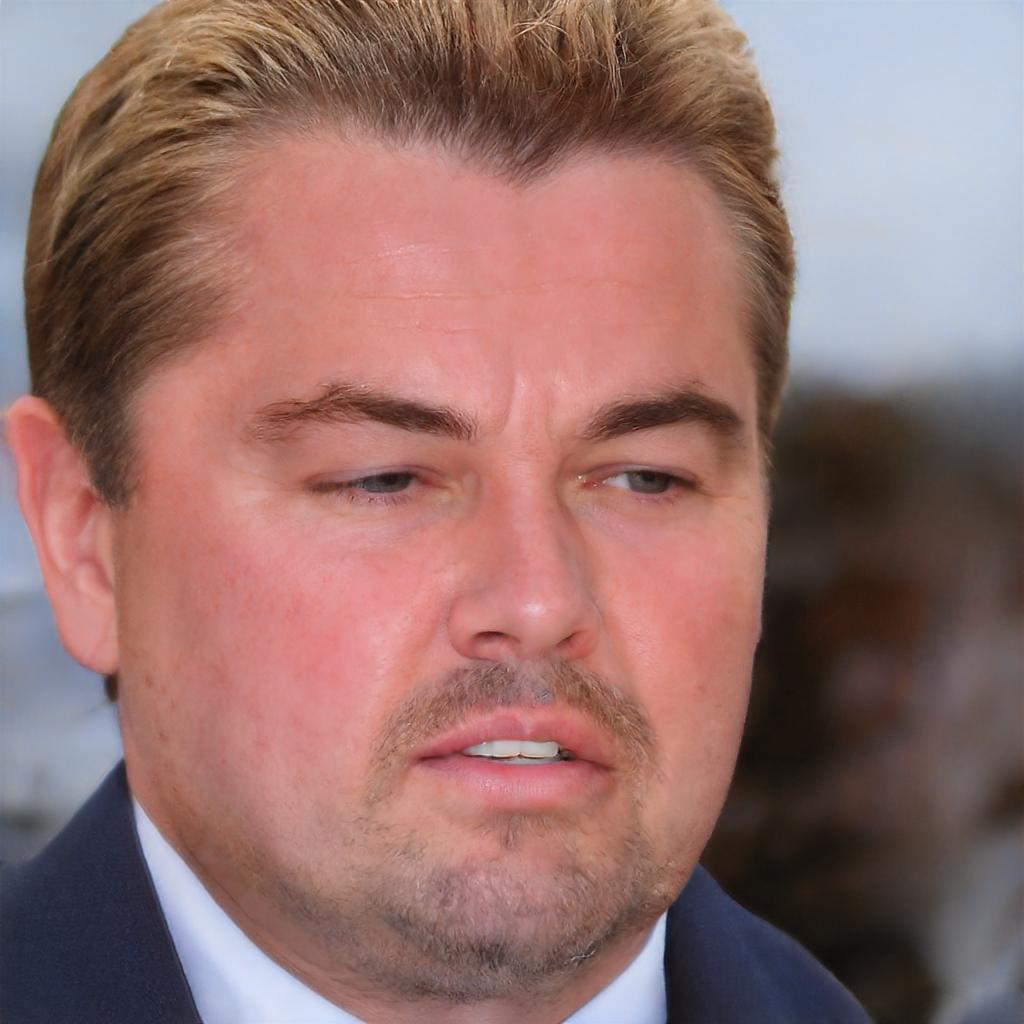} &
		\includegraphics[width=0.079\textwidth]{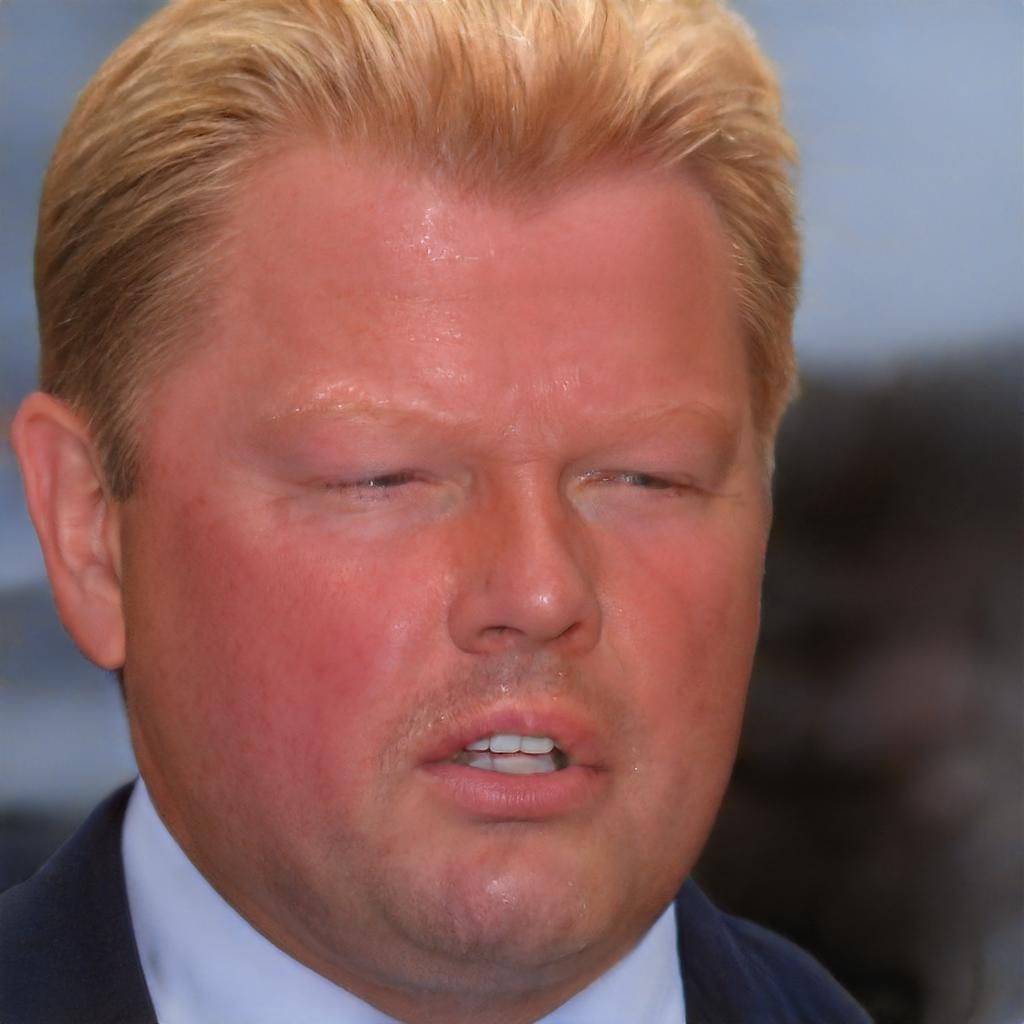} &
		
		\includegraphics[width=0.079\textwidth]{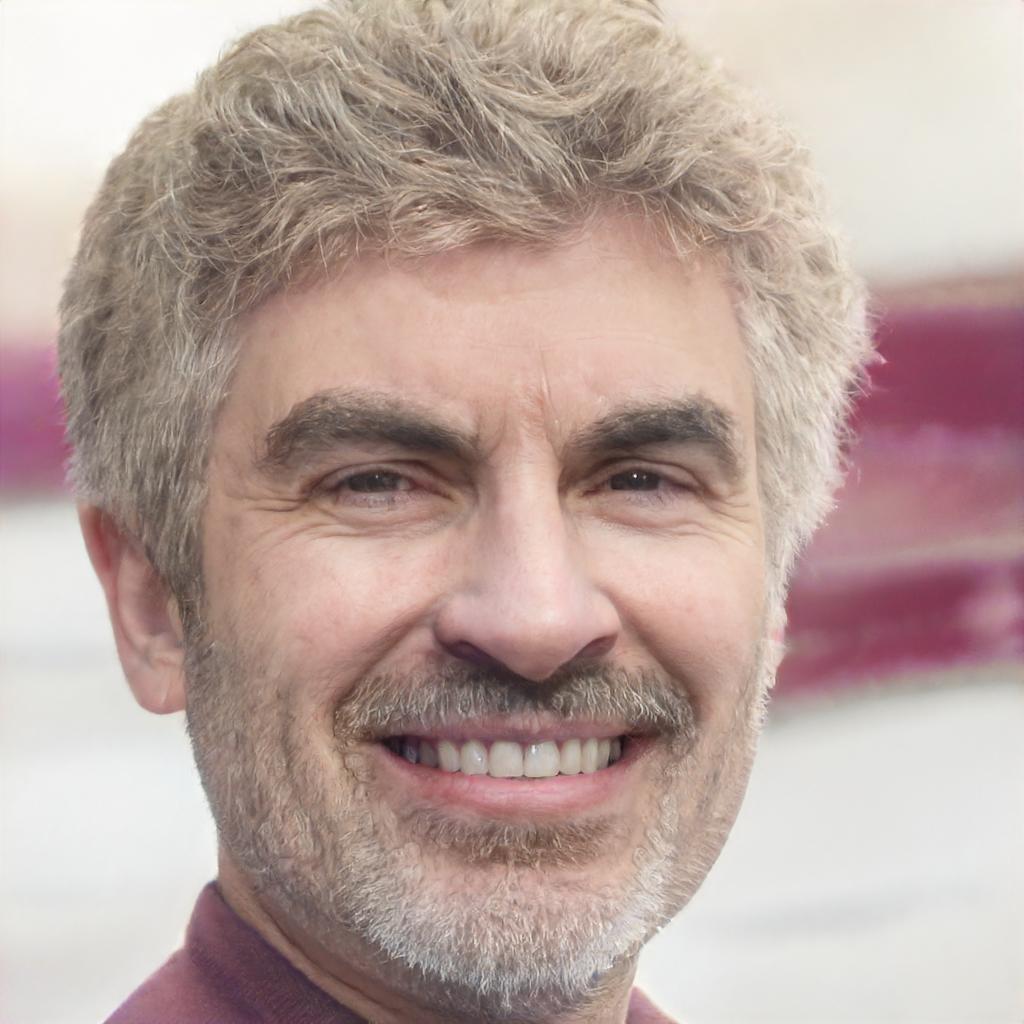} &
		\includegraphics[width=0.079\textwidth]{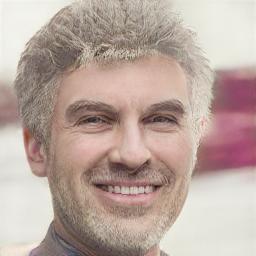} &
		\includegraphics[width=0.079\textwidth]{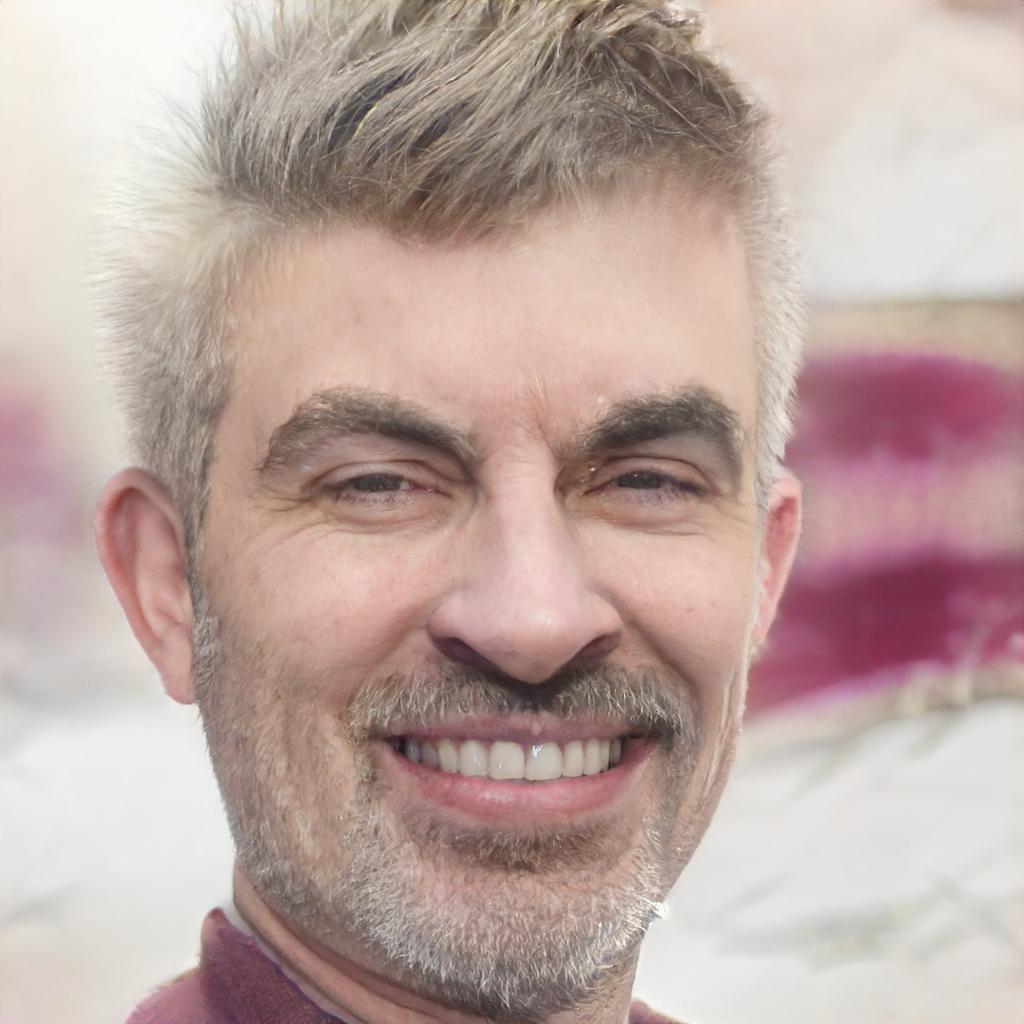} &
		\includegraphics[width=0.079\textwidth]{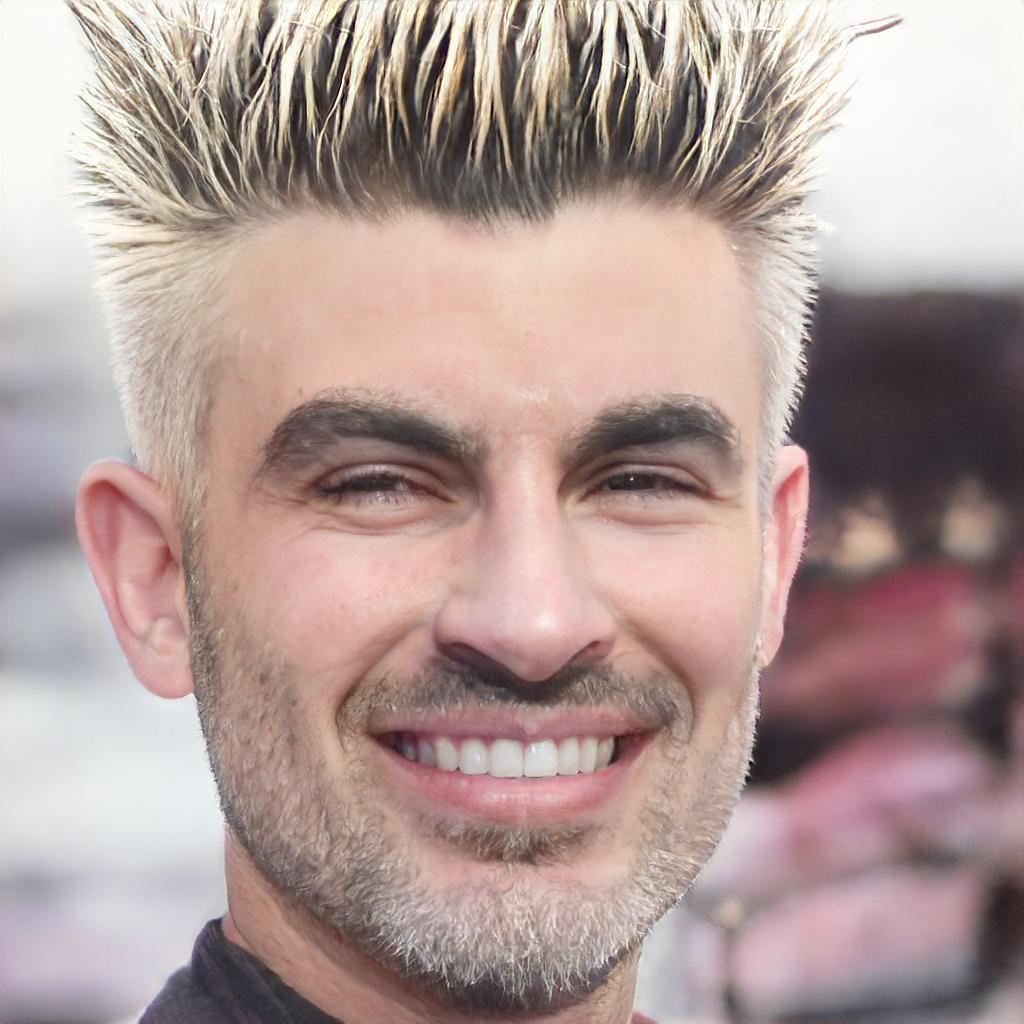} &
		
		\includegraphics[width=0.079\textwidth]{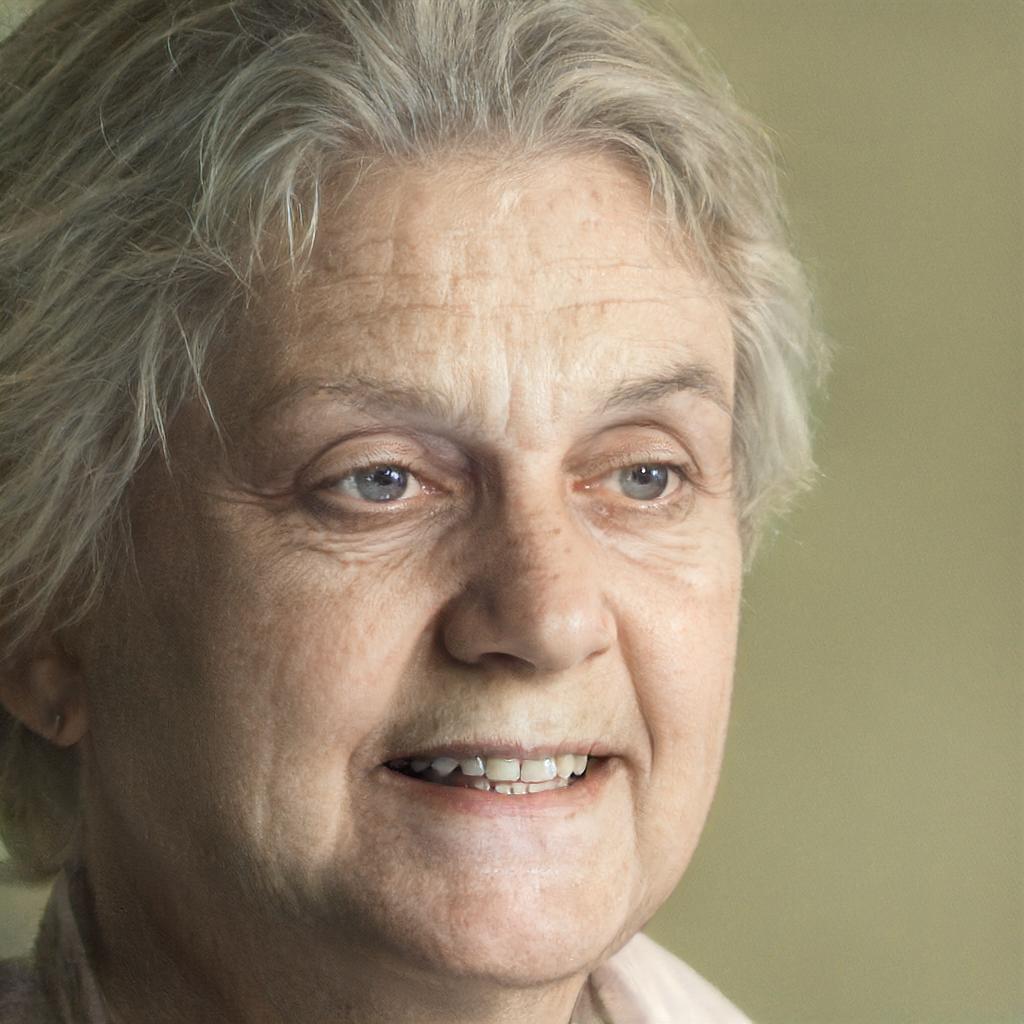} &
		\includegraphics[width=0.079\textwidth]{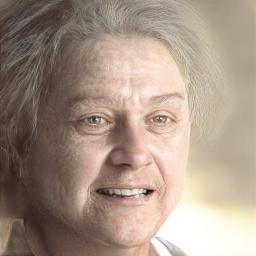} &
		\includegraphics[width=0.079\textwidth]{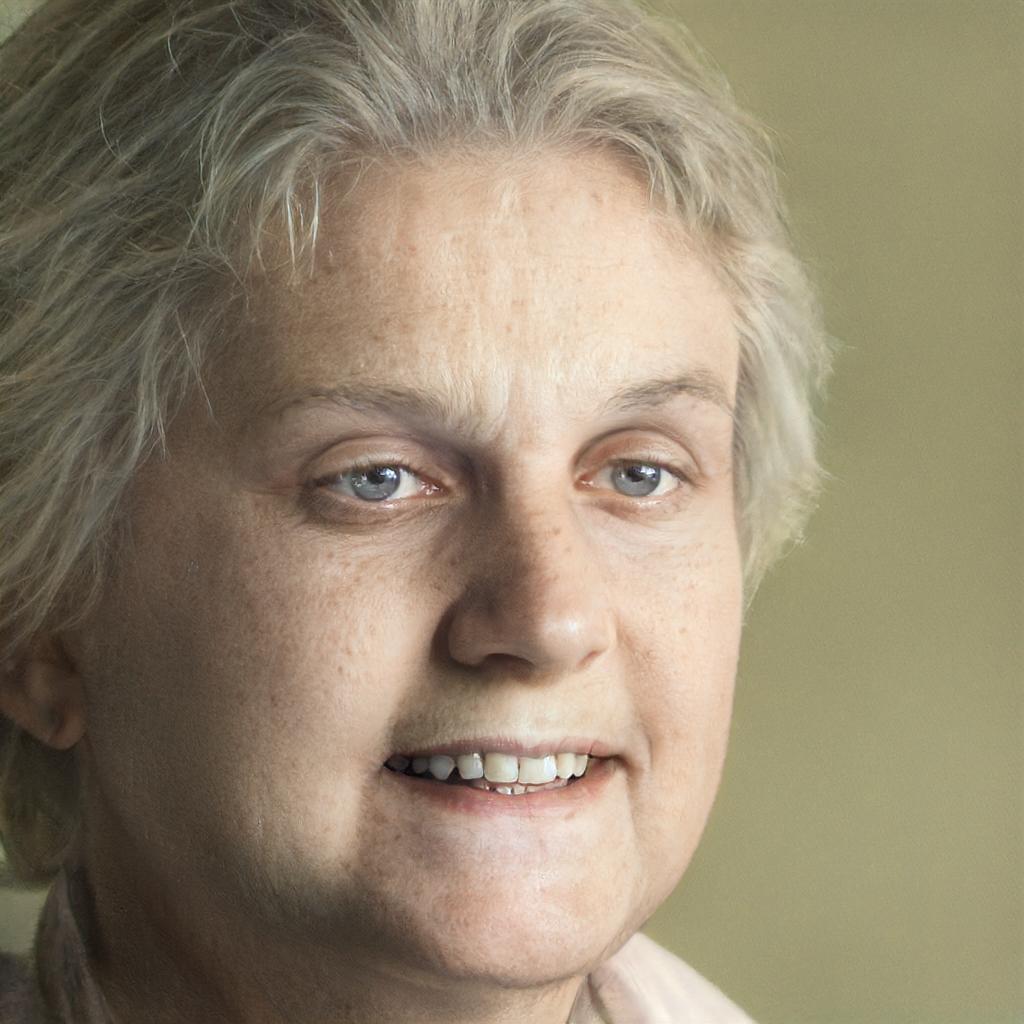} &
		\includegraphics[width=0.079\textwidth]{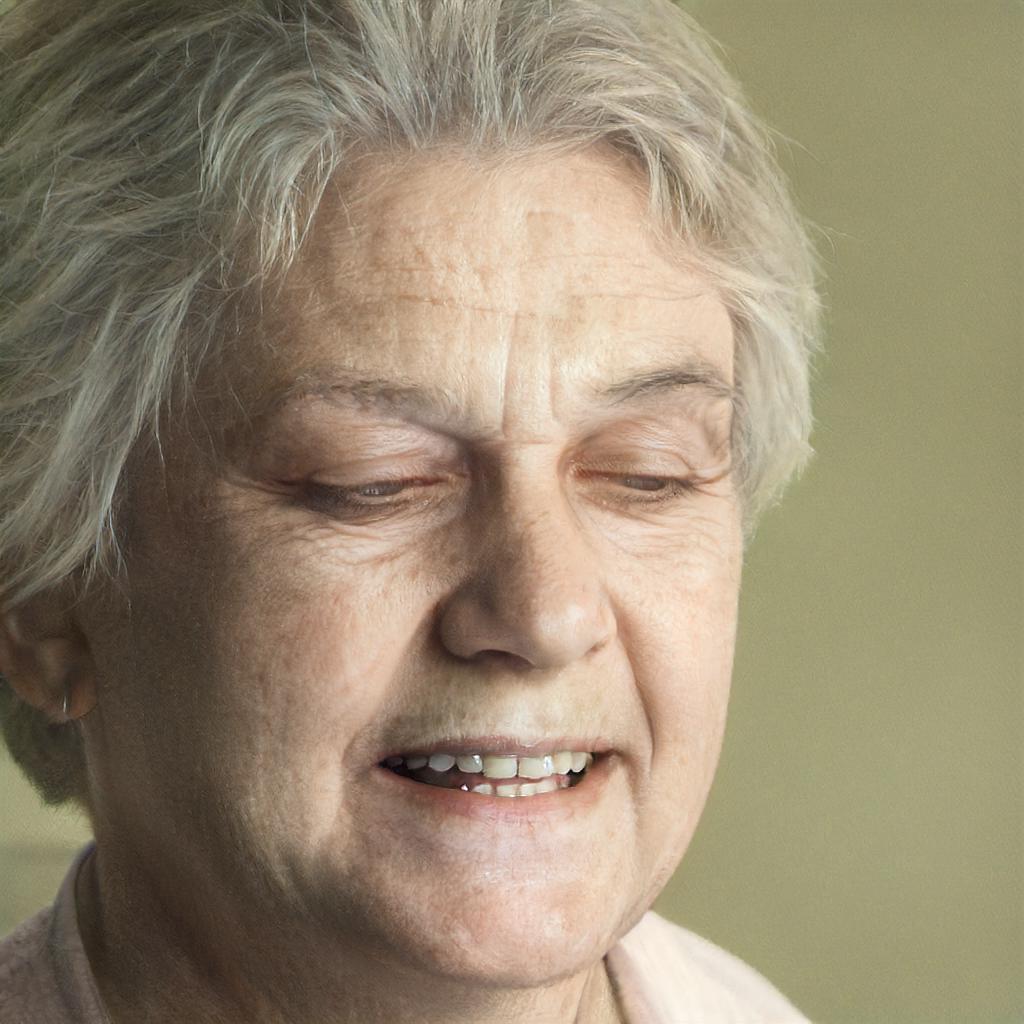} \\

		\includegraphics[width=0.079\textwidth]{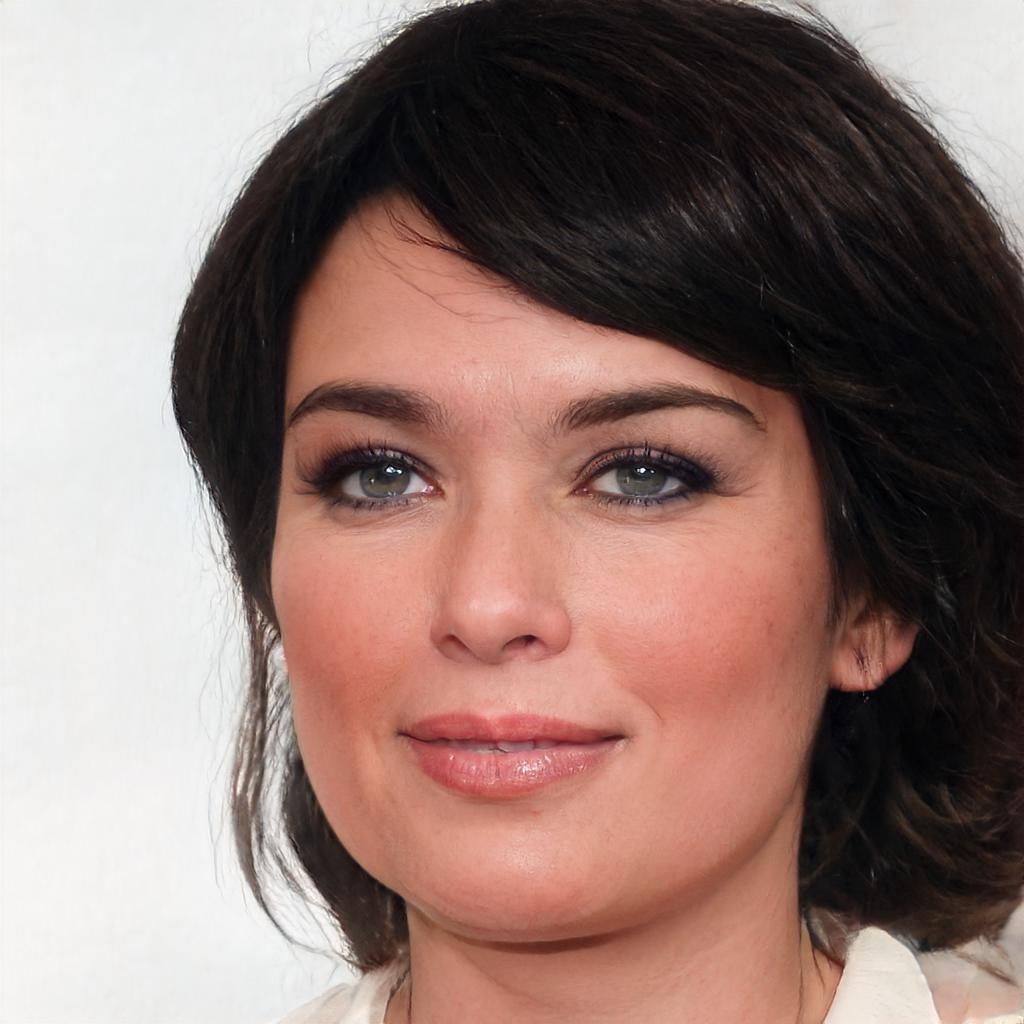} &
		\includegraphics[width=0.079\textwidth]{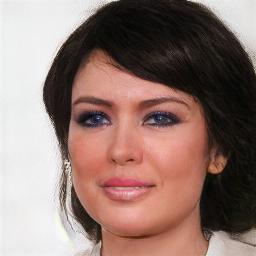} &
		\includegraphics[width=0.079\textwidth]{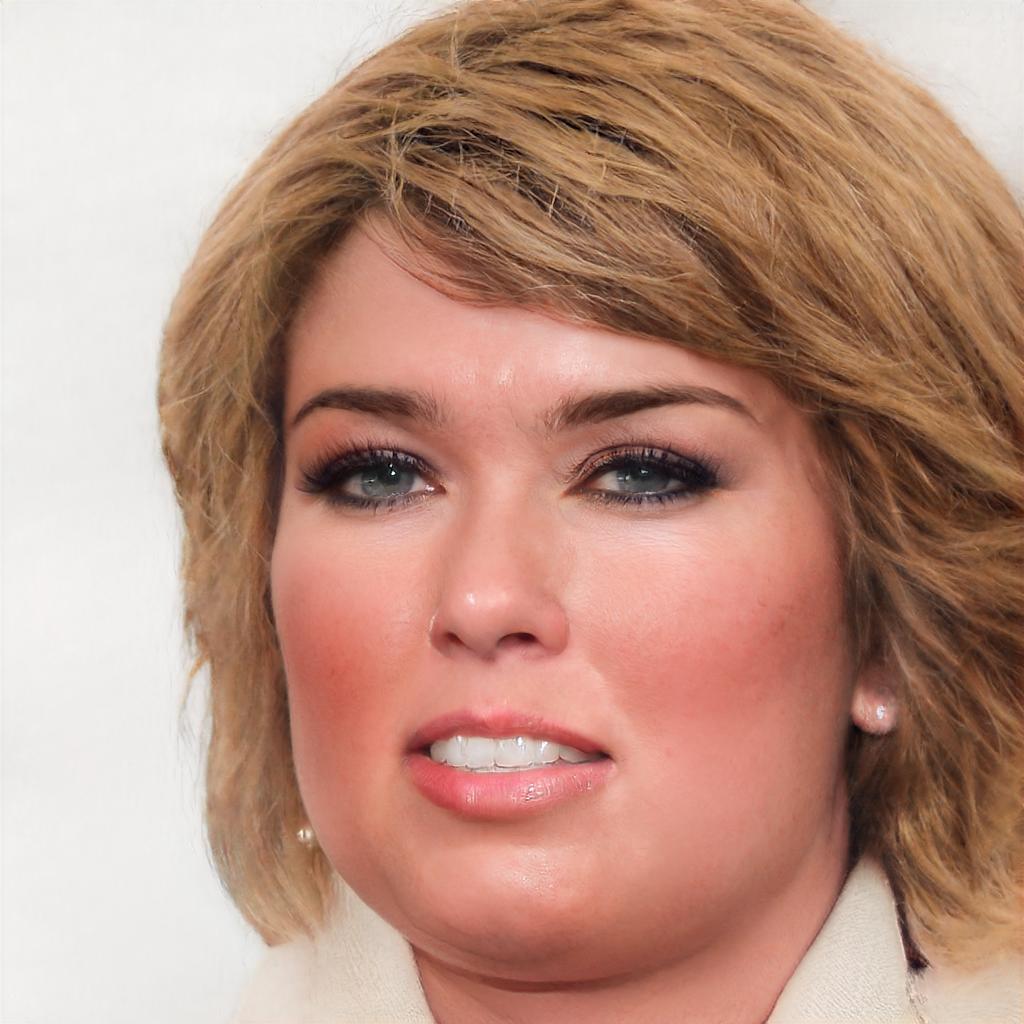} &
		\includegraphics[width=0.079\textwidth]{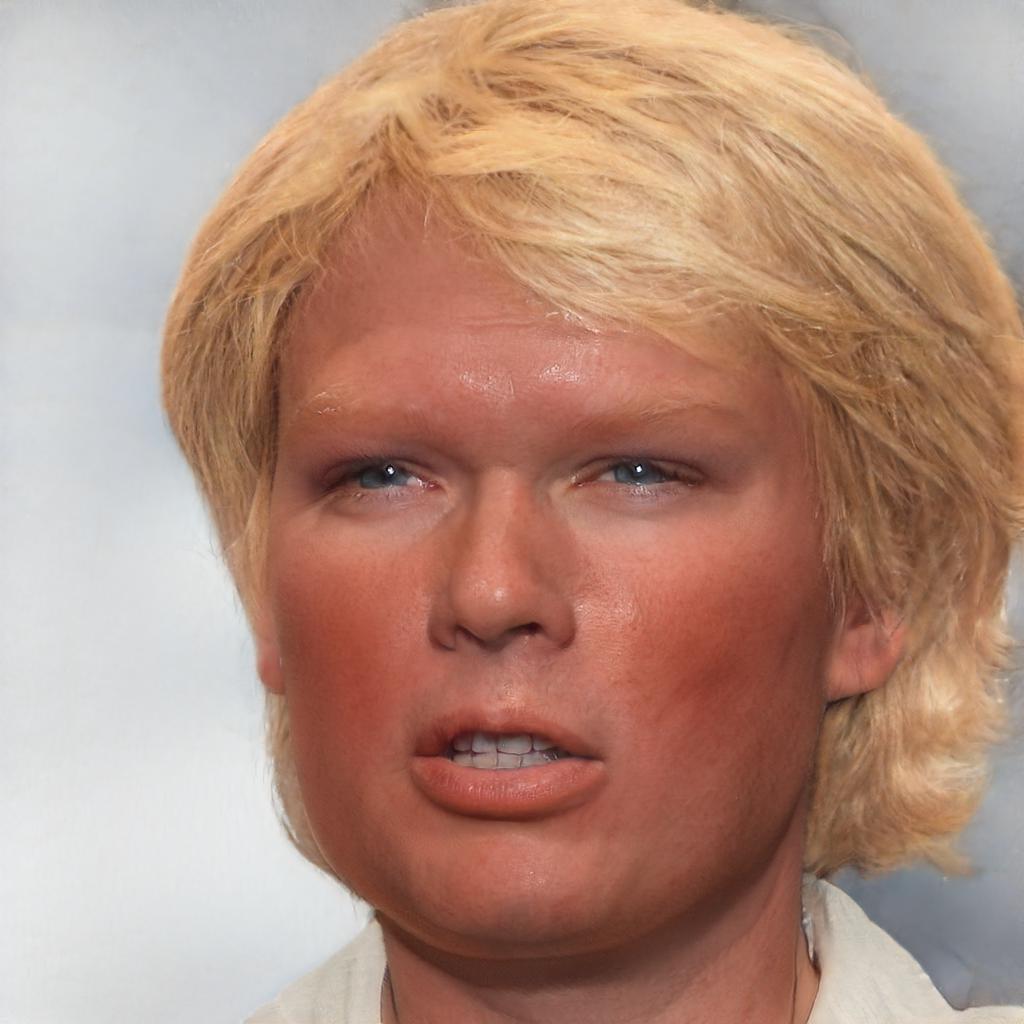} &
		
		\includegraphics[width=0.079\textwidth]{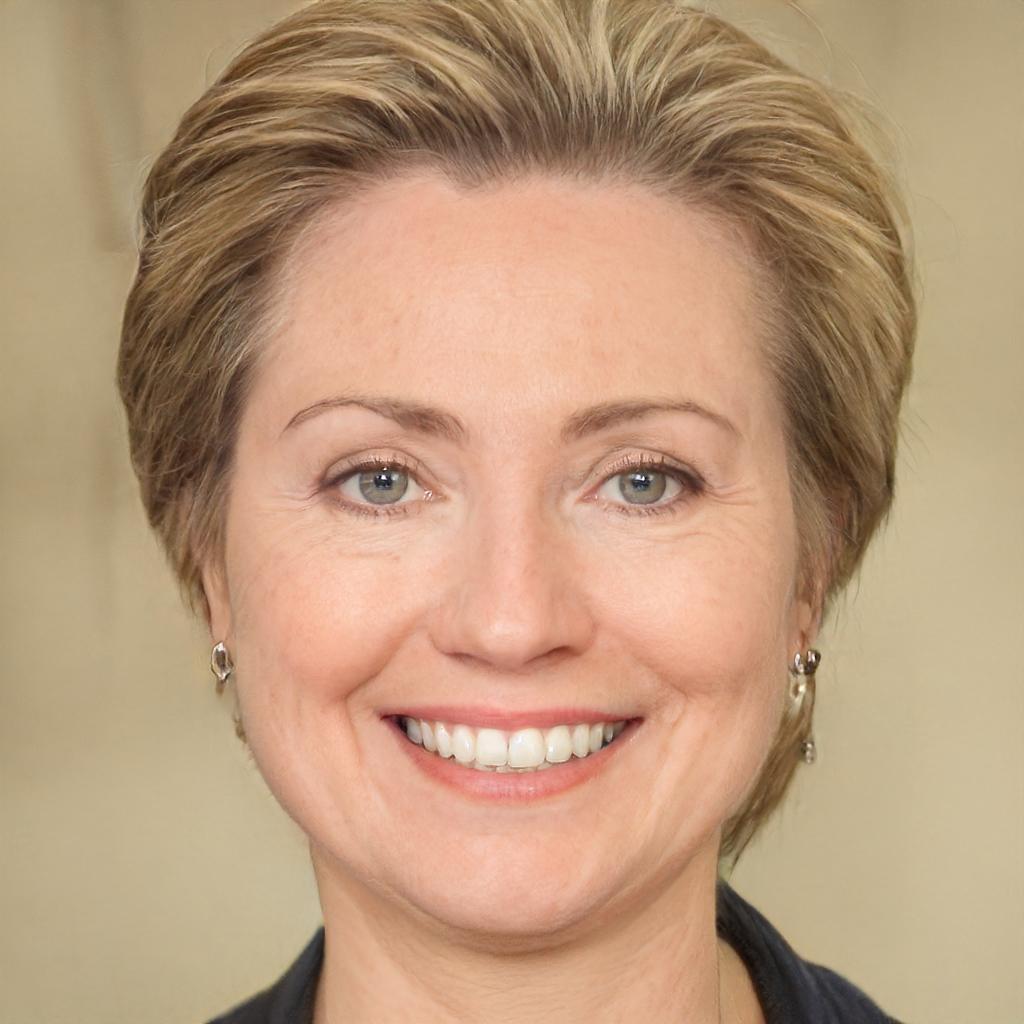} &
		\includegraphics[width=0.079\textwidth]{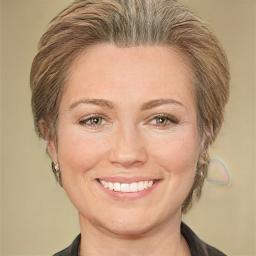} &
		\includegraphics[width=0.079\textwidth]{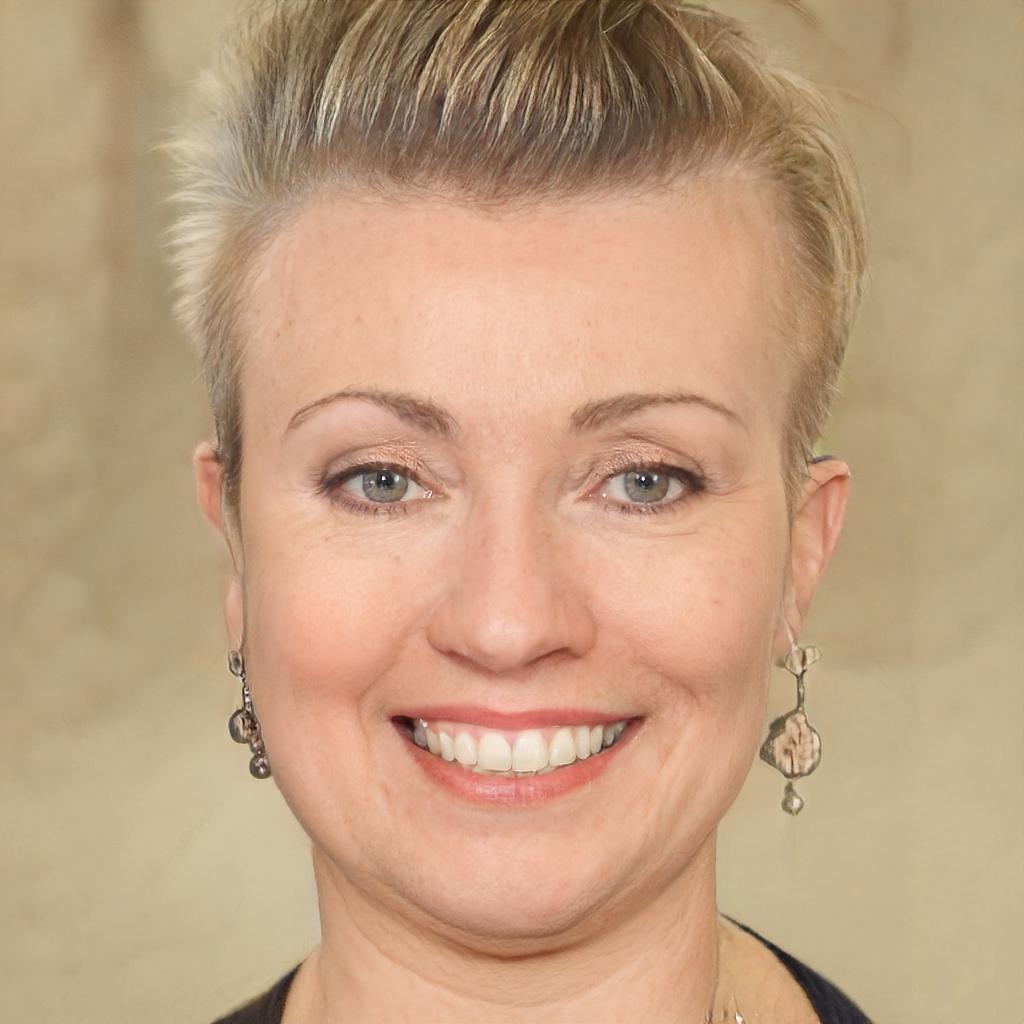} &
		\includegraphics[width=0.079\textwidth]{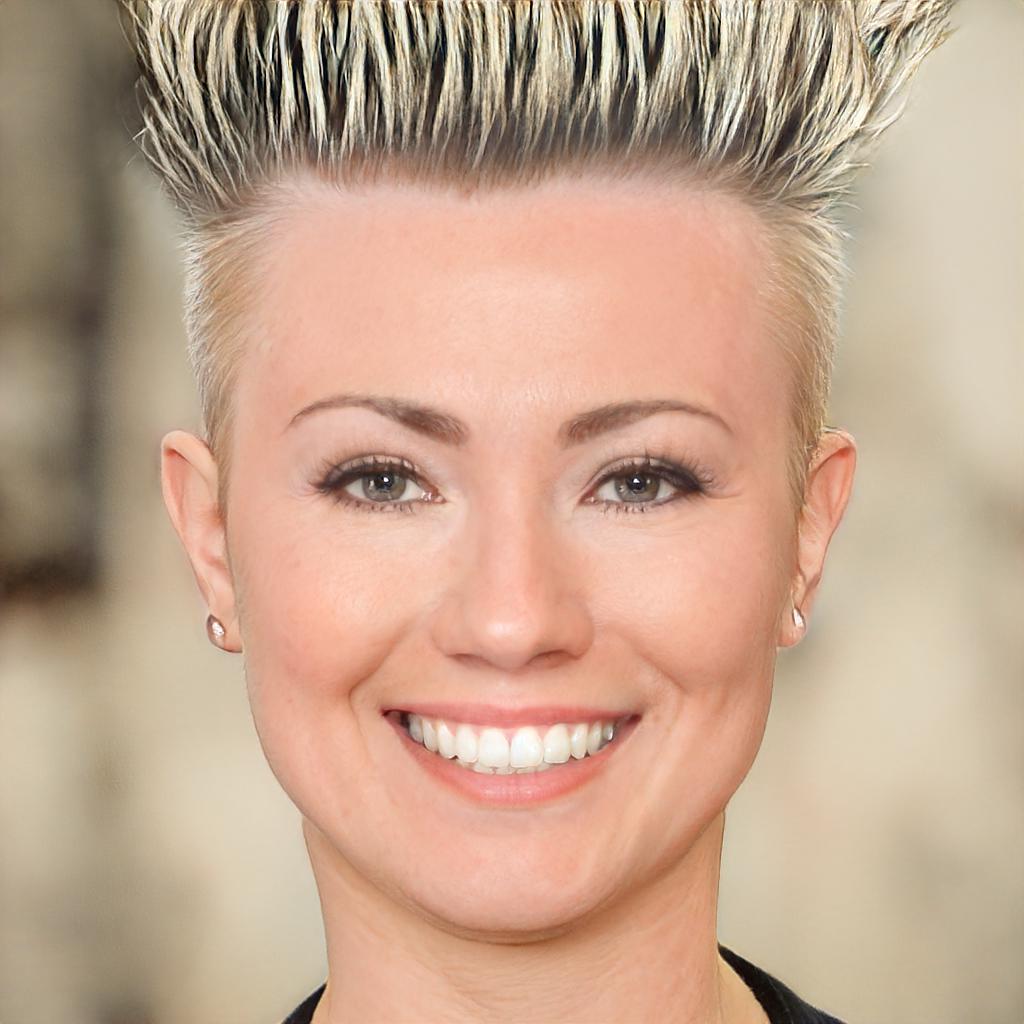} &
		
		\includegraphics[width=0.079\textwidth]{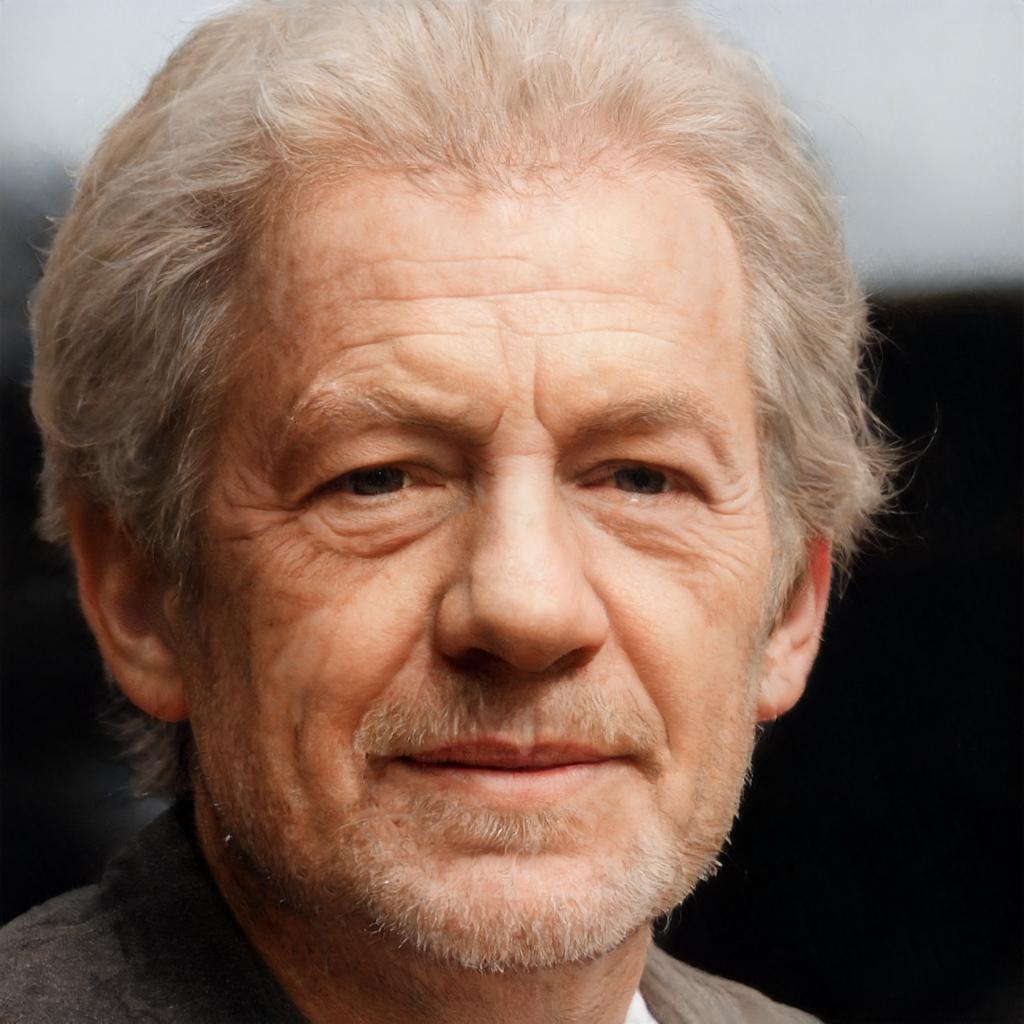} &
		\includegraphics[width=0.079\textwidth]{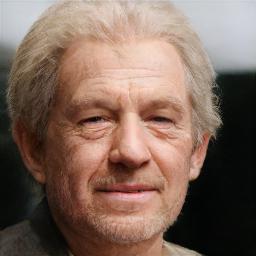} &
		\includegraphics[width=0.079\textwidth]{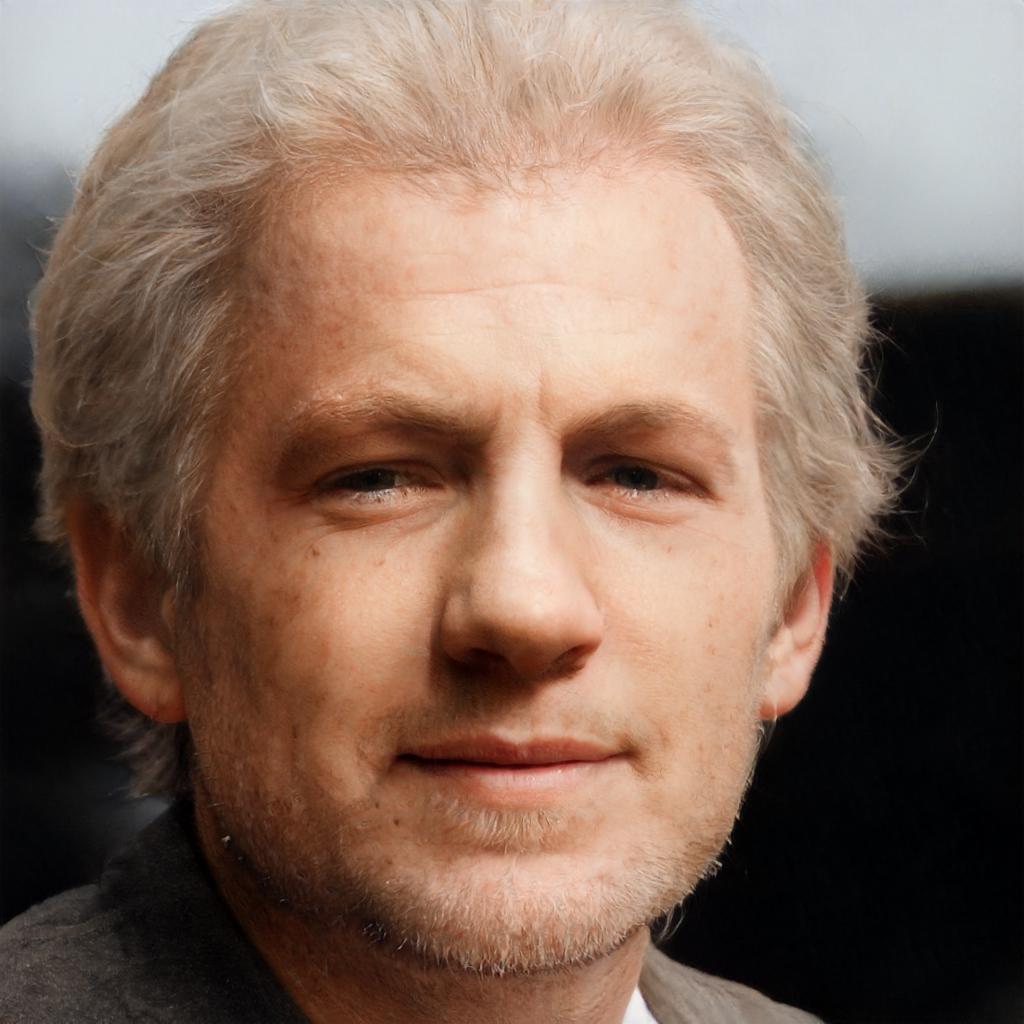} &
		\includegraphics[width=0.079\textwidth]{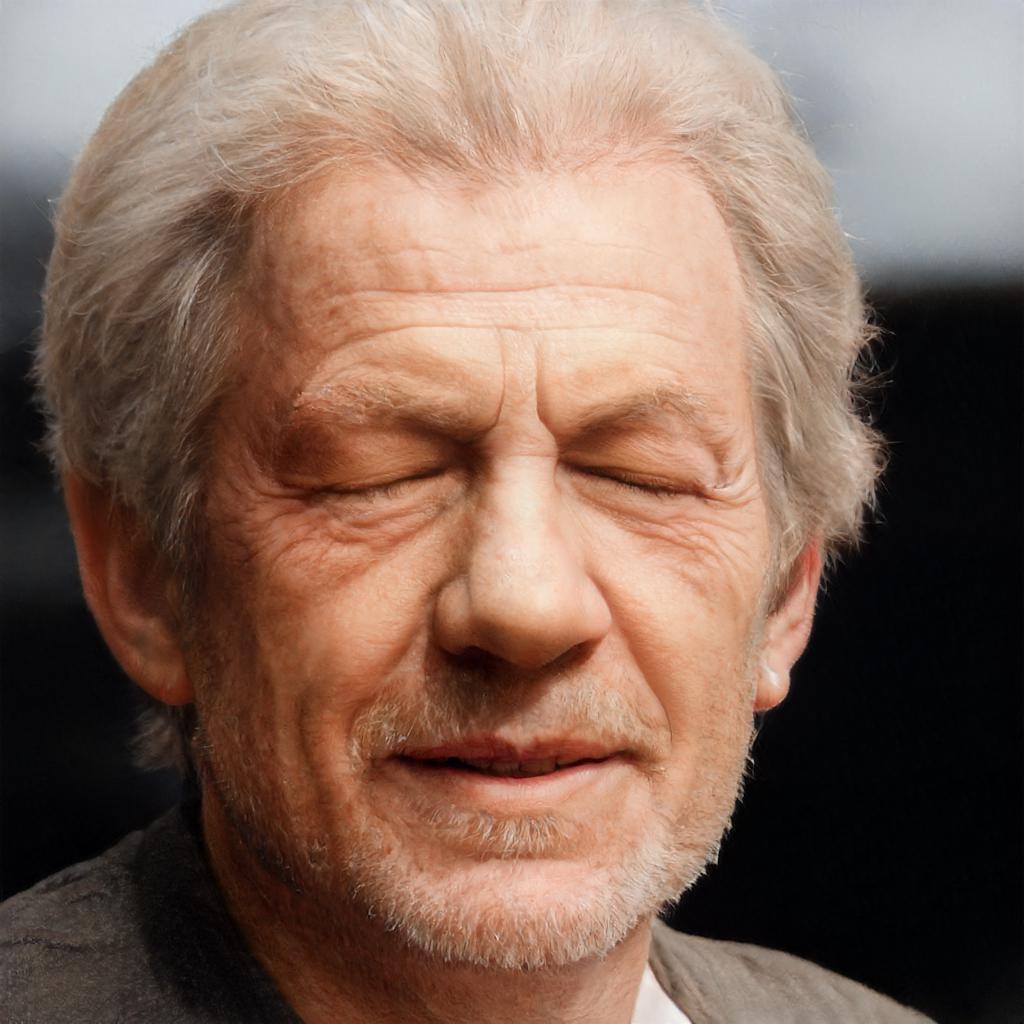} \\
		
	\multicolumn{4}{c}{\ruleline{0.30\textwidth}{Trump}} &\multicolumn{4}{c}{\ruleline{0.30\textwidth}{Mohawk}} &\multicolumn{4}{c}{\ruleline{0.30\textwidth}{Without wrinkles}} 
	\end{tabular}}
	\end{center}
	\caption{\label{fig:global-vs-mapper} We compare three methods that utilize StyleGAN and CLIP using three different kinds of attributes.}
\end{figure*}

\section{Comparisons and Evaluation}~\label{experiments}
\label{sec:experiments}
\vspace{-4mm}

We now turn to compare the three methods presented and analyzed in the previous sections among themselves and to other methods. All the real images that we manipulate are inverted using the e4e encoder~\cite{tov2021designing}. 

\textbf{Text-driven image manipulation methods:}
We begin by comparing several text-driven facial image manipulation methods 
in Figure~\ref{fig:global-vs-mapper}.
We compare between our latent mapper method (Section~\ref{sec:mapper}), our global direction method (Section~\ref{sec:global}), and TediGAN~\cite{xia2020tedigan}. For TediGAN, we use the authors' official implementation, which has been recently updated to utilize CLIP for image manipulation, and thus is somewhat different from the method presented in their paper.
We do not include results of the optimization method presented in Section~\ref{sec:opt}, since
its sensitivity to hyperparameters makes it time-consuming, and therefore not scalable.

We perform the comparison using three kinds of attributes ranging from complex, yet specific (e.g., ``Trump''), less complex and less specific (e.g., ``Mohawk''), to simpler and more common (e.g., ``without wrinkles''). The complex ``Trump'' manipulation, involves several attributes such as blonde hair, squinting eyes, open mouth, somewhat swollen face and Trump's identity. 
While a global latent direction is able to capture the main visual attributes, which are not specific to Trump, it fails to capture the specific identity. In contrast, the latent mapper is more successful.
The ``Mohawk hairstyle'' is a less complex attribute, as it involves only hair, and it isn't as specific.
Thus, both our methods are able to generate satisfactory manipulations. The manipulation generated by the global direction is slightly less pronounced, since the direction in CLIP space is an average one.
Finally, for the ``without wrinkles'' prompt, the global direction succeeds in removing the wrinkles, while keeping other attributes mostly unaffected, while the mapper fails. We attribute this to $\Wplus$ being less disentangled.
We observed similar behavior on another set of attributes (``Obama'',``Angry'',``beard''). We conclude that for complex and specific attributes (especially those that involve identity), the mapper is able to produce better manipulations. For simpler and/or more common attributes, a global direction suffices, while offering more disentangled manipulations. We note that the results produced by TediGAN fail in all three manipulations shown in Figure~\ref{fig:global-vs-mapper}.  

\begin{figure}[tb]
	\setlength{\tabcolsep}{1pt}	
	\begin{tabular}{cccccc}
		& {\footnotesize Original} & {\footnotesize GANSpace} & {\footnotesize InterFaceGAN} & {\footnotesize StyleSpace} & {\footnotesize Ours} \\
		
		\rotatebox{90}{\footnotesize \phantom{kk}Gender} &
		\includegraphics[width=0.18\columnwidth]{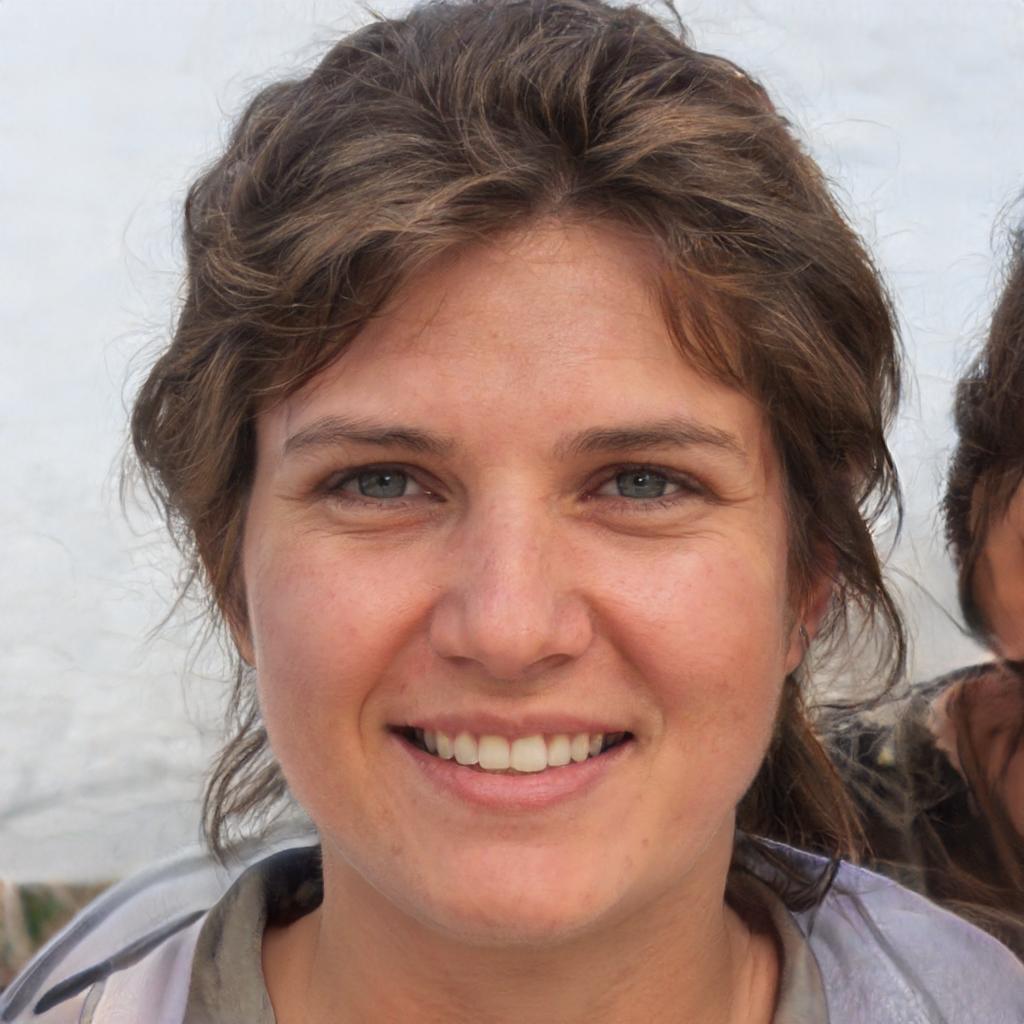} &
		\includegraphics[width=0.18\columnwidth]{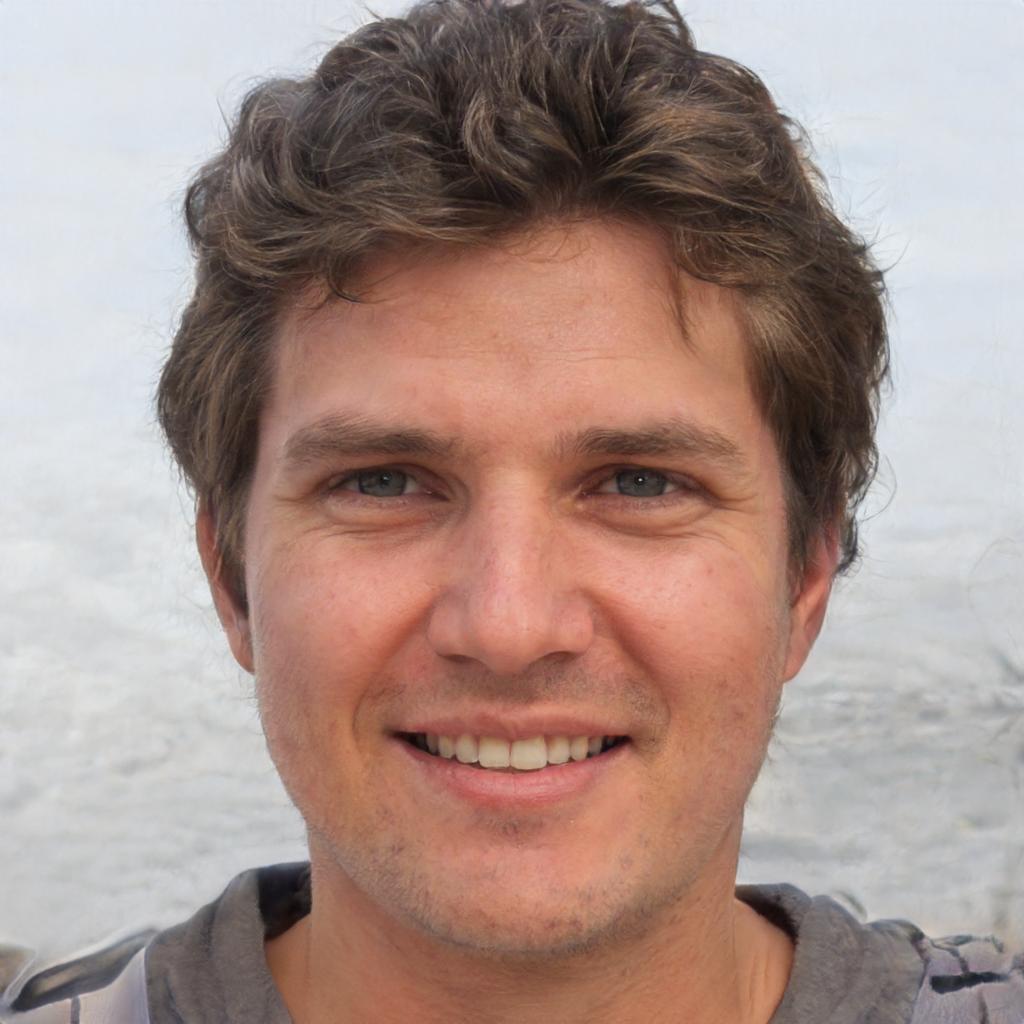} &
		\includegraphics[width=0.18\columnwidth]{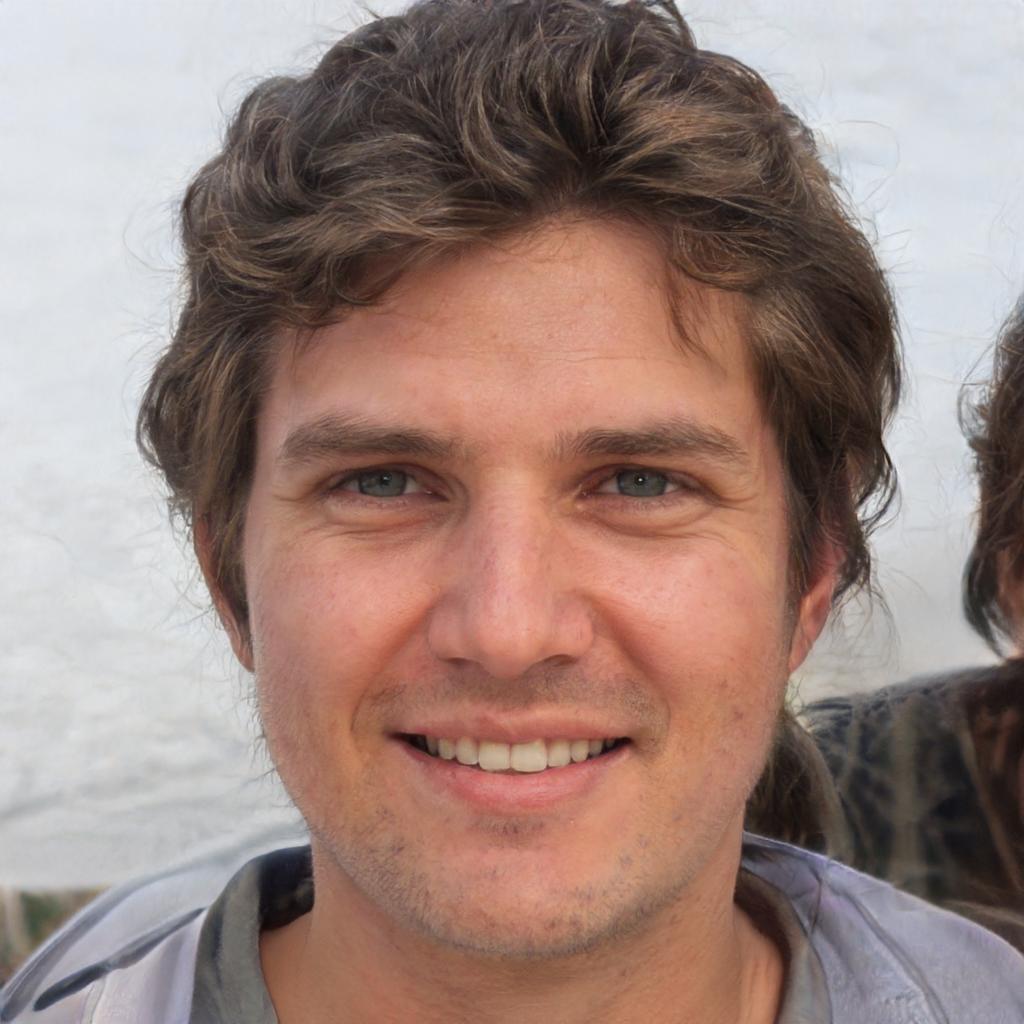} &
		\includegraphics[width=0.18\columnwidth]{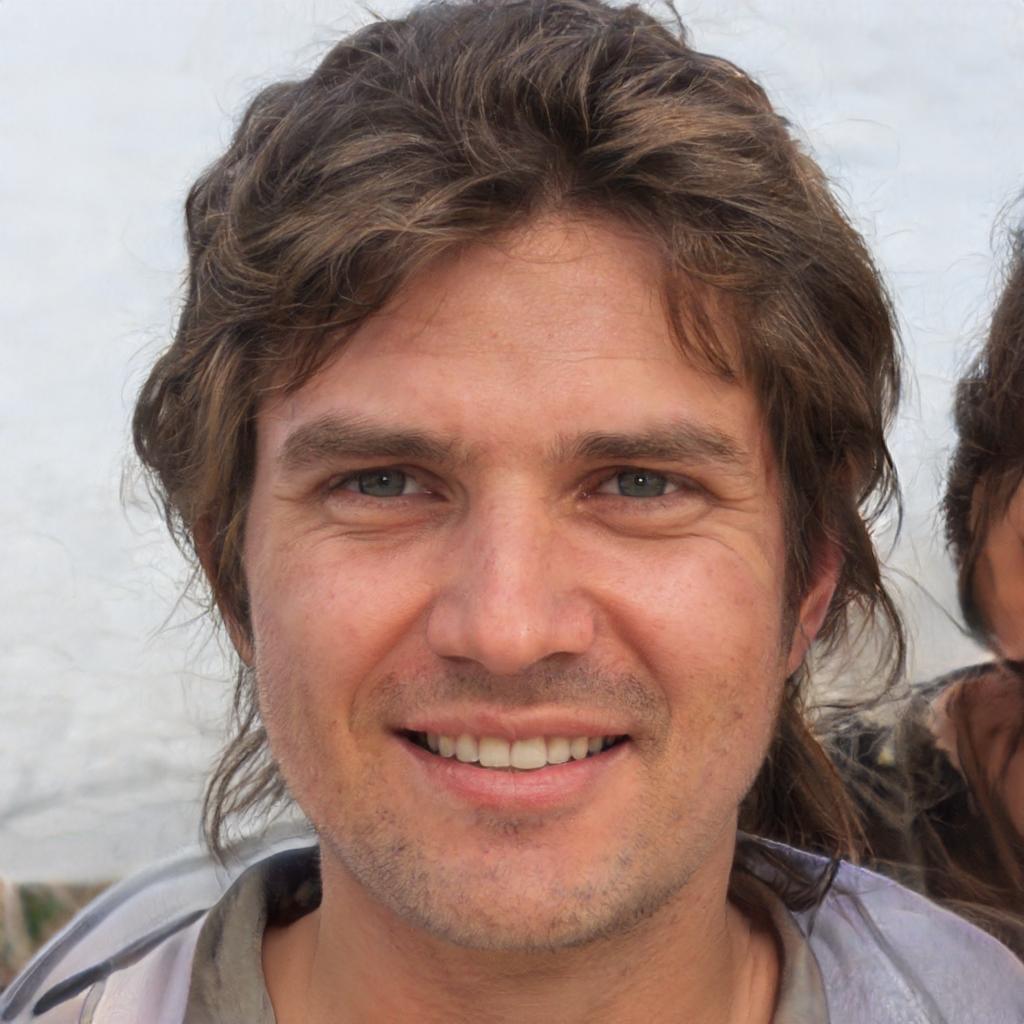} &
		\includegraphics[width=0.18\columnwidth]{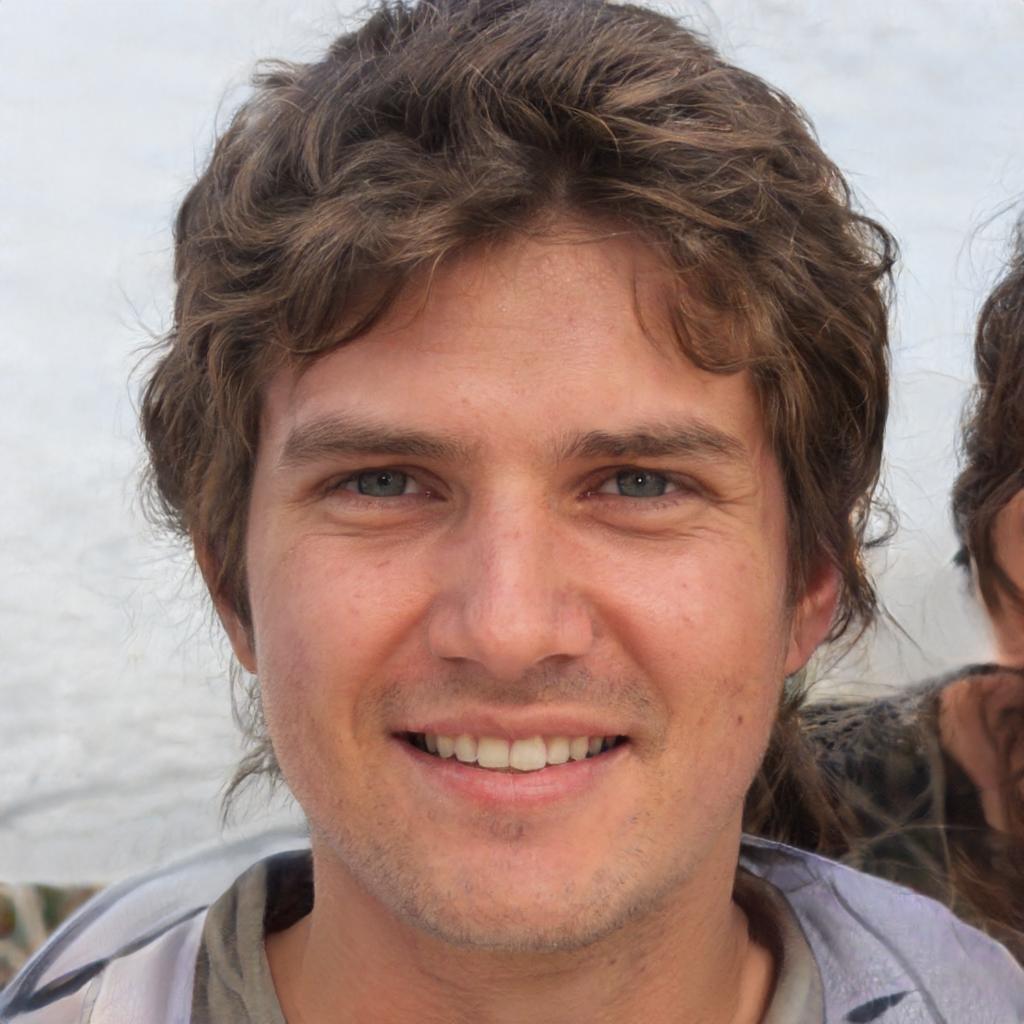} 
		\\
		
		\rotatebox{90}{\footnotesize \phantom{k}Gray hair} &
		\includegraphics[width=0.18\columnwidth]{./fig4/original.jpg} &
		\includegraphics[width=0.18\columnwidth]{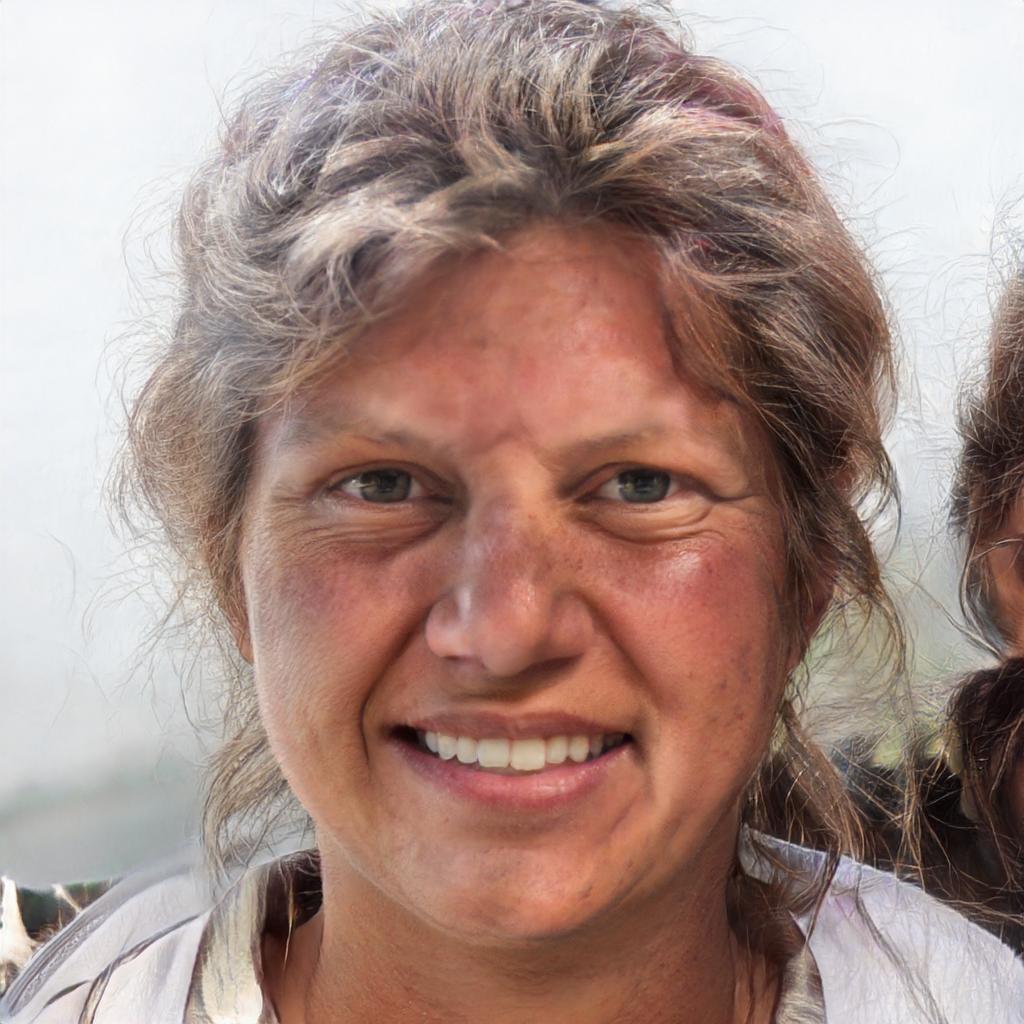} &
		\includegraphics[width=0.18\columnwidth]{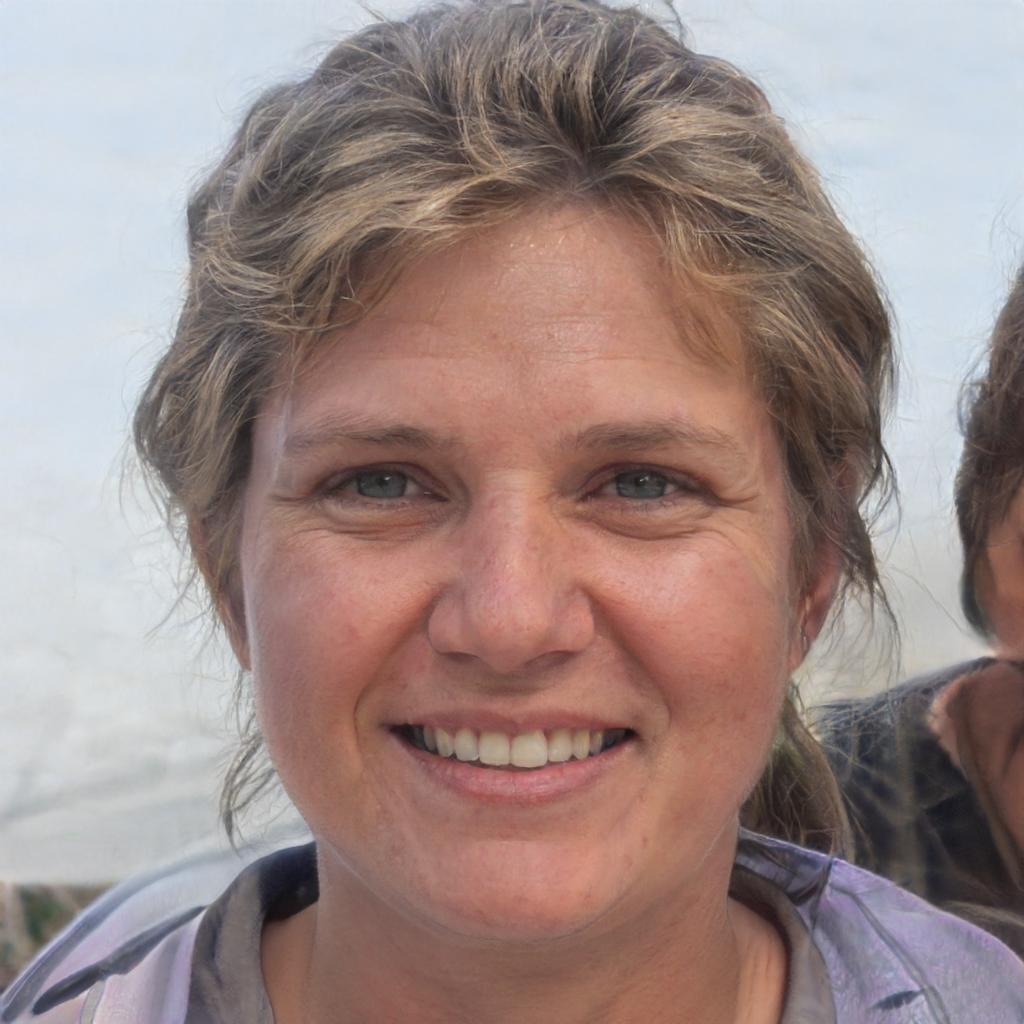} &
		\includegraphics[width=0.18\columnwidth]{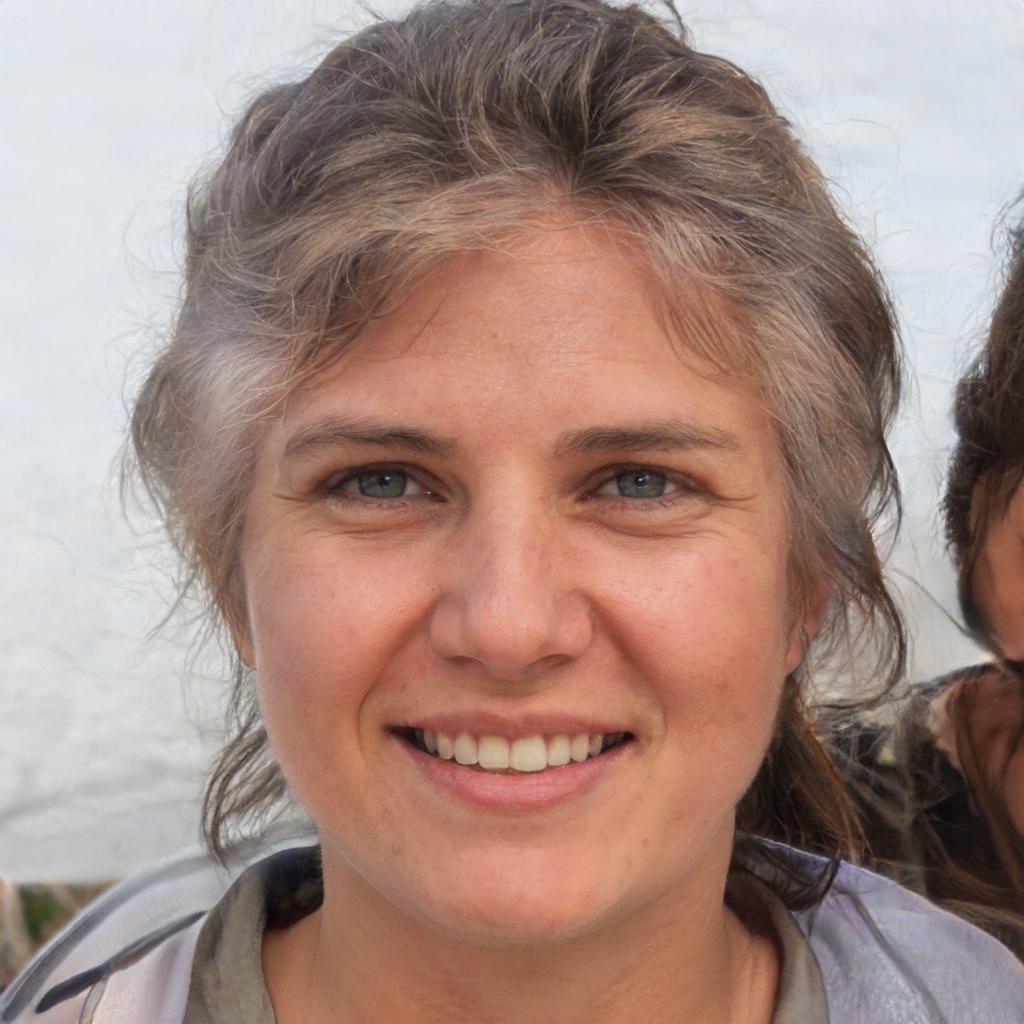}&
		\includegraphics[width=0.18\columnwidth]{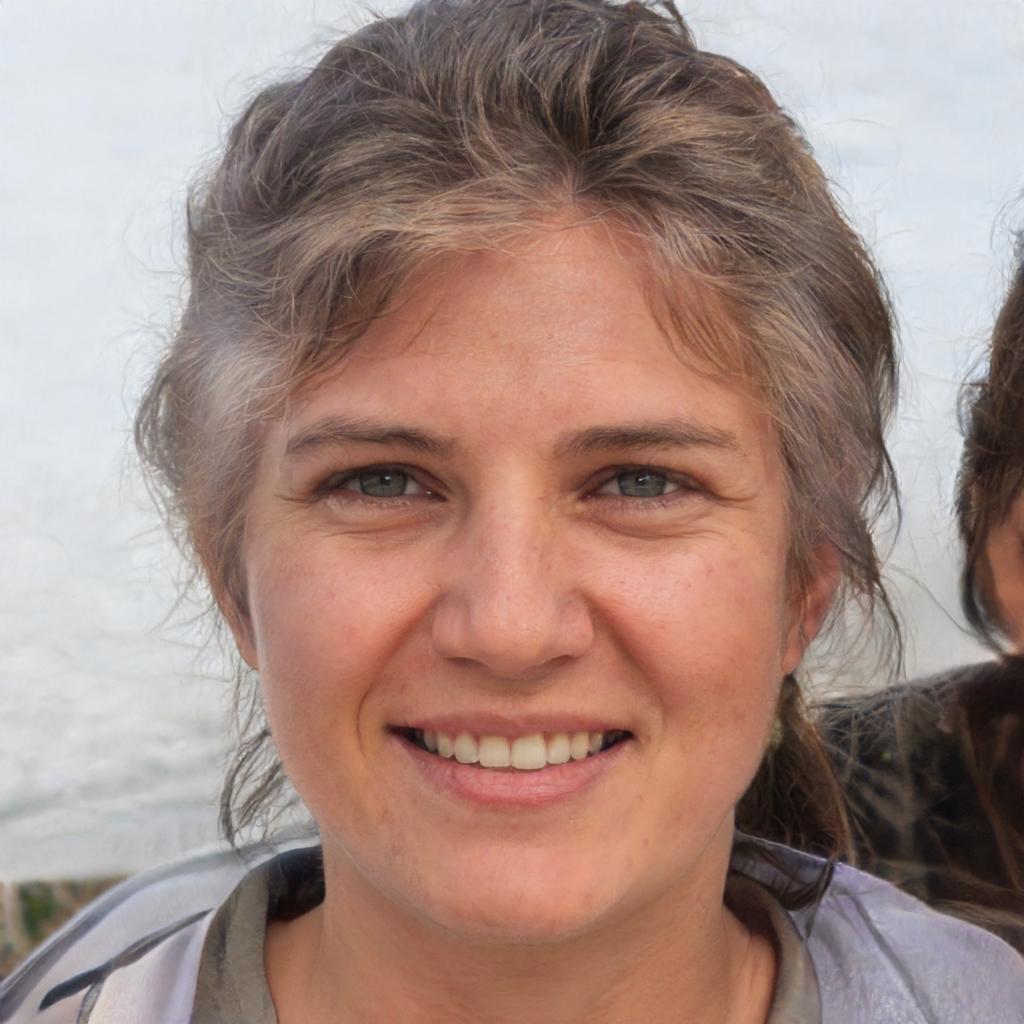} 
		\\
		
		\rotatebox{90}{\footnotesize \phantom{kk}Lipstick} &
		\includegraphics[width=0.18\columnwidth]{./fig4/original.jpg} &
		\includegraphics[width=0.18\columnwidth]{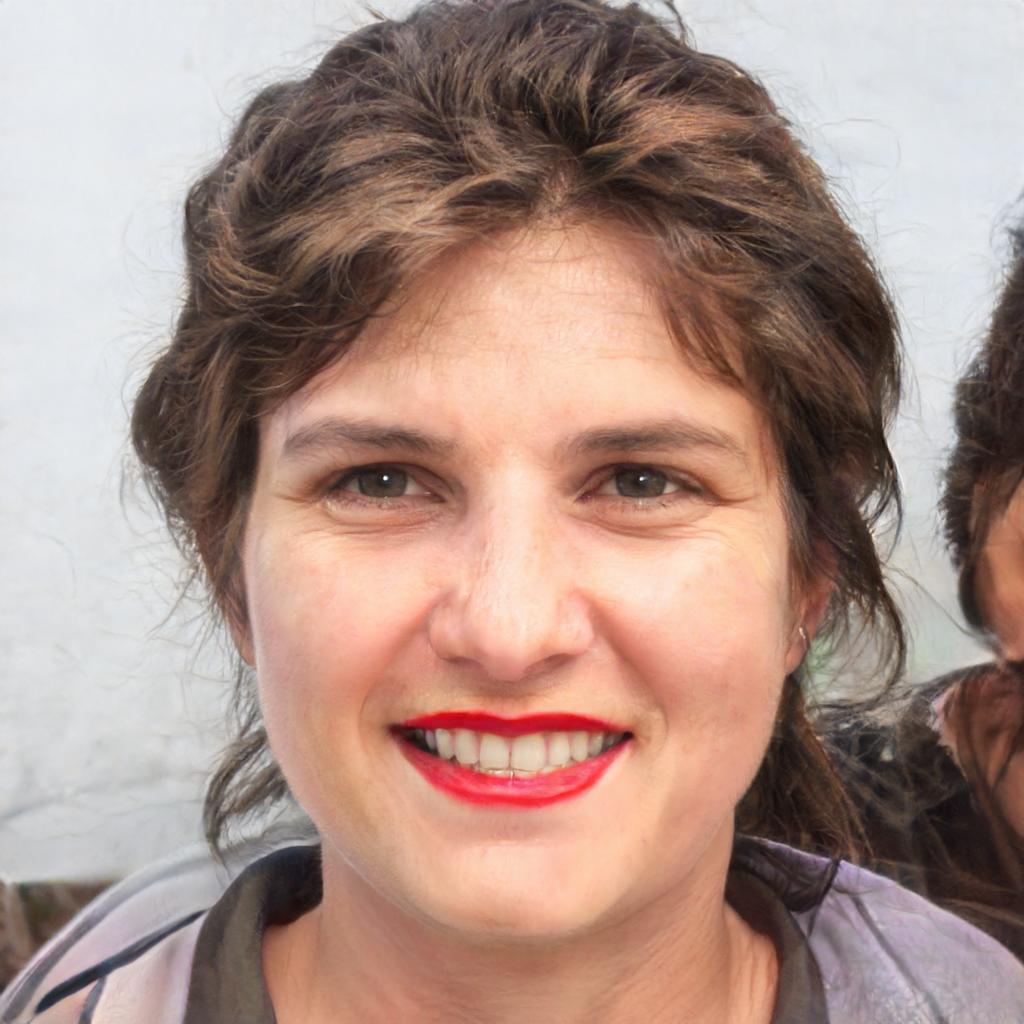} &
		\includegraphics[width=0.18\columnwidth]{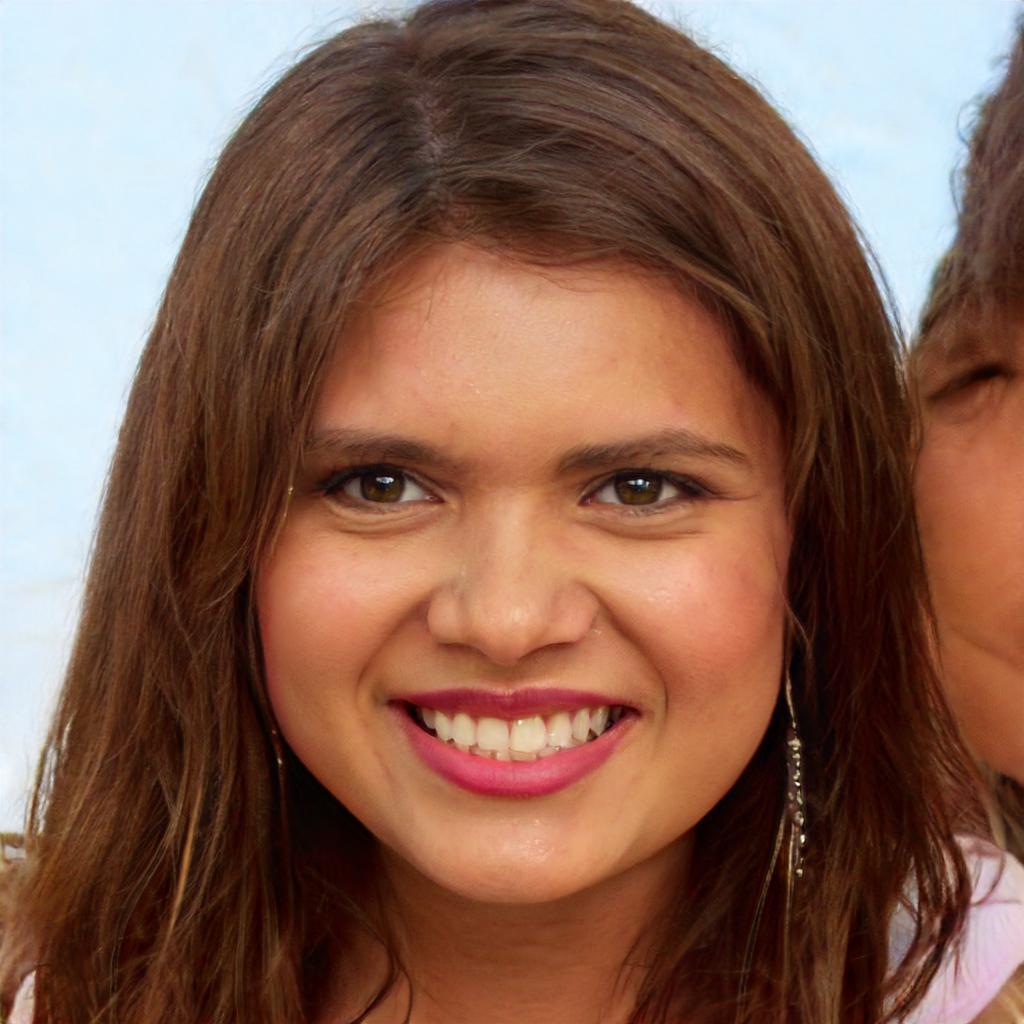} &
		\includegraphics[width=0.18\columnwidth]{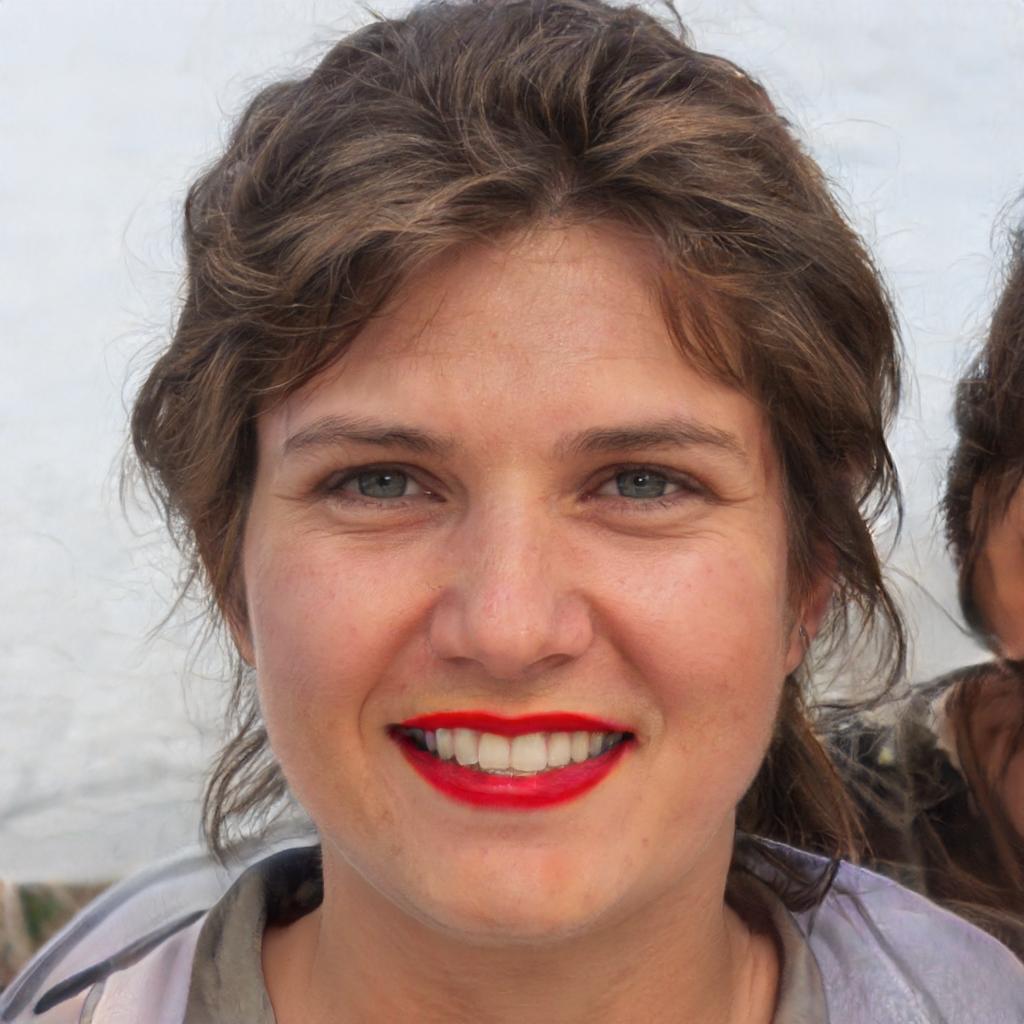} &
		\includegraphics[width=0.18\columnwidth]{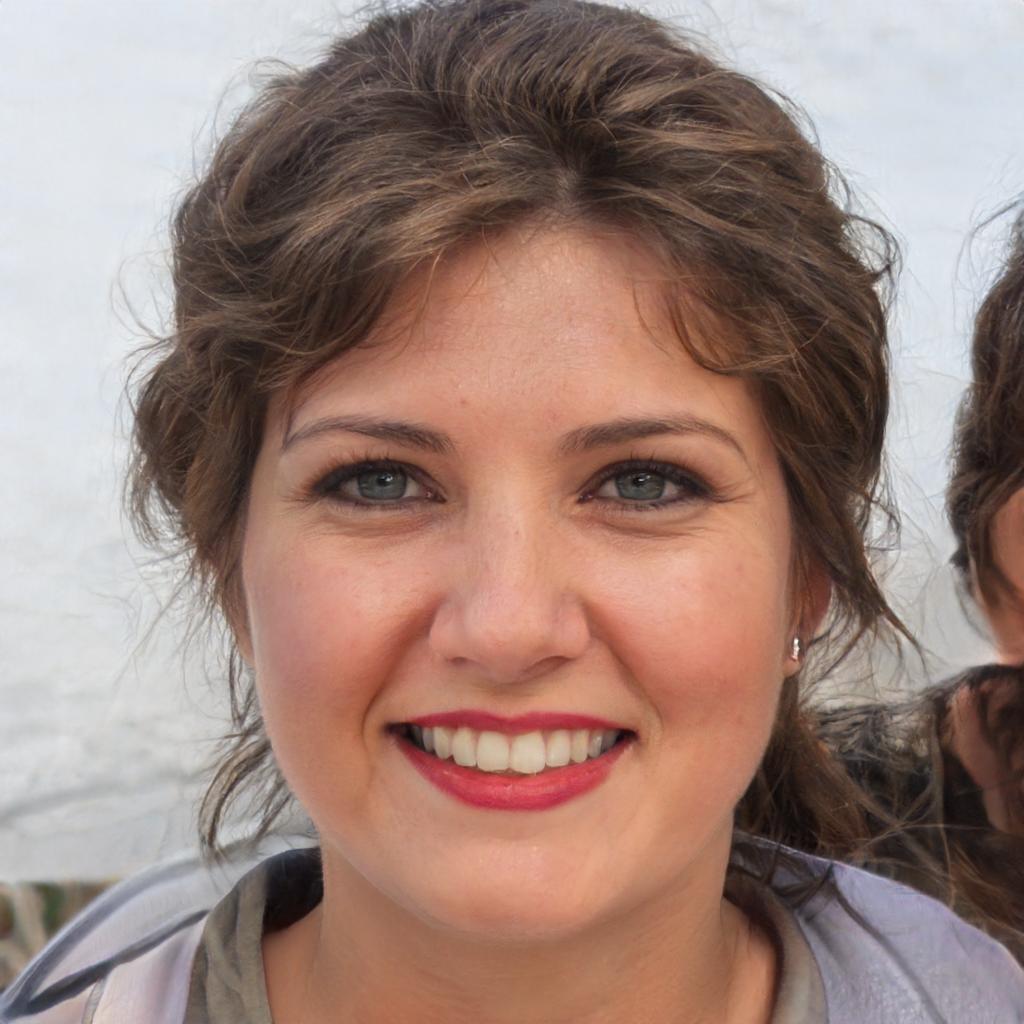} 
	\end{tabular}
	
	\caption{Comparison with state-of-the-art methods using the same amount of manipulation according to a pretrained attribute classifier. 
		\vspace{-3mm}
	}
	\label{fig:compare_linear}
\end{figure}

\textbf{Other StyleGAN manipulation methods:}
In Figure~\ref{fig:compare_linear}, we show a comparison between our global direction method and several state-of-the-art StyleGAN image manipulation methods: GANSpace~\cite{harkonen2020ganspace}, InterFaceGAN \cite{shen2020interfacegan}, and StyleSpace \cite{wu2020stylespace}.
The comparison only examines the attributes which all of the compared methods are able to manipulate (Gender, Grey hair, and Lipstick), and thus it does not include the many novel manipulations enabled by our approach.
Since all of these are common attributes, we do not include our mapper in this comparison.
Following Wu \etal~\cite{wu2020stylespace}, the manipulation step strength is chosen such that it induces the same amount of change in the logit value of the corresponding classifiers (pretrained on CelebA).

It may be seen that in GANSpace~\cite{harkonen2020ganspace} manipulation is entangled with skin color and lighting, while in InterFaceGAN~\cite{shen2020interfacegan} the identity may change significantly (when manipulating Lipstick). Our manipulation is very similar to StyleSpace \cite{wu2020stylespace}, which only changes the target attribute, while all other attributes remain the same. 

In the supplementary material, we also show a comparison with StyleFLow~\cite{abdal2020styleflow}, a state-of-the-art non-linear method. Our method produces results of similar quality, despite the fact that StyleFlow simultaneously uses several attribute classifiers and regressors (from the Microsoft face API), and is thus can manipulate a limited set of attributes. In contrast, our method requires no extra supervision.

%
%
%
%

\paragraph{Limitations.} Our method relies on a pretrained StyleGAN generator and CLIP model for a joint language-vision embedding.
Thus, it cannot be expected to manipulate images to a point where they lie outside the domain of the pretrained generator (or remain inside the domain, but in regions that are less well covered by the generator).
Similarly, text prompts which map into areas of CLIP space that are not well populated by images, cannot be expected to yield a visual manipulation that faithfully reflects the semantics of the prompt.
We have also observed that drastic manipulations in visually diverse datasets are difficult to achieve. For example, while tigers are easily transformed into lions (see Figure~\ref{fig:teaser}), we were less successful when transforming tigers to wolves, as demonstrated in the supplementary material.



\section{Conclusions}

We introduced three novel image manipulation methods, which combine the strong generative powers of StyleGAN with the extraordinary visual concept encoding abilities of CLIP.
We have shown that these techniques enable a wide variety of unique image manipulations, some of which are impossible to achieve with existing methods that rely on annotated data.
We have also demonstrated that CLIP provides fine-grained edit controls, such as specifying a desired hair style, while our method is able to control the manipulation strength and the degree of disentanglement.
In summary, we believe that text-driven manipulation is a powerful image editing tool, whose abilities and importance will only continue to grow.


{\small
\bibliographystyle{ieee_fullname}
\bibliography{egbib}
}

\clearpage
\appendix

\section{Latent Mapper -- Ablation Study}
In this section, we study the importance of various choices in the design of our latent mapper (Section 5). 

\subsection{Architecture} 
\begin{figure}[b]
	\setlength{\tabcolsep}{1pt}
	\centering
	{\footnotesize
		\begin{tabular}{c c c}
			\includegraphics[width=0.32\linewidth]{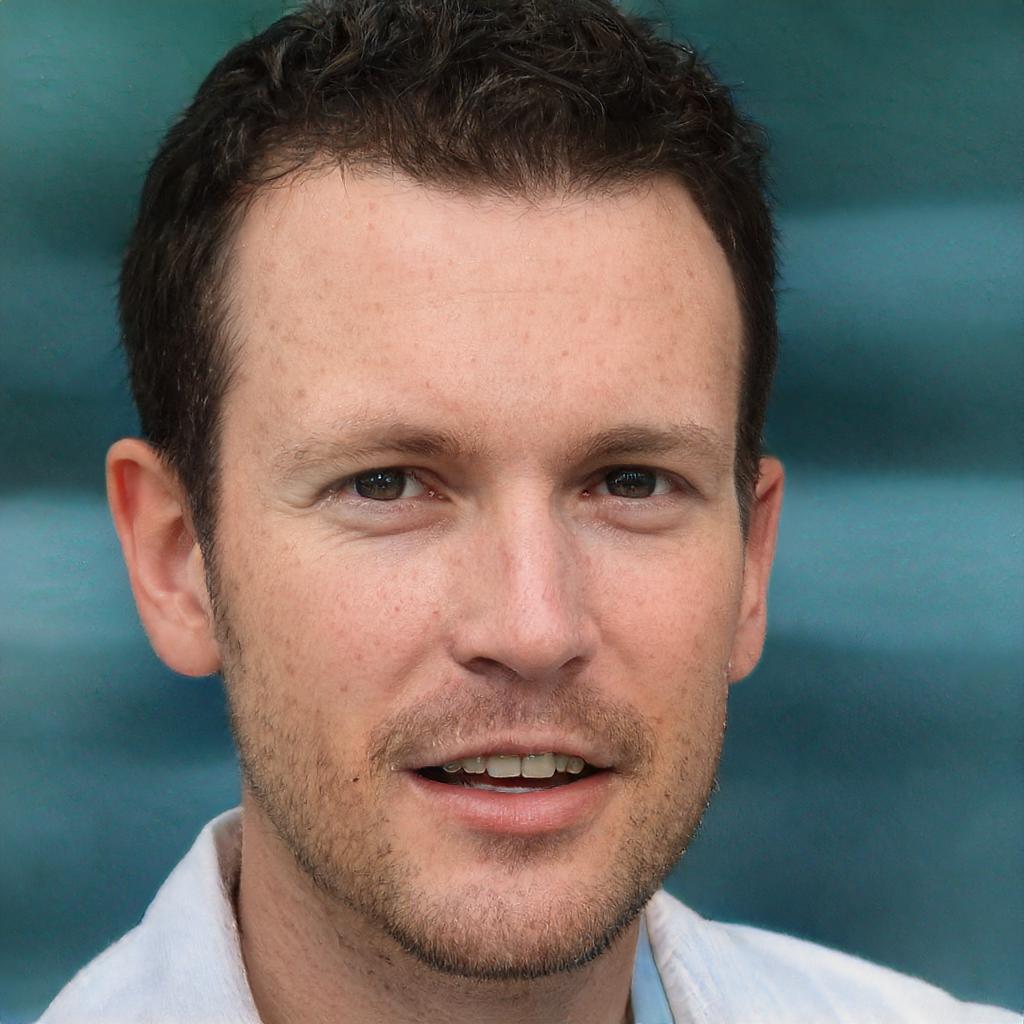} &
            \includegraphics[width=0.32\linewidth]{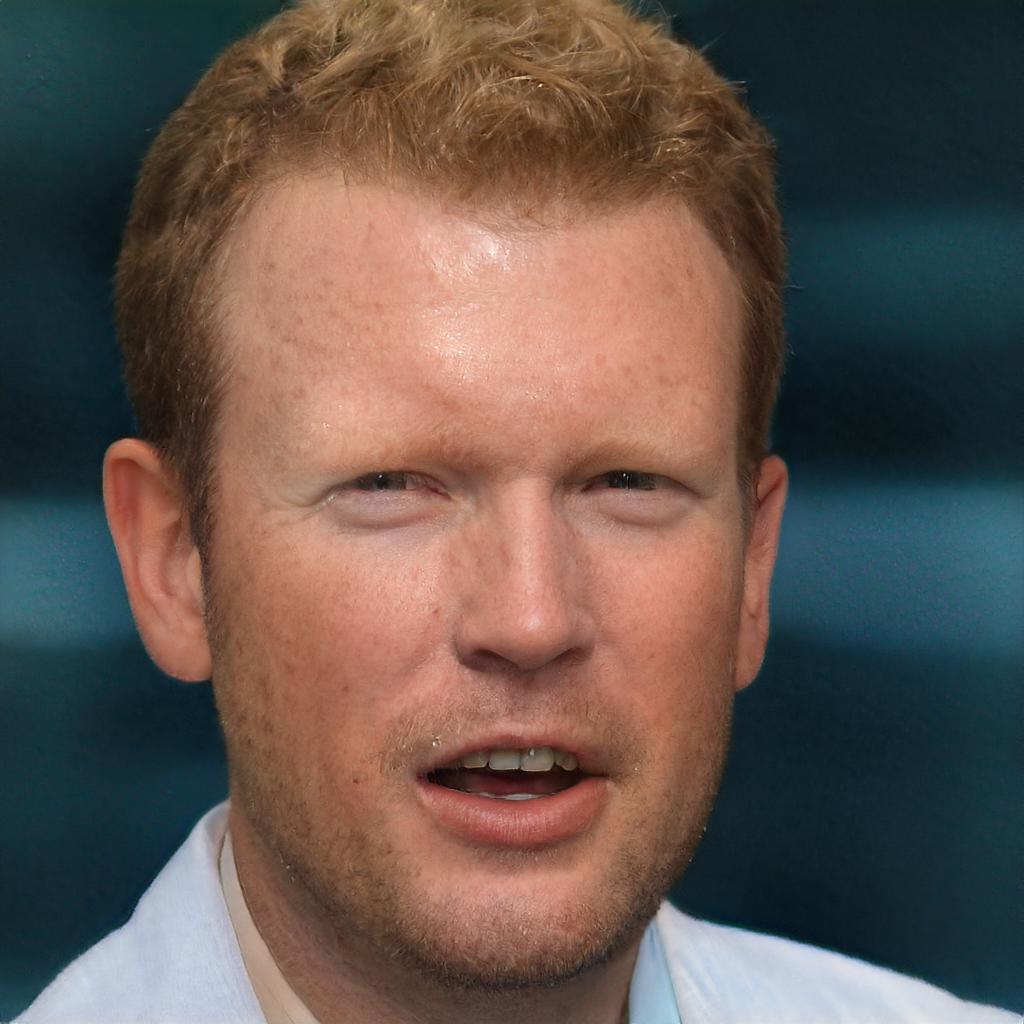} &
			\includegraphics[width=0.32\linewidth]{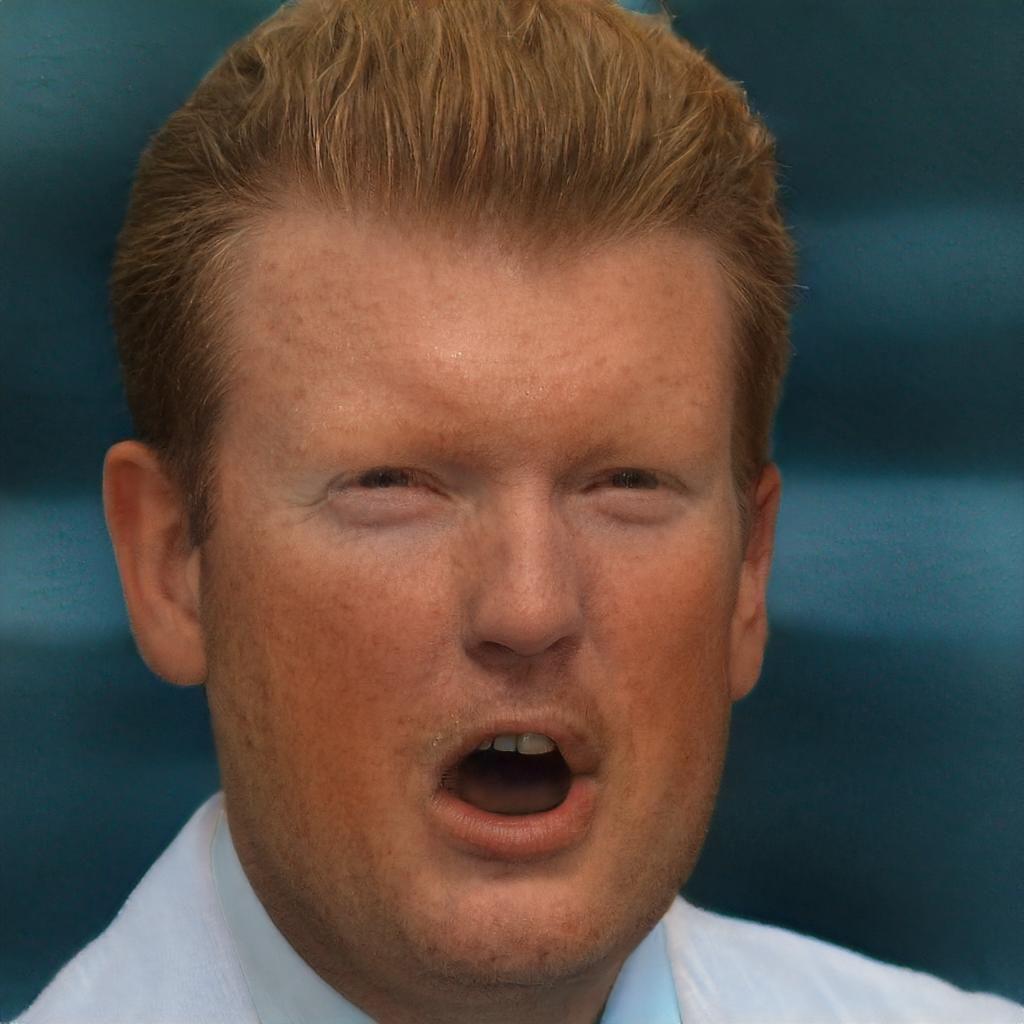}  \\
			\includegraphics[width=0.32\linewidth]{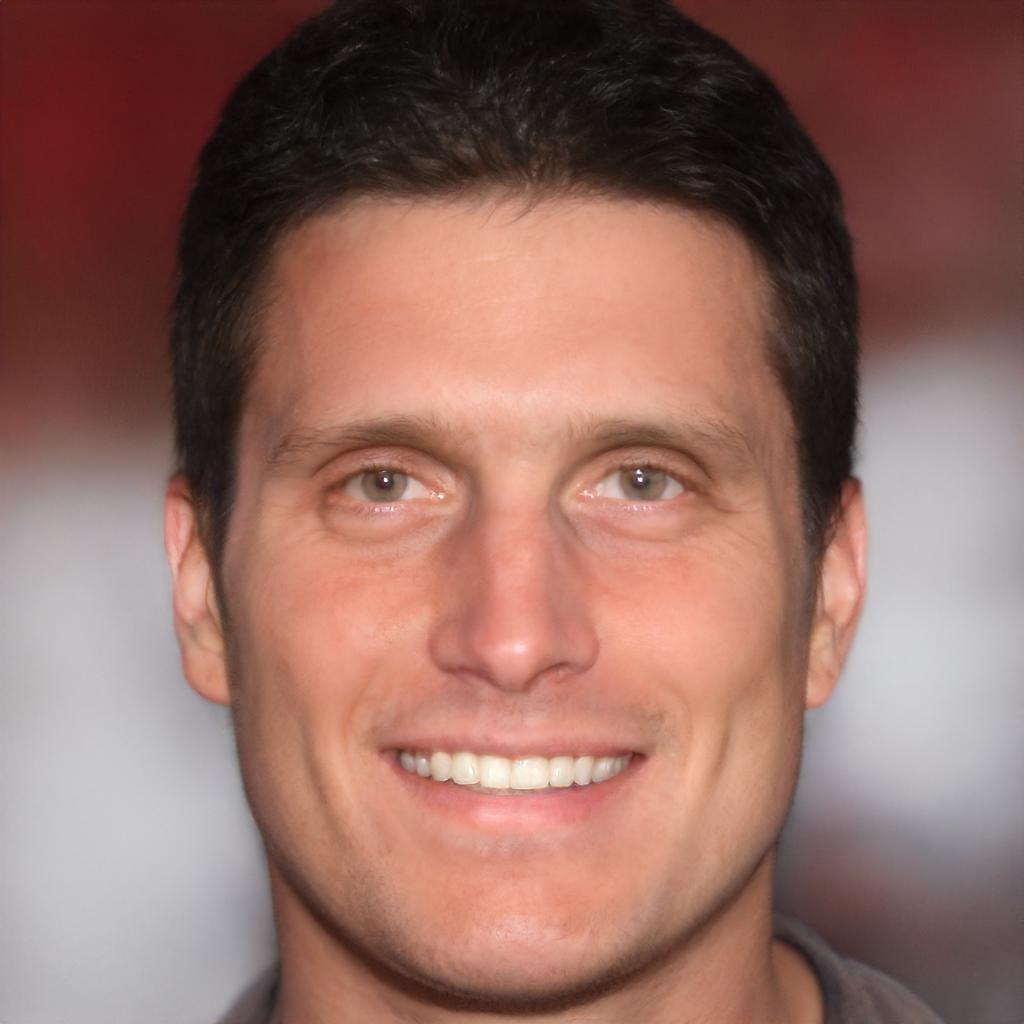} &
            \includegraphics[width=0.32\linewidth]{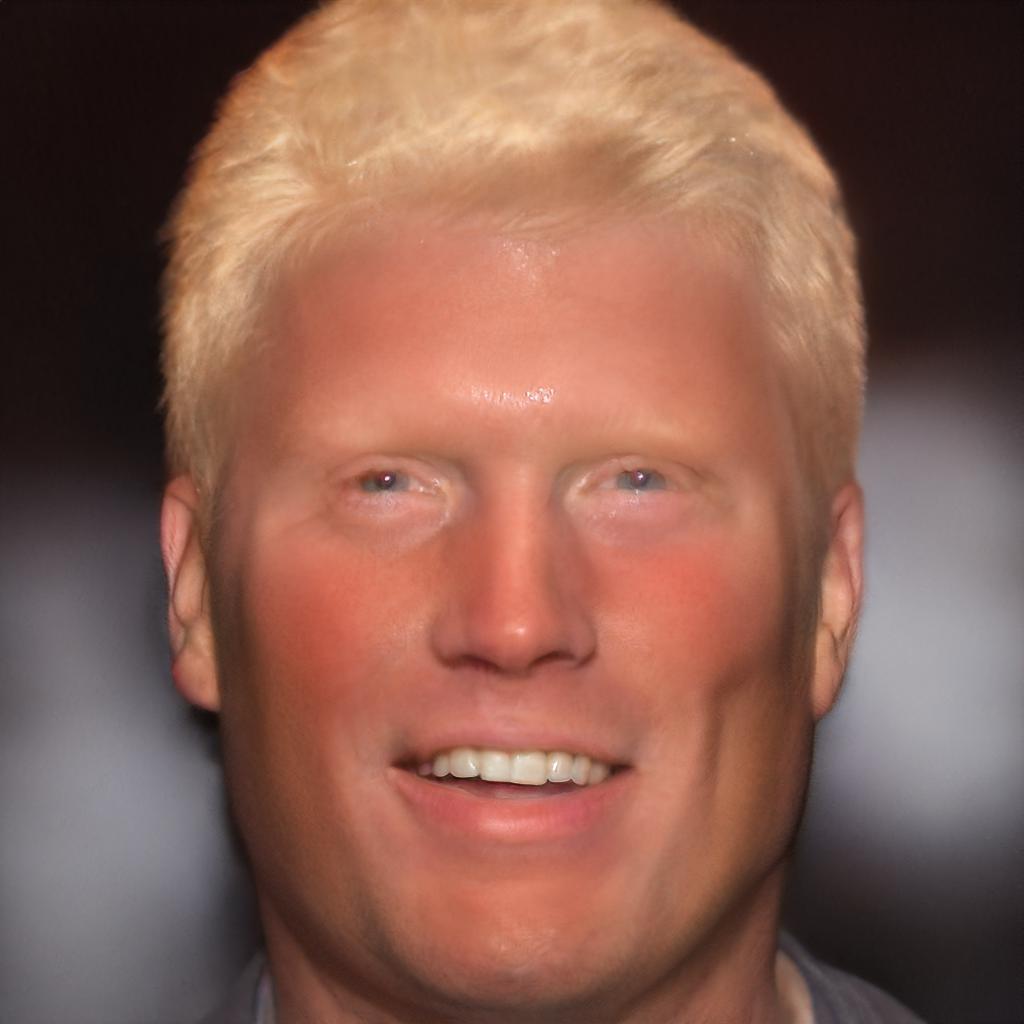} &
			\includegraphics[width=0.32\linewidth]{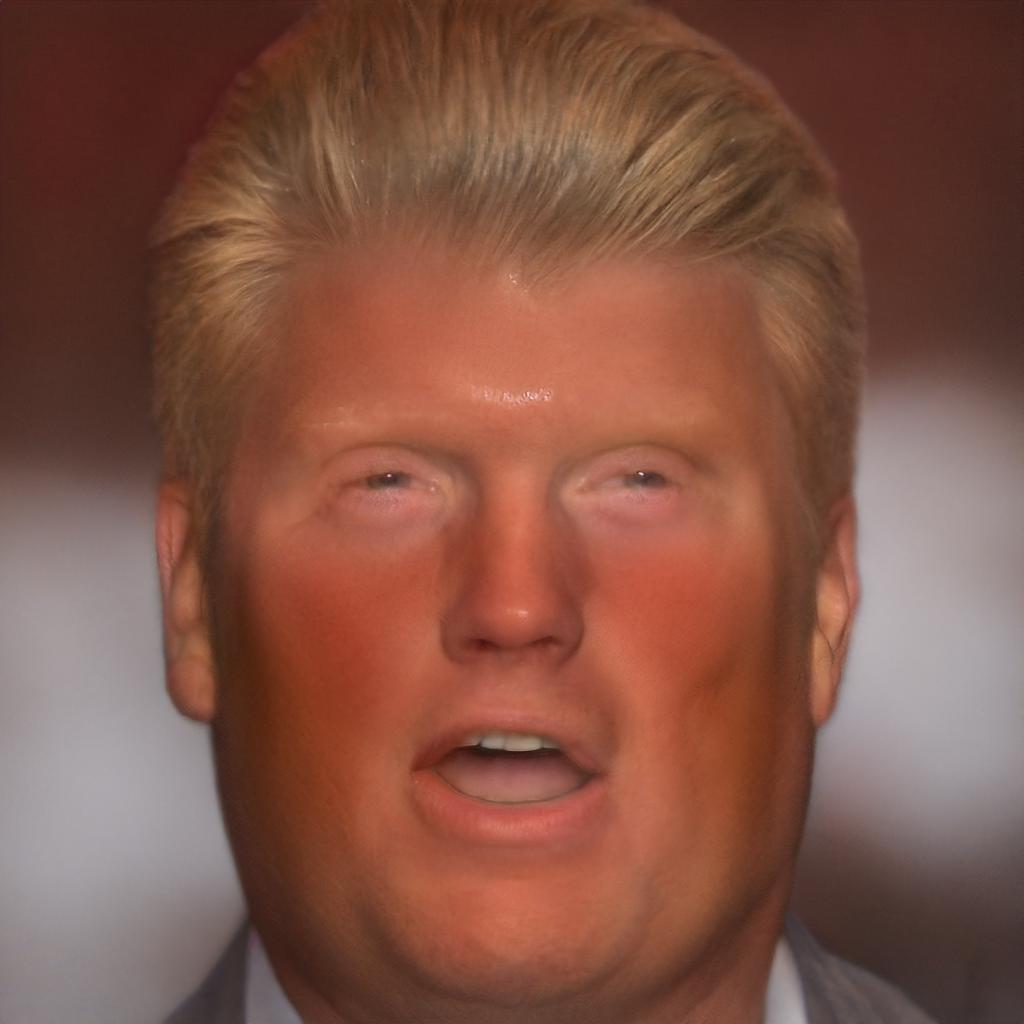}  \\
            Input & Single network & Ours (three networks)
		\end{tabular}
	}
	\caption{Comparing our mapper architecture with a simpler architecture that uses a single mapping network. The simpler mapper fails to infer multiple changes correctly. The changes in the expression and in the hair-style are not strong enough to capture the identity of the target individual. On the other hand, there are unnecessary changes in the background color in the second row when using a single network.}
	\label{fig:ablation-mapper}
\end{figure}

The architecture of the mapper is rather simple and with relatively small number of parameters. Moreover, it has negligible effect on the inference time. Yet, it is natural to compare the presented architecture, which consists of three different mapping networks, to an architecture with a single mapping network. Intuitively, using a separate network for each group of style vector entries should better enable changes at several different levels of detail in the image. Indeed, we find that with driving text that requires such changes, e.g. ``Donald Trump'', a single mapping network does not yield results that are as effective as those produced with three. An example is shown in Figure~\ref{fig:ablation-mapper}.

\begin{figure}[tb]
	\setlength{\tabcolsep}{1pt}
	\centering
	{\footnotesize
		\begin{tabular}{c c c c c}
			\includegraphics[width=0.19\linewidth]{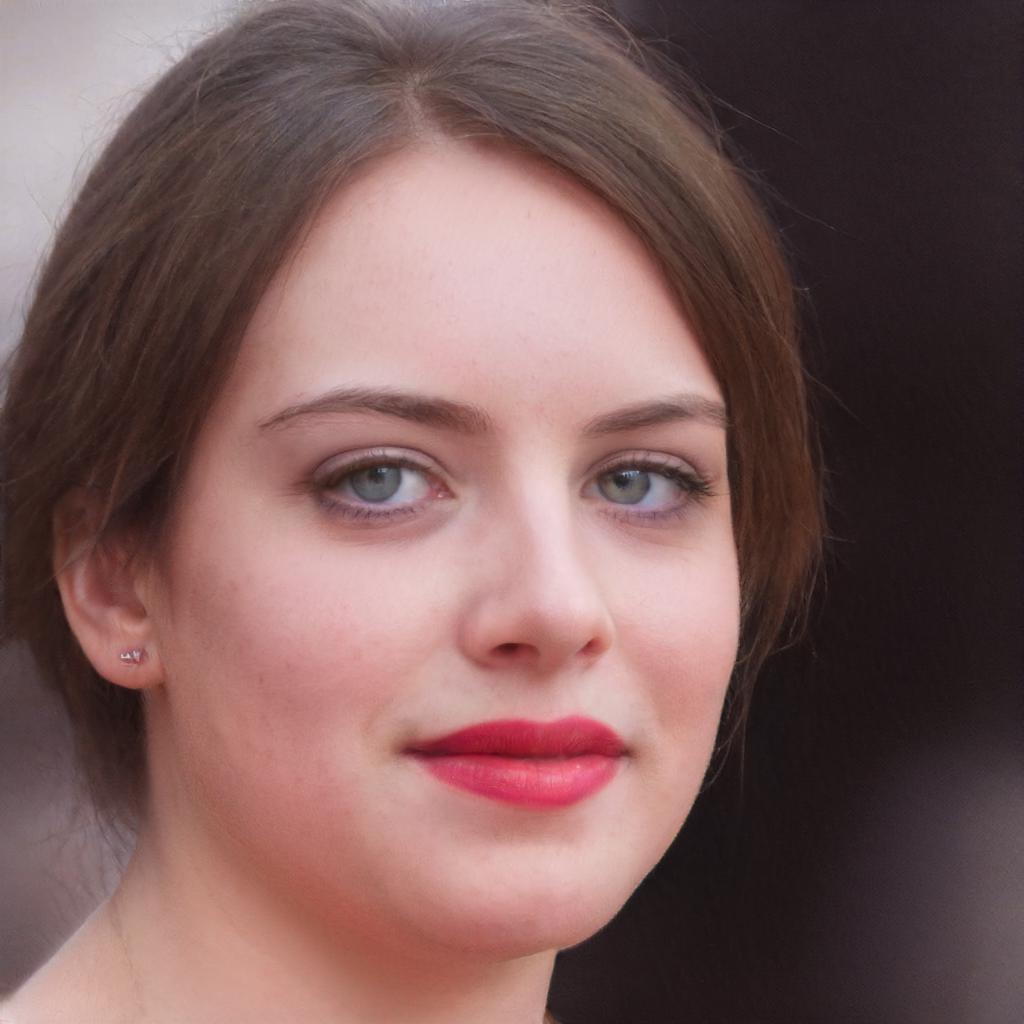} &
            \includegraphics[width=0.19\linewidth]{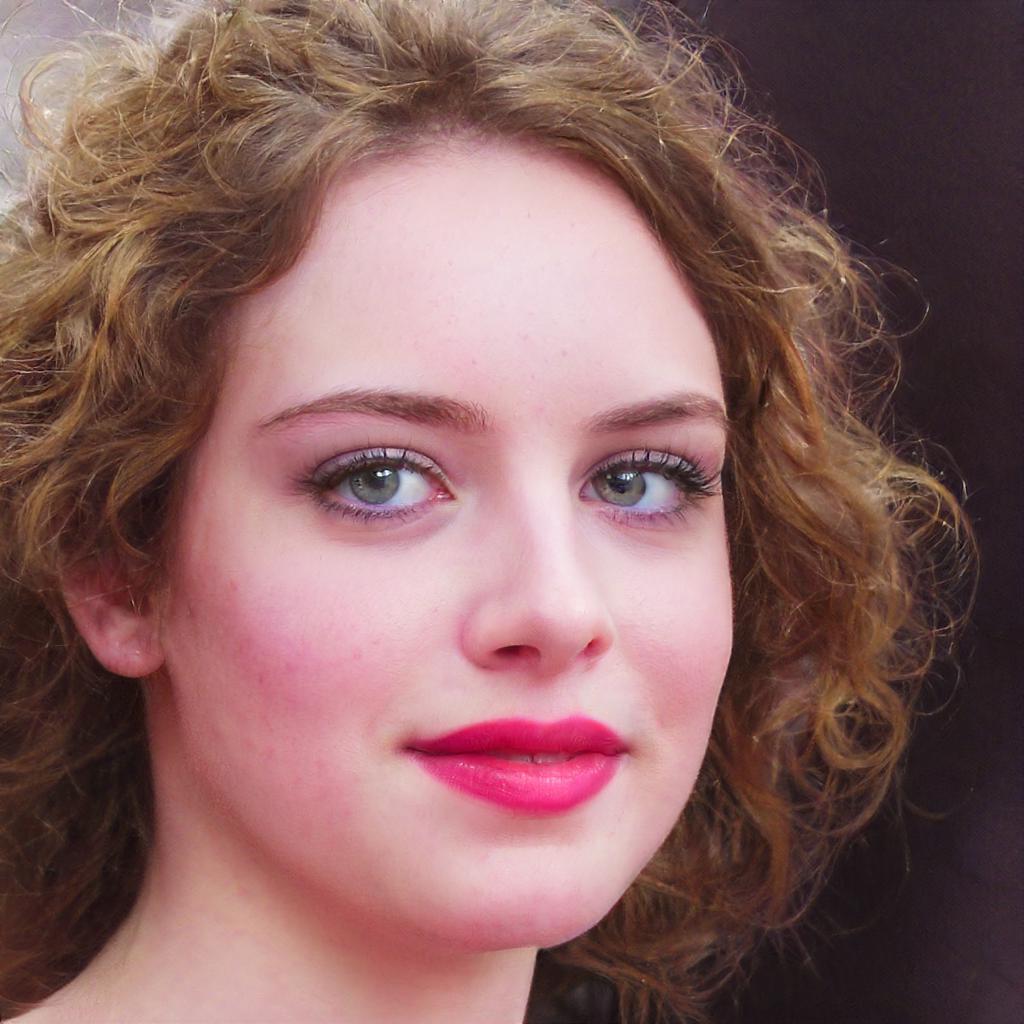} &
			\includegraphics[width=0.19\linewidth]{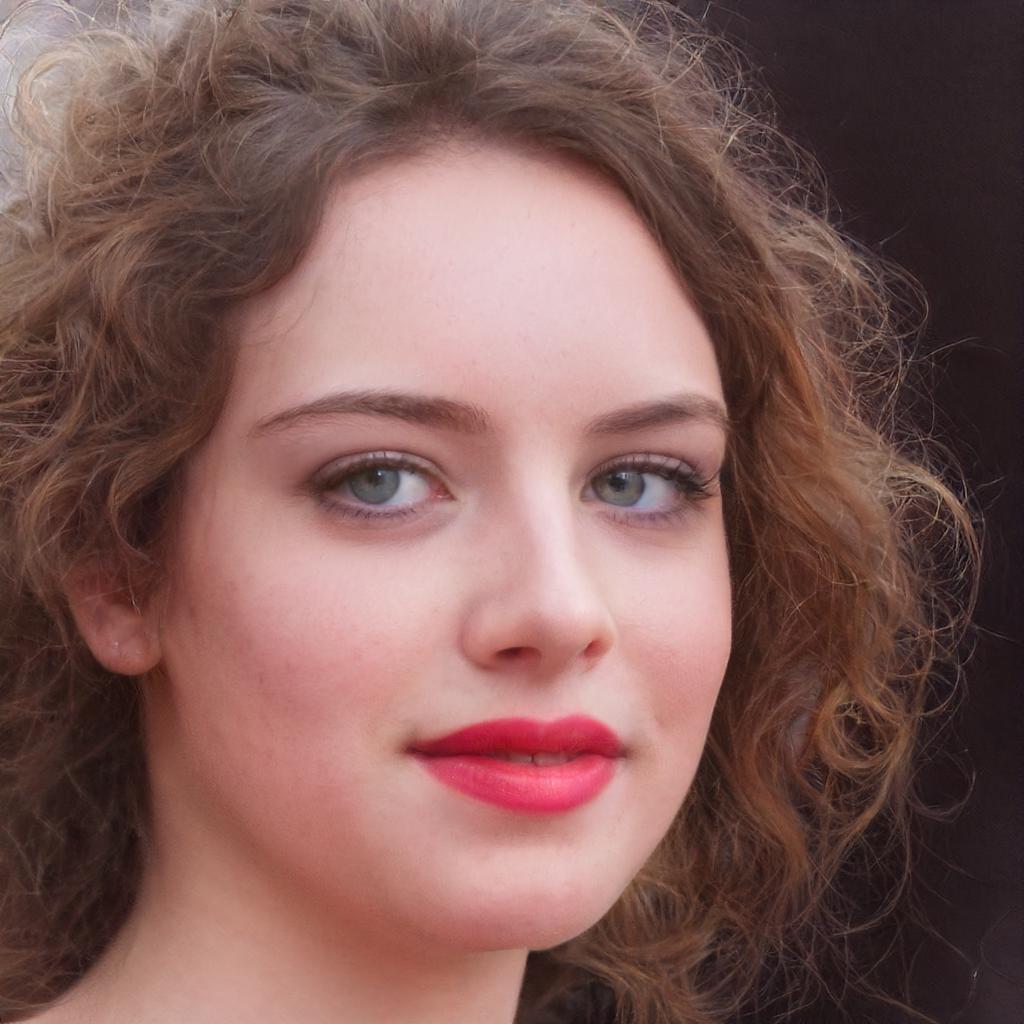} &
			\includegraphics[width=0.19\linewidth]{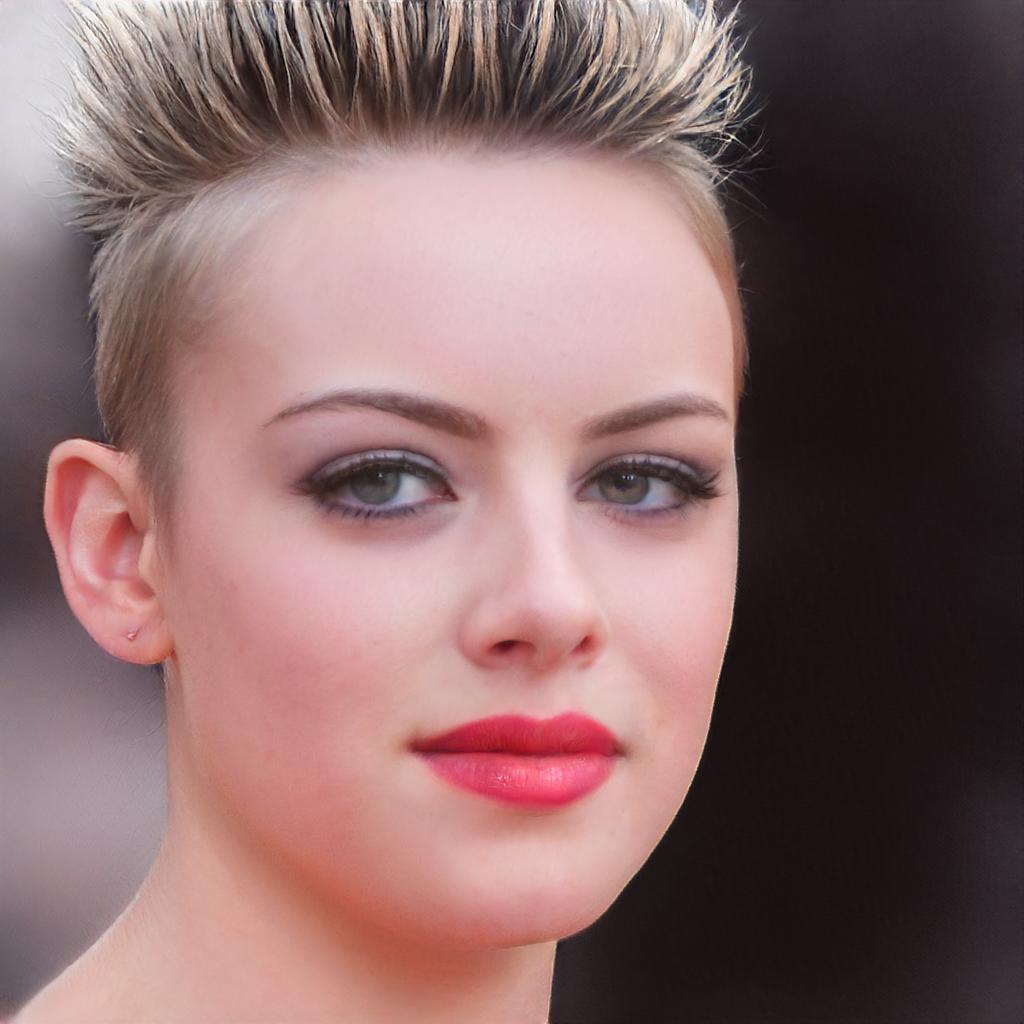} &
            \includegraphics[width=0.19\linewidth]{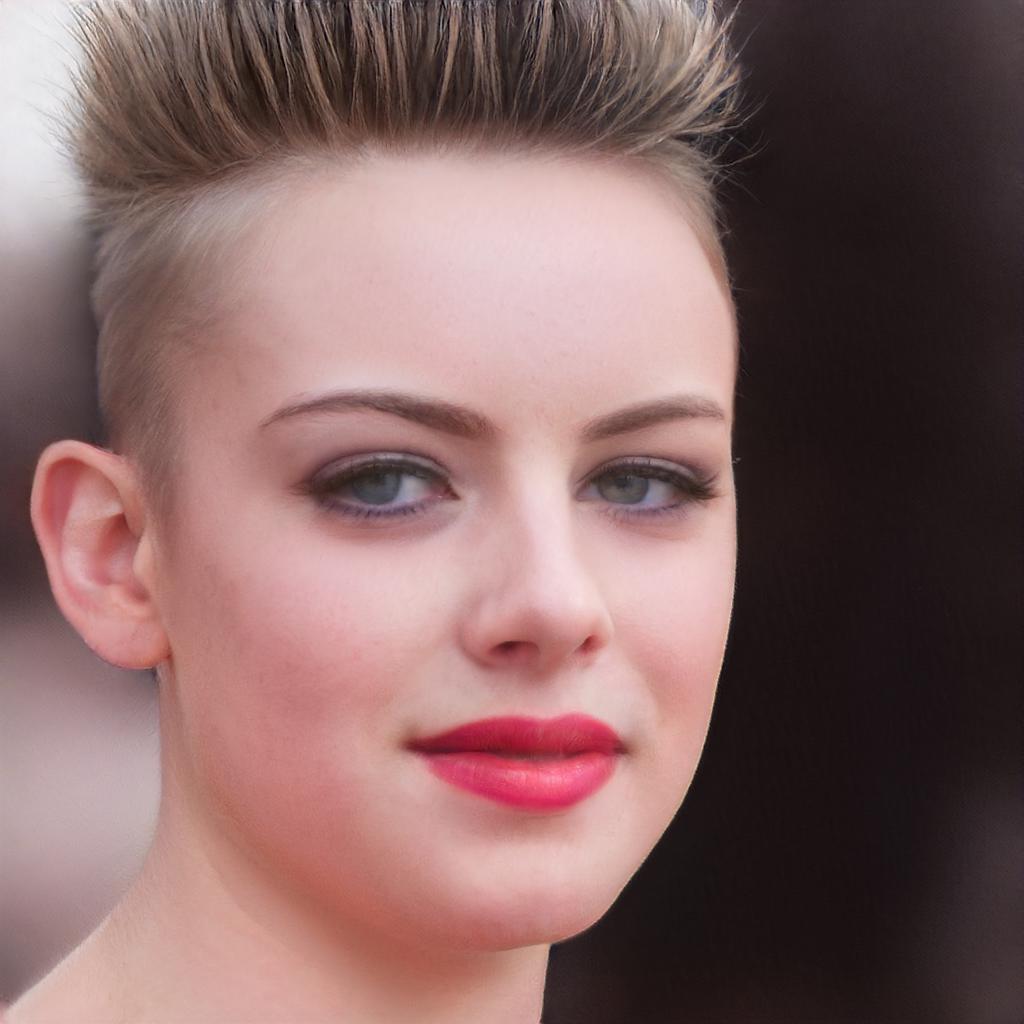} \\
            
			\includegraphics[width=0.19\linewidth]{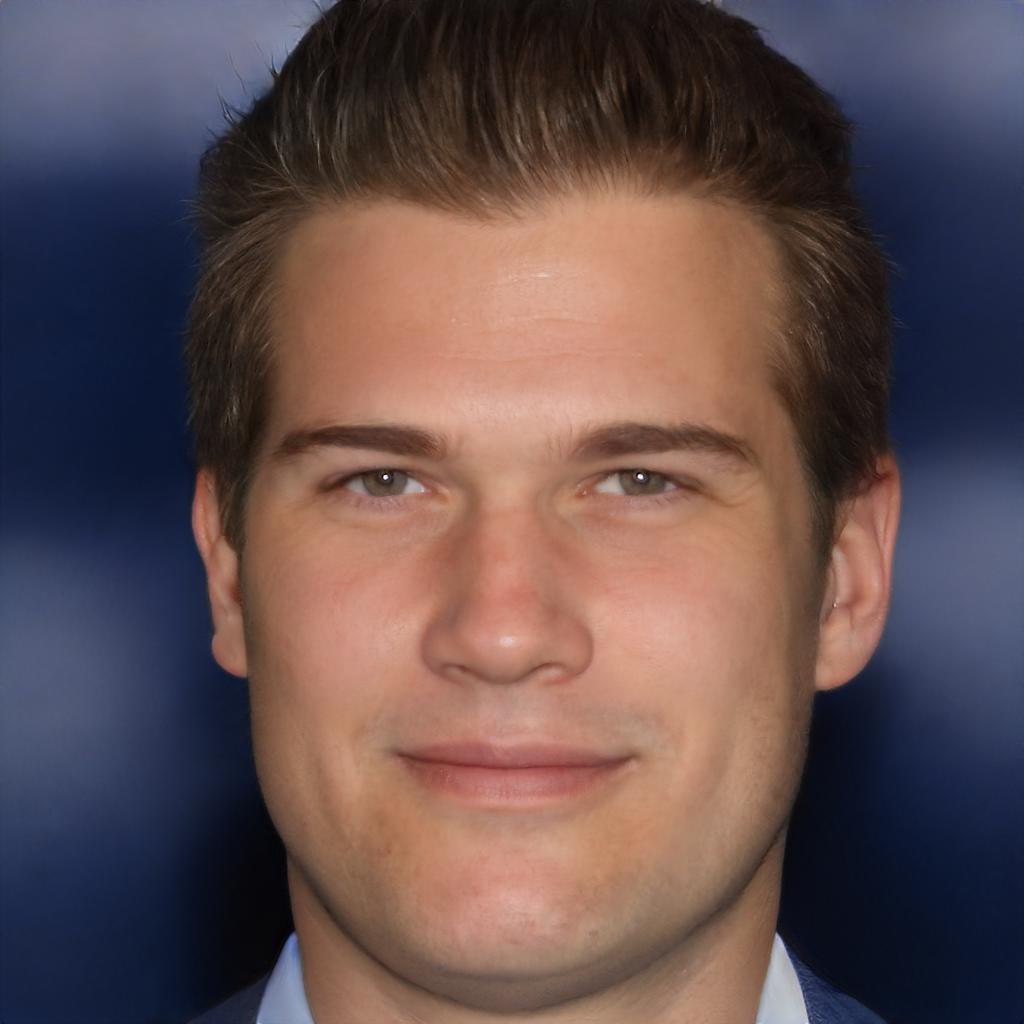} &
            \includegraphics[width=0.19\linewidth]{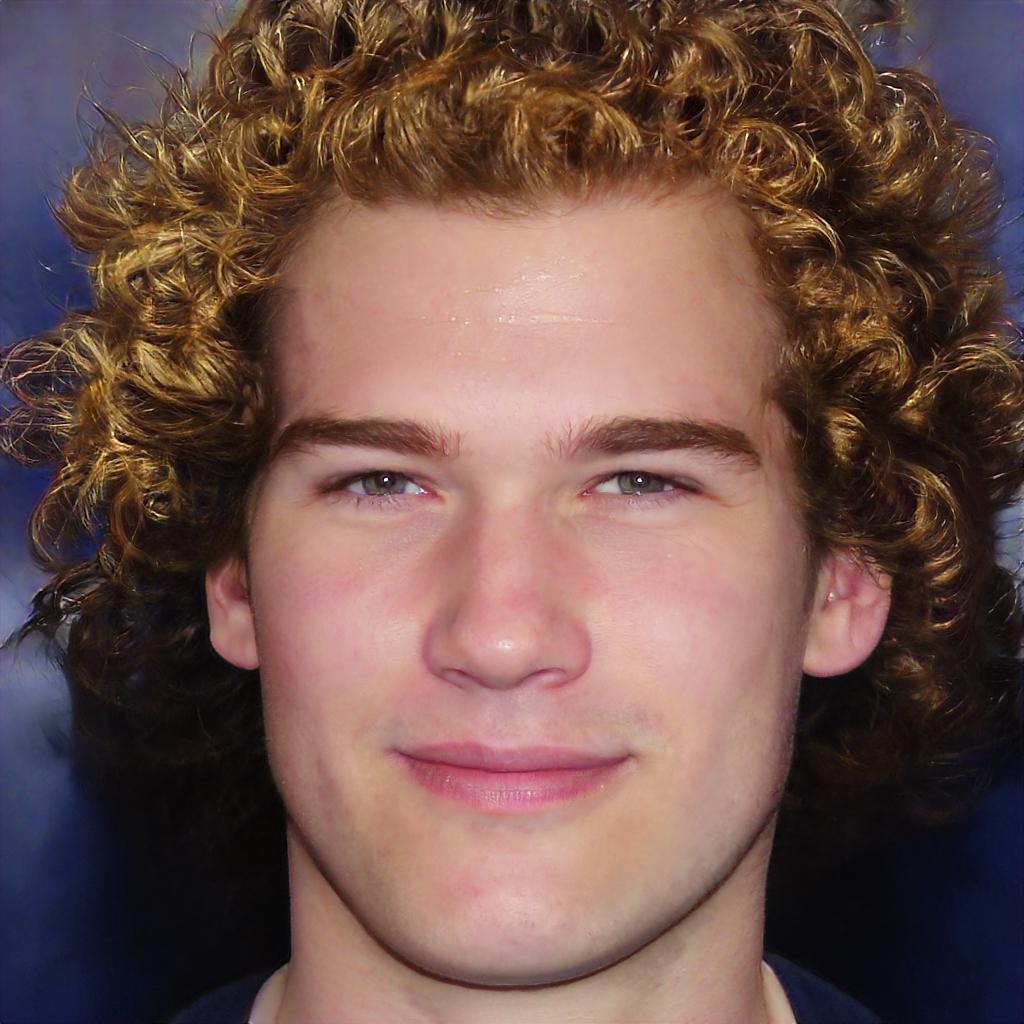} &
			\includegraphics[width=0.19\linewidth]{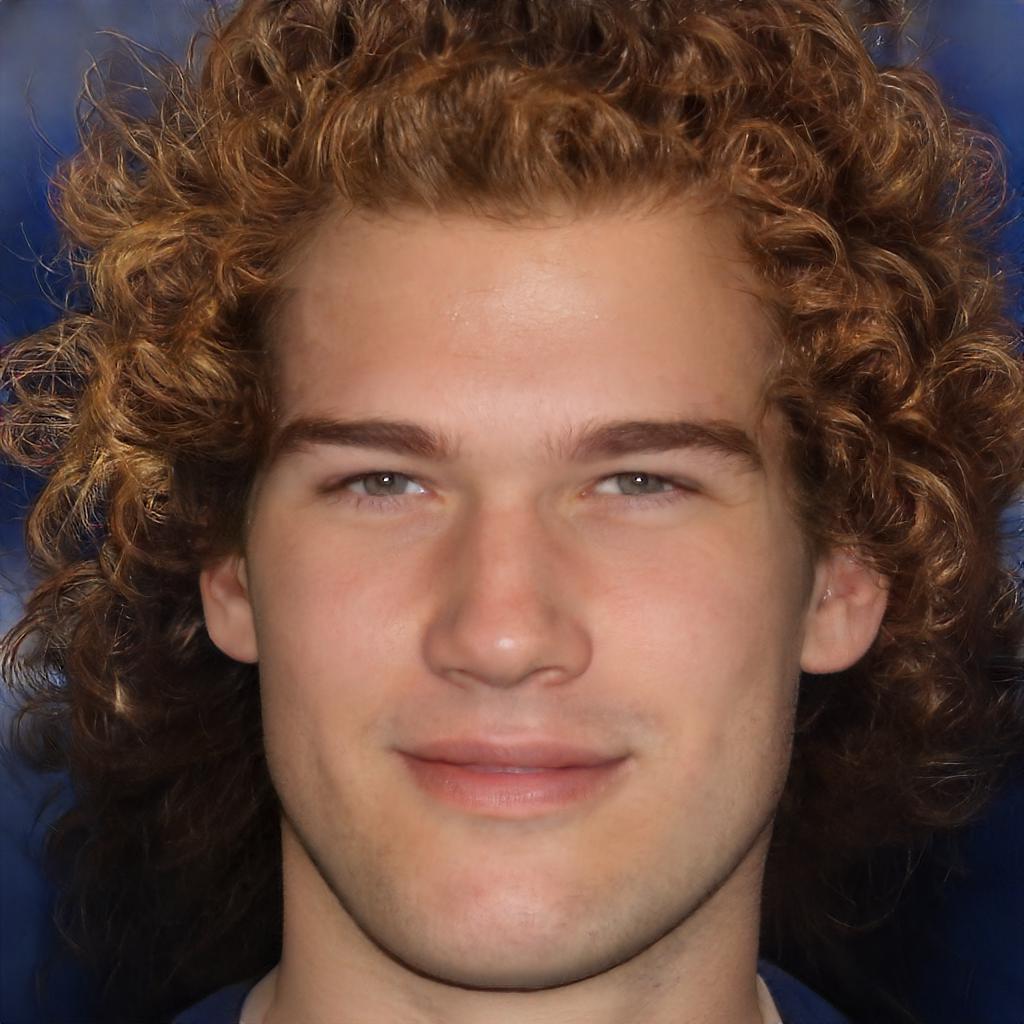} &
			\includegraphics[width=0.19\linewidth]{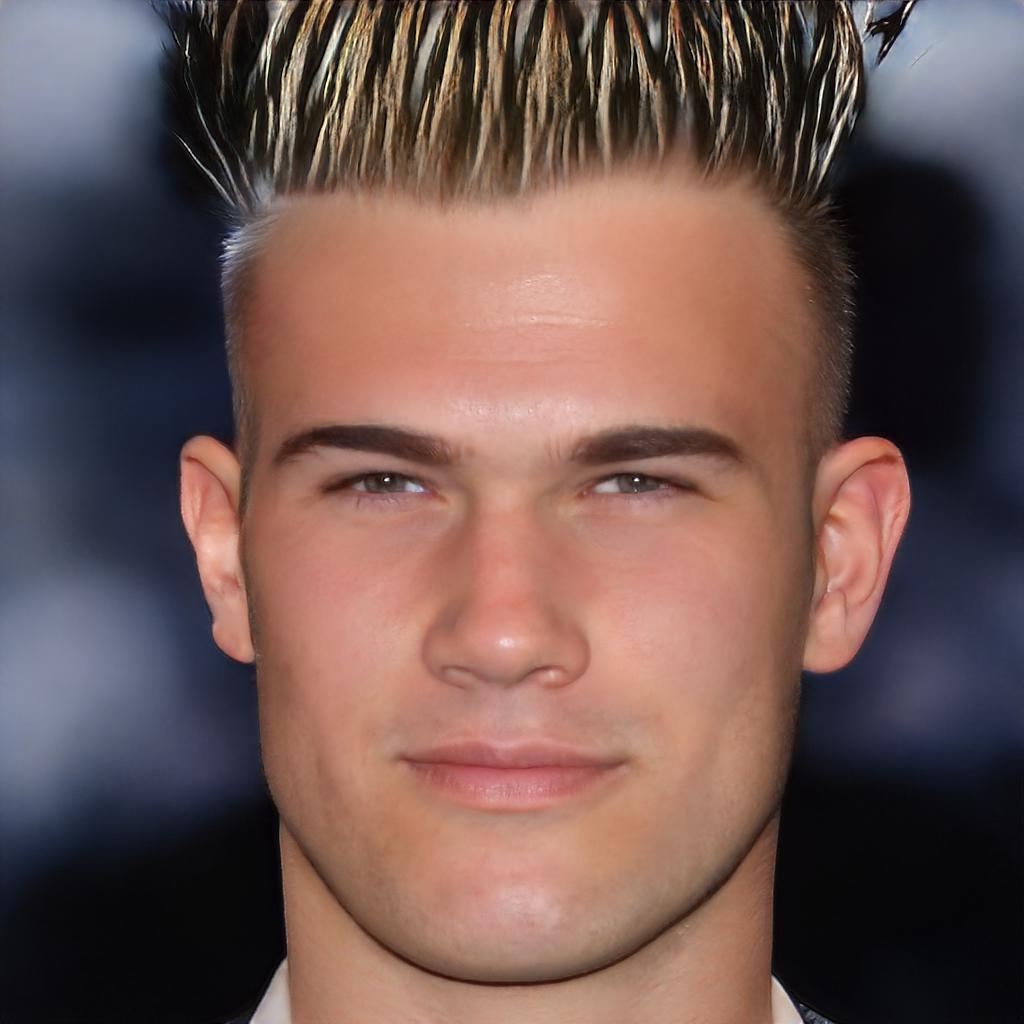} &
            \includegraphics[width=0.19\linewidth]{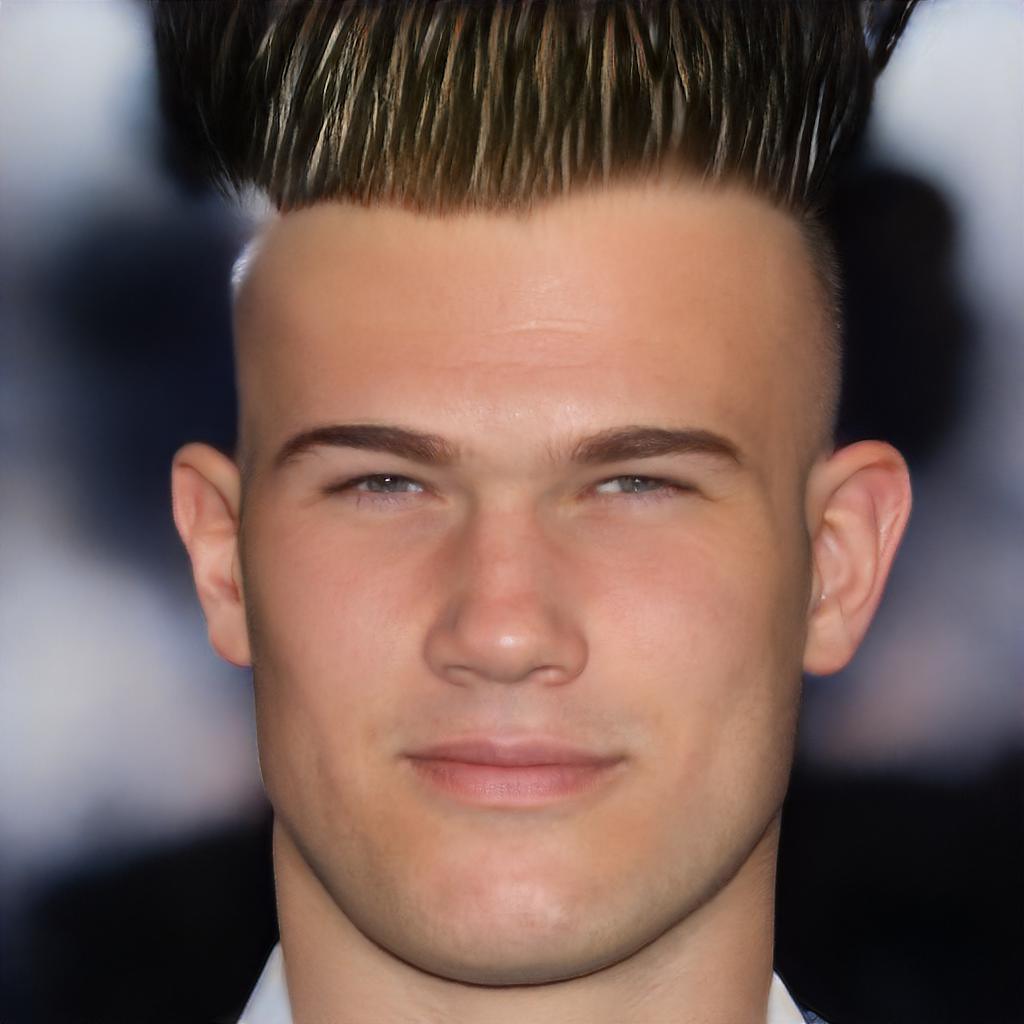} \\
            
            Input & Curly & Curly & Mohawk & Mohawk \\
            & default & without $M^{f}$ & default & without $M^{f}$ \\
		\end{tabular}
	}
	\caption{Removing $M^{f}$ from our full architecture for edits which do not require color scheme manipulation yields slightly better results.}
	\label{fig:ablation-hair}
\end{figure}

Although the full, three network mapper, gives better results for some driving texts, as mentioned in Section 5, we note that not all the three are needed when the manipulation should not affect some attributes. For example, for the hairstyle edits shown in Figure 5, the manipulation should not affect the color scheme of the image. Therefore, we perform these edits when training $M^c$ and $M^m$ only, that is, $M_t(w) = (M^c_t(w_c), M^m_t(w_m), 0)$.
We show a comparison in Figure~\ref{fig:ablation-hair}. As can be seen, by removing $M_f$ from the architecture, we get slightly better results. Therefore, for the sake of simplicity and generalization of the method, we chose to describe the method with all three networks. In the main paper, the results shown were obtained with all three networks, while here we also show results with only two (without $M_f$).

\subsection{Losses}

\begin{figure}[tb]
	\setlength{\tabcolsep}{1pt}
	\centering
	{\footnotesize
		\begin{tabular}{c c c}
			\includegraphics[width=0.32\linewidth]{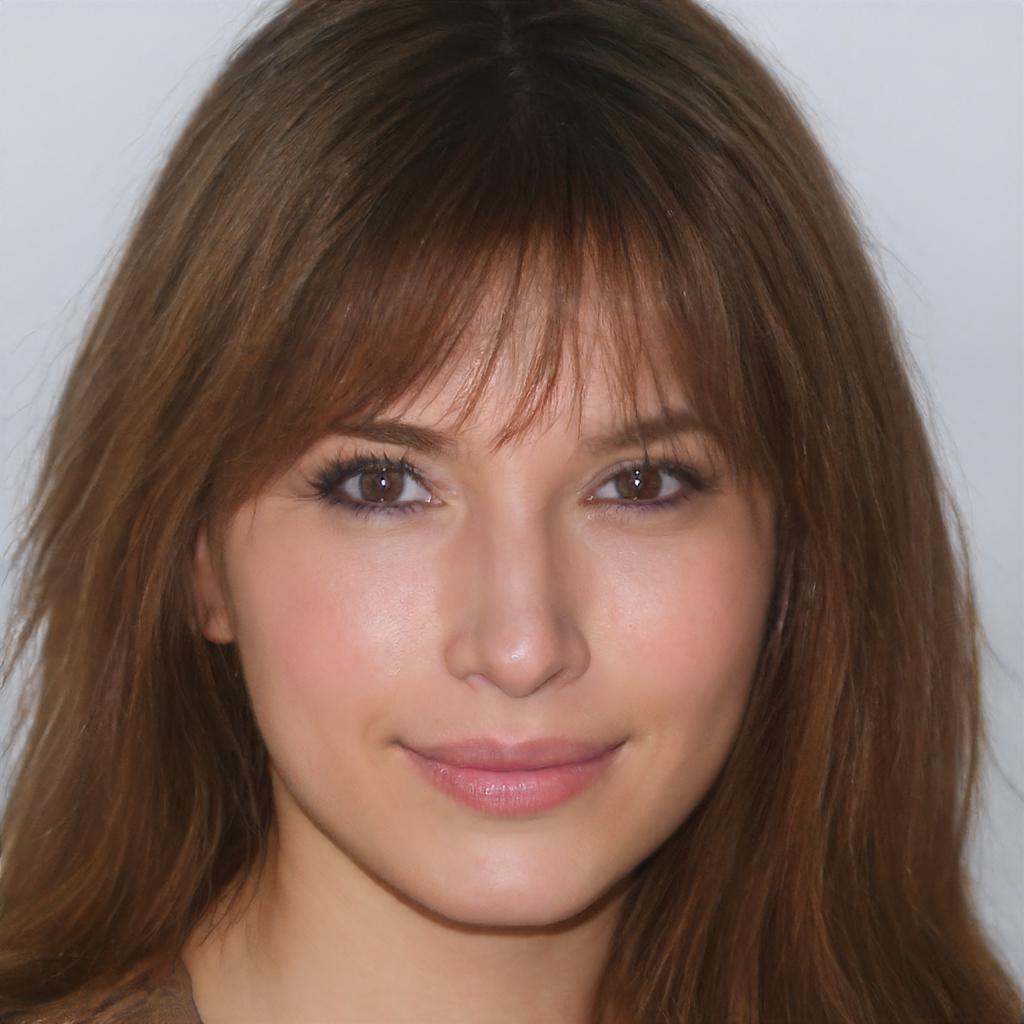} &
            \includegraphics[width=0.32\linewidth]{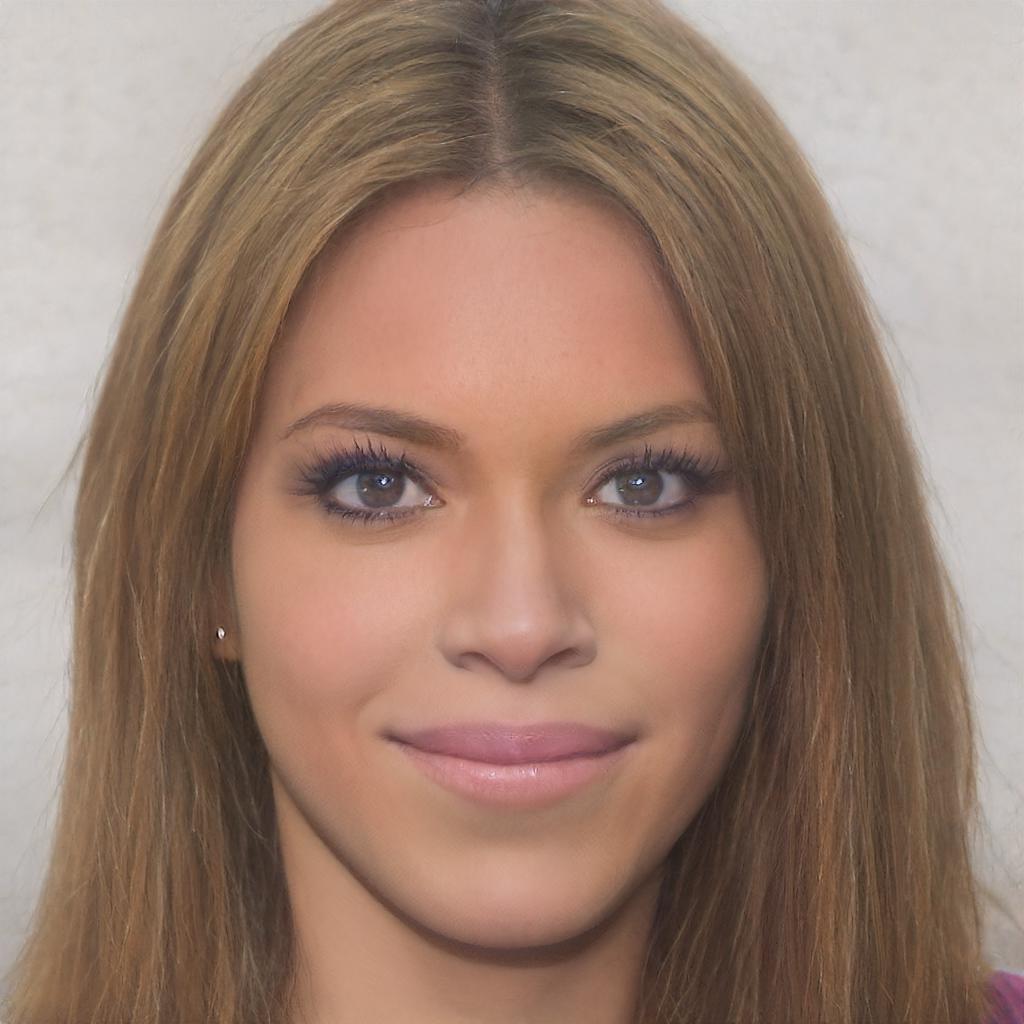} &
			\includegraphics[width=0.32\linewidth]{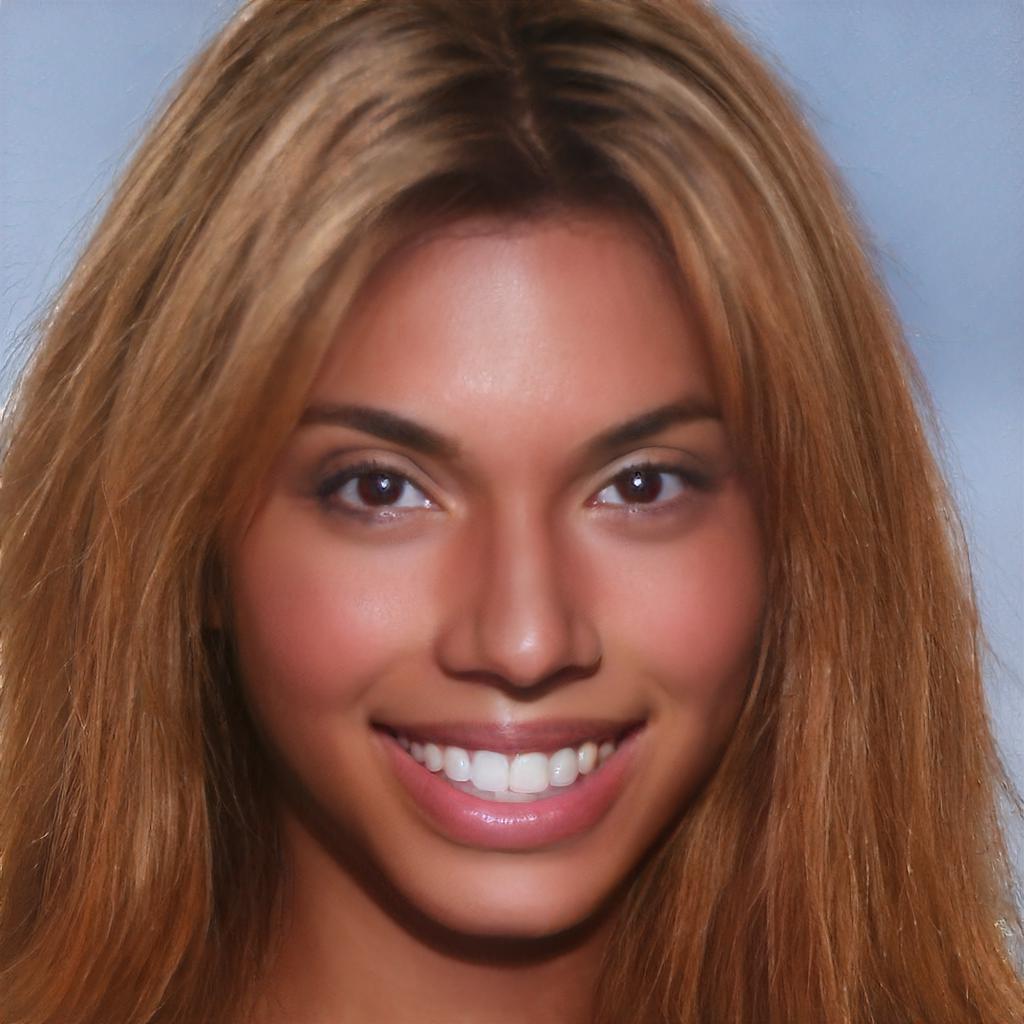}  \\
			
			\includegraphics[width=0.32\linewidth]{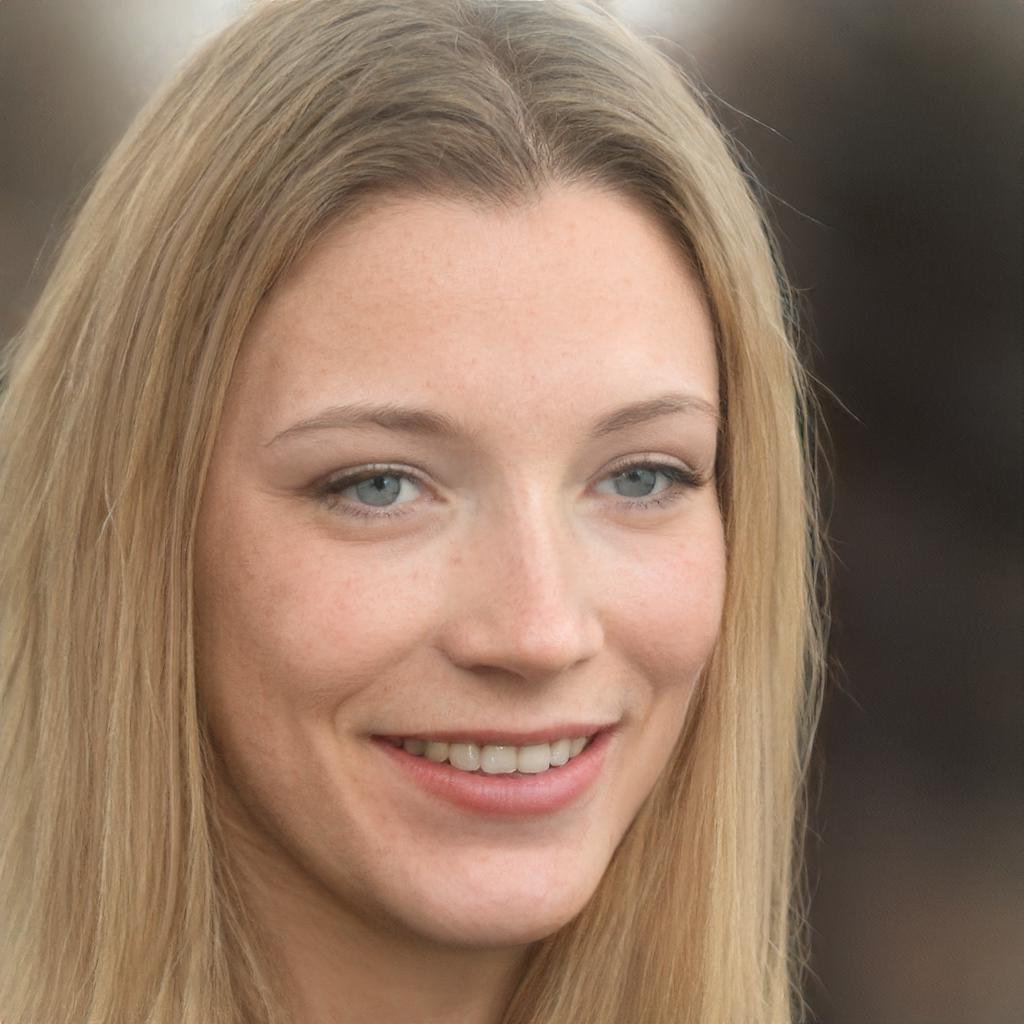} &
            \includegraphics[width=0.32\linewidth]{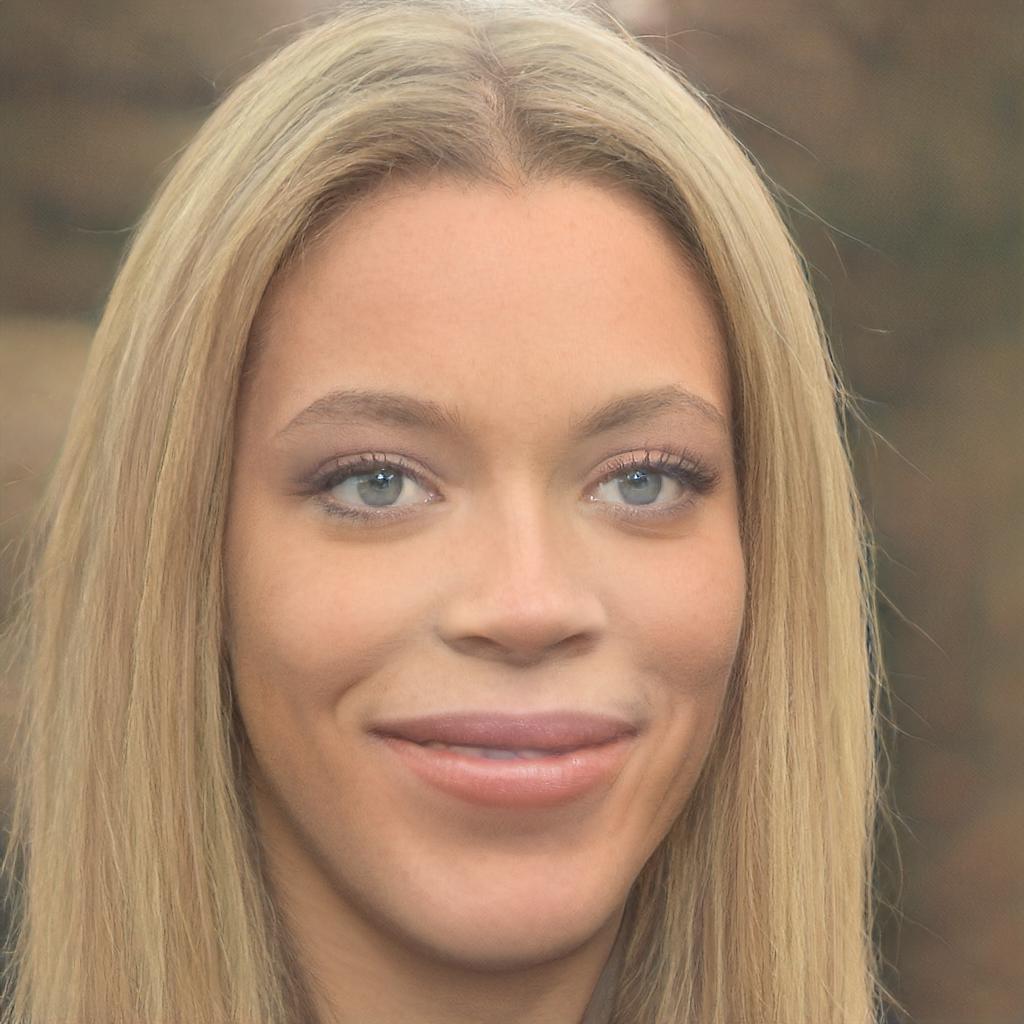} &
			\includegraphics[width=0.32\linewidth]{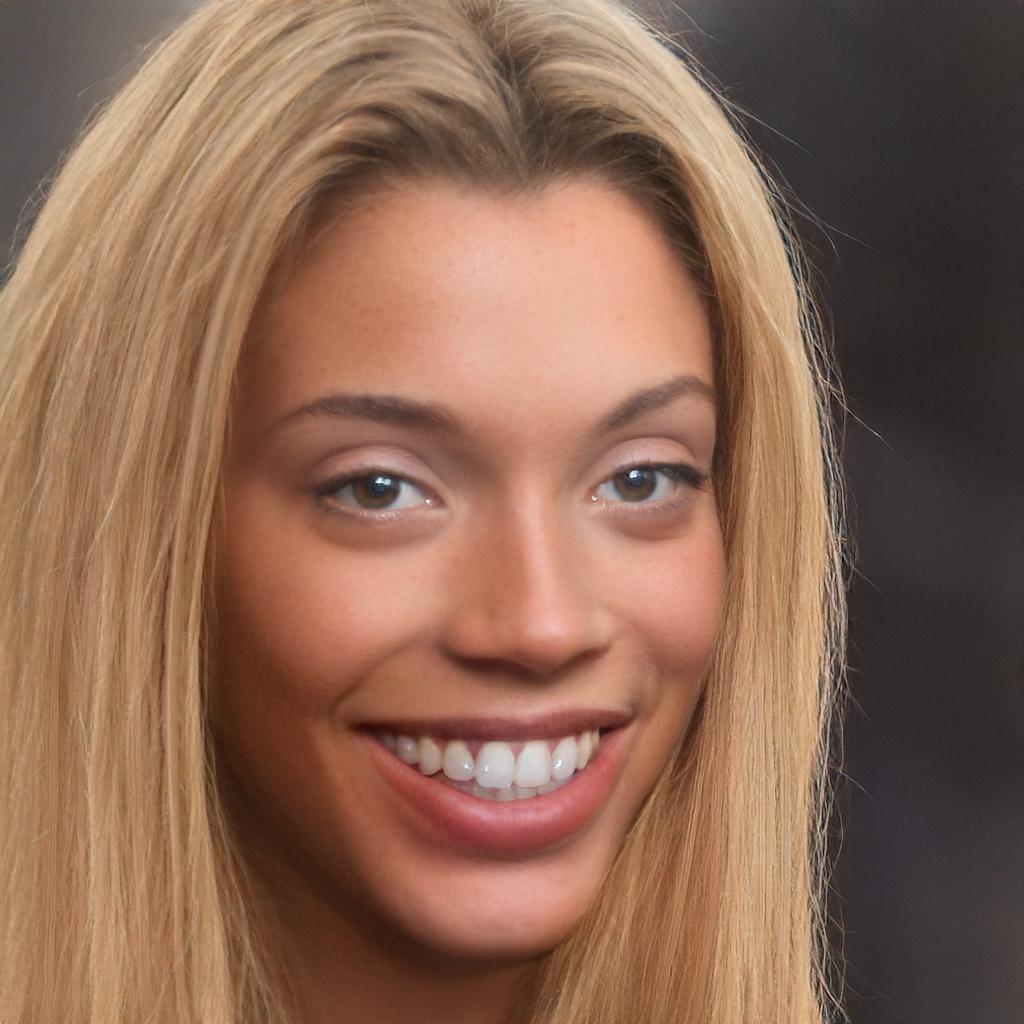}  \\

            Input & ID Loss & CLIP Loss
		\end{tabular}
	}
	\caption{Replacing the CLIP loss with identity loss for the Beyonce edit. The identity loss is computed with respect to an image of Beyonce.}
	\label{fig:ablation-beyonce}
\end{figure}

\paragraph{CLIP Loss}
To show the uniqueness of using a ``celeb edit'' with CLIP, we perform the following experiment. Instead of using the CLIP loss, we use the identity loss with respect to a single image of the desired celeb. Specifically, we perform this experiment by using an image of Beyonce. The results are shown in Figure~\ref{fig:ablation-beyonce}. As can be seen, CLIP guides the mapper to perform a unique edit which cannot be achieved by simply using a facial recognition network.

\begin{figure}[tb]
	\setlength{\tabcolsep}{1pt}
	\centering
	{\footnotesize
		\begin{tabular}{c c c c}
			\includegraphics[width=0.24\linewidth]{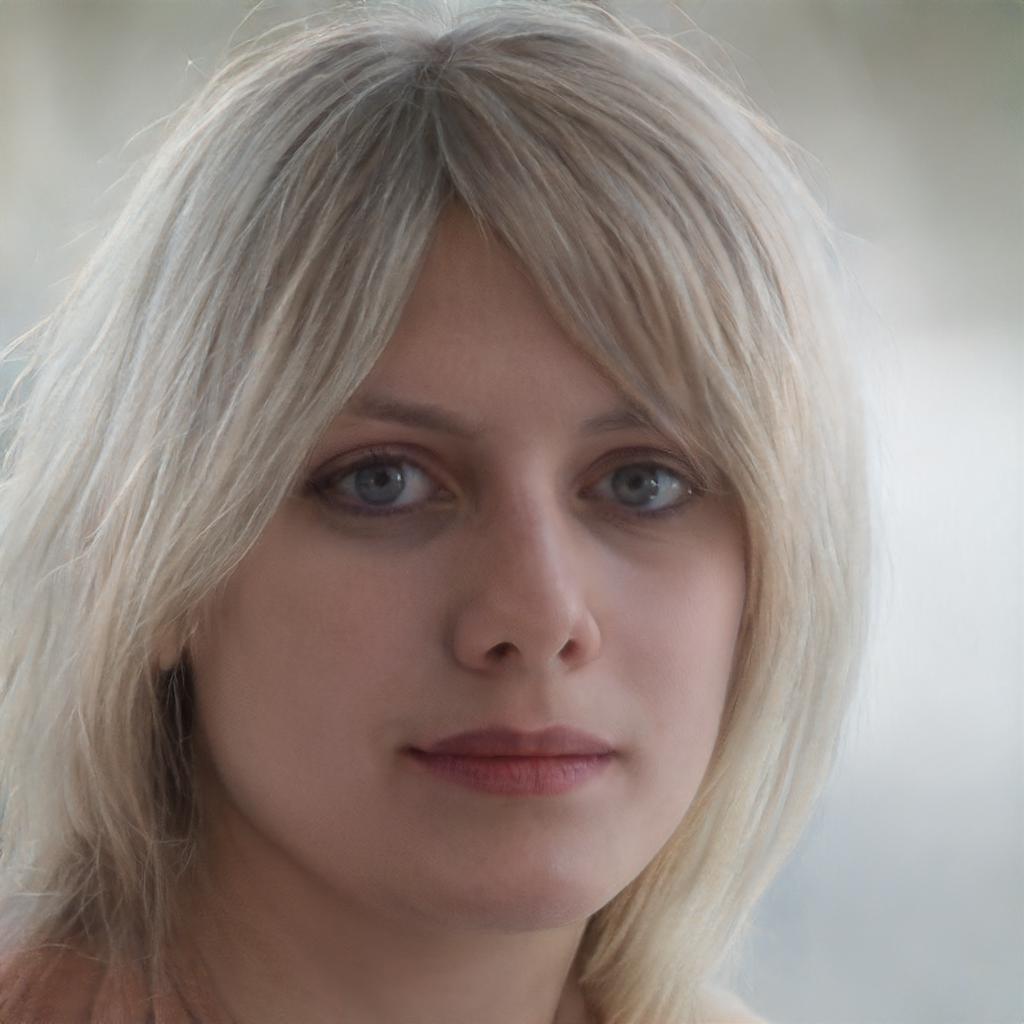} &
            \includegraphics[width=0.24\linewidth]{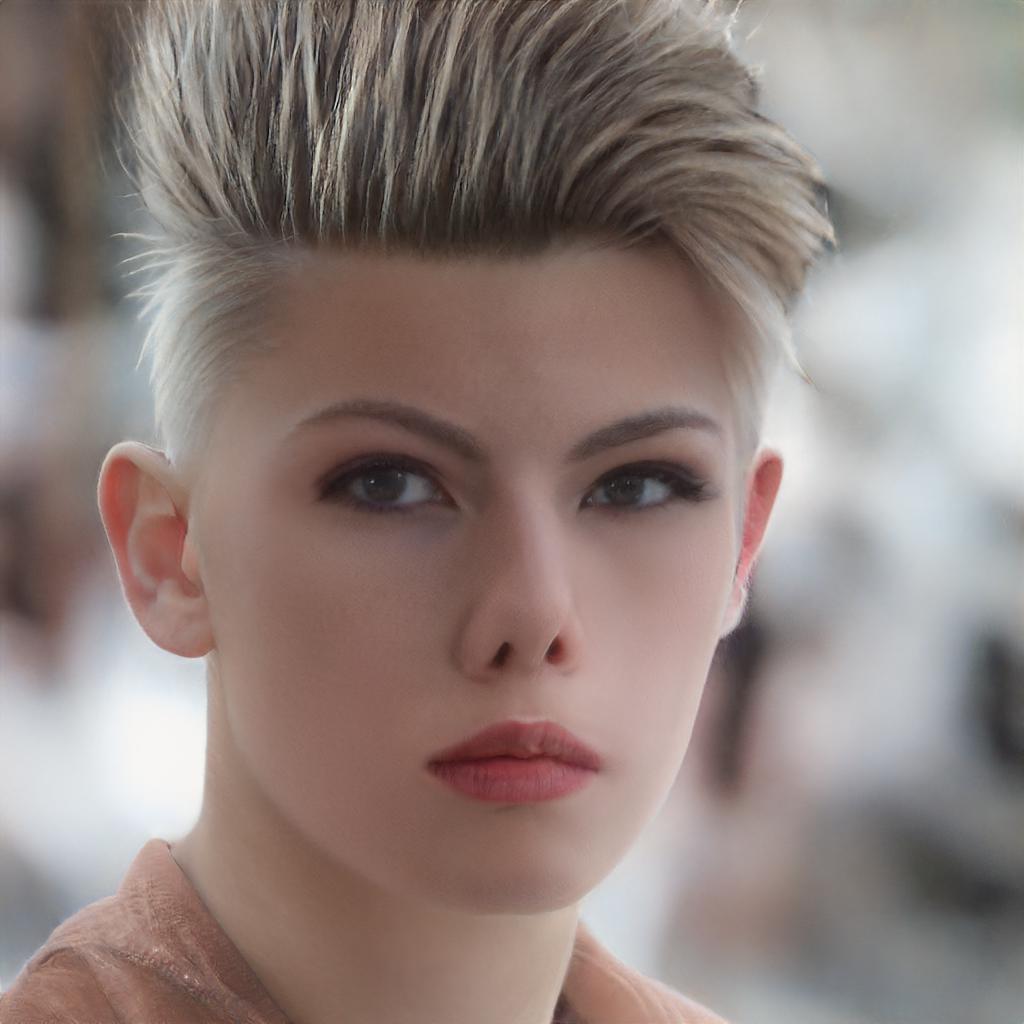} &
			\includegraphics[width=0.24\linewidth]{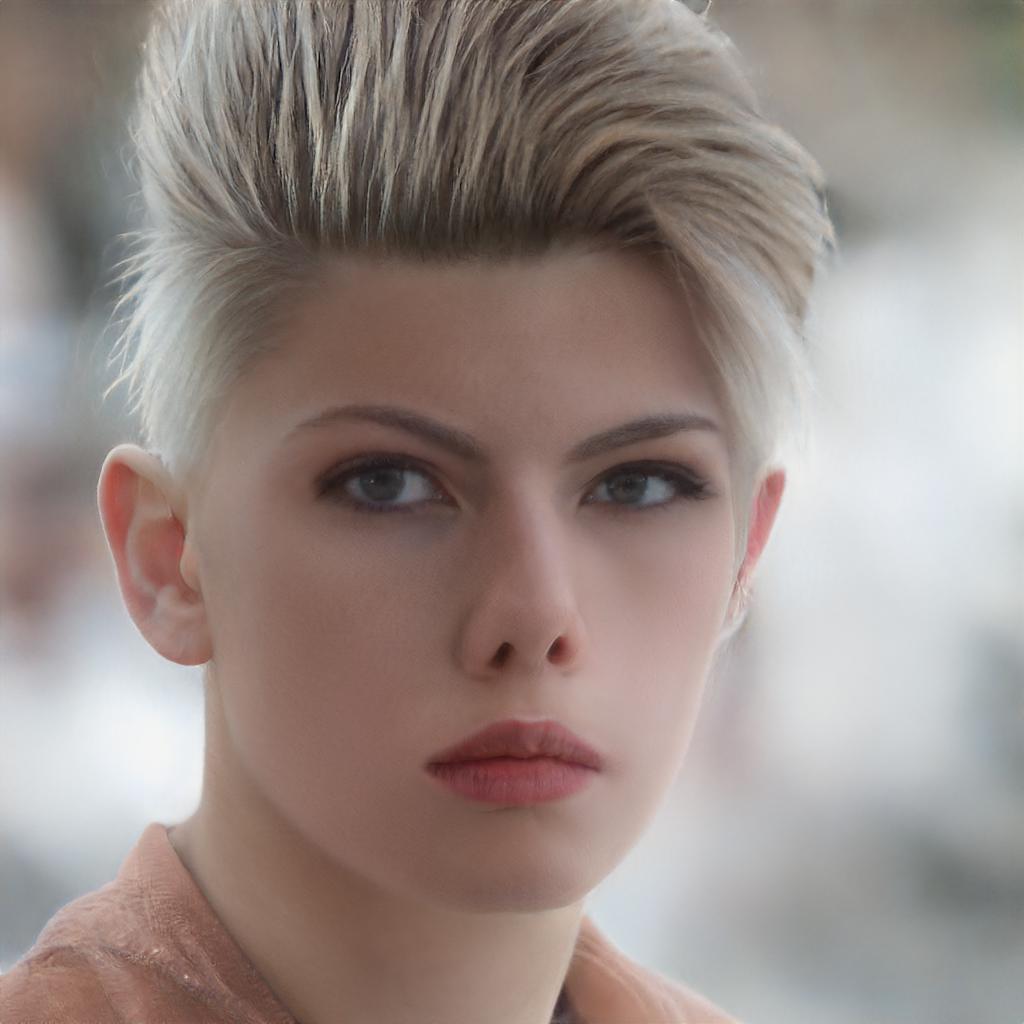} 
			&
			\includegraphics[width=0.24\linewidth]{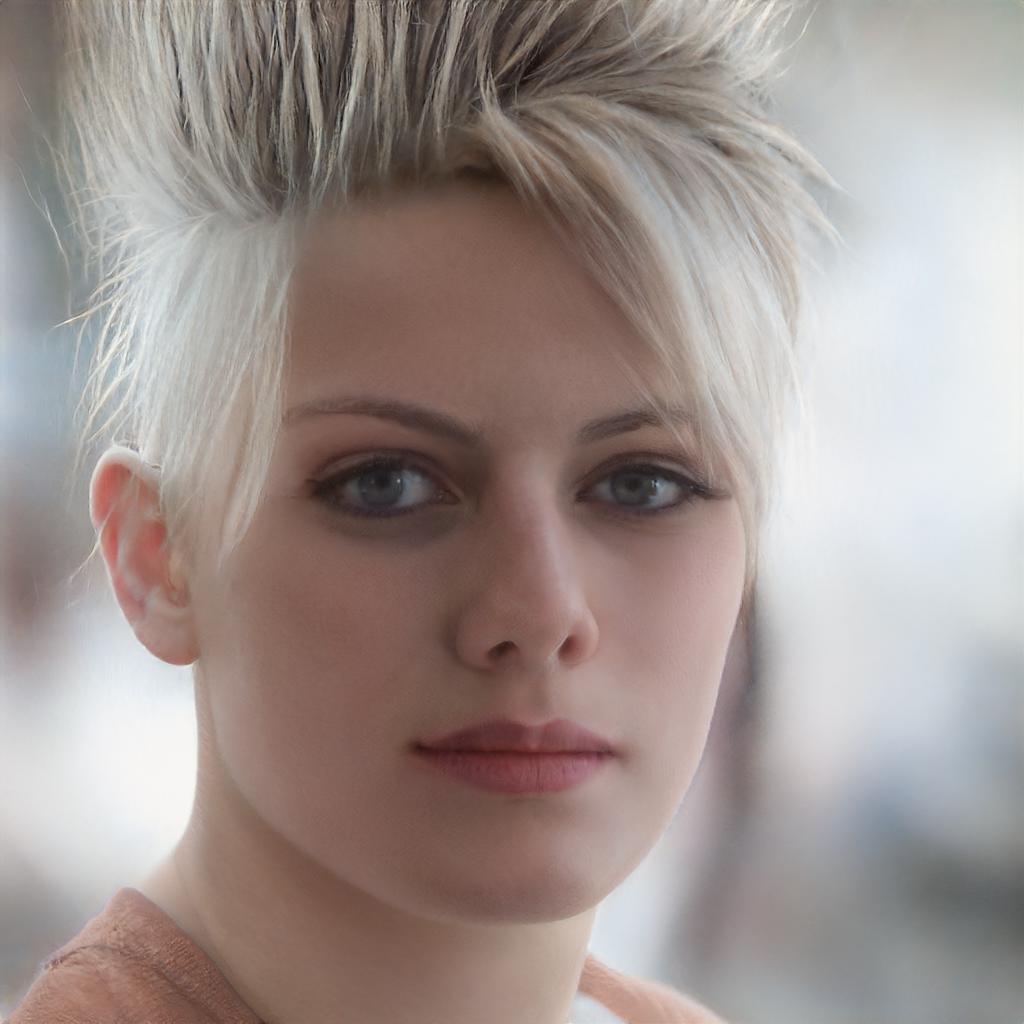} 
			\\

			\includegraphics[width=0.24\linewidth]{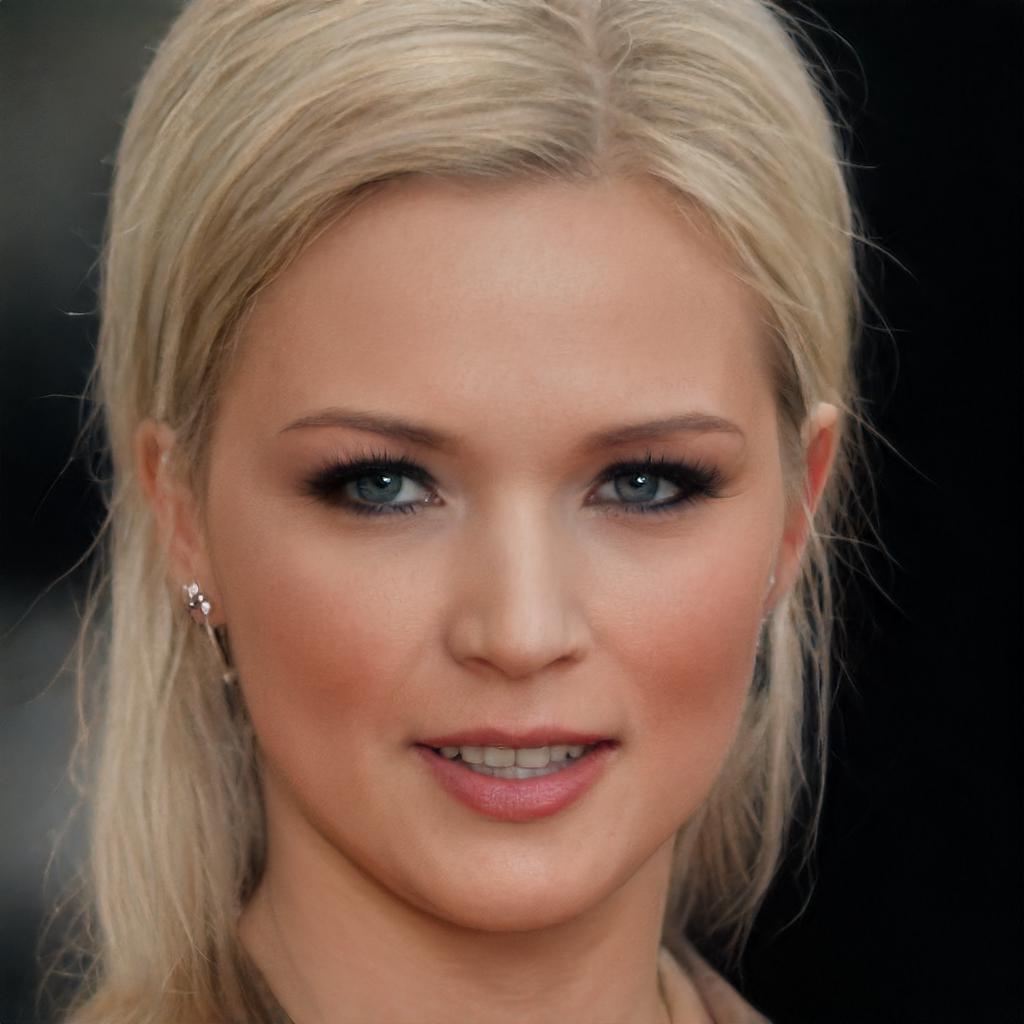} &
            \includegraphics[width=0.24\linewidth]{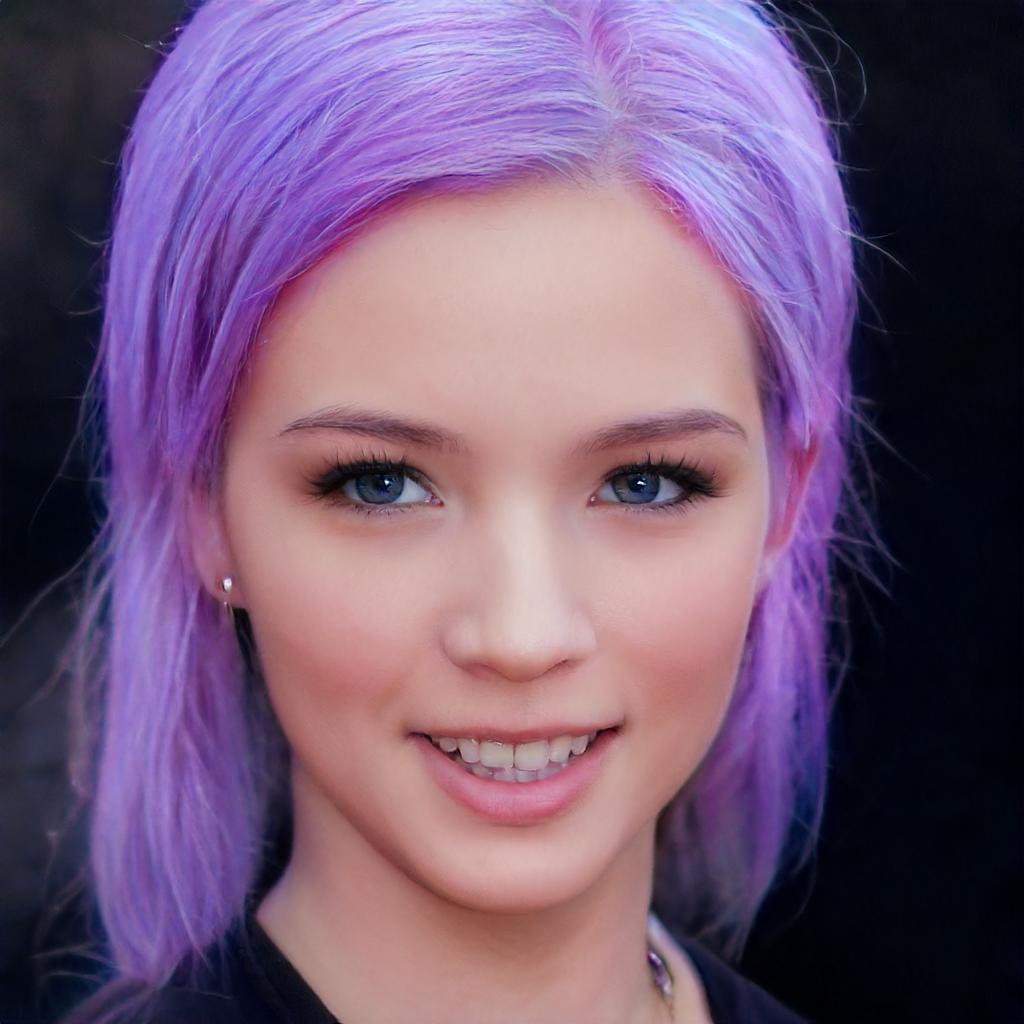} &
			\includegraphics[width=0.24\linewidth]{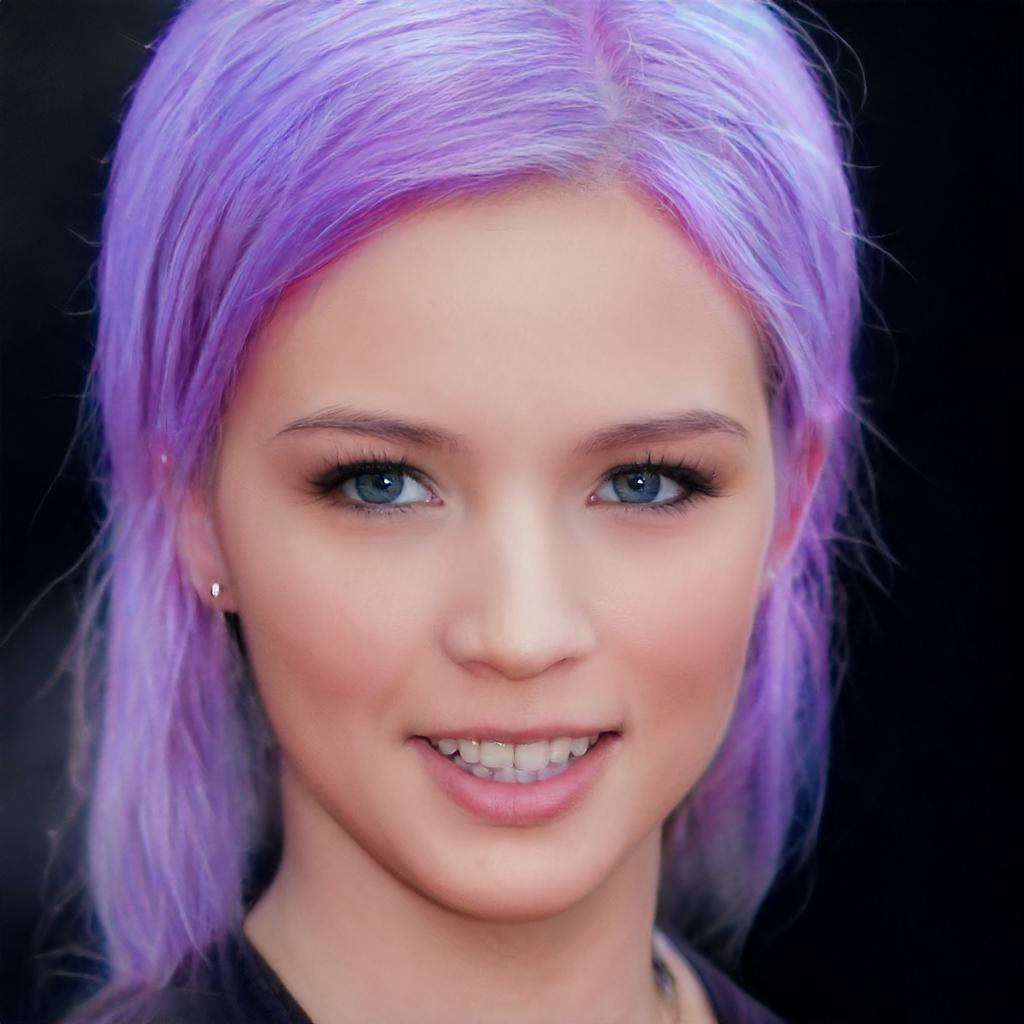} 
			&
			\includegraphics[width=0.24\linewidth]{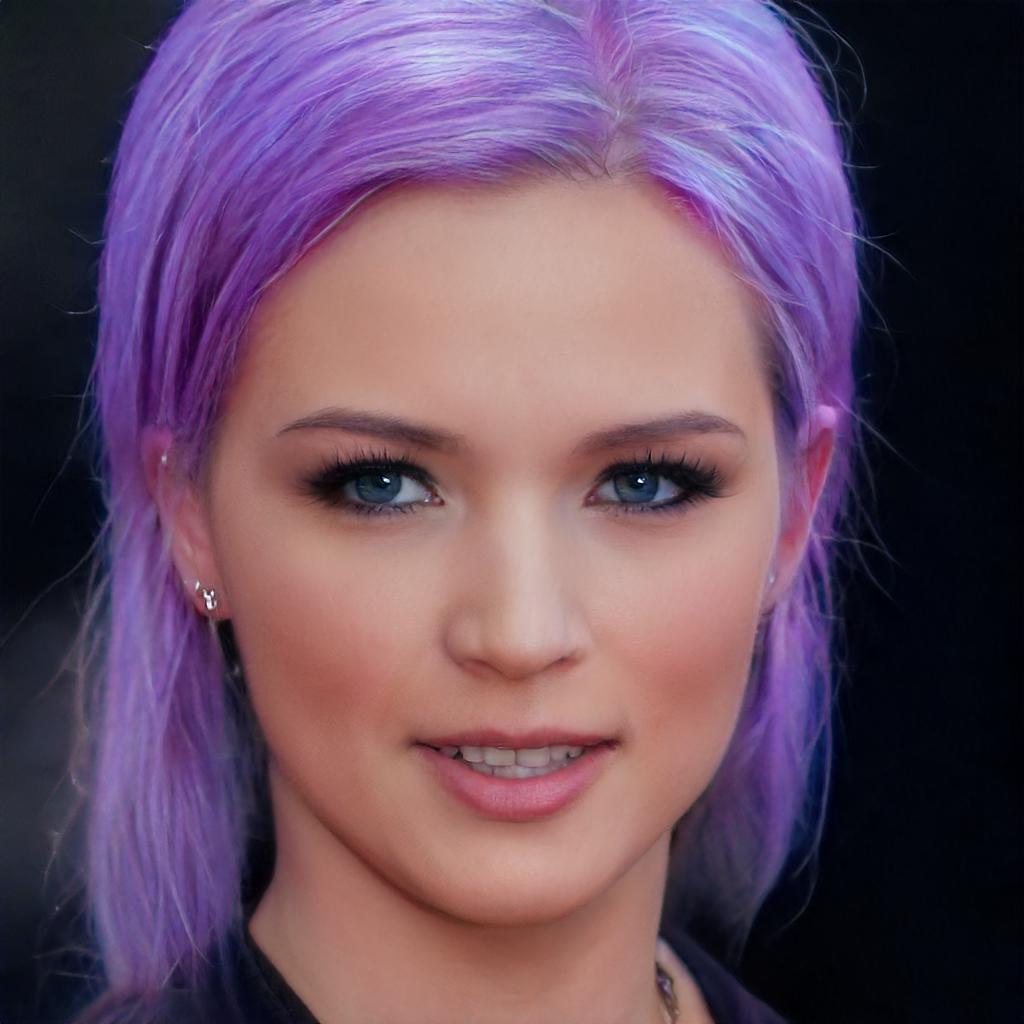} 
			\\
			
			\includegraphics[width=0.24\linewidth]{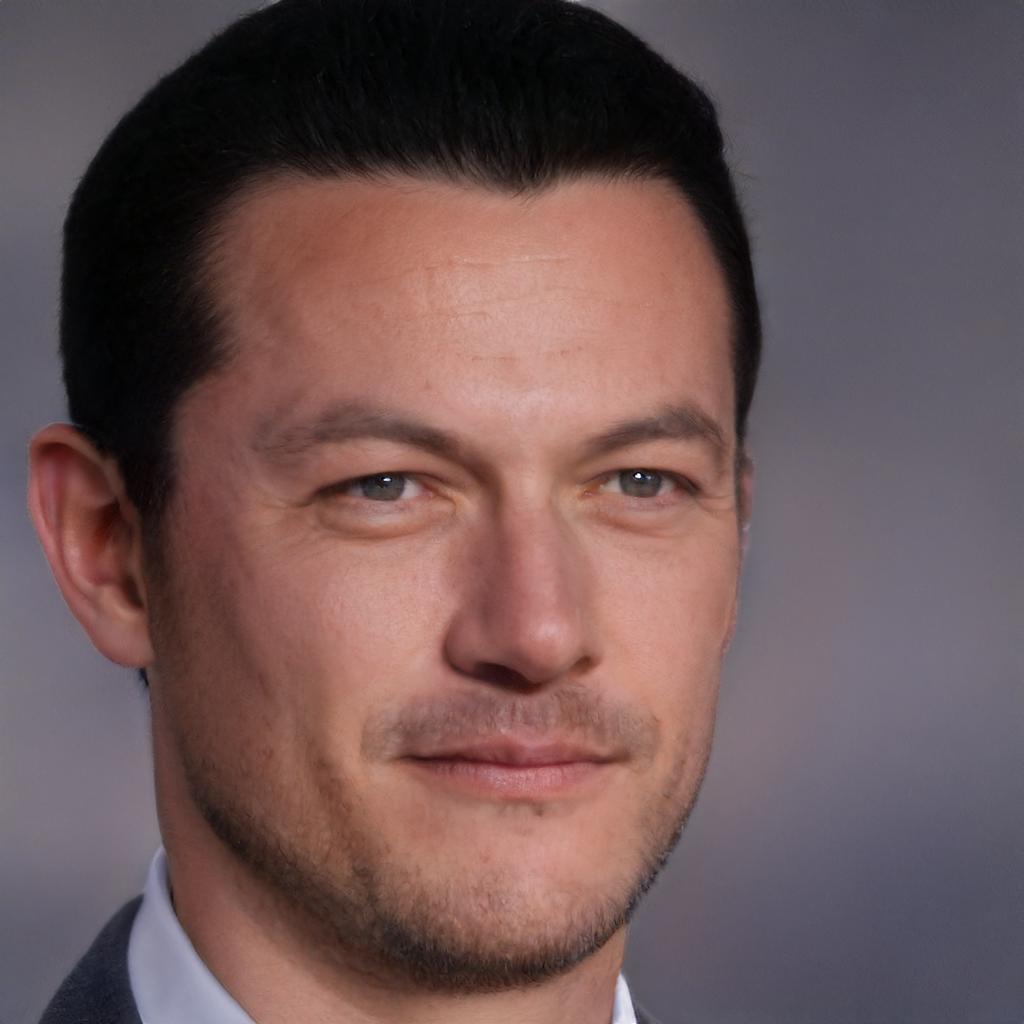} &
            \includegraphics[width=0.24\linewidth]{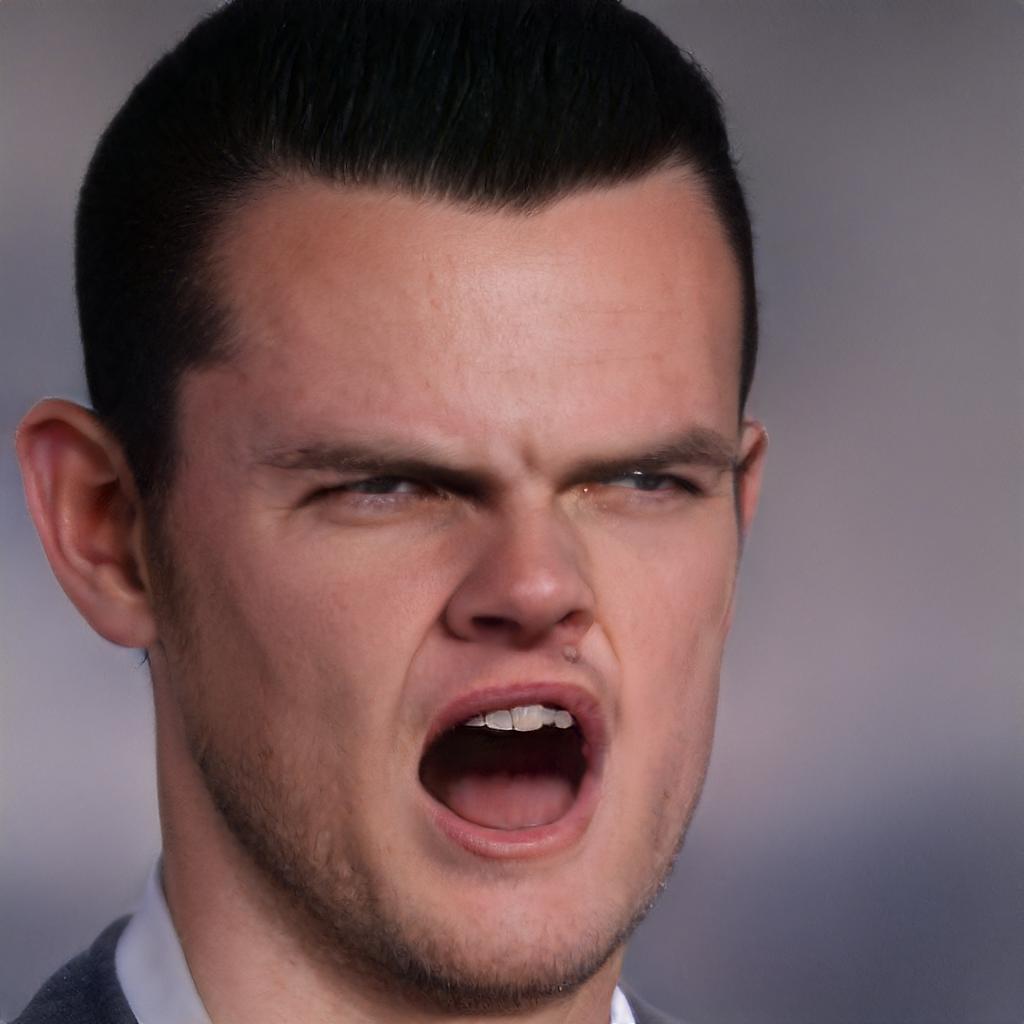} &
			\includegraphics[width=0.24\linewidth]{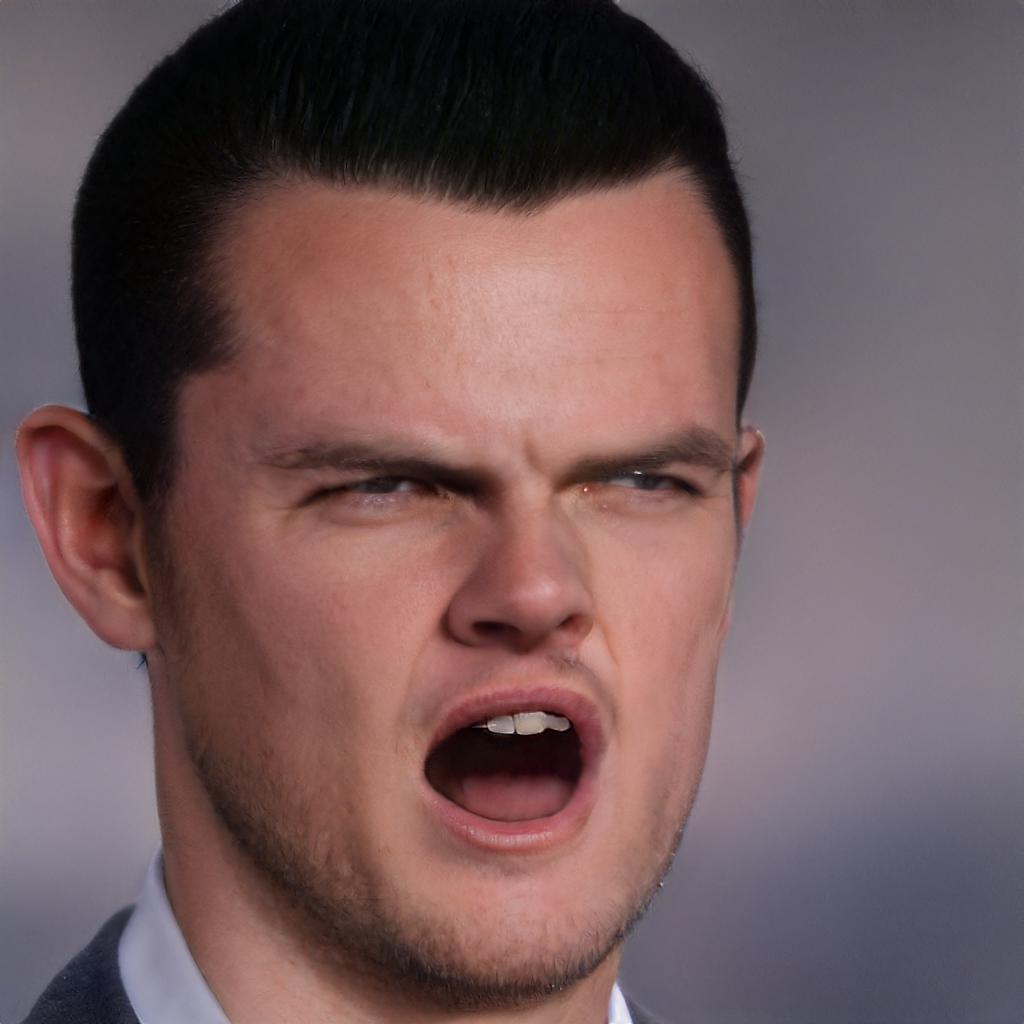} 
			&
			\includegraphics[width=0.24\linewidth]{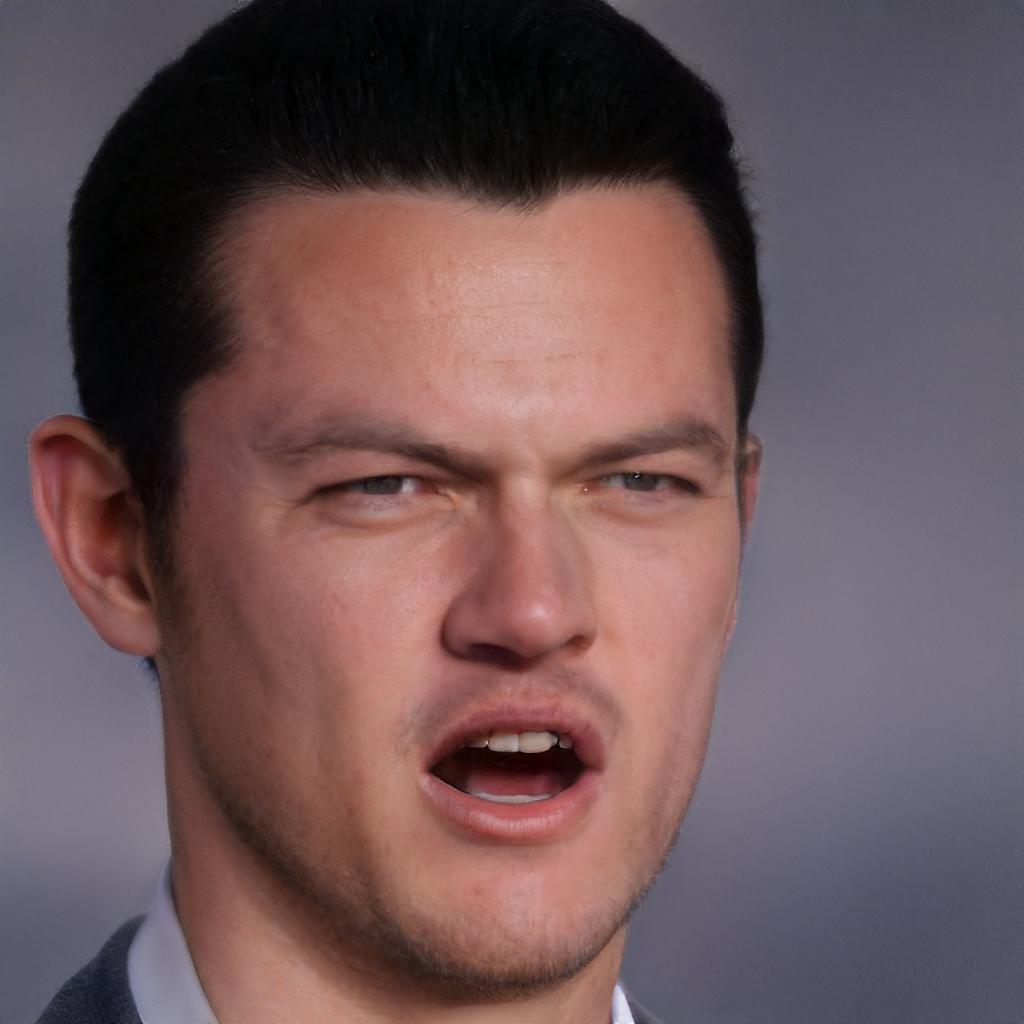} 
			\\
			
			\includegraphics[width=0.24\linewidth]{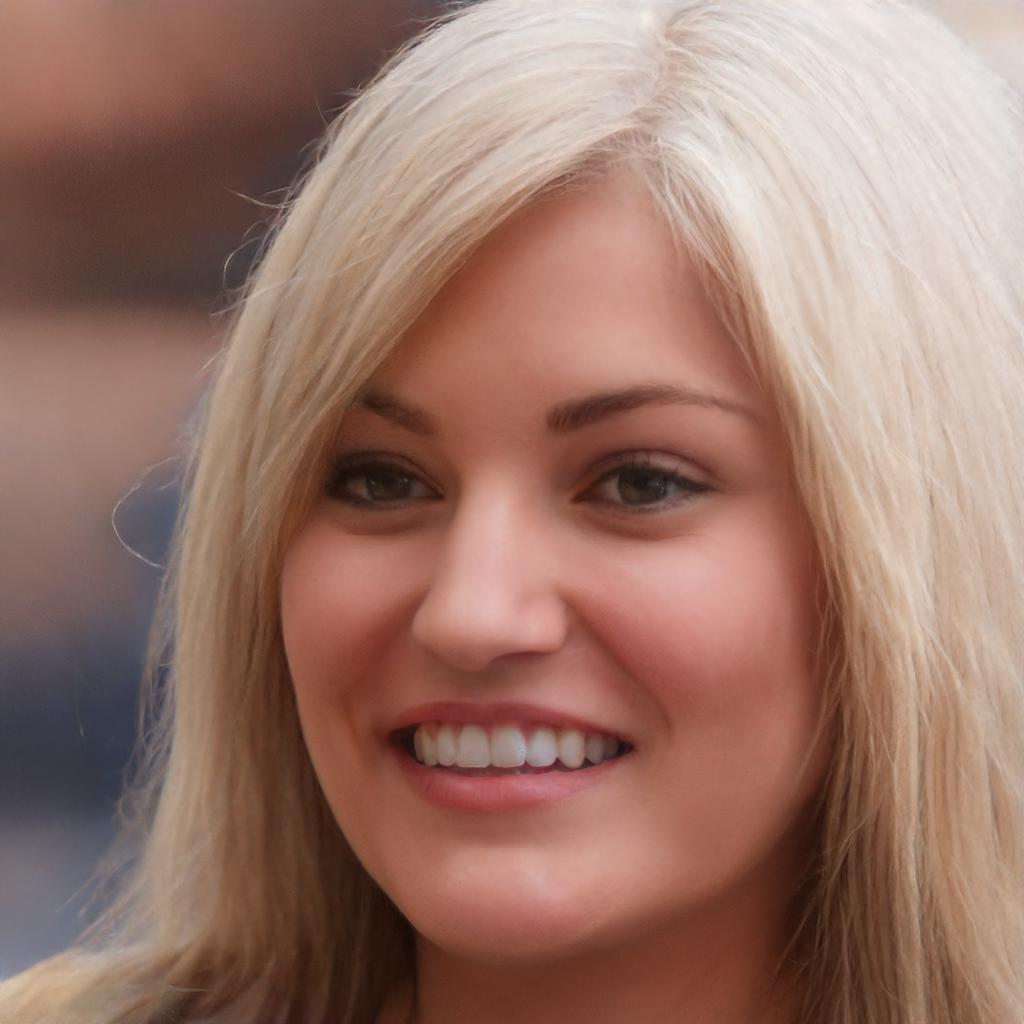} &
            \includegraphics[width=0.24\linewidth]{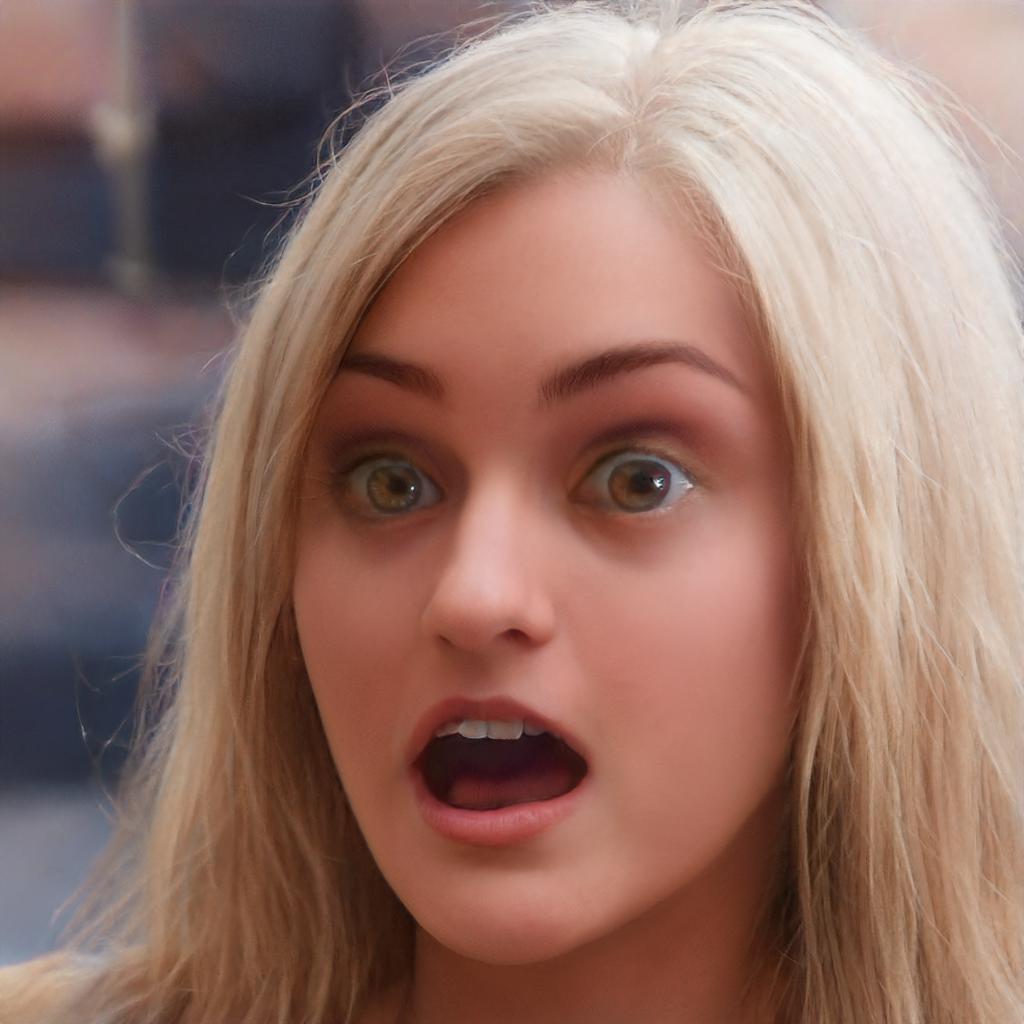} &
			\includegraphics[width=0.24\linewidth]{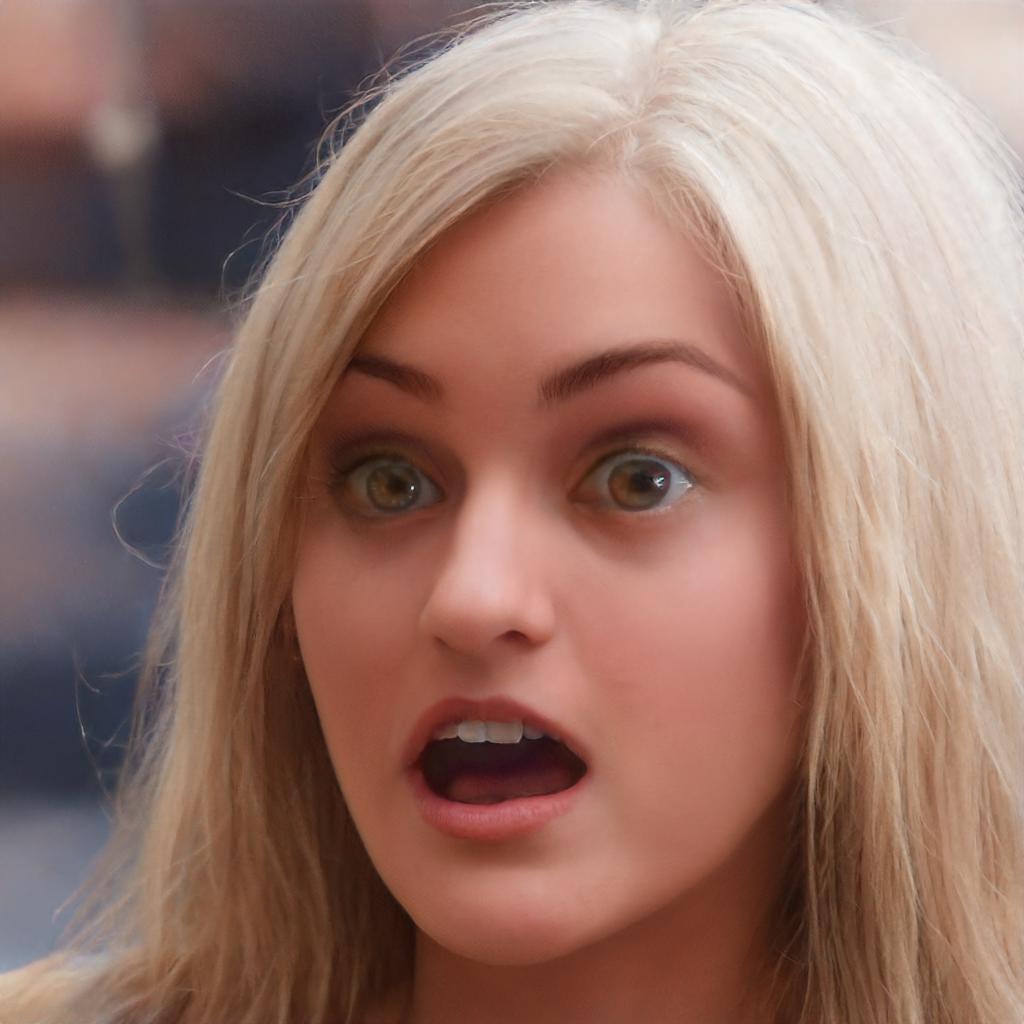} 
			&
			\includegraphics[width=0.24\linewidth]{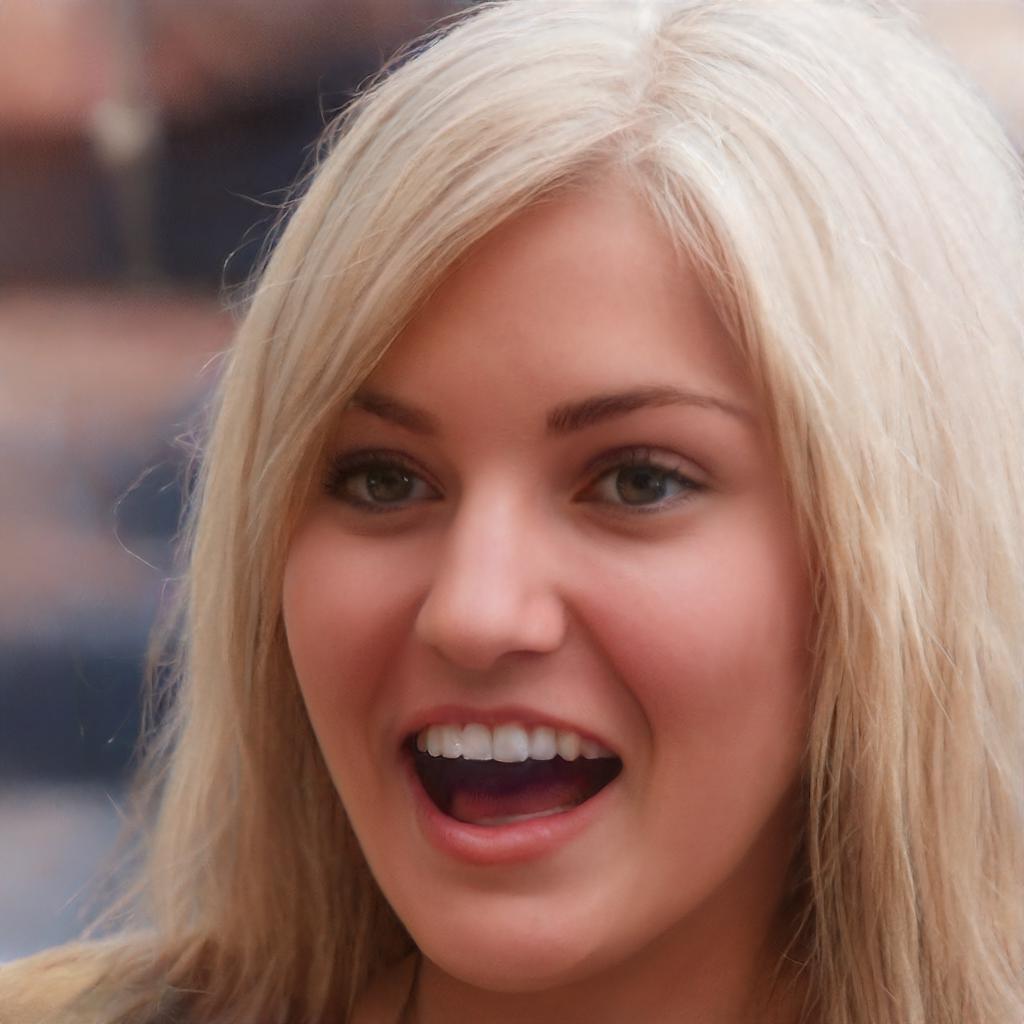} 
			\\
            Input & (0.8, 0) & (1.6, 0) & (0.8, 0.1)
		\end{tabular}
	}
	\caption{Identity loss ablation study. Under each column we specify $(\lambda_{\text{L2}}, \lambda_{\text{ID}})$. In the second and the third columns we did not use the identity loss. As can be seen, the identity of individual in the input image is not preserved.}
	\label{fig:ablation-id}
\end{figure}

\paragraph{ID Loss}
Here we show that the identity loss is significant for preserving the identity of the person in the input image. When using the default parameter setting of $\lambda_{\text{L2}} = 0.8$ with $\lambda_{\text{ID}} = 0$ (i.e., no identity loss), we observe that the mapper fails to preserve the identity, and introduces large changes. Therefore, we also experiment with $\lambda_{\text{L2}} = 1.6$, however, this still does not preserve the original identity well enough. The results are shown in Figure~\ref{fig:ablation-id}.

\section{Additional Results}

\begin{figure*}[tb]
	\setlength{\tabcolsep}{1pt}
	\centering
	{\footnotesize
		\begin{tabular}{c c c c c c c}
			\includegraphics[width=0.14\textwidth]{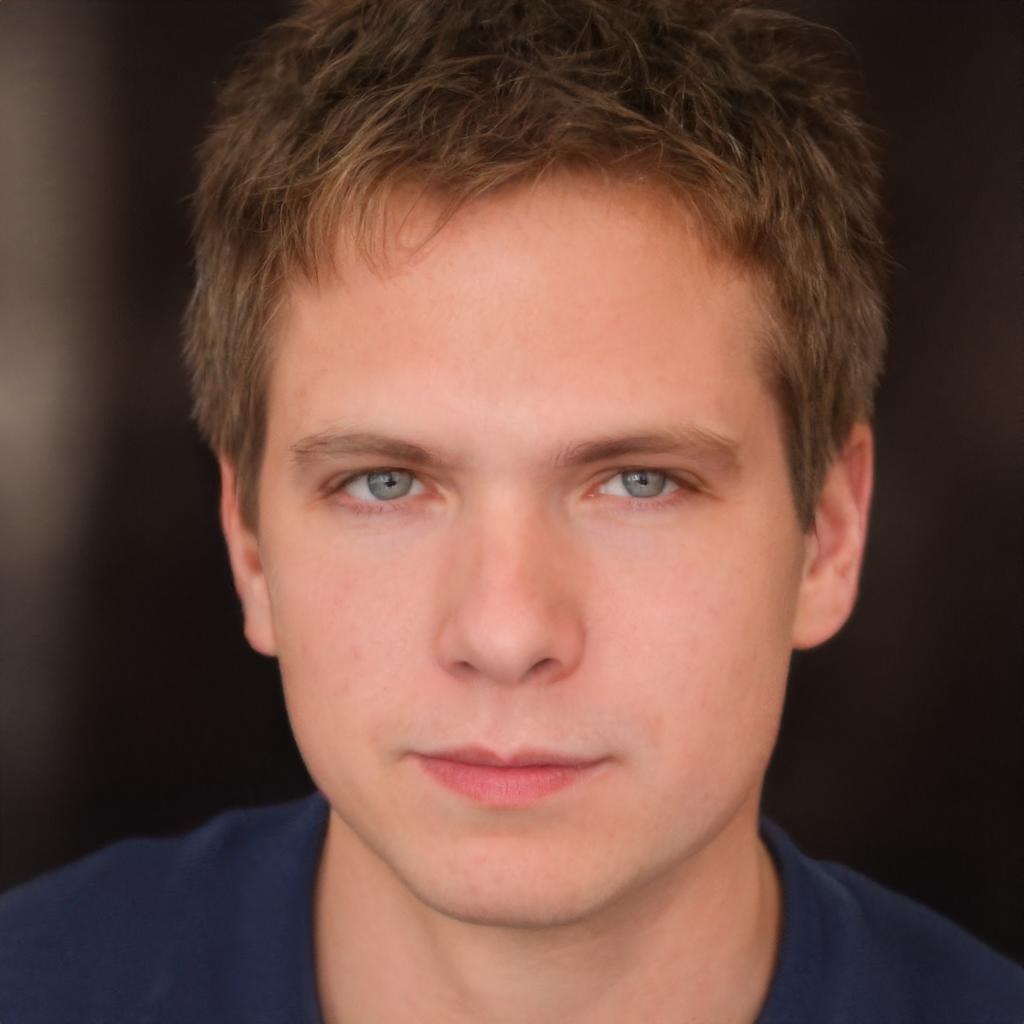} &
 			\includegraphics[width=0.14\textwidth]{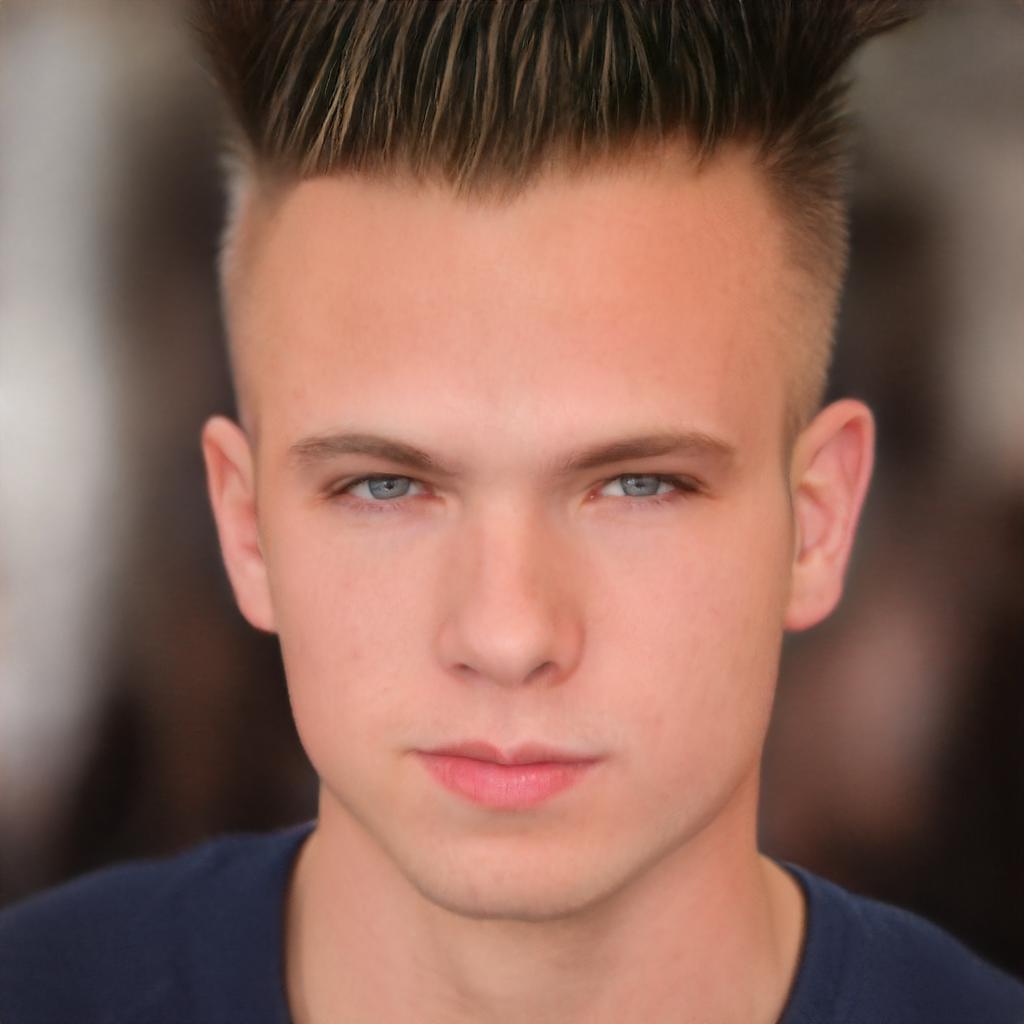} &
 			\includegraphics[width=0.14\textwidth]{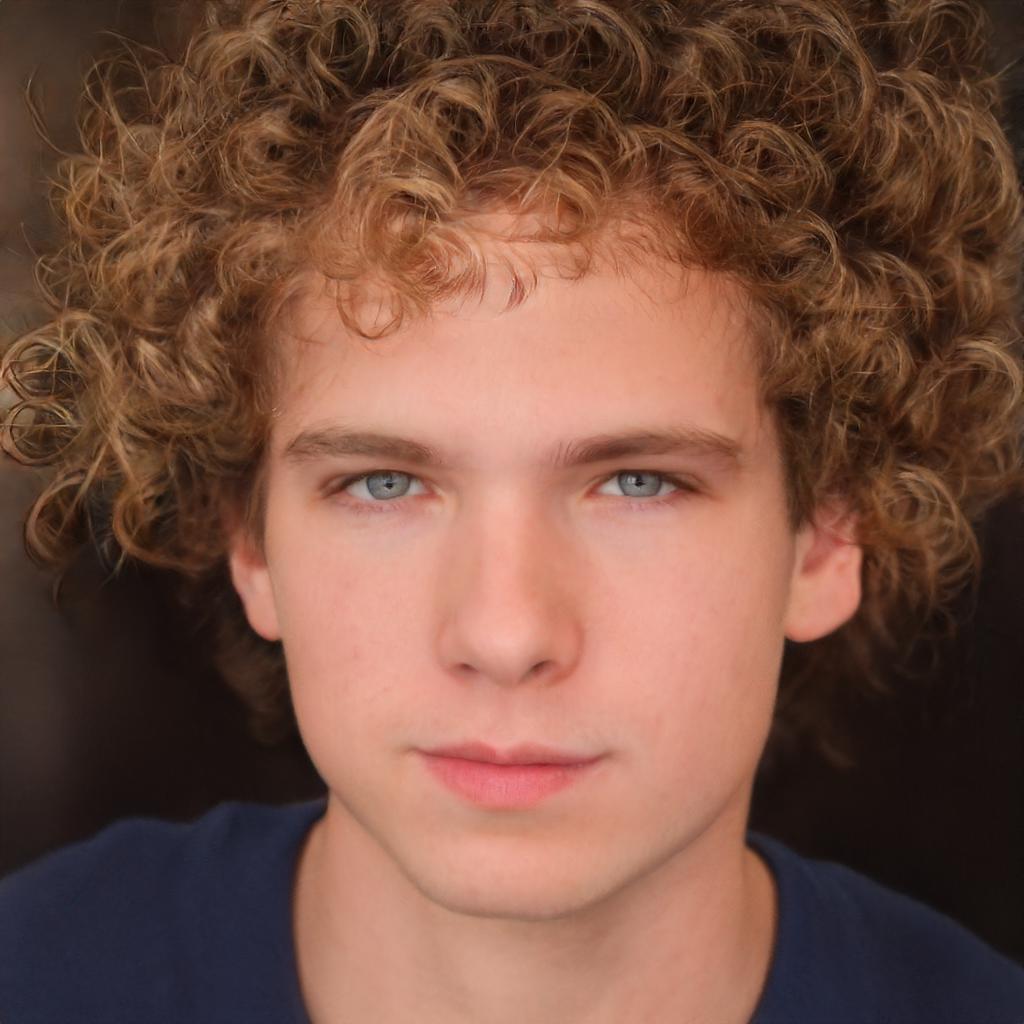} &
 			\includegraphics[width=0.14\textwidth]{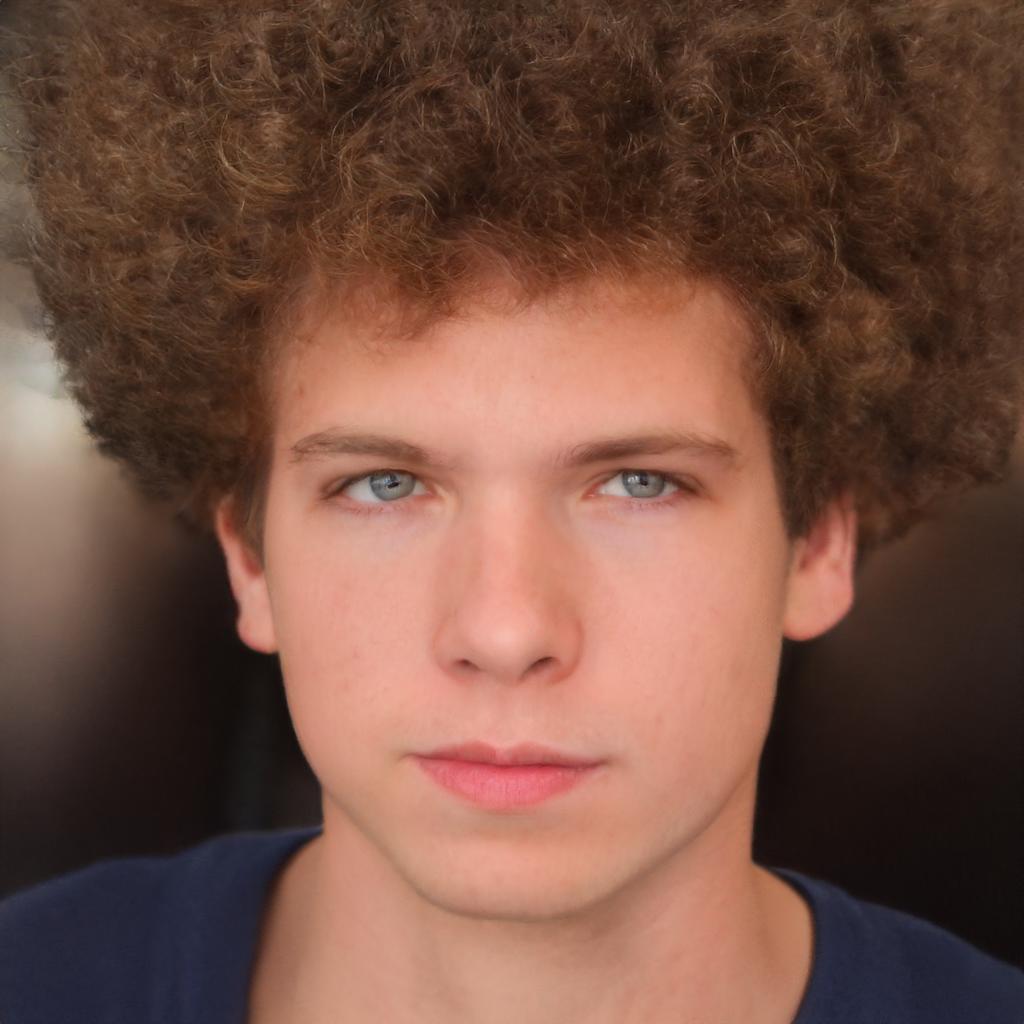} &
 			\includegraphics[width=0.14\textwidth]{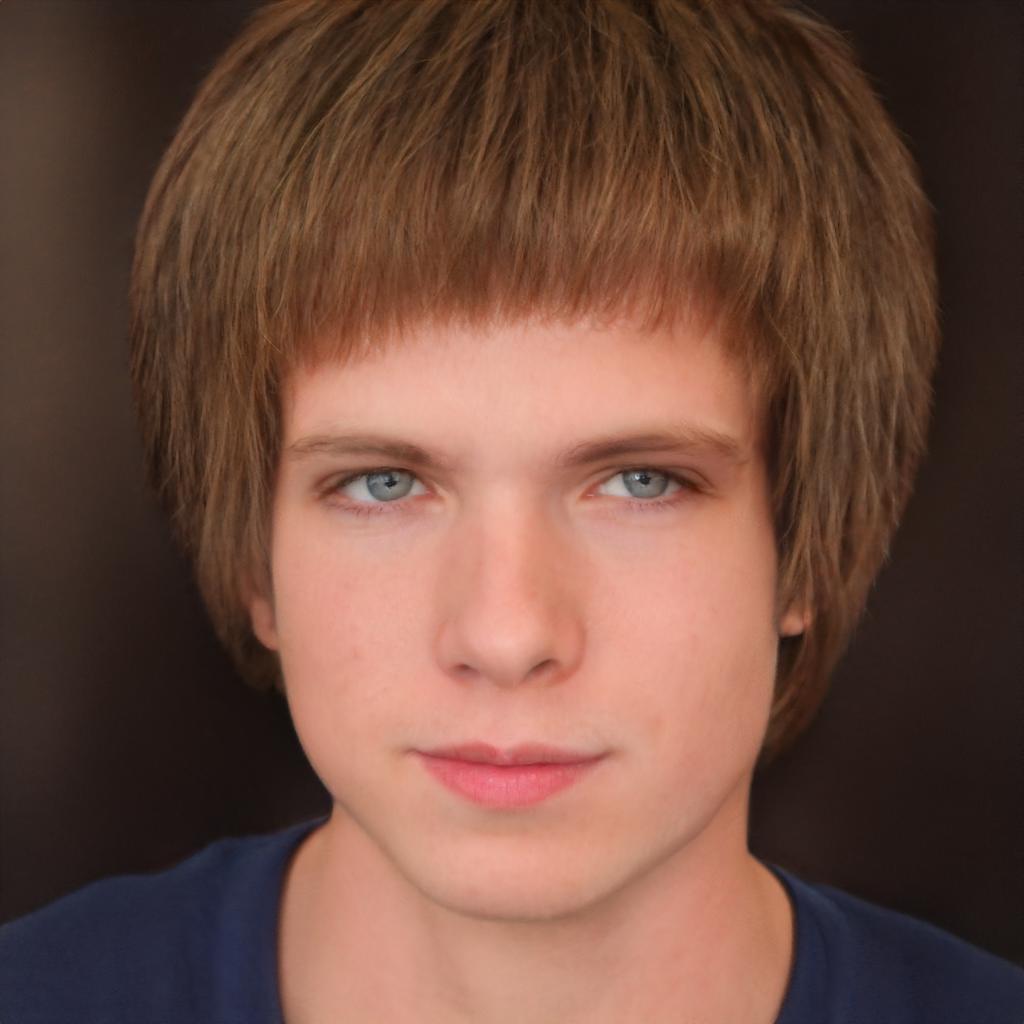} &
 			\includegraphics[width=0.14\textwidth]{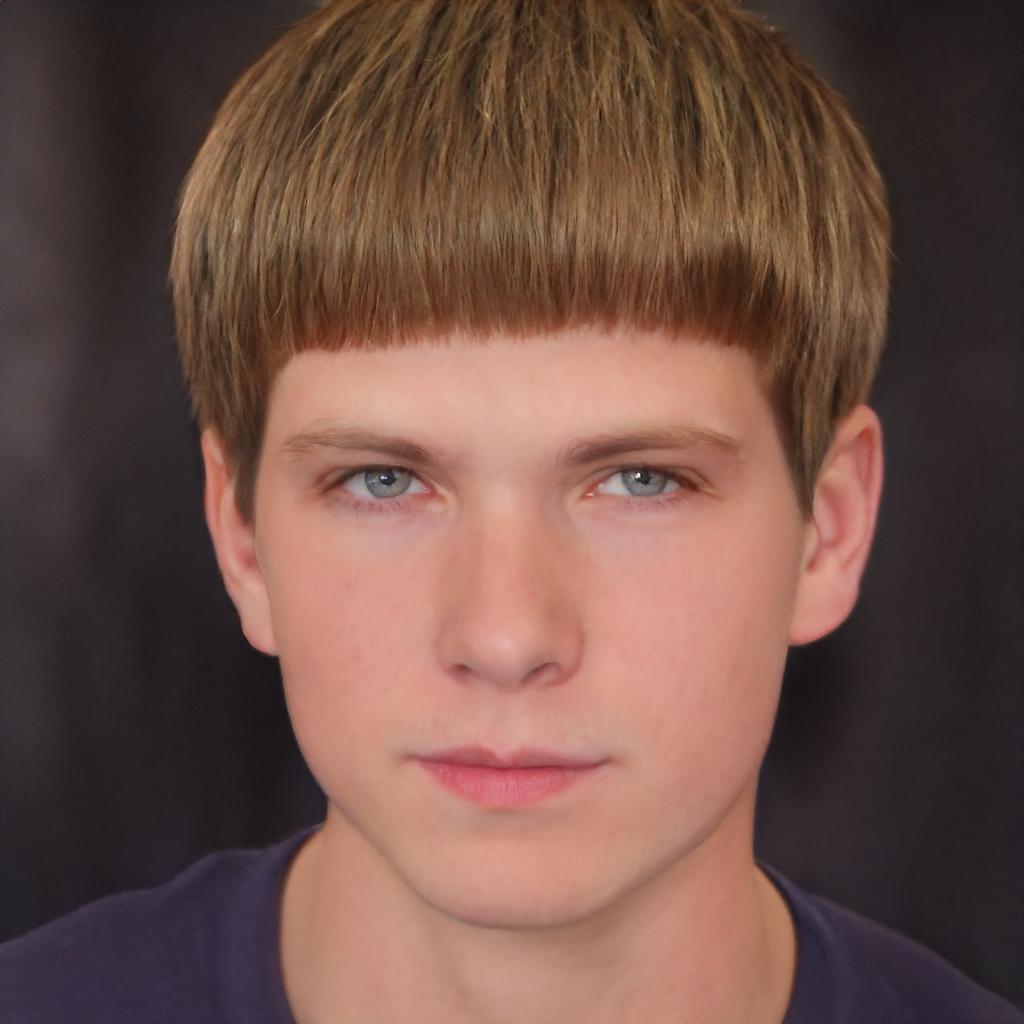} &
            \includegraphics[width=0.14\textwidth]{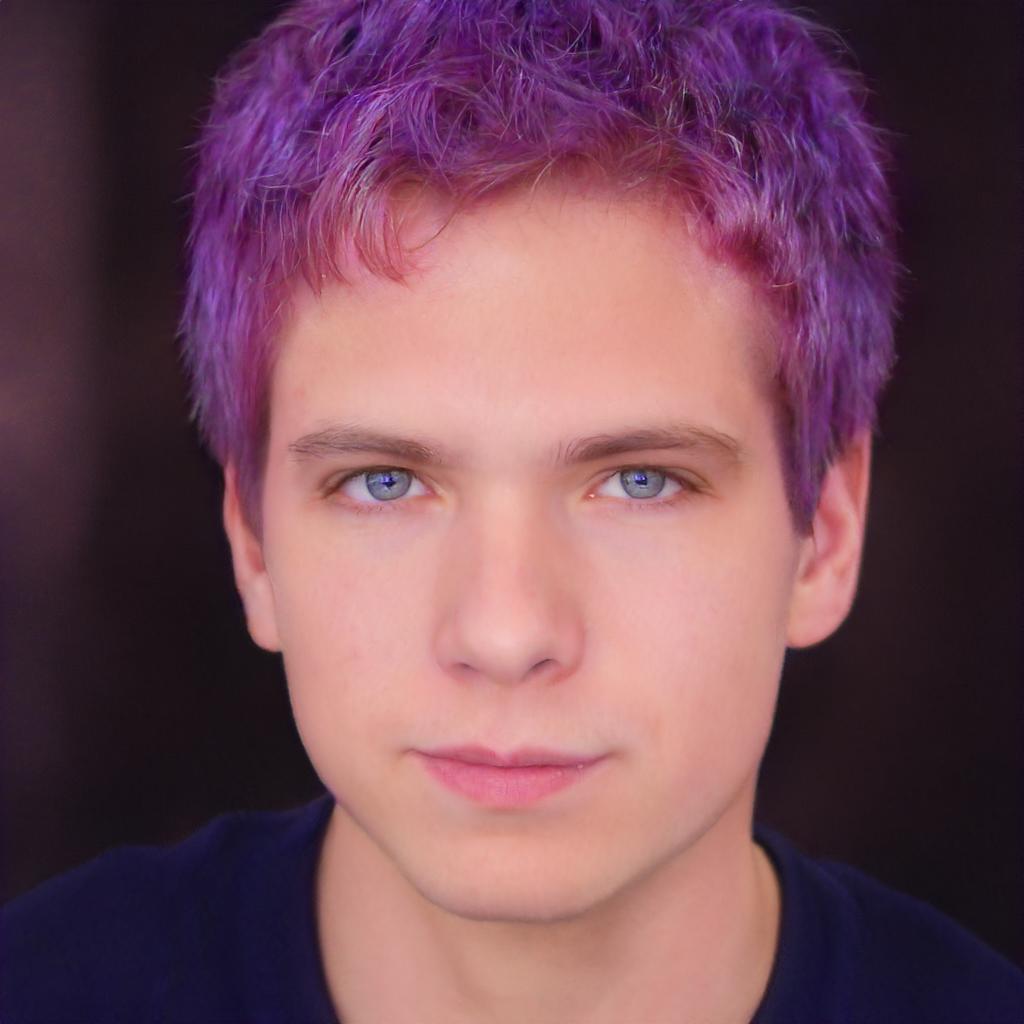} \\
            
            \includegraphics[width=0.14\textwidth]{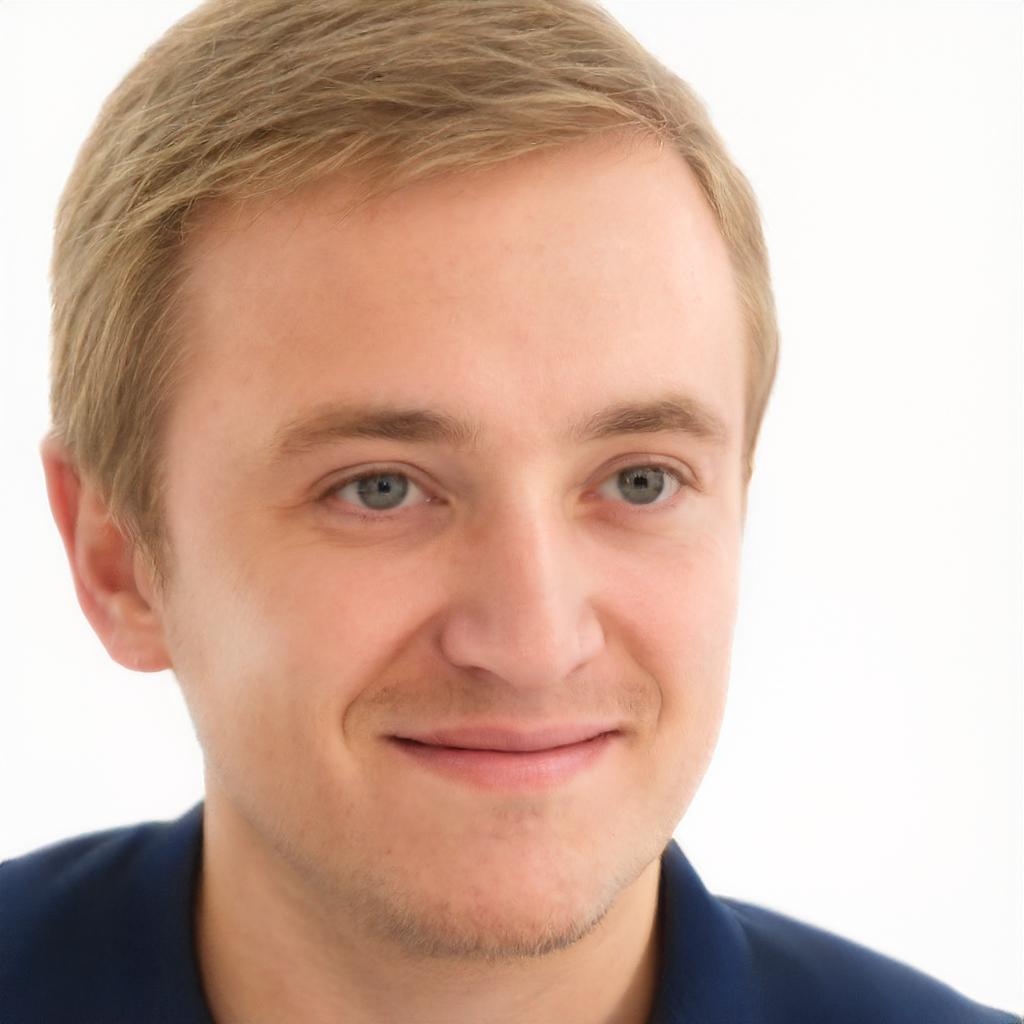} &
			\includegraphics[width=0.14\textwidth]{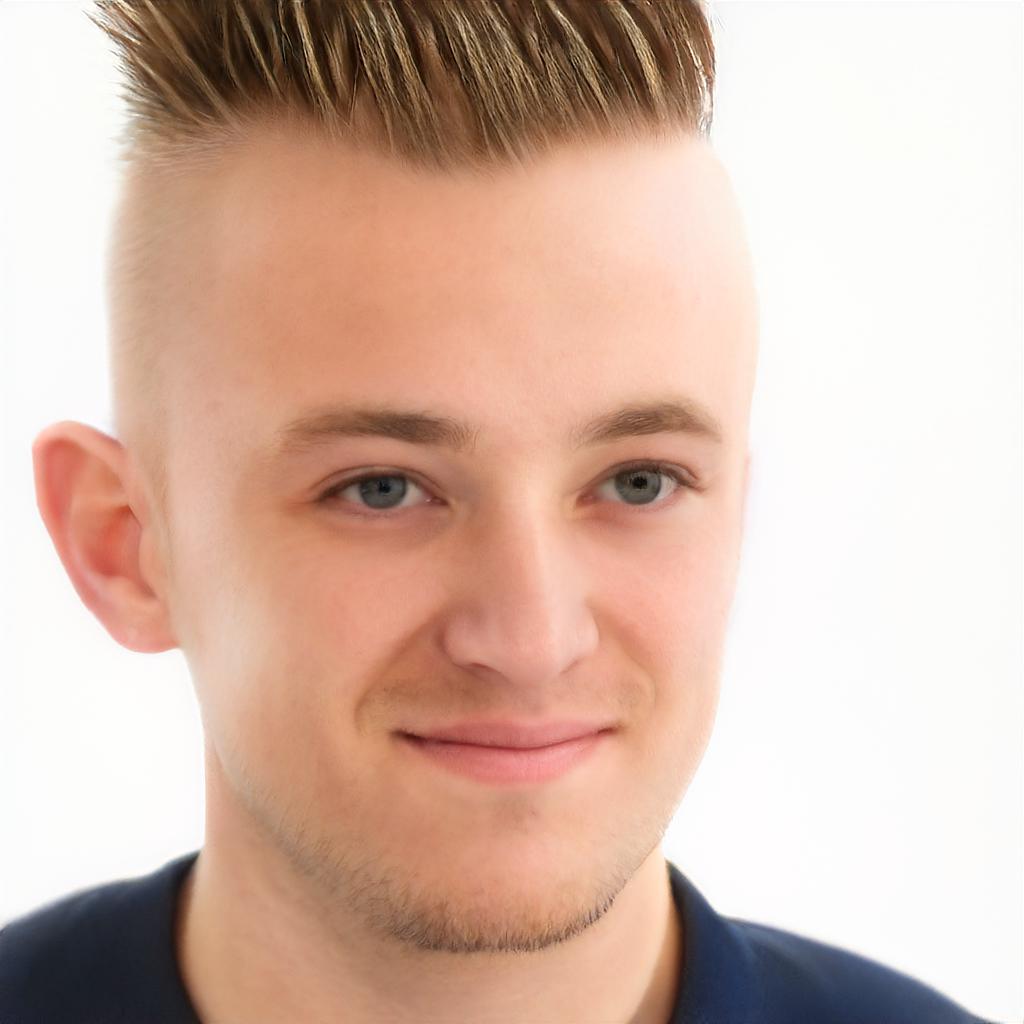} &
			\includegraphics[width=0.14\textwidth]{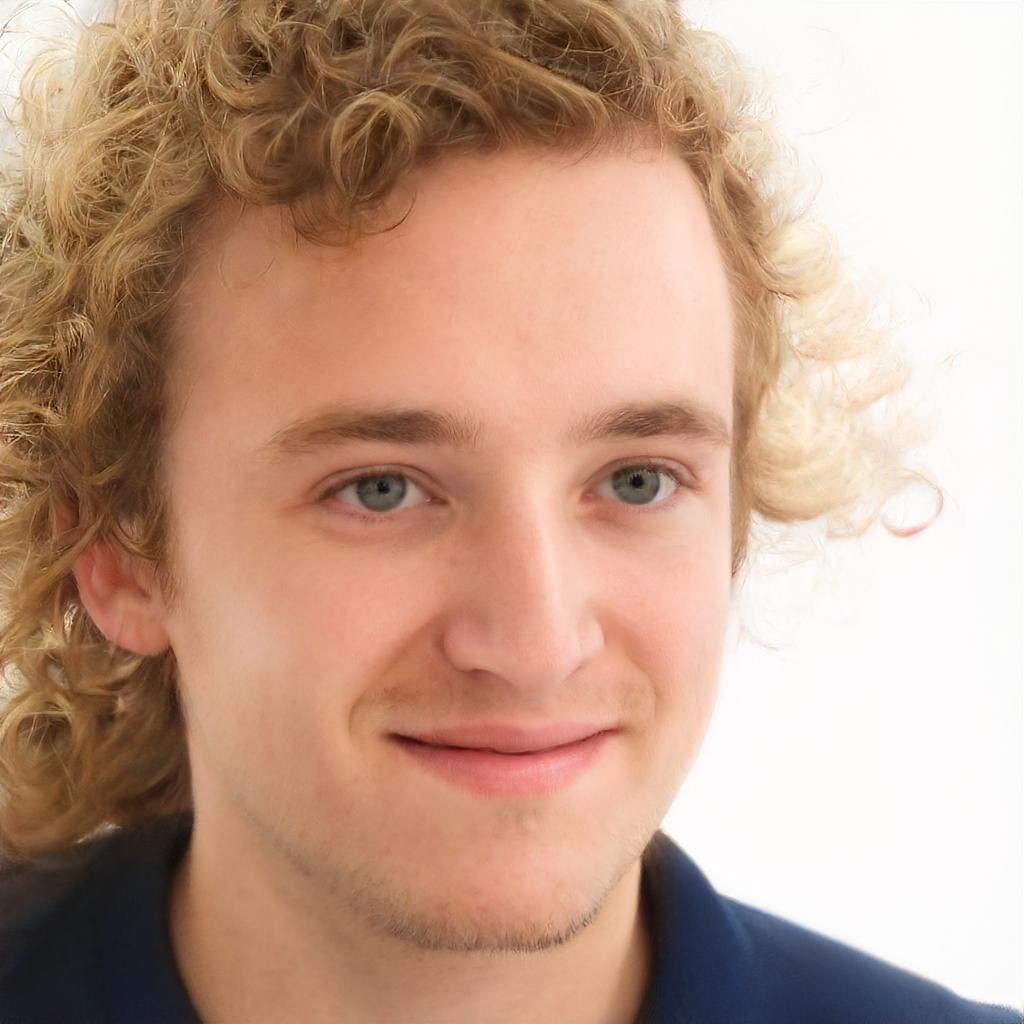} &
			\includegraphics[width=0.14\textwidth]{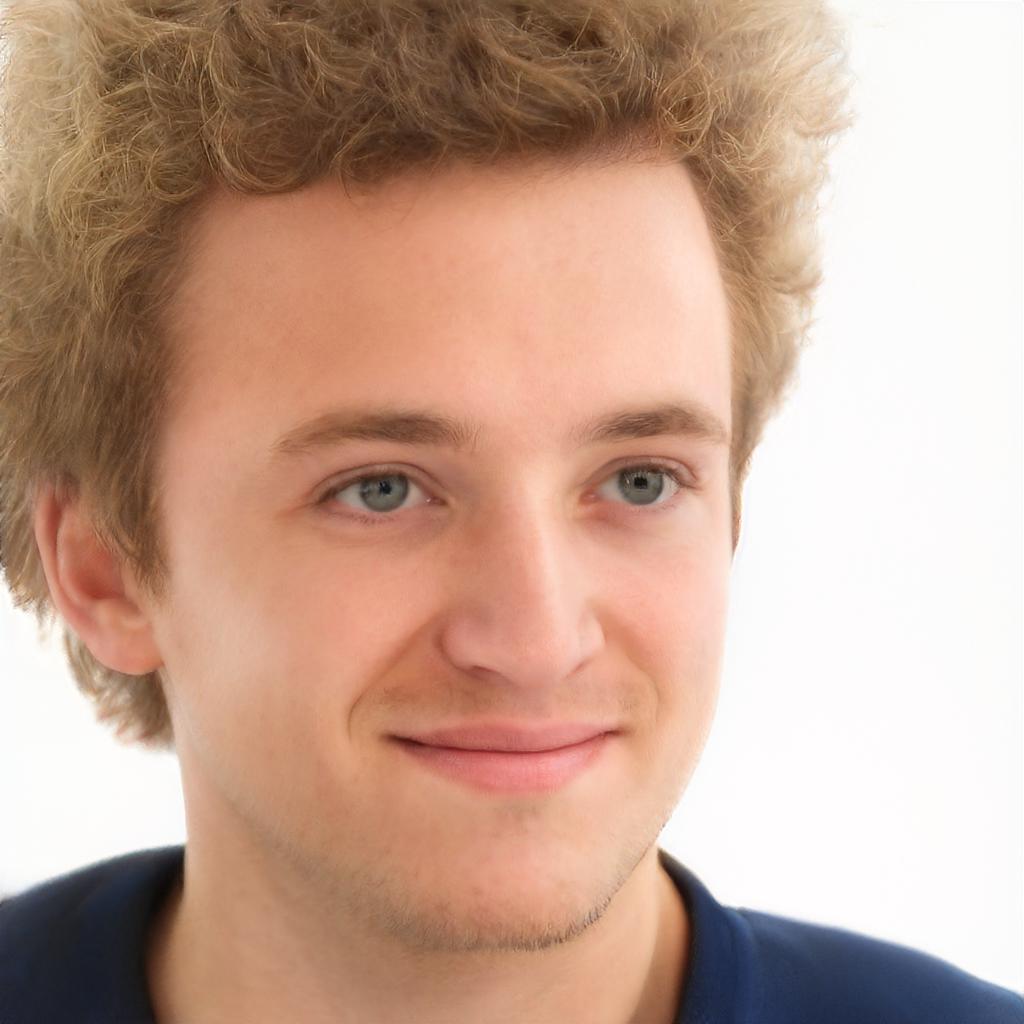} &
			\includegraphics[width=0.14\textwidth]{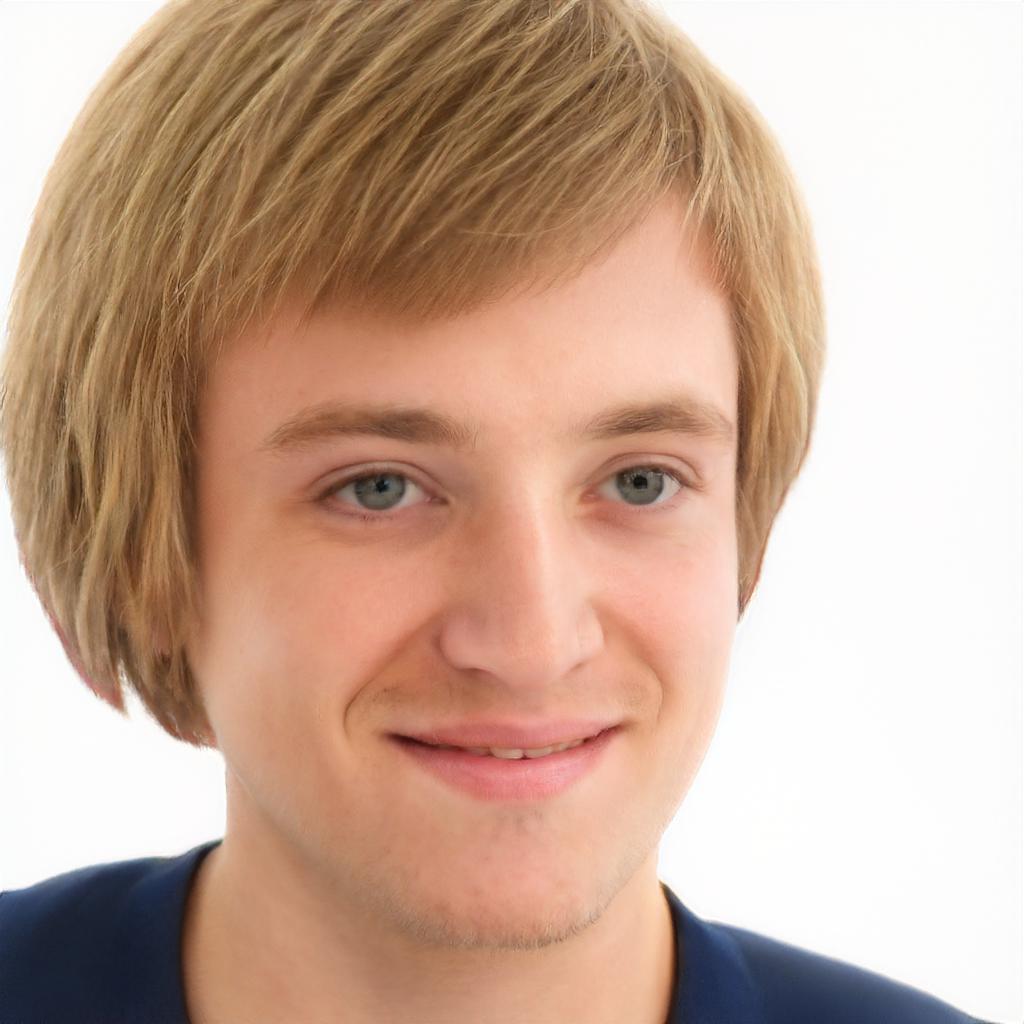} &
			\includegraphics[width=0.14\textwidth]{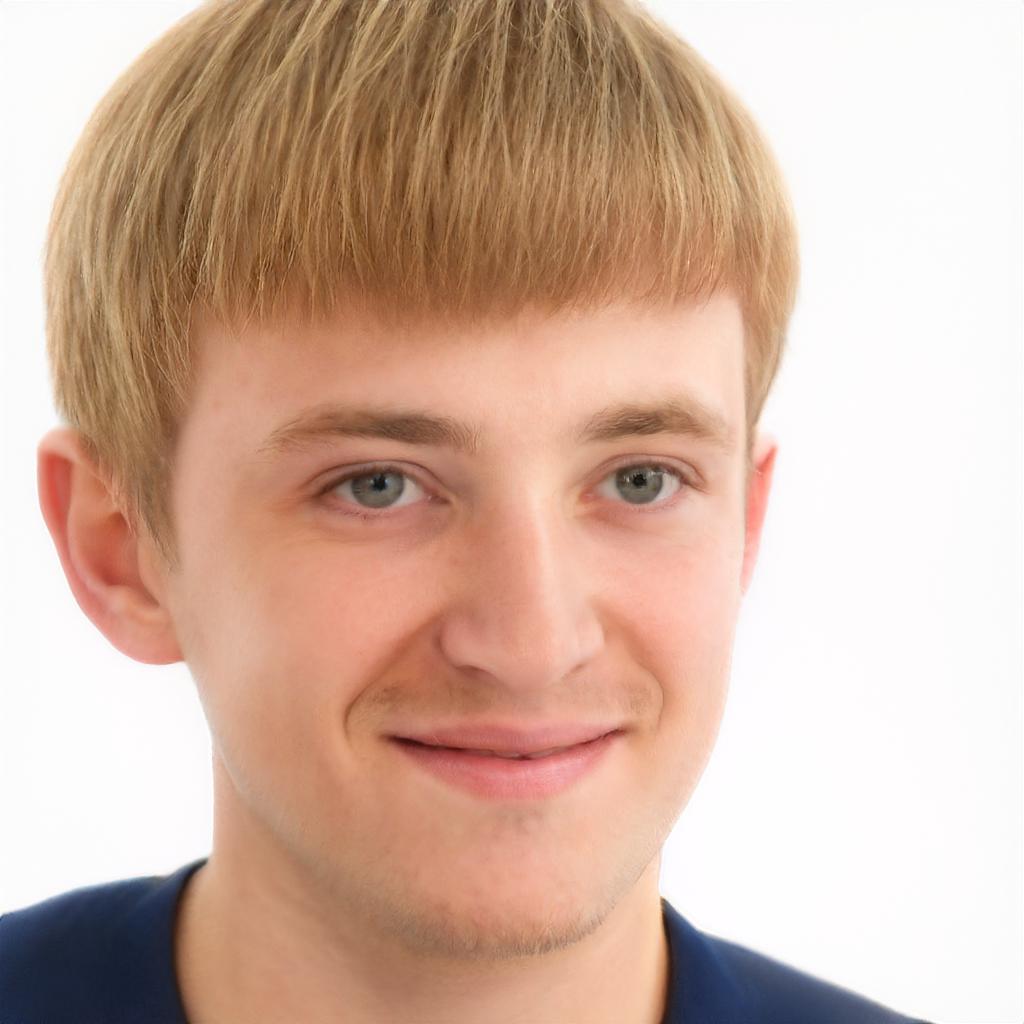} &
            \includegraphics[width=0.14\textwidth]{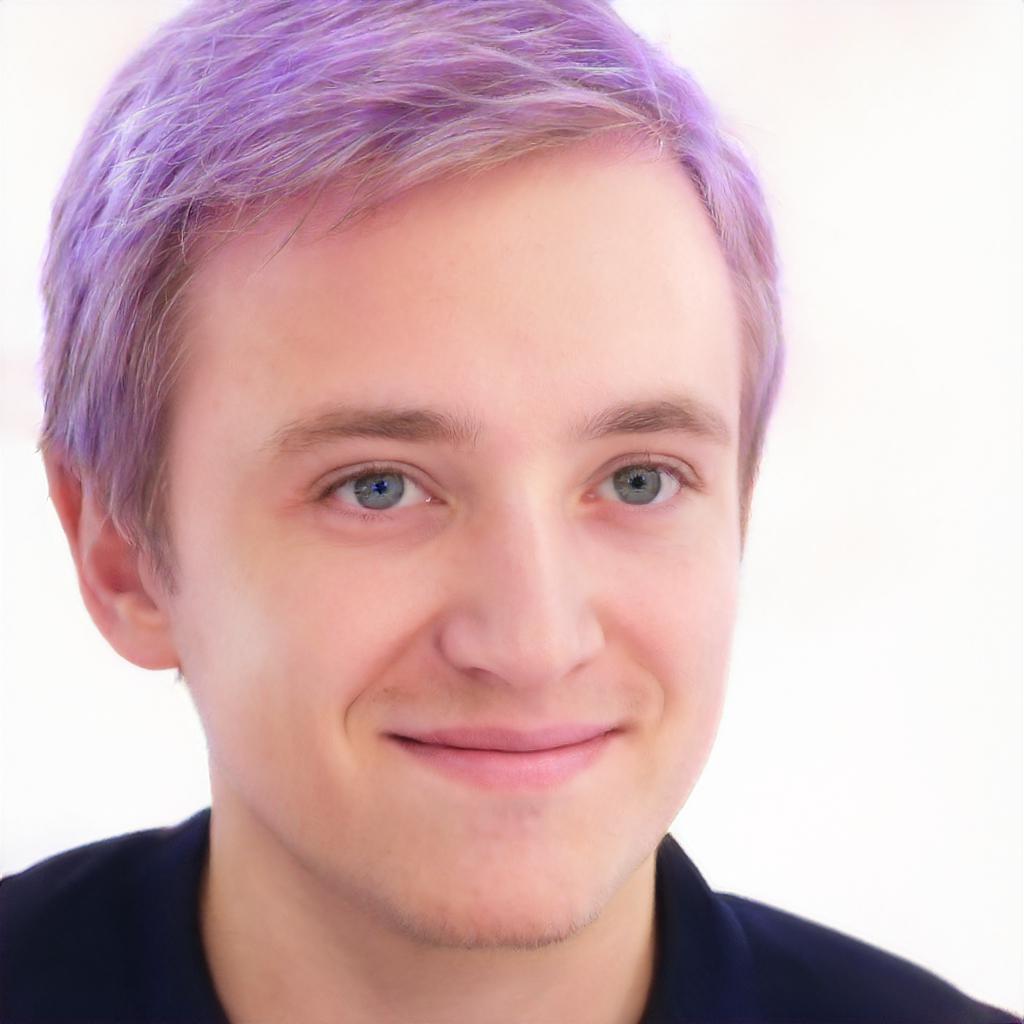} \\
            
			\includegraphics[width=0.14\textwidth]{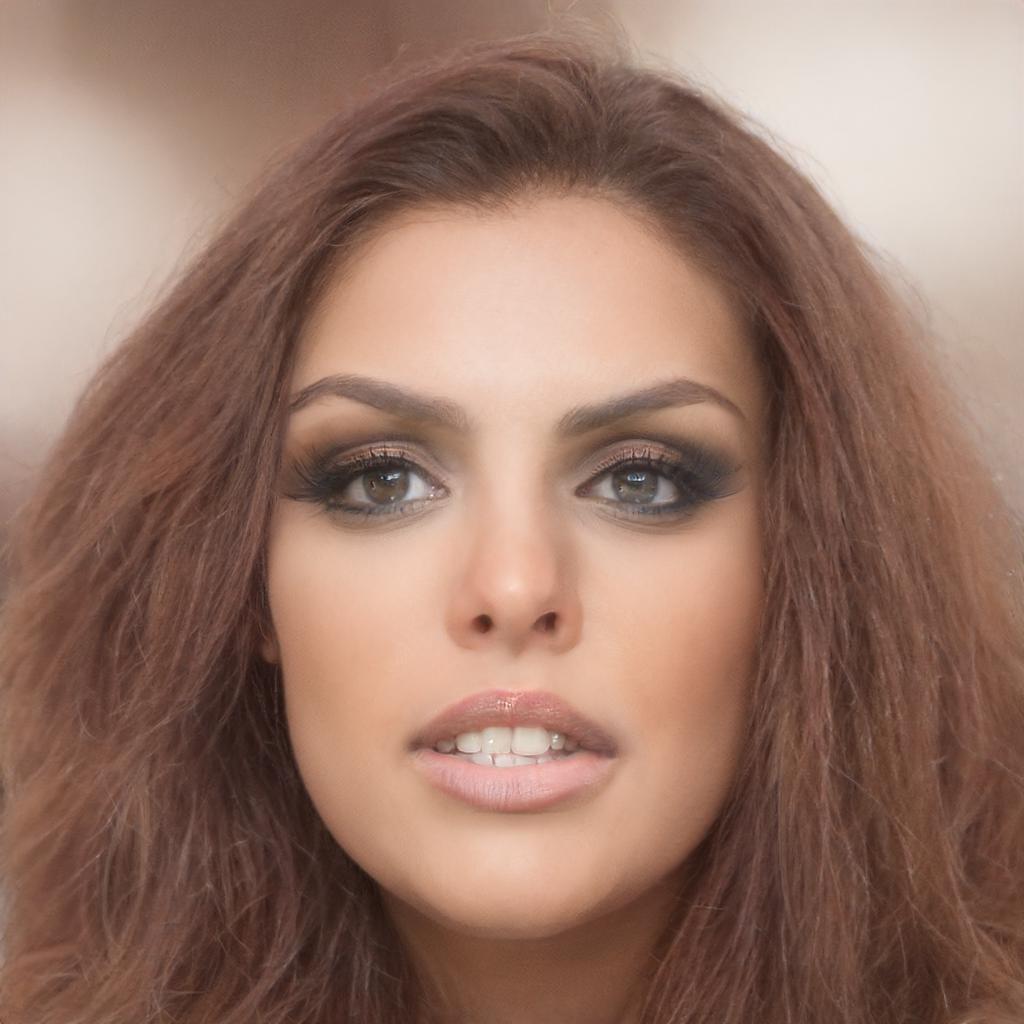} &
			\includegraphics[width=0.14\textwidth]{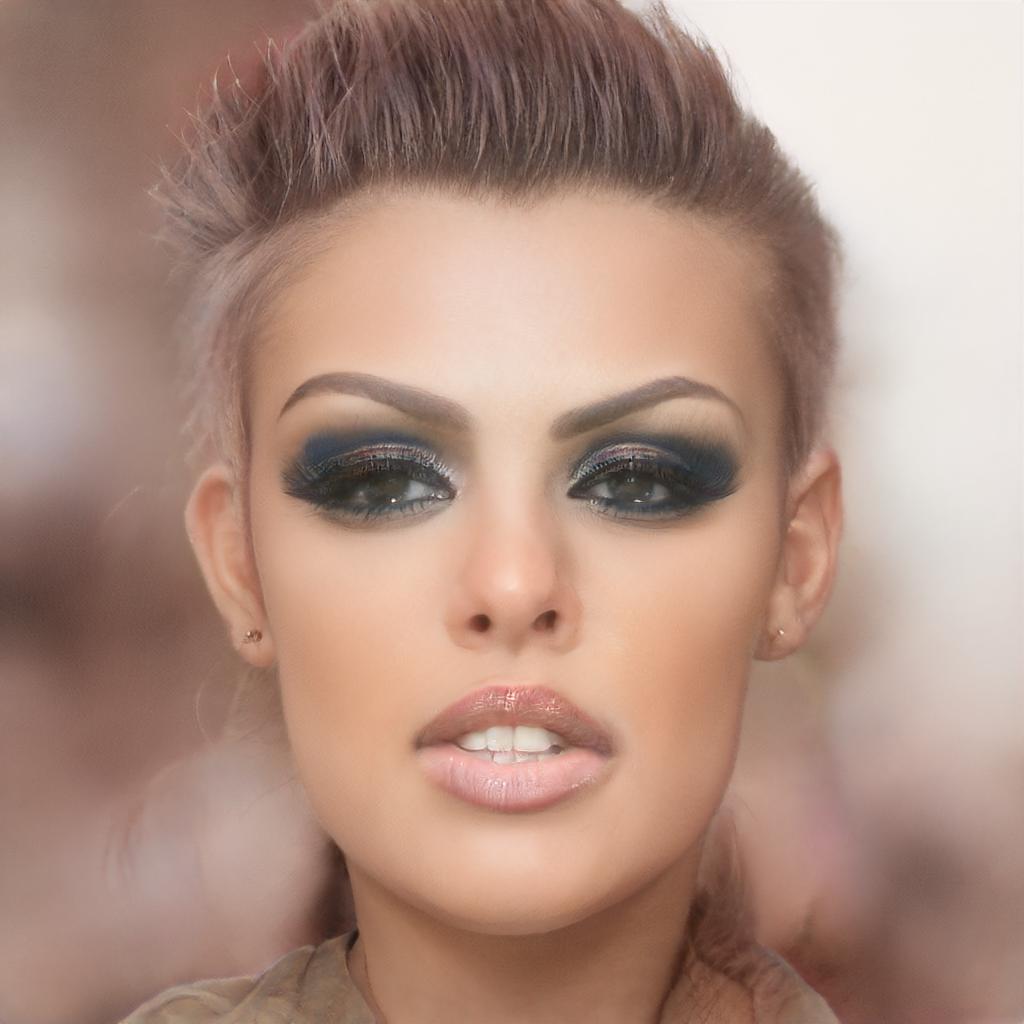} &
			\includegraphics[width=0.14\textwidth]{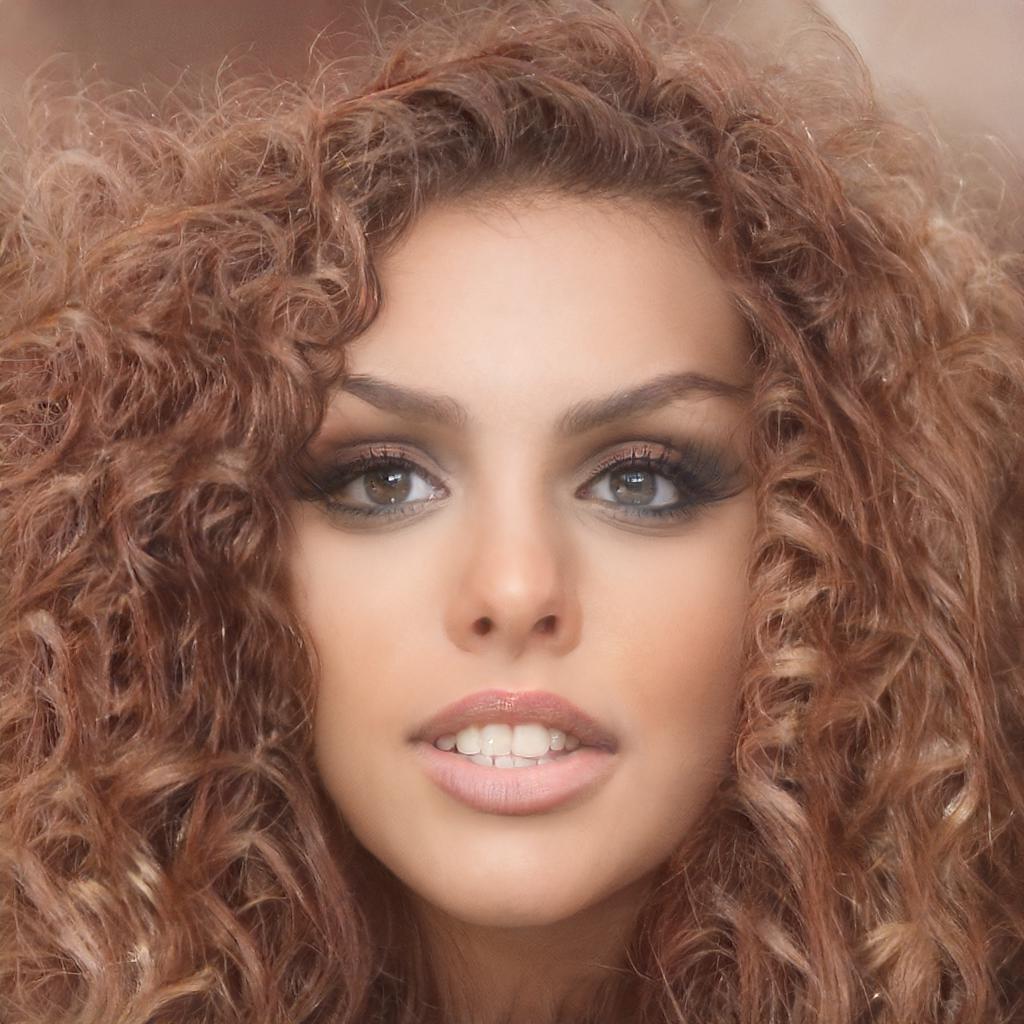} &
			\includegraphics[width=0.14\textwidth]{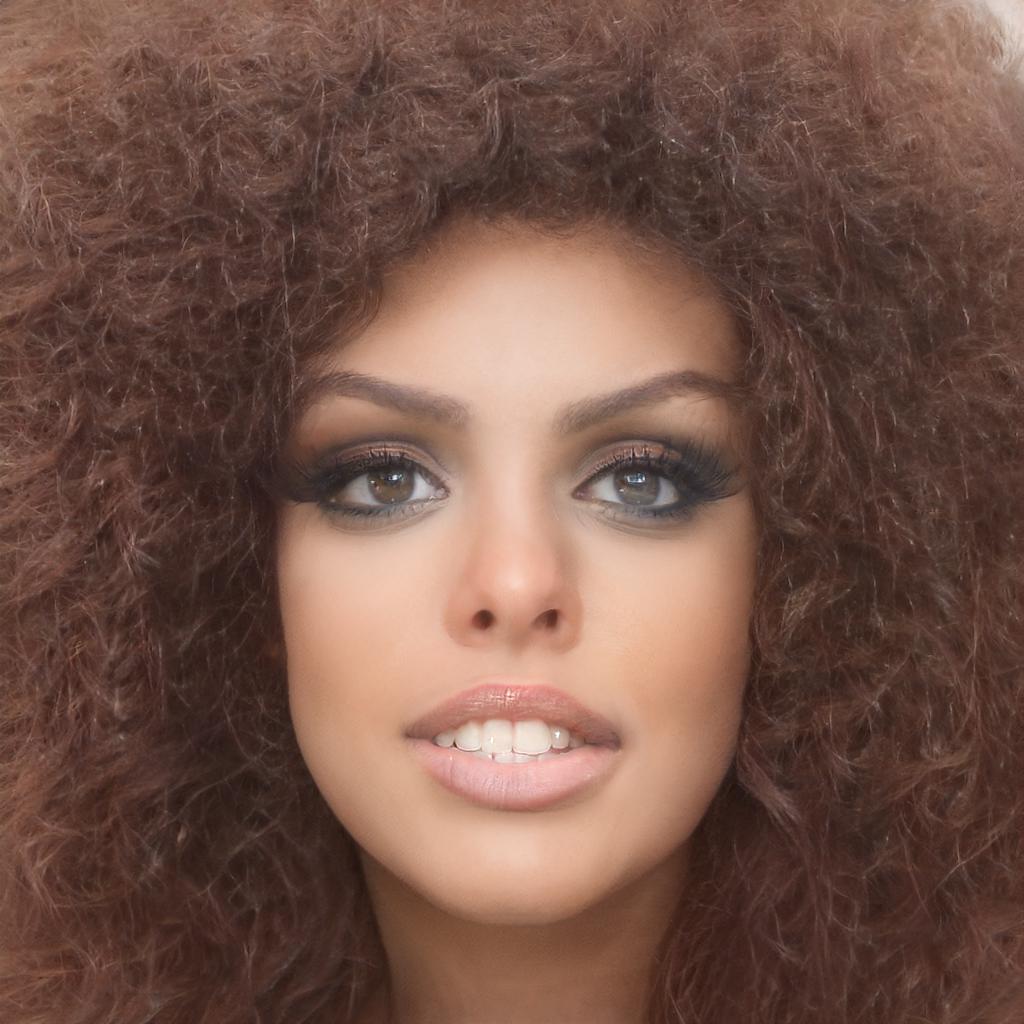} &
			\includegraphics[width=0.14\textwidth]{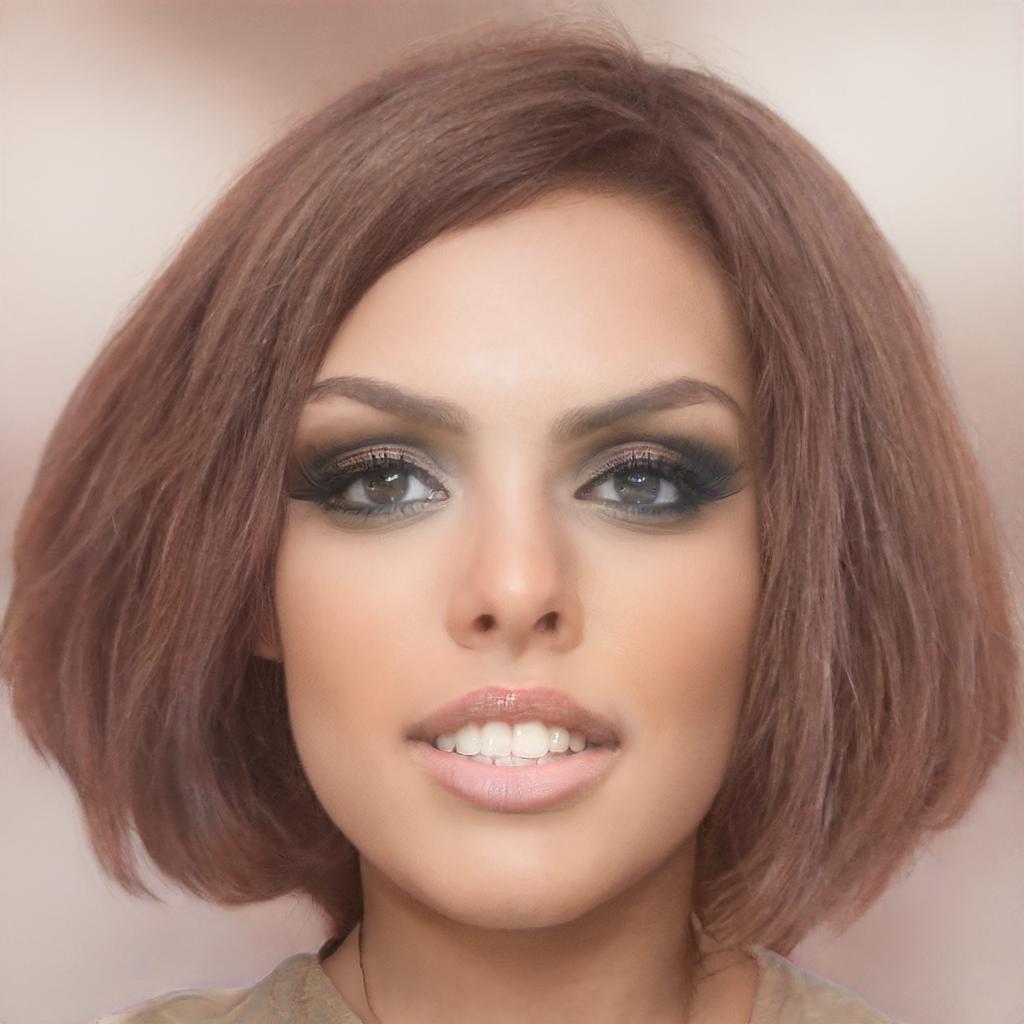} &
			\includegraphics[width=0.14\textwidth]{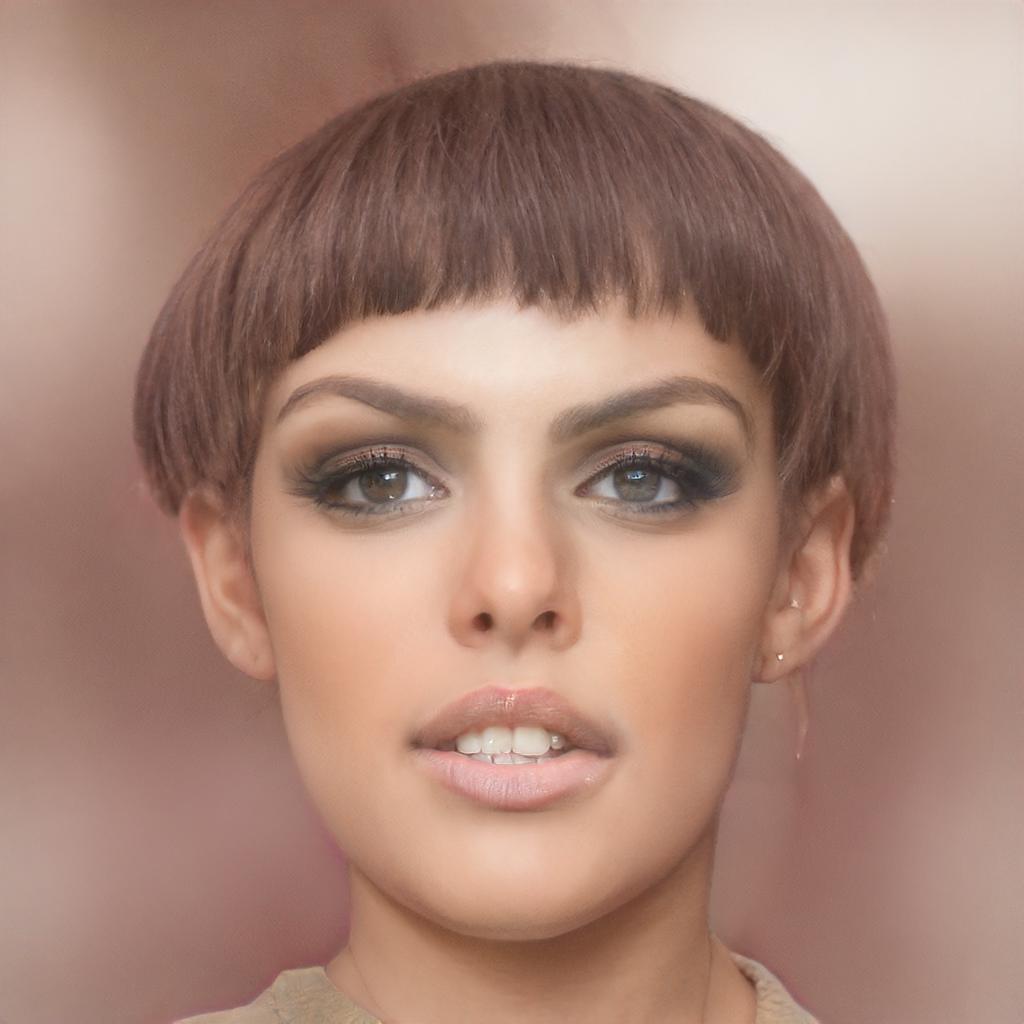} &
            \includegraphics[width=0.14\textwidth]{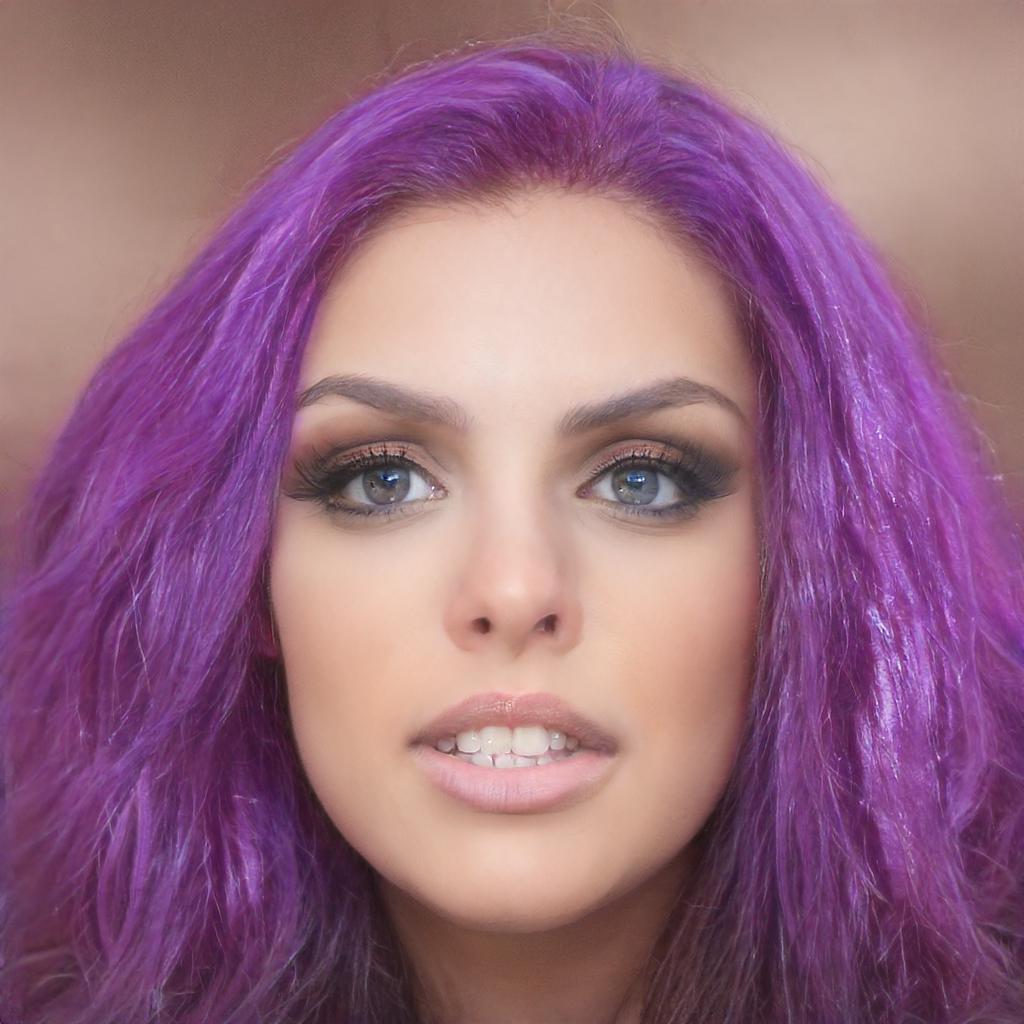} \\
            
			\includegraphics[width=0.14\textwidth]{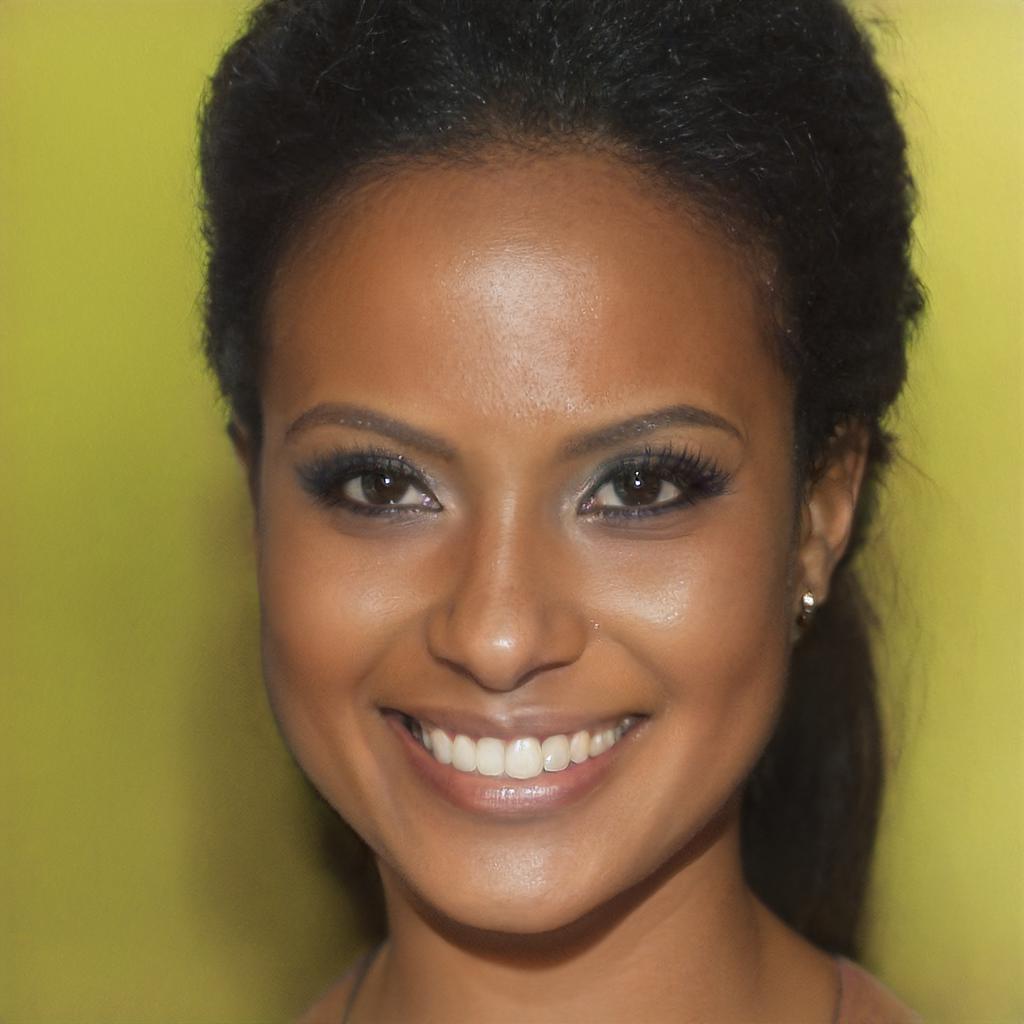} &
			\includegraphics[width=0.14\textwidth]{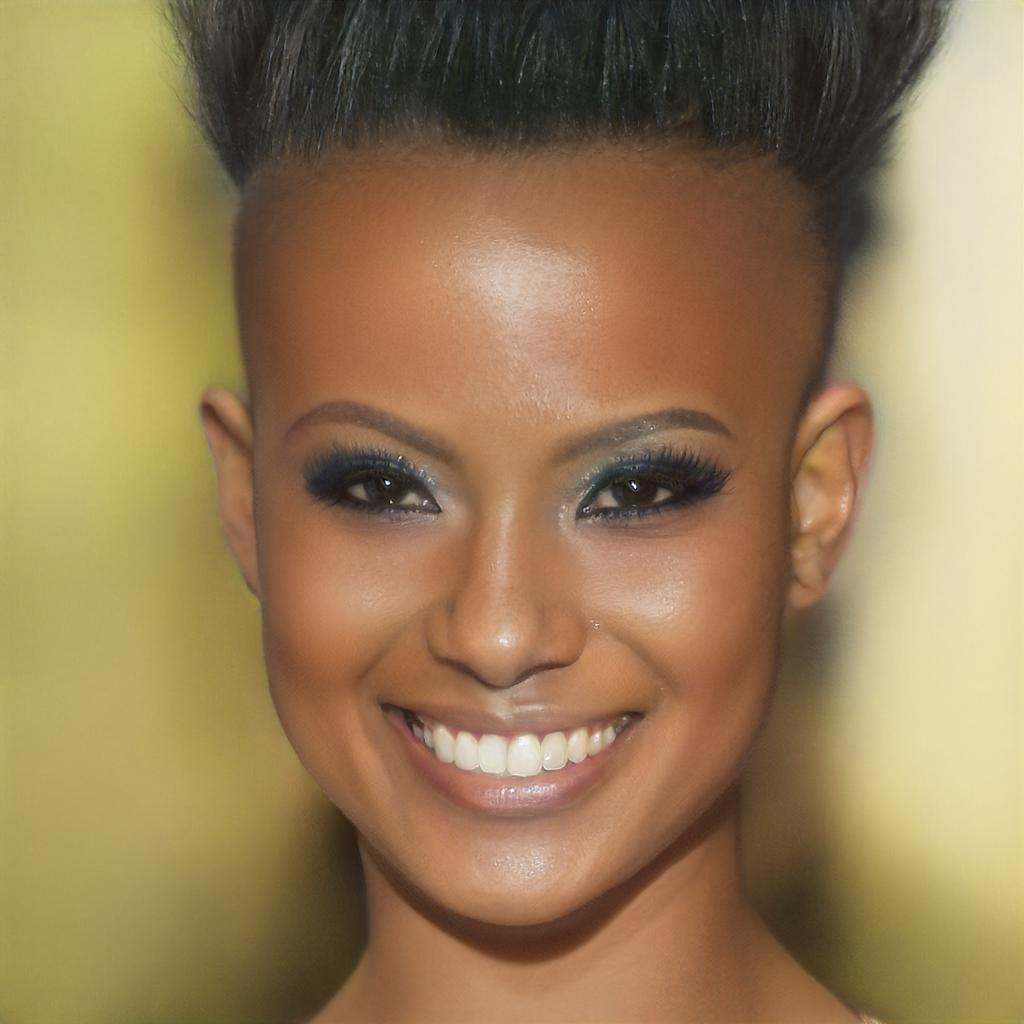} &
			\includegraphics[width=0.14\textwidth]{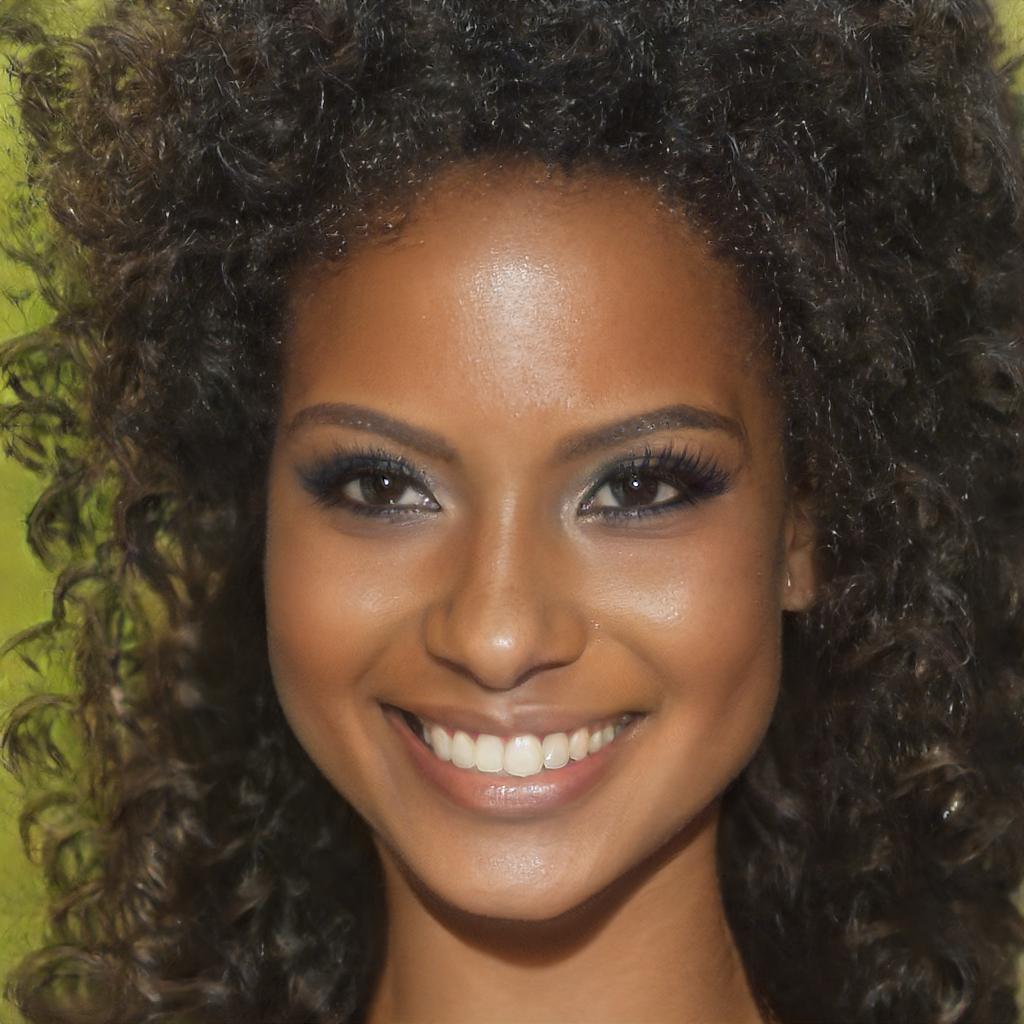} &
			\includegraphics[width=0.14\textwidth]{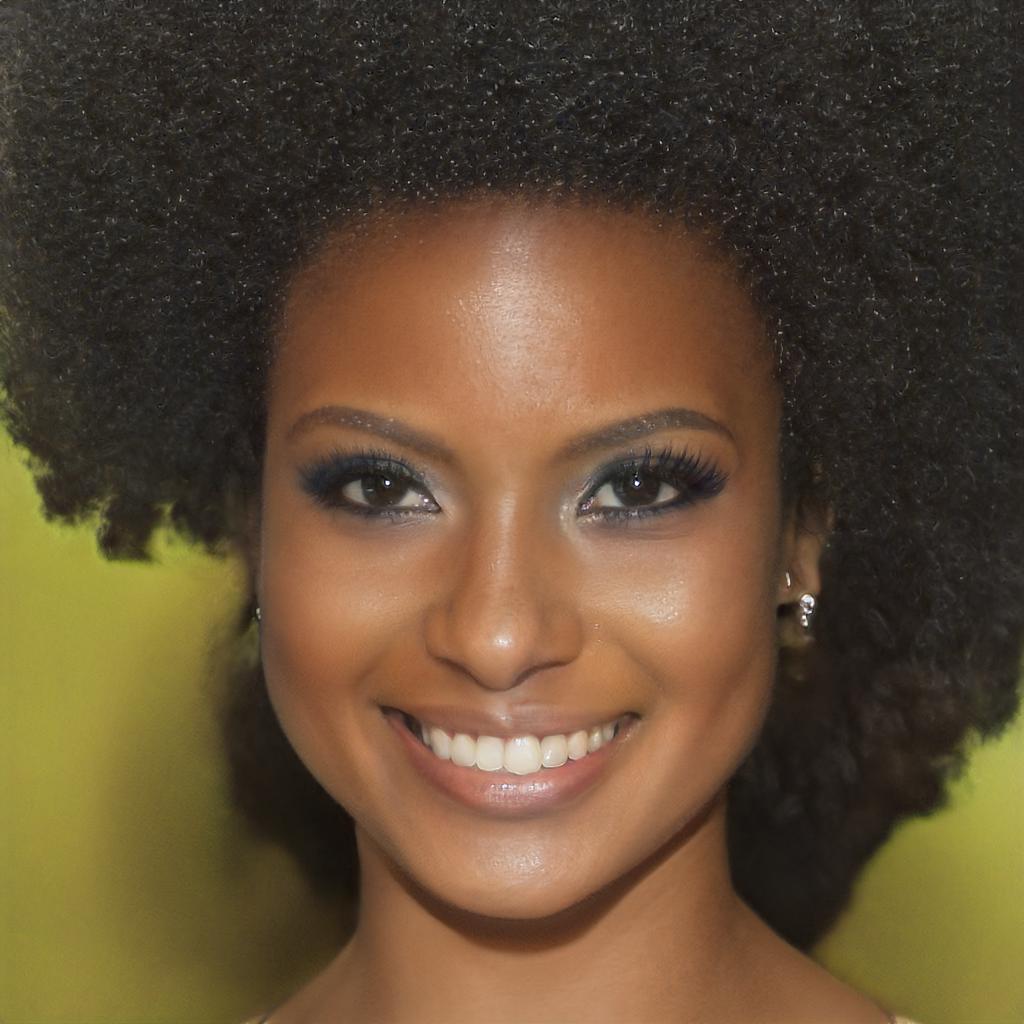} &
			\includegraphics[width=0.14\textwidth]{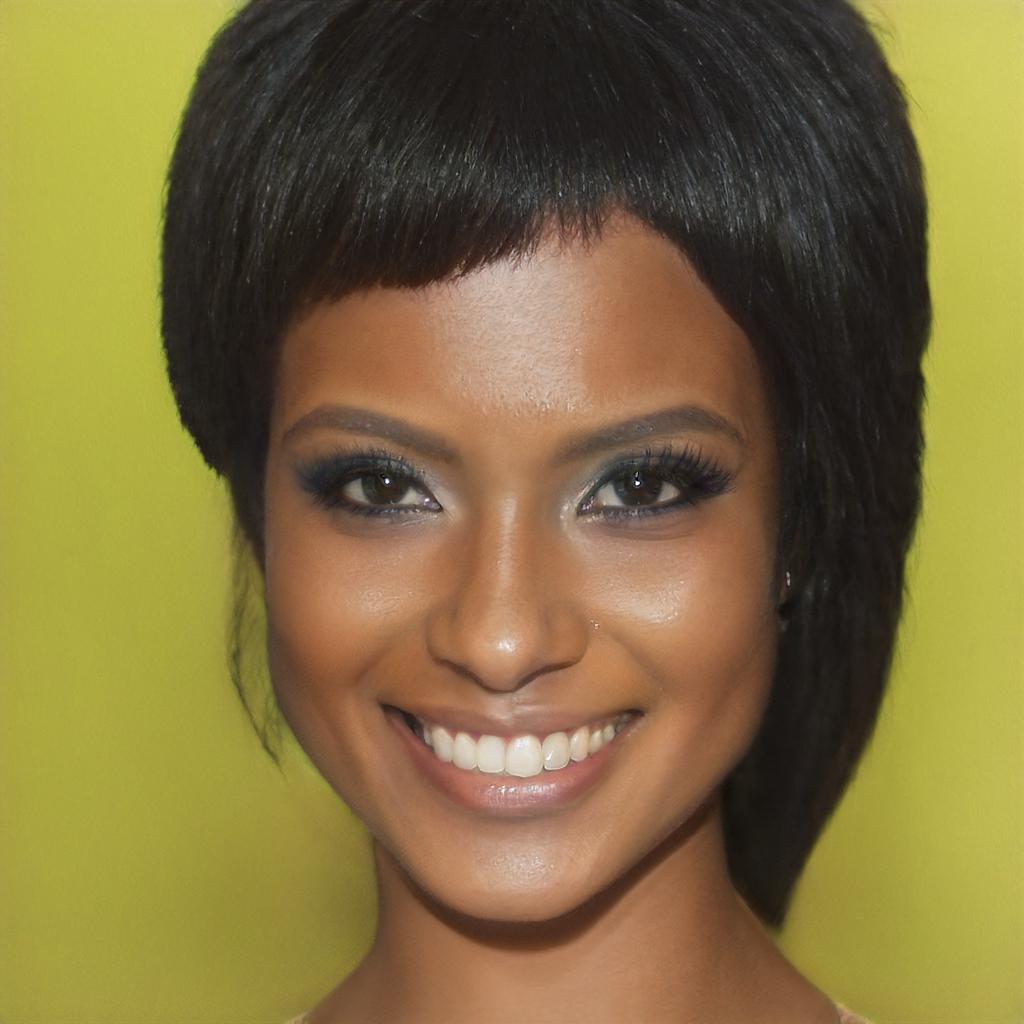} &
			\includegraphics[width=0.14\textwidth]{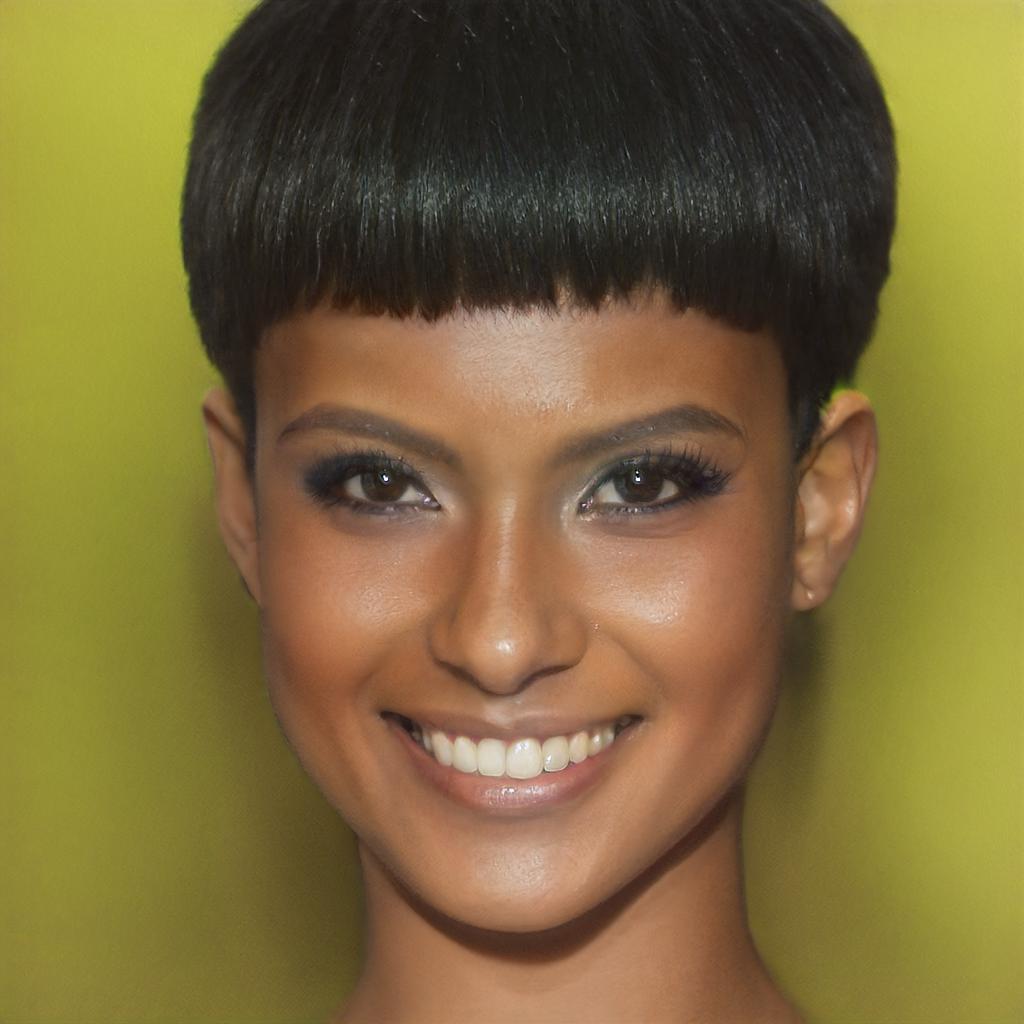} &
            \includegraphics[width=0.14\textwidth]{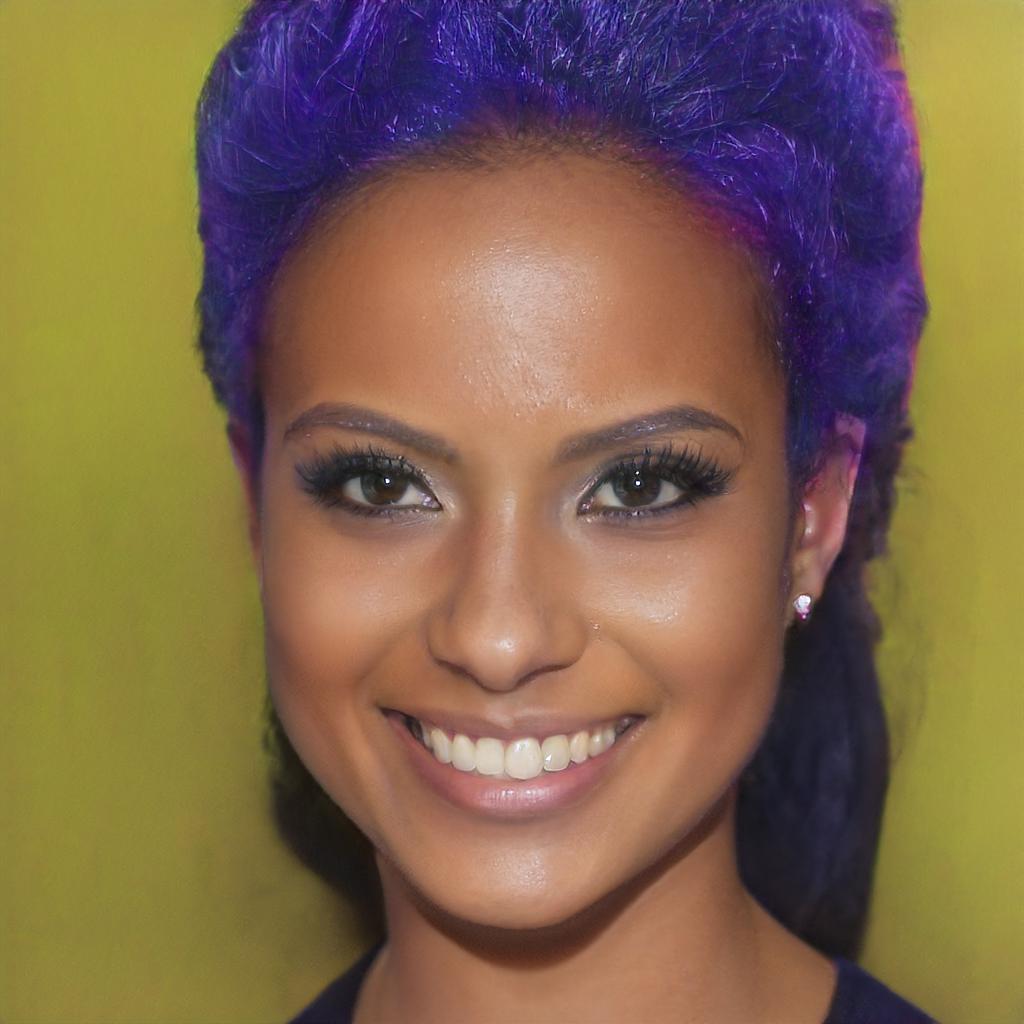} \\
            
            \includegraphics[width=0.14\textwidth]{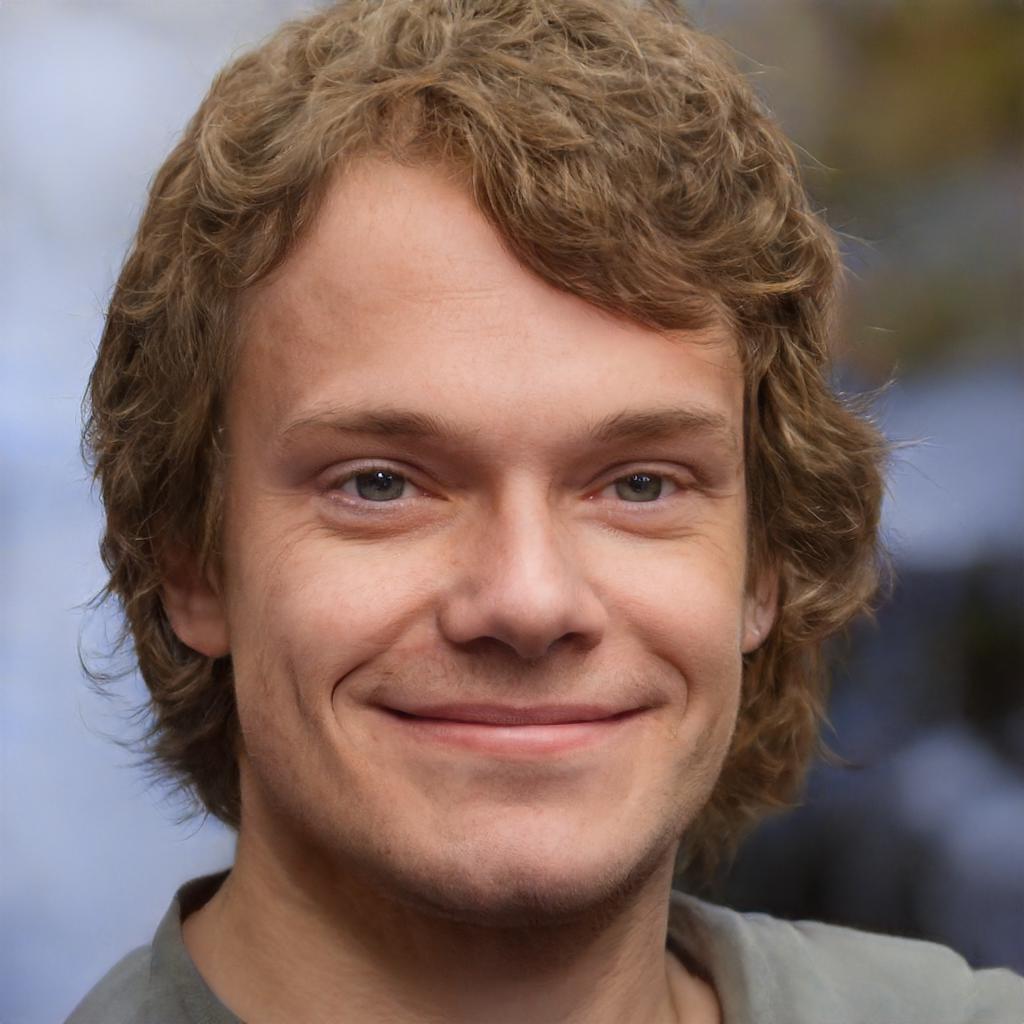} &
 			\includegraphics[width=0.14\textwidth]{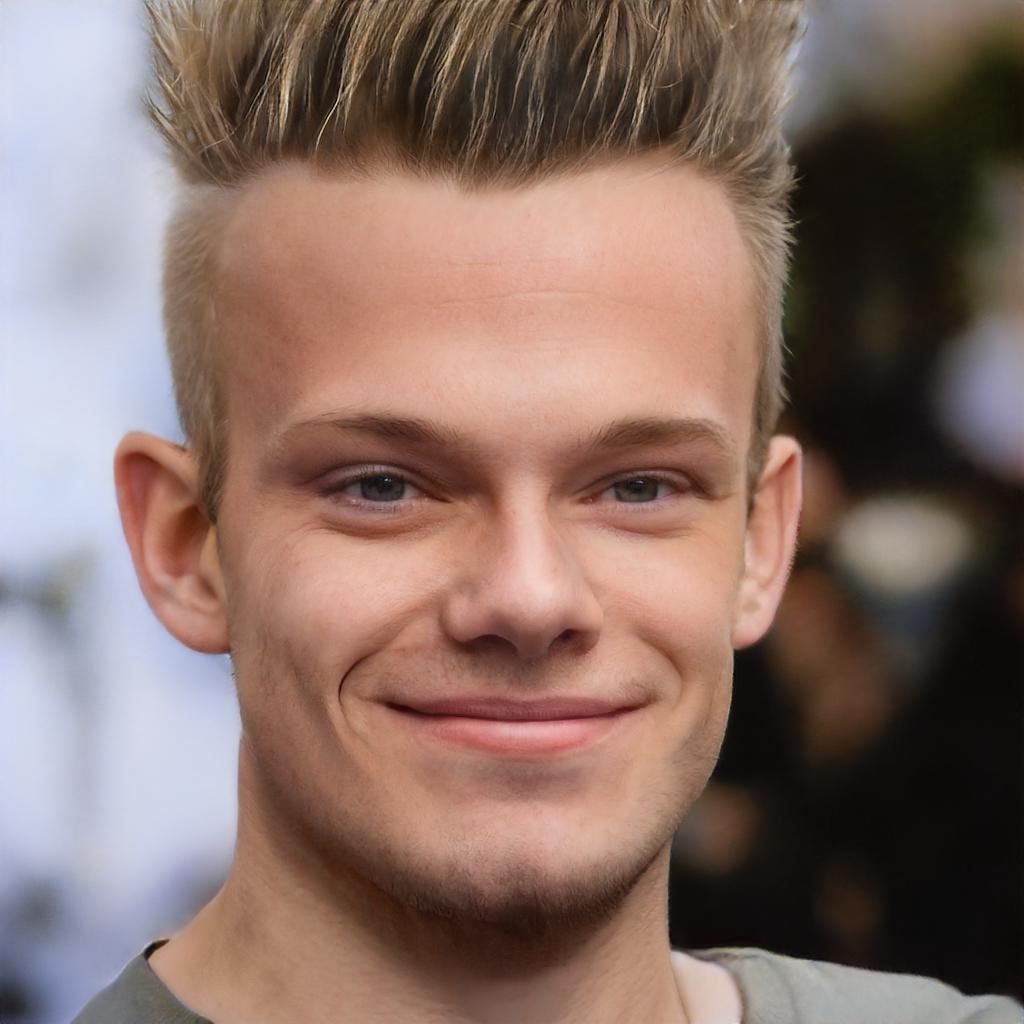} &
 			\includegraphics[width=0.14\textwidth]{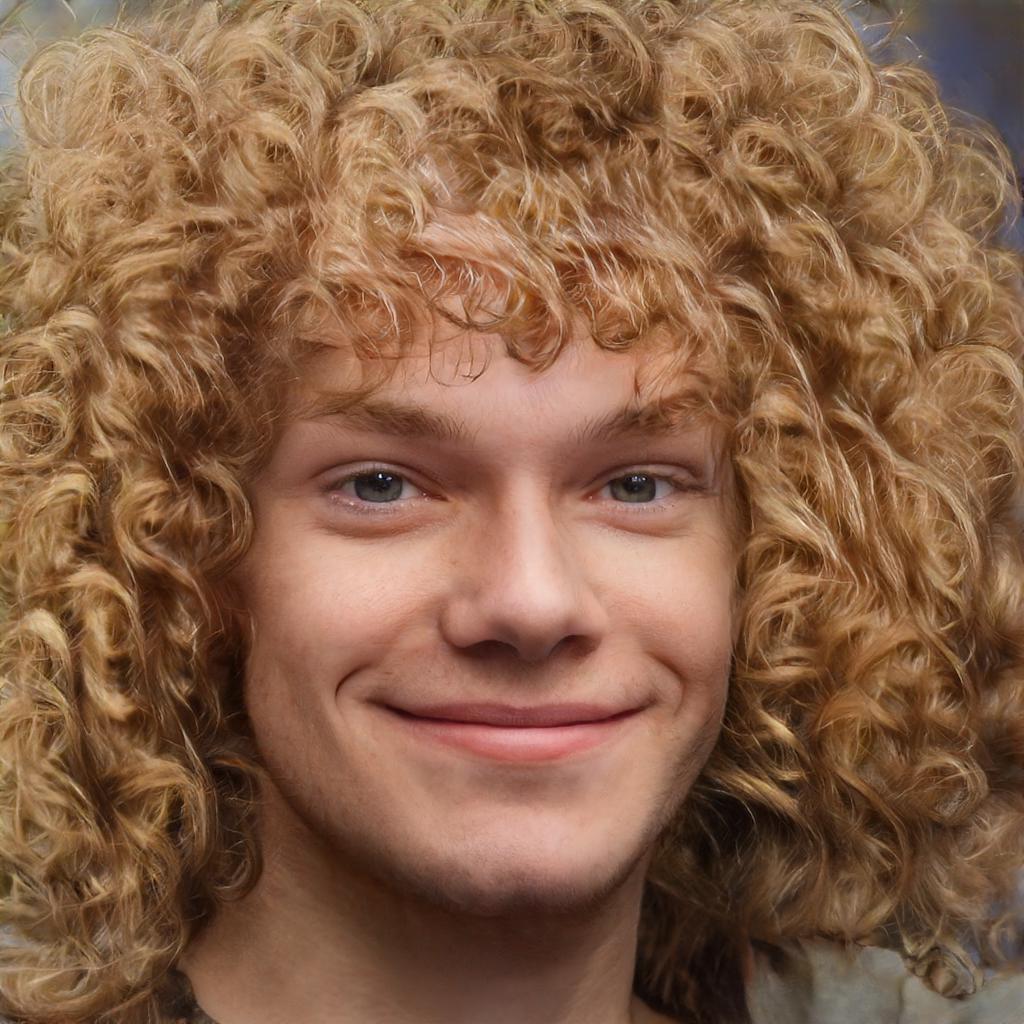} &
 			\includegraphics[width=0.14\textwidth]{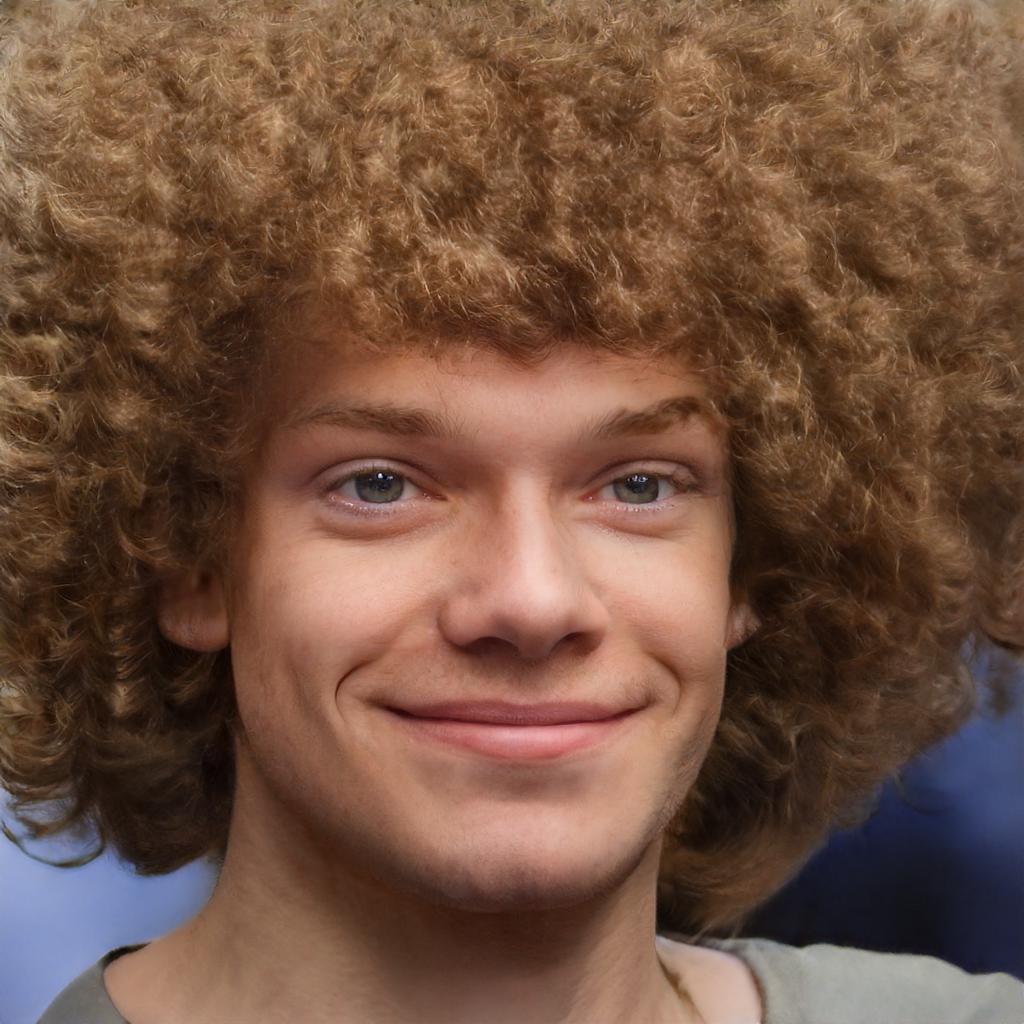} &
 			\includegraphics[width=0.14\textwidth]{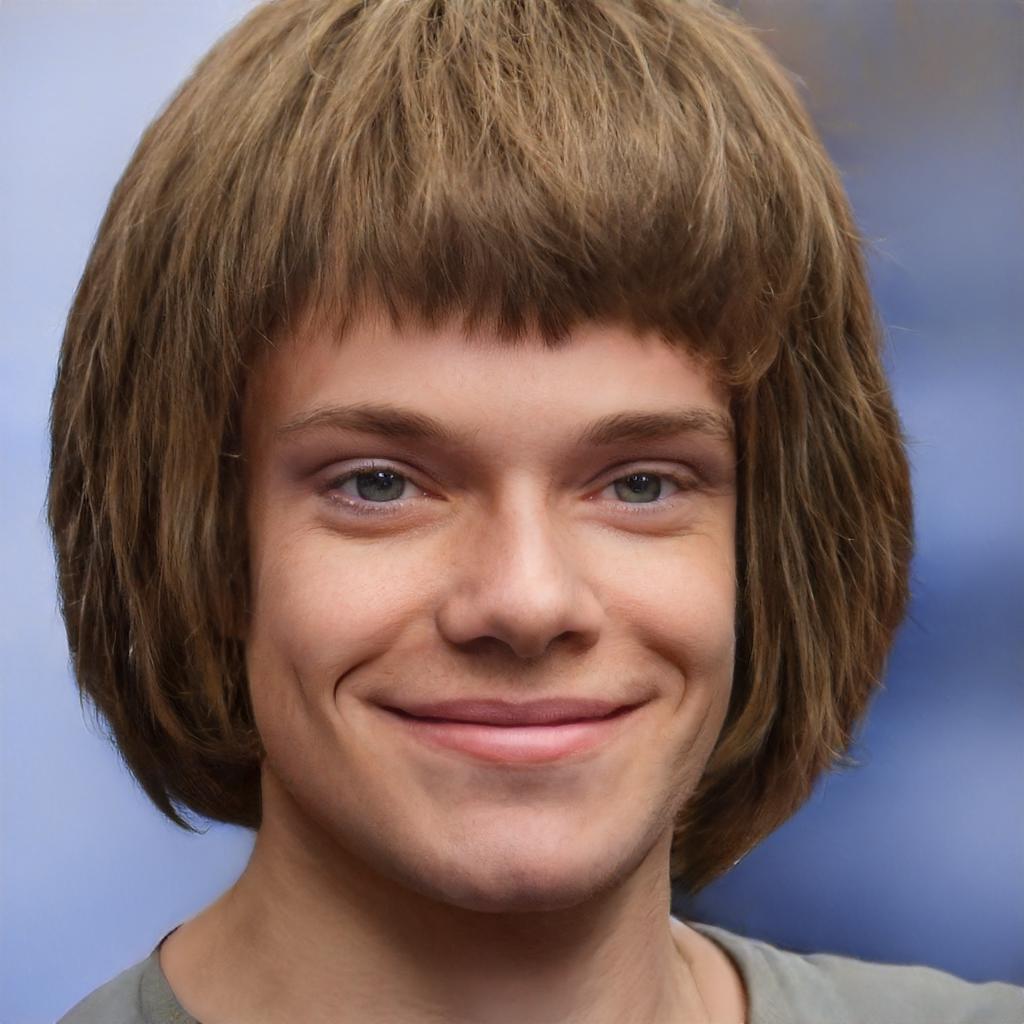} &
 			\includegraphics[width=0.14\textwidth]{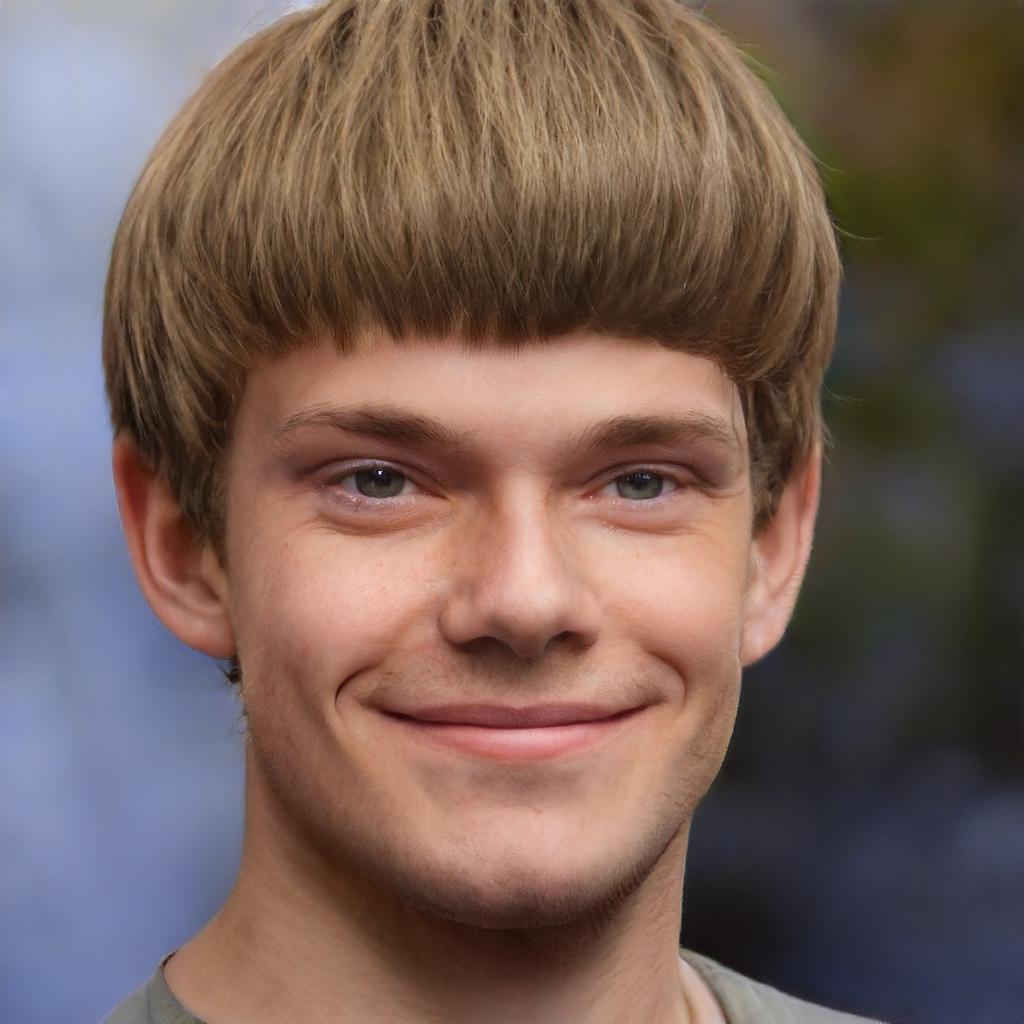} &
            \includegraphics[width=0.14\textwidth]{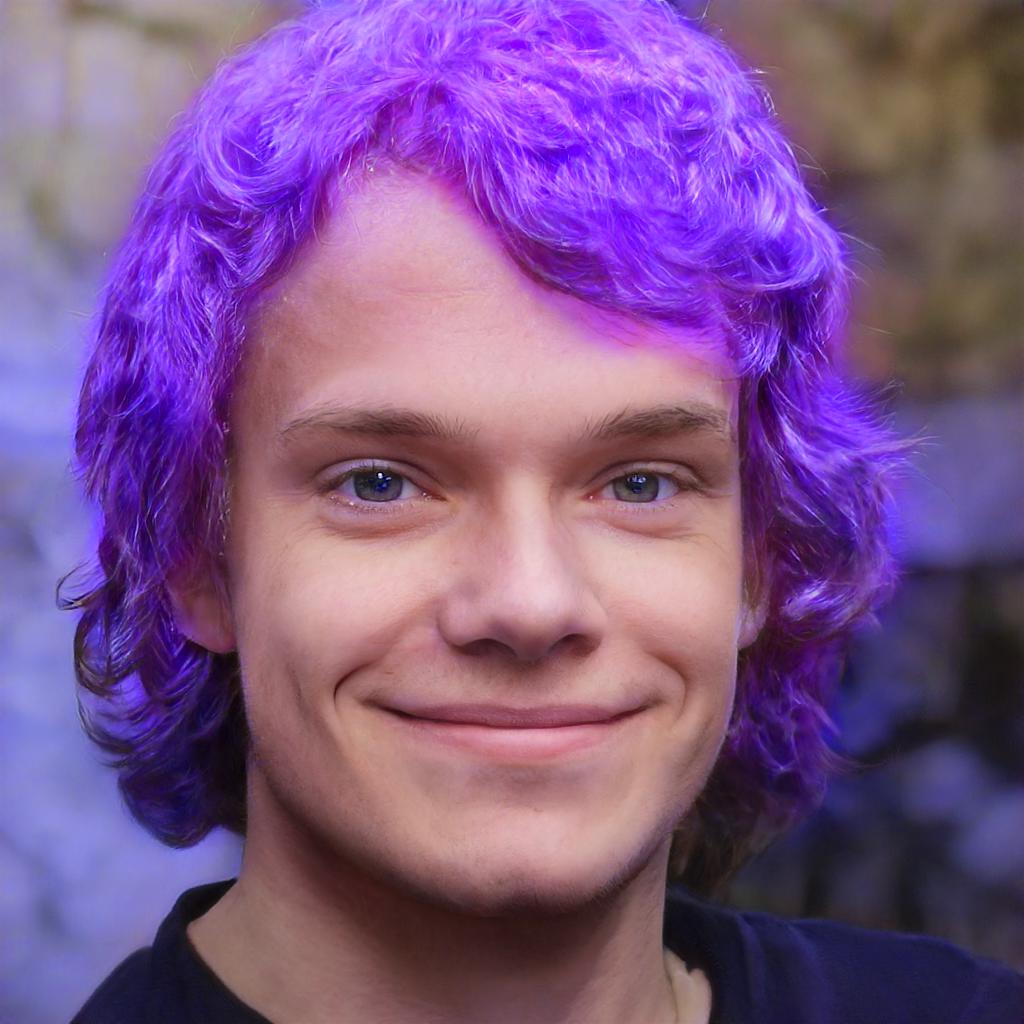} \\
            
            \includegraphics[width=0.14\textwidth]{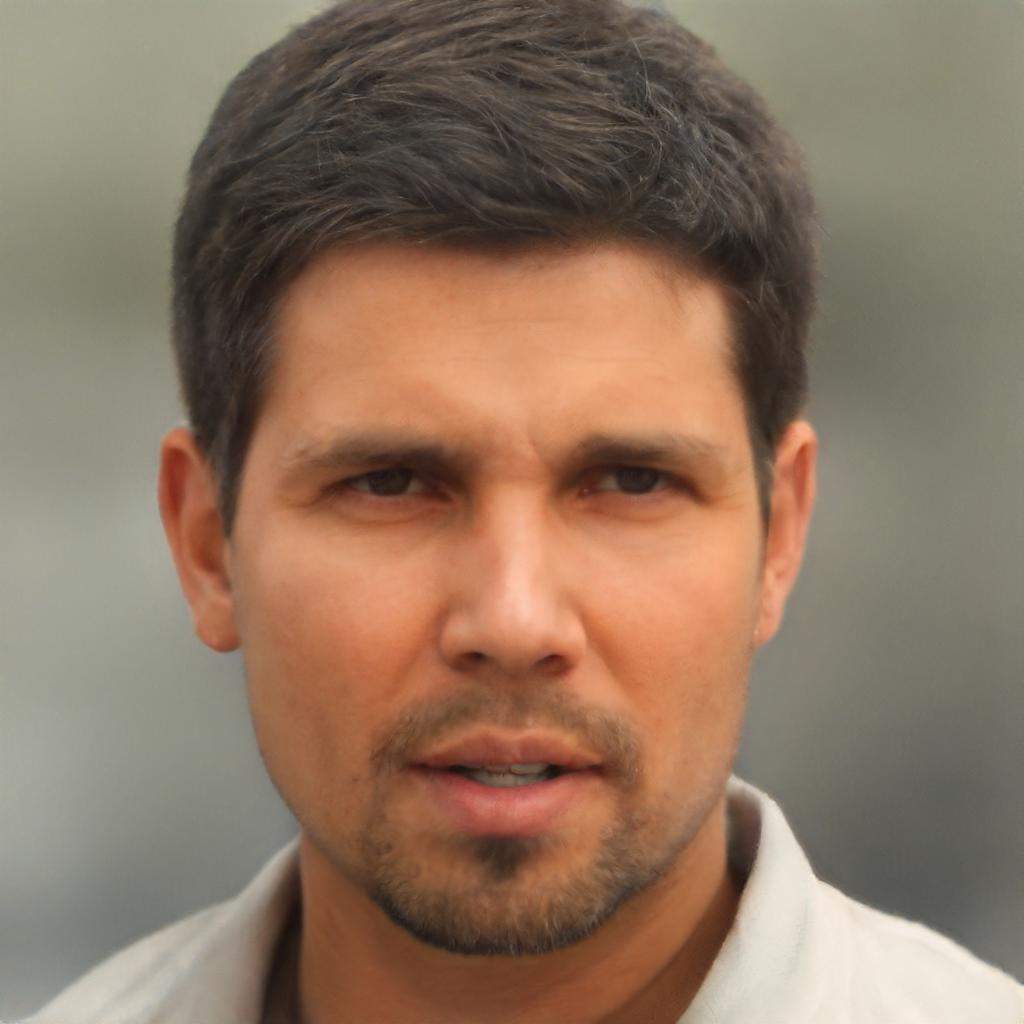} &
			\includegraphics[width=0.14\textwidth]{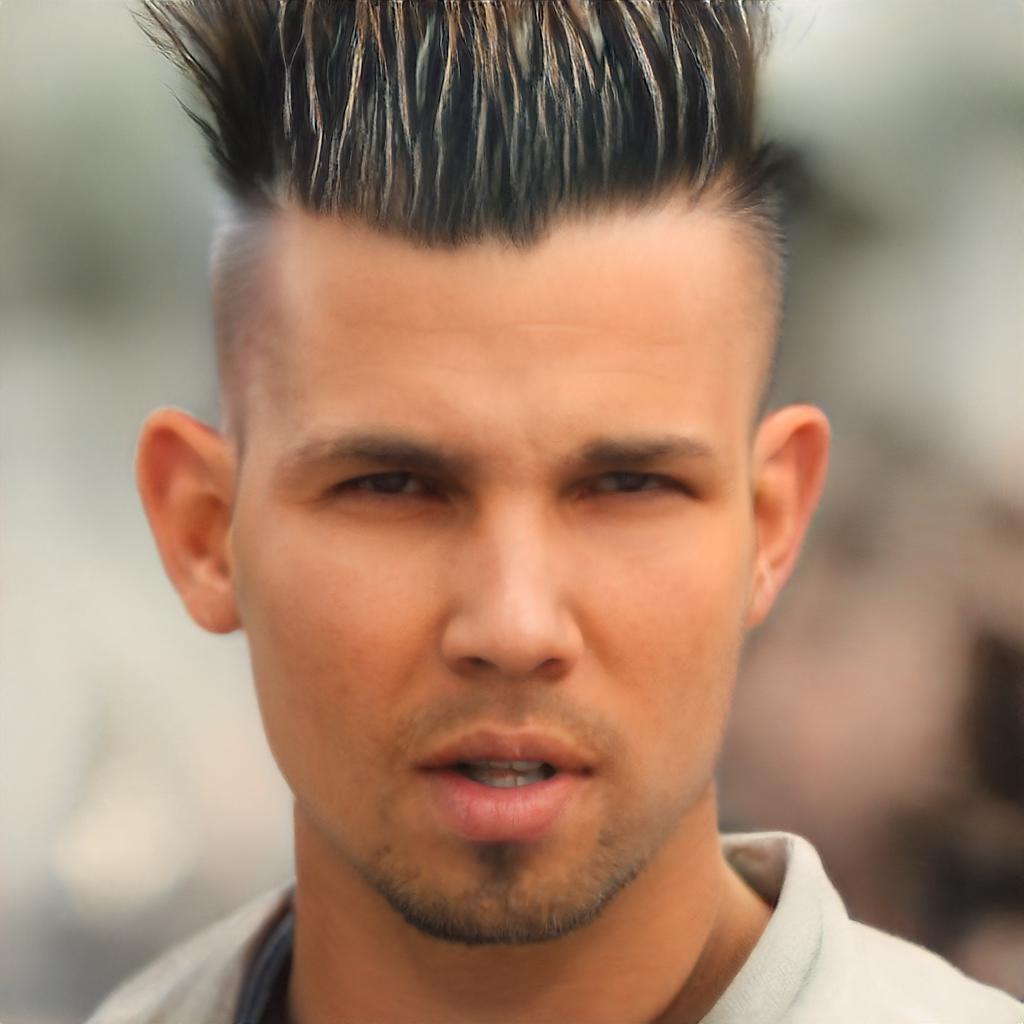} &
			\includegraphics[width=0.14\textwidth]{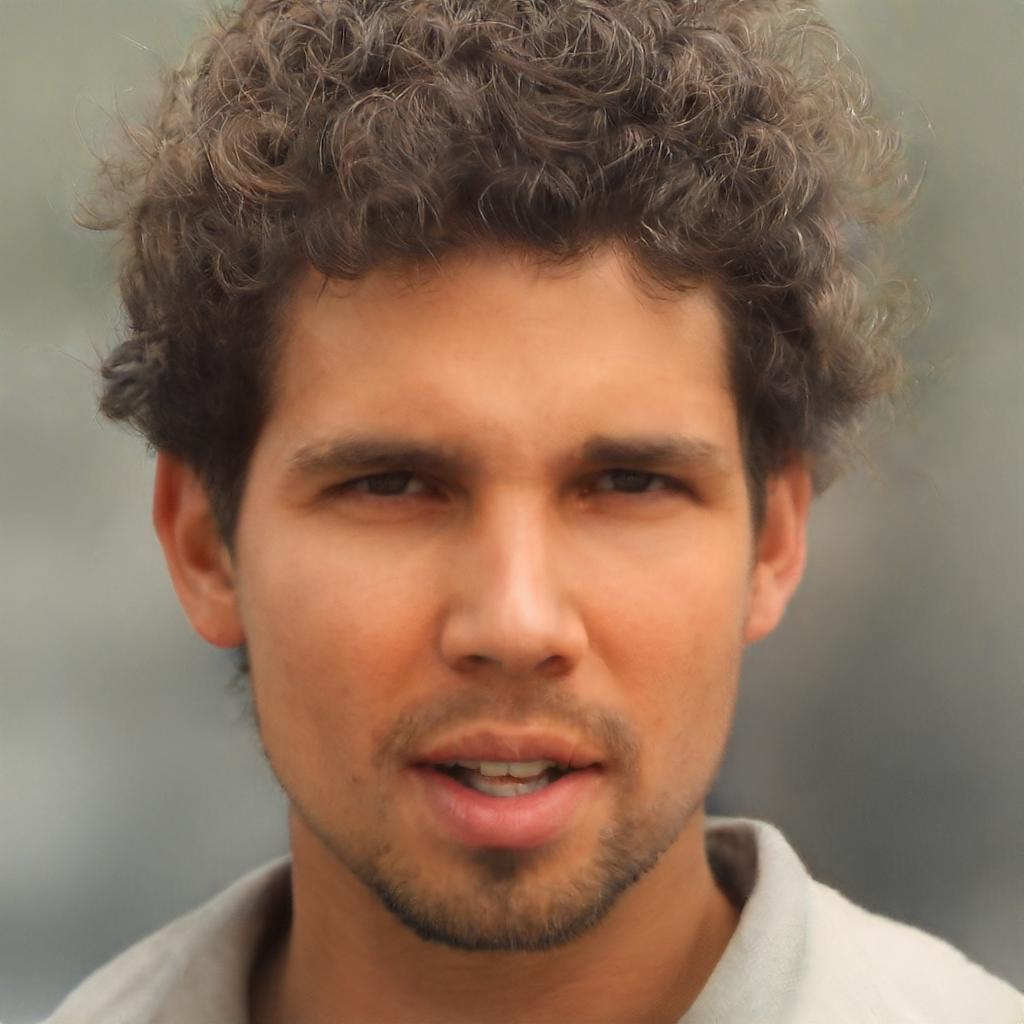} &
			\includegraphics[width=0.14\textwidth]{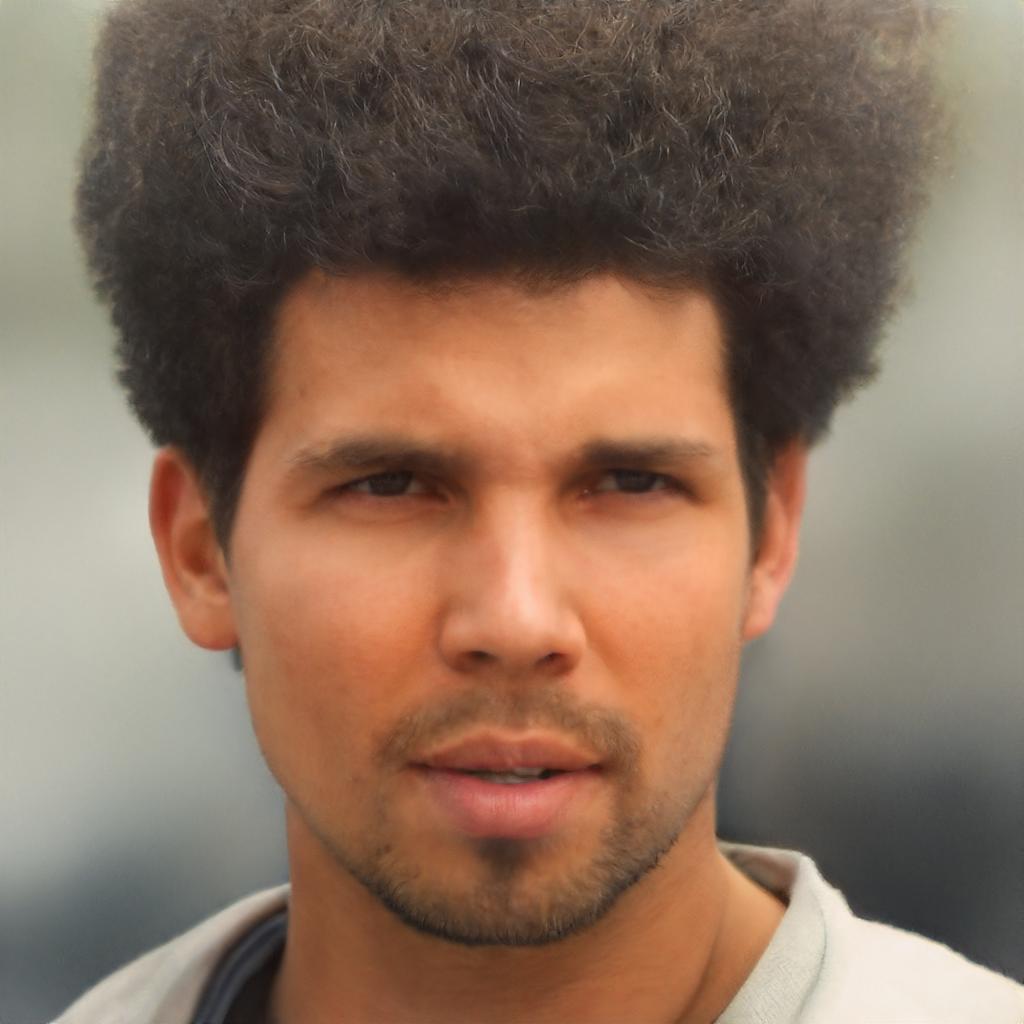} &
			\includegraphics[width=0.14\textwidth]{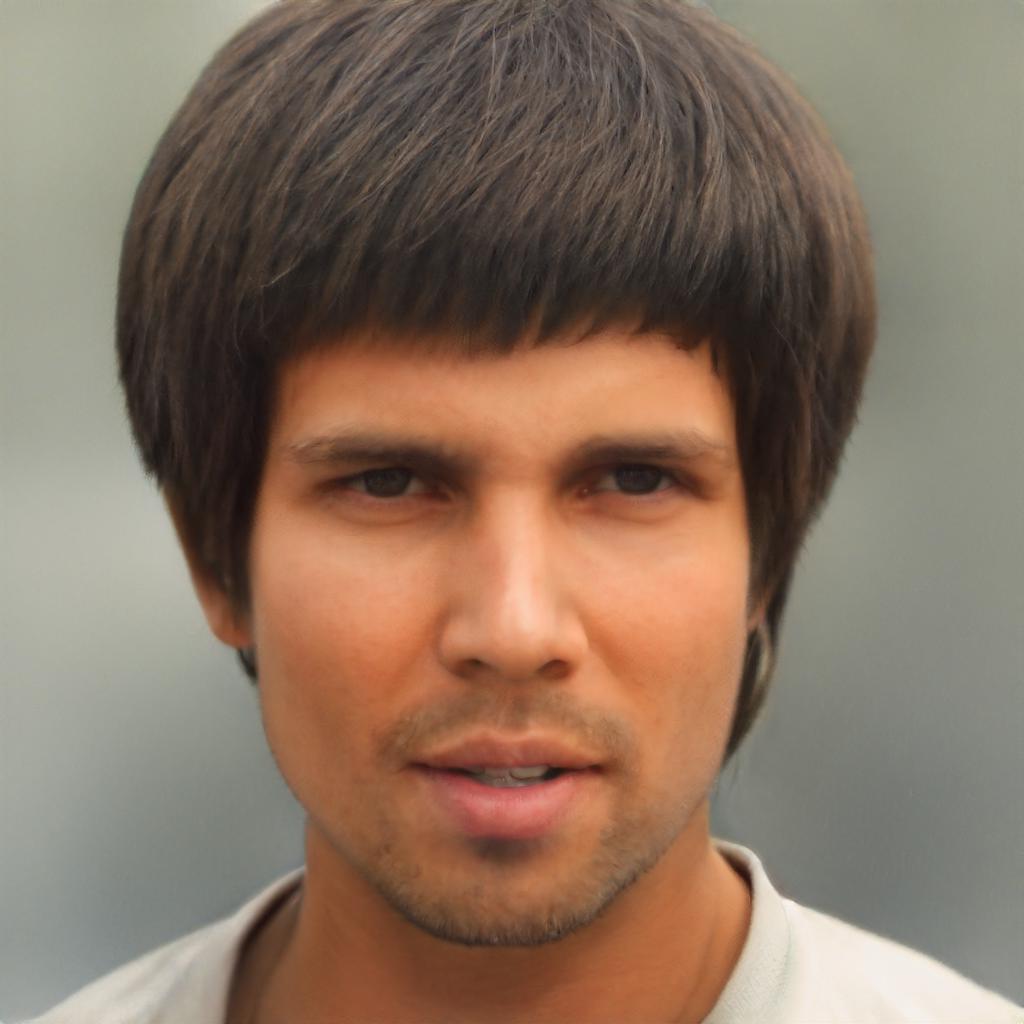} &
			\includegraphics[width=0.14\textwidth]{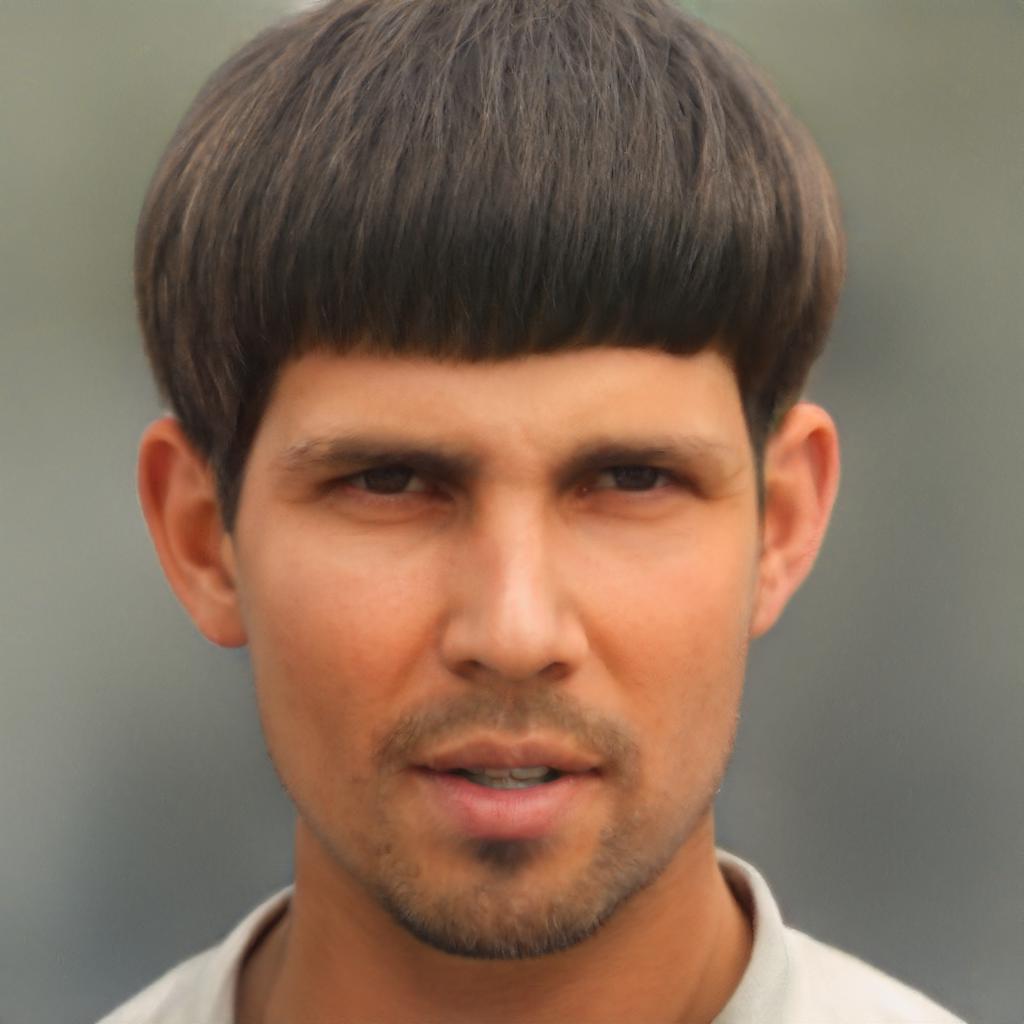} &
            \includegraphics[width=0.14\textwidth]{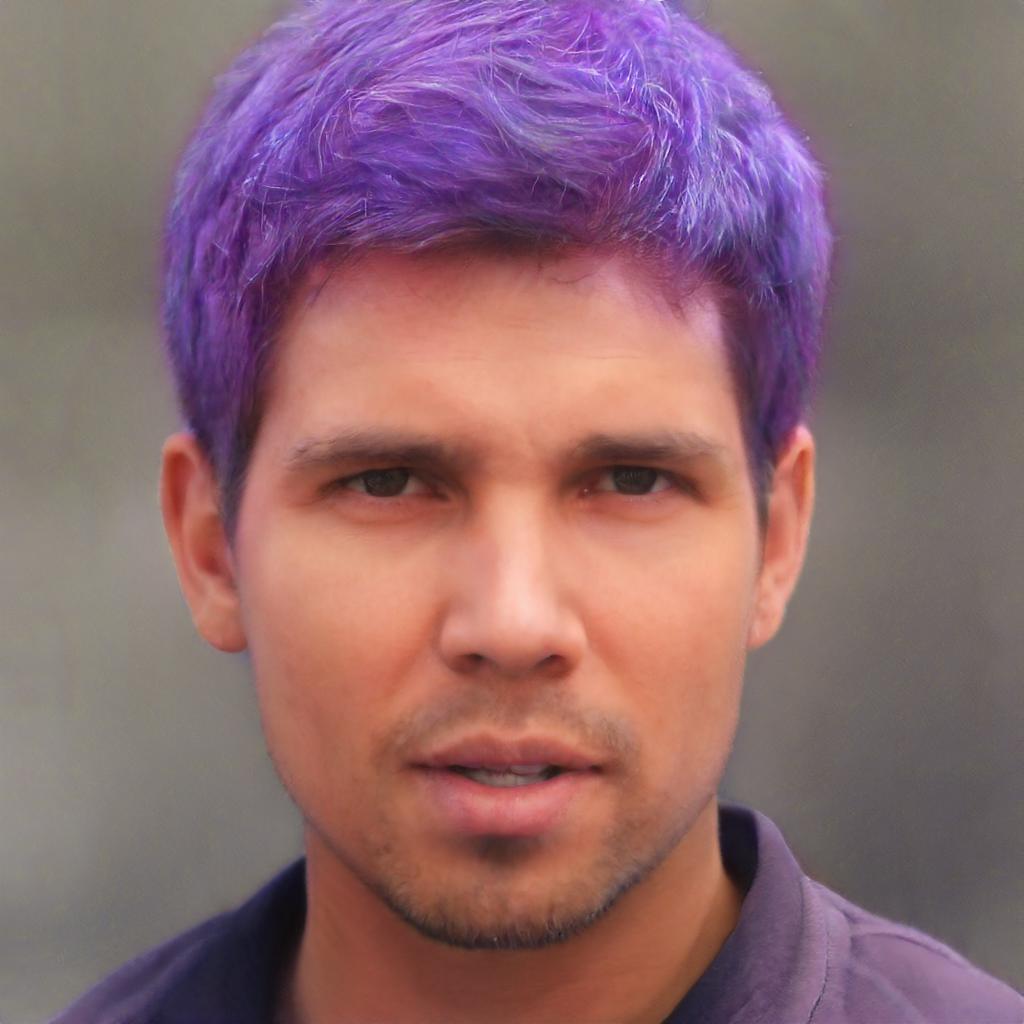} \\
            
			\includegraphics[width=0.14\textwidth]{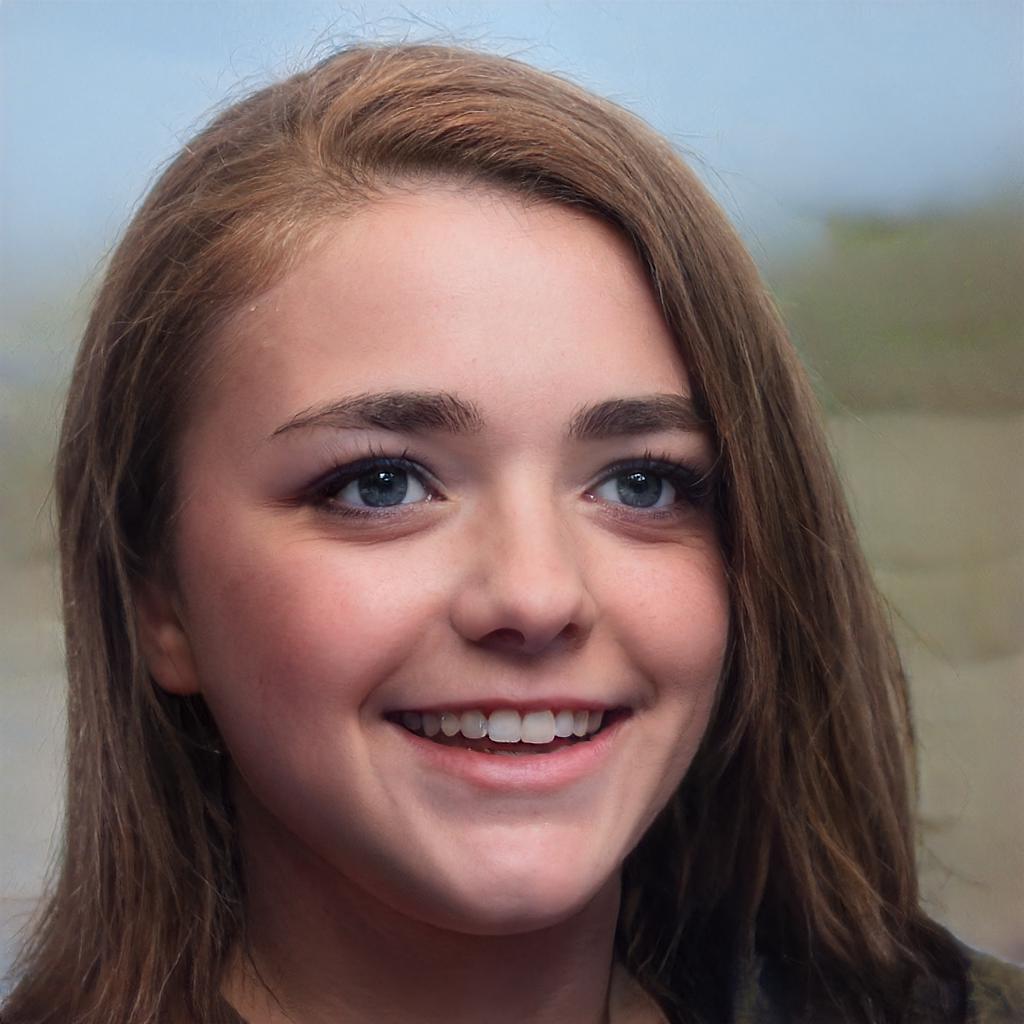} &
			\includegraphics[width=0.14\textwidth]{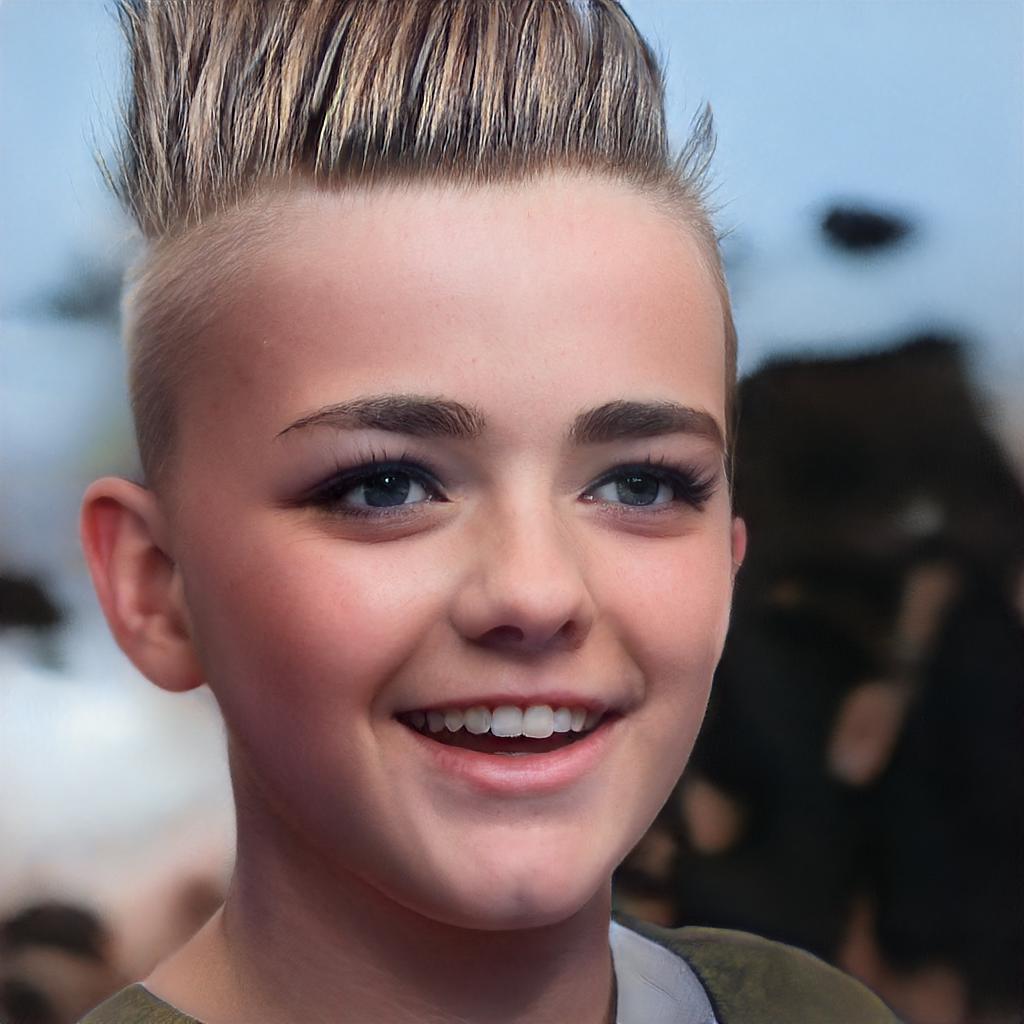} &
			\includegraphics[width=0.14\textwidth]{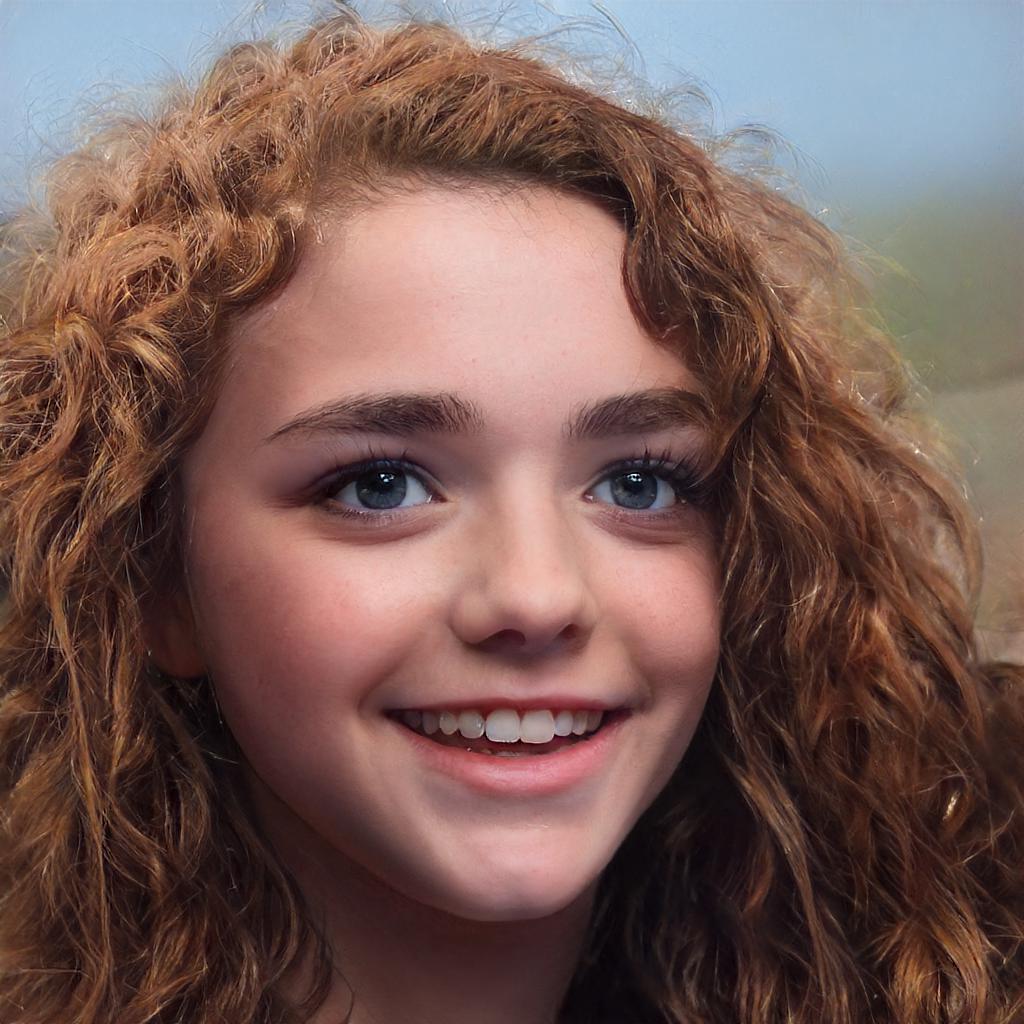} &
			\includegraphics[width=0.14\textwidth]{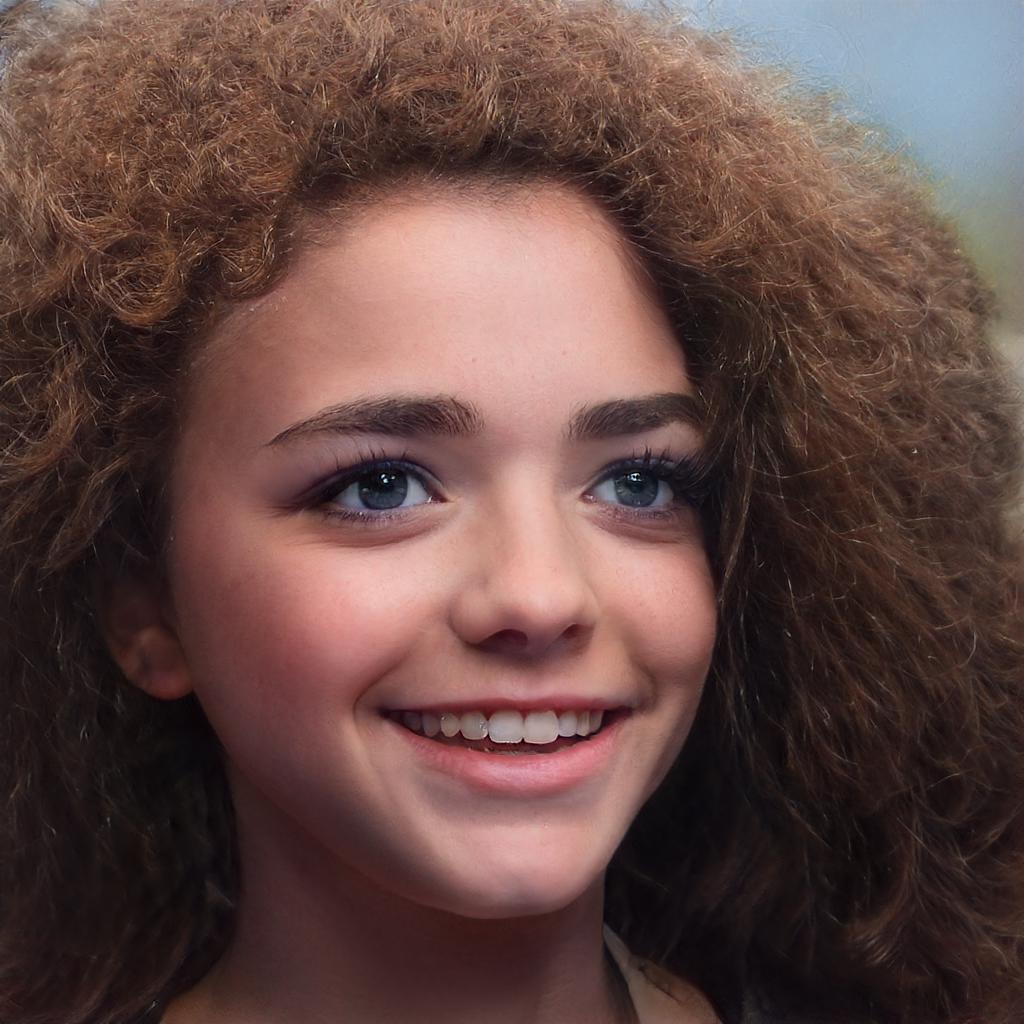} &
			\includegraphics[width=0.14\textwidth]{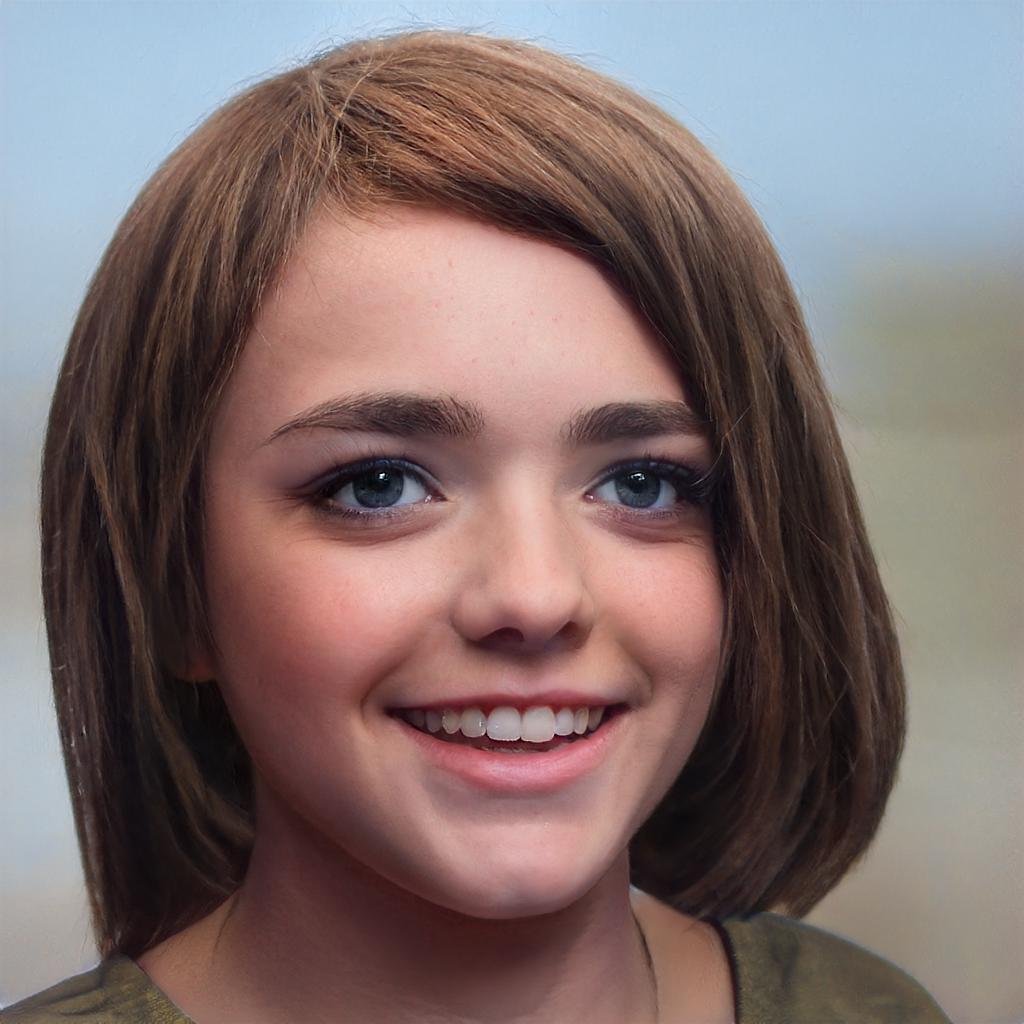} &
			\includegraphics[width=0.14\textwidth]{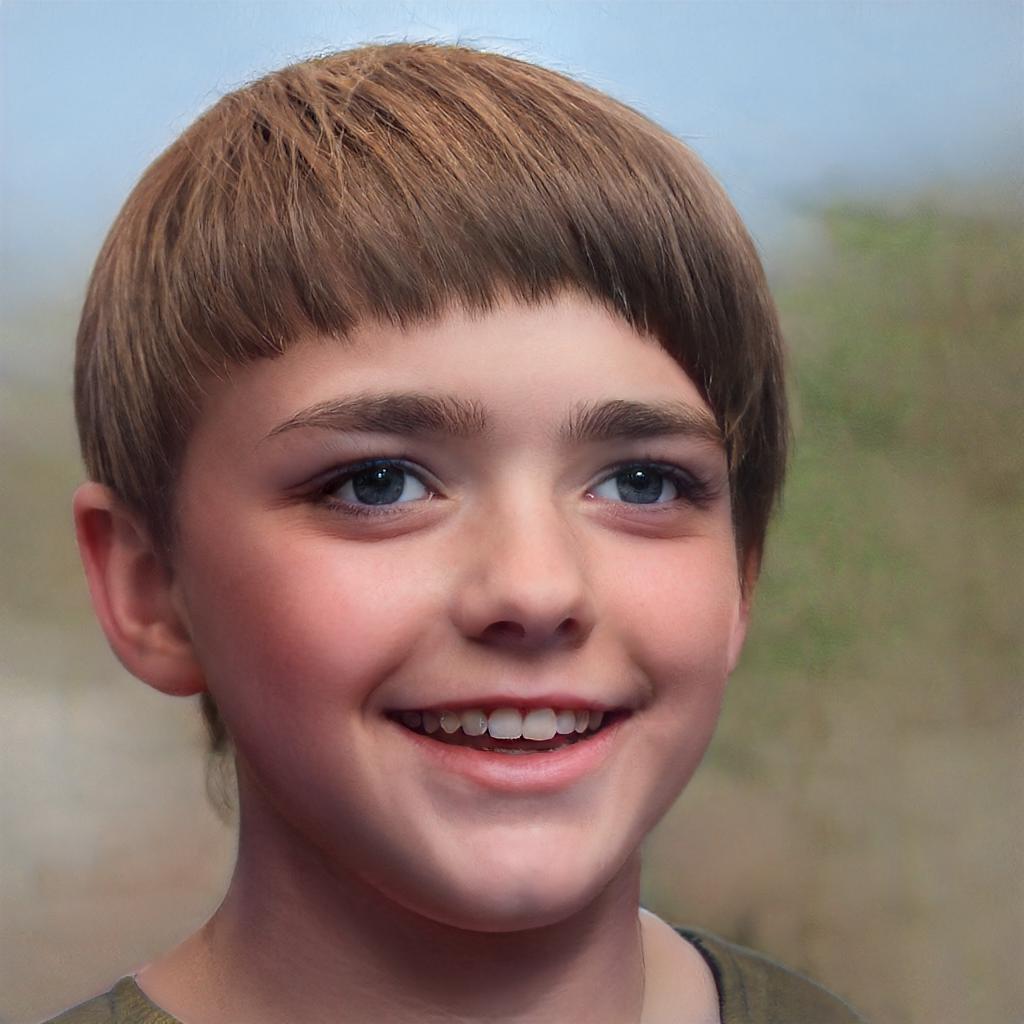} &
            \includegraphics[width=0.14\textwidth]{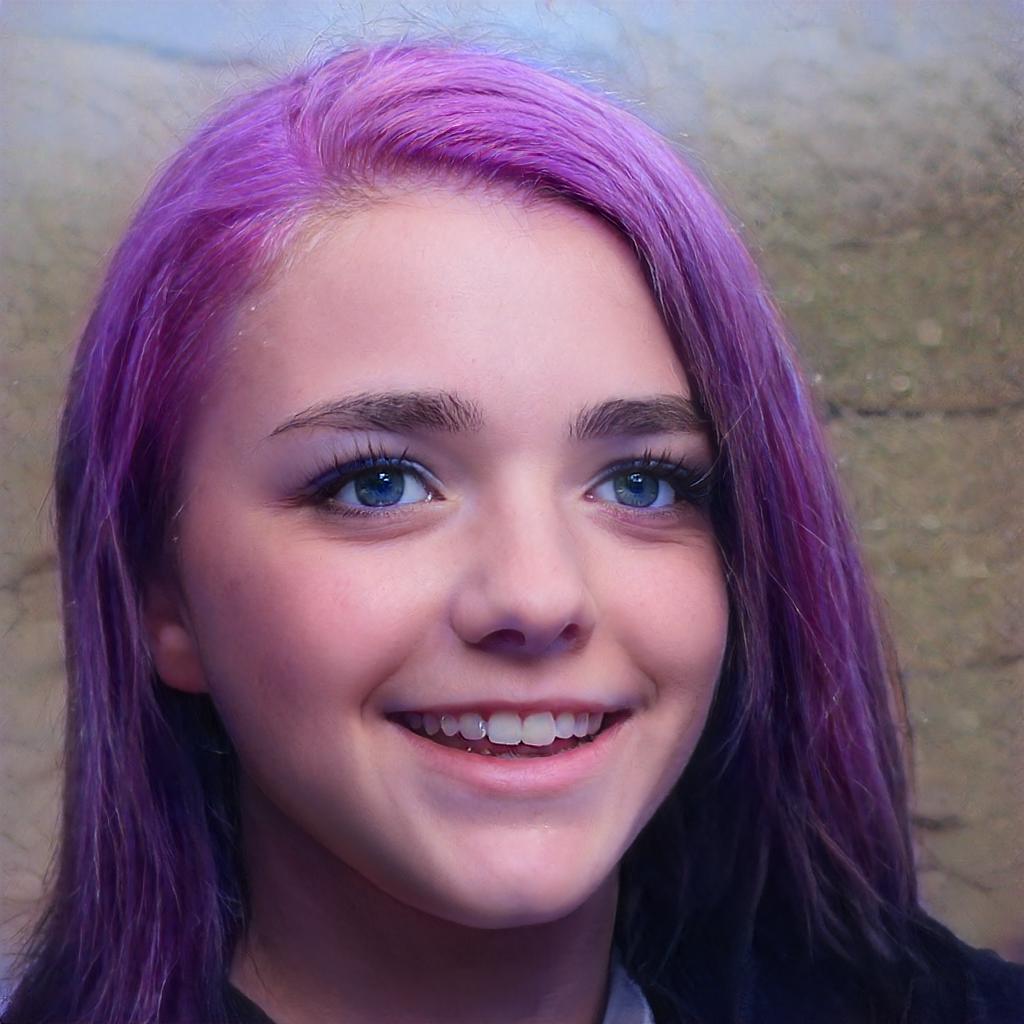} \\
            
			\includegraphics[width=0.14\textwidth]{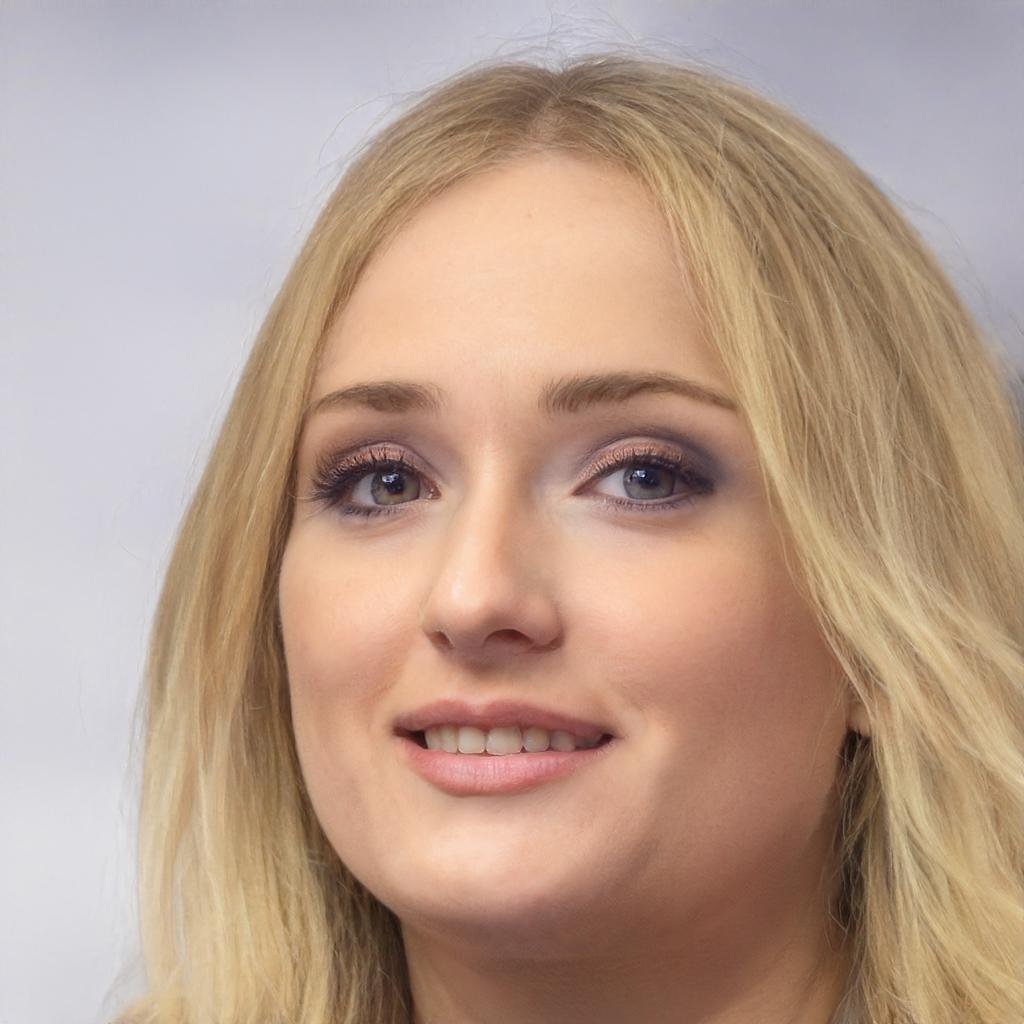} &
			\includegraphics[width=0.14\textwidth]{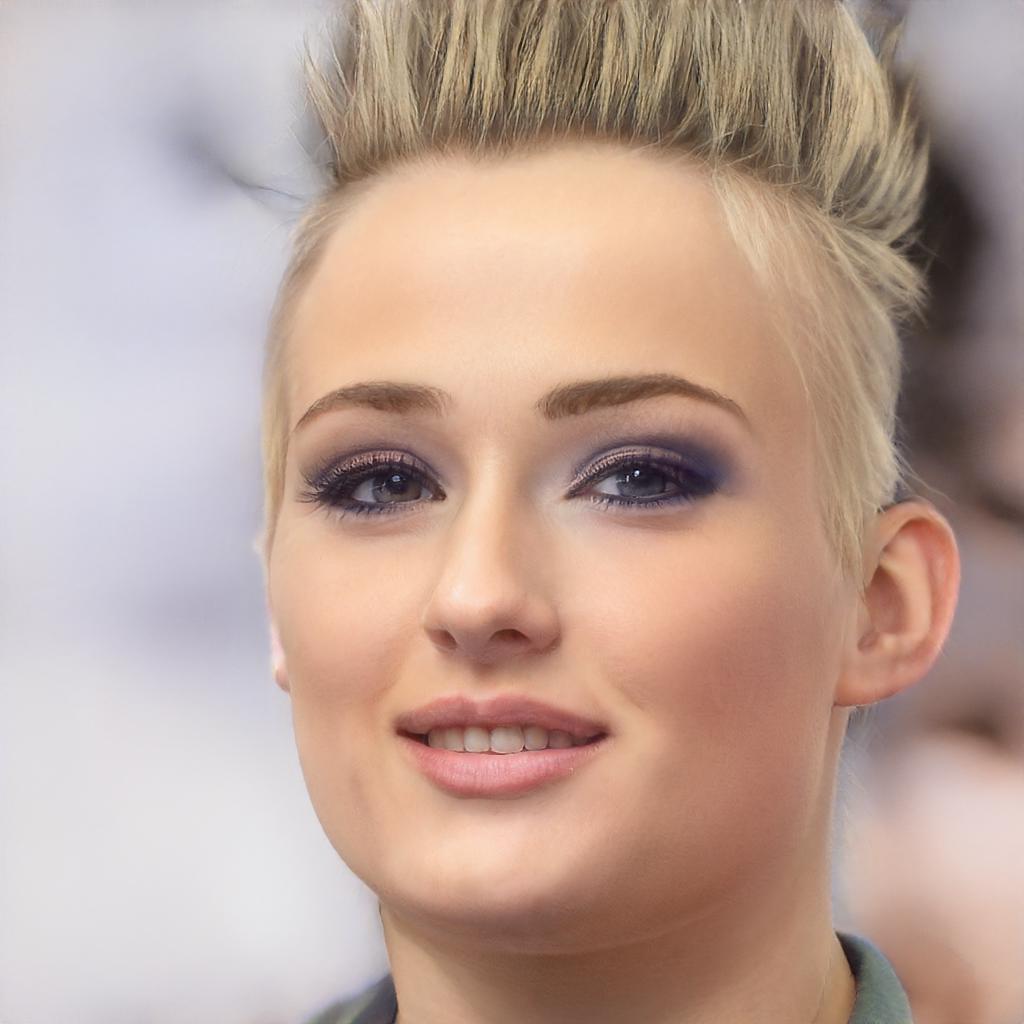} &
			\includegraphics[width=0.14\textwidth]{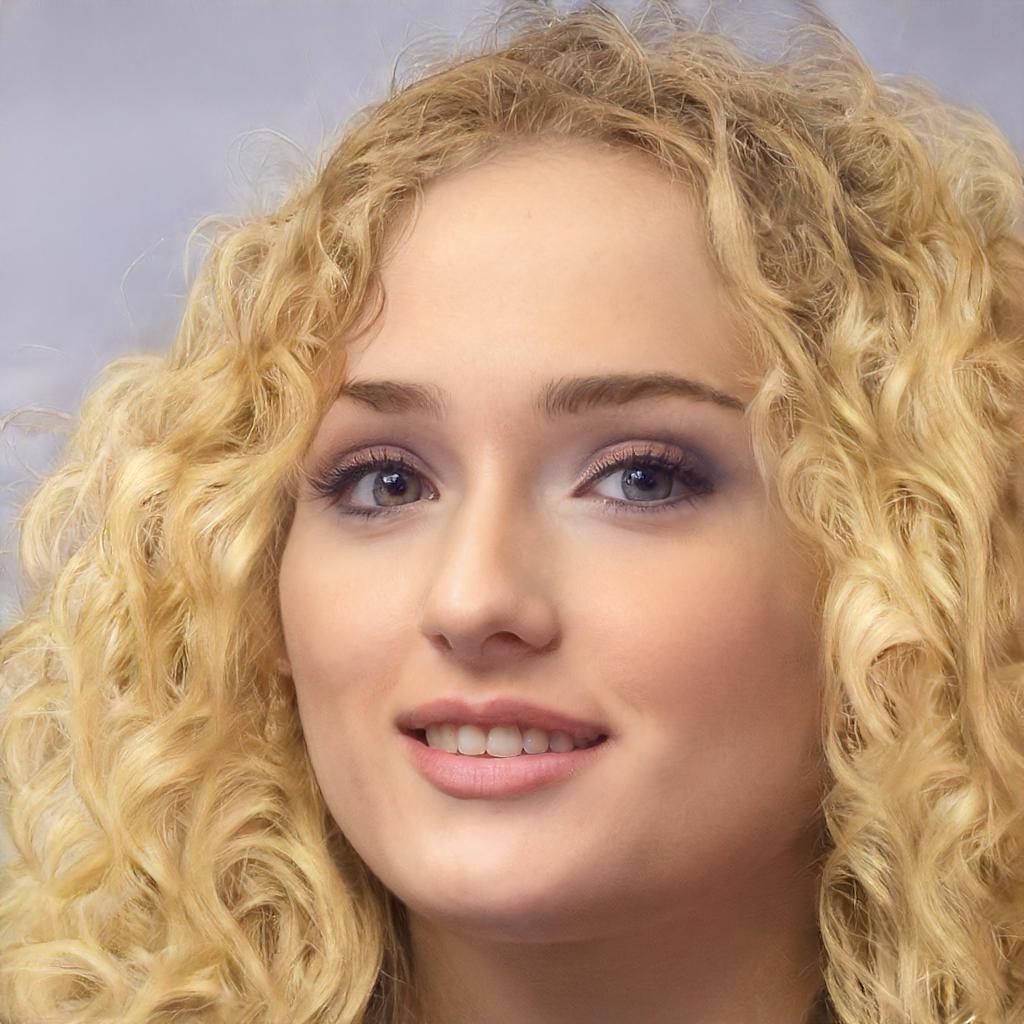} &
			\includegraphics[width=0.14\textwidth]{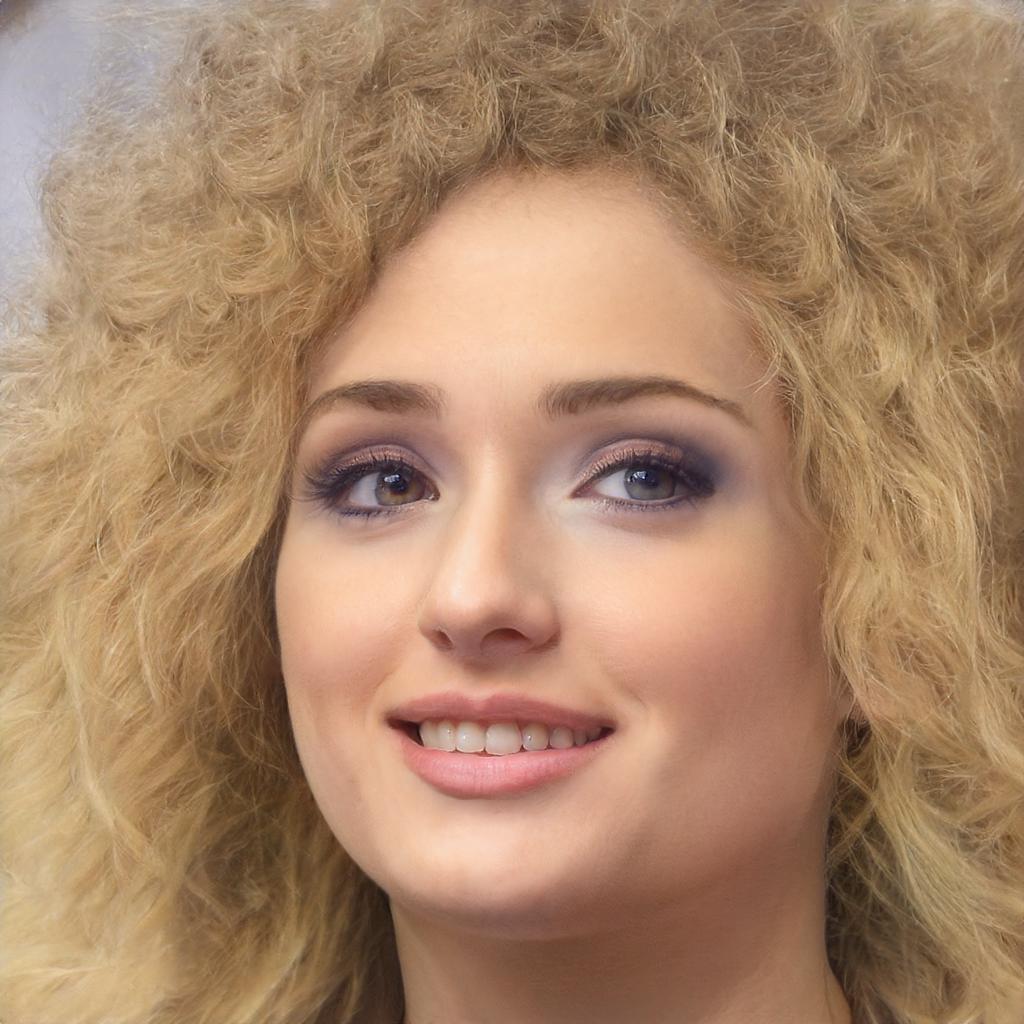} &
			\includegraphics[width=0.14\textwidth]{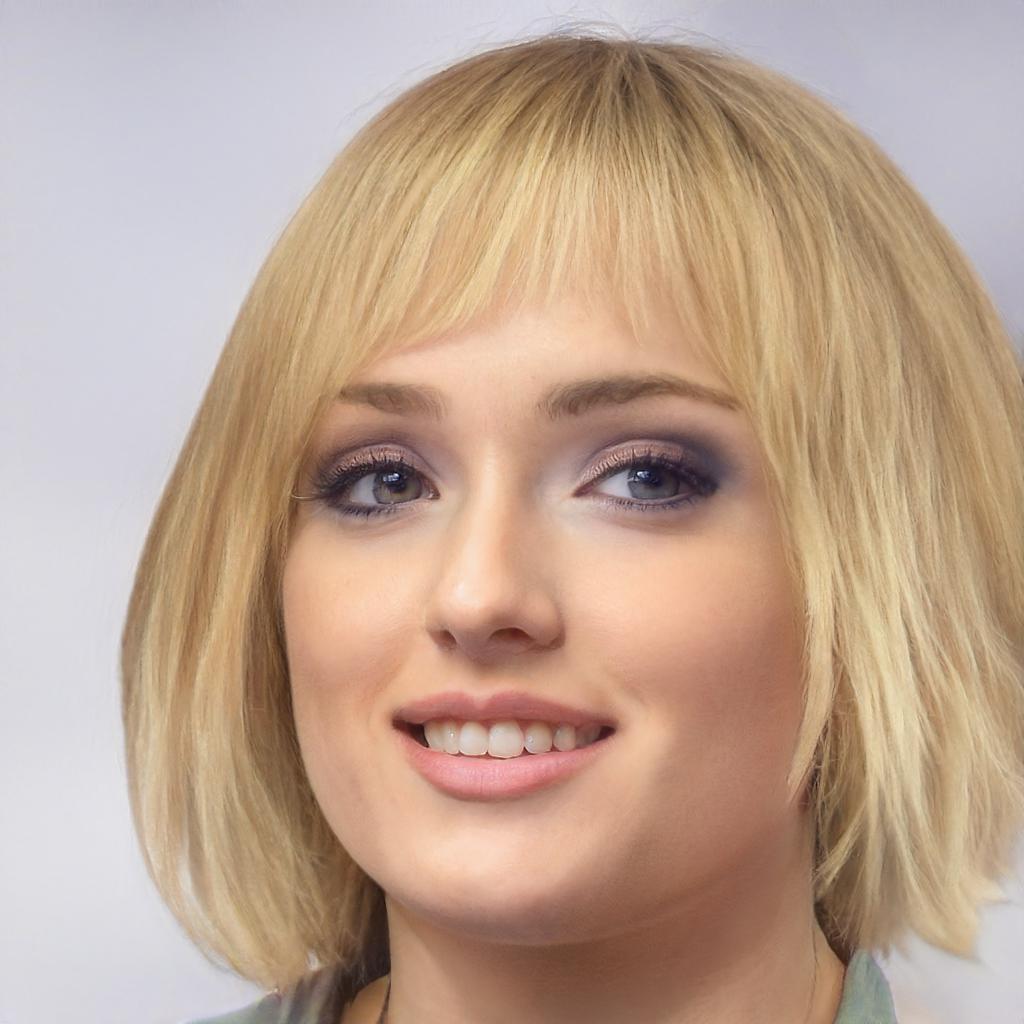} &
			\includegraphics[width=0.14\textwidth]{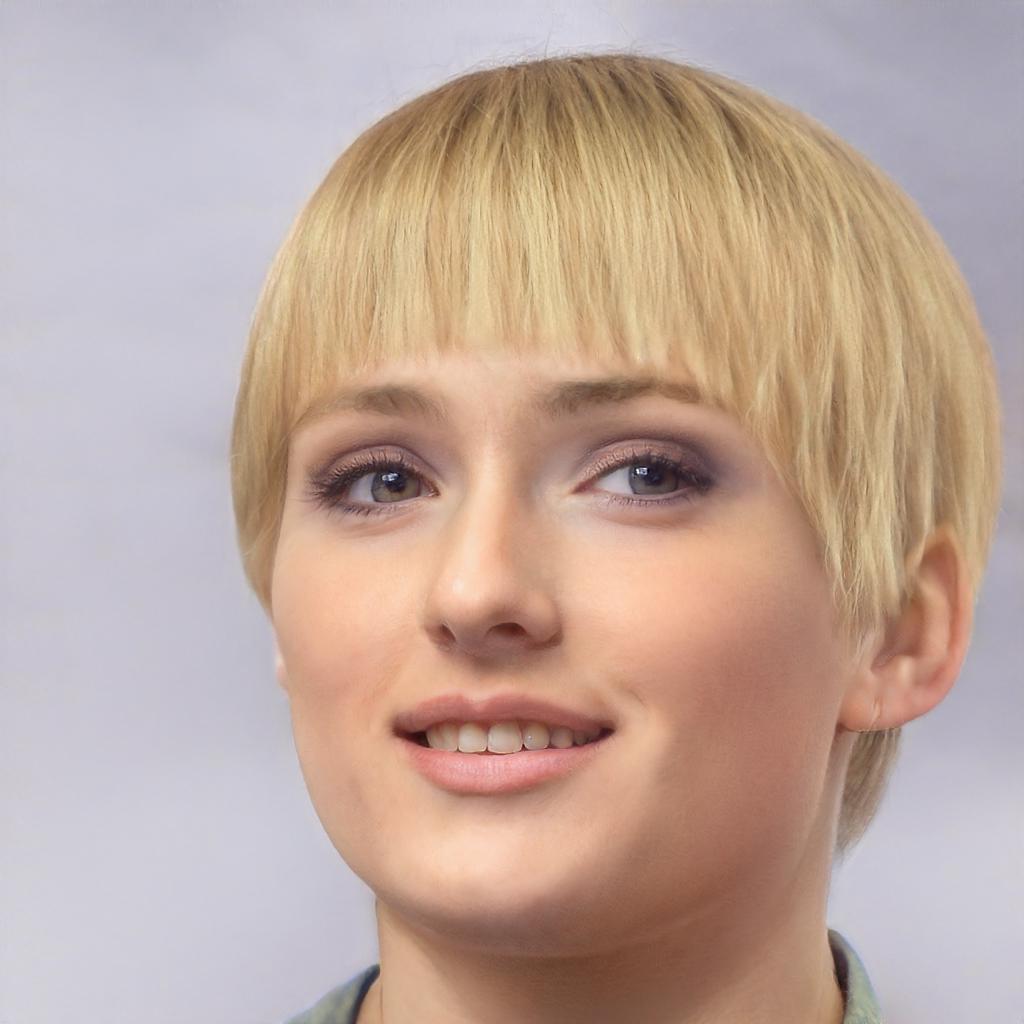} &
            \includegraphics[width=0.14\textwidth]{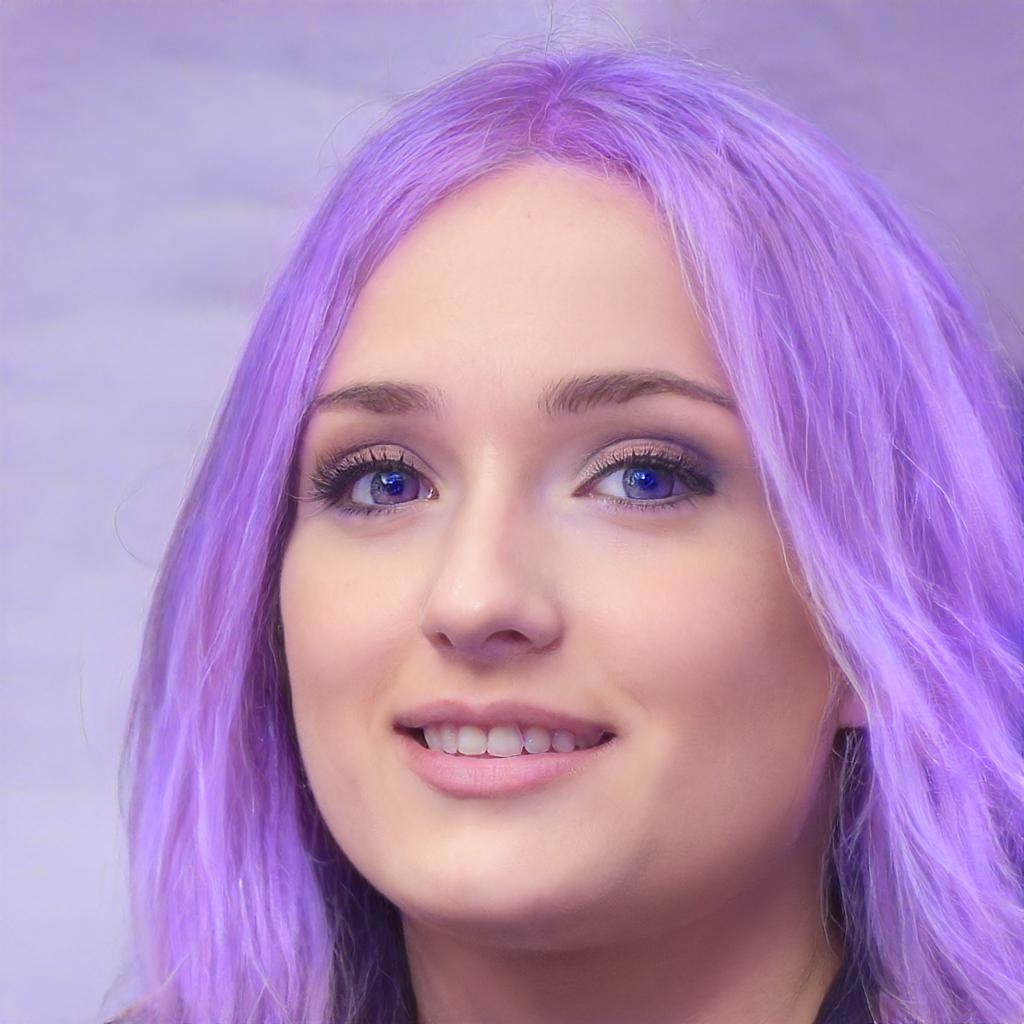} \\
            
            Input & Mohawk & Curly & Afro & Bobcut & Bowlcut & Purple
		\end{tabular}
	}
	\caption{Hair style manipulations obtained by the latent mapper. Except for the purple hair, all mappers were trained without $M^{f}$. \vspace{5mm}}
	\label{fig:supp-hair}

\end{figure*}

\begin{figure}[p]
	\setlength{\tabcolsep}{1pt}
	\centering
	{\footnotesize
		\begin{tabular}{c c c c}
			\includegraphics[width=0.23\linewidth]{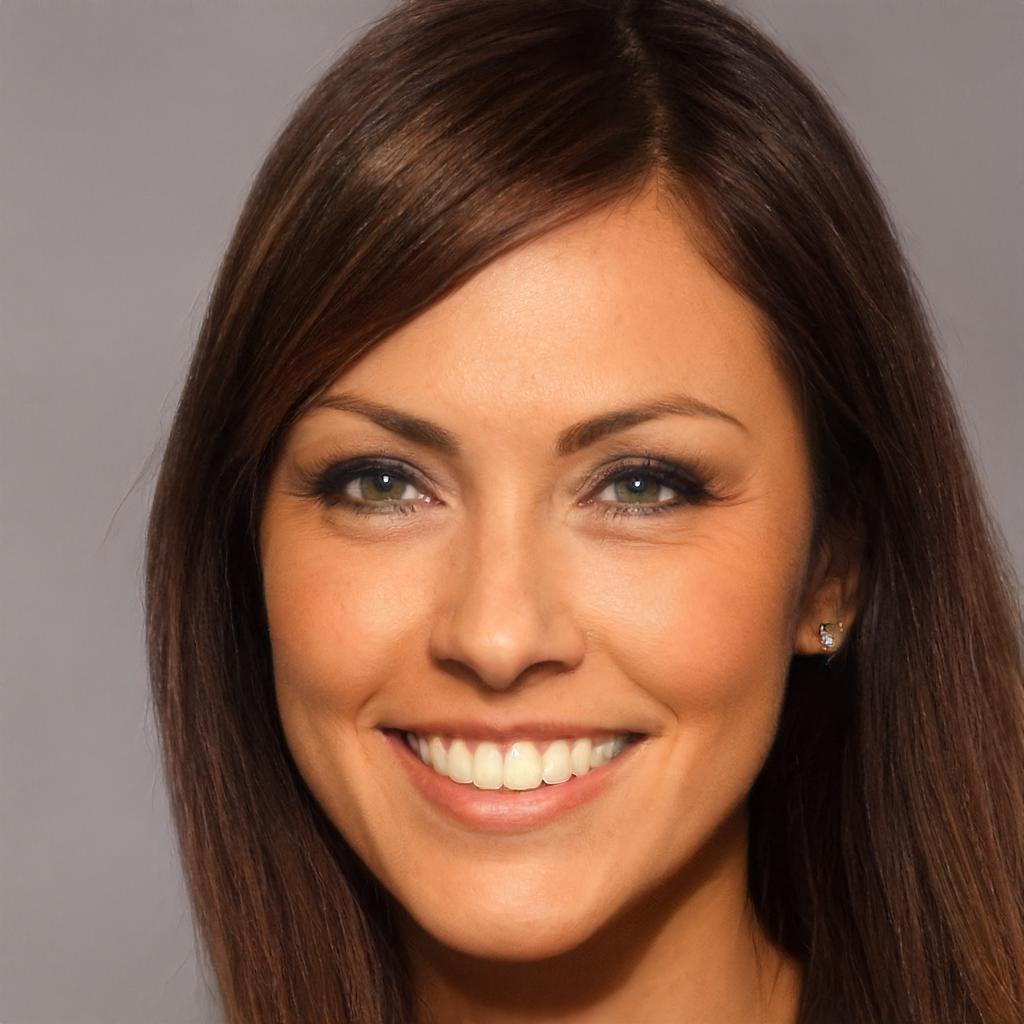} &
 			\includegraphics[width=0.23\linewidth]{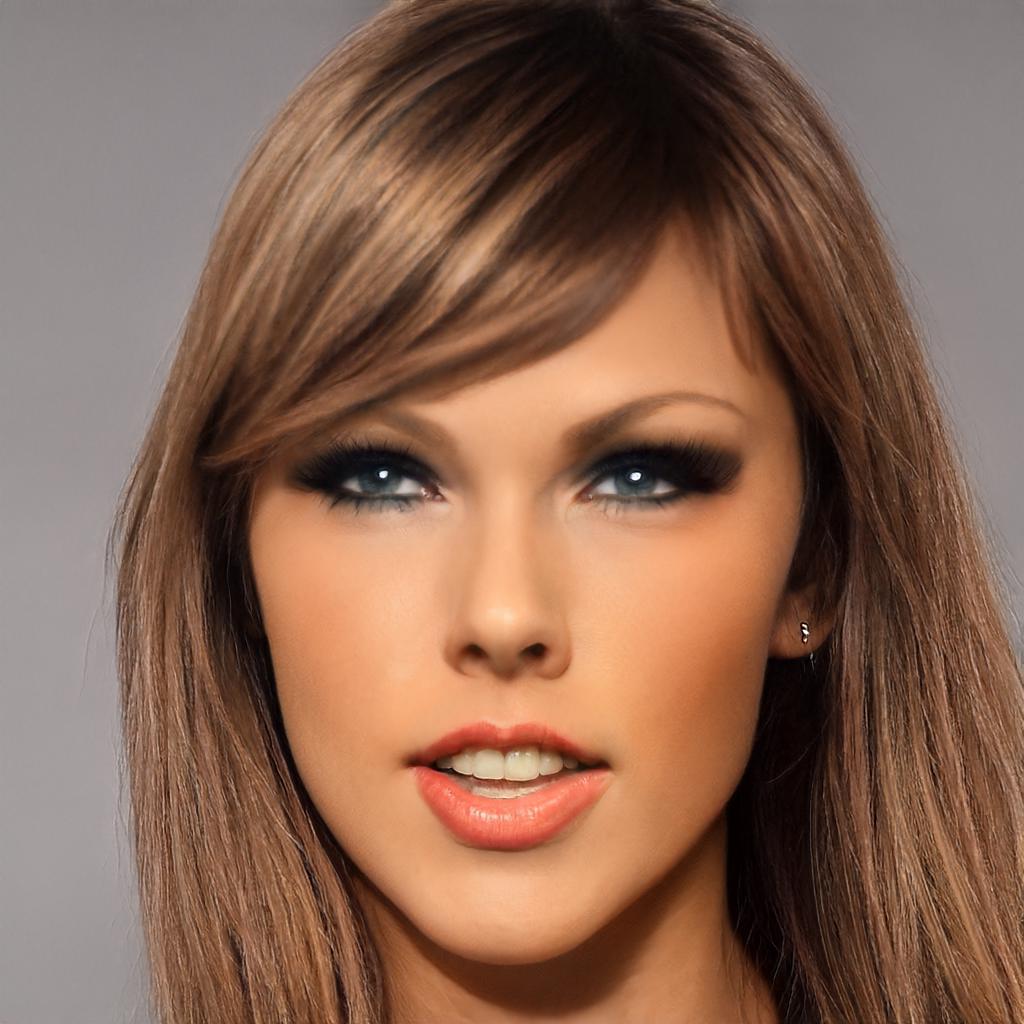} &
 			\includegraphics[width=0.23\linewidth]{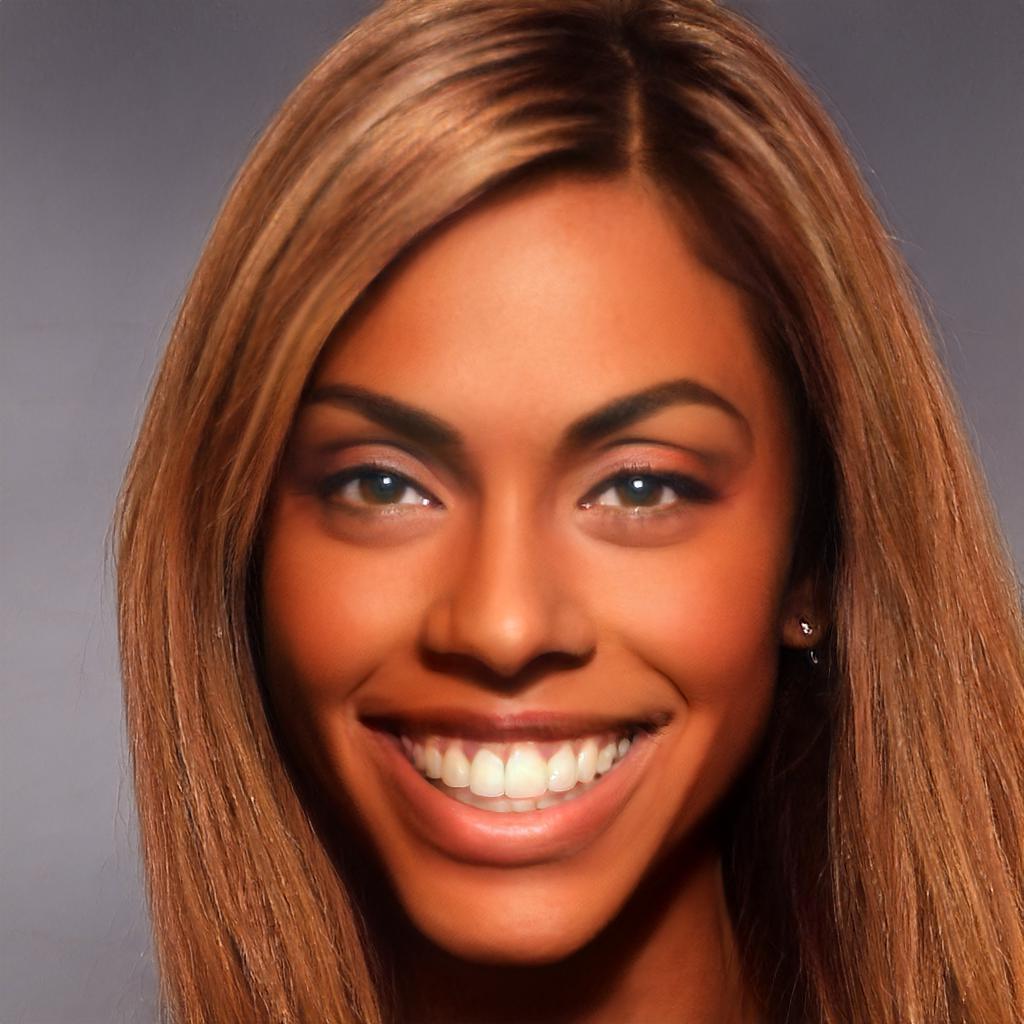} &
 			 \includegraphics[width=0.23\linewidth]{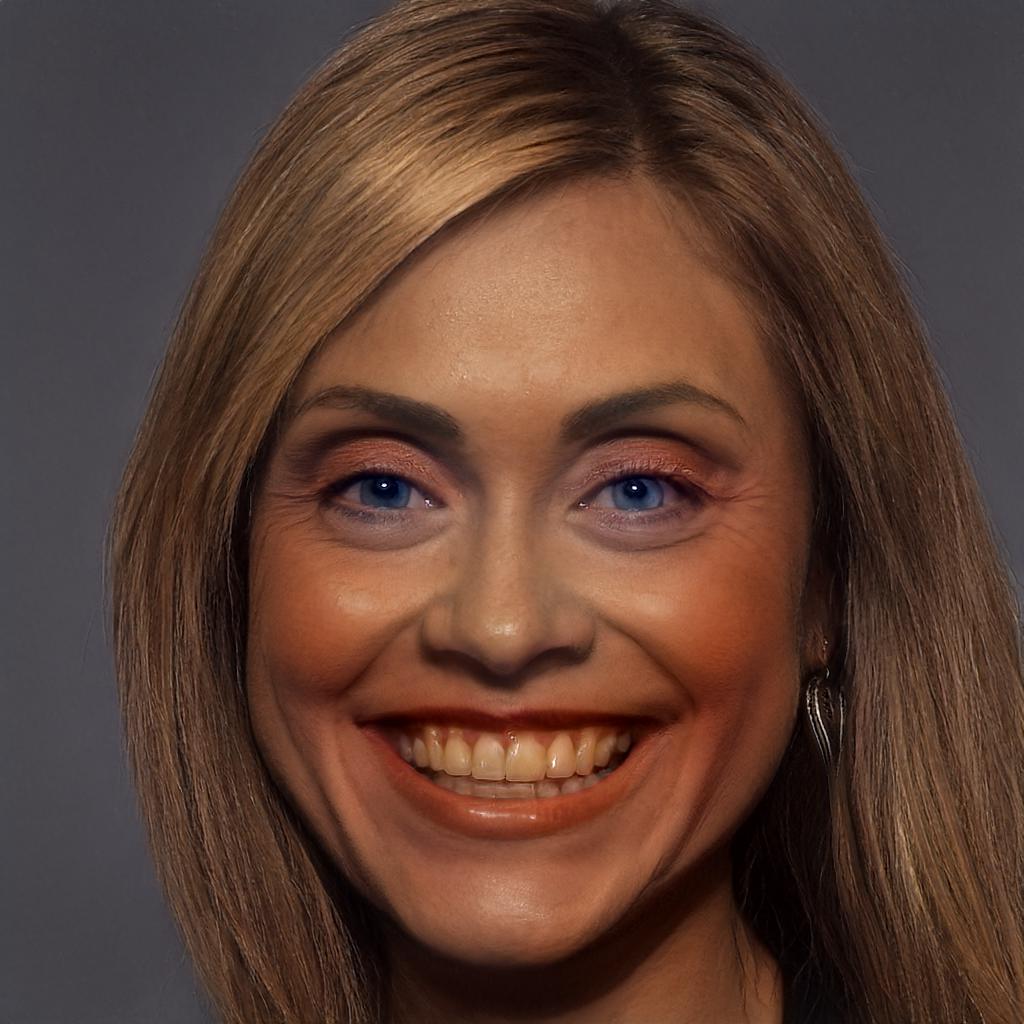}\\
 			
 			\includegraphics[width=0.23\linewidth]{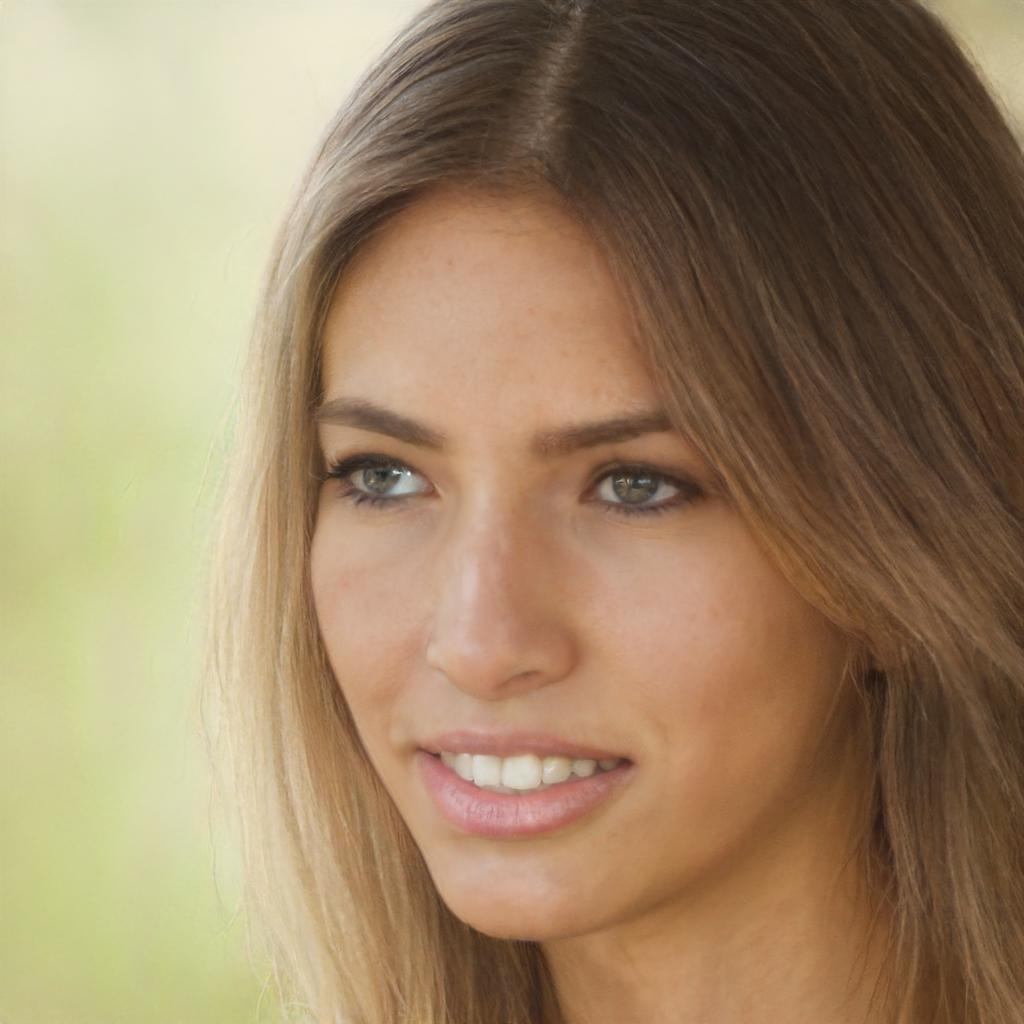} &
 			\includegraphics[width=0.23\linewidth]{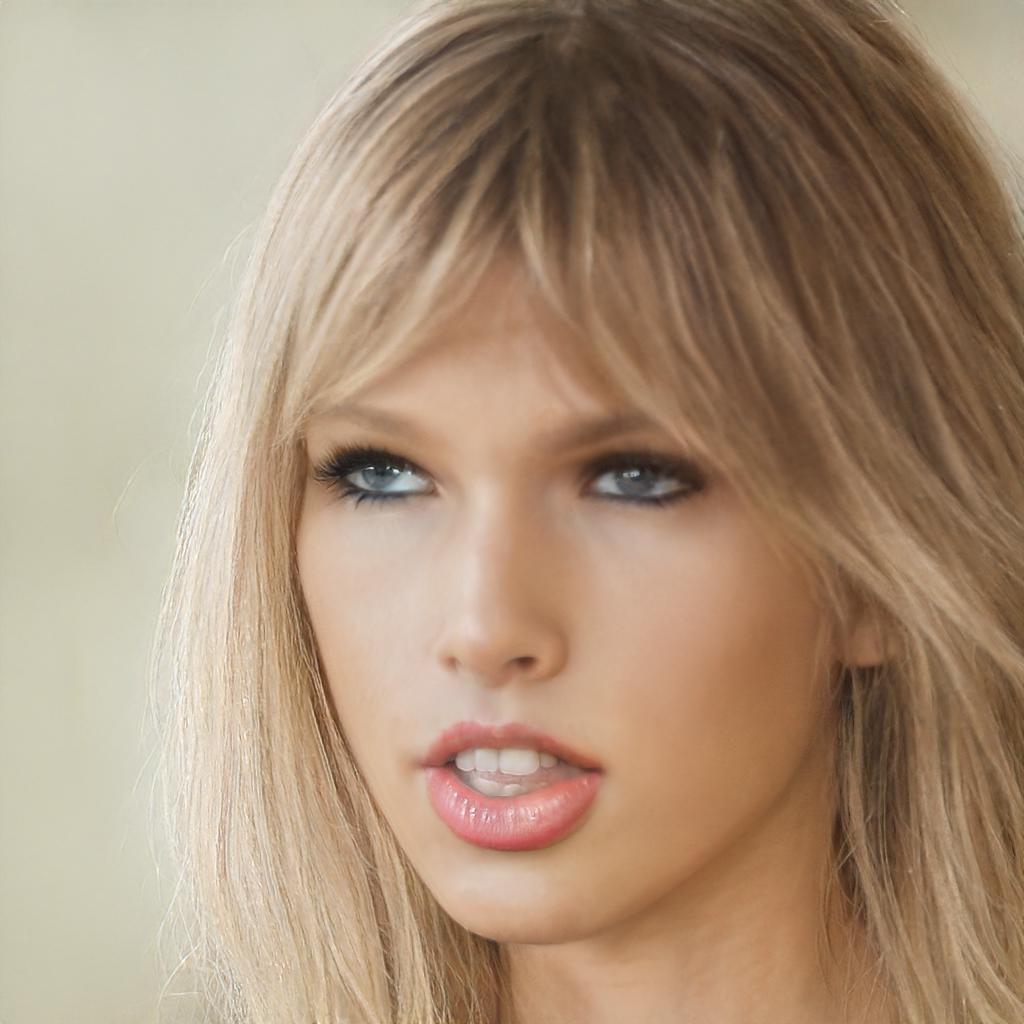} &
 			\includegraphics[width=0.23\linewidth]{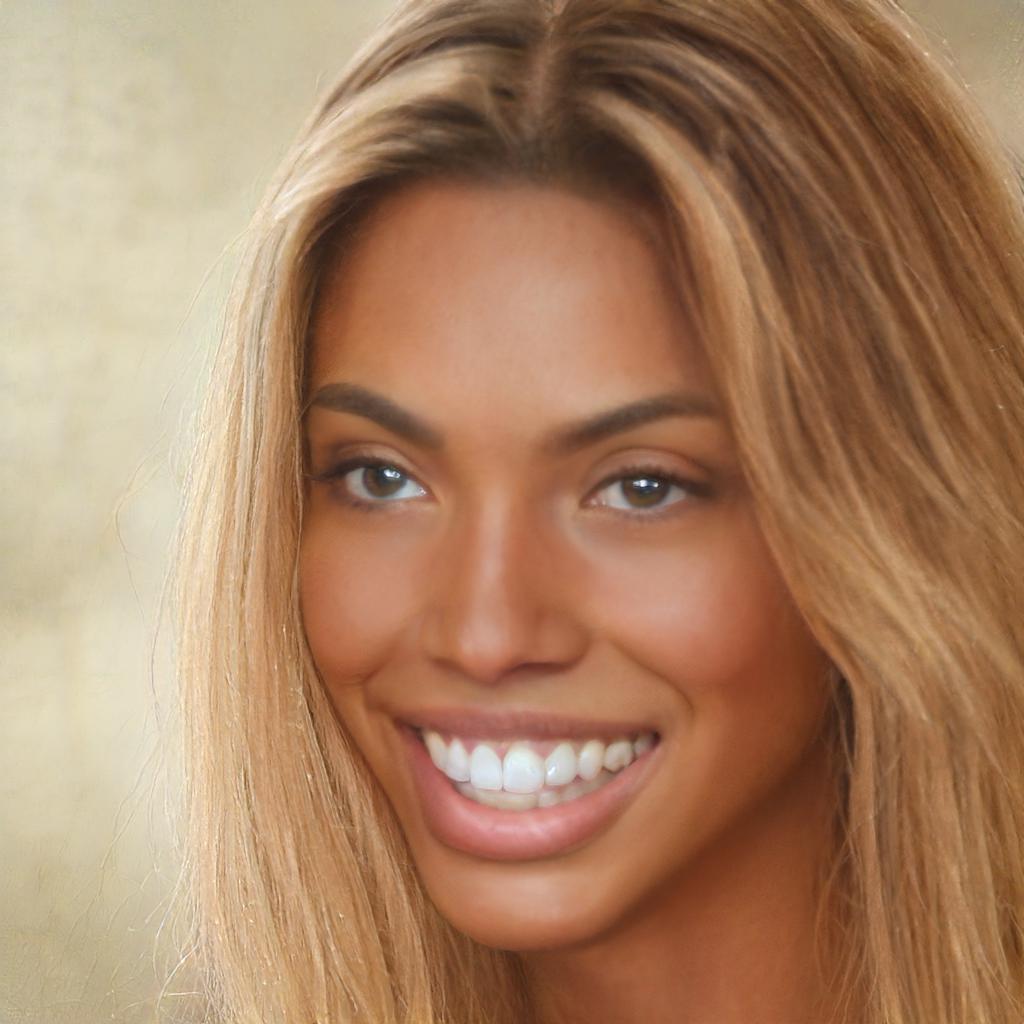} &
 			\includegraphics[width=0.23\linewidth]{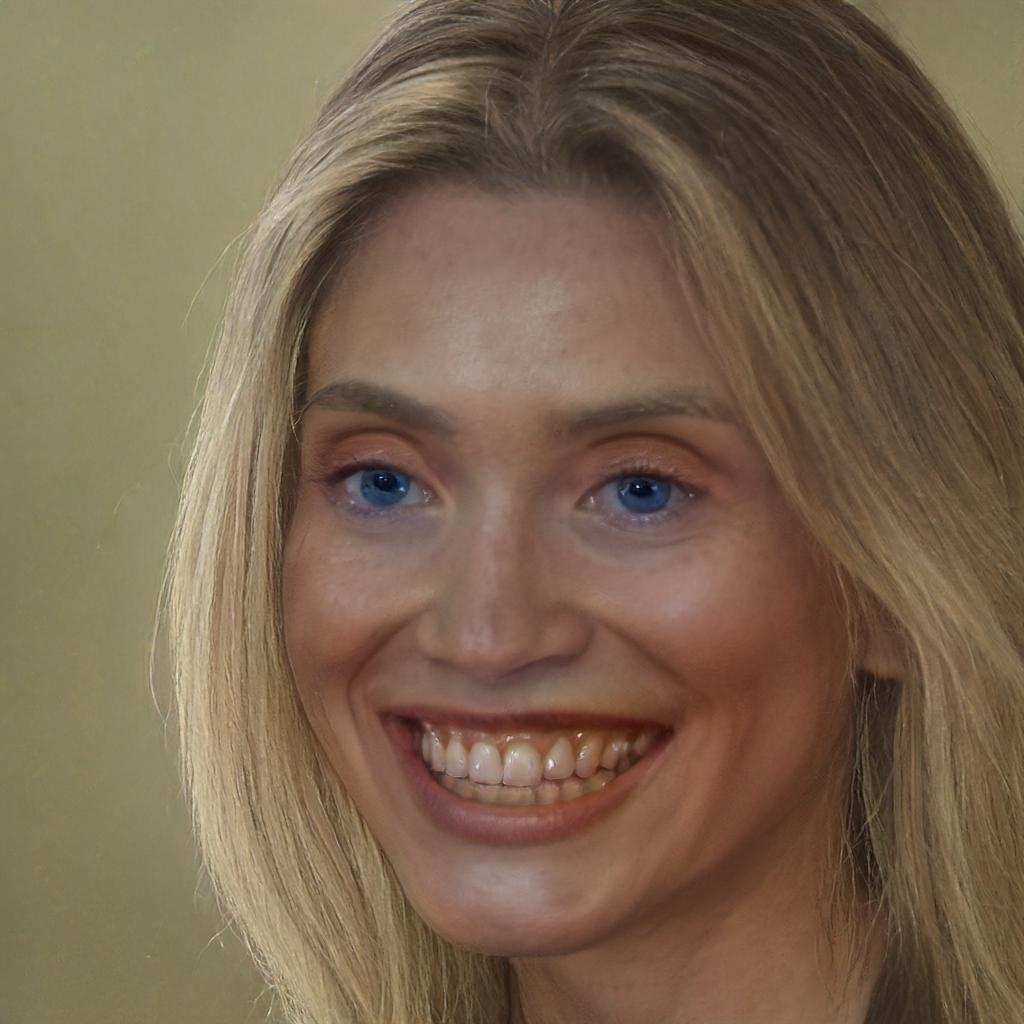}
 			\\

			\includegraphics[width=0.23\linewidth]{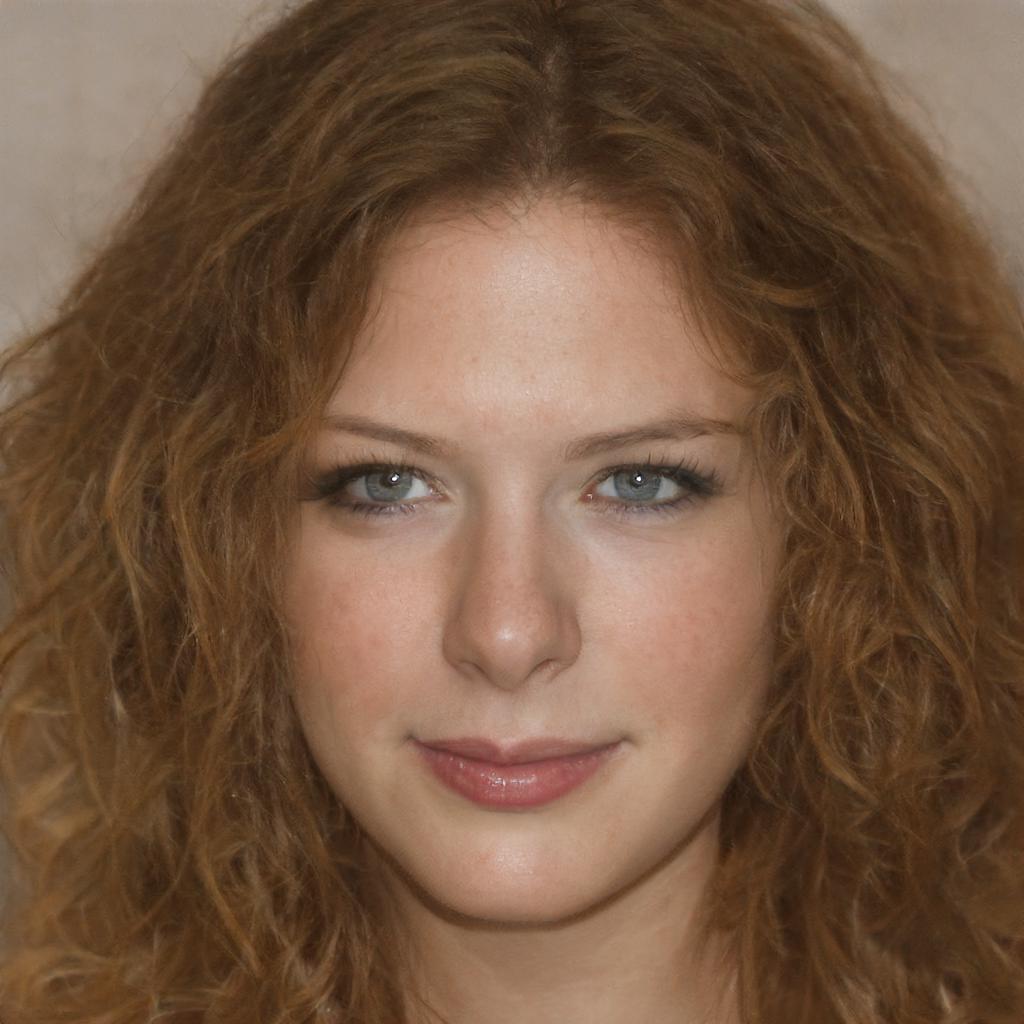} &
 			\includegraphics[width=0.23\linewidth]{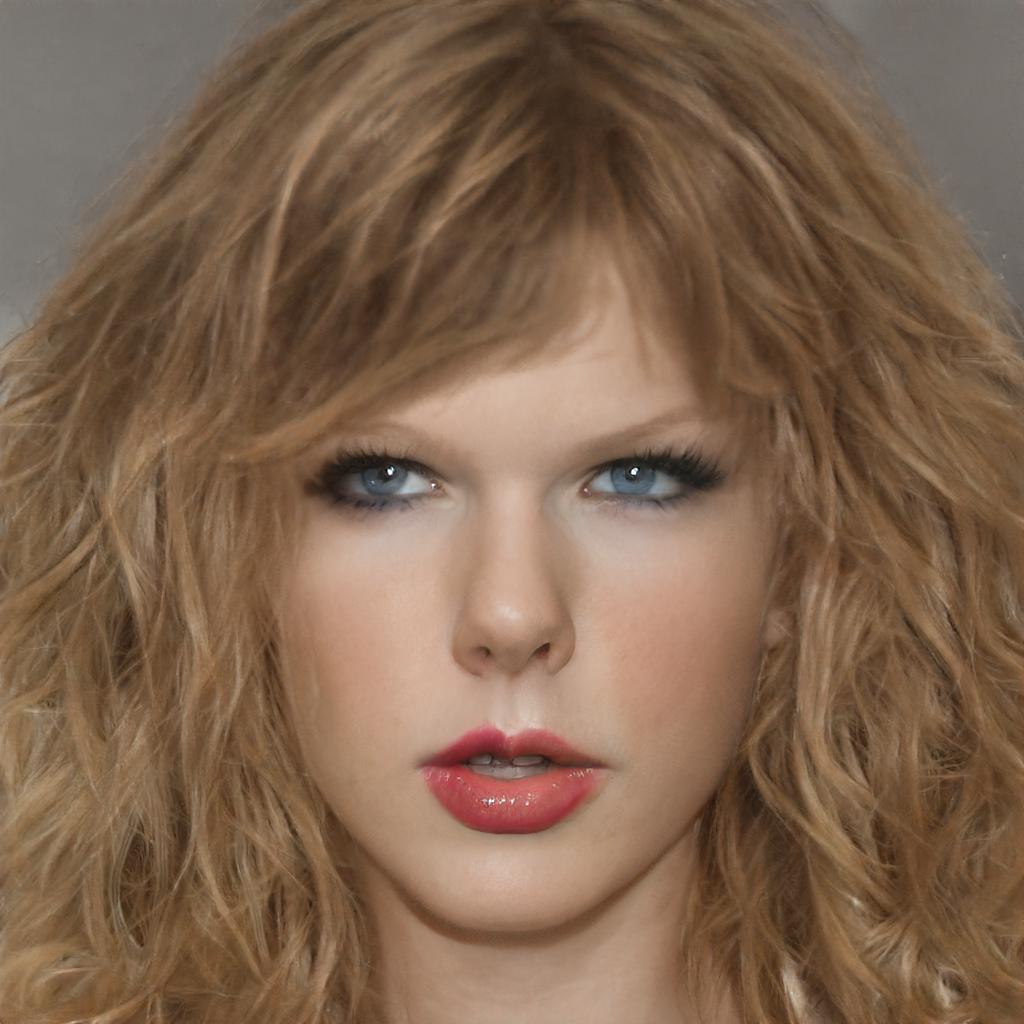} &
 			\includegraphics[width=0.23\linewidth]{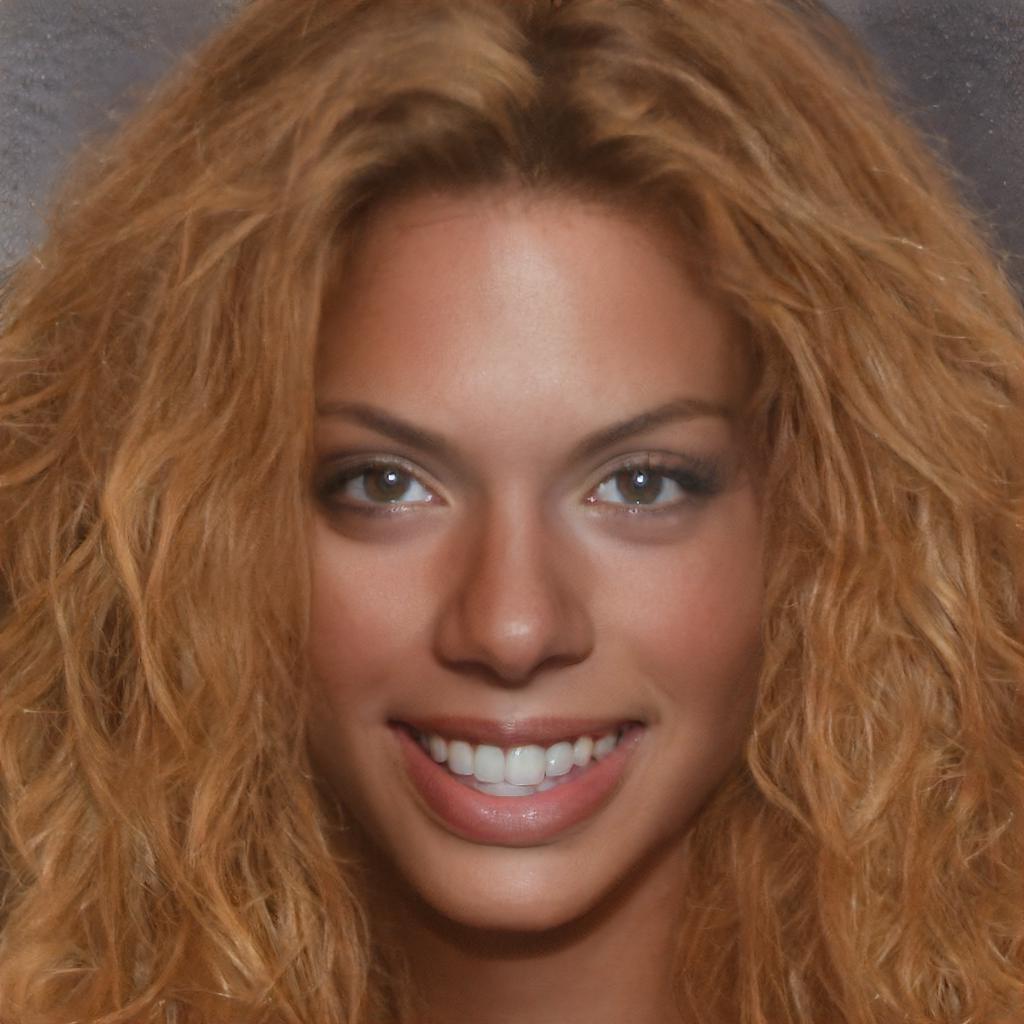} &
 			\includegraphics[width=0.23\linewidth]{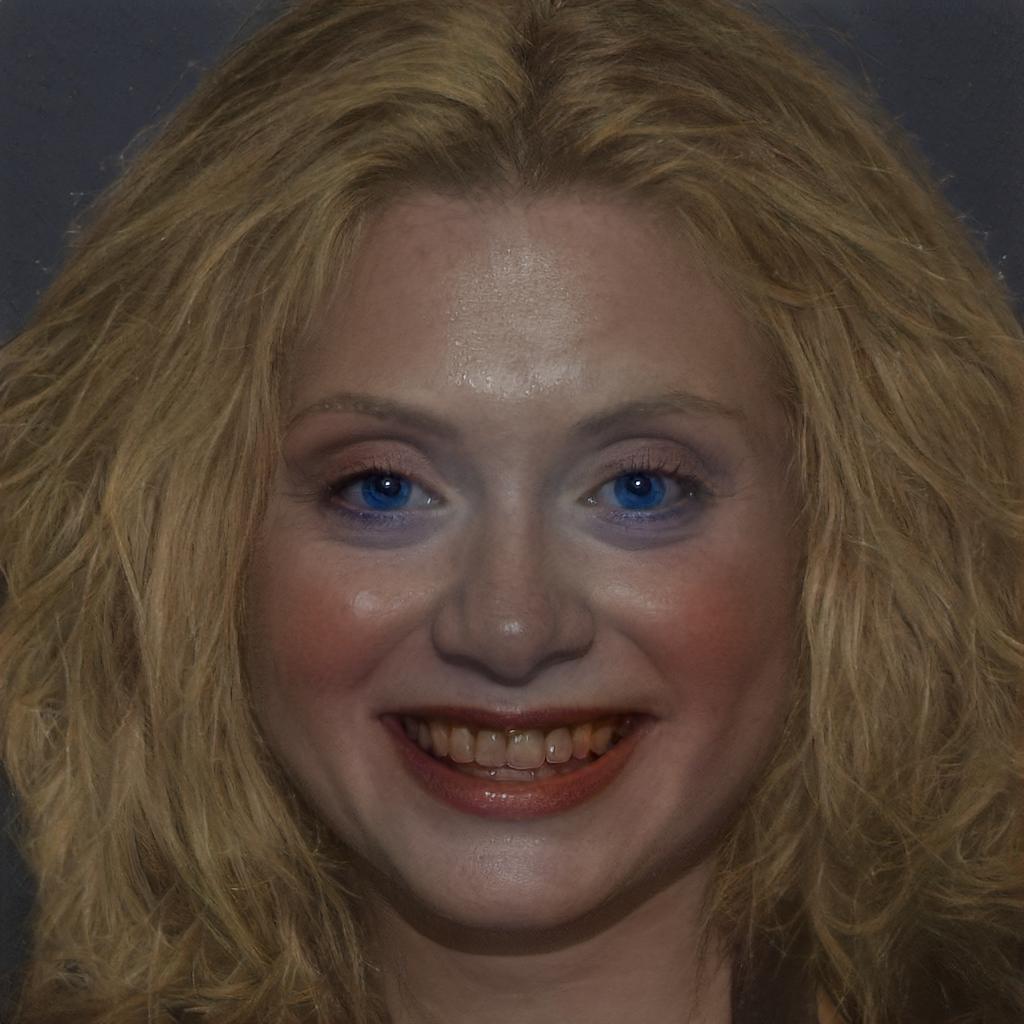}
 			\\

			\includegraphics[width=0.23\linewidth]{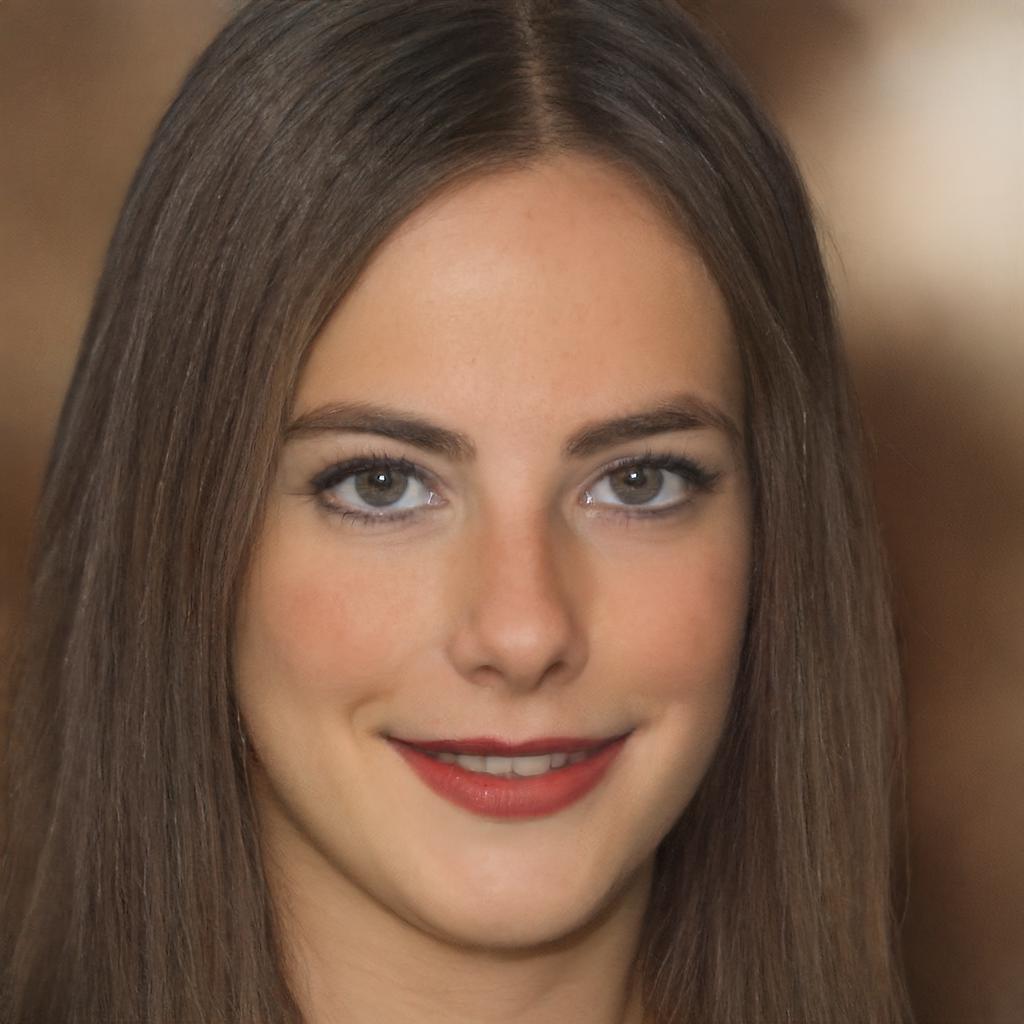} &
 			\includegraphics[width=0.23\linewidth]{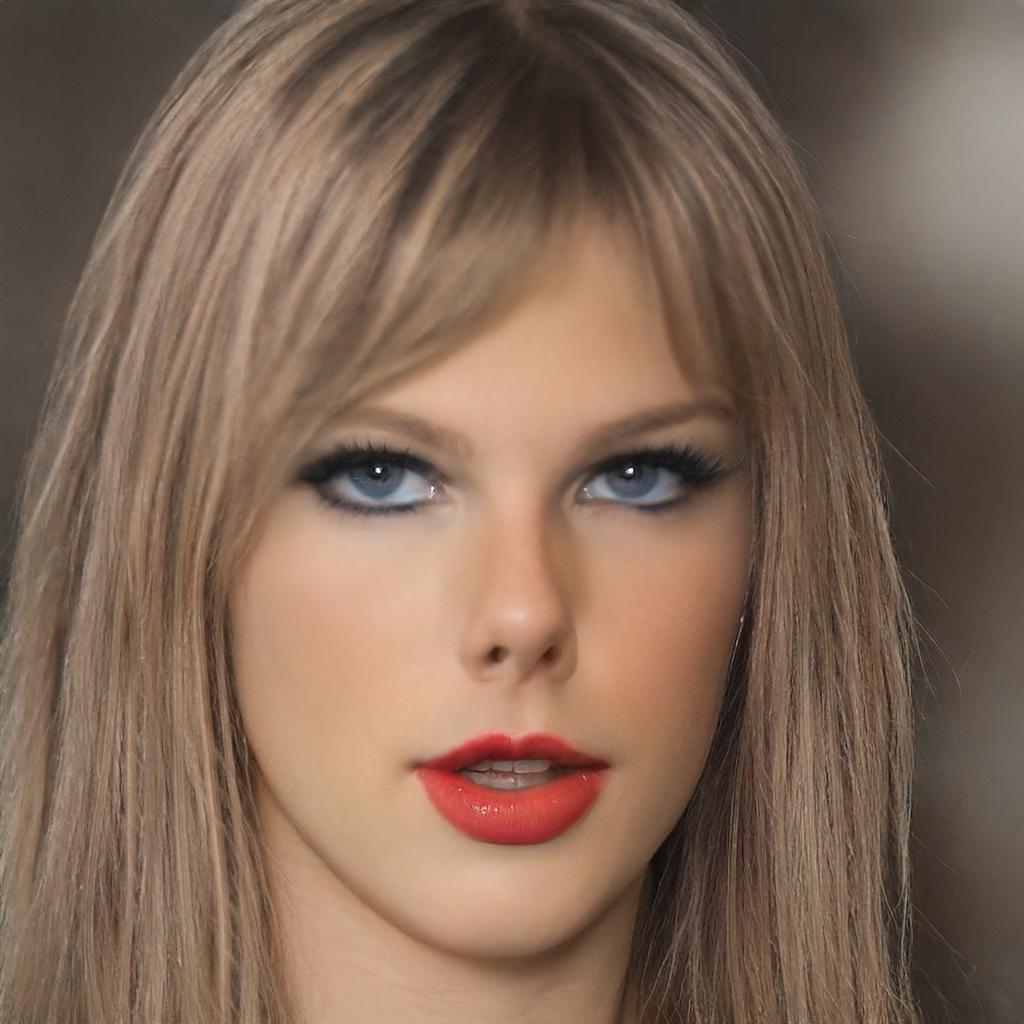} &
 			\includegraphics[width=0.23\linewidth]{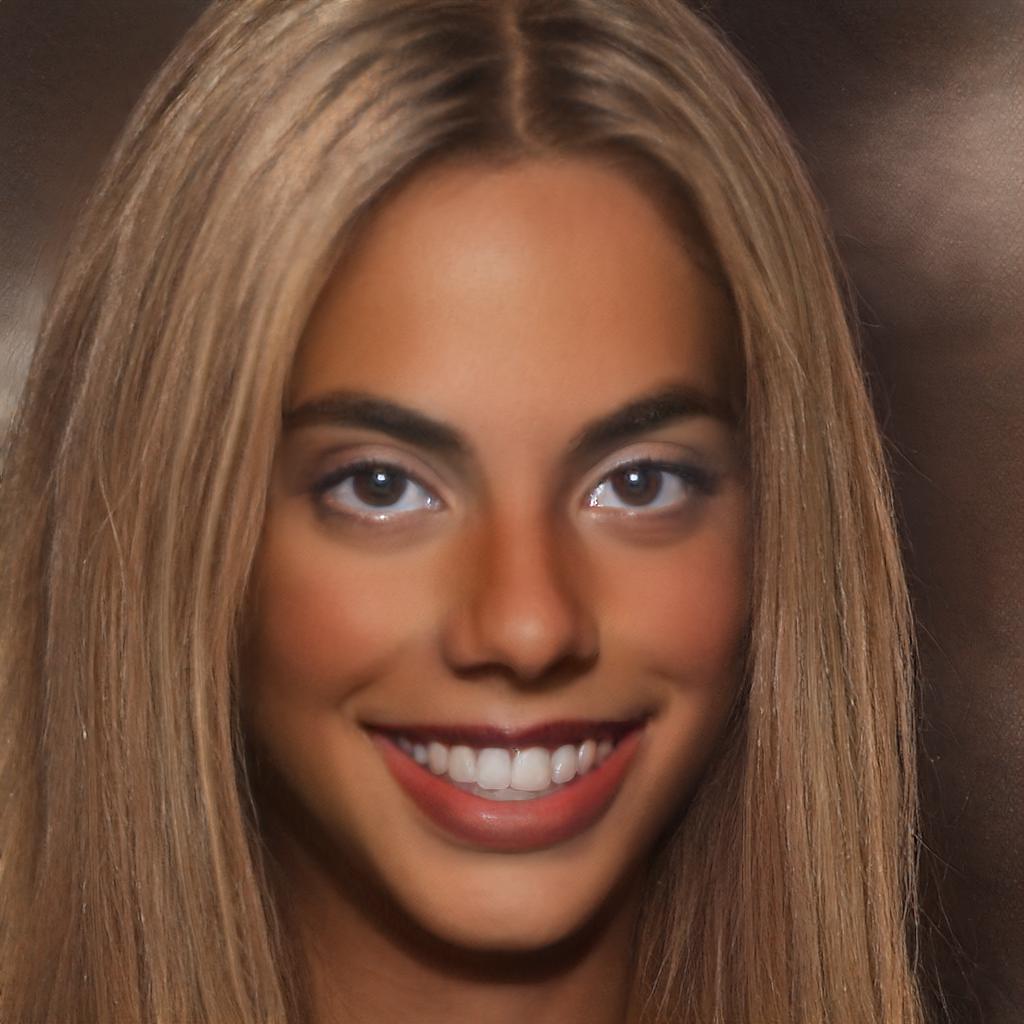} &
 			\includegraphics[width=0.23\linewidth]{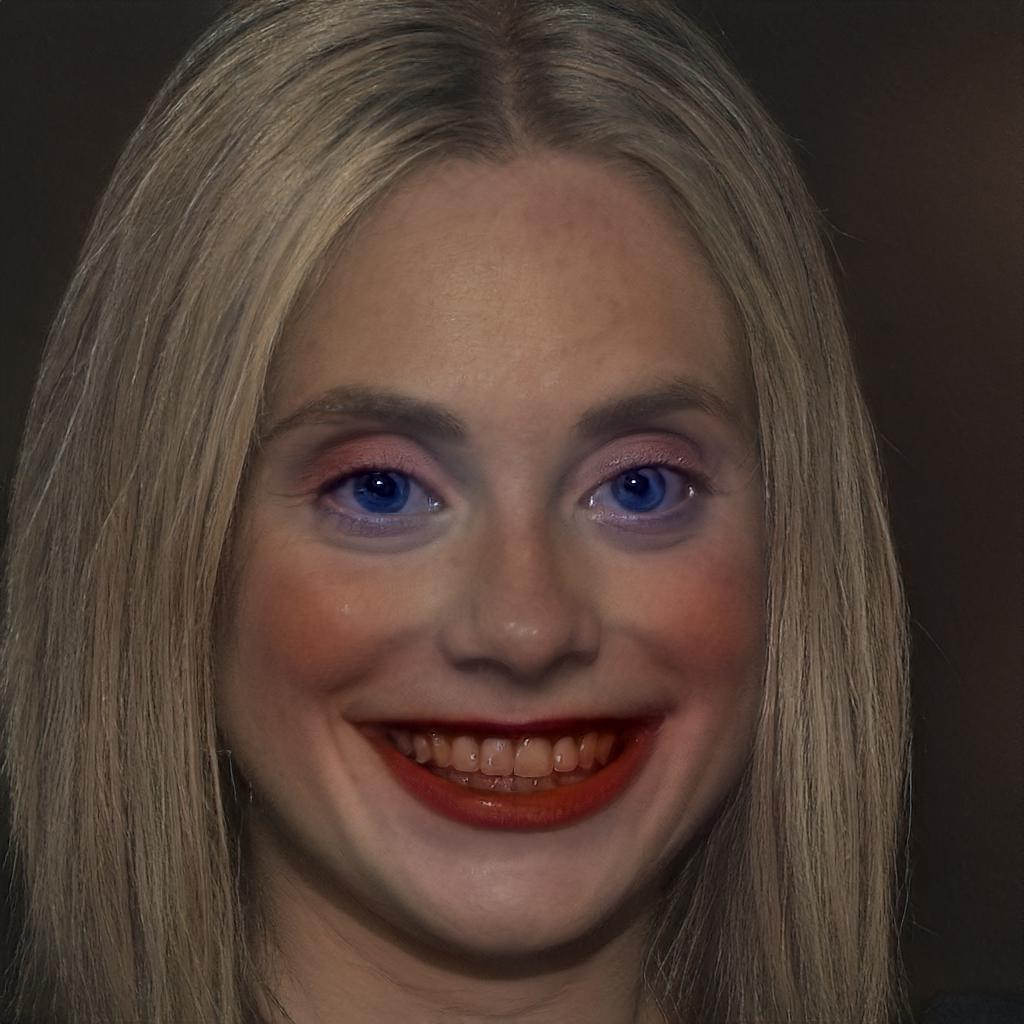}
 			\\
 			Input & Taylor Swift & Beyonce & Hillary Clinton
		\end{tabular}
	}
	\caption{Celeb edits performed by the latent mapper.}
	\label{fig:supp-women}
\end{figure}

\begin{figure}[p]
	\setlength{\tabcolsep}{1pt}
	\centering
	{\footnotesize
		\begin{tabular}{c c c c c}
			\includegraphics[width=0.23\linewidth]{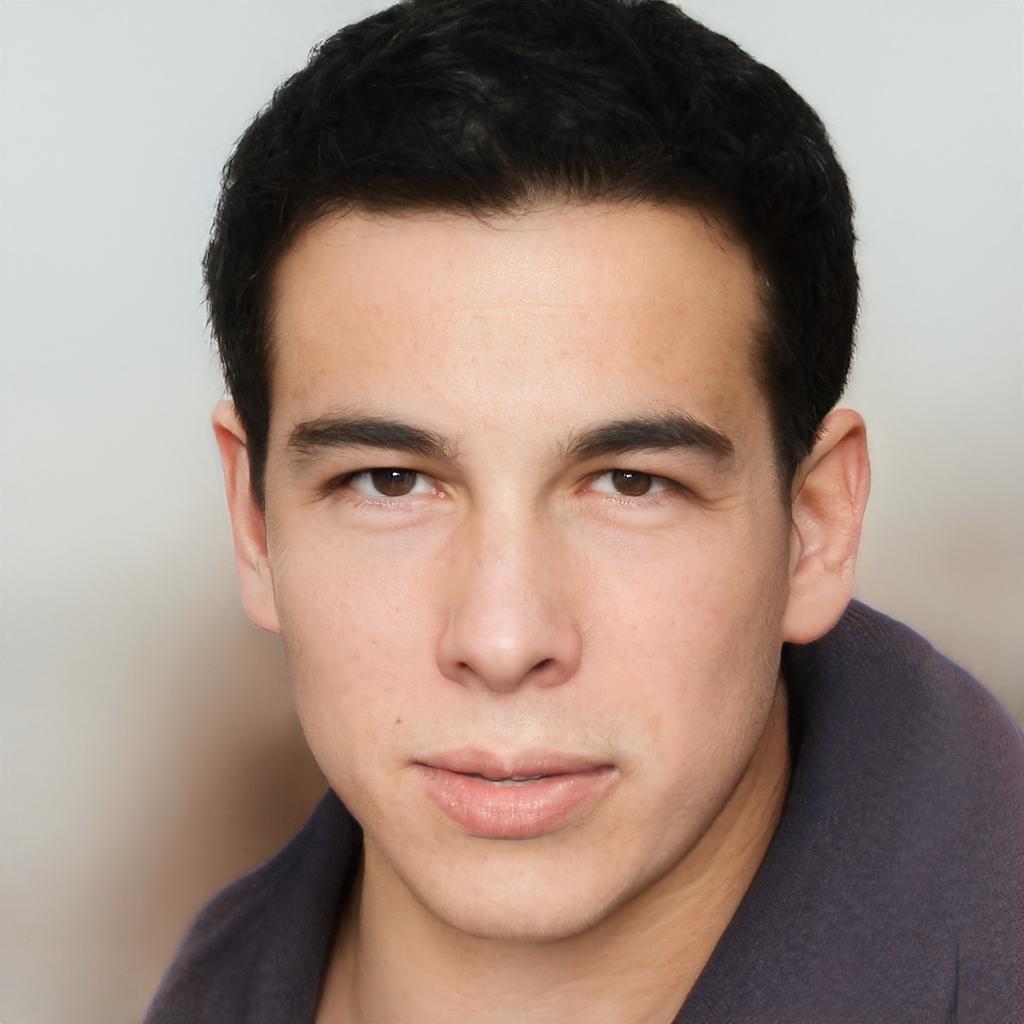} &
 			\includegraphics[width=0.23\linewidth]{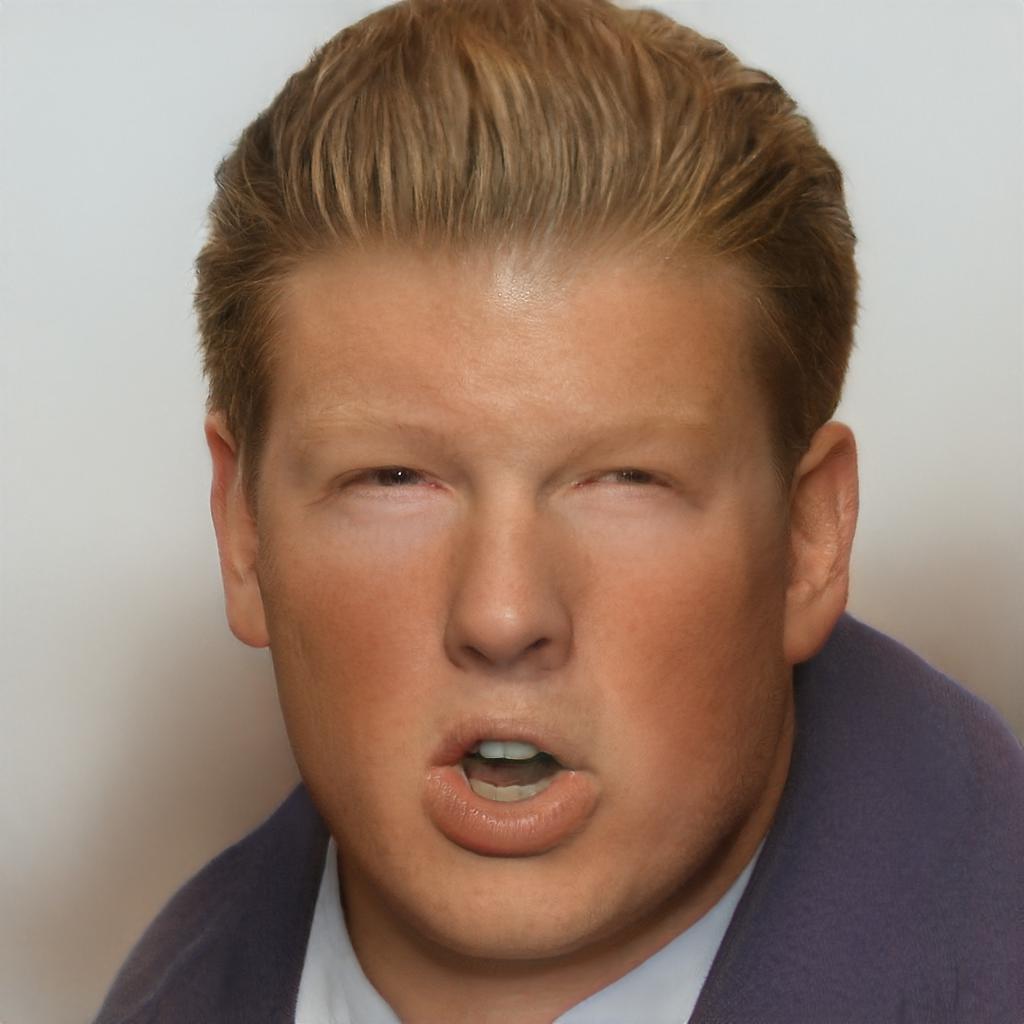} &
 			 \includegraphics[width=0.23\linewidth]{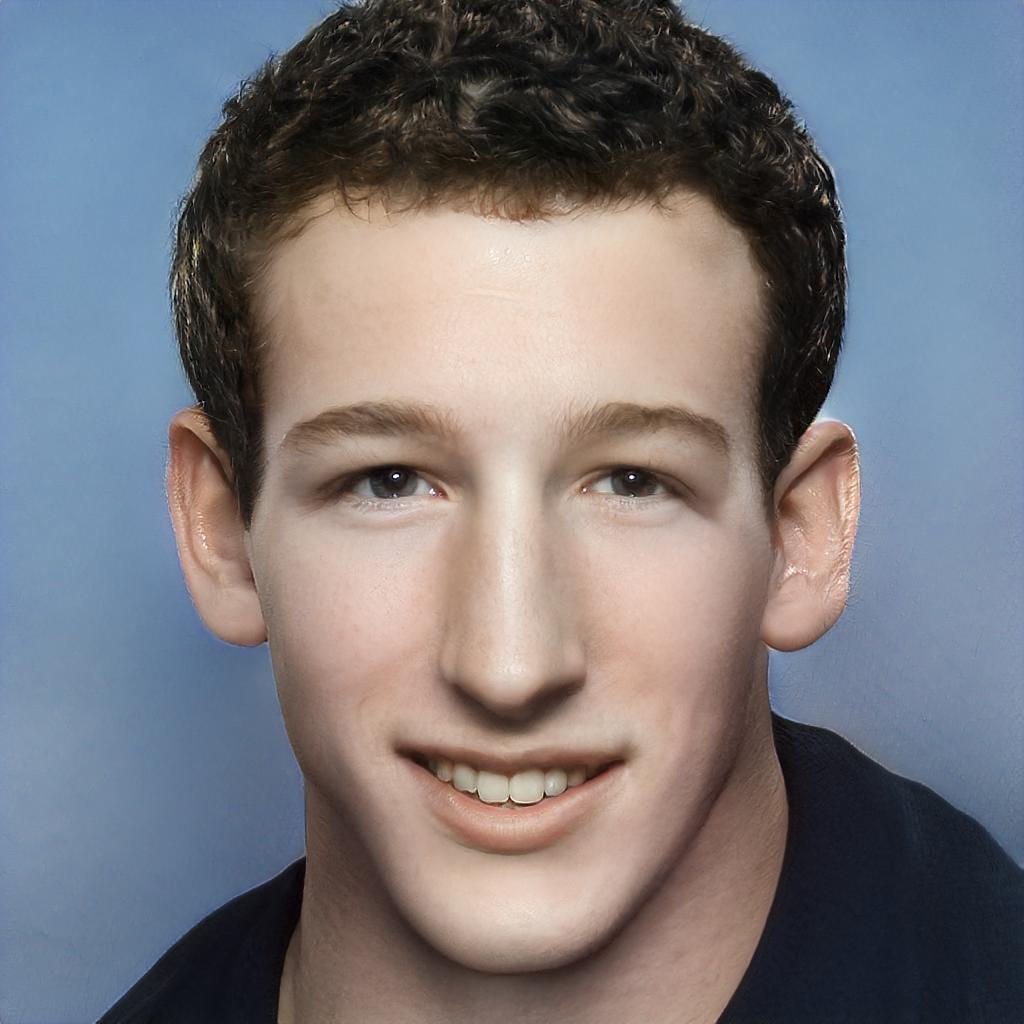} & 		\includegraphics[width=0.23\linewidth]{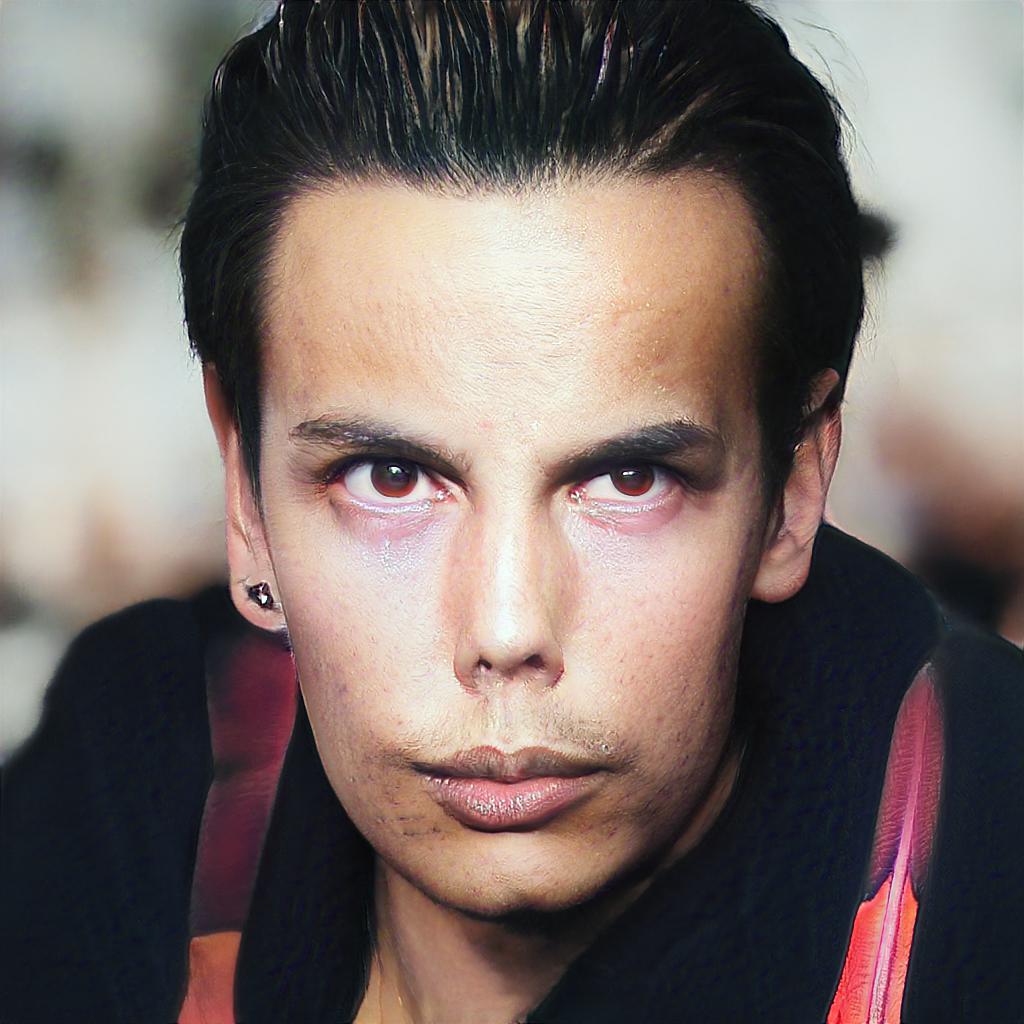}
 			\\
 			
 			\includegraphics[width=0.23\linewidth]{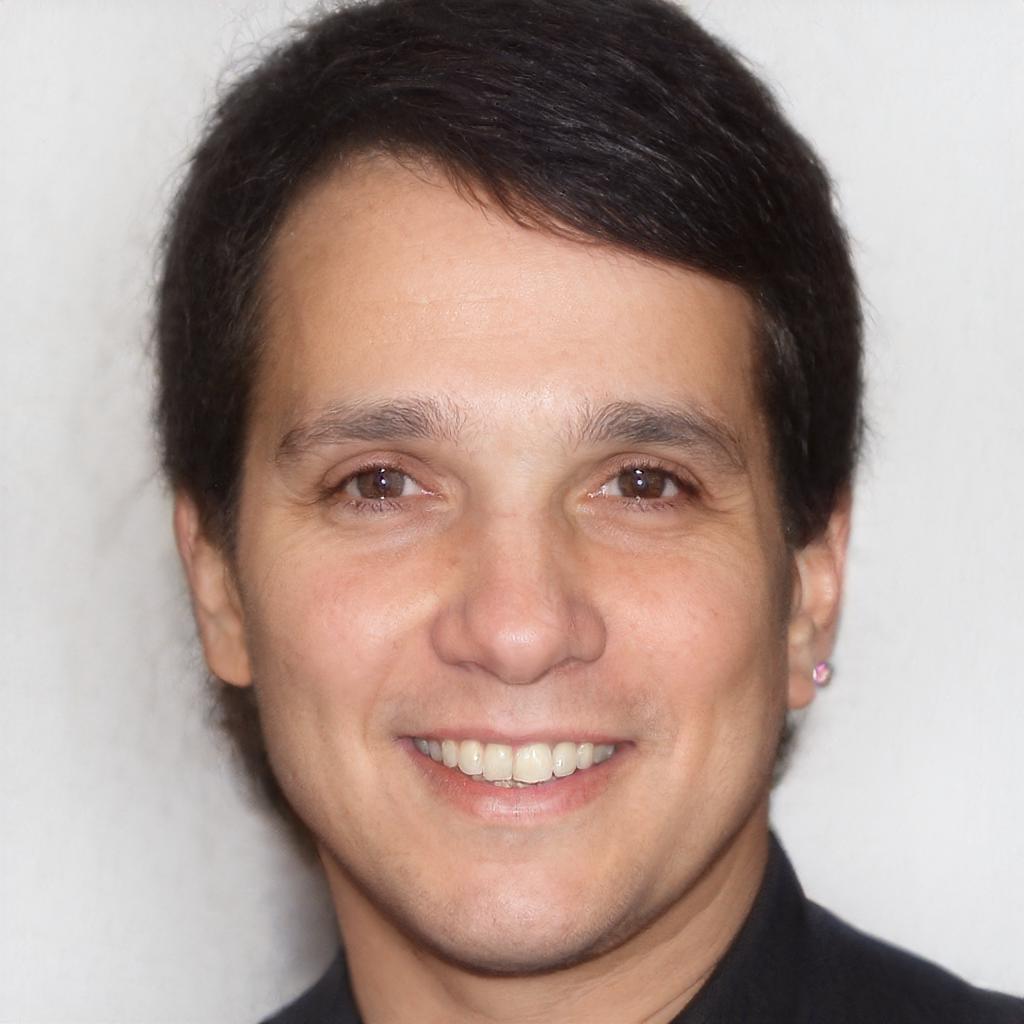} &
 			\includegraphics[width=0.23\linewidth]{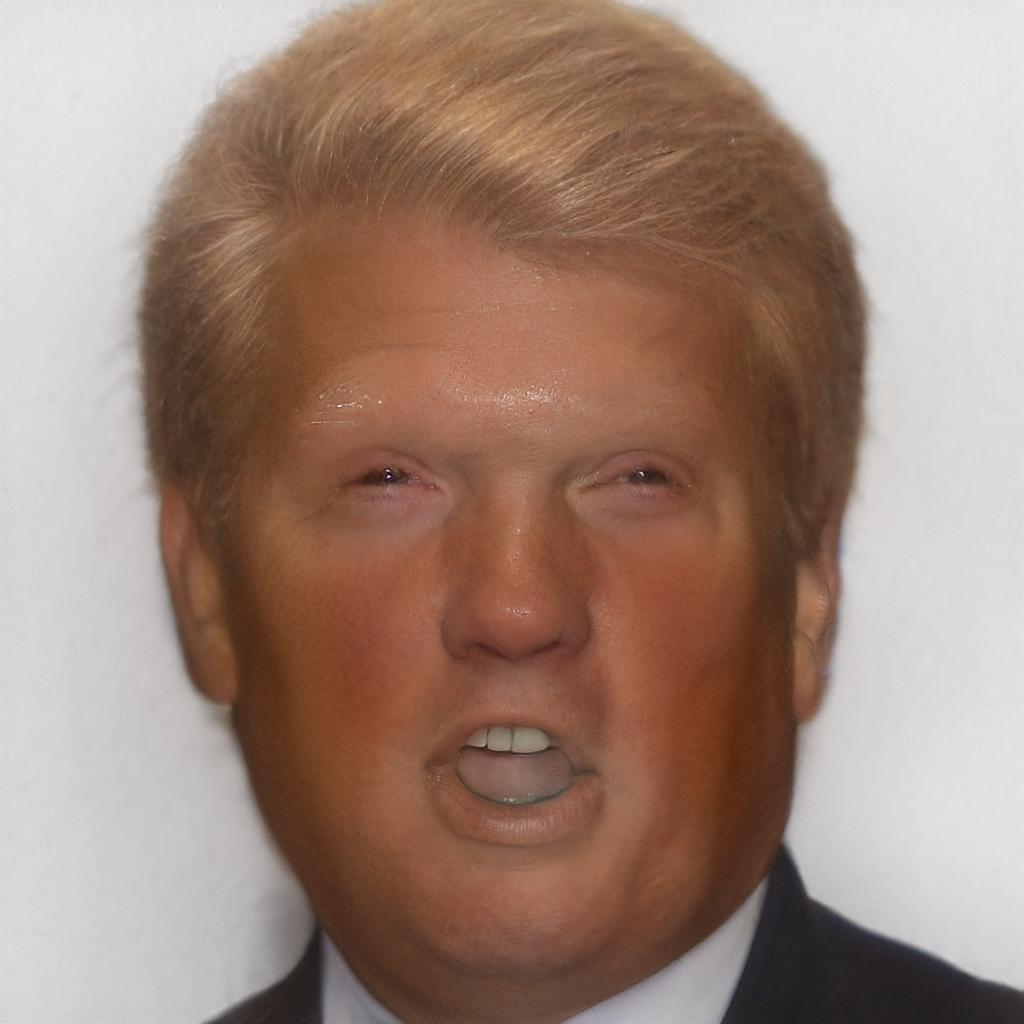} &
 			\includegraphics[width=0.23\linewidth]{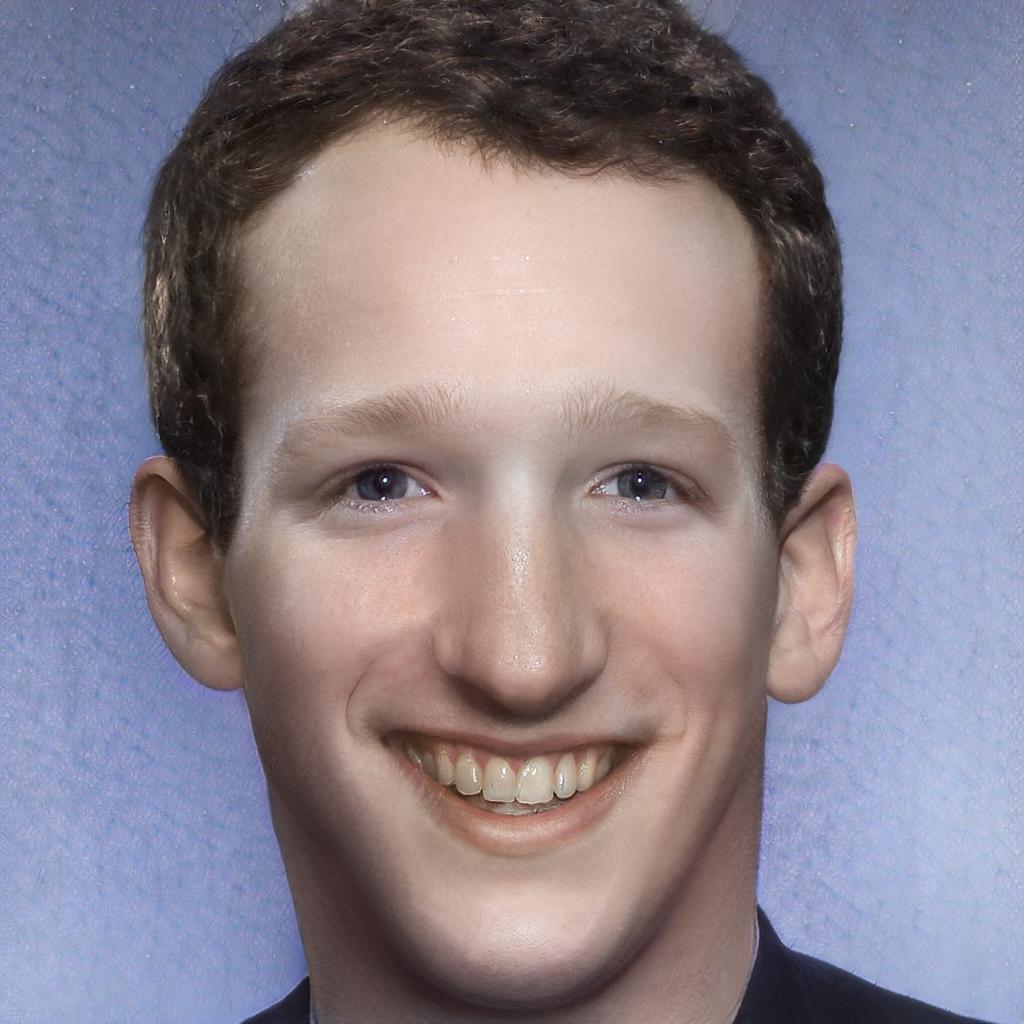} &
 			\includegraphics[width=0.23\linewidth]{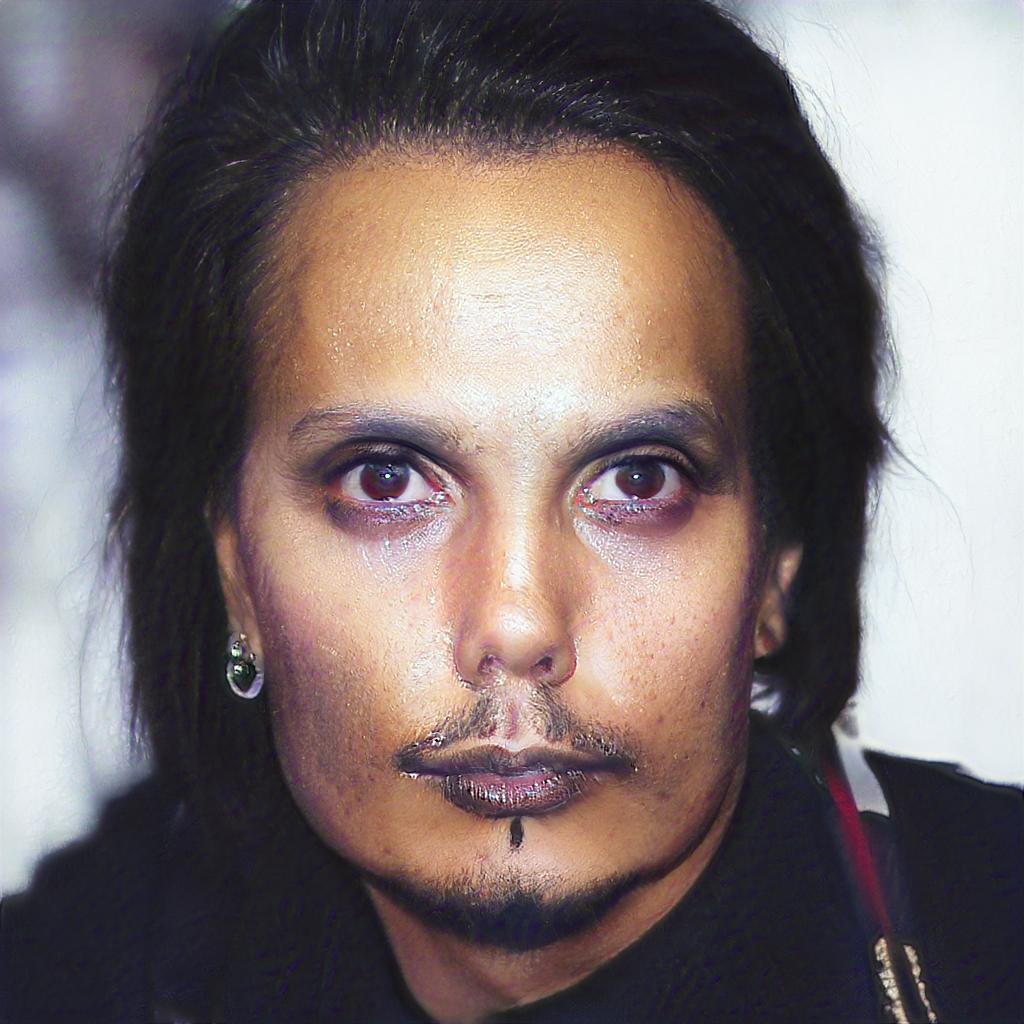}
 			\\

			\includegraphics[width=0.23\linewidth]{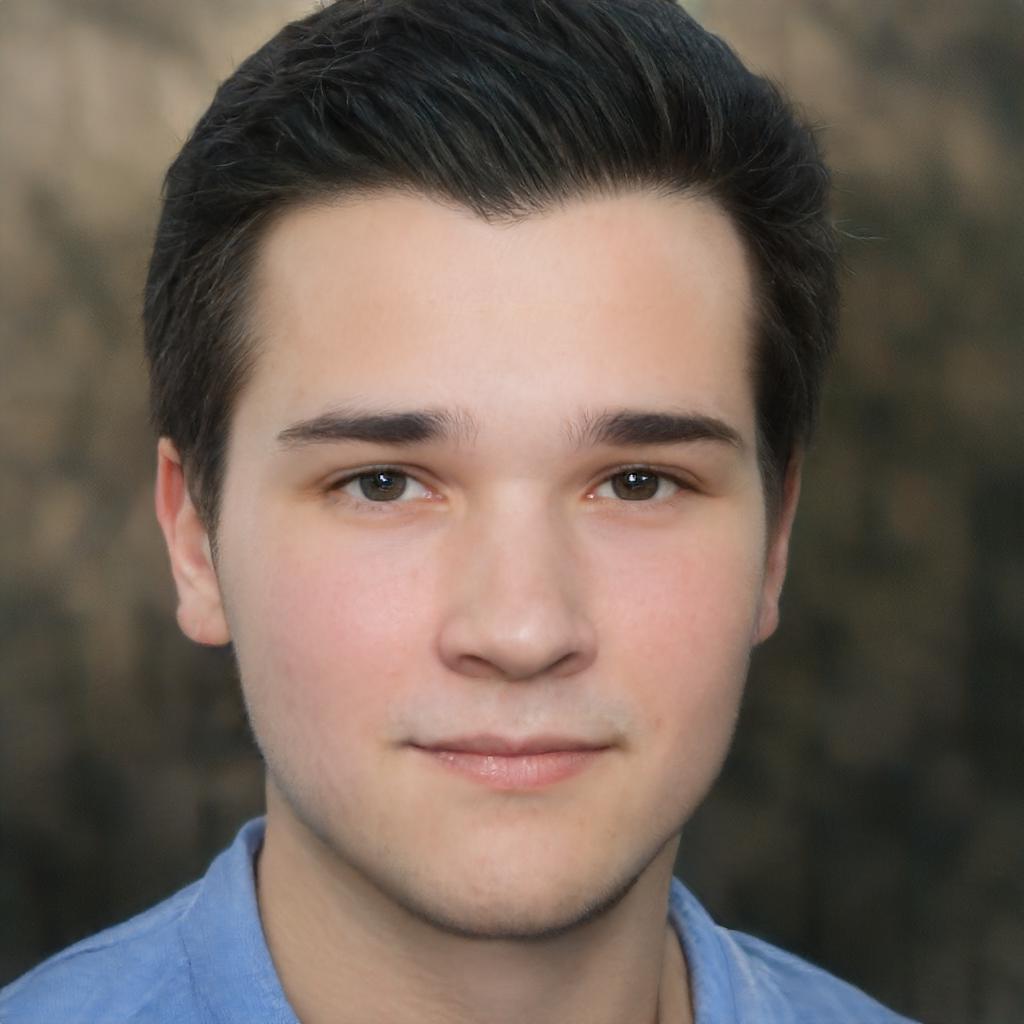} &
 			\includegraphics[width=0.23\linewidth]{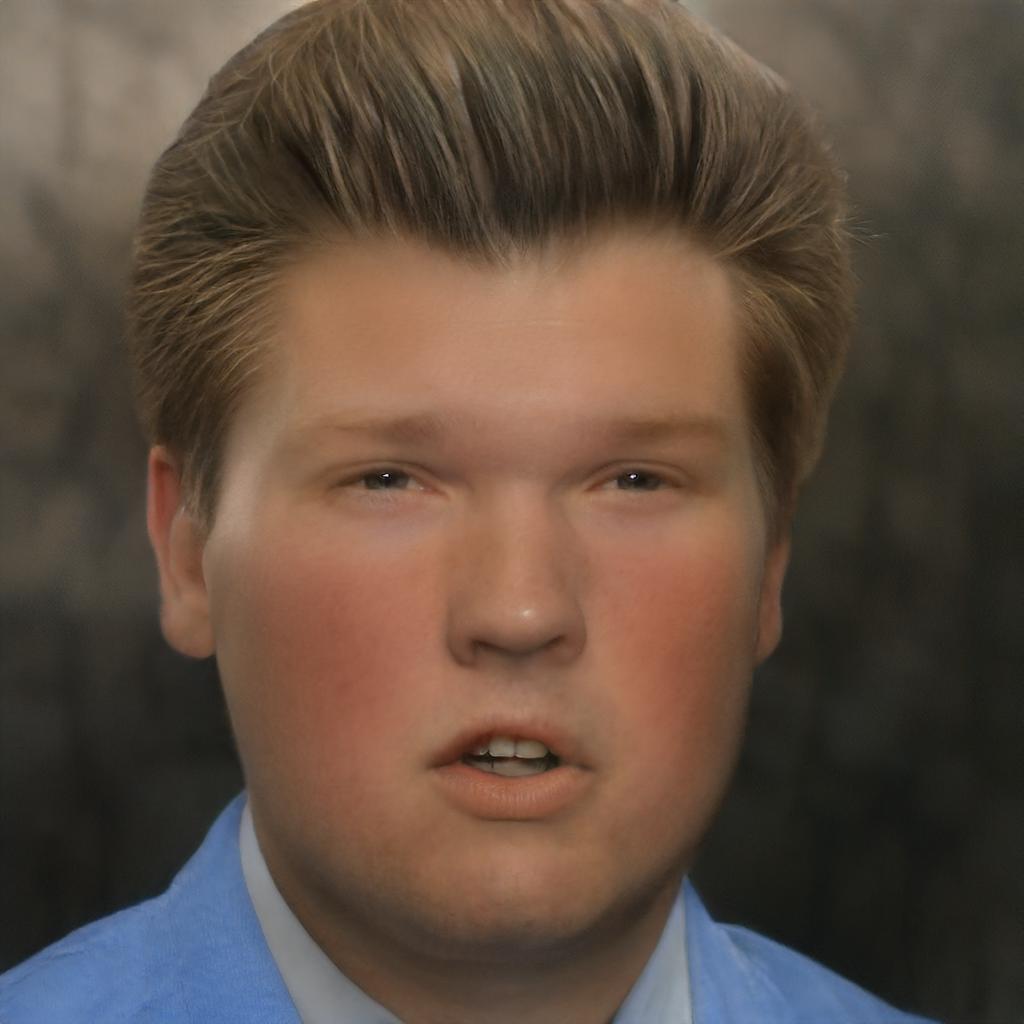} &
 			\includegraphics[width=0.23\linewidth]{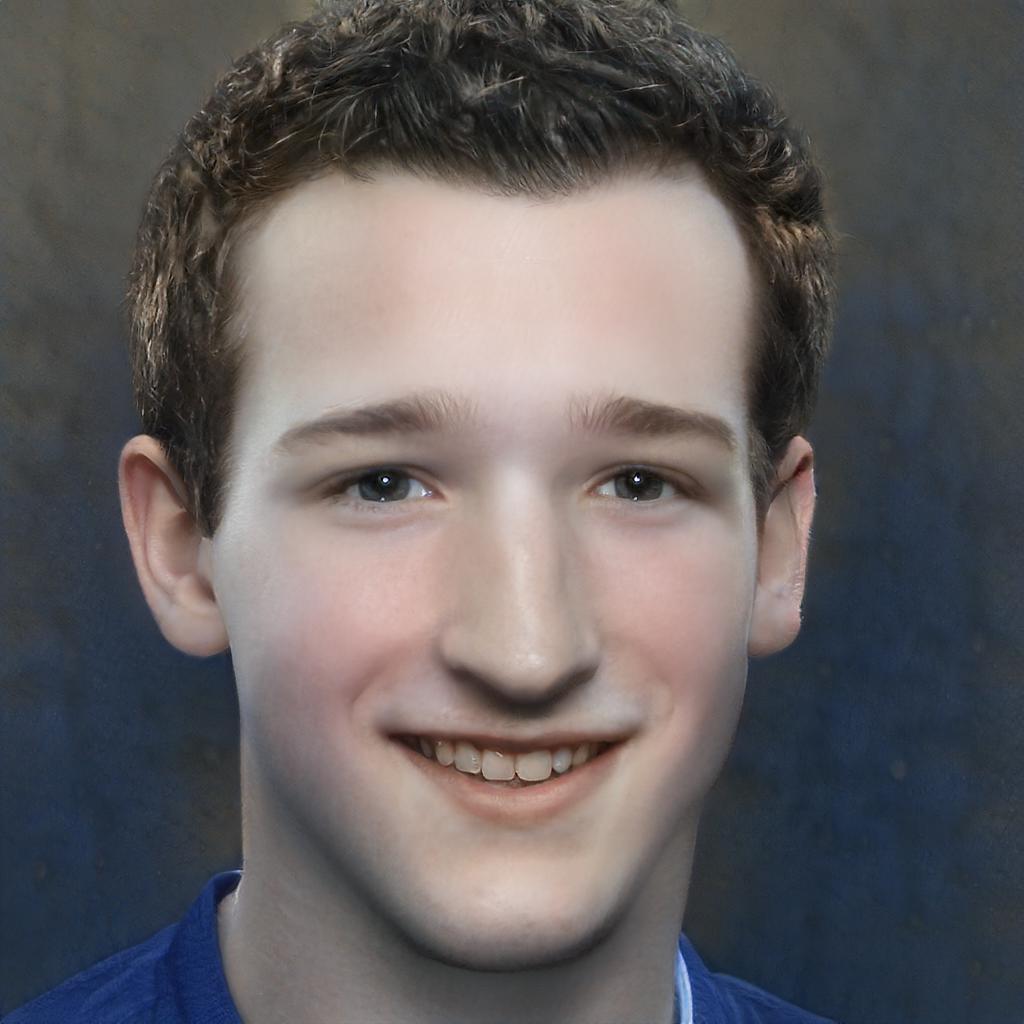} &
 			\includegraphics[width=0.23\linewidth]{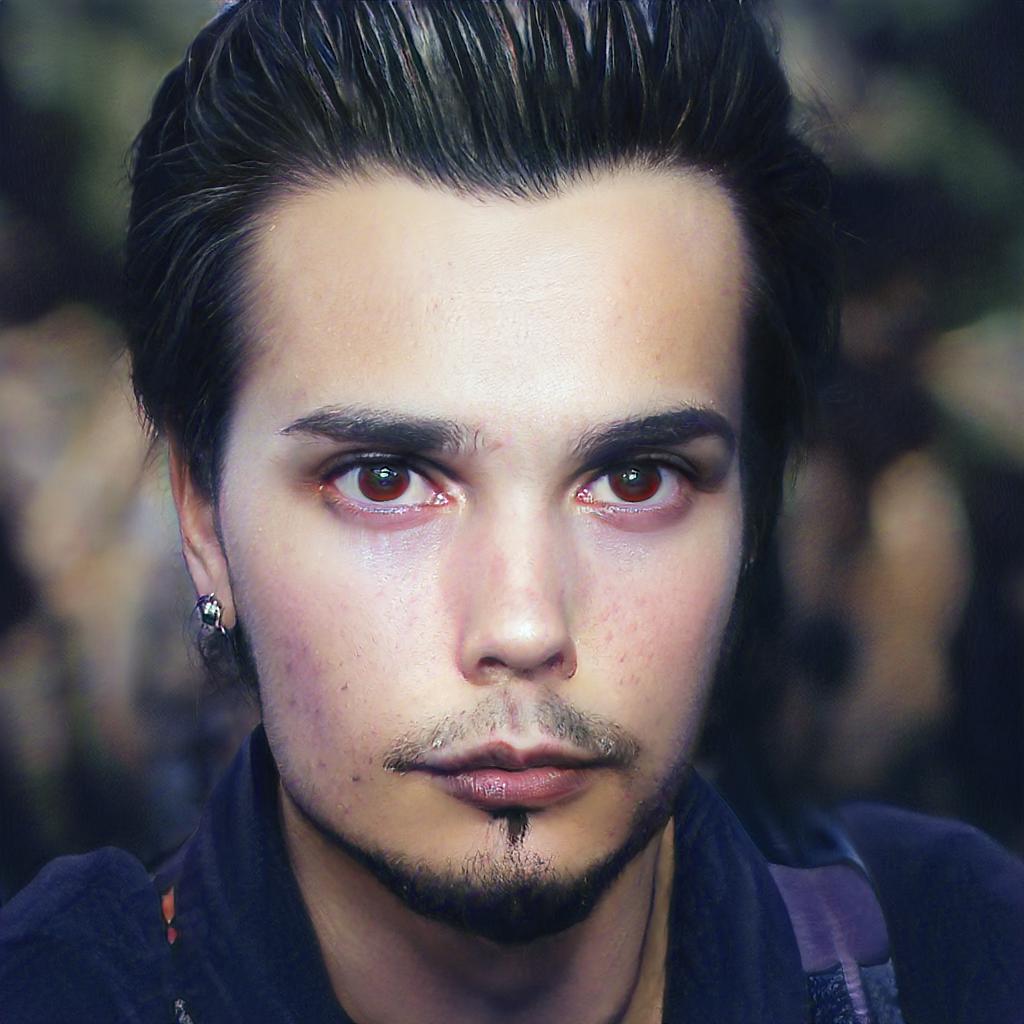}
 			\\
 			Input & Trump & Mark Zuckerberg & Johnny Depp
		\end{tabular}
	}
	\caption{Celeb edits performed by the latent mapper.}
	\label{fig:supp-men}
\end{figure}

\begin{figure}[p]
	\setlength{\tabcolsep}{1pt}
	\centering
	{\footnotesize
		\begin{tabular}{c c c}
			\includegraphics[width=0.32\linewidth]{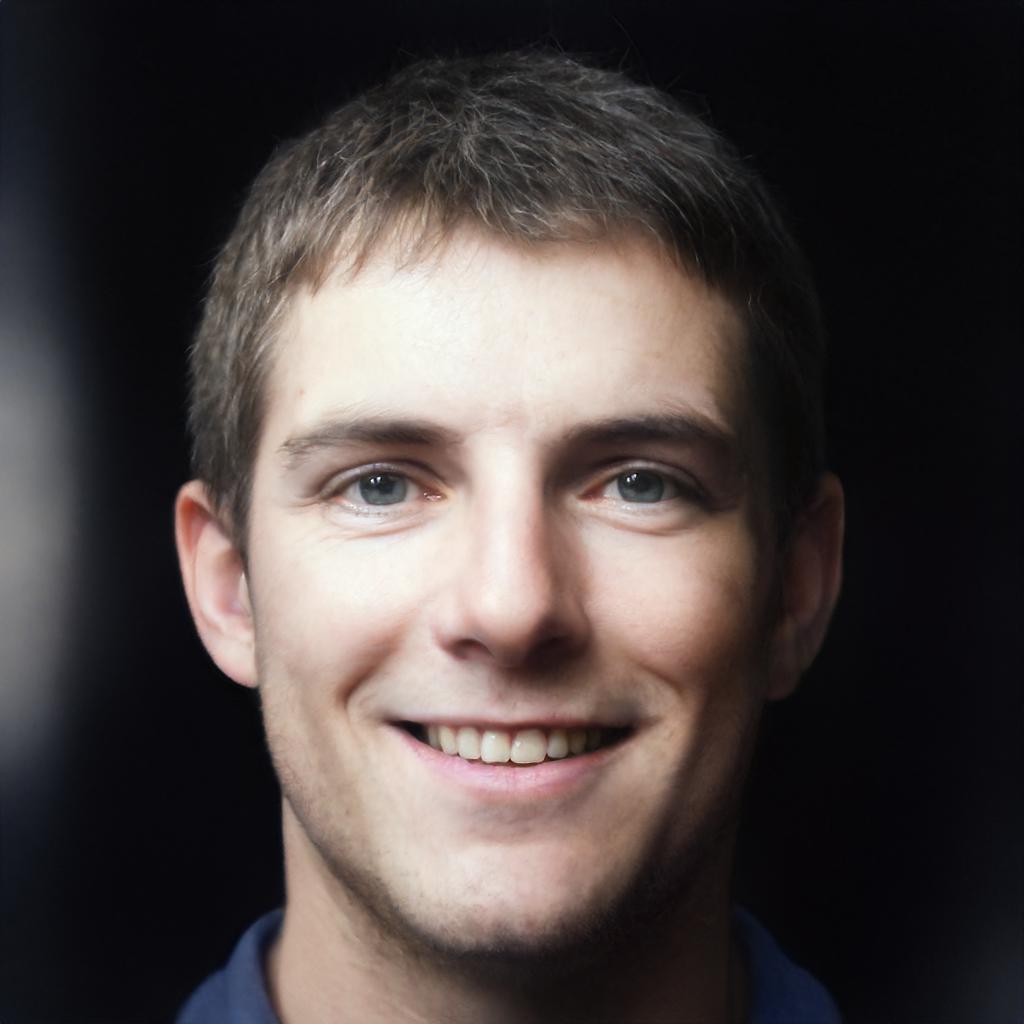} &
 			\includegraphics[width=0.32\linewidth]{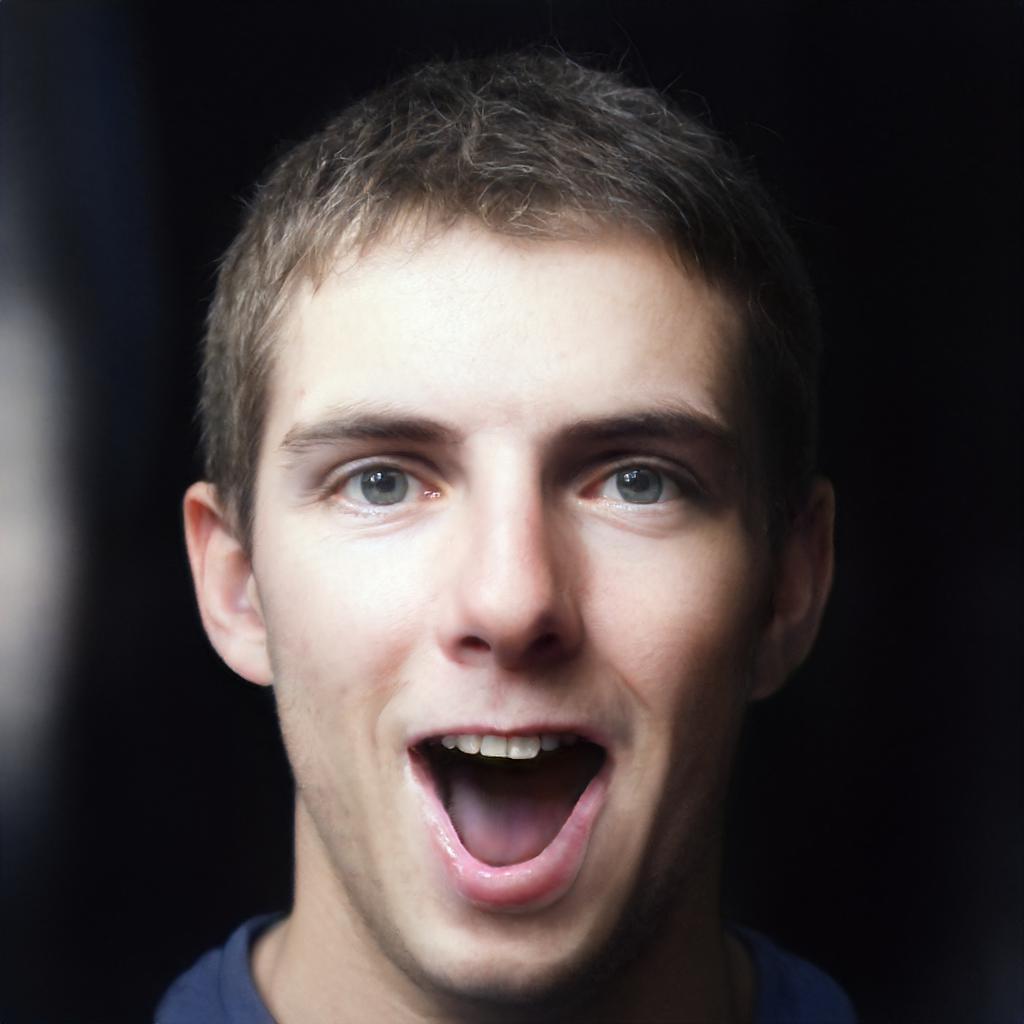} &
 			\includegraphics[width=0.32\linewidth]{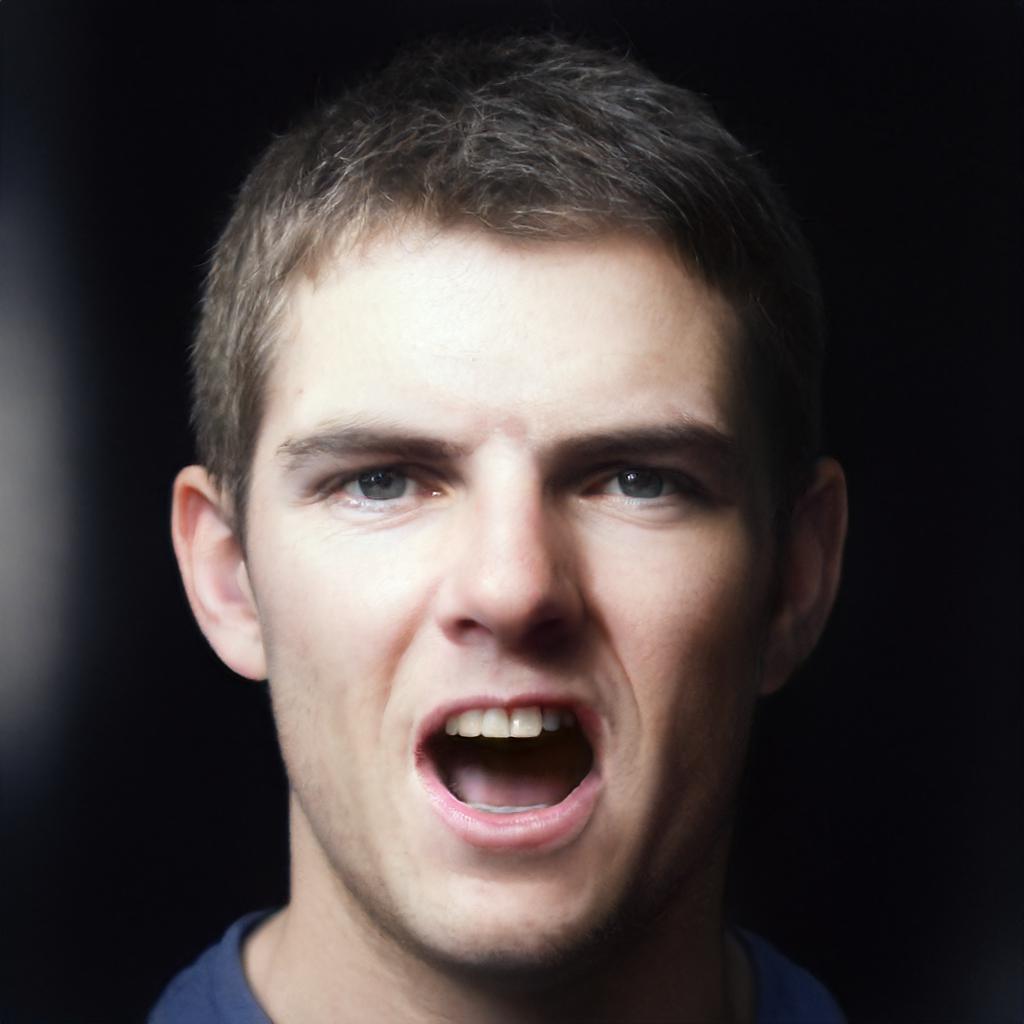}\\

            \includegraphics[width=0.32\linewidth]{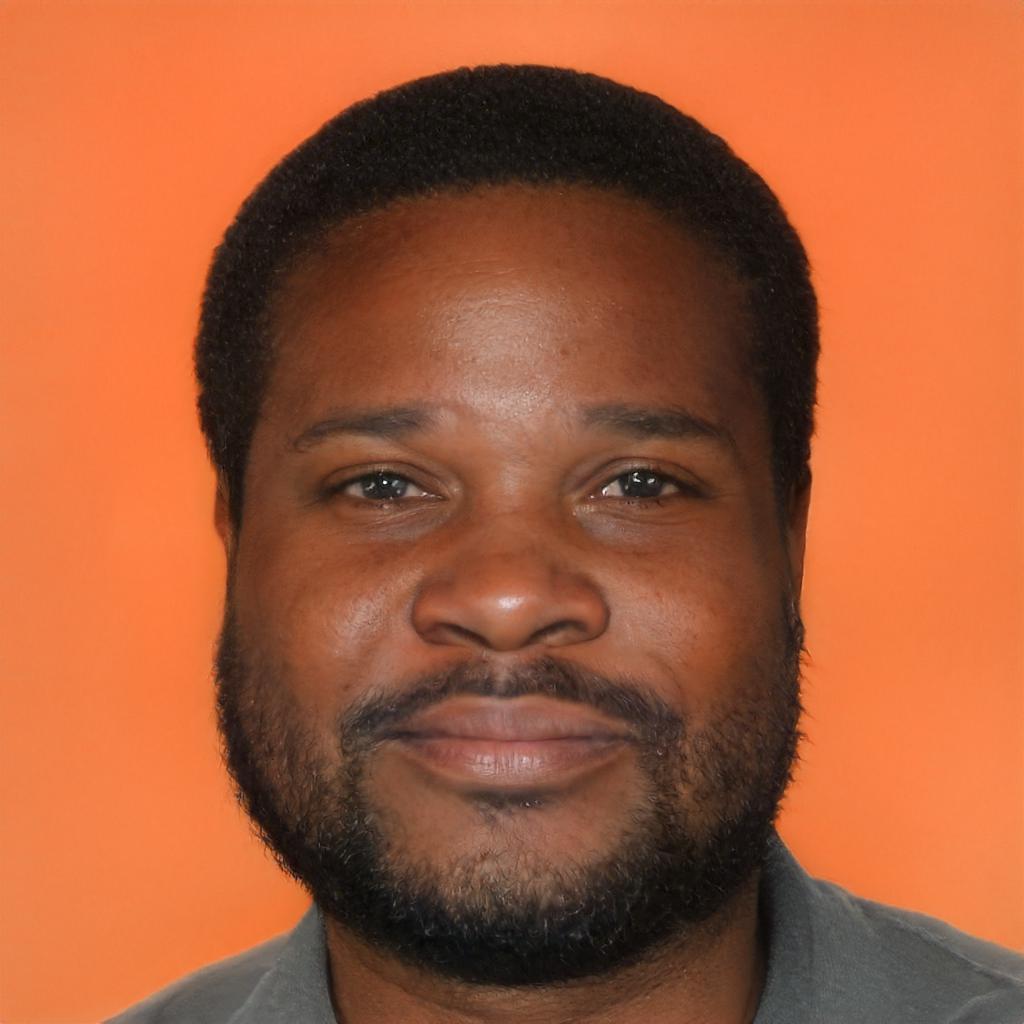} &
 			\includegraphics[width=0.32\linewidth]{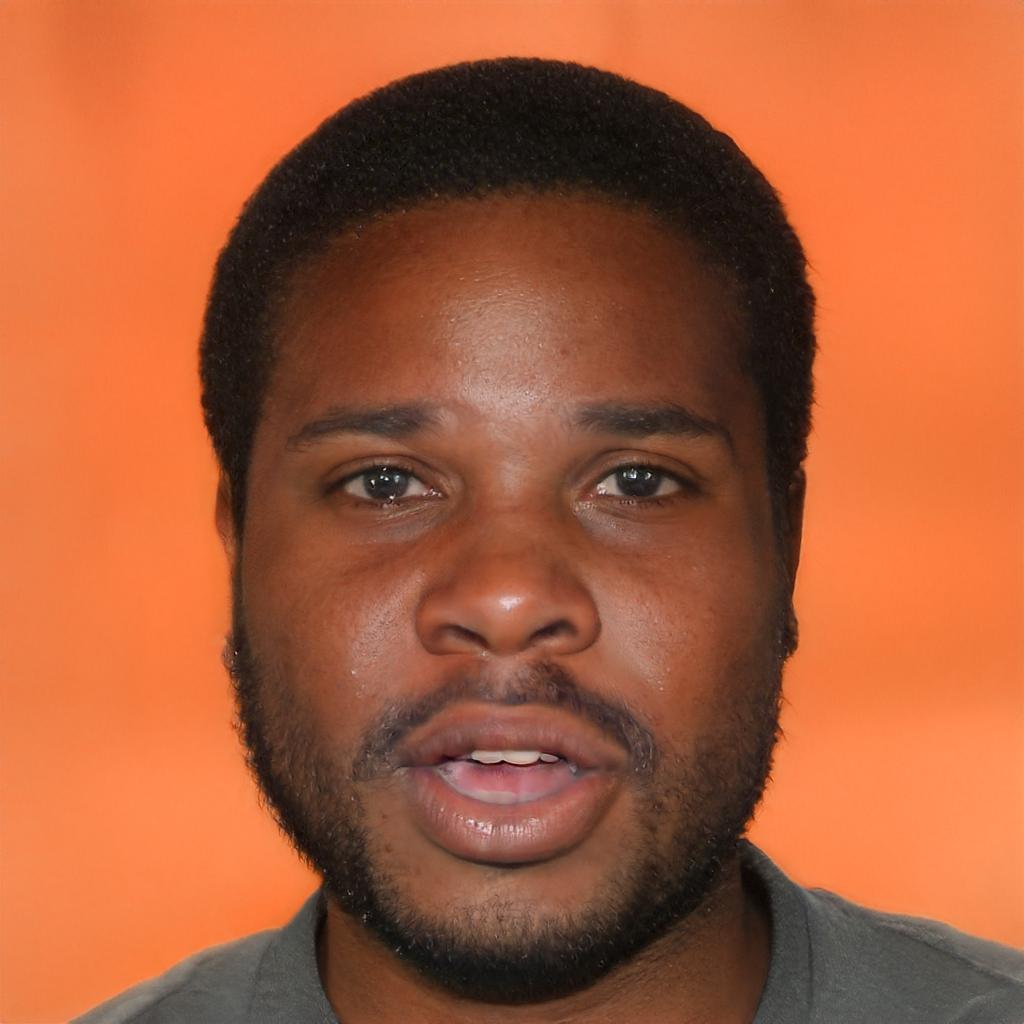} &
 			\includegraphics[width=0.32\linewidth]{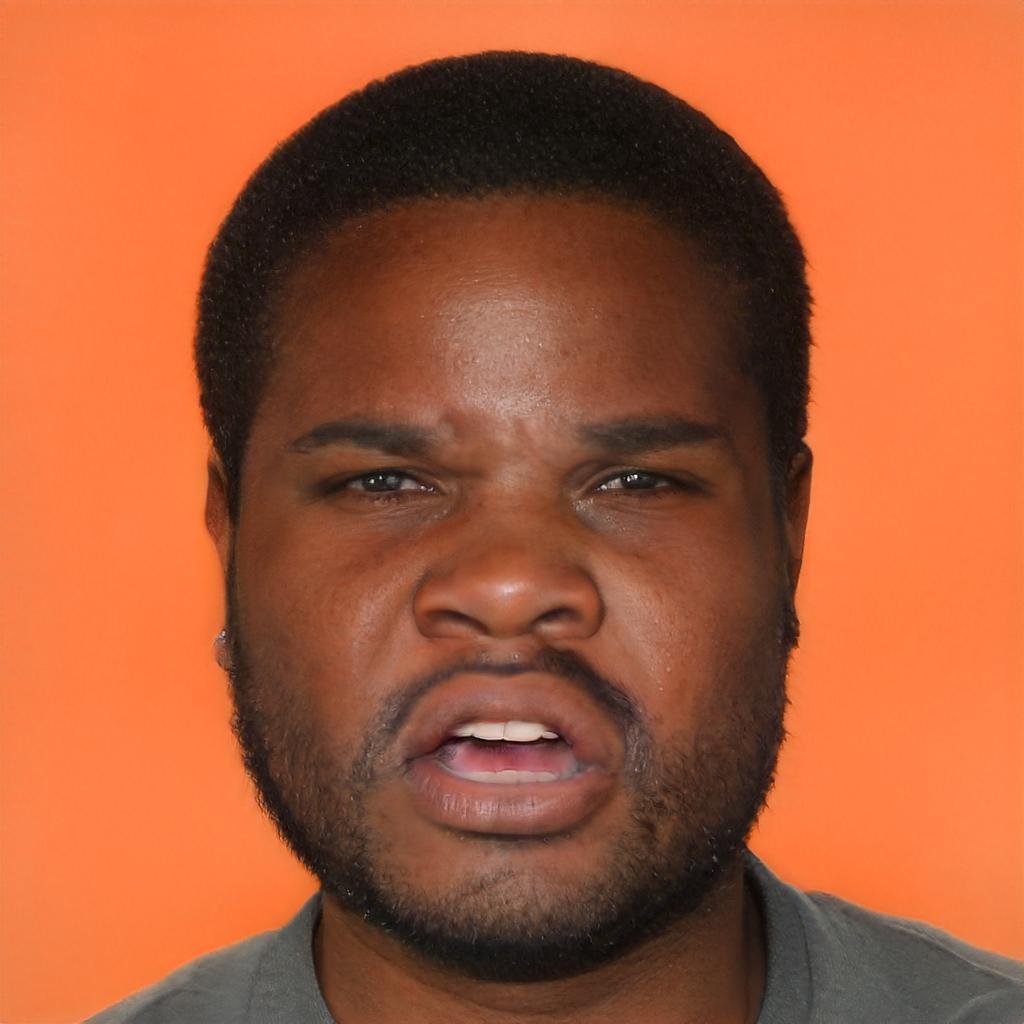}\\

            \includegraphics[width=0.32\linewidth]{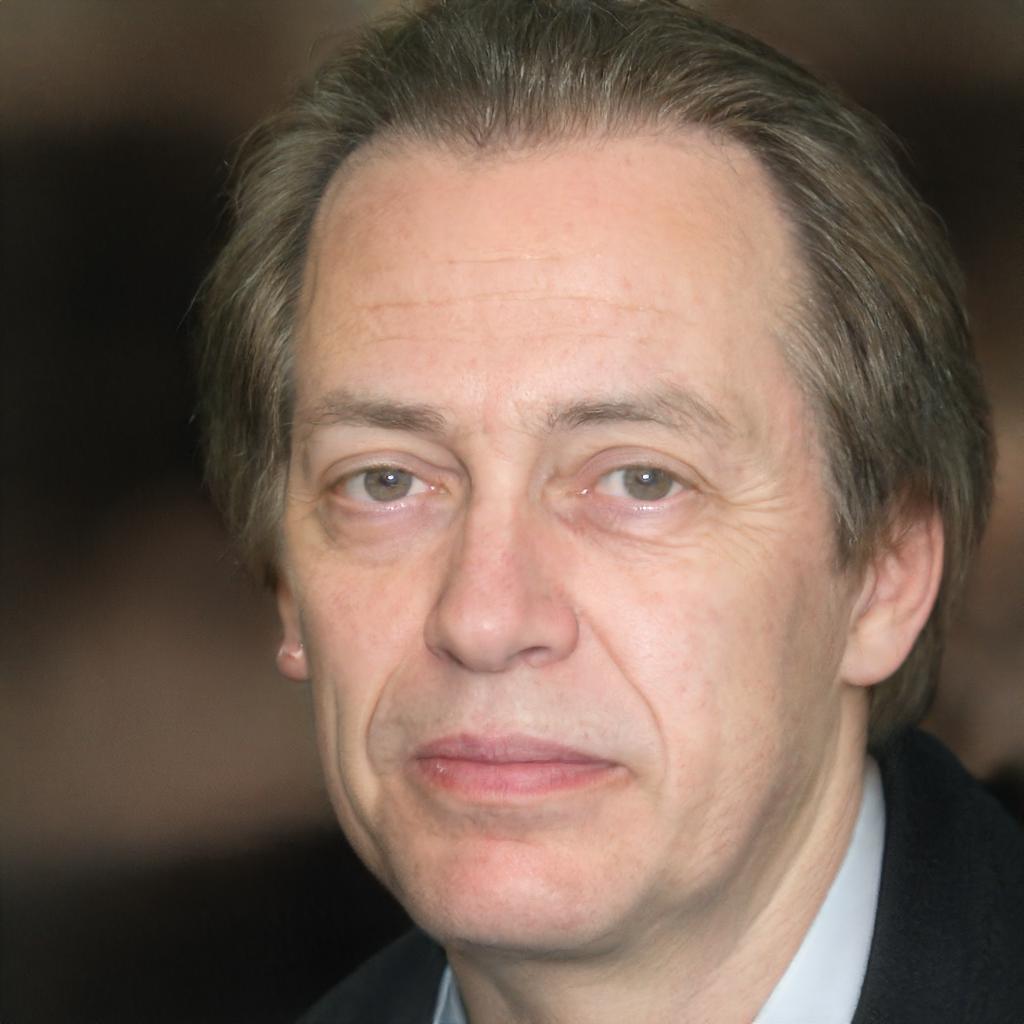} &
 			\includegraphics[width=0.32\linewidth]{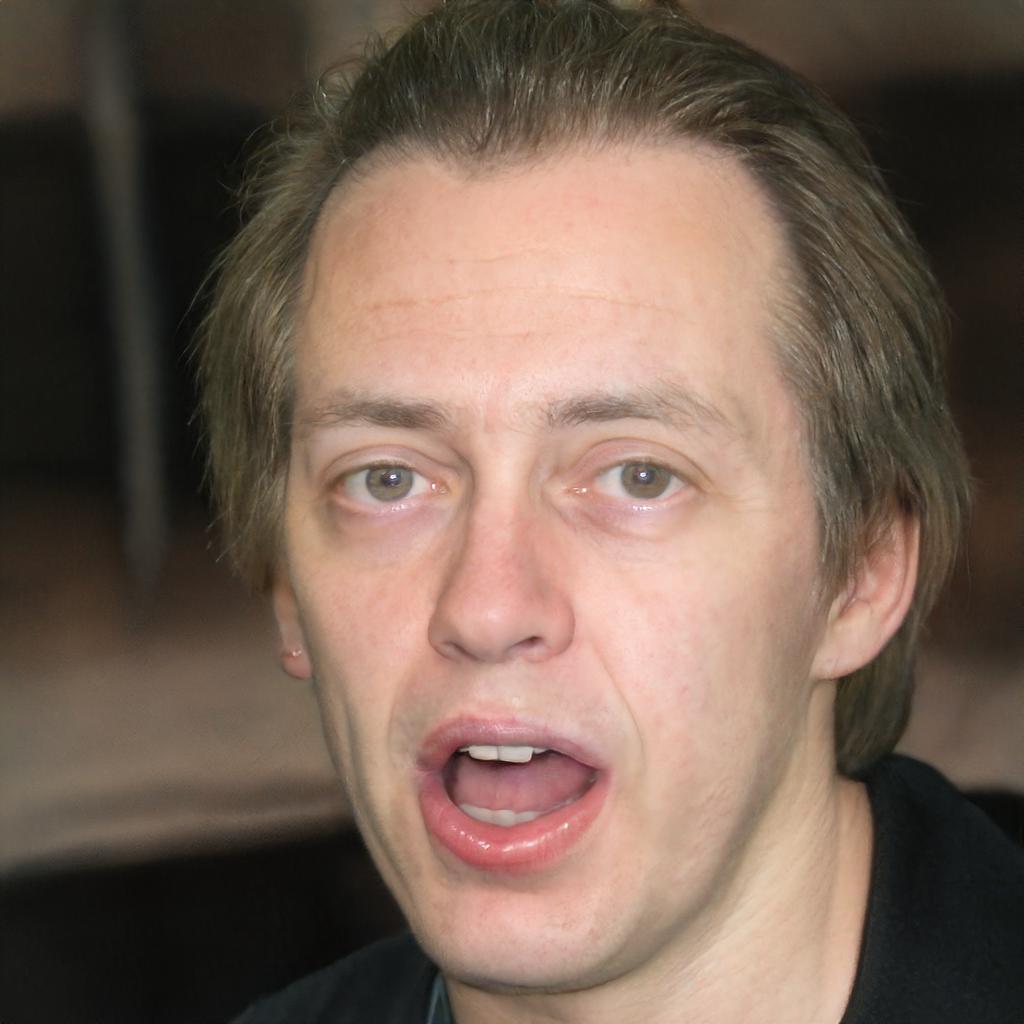} &
 			\includegraphics[width=0.32\linewidth]{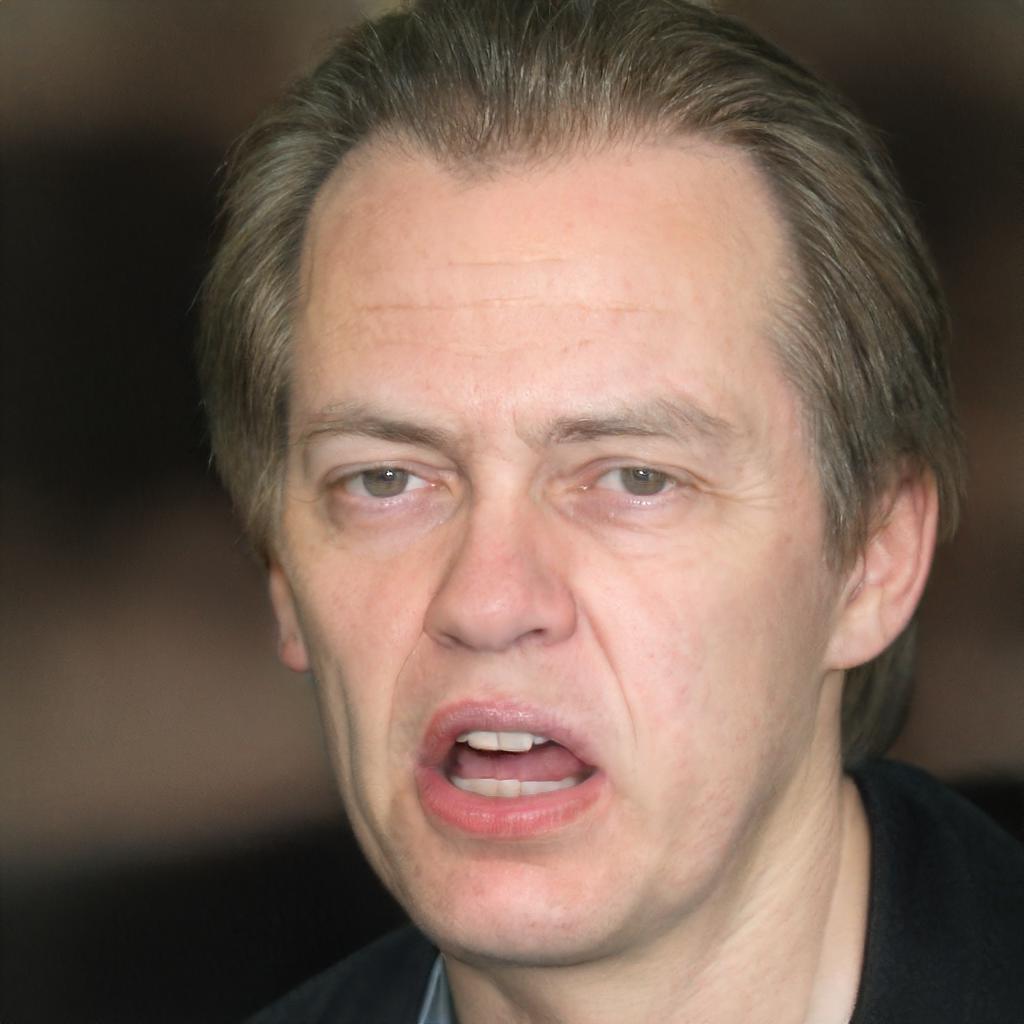}\\

            \includegraphics[width=0.32\linewidth]{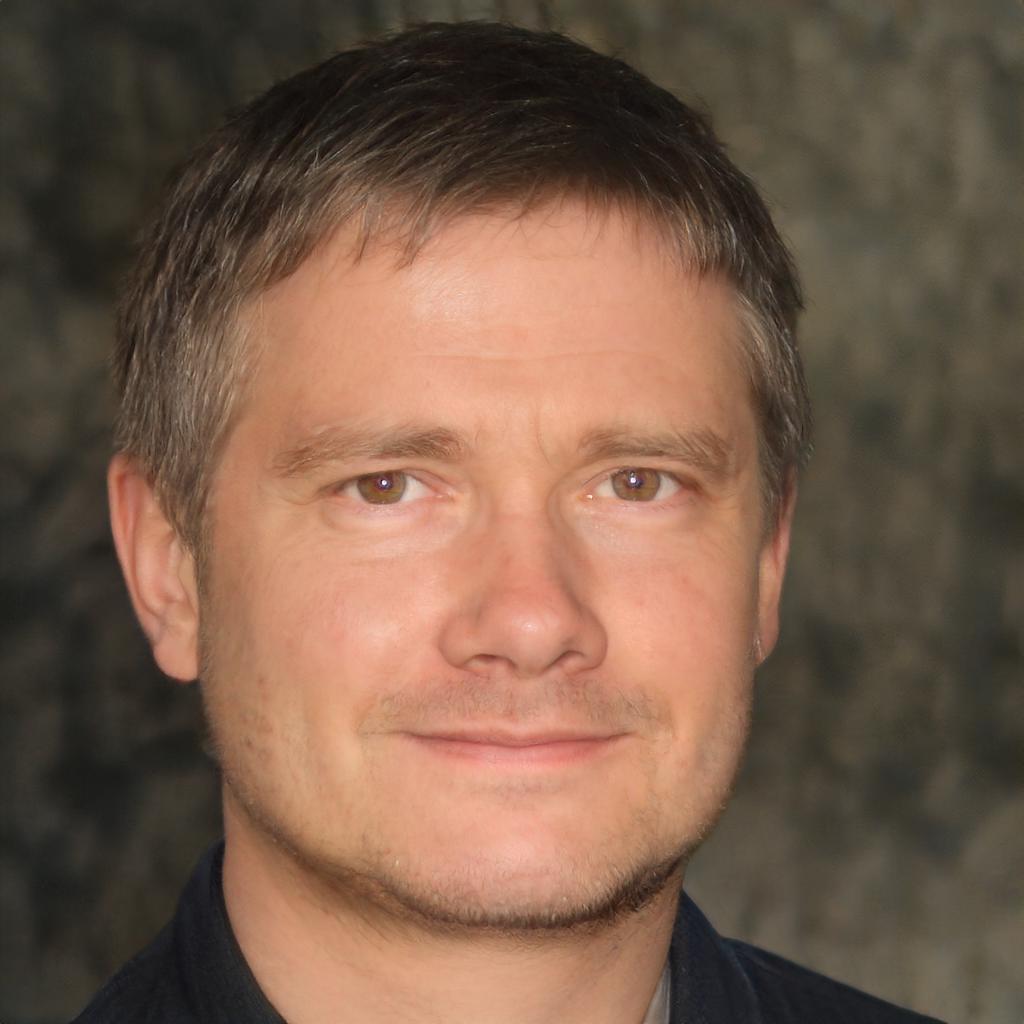} &
 			\includegraphics[width=0.32\linewidth]{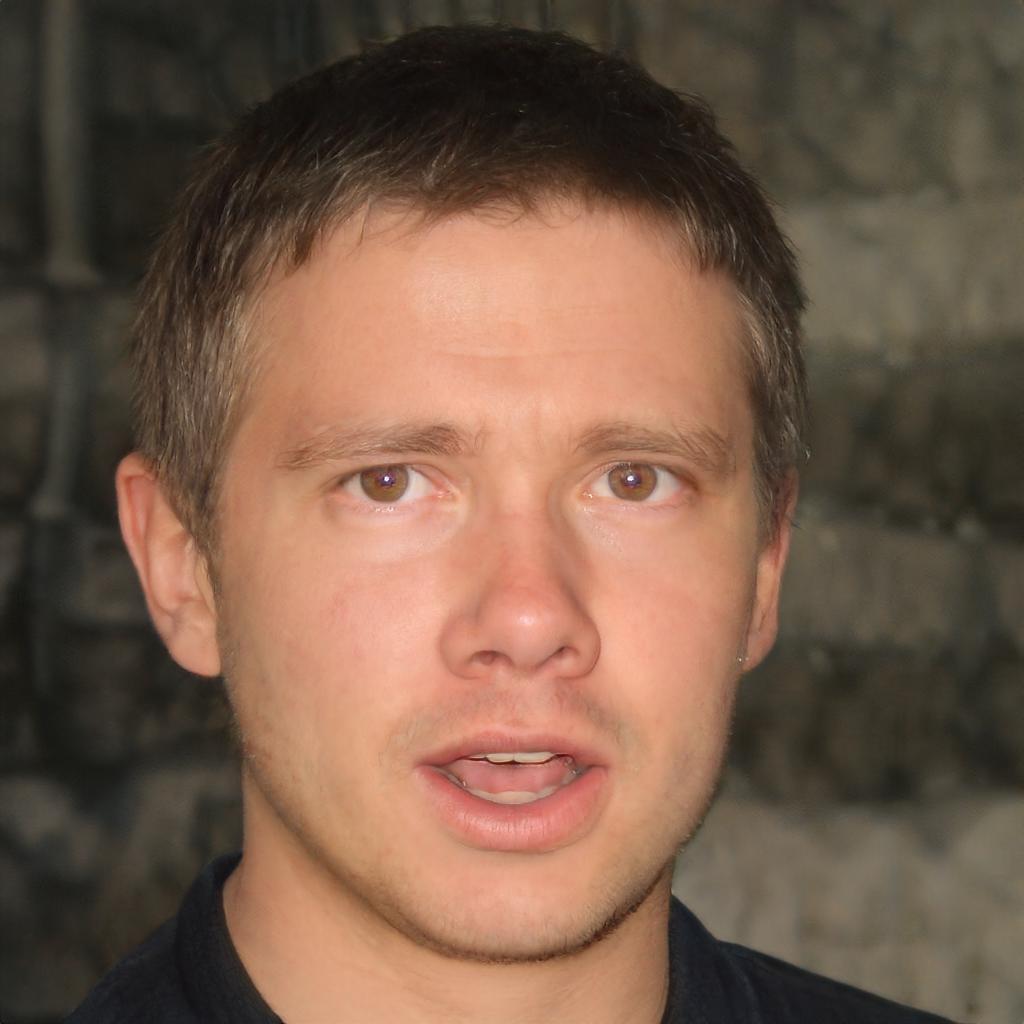} &
 			\includegraphics[width=0.32\linewidth]{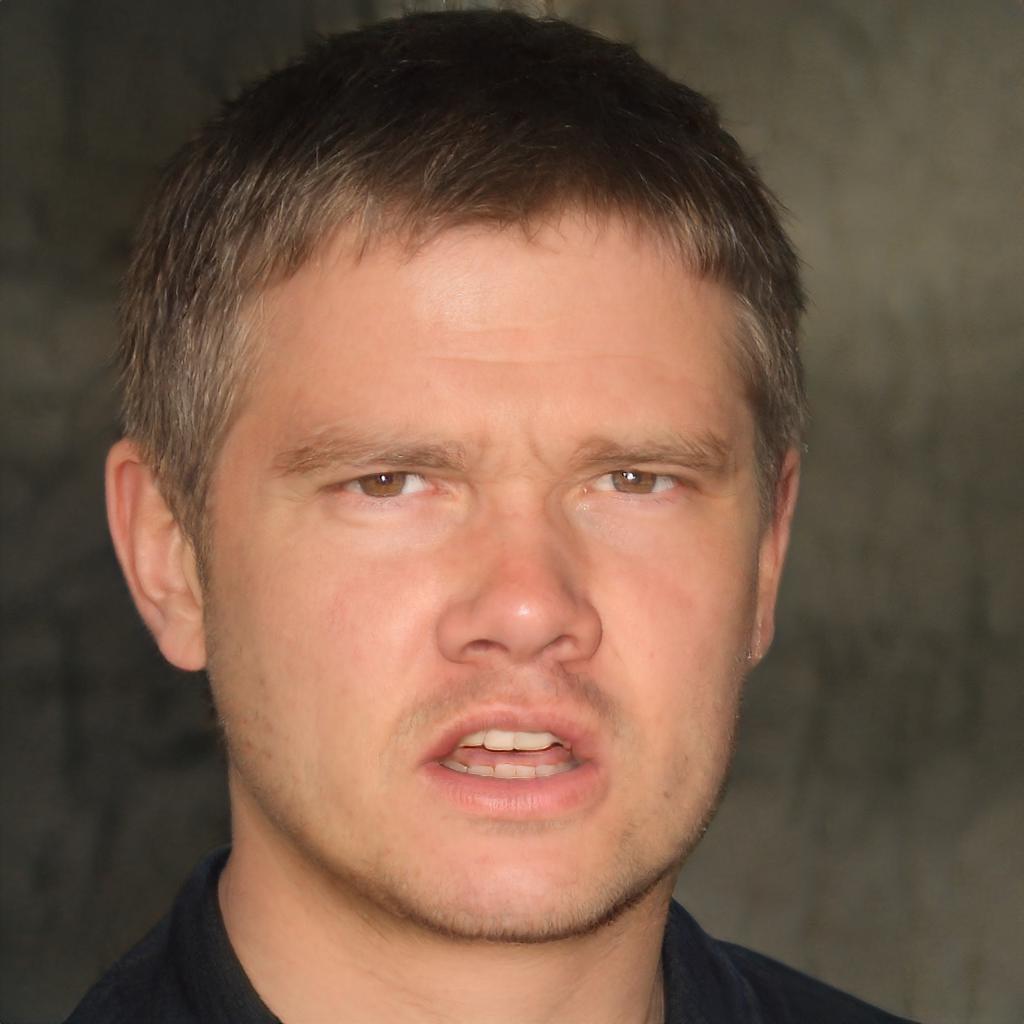}\\

			\includegraphics[width=0.32\linewidth]{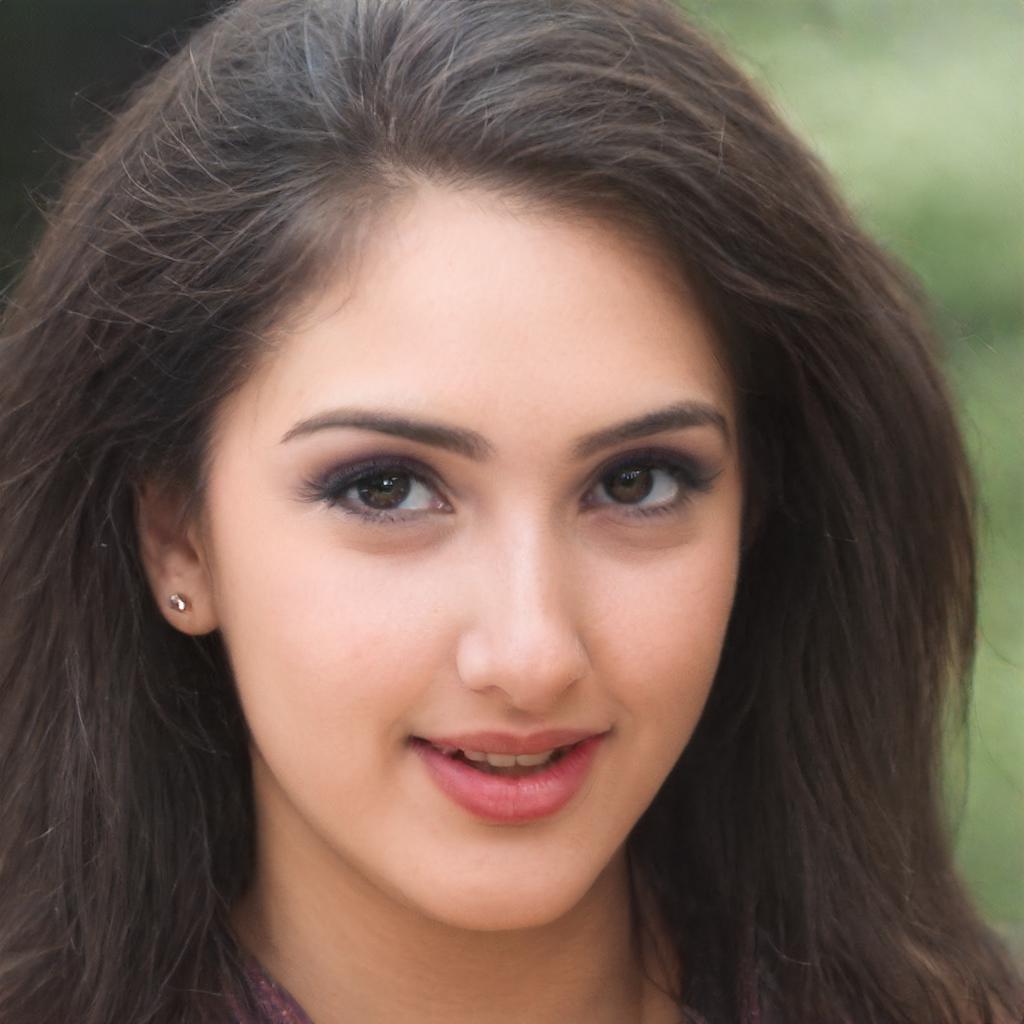} &
 			\includegraphics[width=0.32\linewidth]{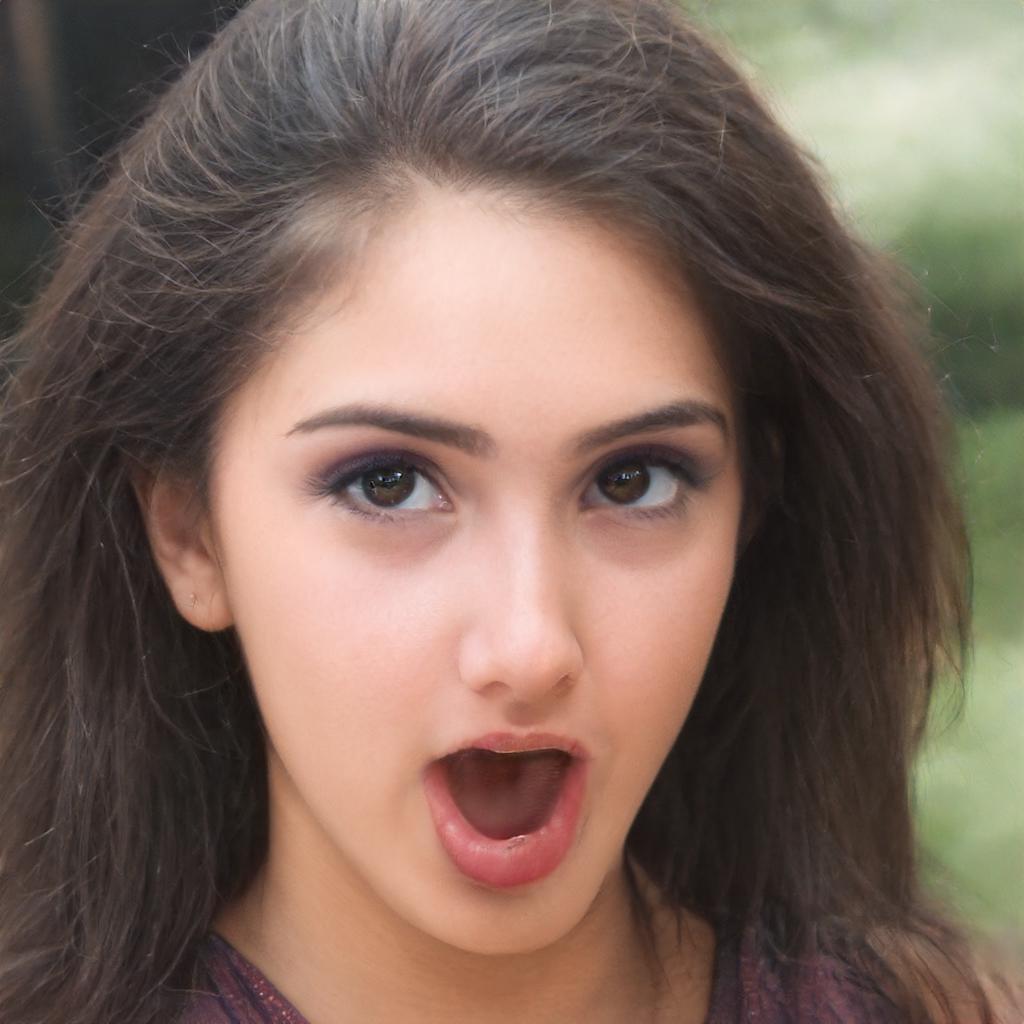} &
 			\includegraphics[width=0.32\linewidth]{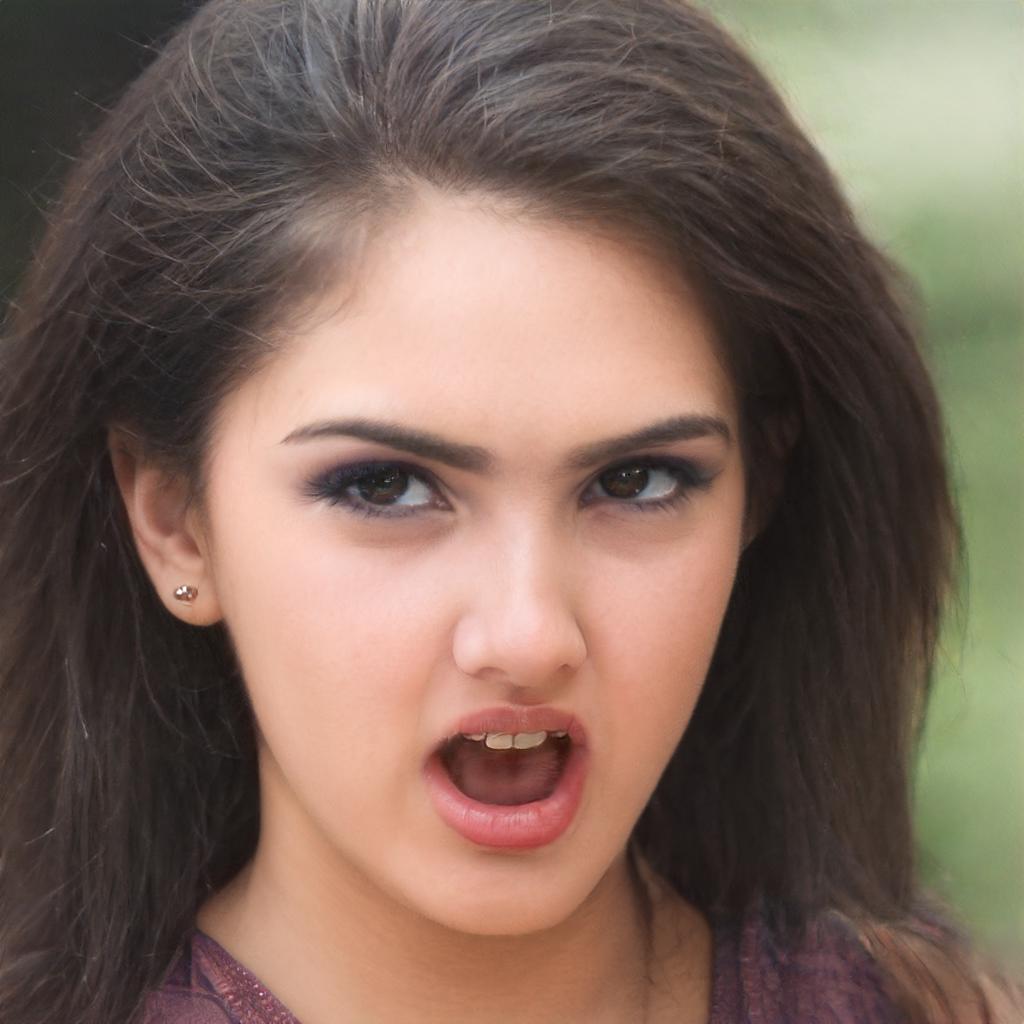}\\

            \includegraphics[width=0.32\linewidth]{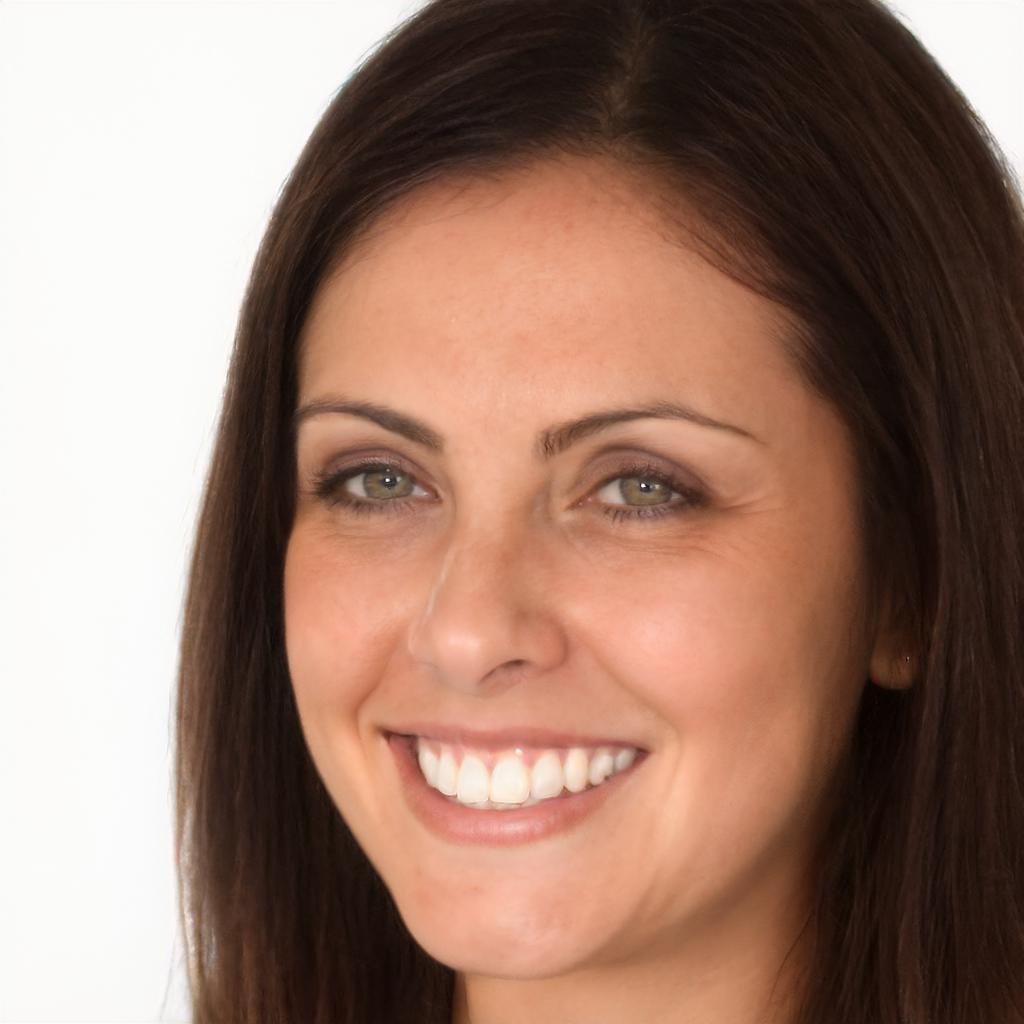} &
 			\includegraphics[width=0.32\linewidth]{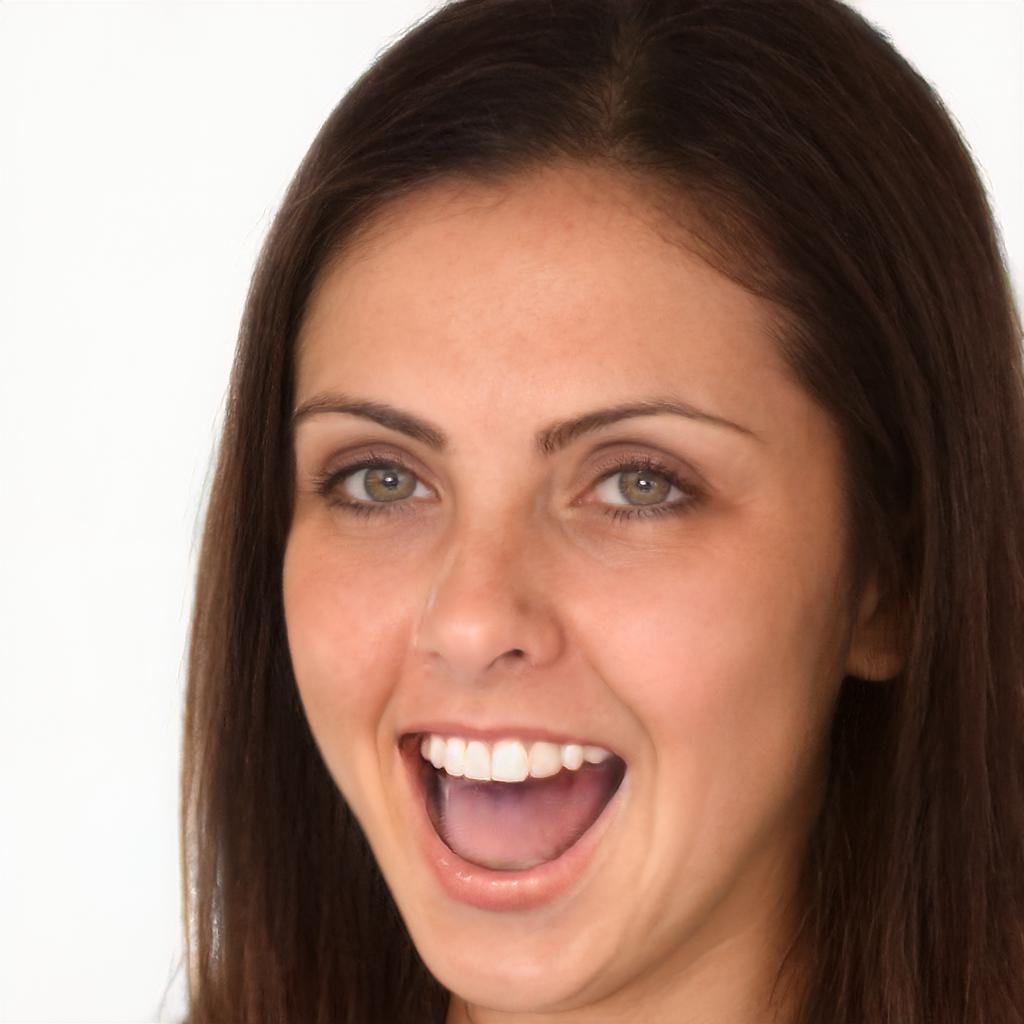} &
 			\includegraphics[width=0.32\linewidth]{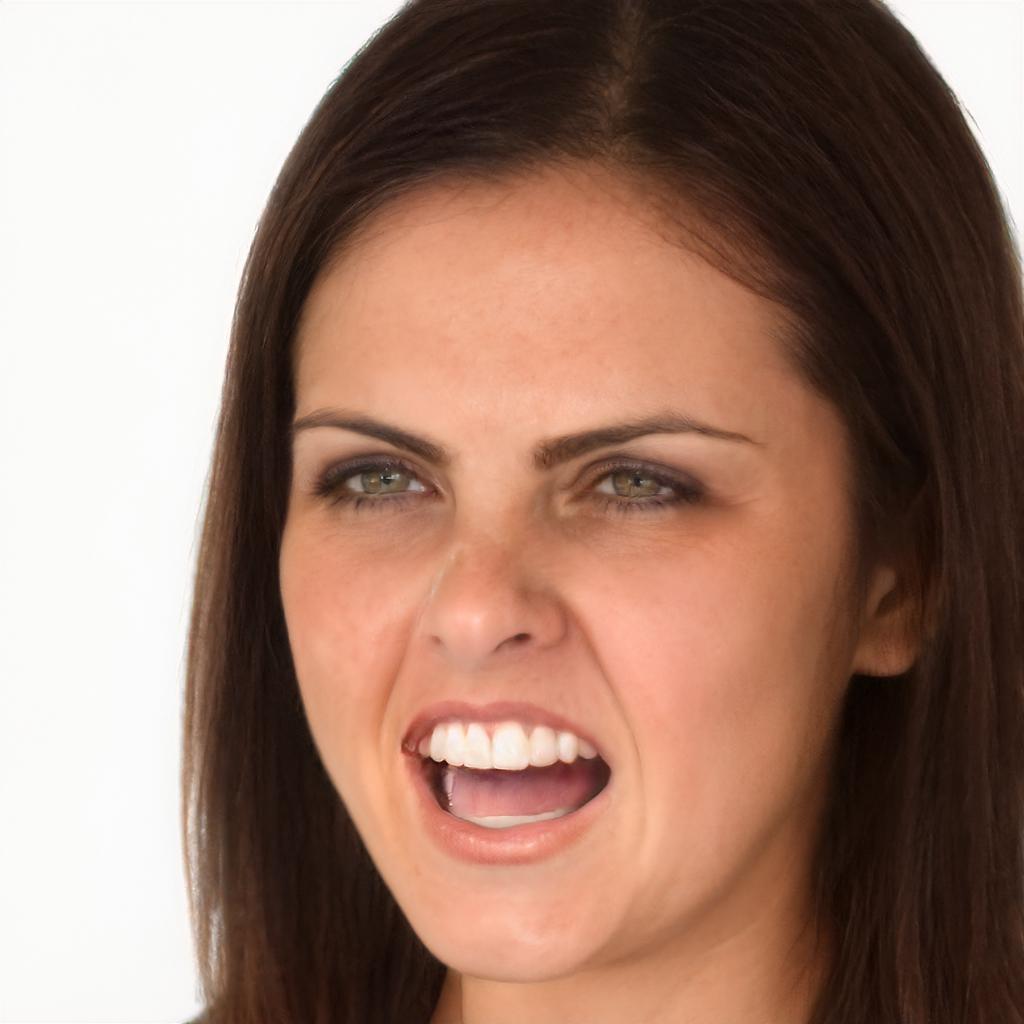}\\

            \includegraphics[width=0.32\linewidth]{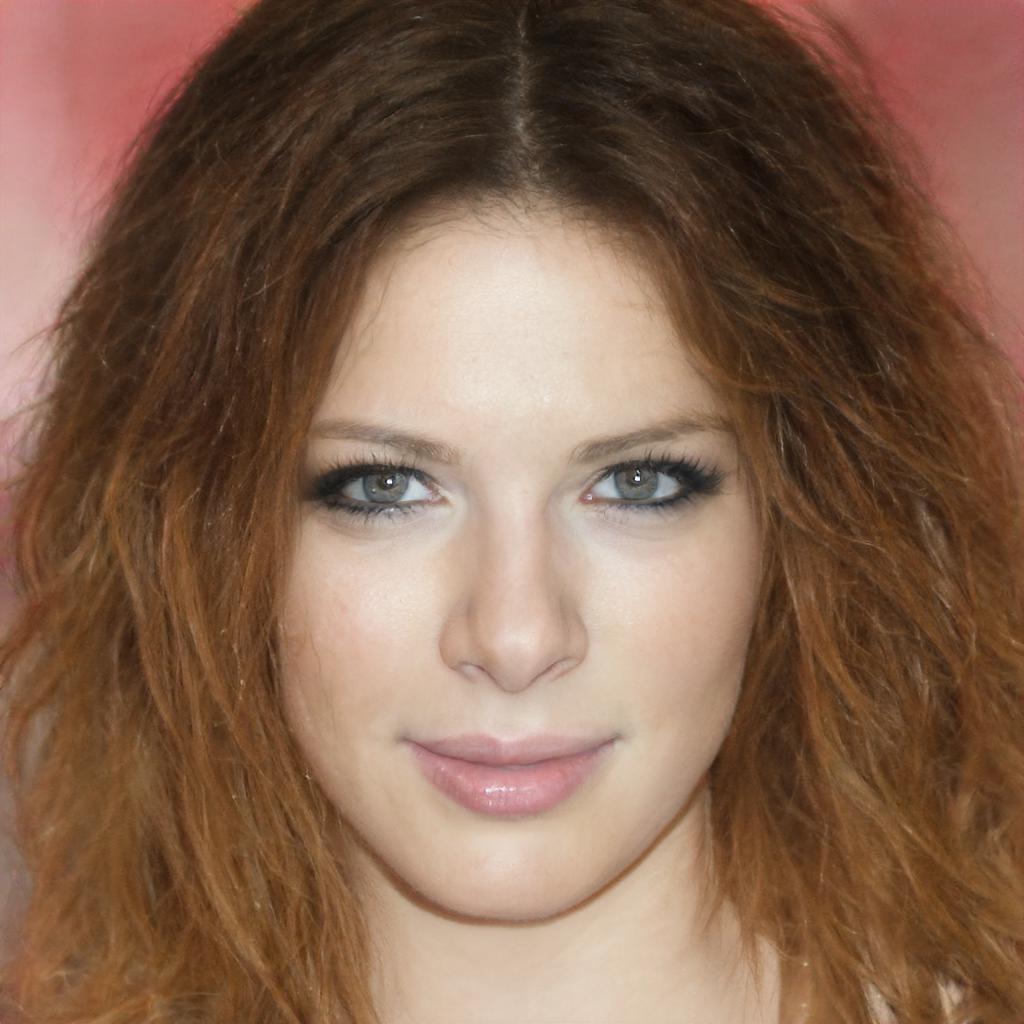} &
 			\includegraphics[width=0.32\linewidth]{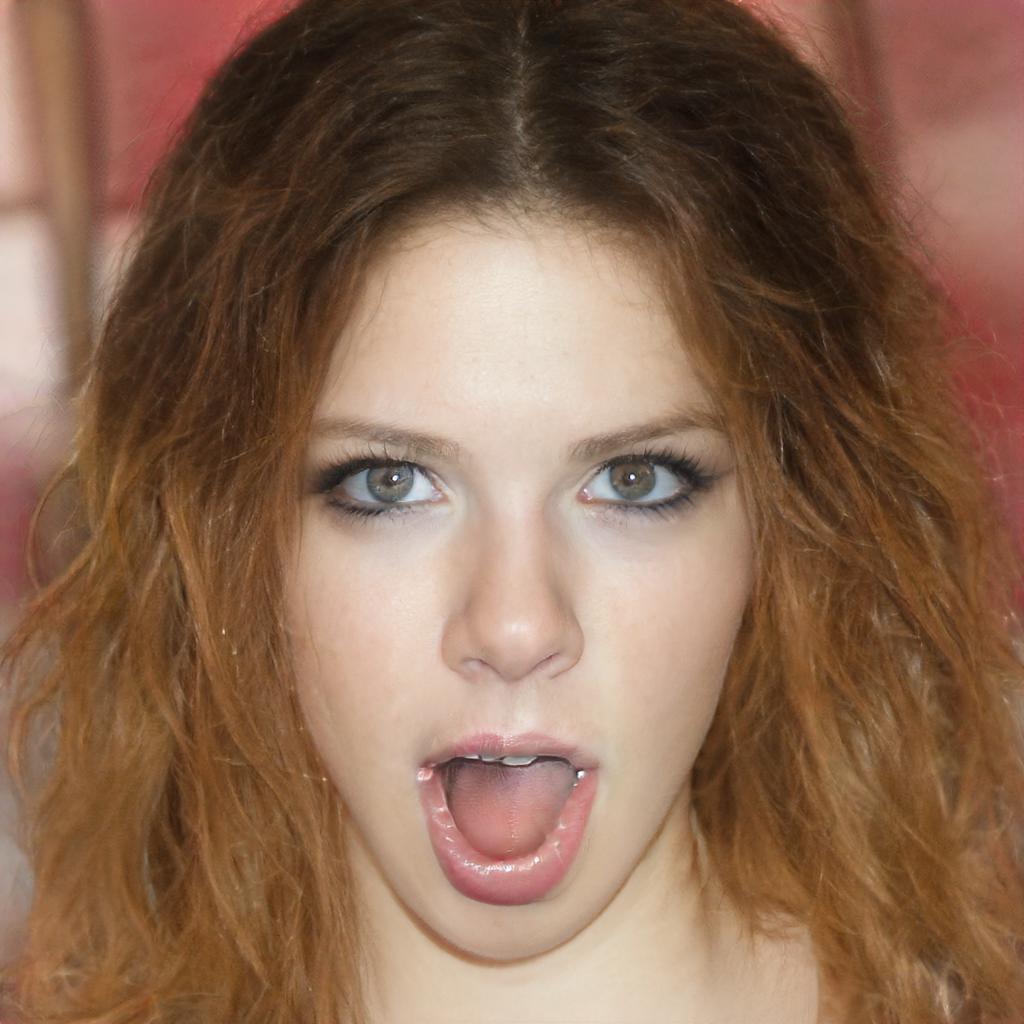} &
 			\includegraphics[width=0.32\linewidth]{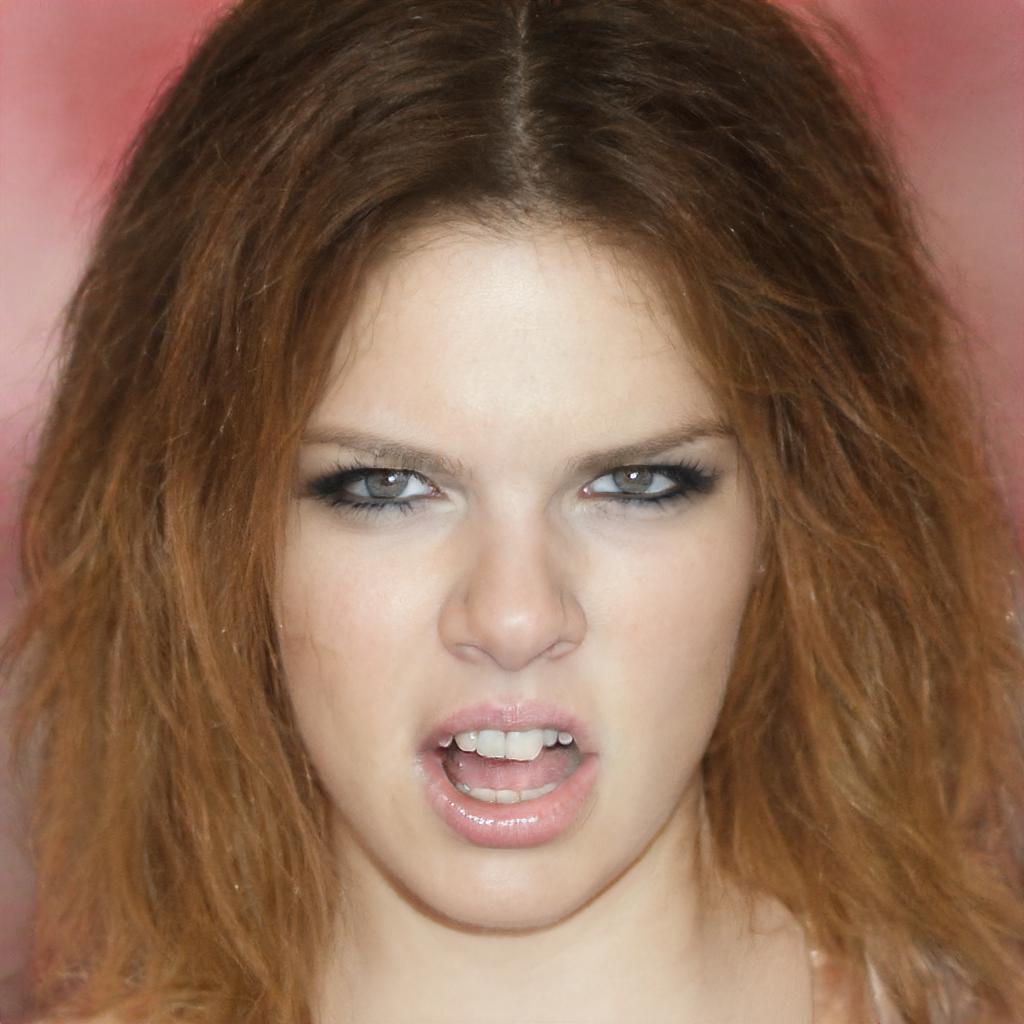}\\

            \includegraphics[width=0.32\linewidth]{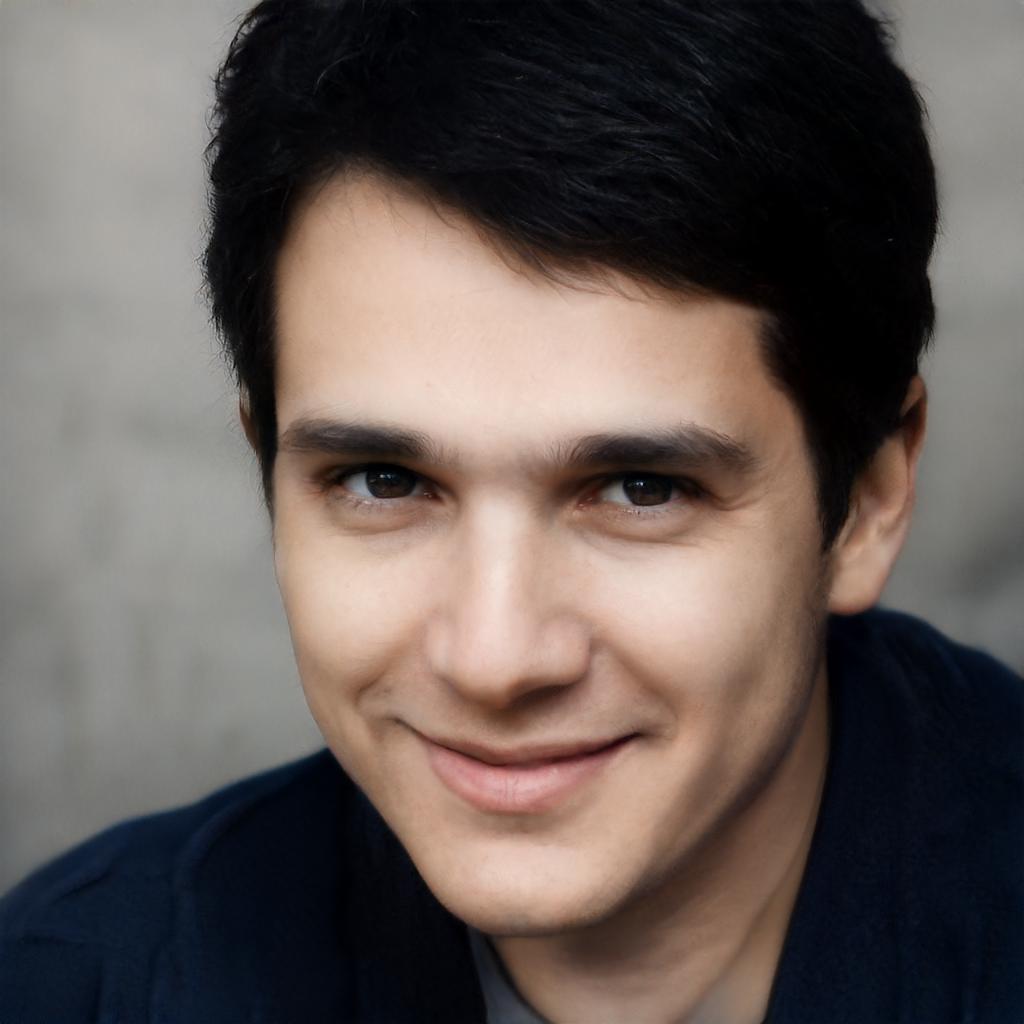} &
 			\includegraphics[width=0.32\linewidth]{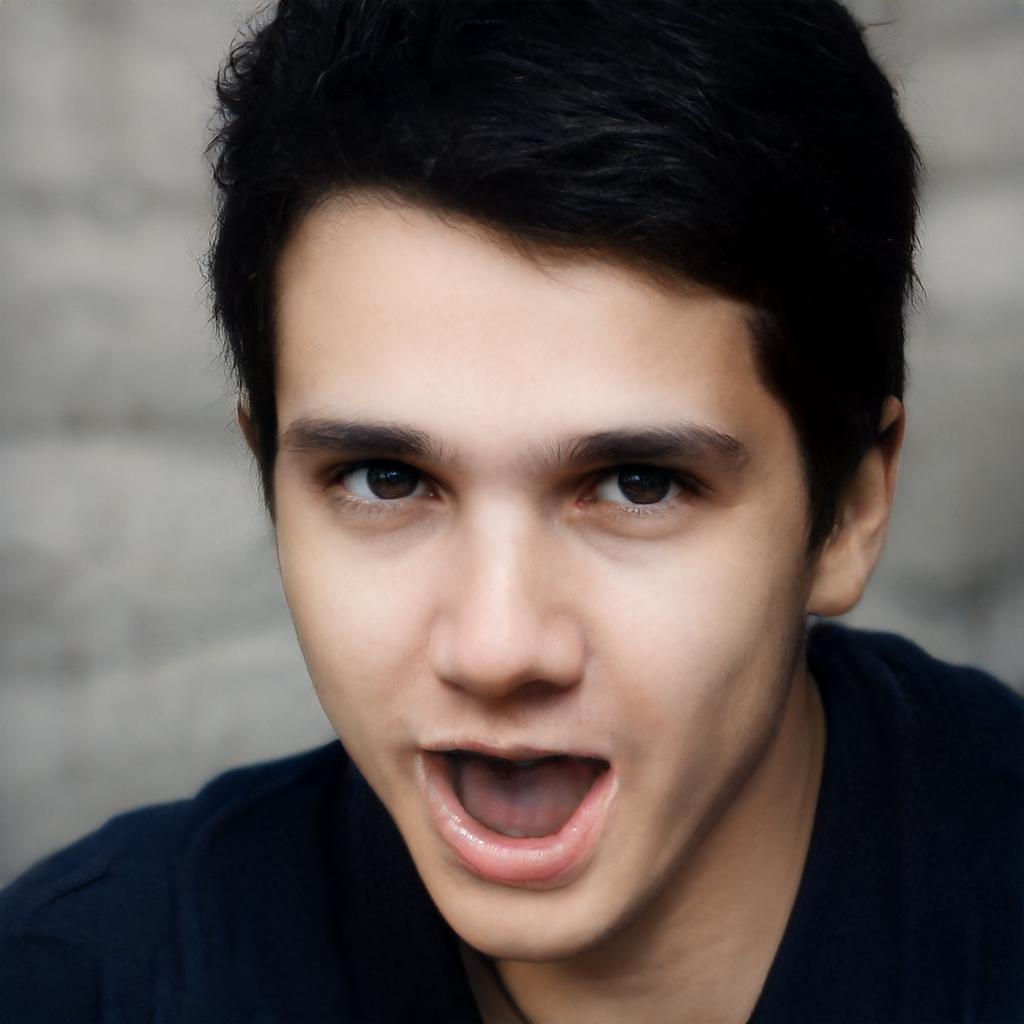} &
 			\includegraphics[width=0.32\linewidth]{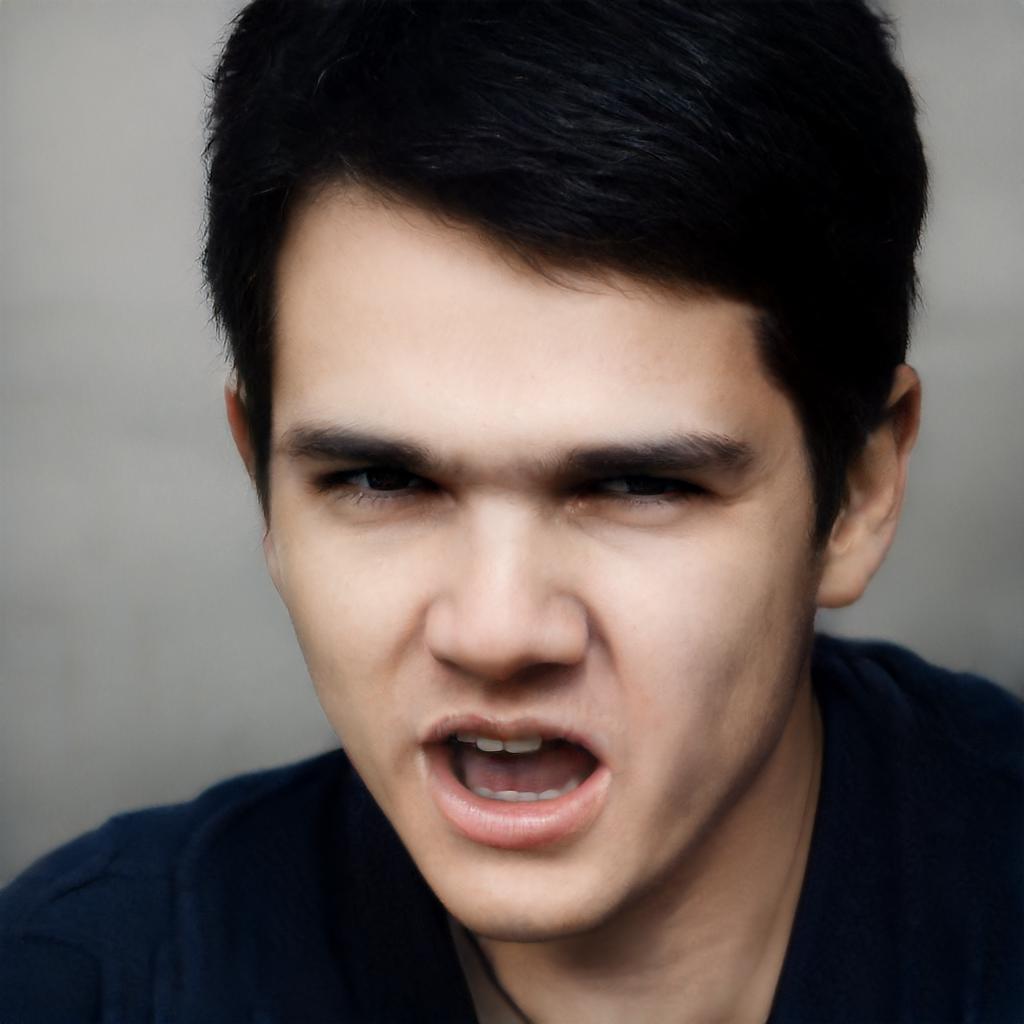}\\
 			Input & Surprised & Angry
		\end{tabular}
	}
	\caption{\label{fig:supp-expressions}Expression edits performed by the latent mapper.}
\end{figure}

In this section we provide additional results to those presented in the paper. Specifically, we begin with a variety of image manipulations obtained using our latent mapper. All manipulated images are taken from the CelebA-HQ and were inverted by e4e~\cite{tov2021designing}.
In Figure~\ref{fig:supp-hair} we show a large gallery of hair style manipulations. 
In Figures~\ref{fig:supp-women} and \ref{fig:supp-men} we show ``celeb'' edits, where the input image is manipulated to resemble a certain target celebrity.
In Figure~\ref{fig:supp-expressions} we show a variety of expression edits.



\begin{figure*}[tb]
	\centering
	\setlength{\tabcolsep}{1.1pt}
	
	{\footnotesize
		\begin{tabular}{ccccc}
			Input & Orange & Big Ears & Big Nose & Cute \\
			\includegraphics[width=0.18\textwidth]{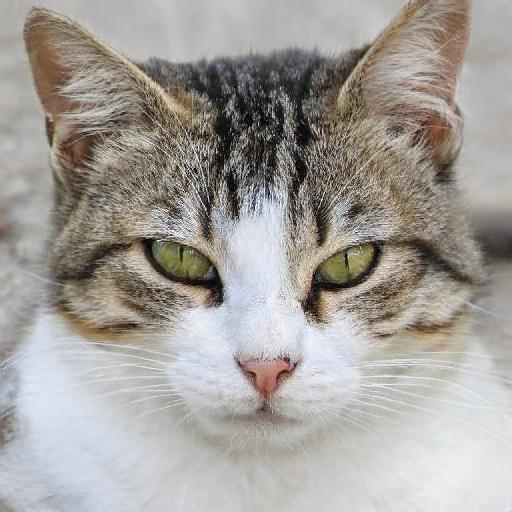} &
			\includegraphics[width=0.18\textwidth]{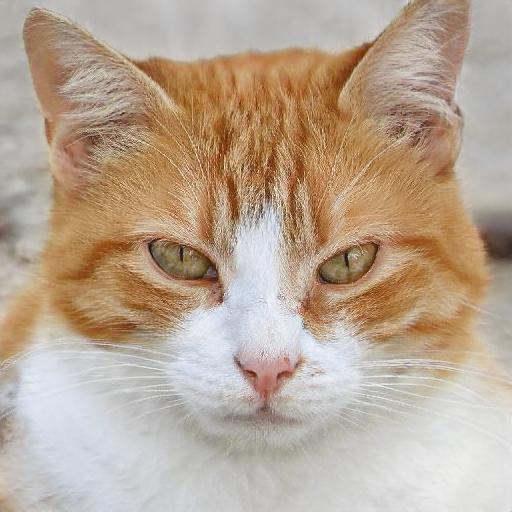} &
			\includegraphics[width=0.18\textwidth]{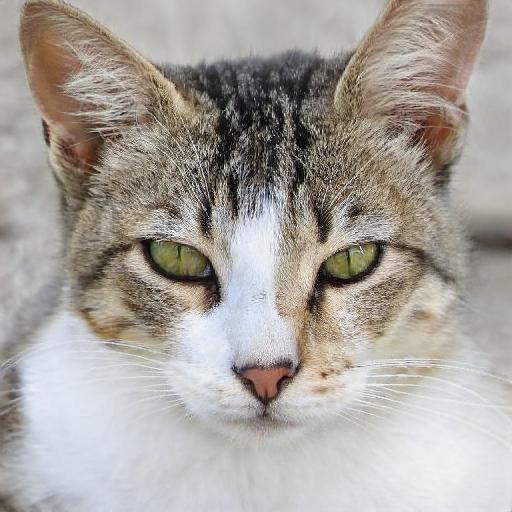} &
			\includegraphics[width=0.18\textwidth]{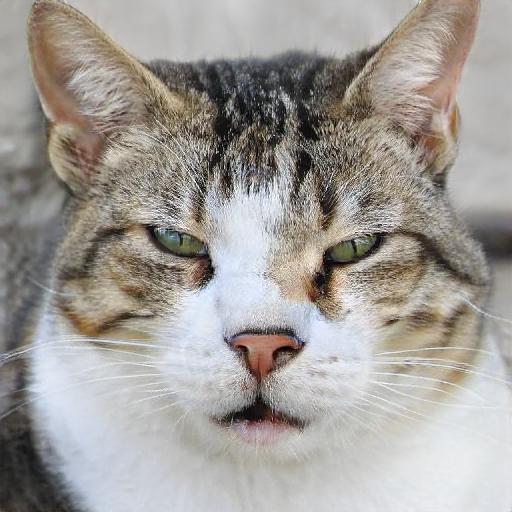} &
			\includegraphics[width=0.18\textwidth]{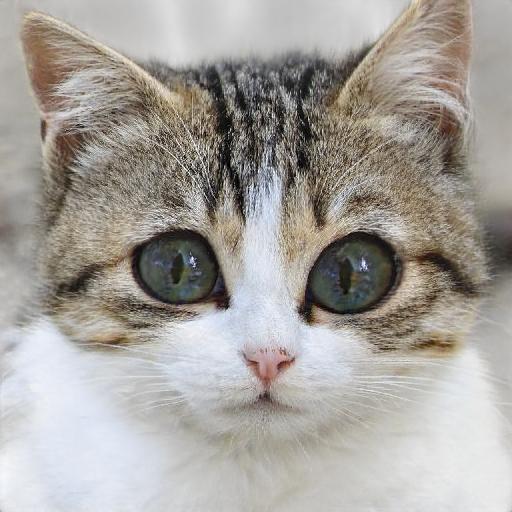}
			\\
			\includegraphics[width=0.18\textwidth]{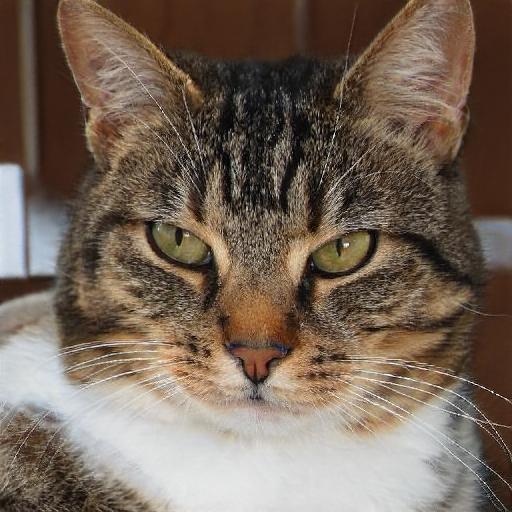} &
			\includegraphics[width=0.18\textwidth]{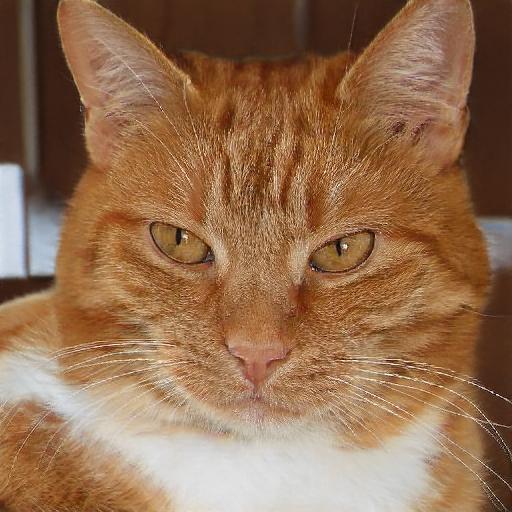} &
			\includegraphics[width=0.18\textwidth]{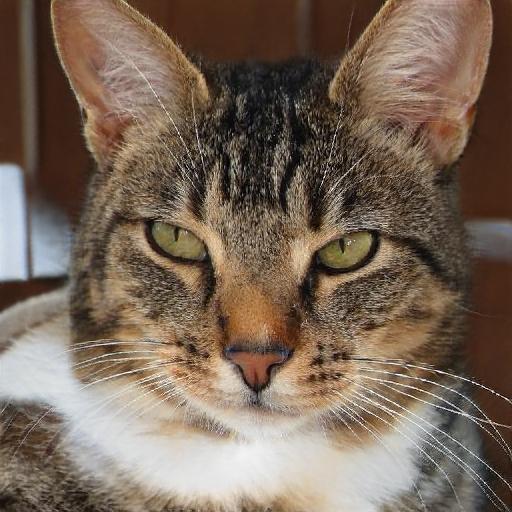} &
			\includegraphics[width=0.18\textwidth]{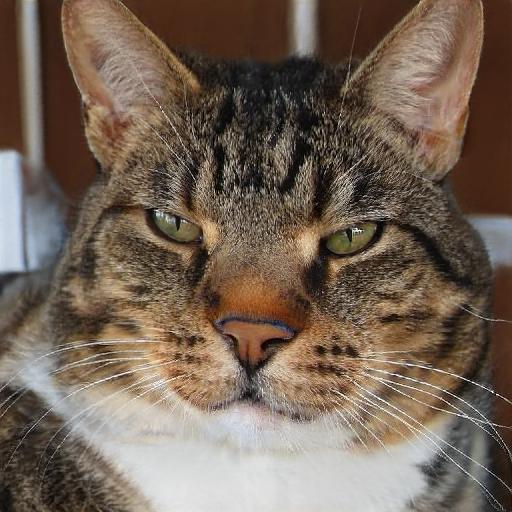} &
			\includegraphics[width=0.18\textwidth]{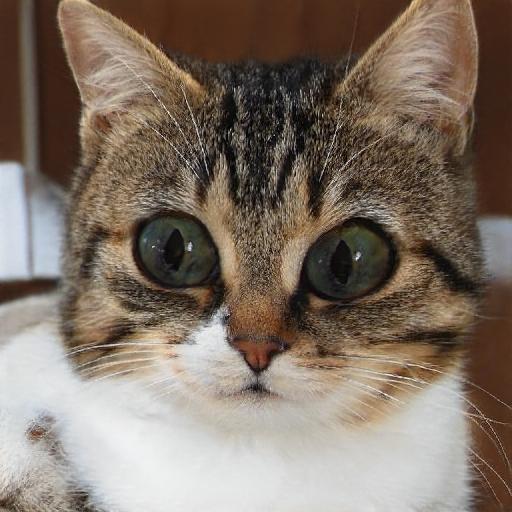}
			\\
			\includegraphics[width=0.18\textwidth]{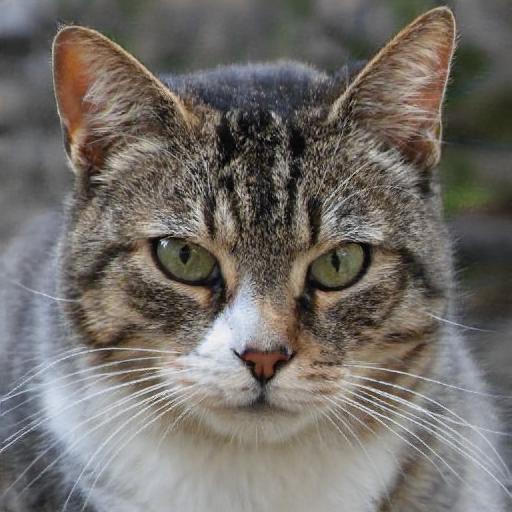} &
			\includegraphics[width=0.18\textwidth]{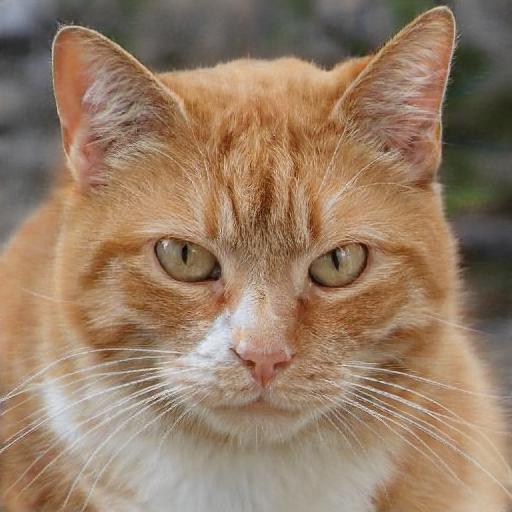} &
			\includegraphics[width=0.18\textwidth]{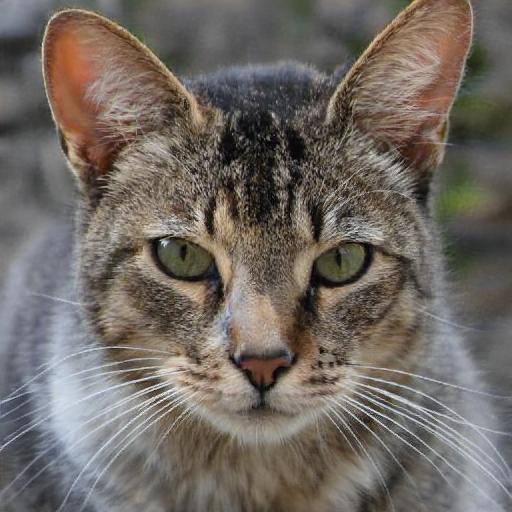} &
			\includegraphics[width=0.18\textwidth]{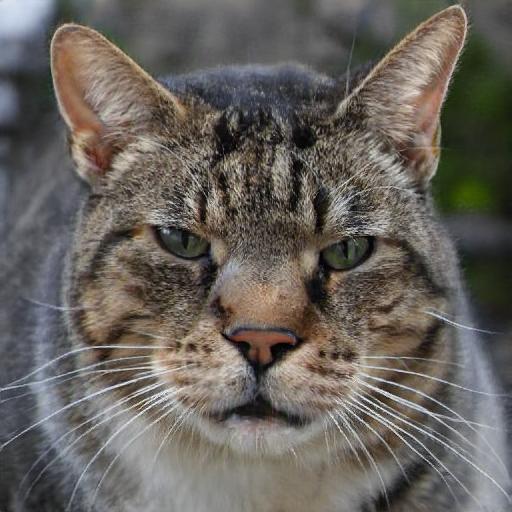} &
			\includegraphics[width=0.18\textwidth]{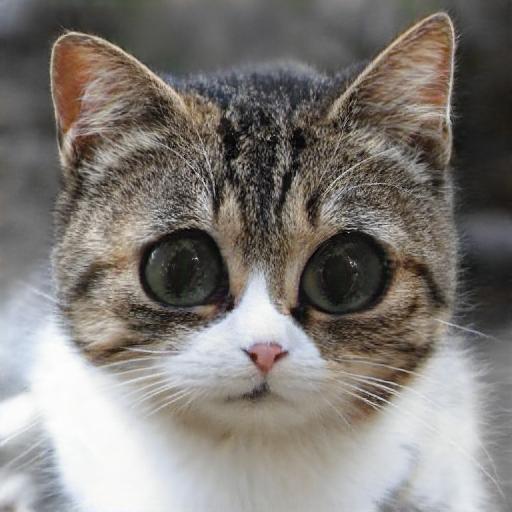}
			\\
			\\
			Input & Trees & Clouds  &  Spires & Round Roof \\
			\includegraphics[width=0.18\textwidth]{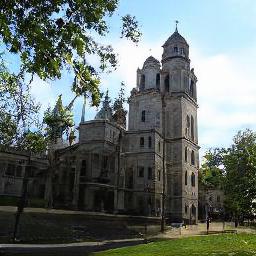} &
			\includegraphics[width=0.18\textwidth]{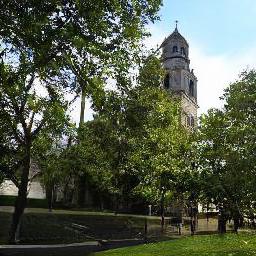} &
			\includegraphics[width=0.18\textwidth]{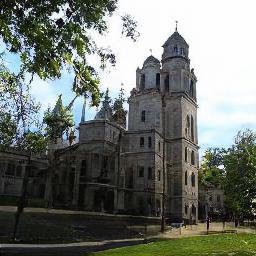} &
			\includegraphics[width=0.18\textwidth]{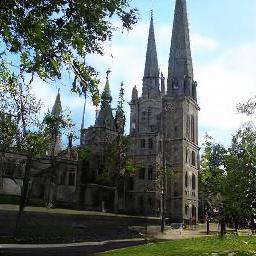} &
			\includegraphics[width=0.18\textwidth]{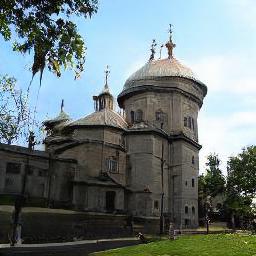} 
			\\
			\includegraphics[width=0.18\textwidth]{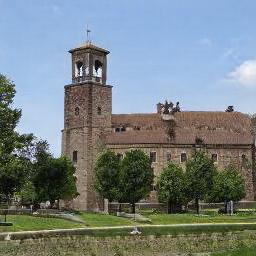} &
			\includegraphics[width=0.18\textwidth]{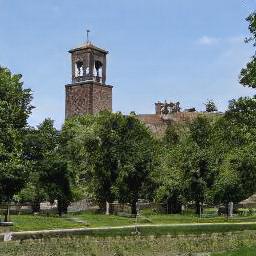} &
			\includegraphics[width=0.18\textwidth]{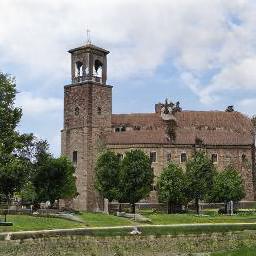} &
			\includegraphics[width=0.18\textwidth]{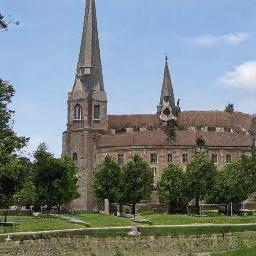} &
			\includegraphics[width=0.18\textwidth]{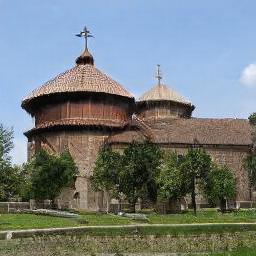} 
			\\
			\includegraphics[width=0.18\textwidth]{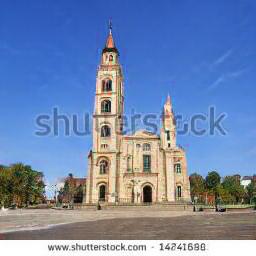} &
			\includegraphics[width=0.18\textwidth]{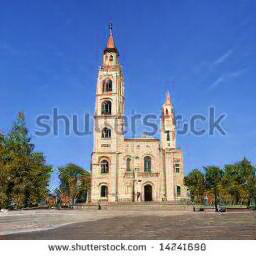} &
			\includegraphics[width=0.18\textwidth]{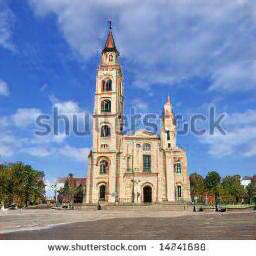} &
			\includegraphics[width=0.18\textwidth]{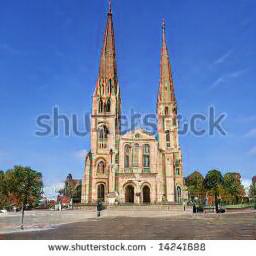} &
			\includegraphics[width=0.18\textwidth]{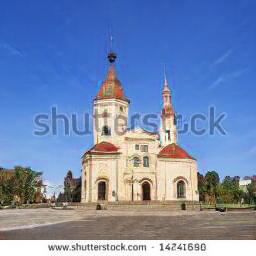} 			
		\end{tabular}
	}
	\caption{A variety of edits for non-face images along text-driven global latent manipulation directions.  Left: using StyleGAN2-ada~\cite{karras2020training} pretrained on AFHQ cats~\cite{choi2020stargan}. Right: using StyleGAN2 pretrained on LSUN Church~\cite{yu2015lsun}. The target attribute used in the text prompt is indicated above each column.}
	\label{fig:alex-nonfaces-supp}
\end{figure*}

Next, Figure~\ref{fig:alex-nonfaces-supp} shows a variety of edits on non-face datasets, performed along text-driven global latent manipulation directions (Section 6).

\begin{figure*}[tb]
    \centering
	\setlength{\tabcolsep}{1pt}	
	\begin{tabular}{cccccc}
		& {\footnotesize $\alpha=-2$} & {\footnotesize $\alpha=-1$} & {\footnotesize Original} &{\footnotesize $\alpha=1$} &{\footnotesize $\alpha=2$} \\
		
		\rotatebox{90}{\footnotesize \phantom{k} $\beta=0.40$} &
		\includegraphics[width=0.25\columnwidth]{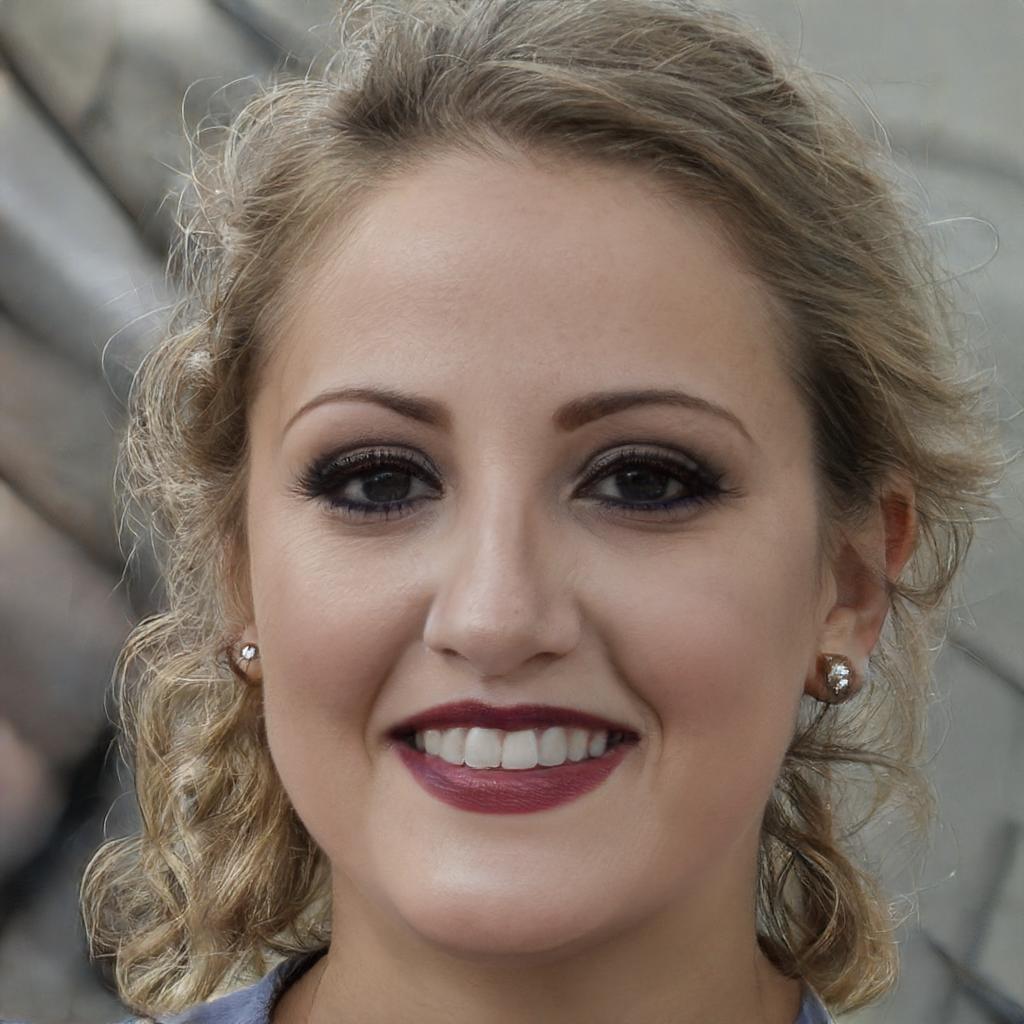} &
		\includegraphics[width=0.25\columnwidth]{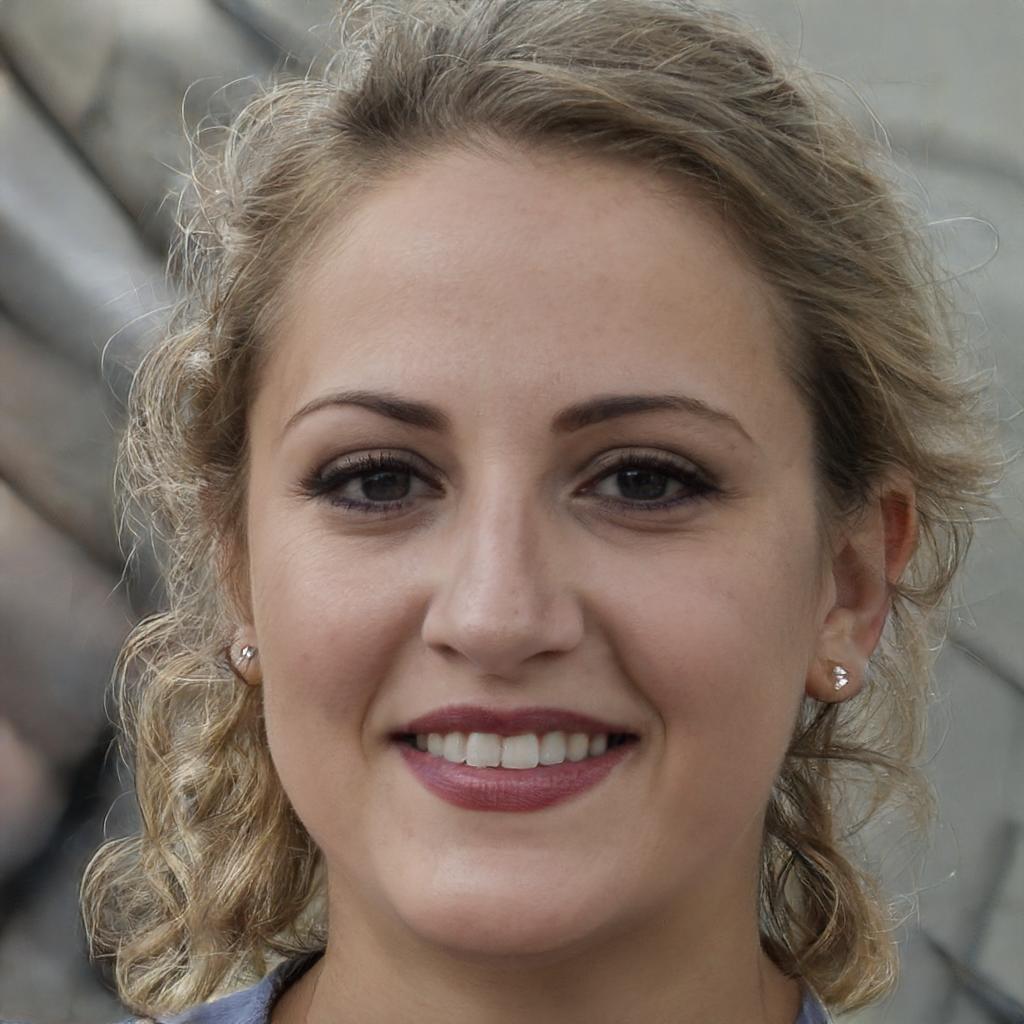} &
		\includegraphics[width=0.25\columnwidth]{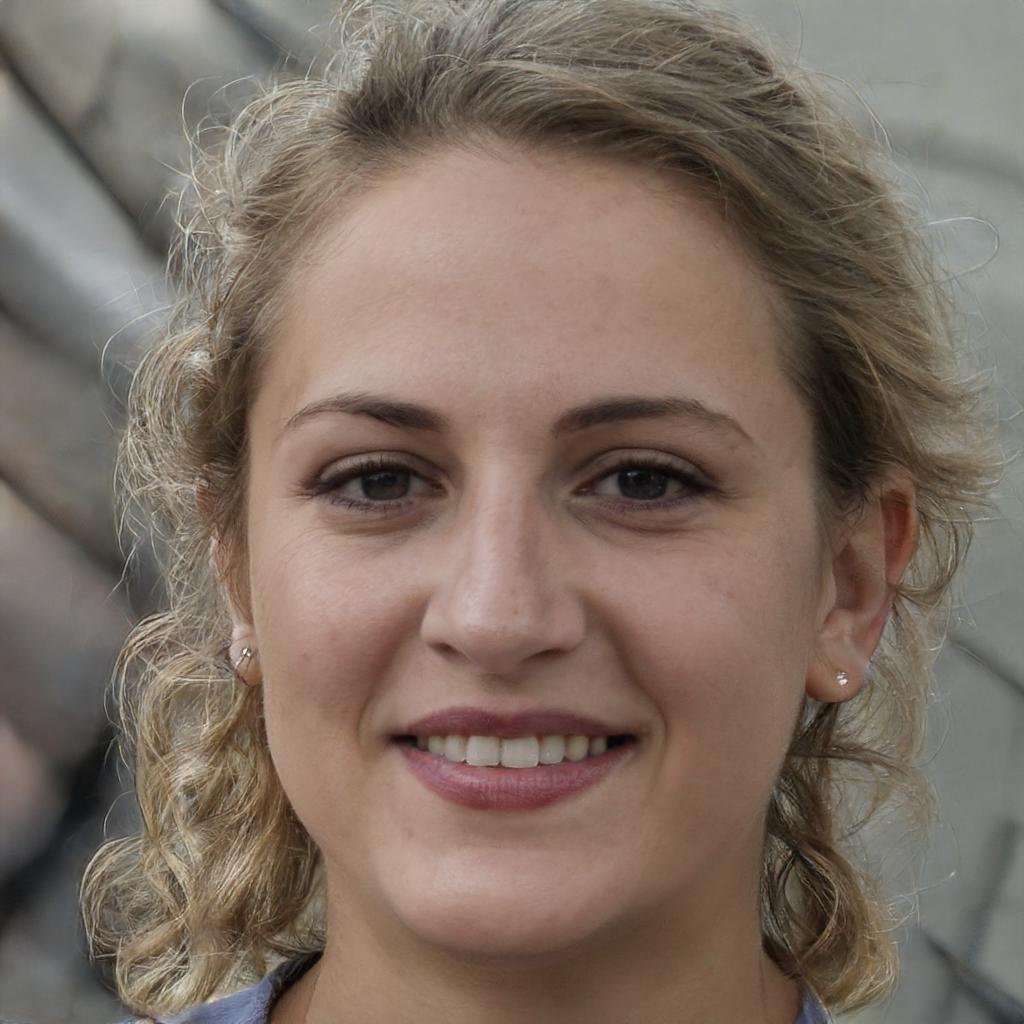} &
		\includegraphics[width=0.25\columnwidth]{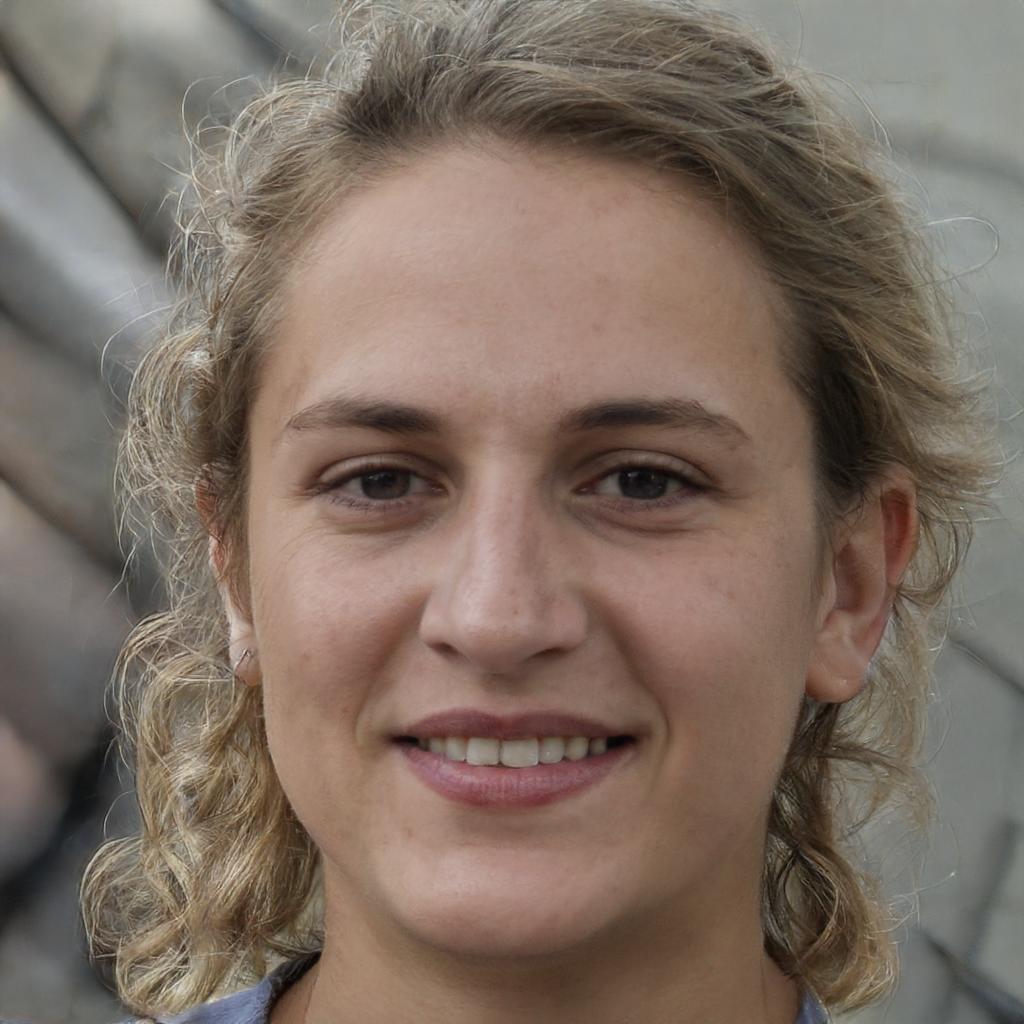} &
		\includegraphics[width=0.25\columnwidth]{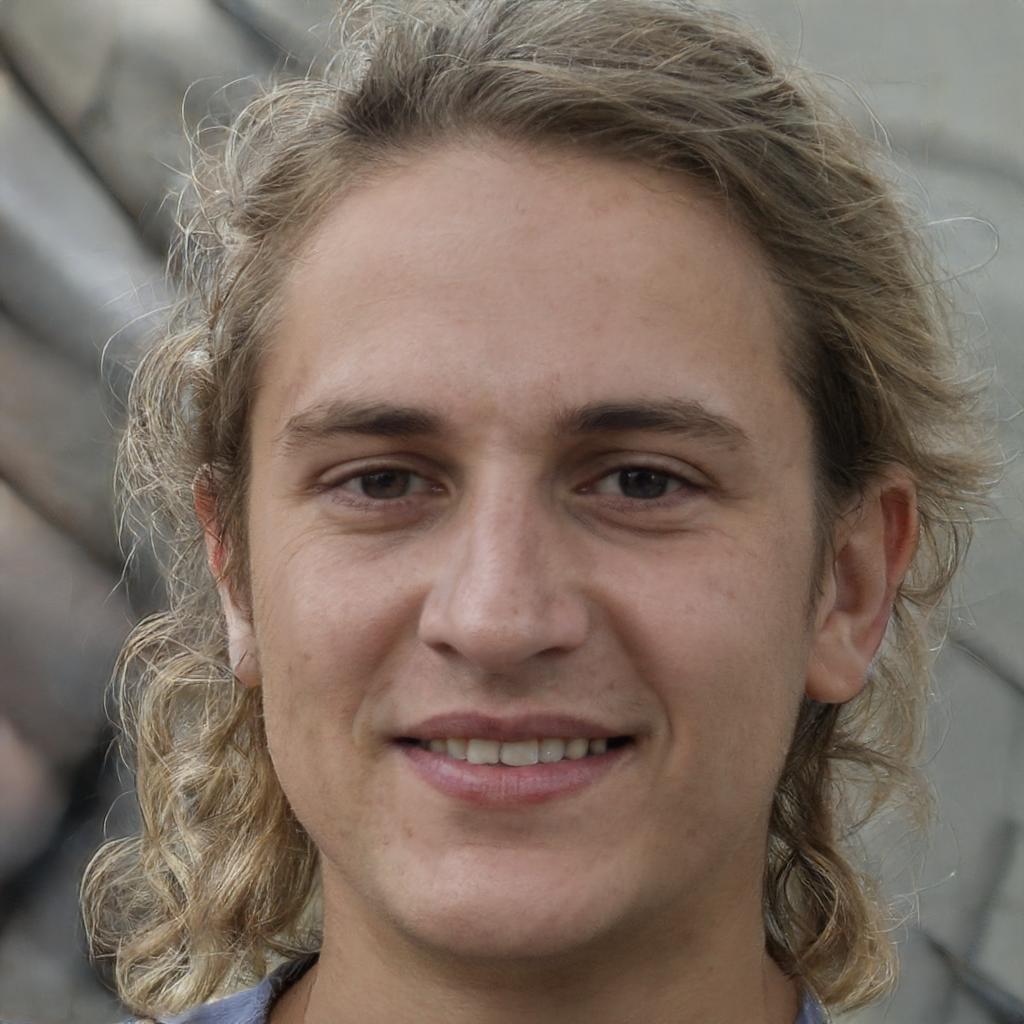} 
		\\
		\rotatebox{90}{\footnotesize \phantom{k} $\beta=0.30$} &
		\includegraphics[width=0.25\columnwidth]{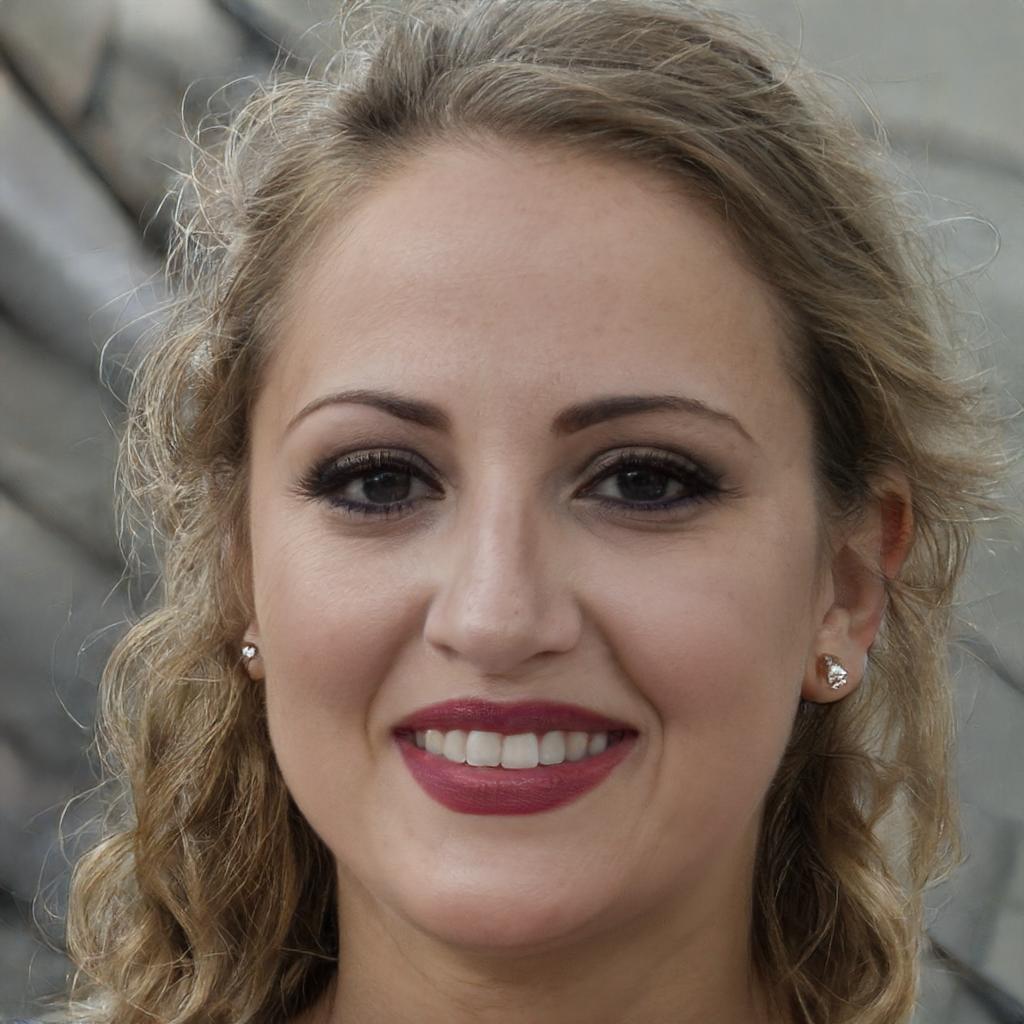} &
		\includegraphics[width=0.25\columnwidth]{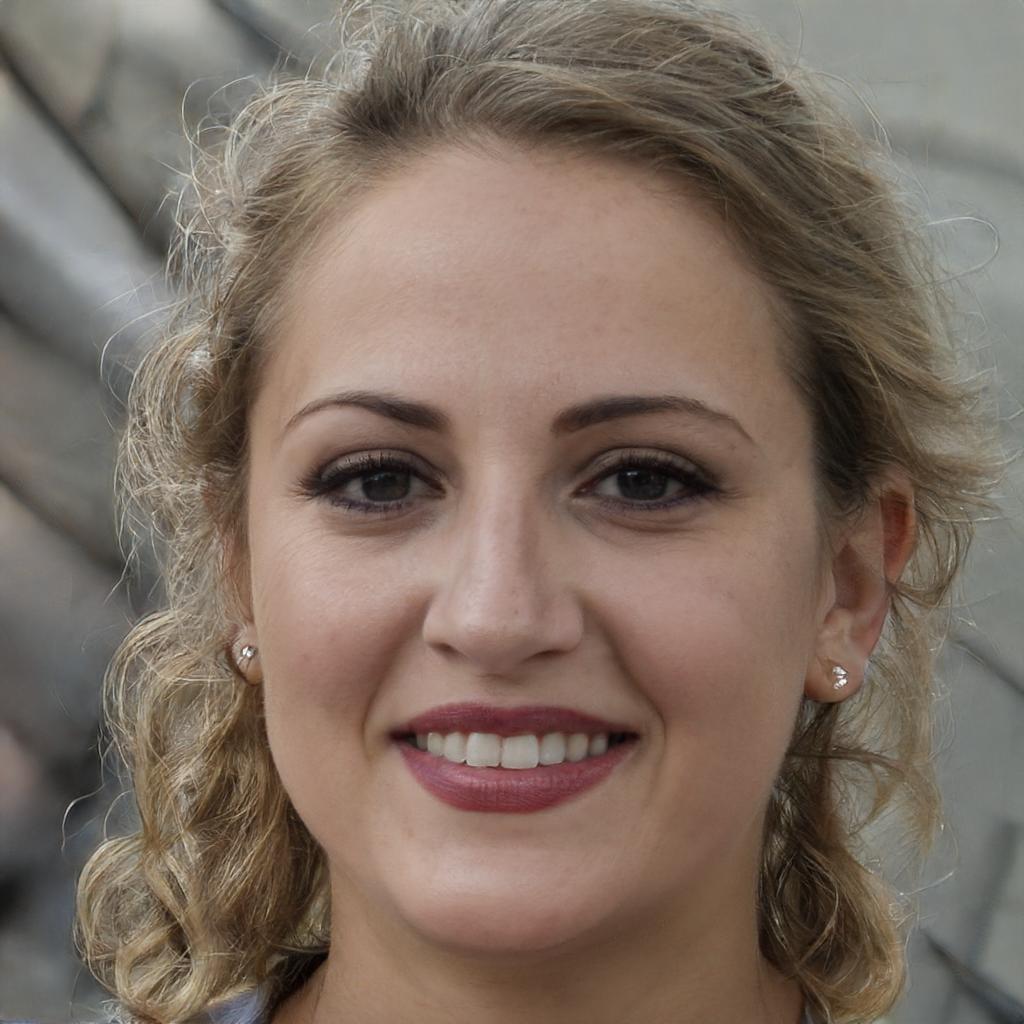} &
		\includegraphics[width=0.25\columnwidth]{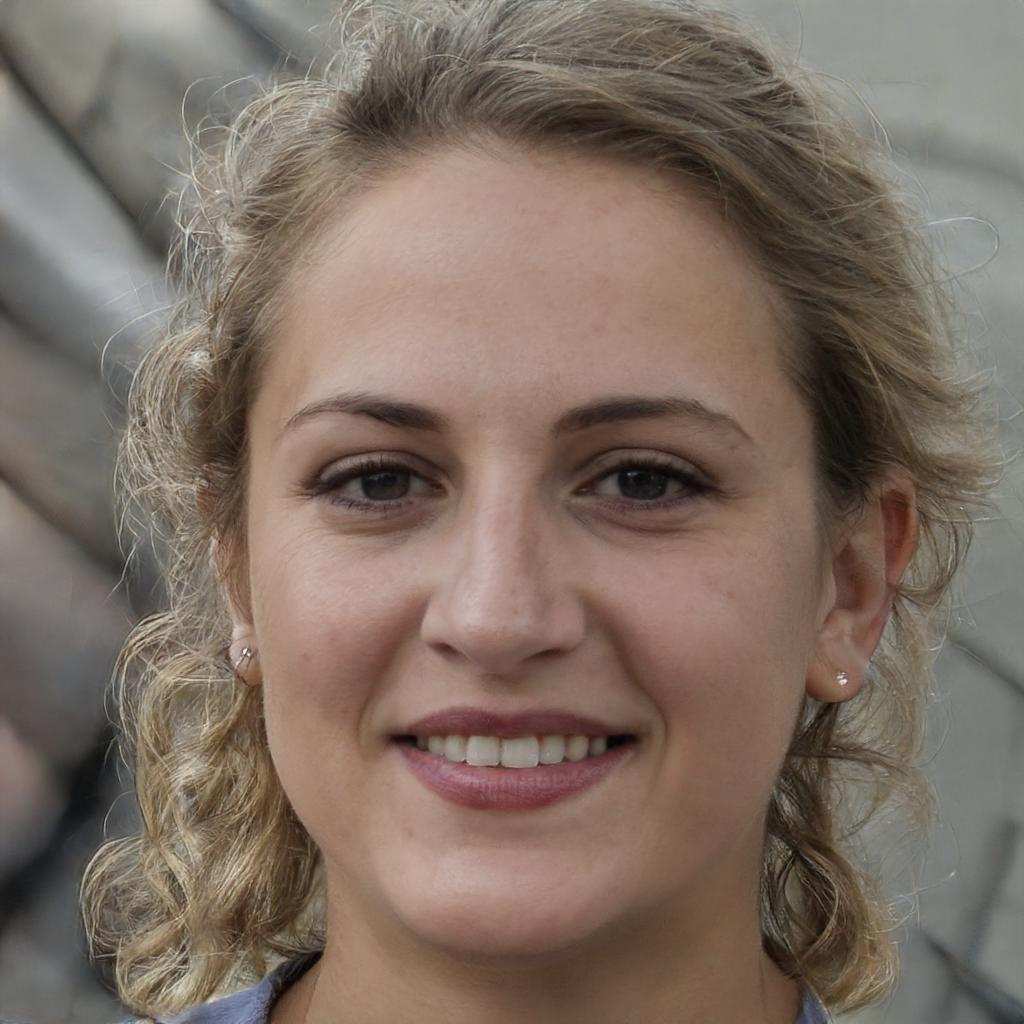} &
		\includegraphics[width=0.25\columnwidth]{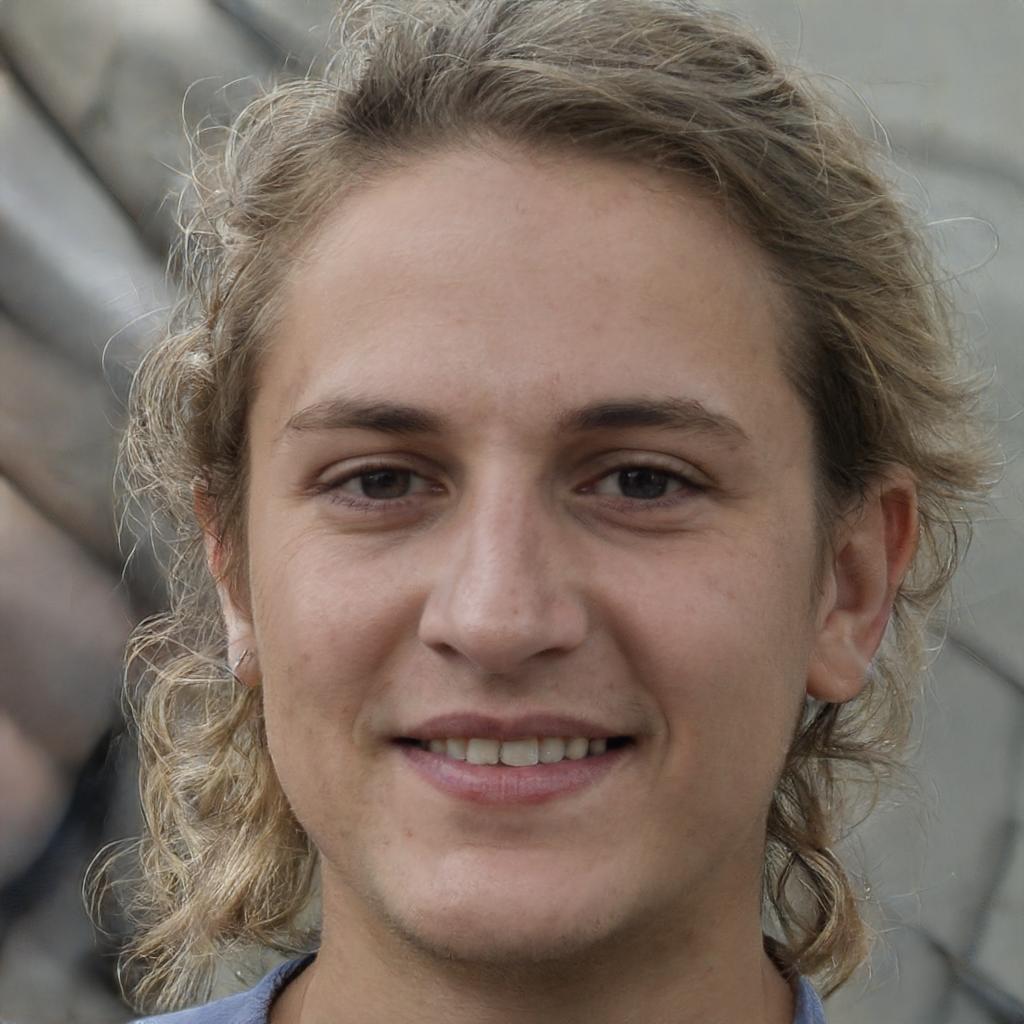} &
		\includegraphics[width=0.25\columnwidth]{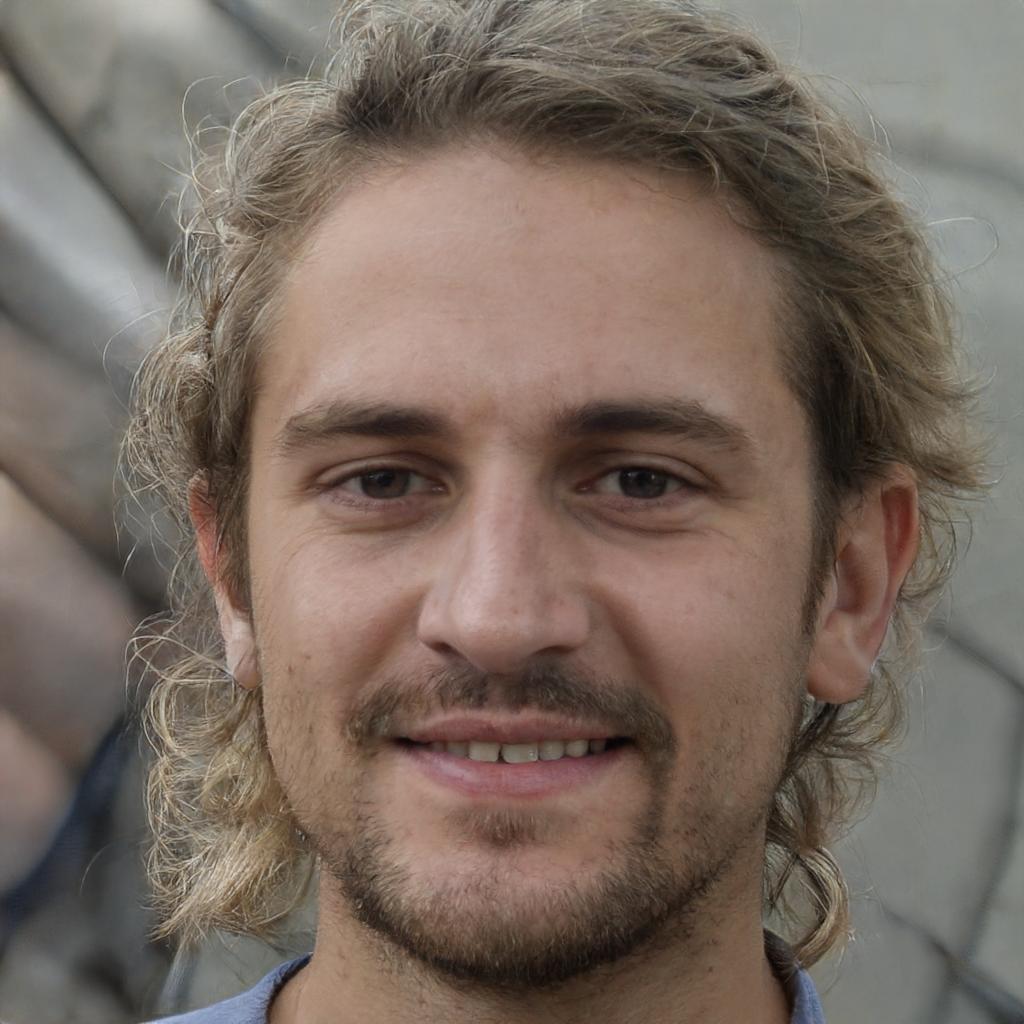} 
		\\
		\rotatebox{90}{\footnotesize \phantom{k} $\beta=0.20$} &
		\includegraphics[width=0.25\columnwidth]{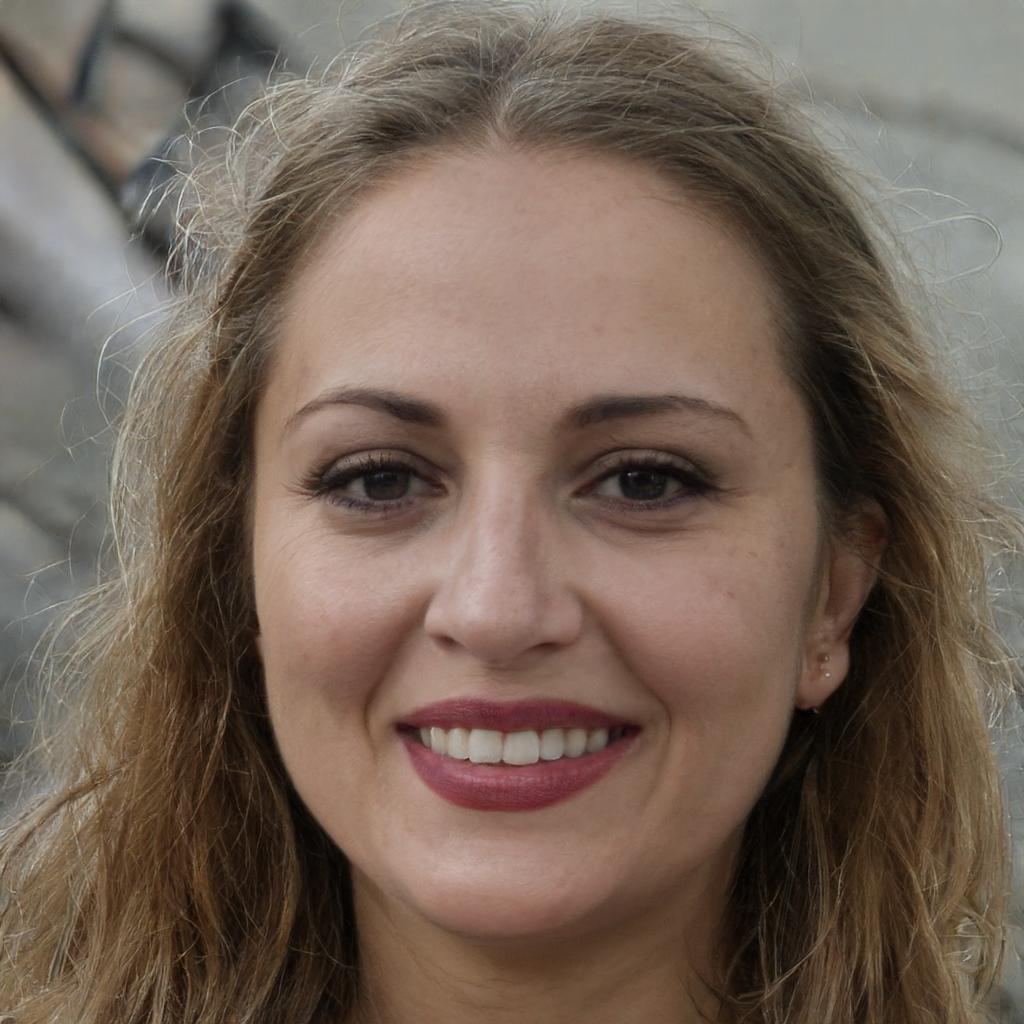} &
		\includegraphics[width=0.25\columnwidth]{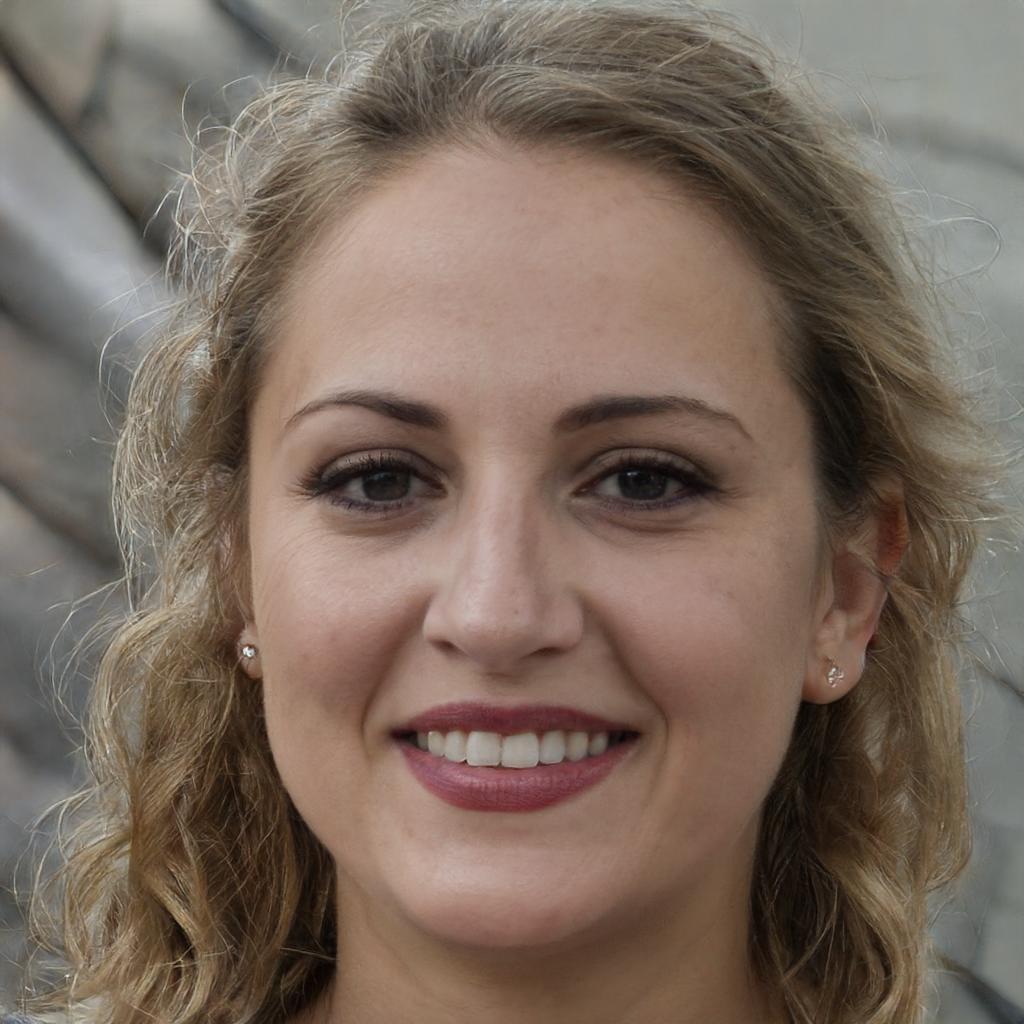} &
		\includegraphics[width=0.25\columnwidth]{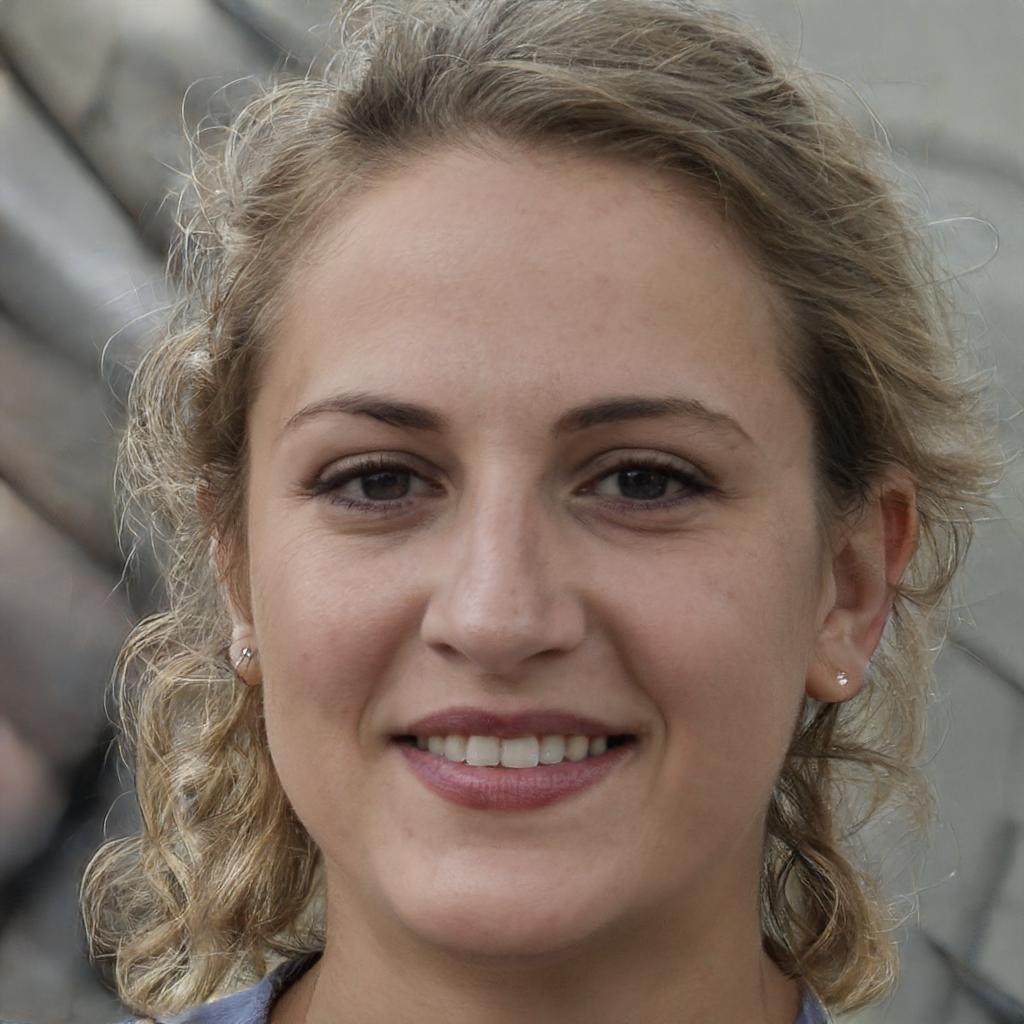} &
		\includegraphics[width=0.25\columnwidth]{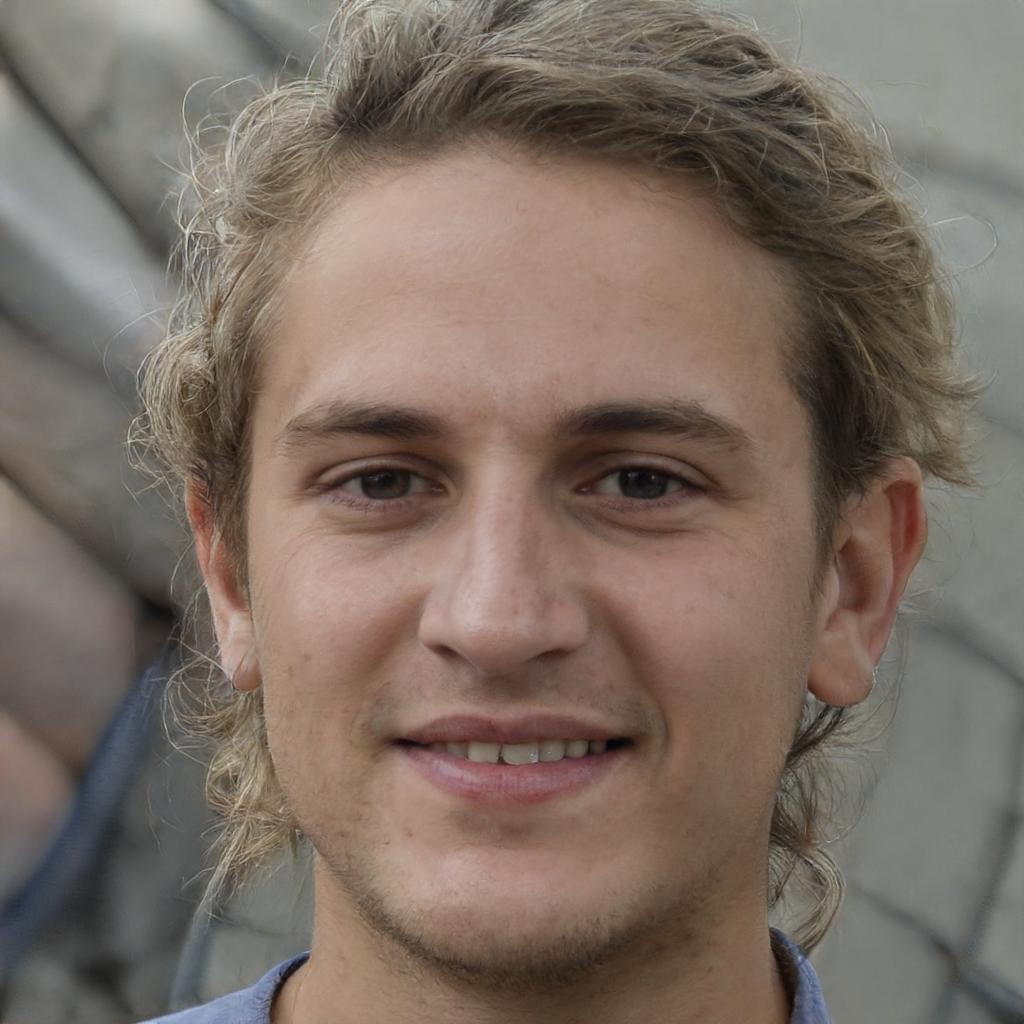} &
		\includegraphics[width=0.25\columnwidth]{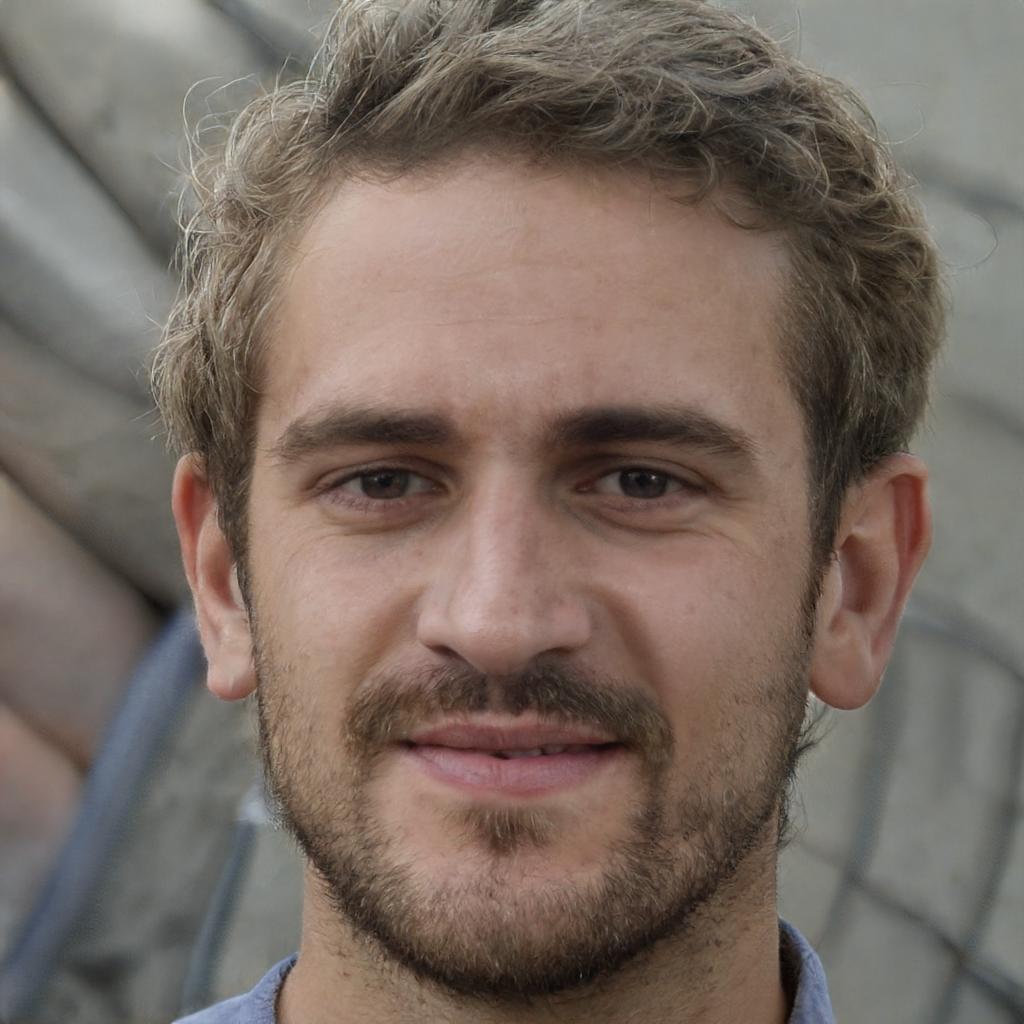} 
		\\
	\end{tabular}
	\caption{We demonstrate gender manipulation (driven by the prompt ``a photo of a male face'') for different manipulation strengths and disentanglement thresholds. Moving along the global direction, causes the facial features to become more masculine, while steps in the opposite direction yields more feminine features. The effect becomes stronger as the strength $\alpha$ increases. When the disentanglement threshold $\beta$ is high, only the facial features are affected, and as $\beta$ is lowered, additional correlated attributes, such as hair length and facial hair are affected as well.
	}
	\label{fig:disentanglement_strength2}
\end{figure*}

Figure~\ref{fig:disentanglement_strength2} shows image manipulations driven by the prompt ``a photo of a male face'' for different manipulation strengths and disentanglement thresholds. Moving along the global direction, causes the facial features to become more masculine, while steps in the opposite direction yields more feminine features. The effect becomes stronger as the strength $\alpha$ increases. When the disentanglement threshold $\beta$ is high, only the facial features are affected, and as $\beta$ is lowered, additional correlated attributes, such as hair length and facial hair are affected as well.

\begin{figure*}[tb]
	\centering
	\setlength{\tabcolsep}{1pt}	
	\begin{tabular}{cccccc}
		& {\footnotesize Original} & {\footnotesize GANSpace} & {\footnotesize InterfaceGAN} & {\footnotesize StyleSpace} & {\footnotesize Ours} \\
		
		\rotatebox{90}{\footnotesize \phantom{kk}Gender} &
		\includegraphics[width=0.25\columnwidth]{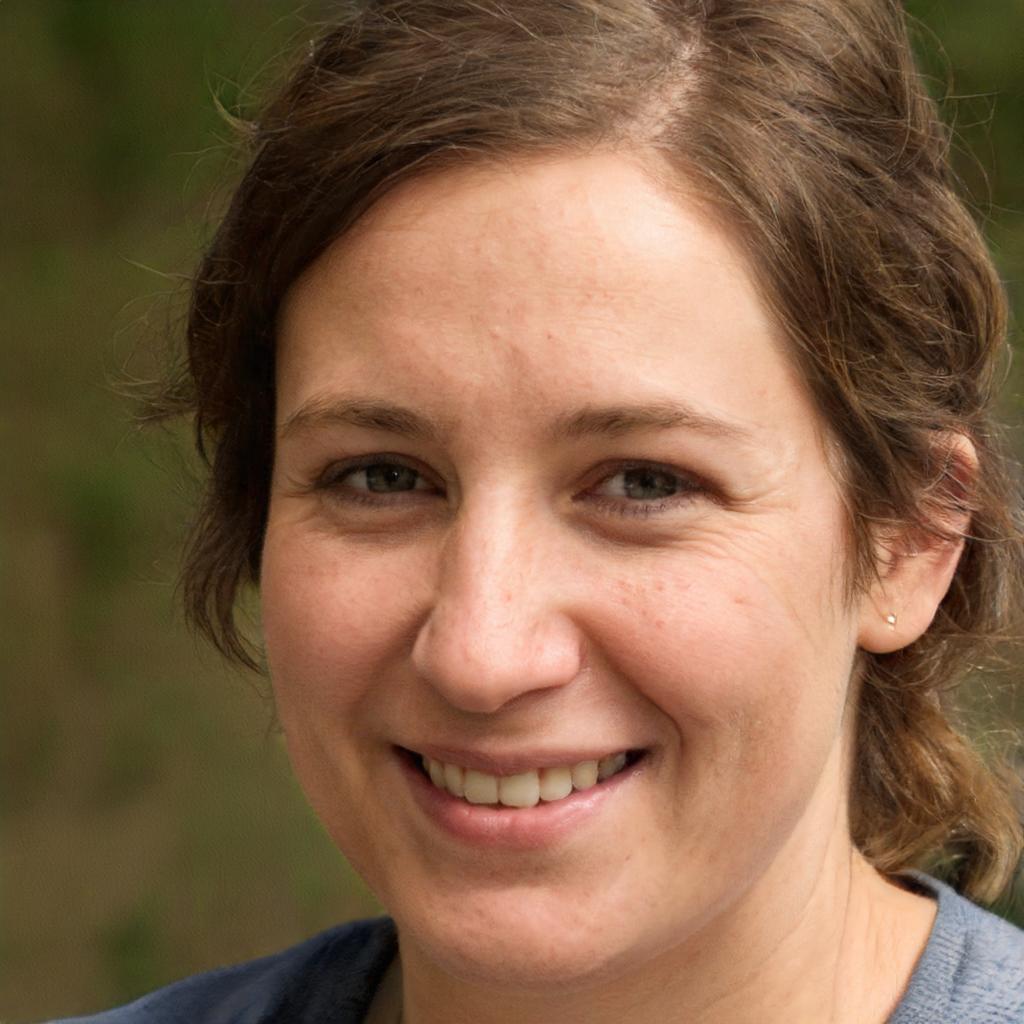} &
		\includegraphics[width=0.25\columnwidth]{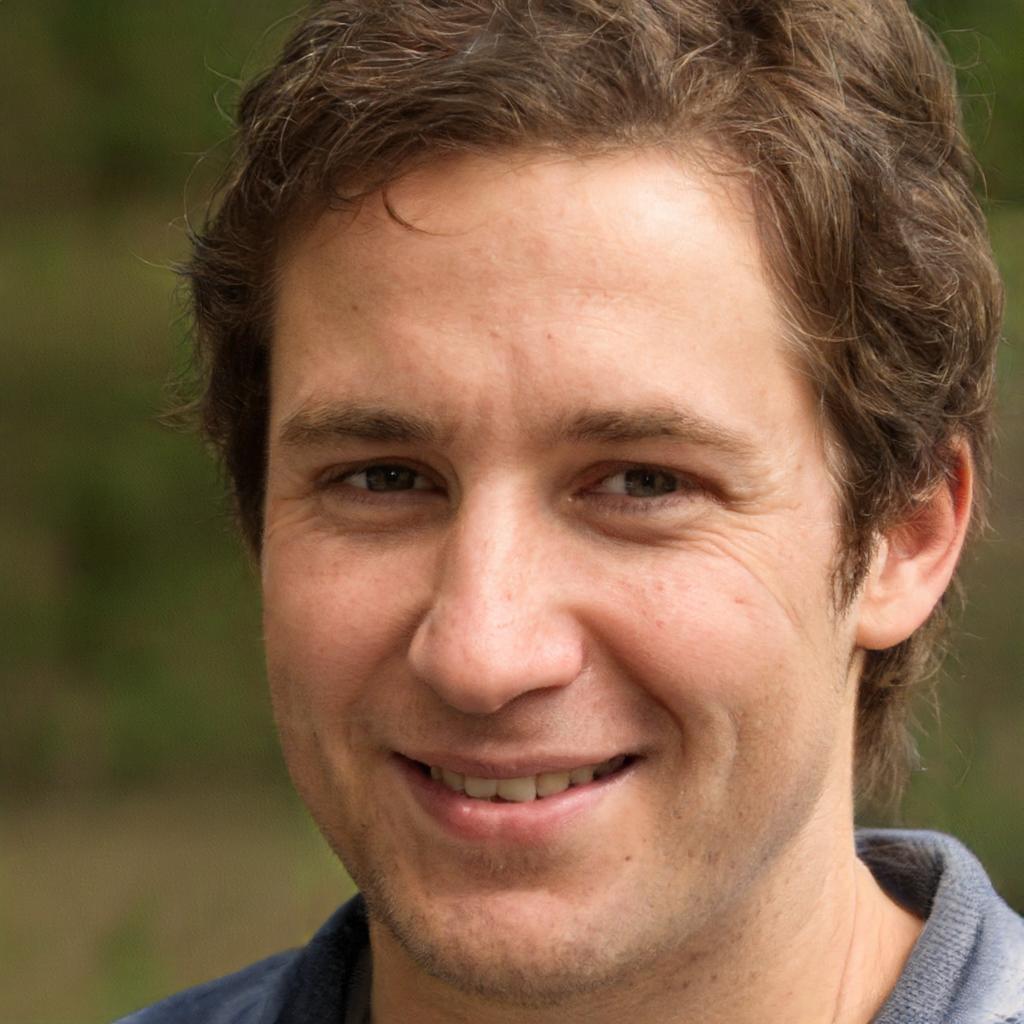} &
		\includegraphics[width=0.25\columnwidth]{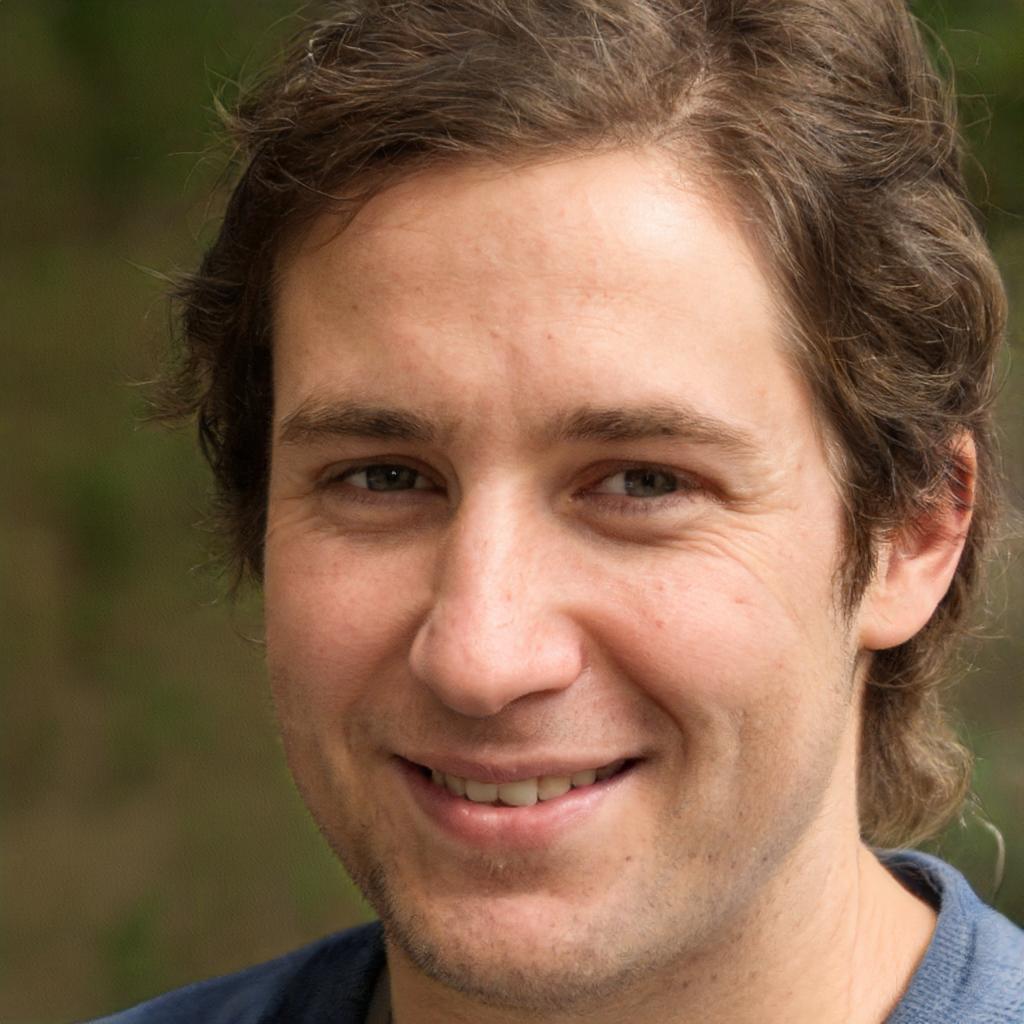} &
		\includegraphics[width=0.25\columnwidth]{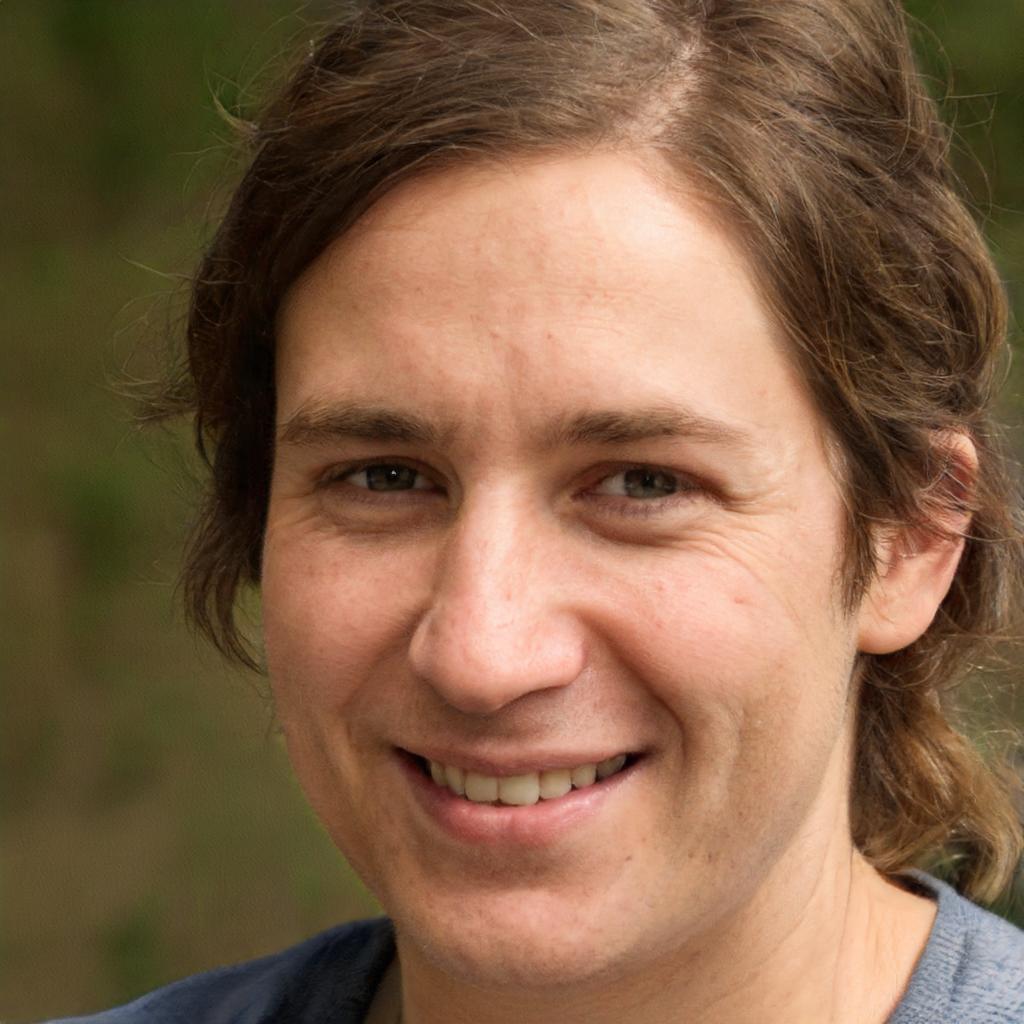} &
		\includegraphics[width=0.25\columnwidth]{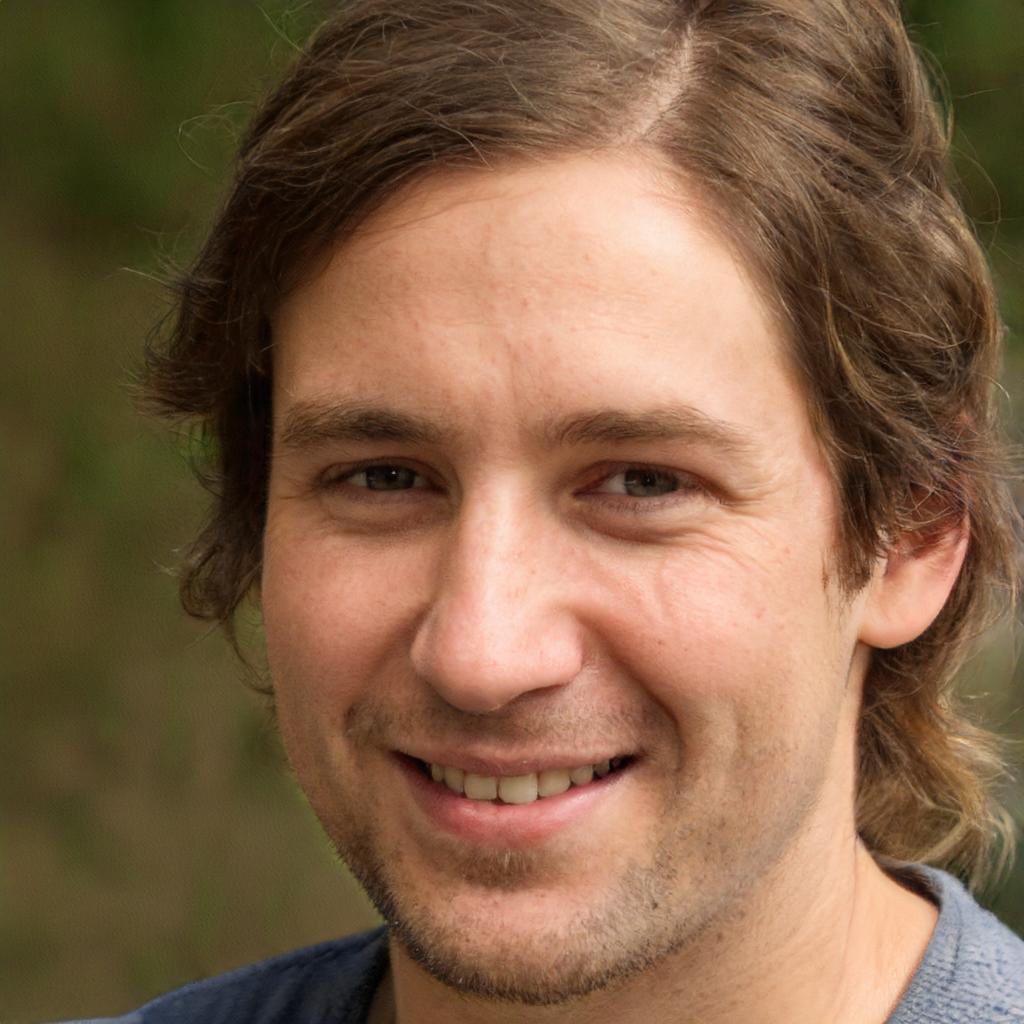} 
		\\
		
		\rotatebox{90}{\footnotesize \phantom{k}Gray hair} &
		\includegraphics[width=0.25\columnwidth]{./fig3/original.jpg} &
		\includegraphics[width=0.25\columnwidth]{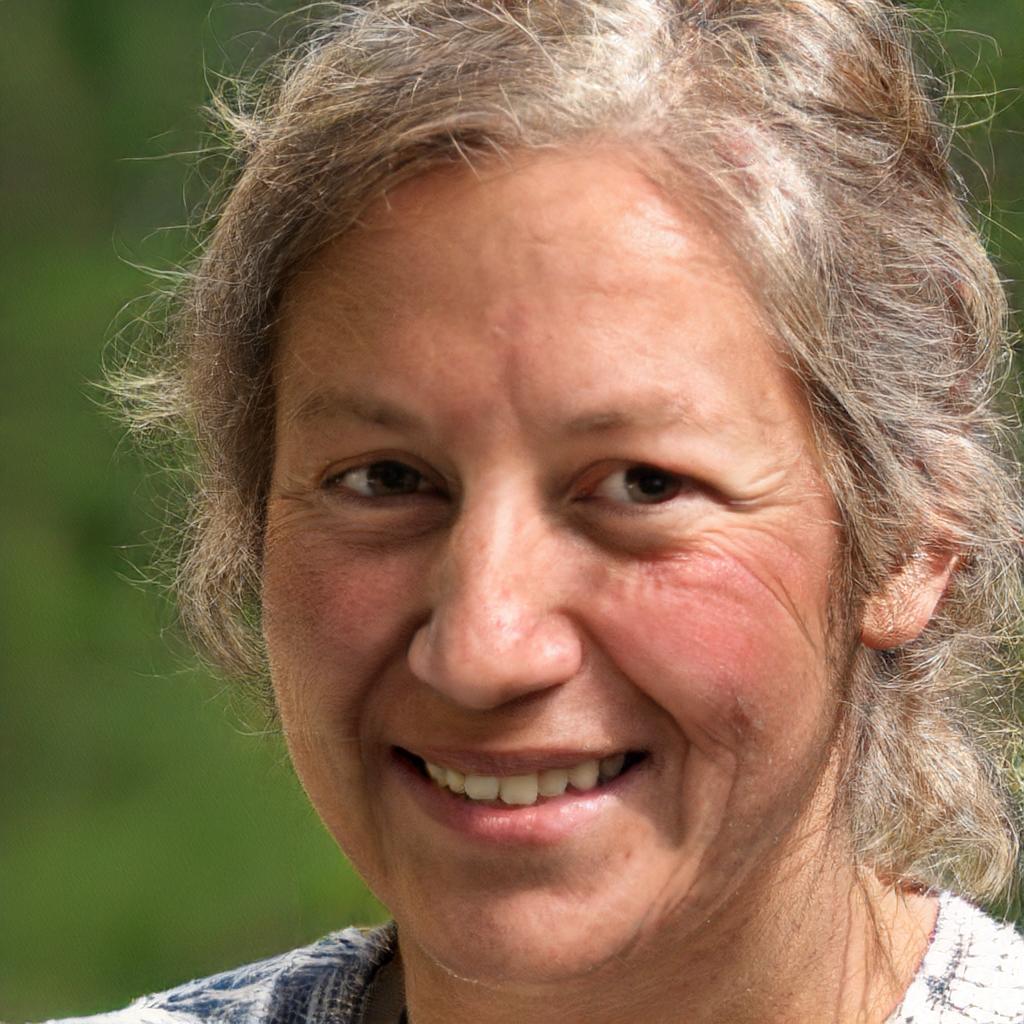} &
		\includegraphics[width=0.25\columnwidth]{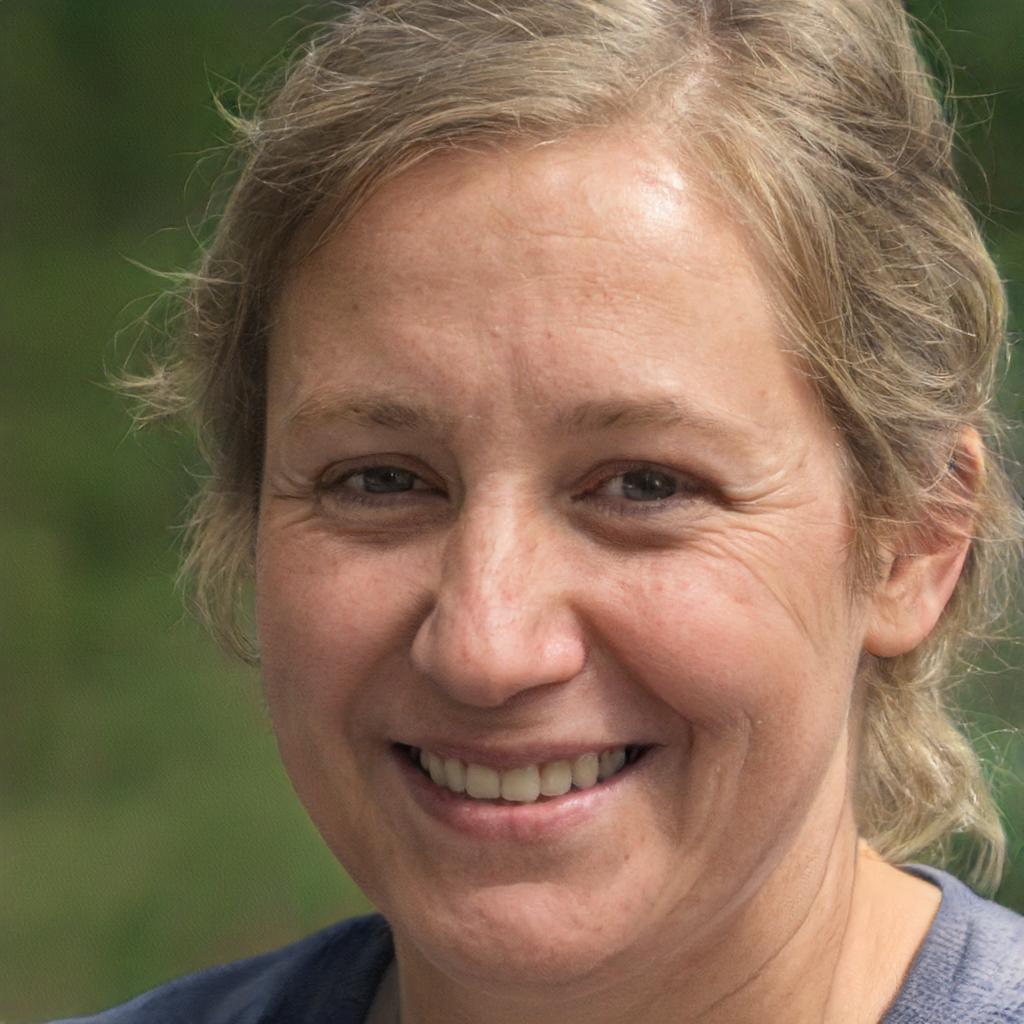} &
		\includegraphics[width=0.25\columnwidth]{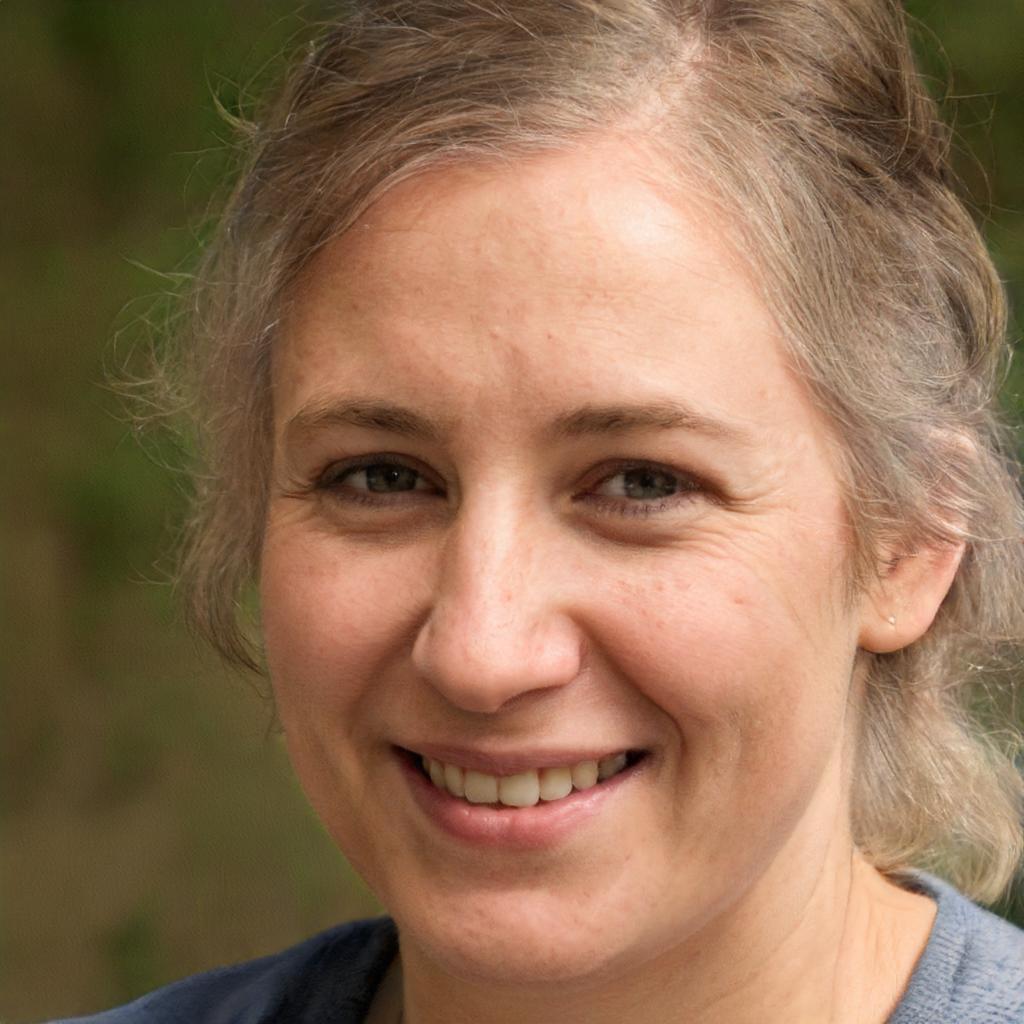}&
		\includegraphics[width=0.25\columnwidth]{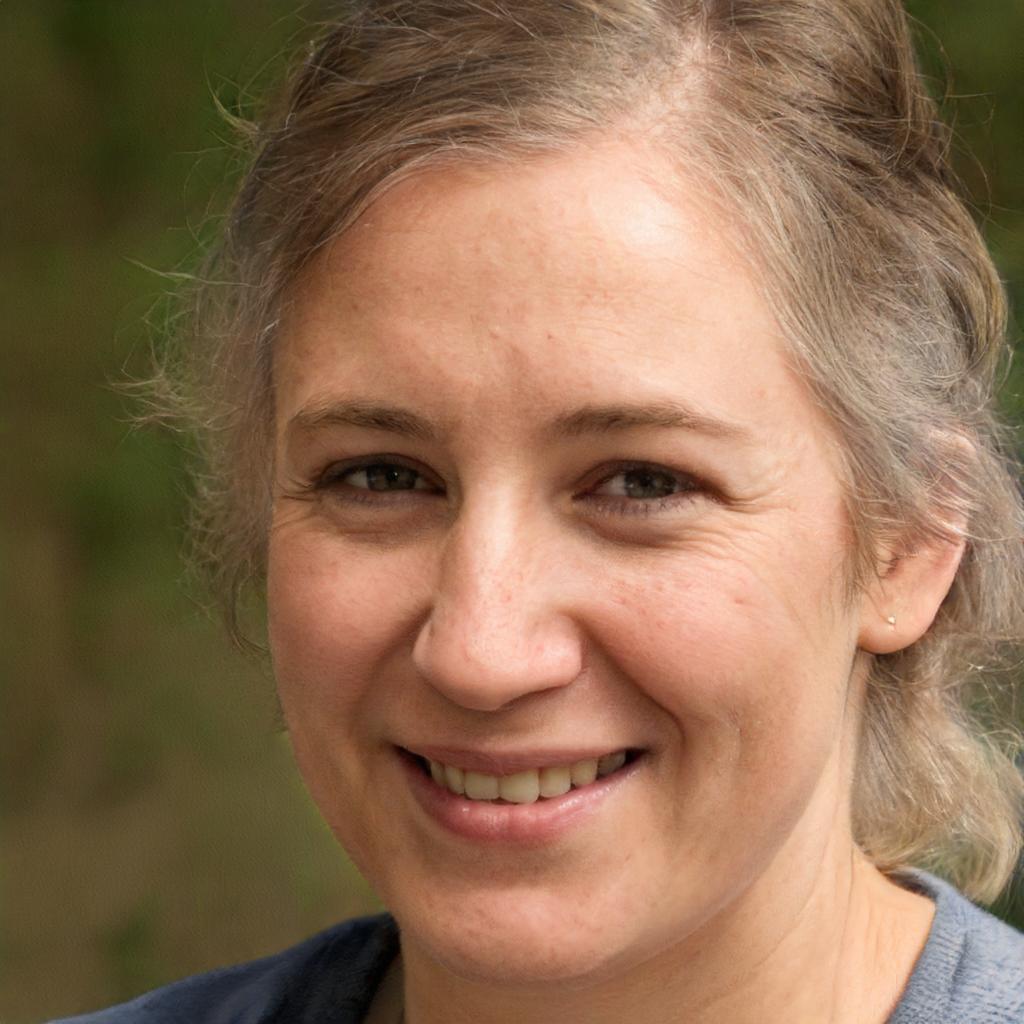} 
		\\
		
		\rotatebox{90}{\footnotesize \phantom{kk}Lipstick} &
		\includegraphics[width=0.25\columnwidth]{./fig3/original.jpg} &
		\includegraphics[width=0.25\columnwidth]{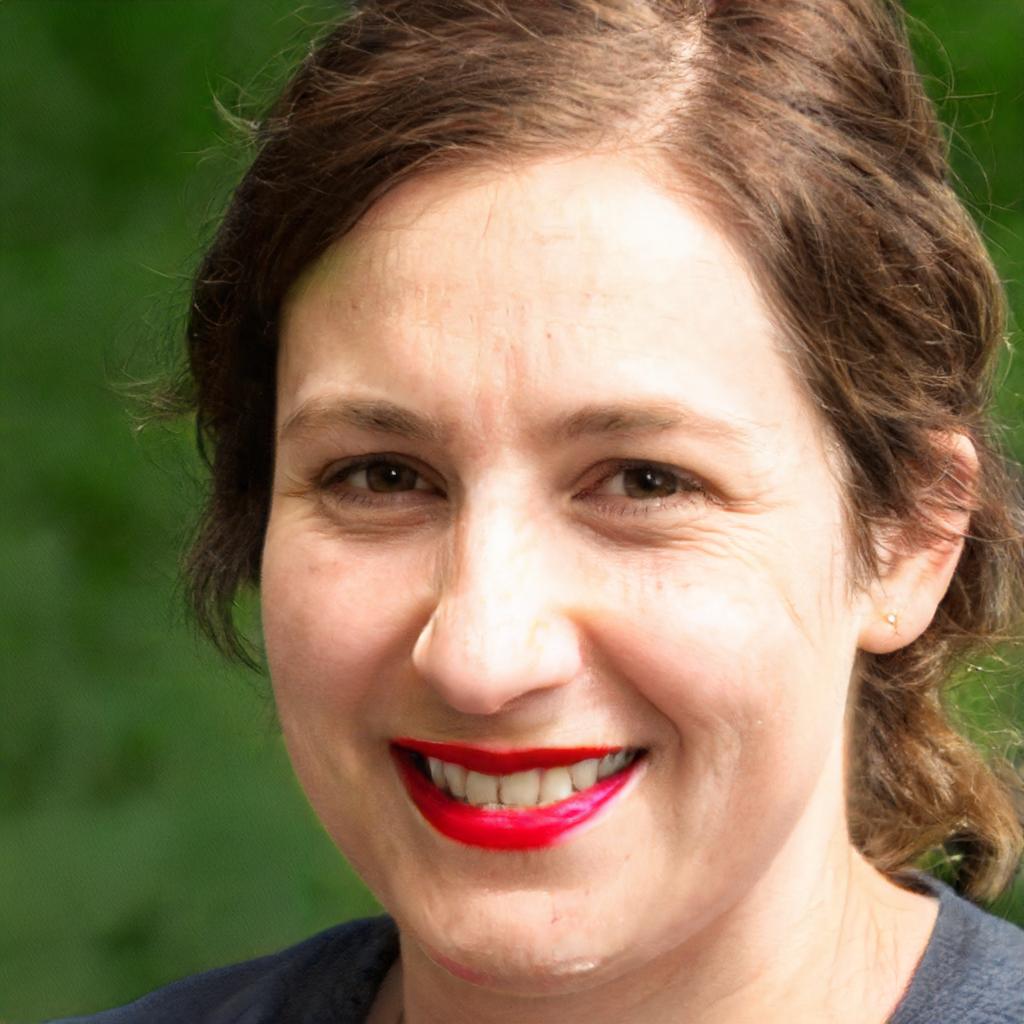} &
		\includegraphics[width=0.25\columnwidth]{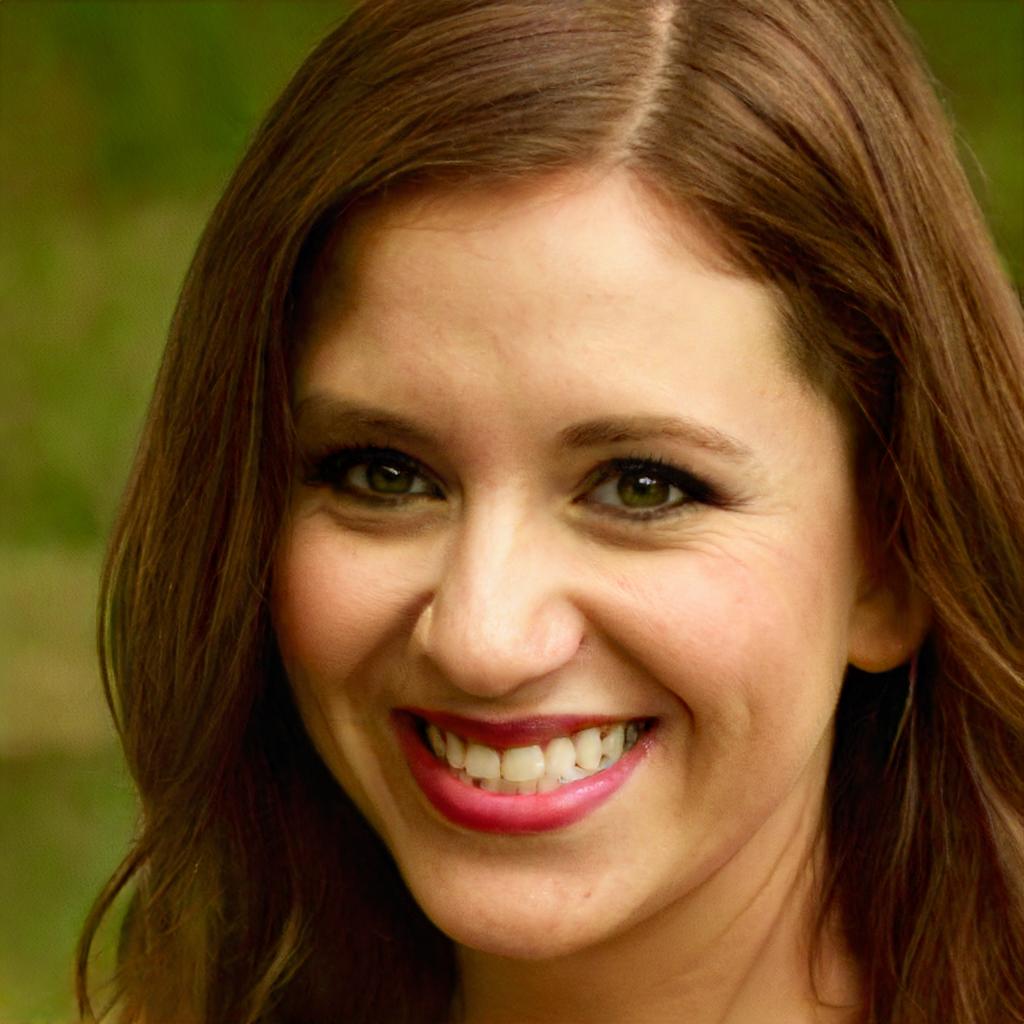} &
		\includegraphics[width=0.25\columnwidth]{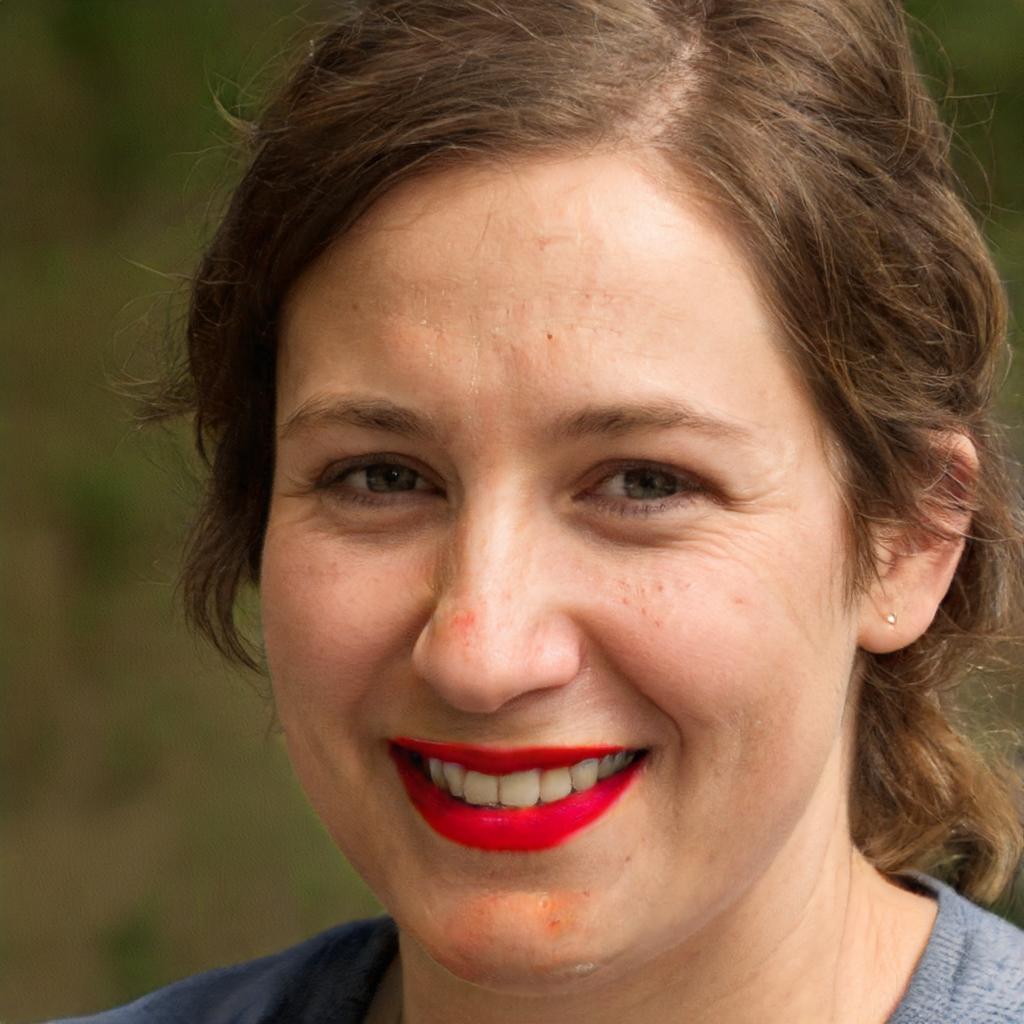} &
		\includegraphics[width=0.25\columnwidth]{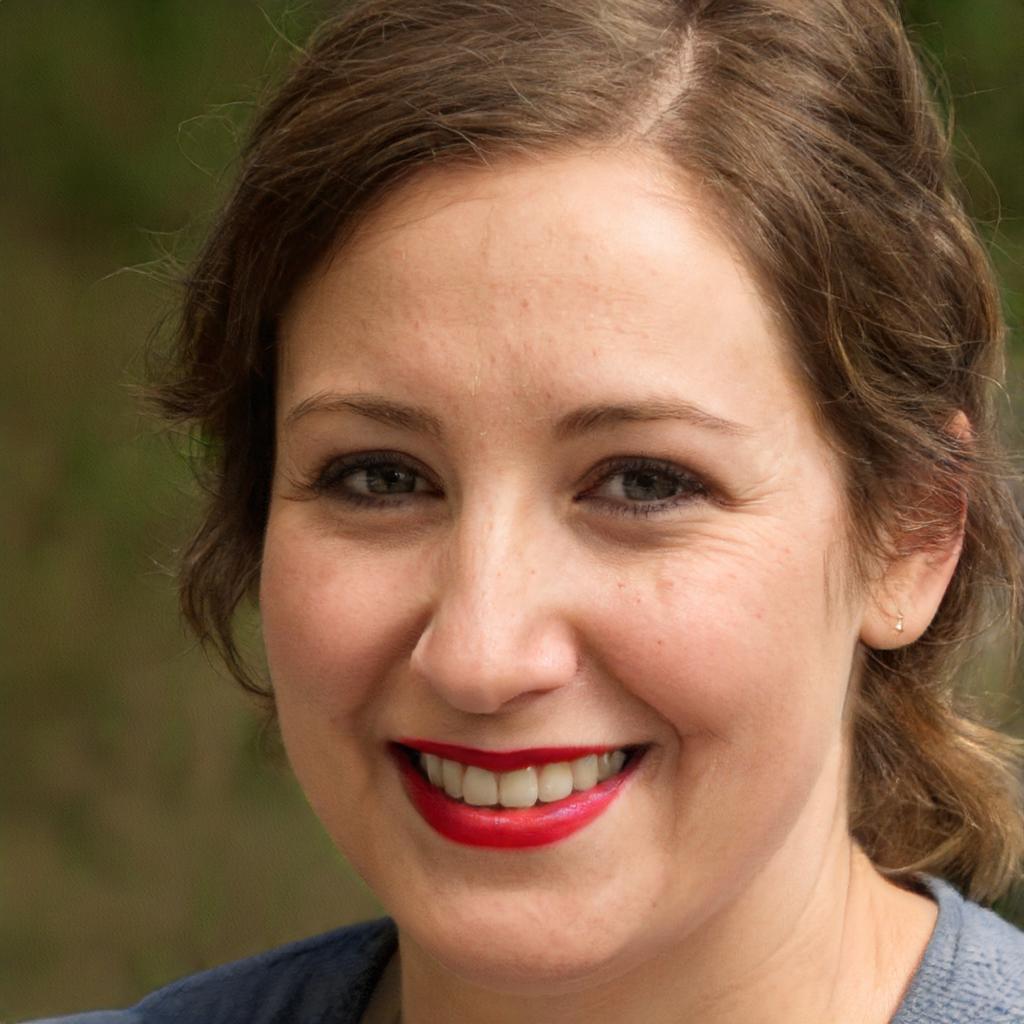} 
	\end{tabular}
	
	\caption{Comparison with state-of-the-art methods using the same amount of manipulation according to a pretrained attribute classifier.
	}
	\label{fig:compare_linear2}
\end{figure*}

In Figure~\ref{fig:compare_linear2}, we show another comparison between our global direction method and several state-of-the-art StyleGAN image manipulation methods: GANSpace~\cite{harkonen2020ganspace}, InterFaceGAN \cite{shen2020interfacegan}, and StyleSpace \cite{wu2020stylespace}.
The comparison only examines the attributes which all of the compared methods are able to manipulate (Gender, Grey hair, and Lipstick), and thus it does not include the many novel manipulations enabled by our approach.
Following Wu \etal~\cite{wu2020stylespace}, the manipulation step strength is chosen such that it induces the same amount of change in the logit value of the corresponding classifiers (pretrained on CelebA).
It may be seen that in GANSpace~\cite{harkonen2020ganspace} manipulation is entangled with skin color and lighting, while in InterFaceGAN~\cite{shen2020interfacegan} the identity may change significantly (when manipulating Lipstick). Our manipulation is very similar to StyleSpace \cite{wu2020stylespace}, which only changes the target attribute, while all other attributes remain the same. 

\begin{figure*}[tb]
    
	\setlength{\tabcolsep}{1pt}	
	\centering
	\begin{tabular}{ccccc}
		& {\footnotesize Original} & {\footnotesize Beard} & {\footnotesize Glasses} & {\footnotesize Bald}   \\
		
		\rotatebox{90}{\footnotesize \phantom{kk}Global} &
		\includegraphics[width=0.34\columnwidth]{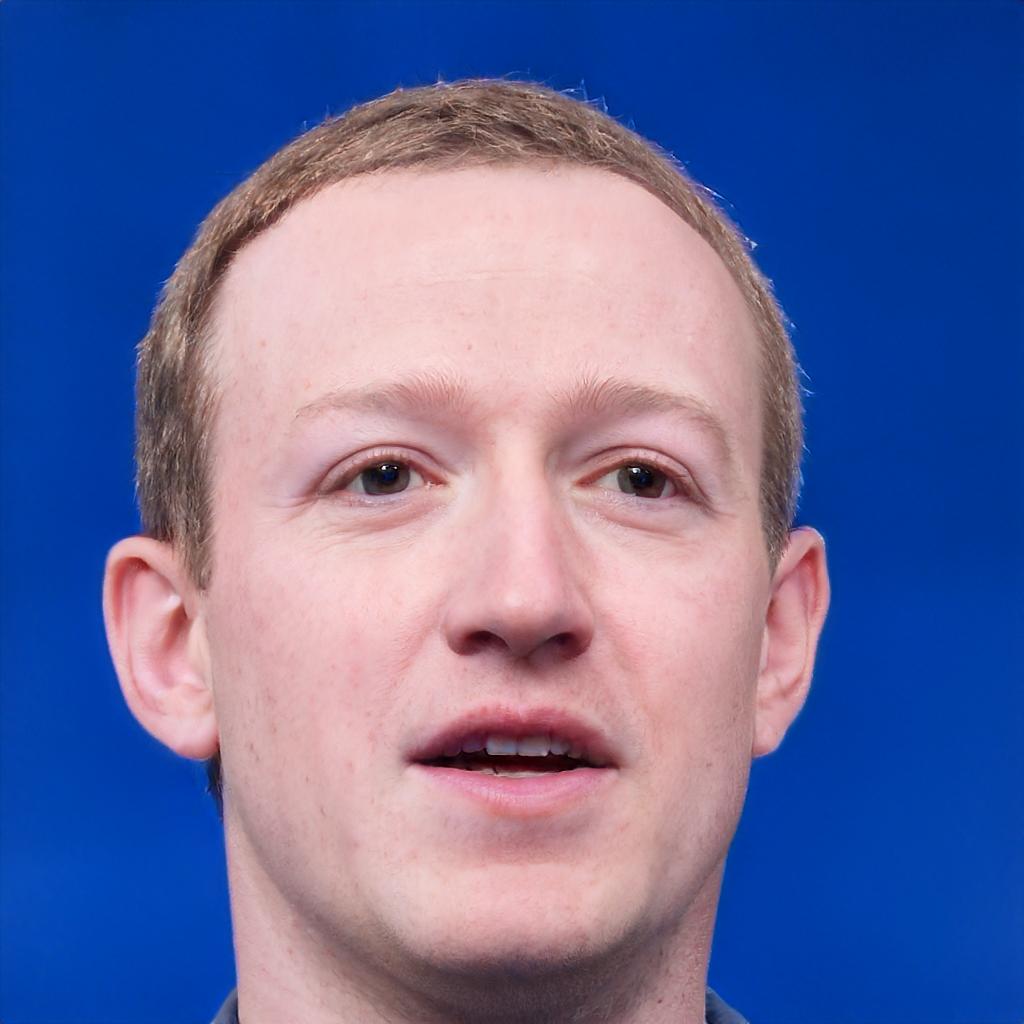} &
		\includegraphics[width=0.34\columnwidth]{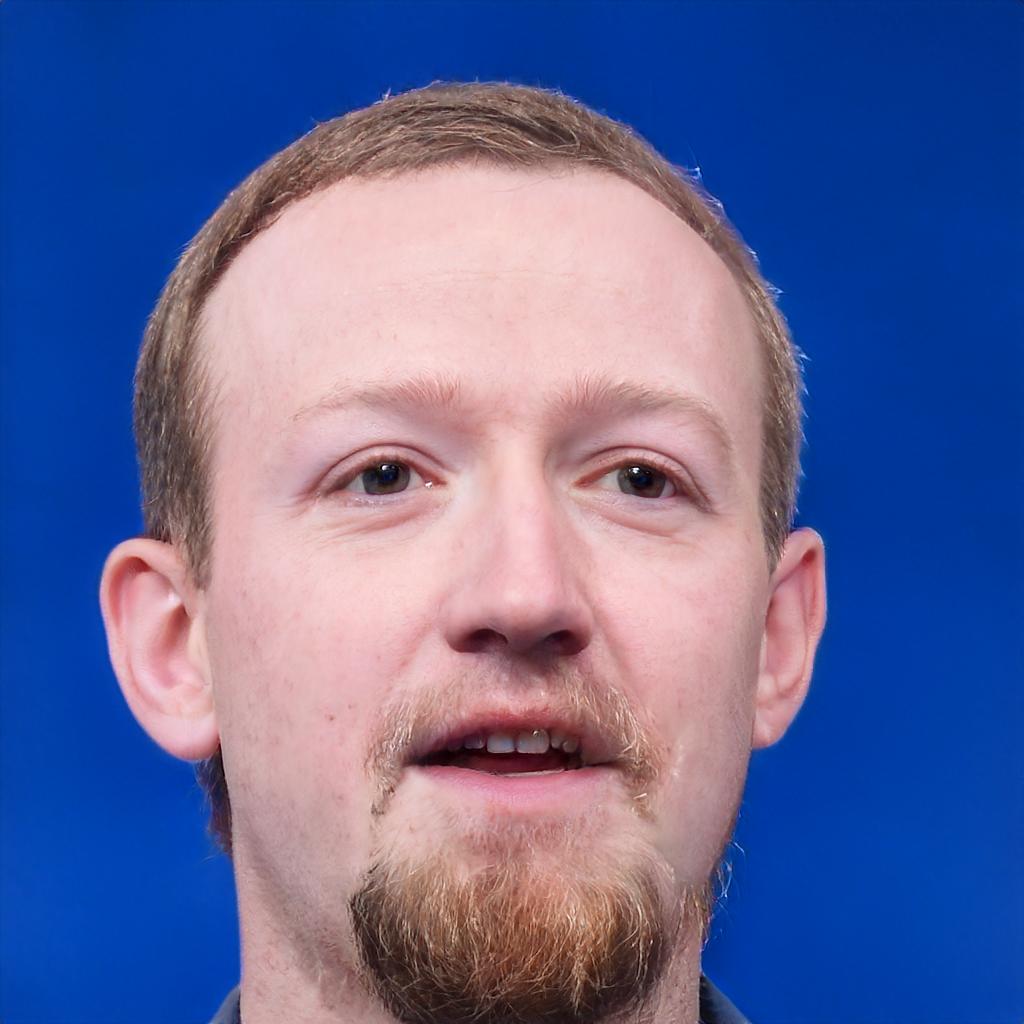} &
		\includegraphics[width=0.34\columnwidth]{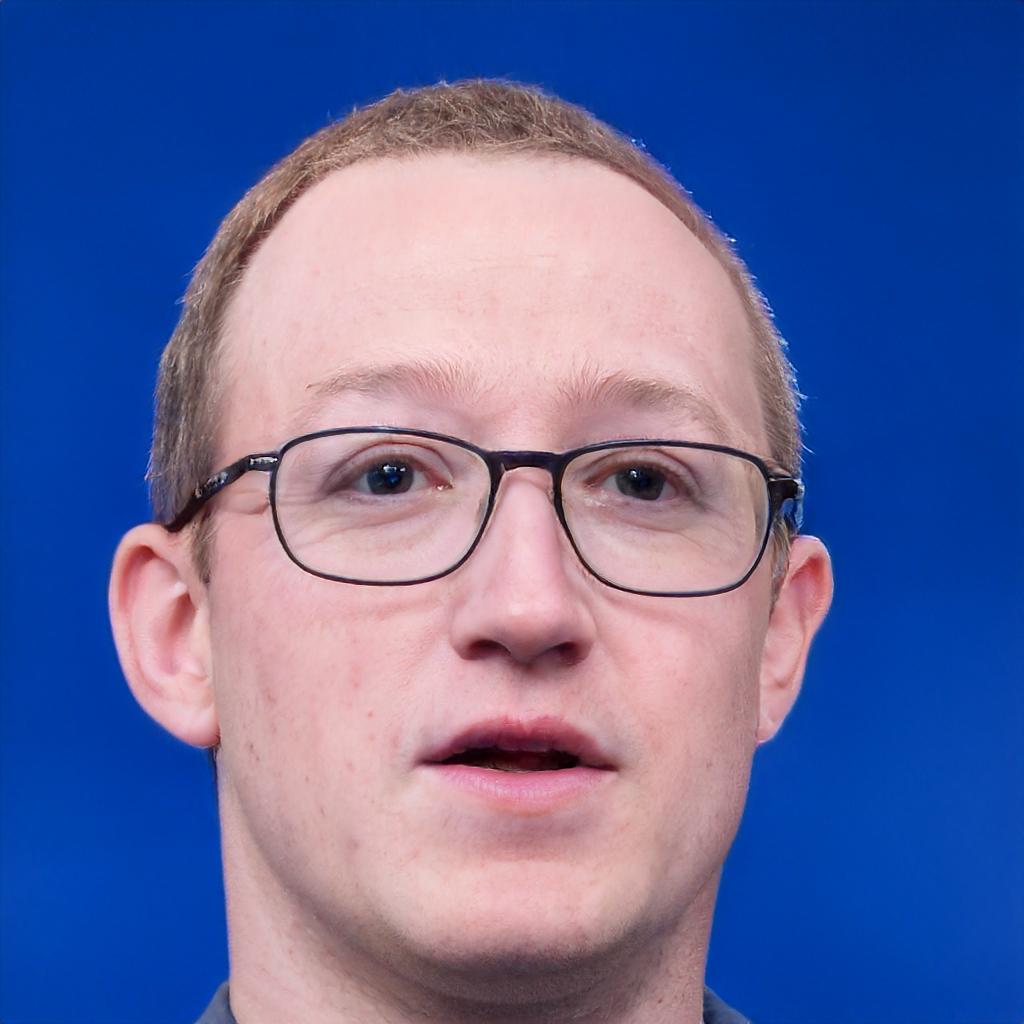} &
	    \includegraphics[width=0.34\columnwidth]{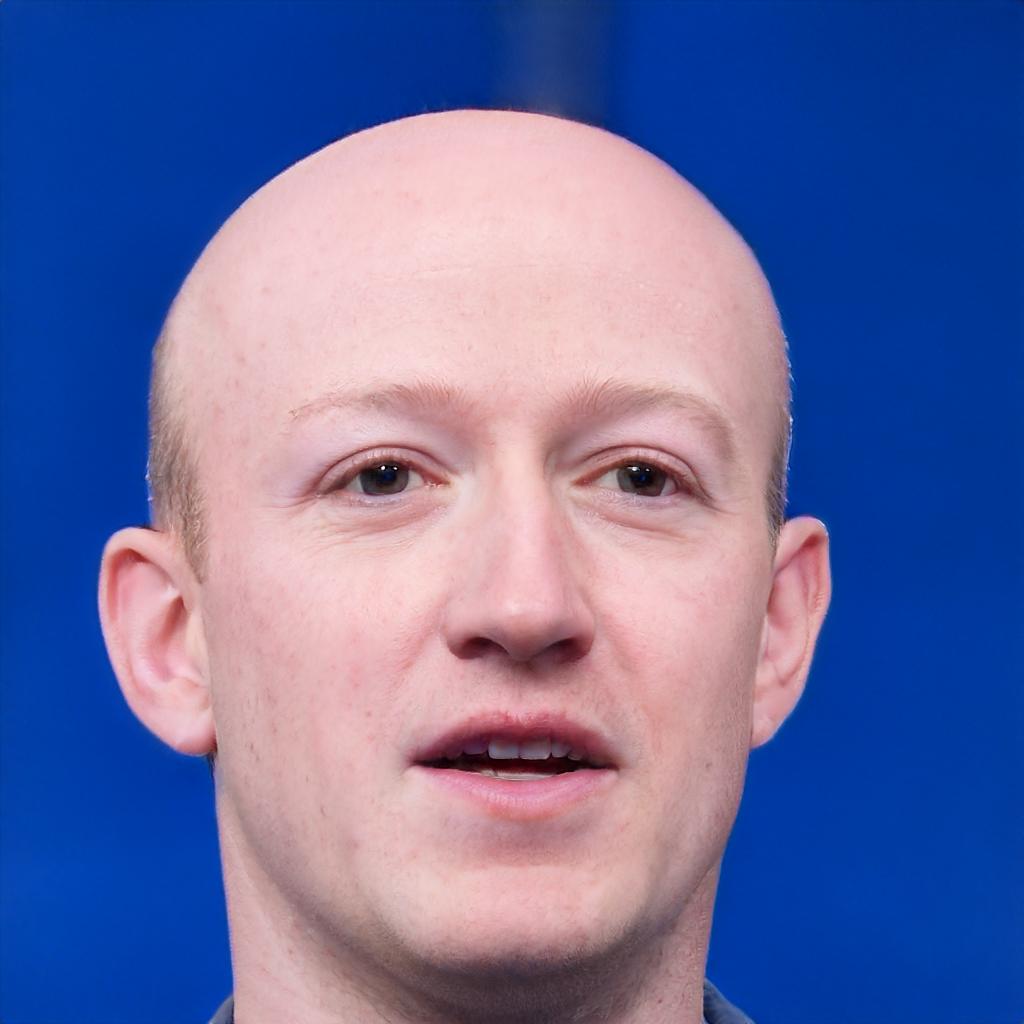} \\
	    
	    \rotatebox{90}{\footnotesize \phantom{kk}StyleFlow} &
		\includegraphics[width=0.34\columnwidth]{./compare/styleflow/0_0.jpg} &
		\includegraphics[width=0.34\columnwidth]{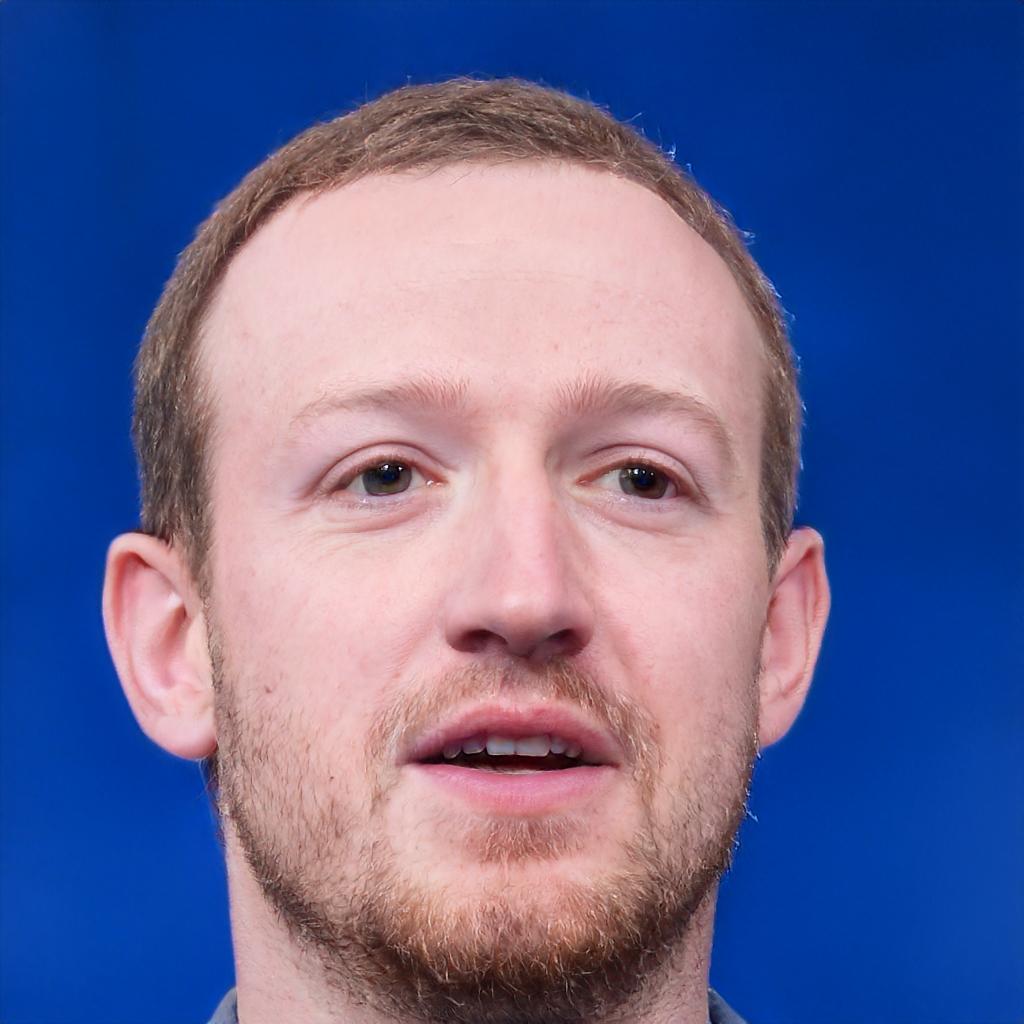} &
		\includegraphics[width=0.34\columnwidth]{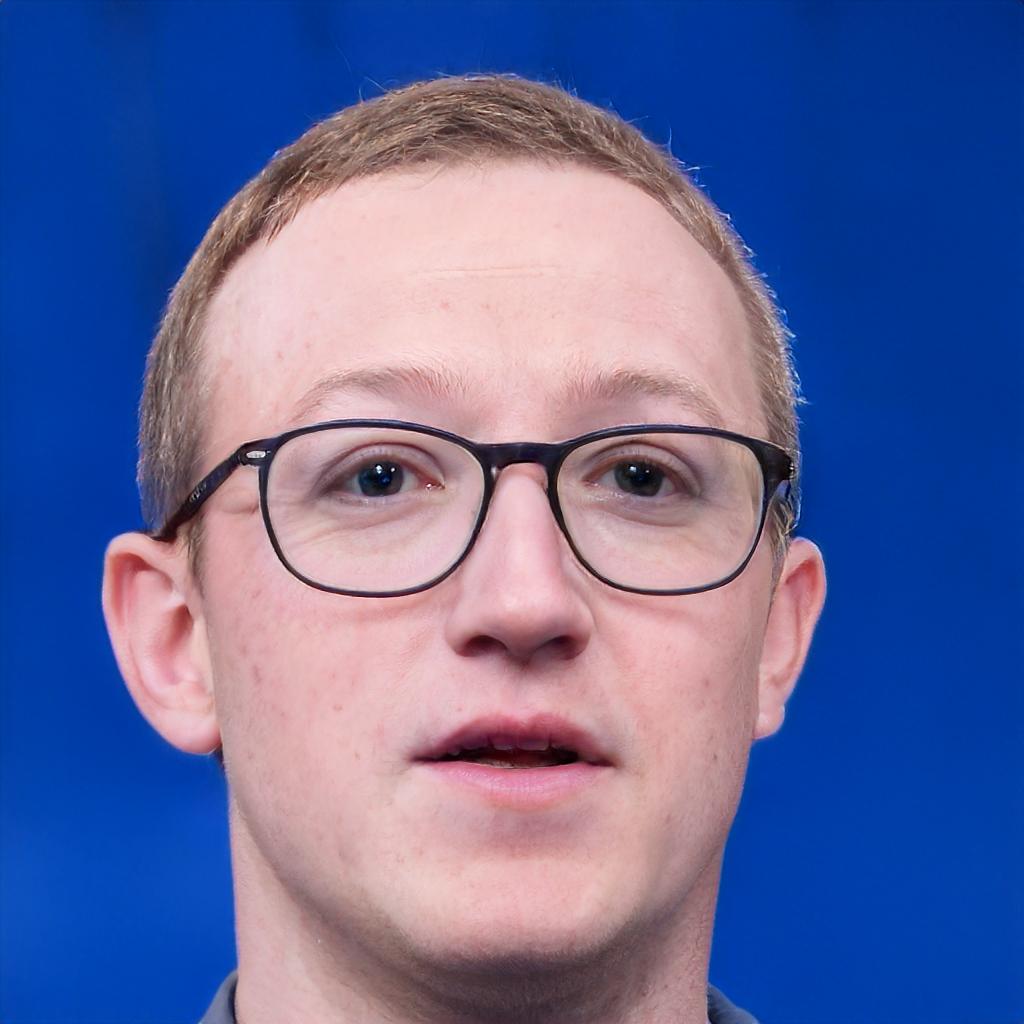} &
	    \includegraphics[width=0.34\columnwidth]{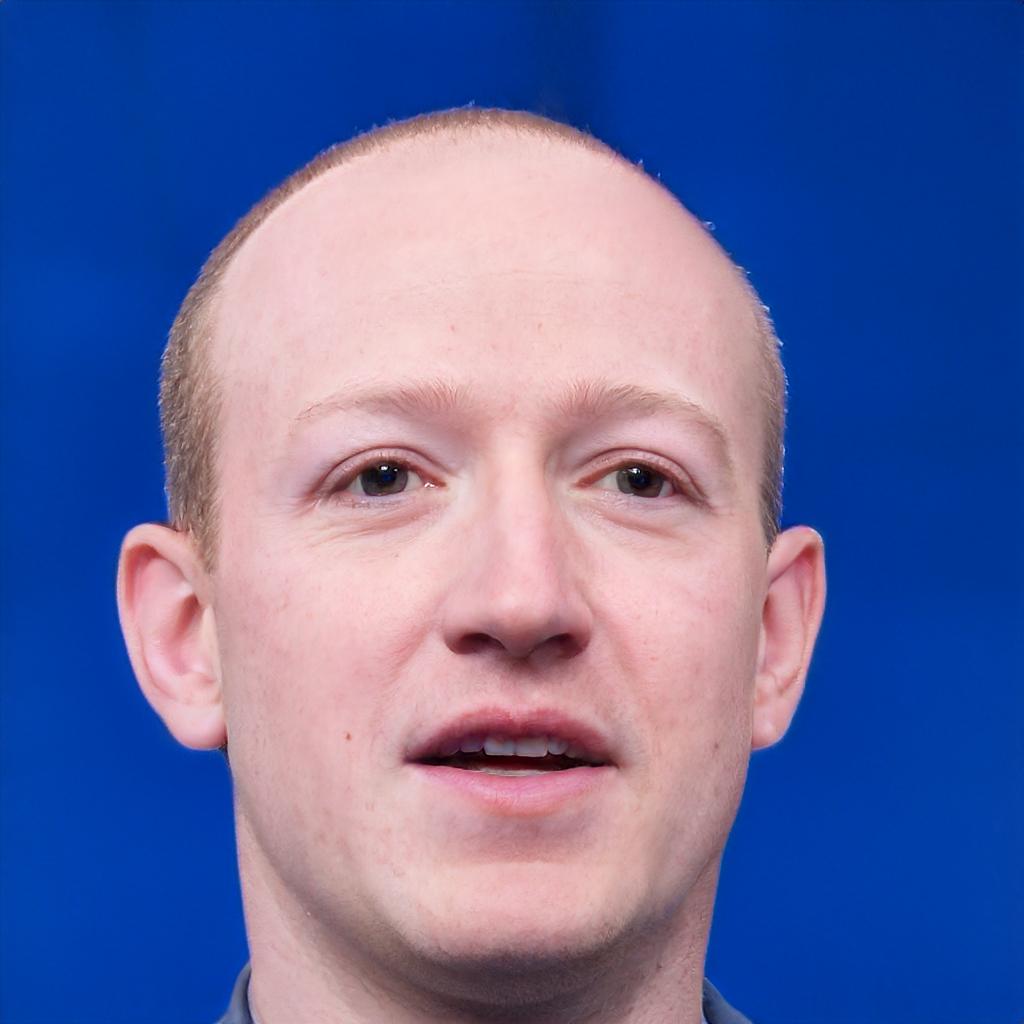} \\
	    
	    \\
	    \rotatebox{90}{\footnotesize \phantom{kk}Global} &
		\includegraphics[width=0.34\columnwidth]{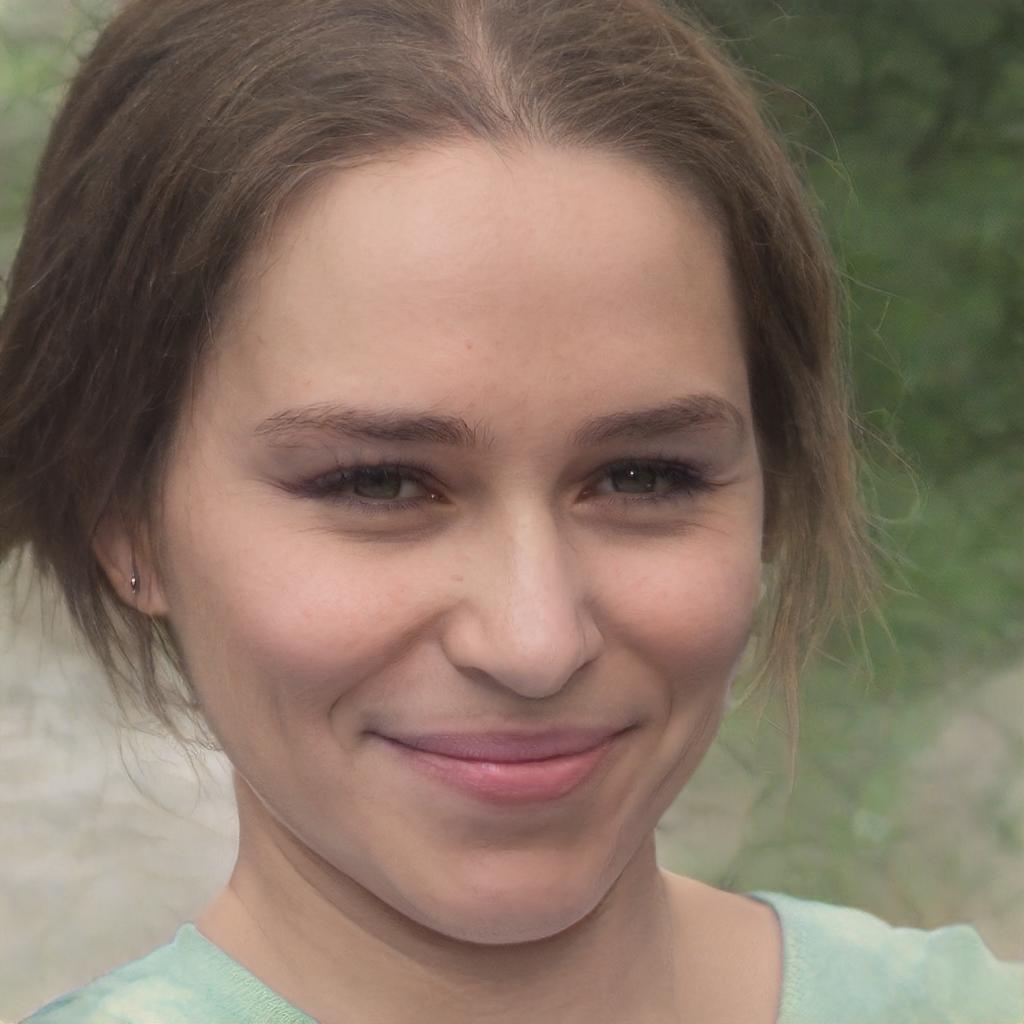} &
		\includegraphics[width=0.34\columnwidth]{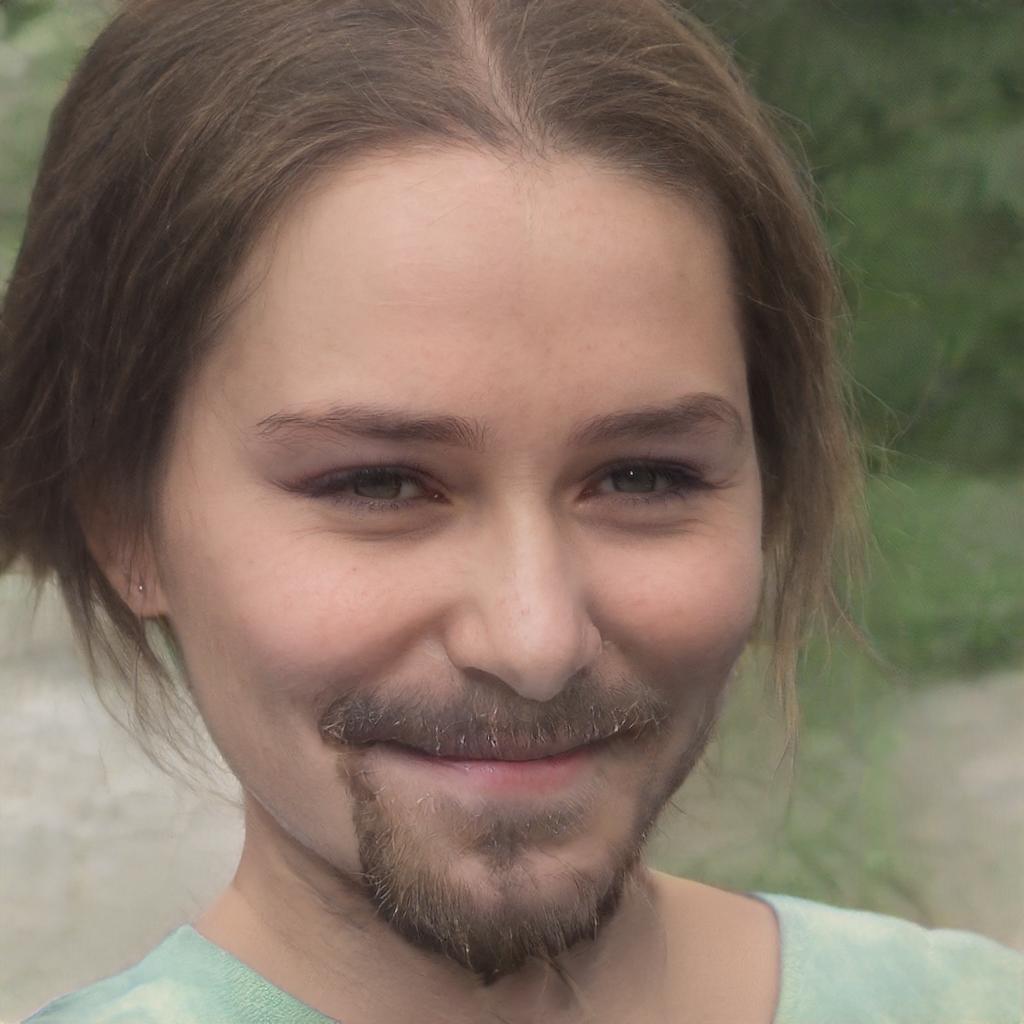} &
		\includegraphics[width=0.34\columnwidth]{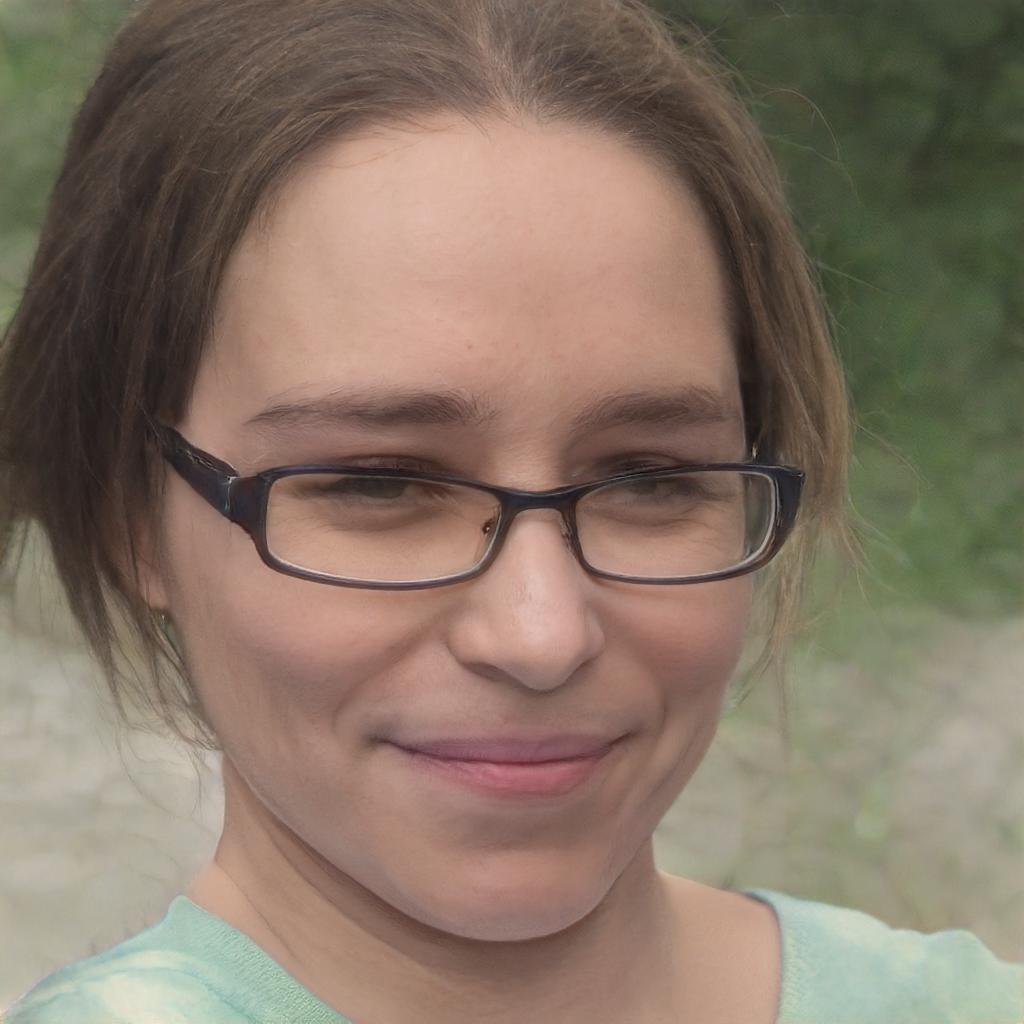} &
	    \includegraphics[width=0.34\columnwidth]{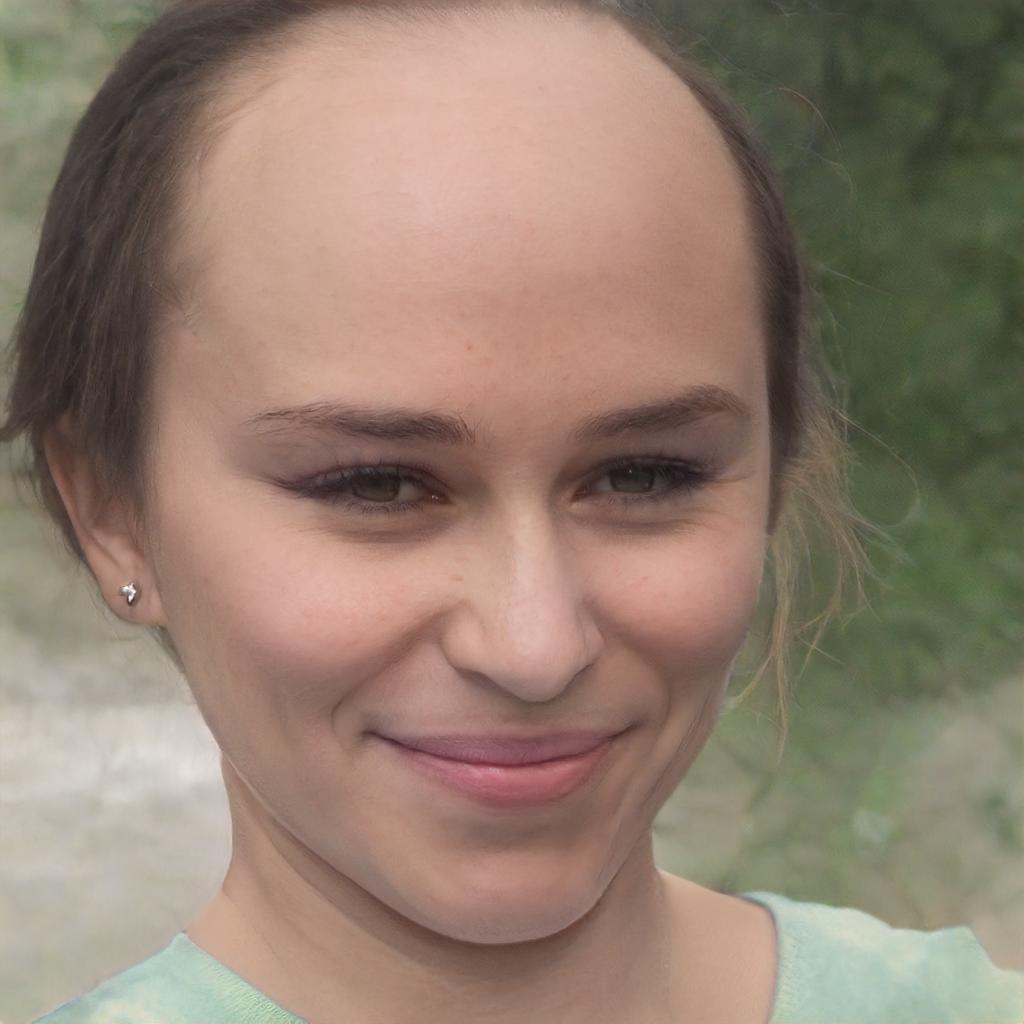} \\
	    
	    \rotatebox{90}{\footnotesize \phantom{kk}StyleFlow} &
		\includegraphics[width=0.34\columnwidth]{./compare/styleflow/1_0.jpg} &
		\includegraphics[width=0.34\columnwidth]{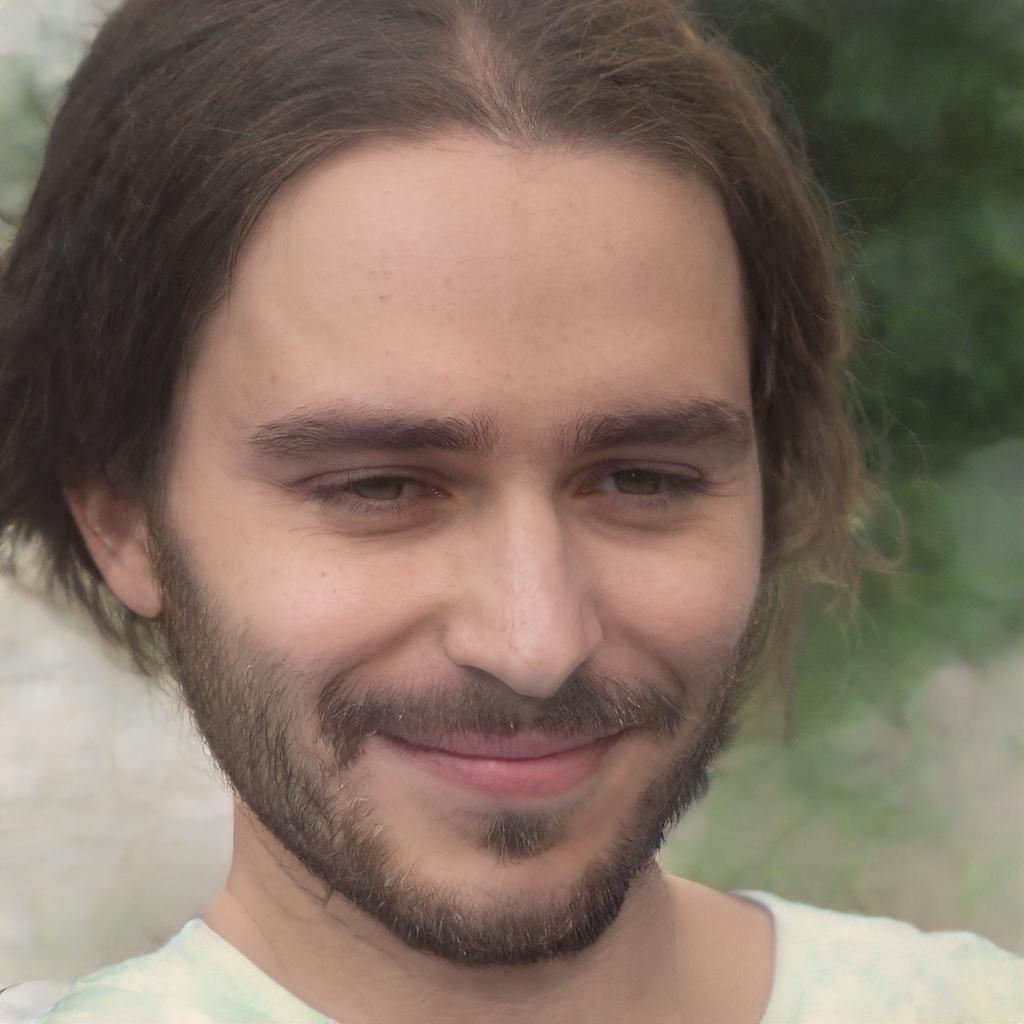} &
		\includegraphics[width=0.34\columnwidth]{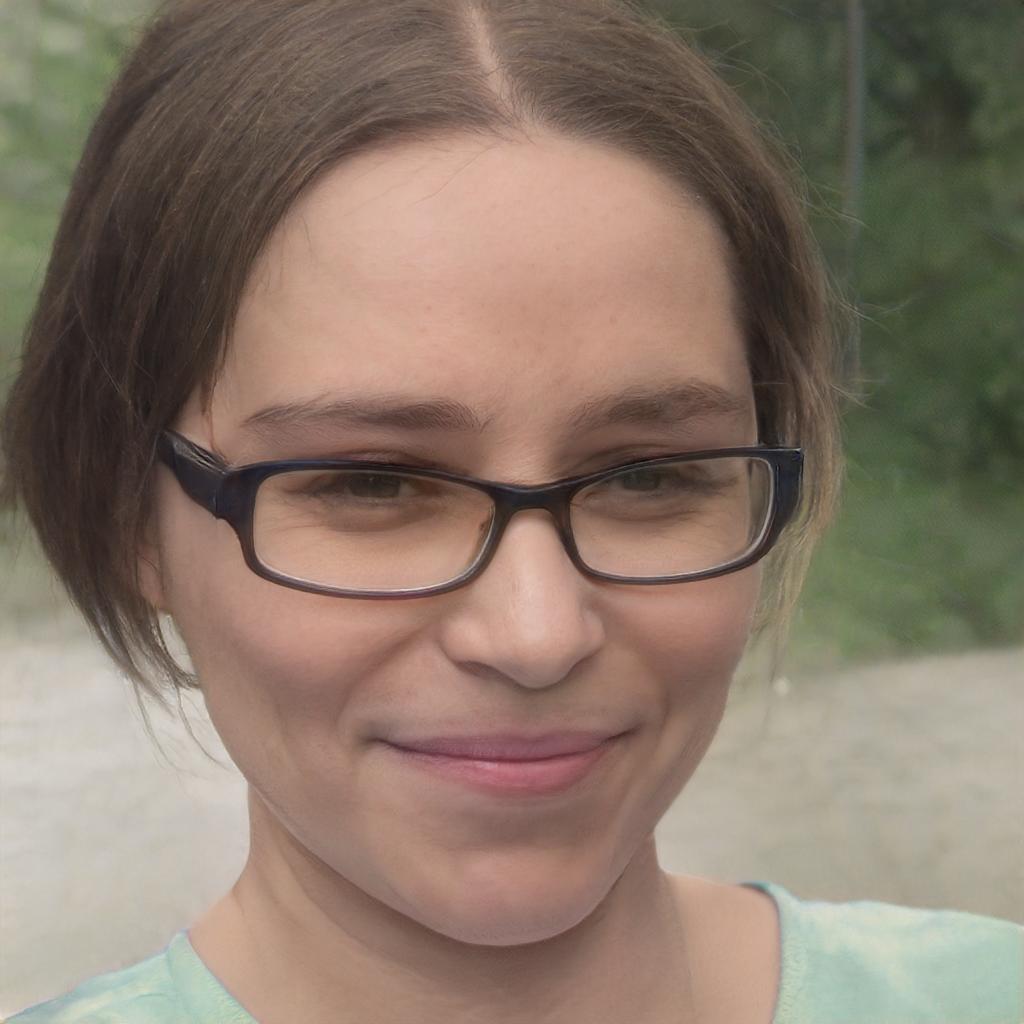} &
	    \includegraphics[width=0.34\columnwidth]{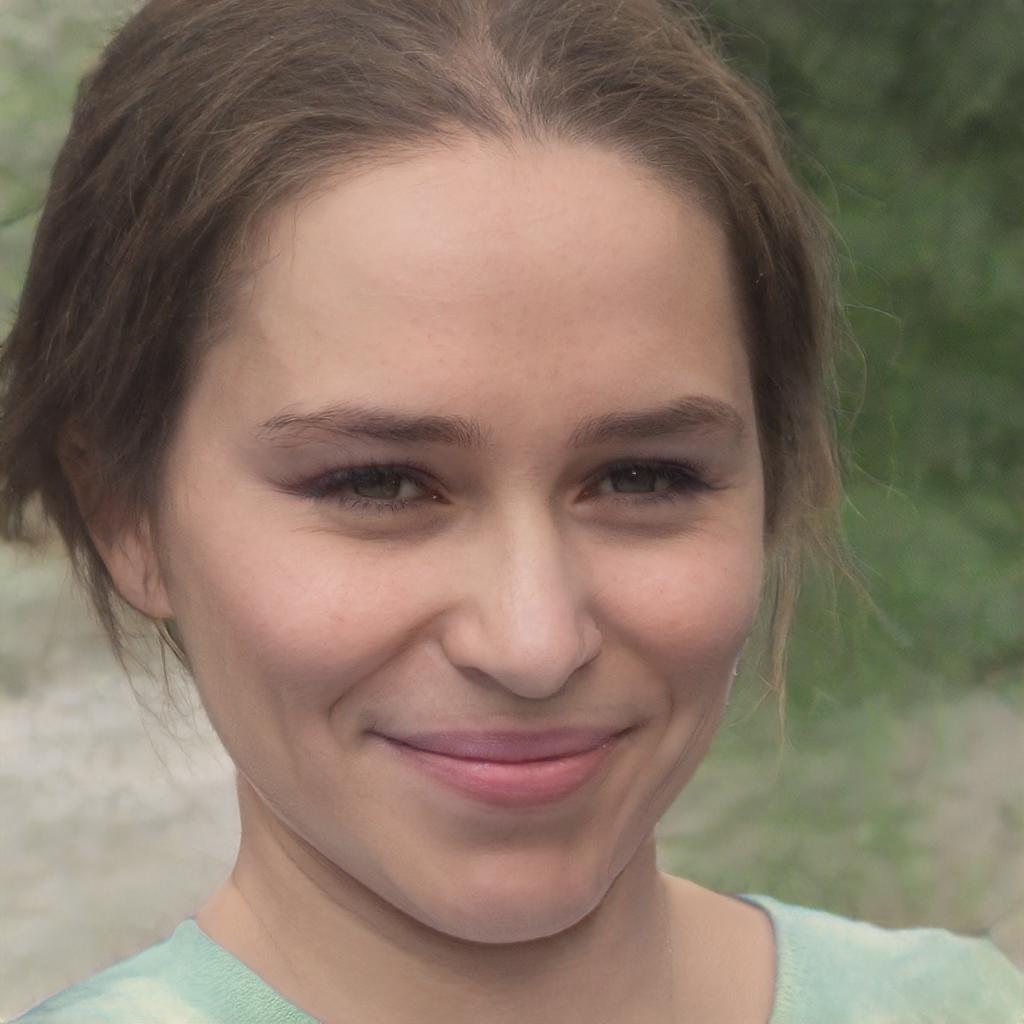} \\
	\end{tabular}
	
	\caption{Comparison between StyleFlow~\cite{abdal2020styleflow} and our global directions. Our method produces results of similar quality, despite the fact that StyleFlow simultaneously uses several attribute classifiers and regressors (from the Microsoft face API), and is thus able to manipulate a limited set of attributes. In contrast, our method requires no extra supervision.
	}
	\label{fig:styleflow}
\end{figure*}

Figure~\ref{fig:styleflow} shows a comparison between StyleFlow~\cite{abdal2020styleflow} and our global directions method. It may be seen that our method is able to produce results of comparable visual quality, despite the fact that StyleFlow requires the simultaneous use of several attribute classifiers and regressors (from the Microsoft face API), and is thus able to manipulate a limited set of attributes. In contrast, our method required no extra supervision to produce these and all of the other manipulations demonstrated in this work.

\begin{figure*}[]
	\setlength{\tabcolsep}{1pt}
	\begin{center}
	{\footnotesize
	\begin{tabular}{ccccccccc}
		  Input  & Global & Mapper  & Input  & Global & Mapper  & Input  & Global & Mapper \\
	 
		\includegraphics[width=0.11\textwidth]{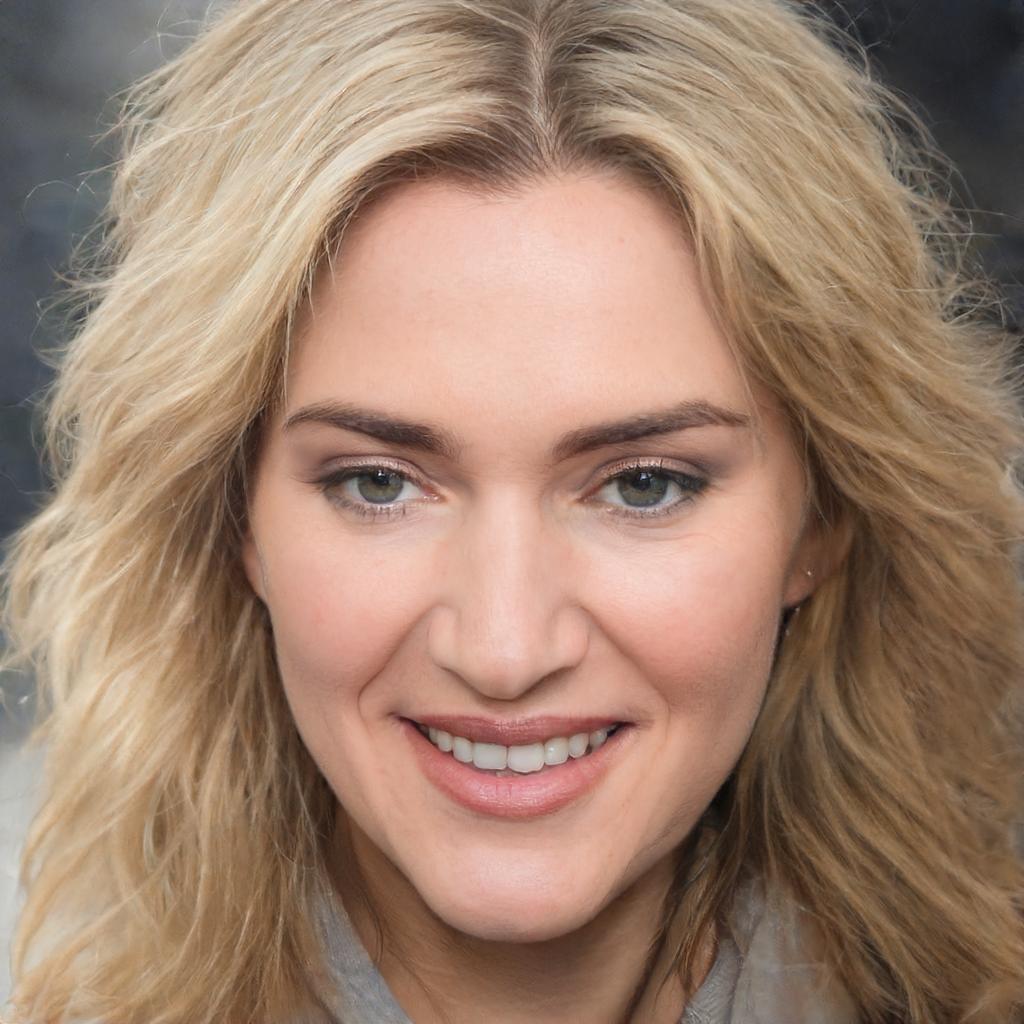} &
		\includegraphics[width=0.11\textwidth]{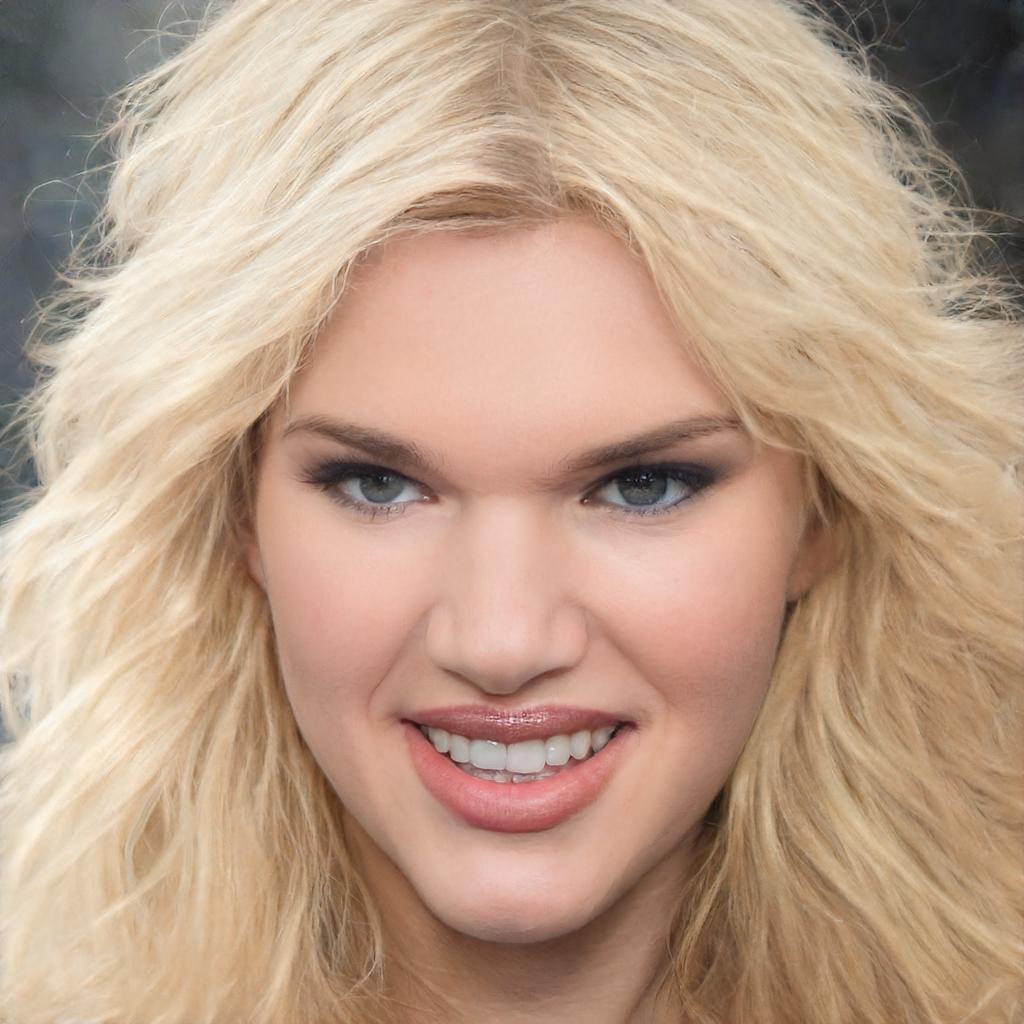} &
		\includegraphics[width=0.11\textwidth]{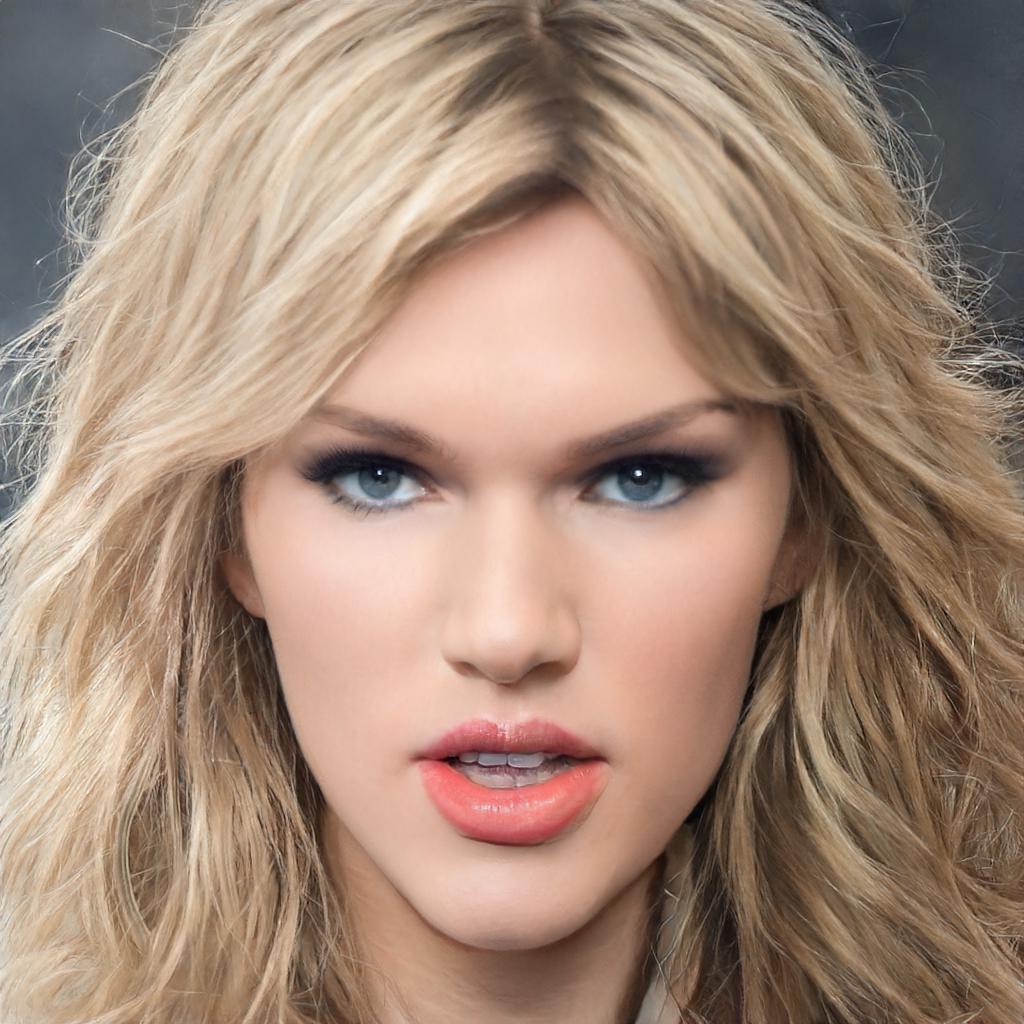} &
		
		\includegraphics[width=0.11\textwidth]{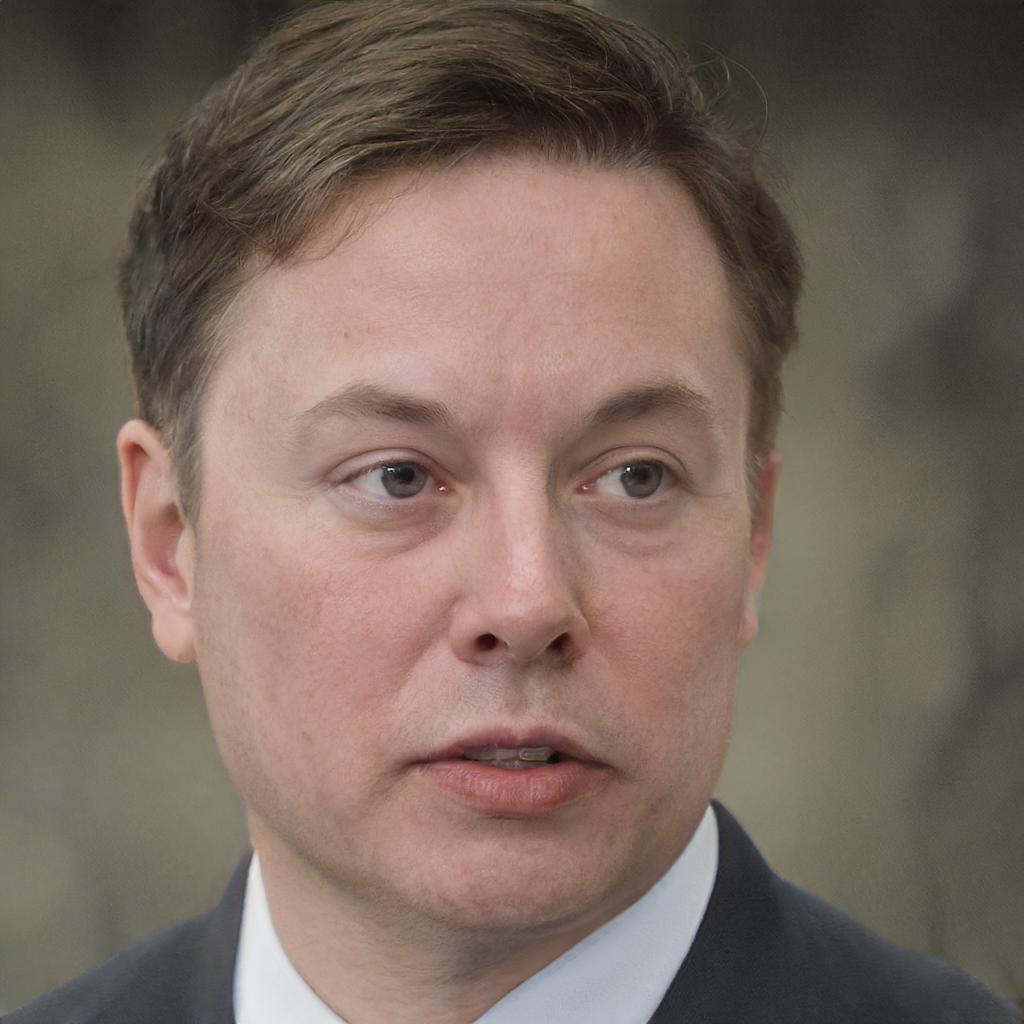} &
		\includegraphics[width=0.11\textwidth]{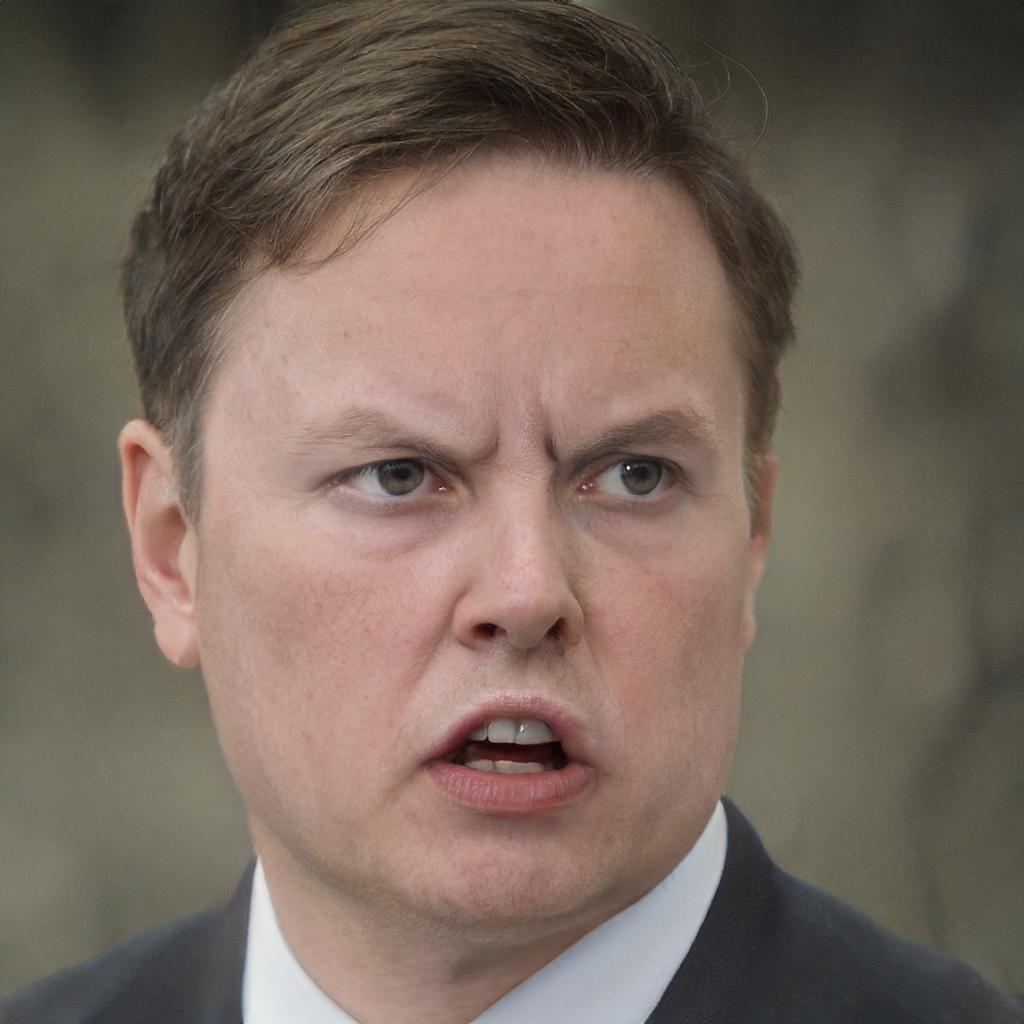} &
		\includegraphics[width=0.11\textwidth]{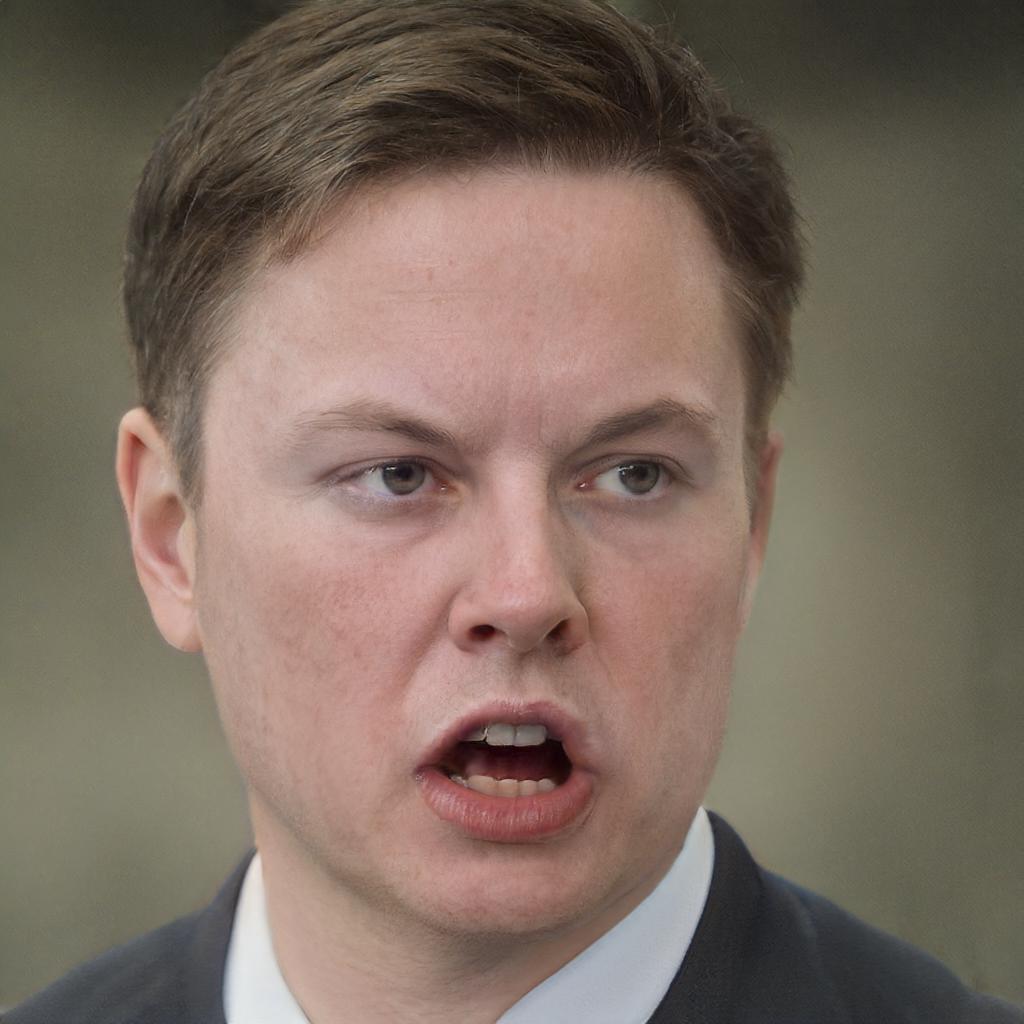} &

		\includegraphics[width=0.11\textwidth]{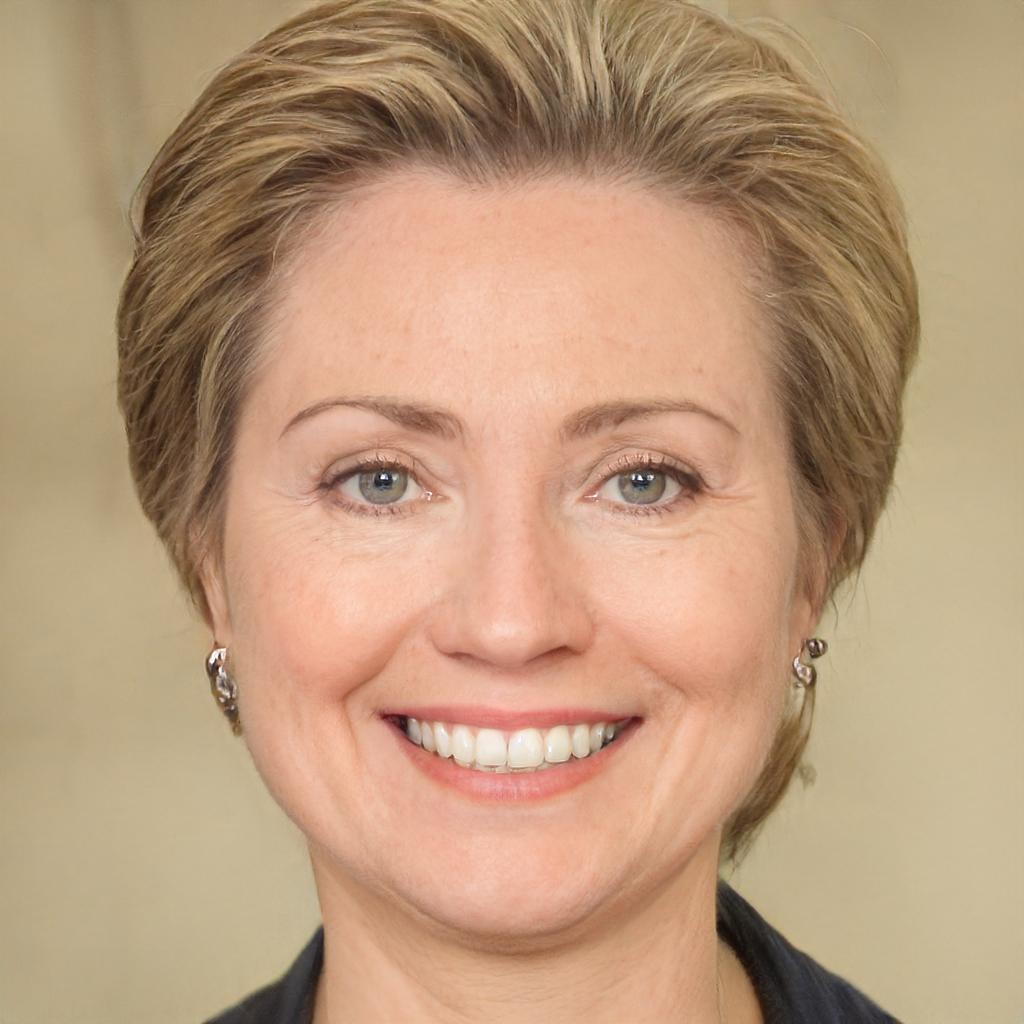} &
		\includegraphics[width=0.11\textwidth]{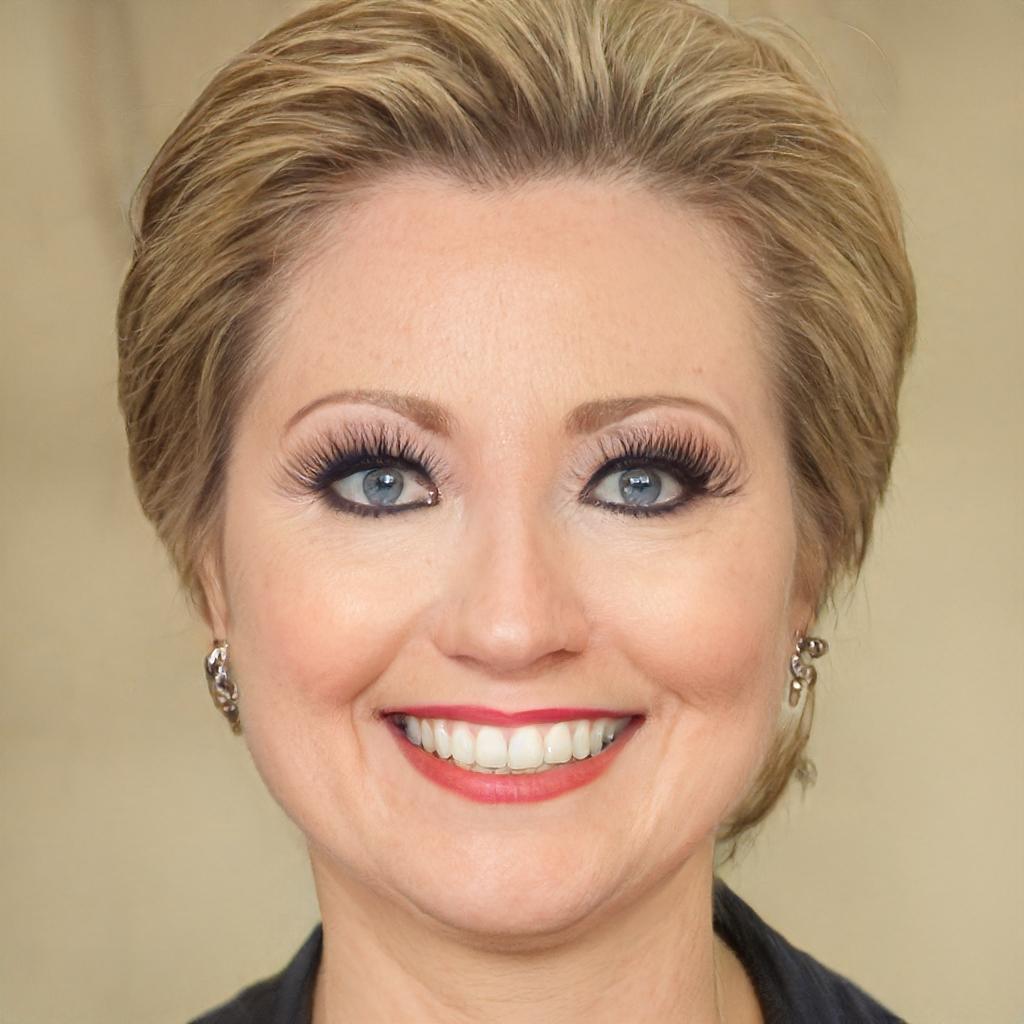} &
		\includegraphics[width=0.11\textwidth]{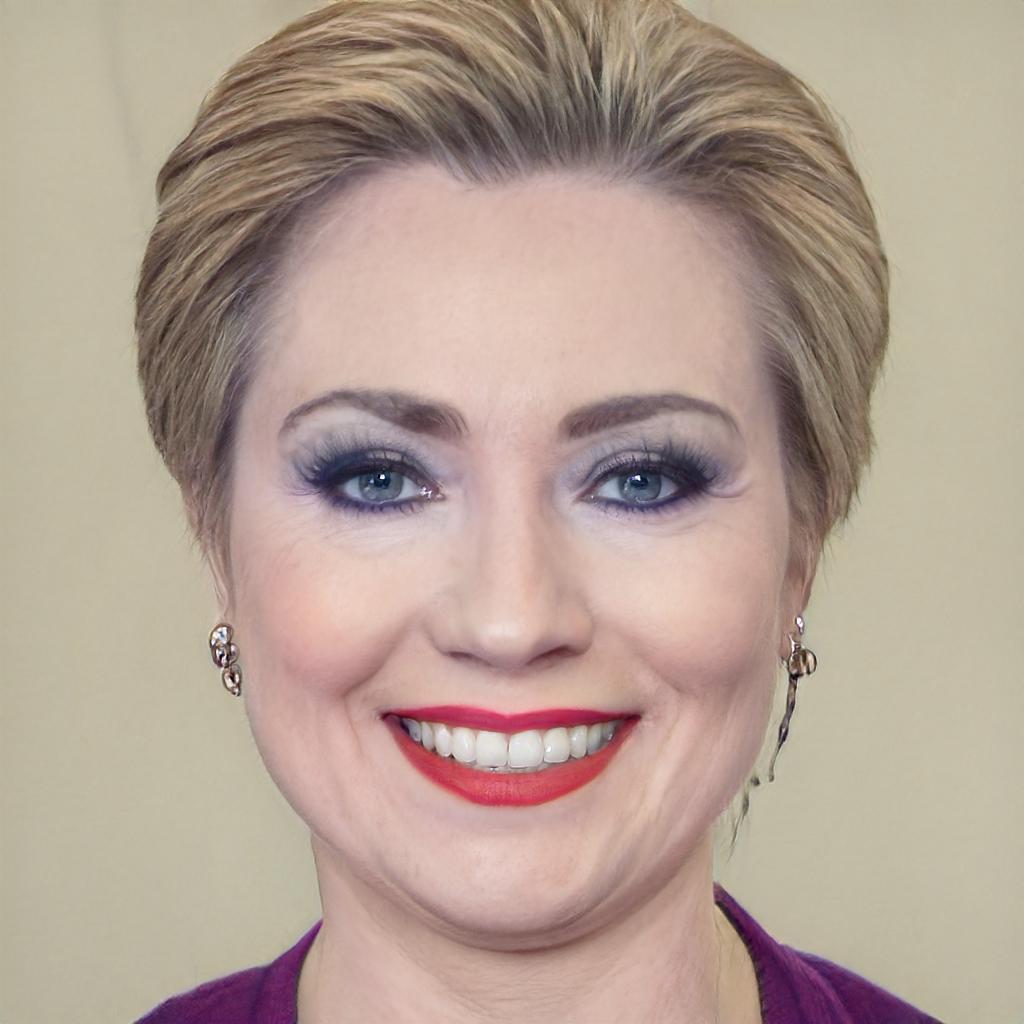} \\

		\includegraphics[width=0.11\textwidth]{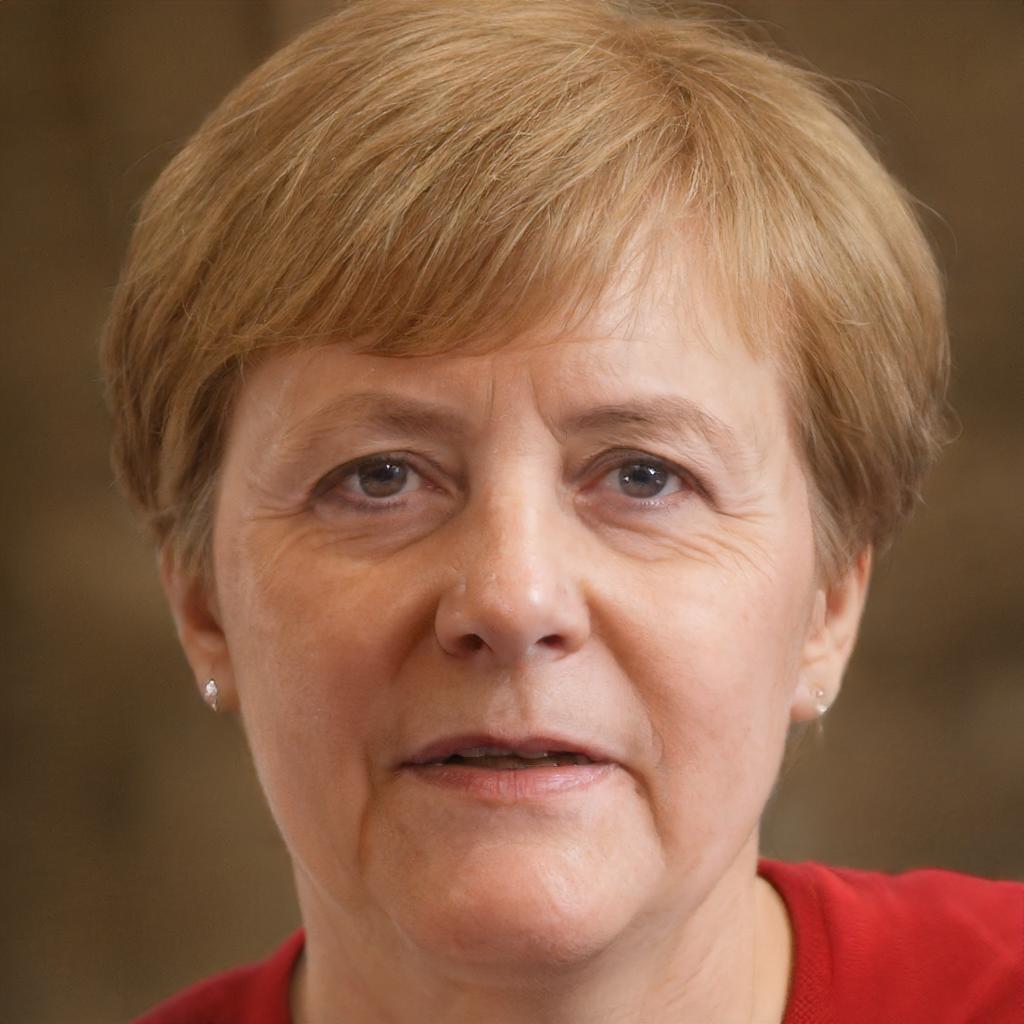} &
		\includegraphics[width=0.11\textwidth]{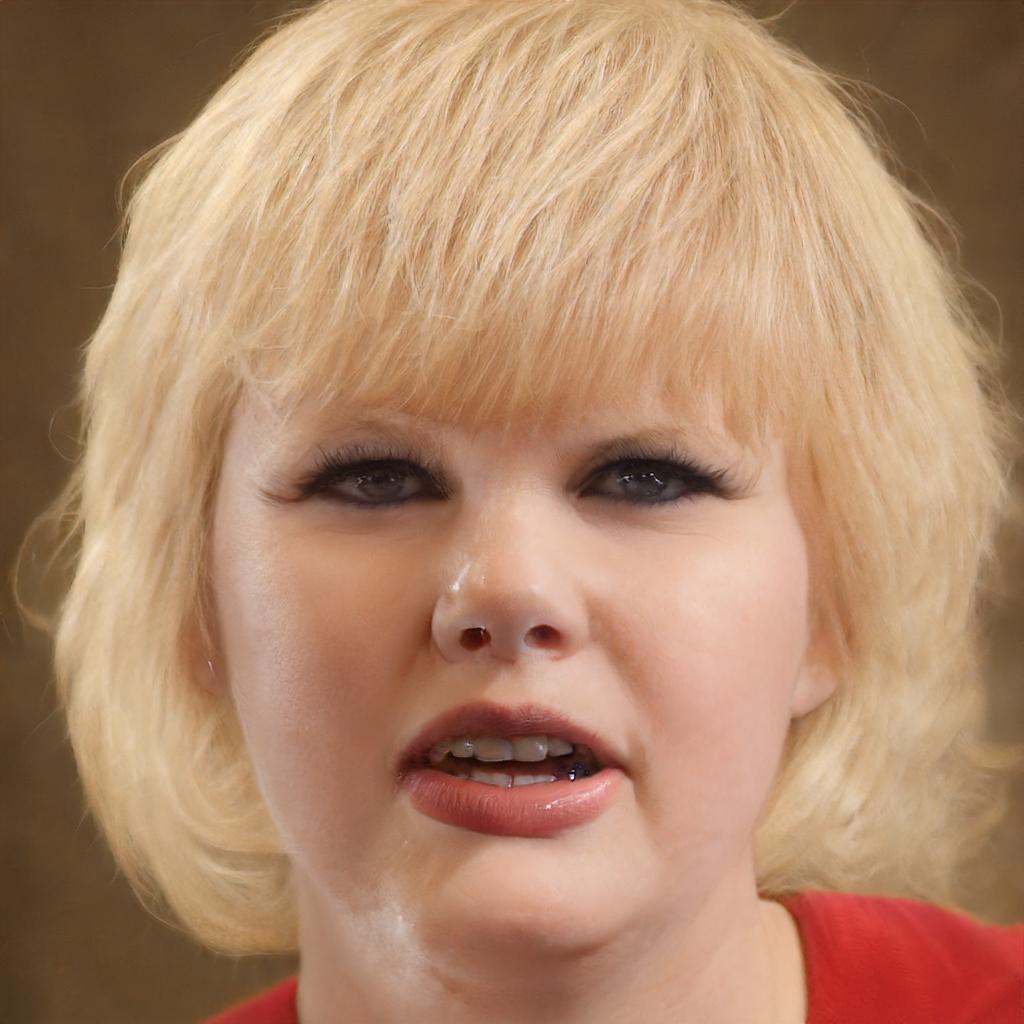} &
		\includegraphics[width=0.11\textwidth]{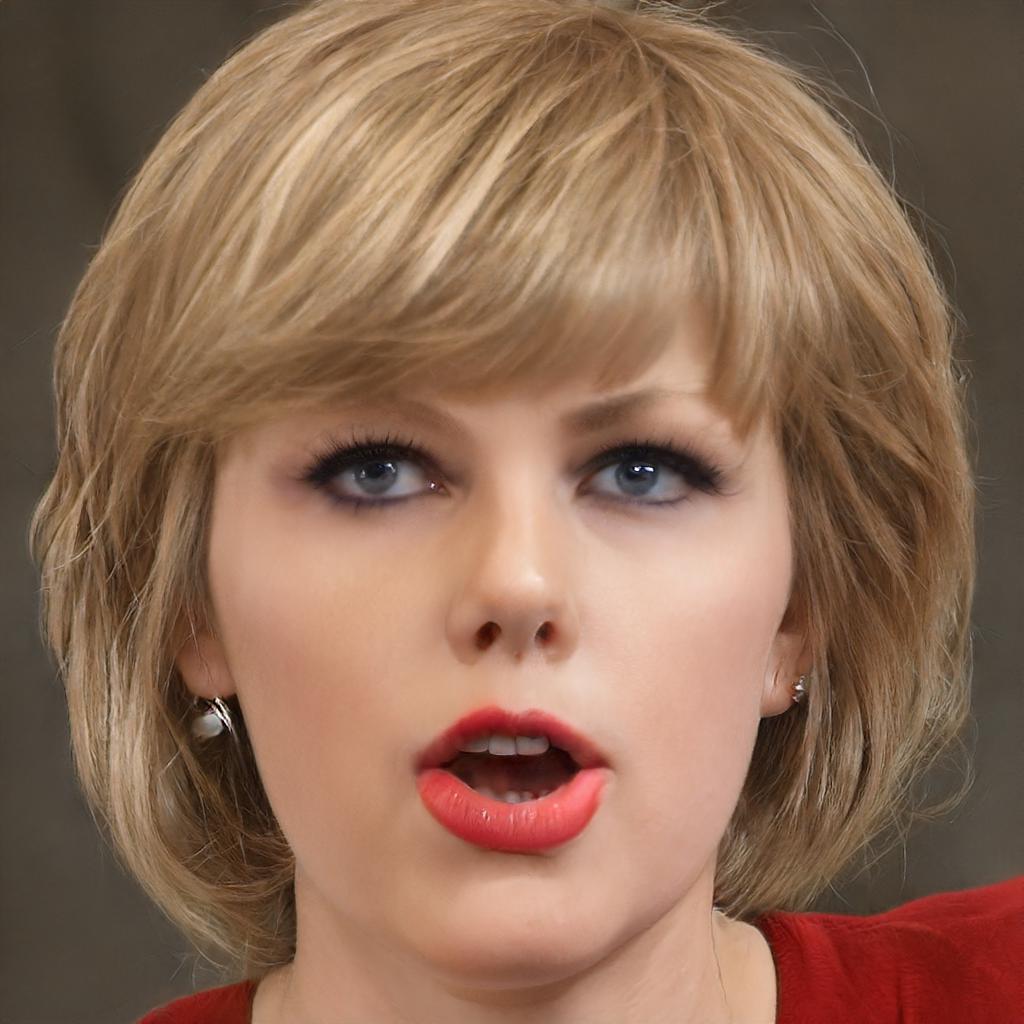} &
		
		\includegraphics[width=0.11\textwidth]{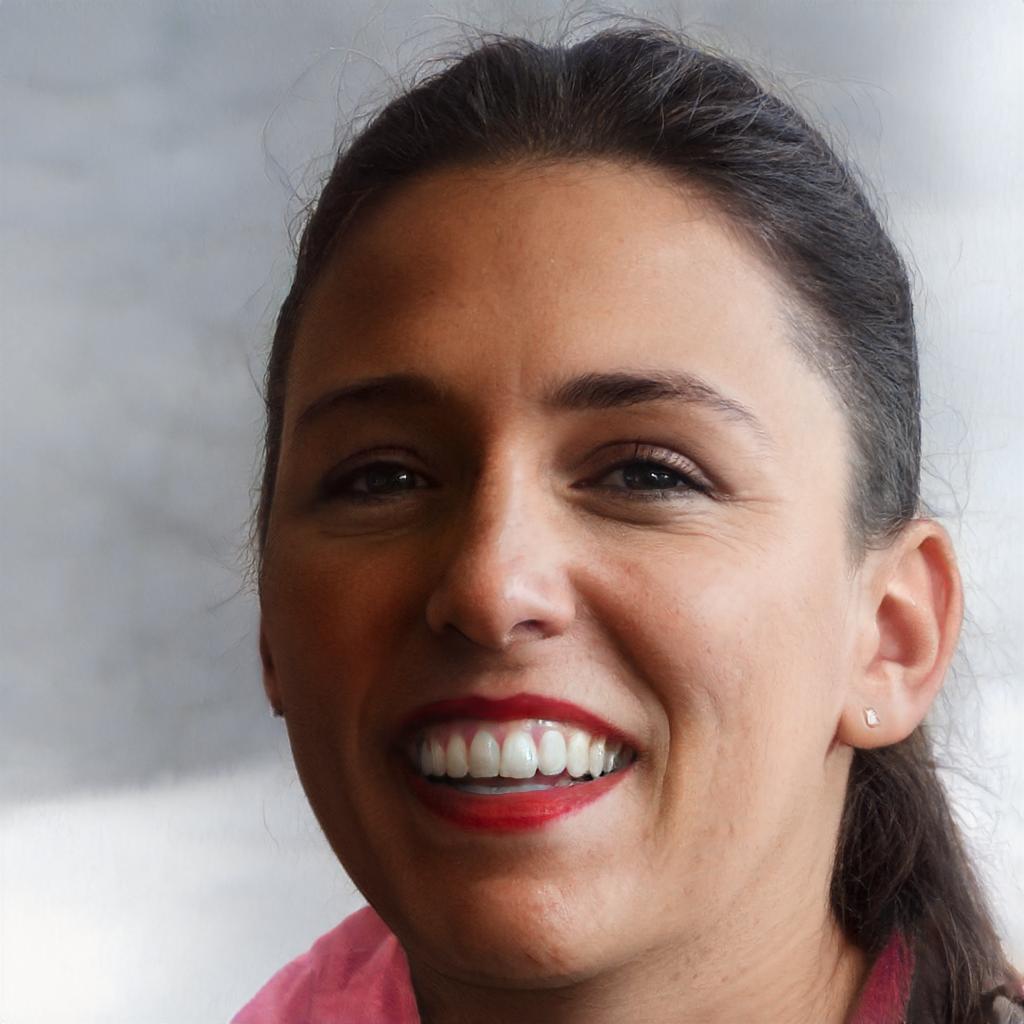} &
		\includegraphics[width=0.11\textwidth]{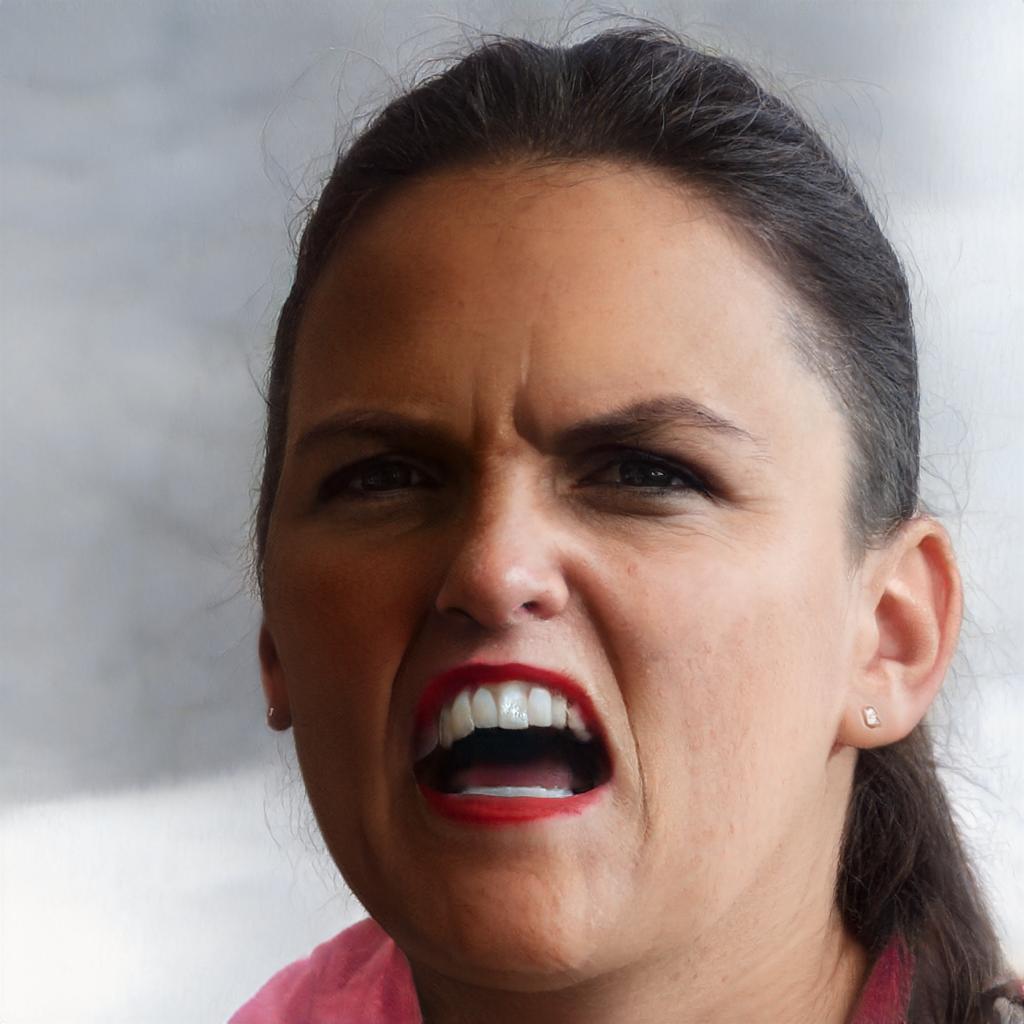} &
		\includegraphics[width=0.11\textwidth]{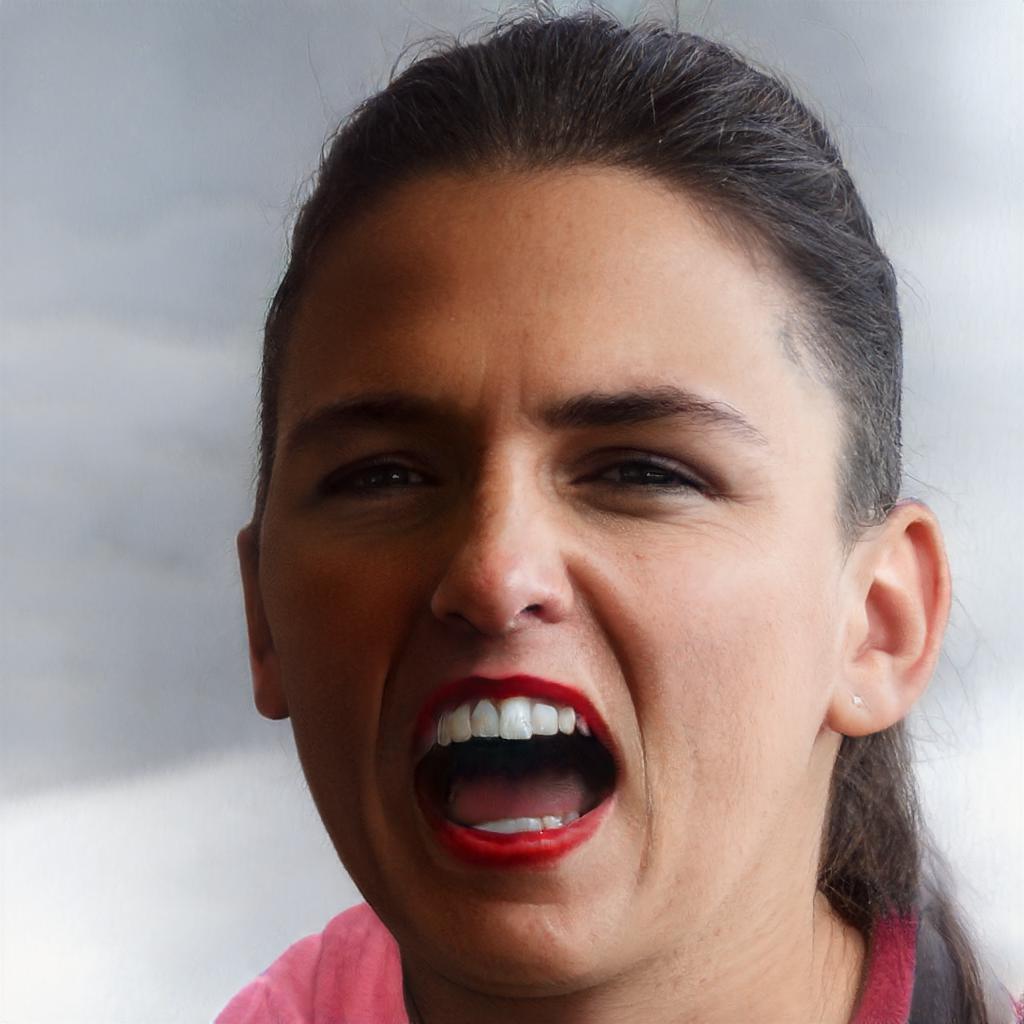} &
		
		\includegraphics[width=0.11\textwidth]{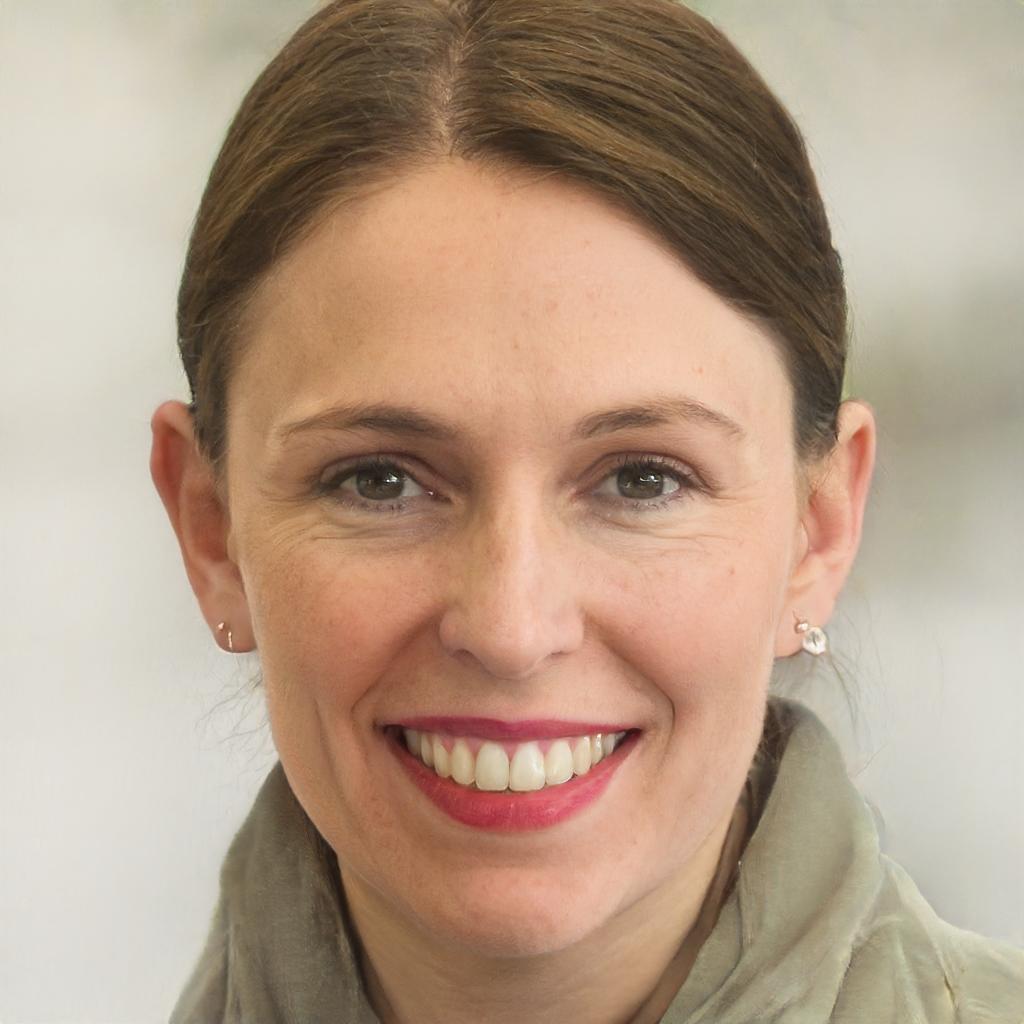} &
		\includegraphics[width=0.11\textwidth]{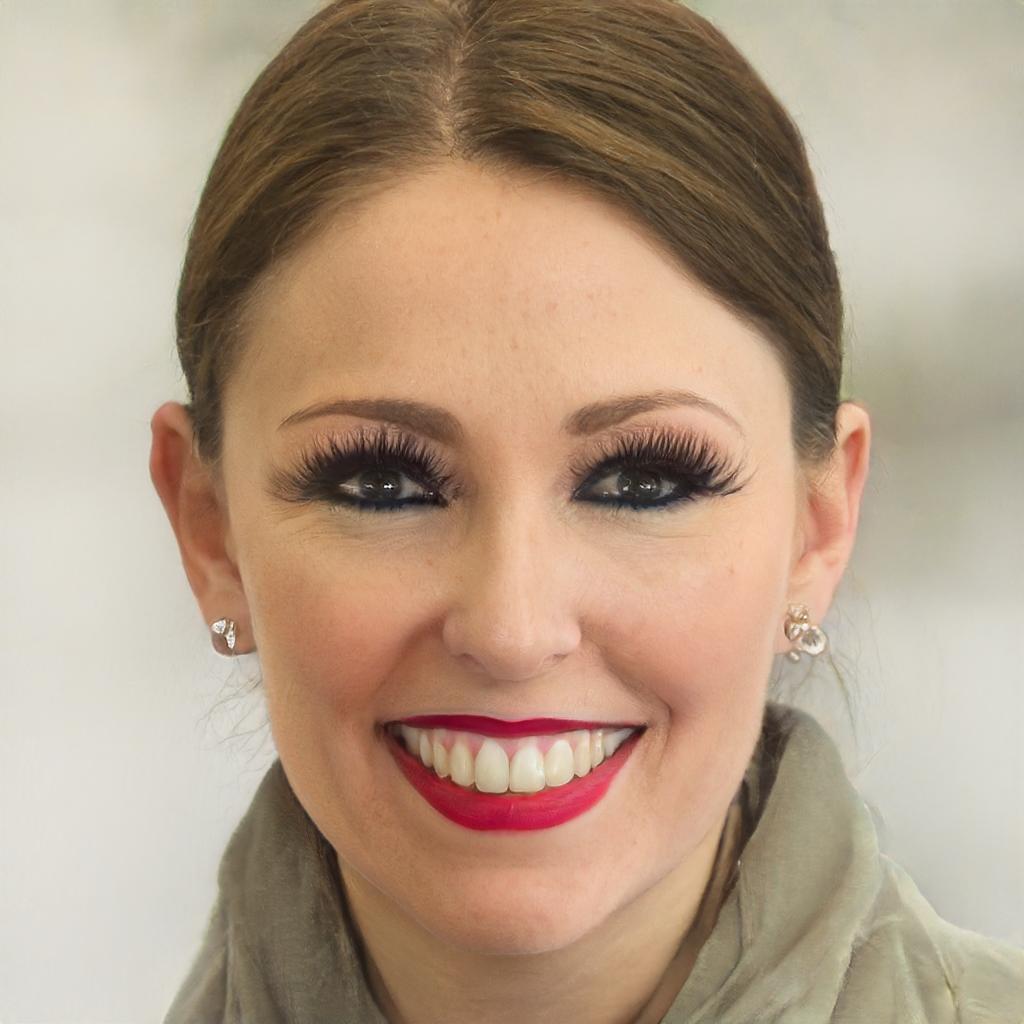} &
		\includegraphics[width=0.11\textwidth]{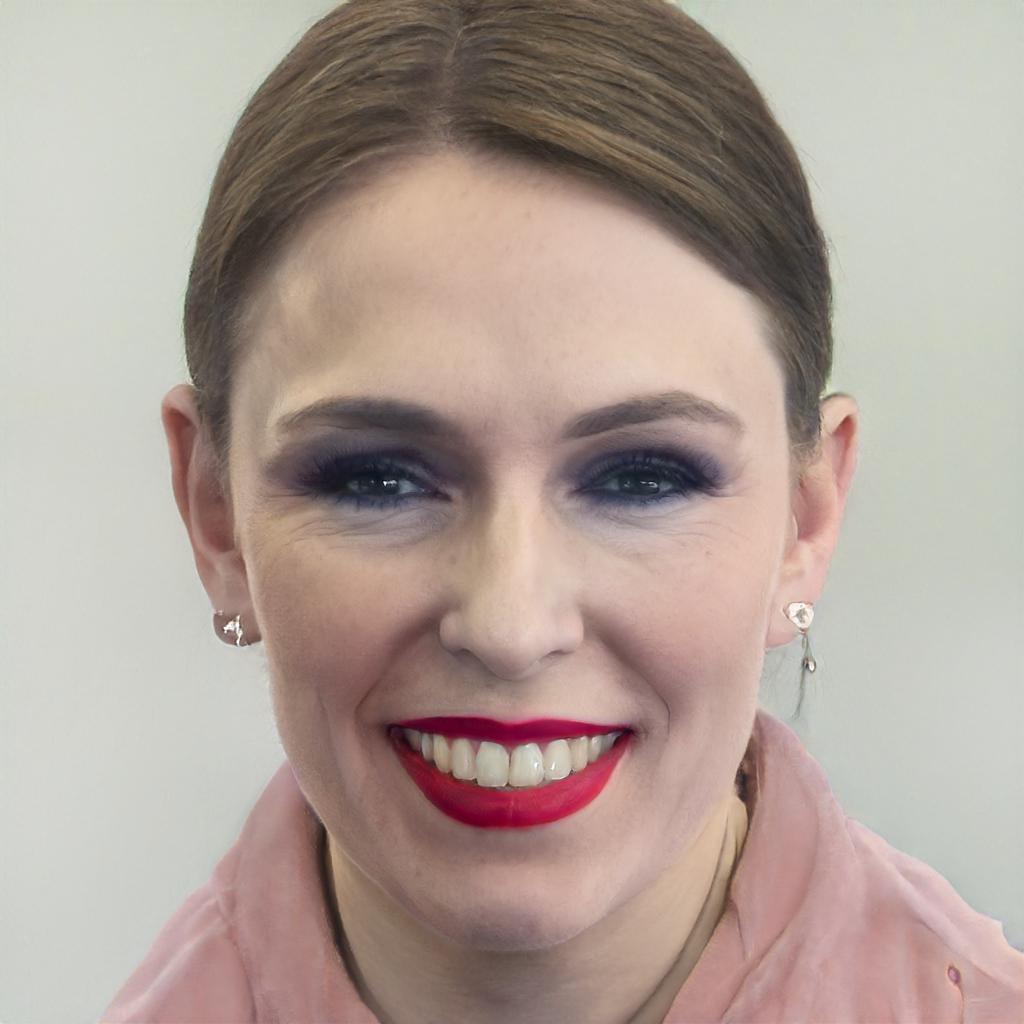} \\

		\includegraphics[width=0.11\textwidth]{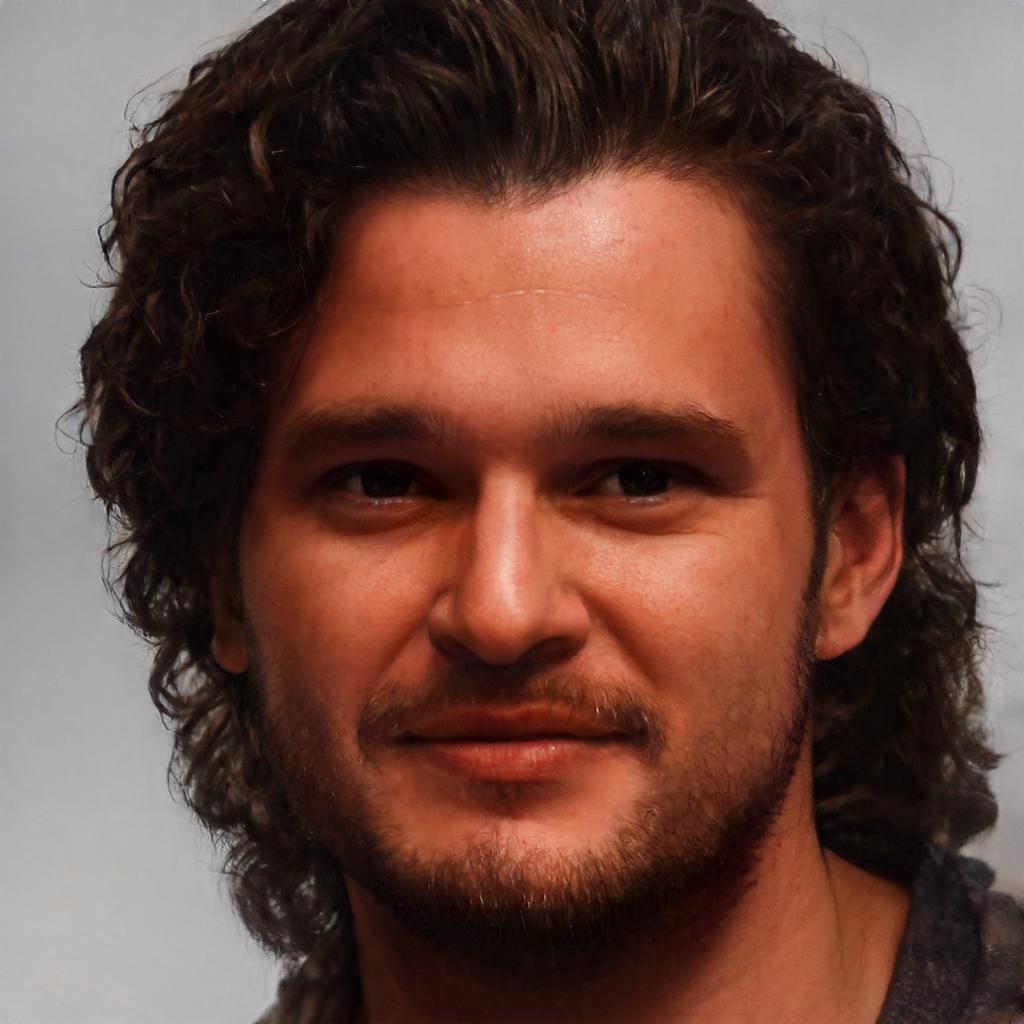} &
		\includegraphics[width=0.11\textwidth]{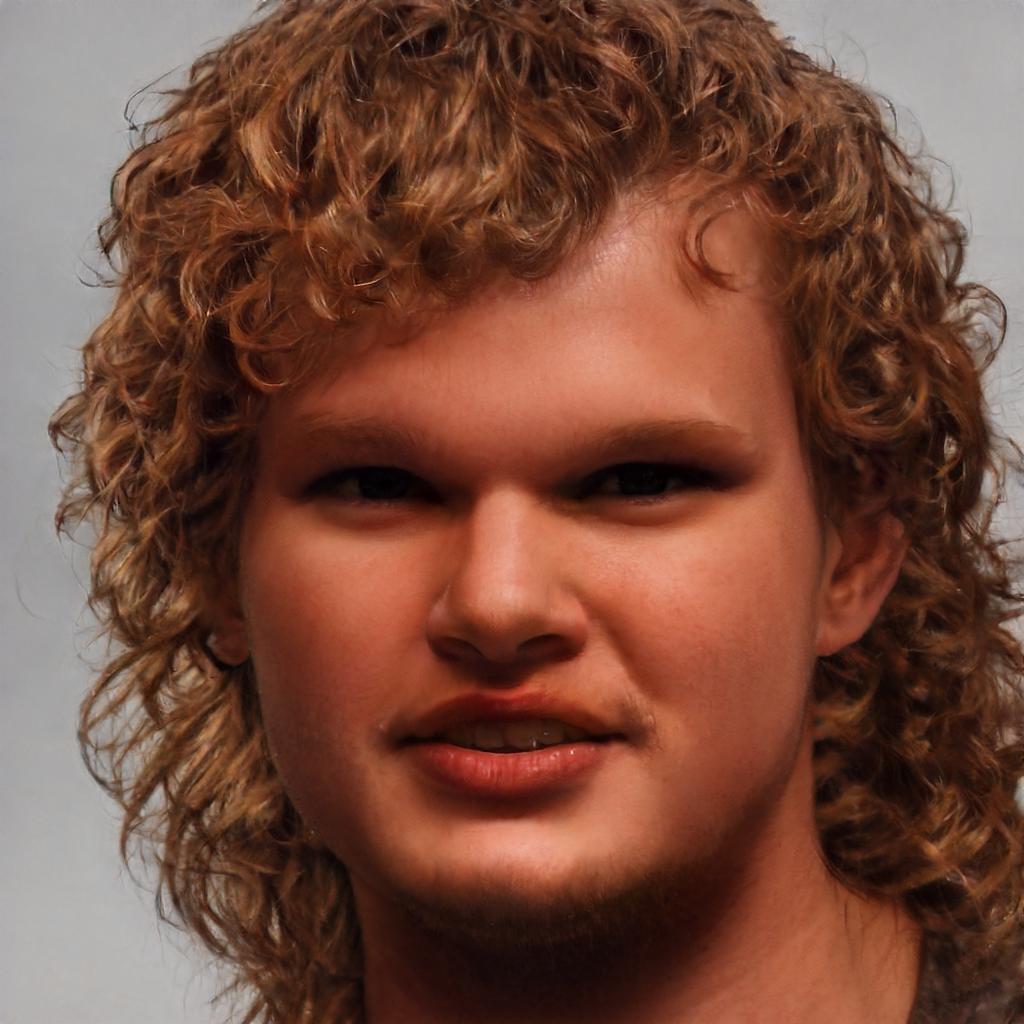} &
		\includegraphics[width=0.11\textwidth]{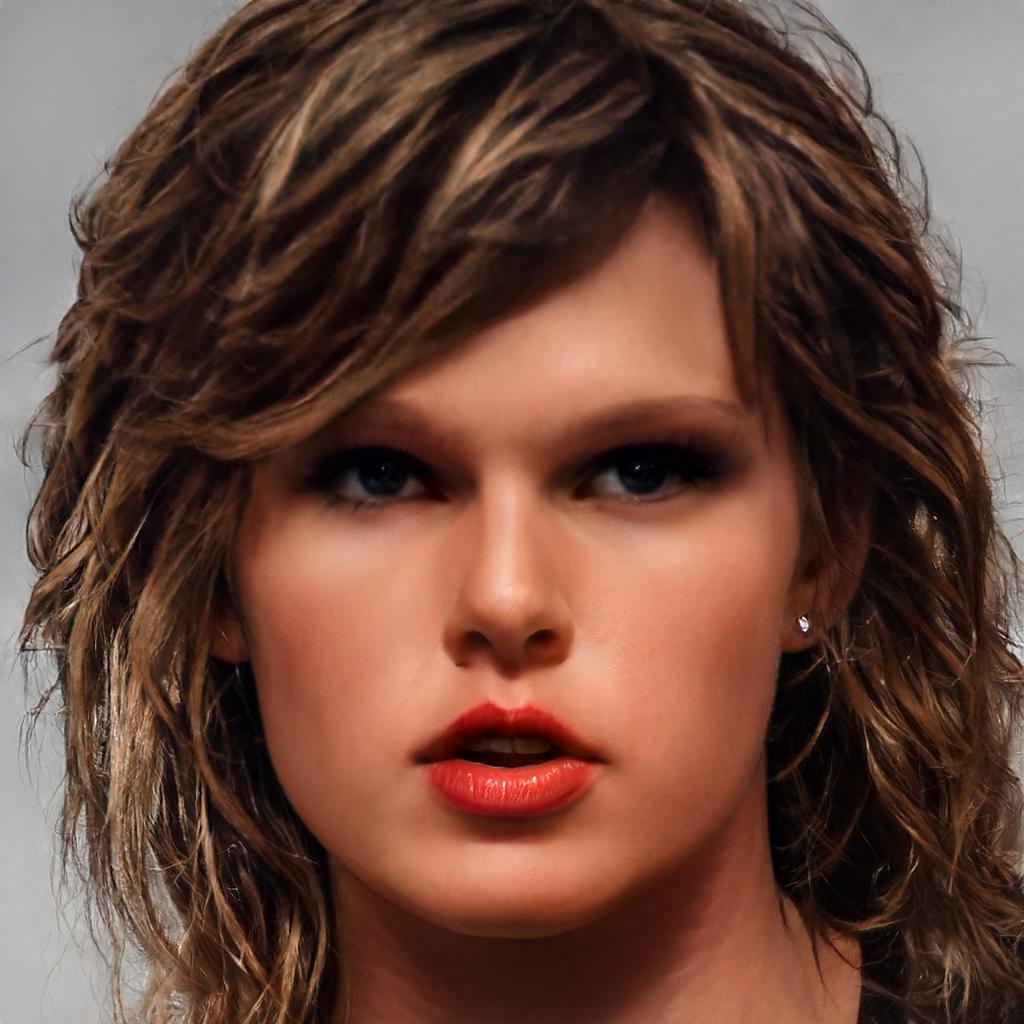} &
		
		\includegraphics[width=0.11\textwidth]{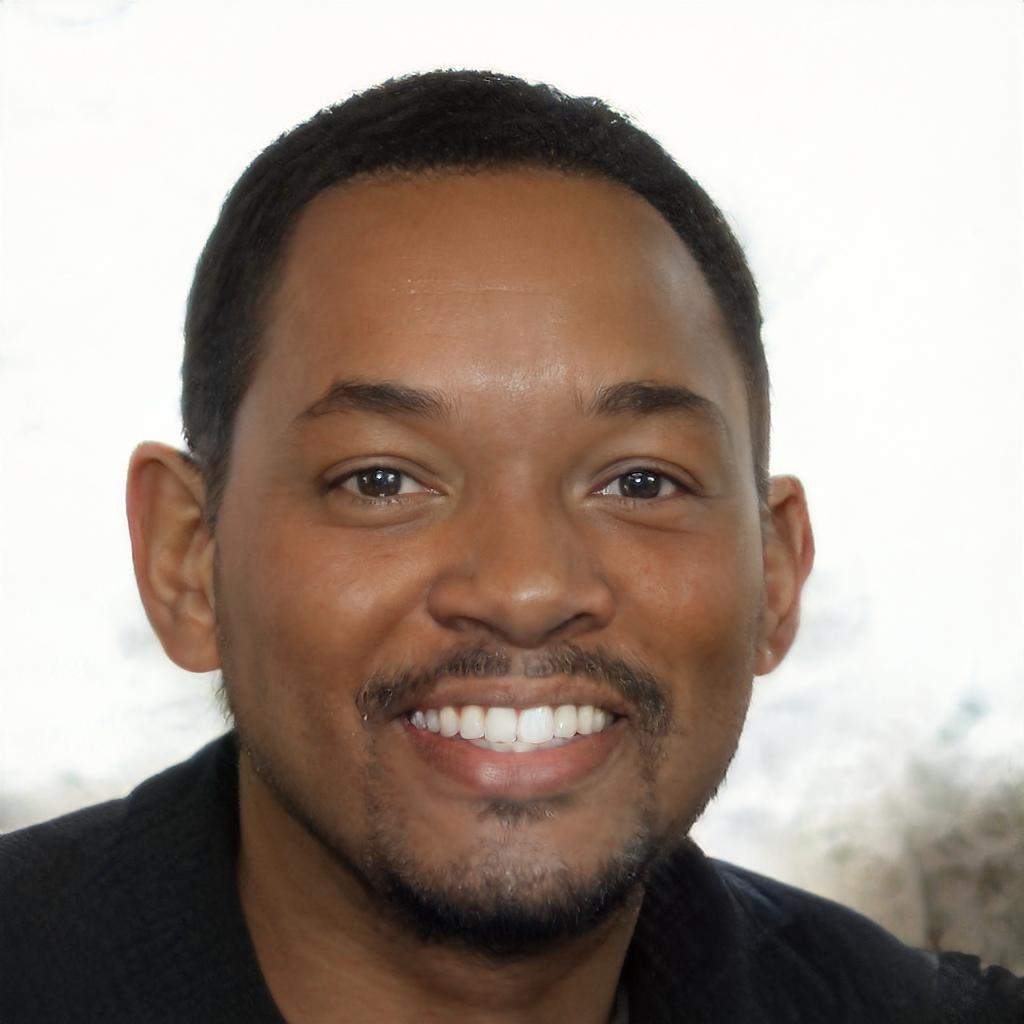} &
		\includegraphics[width=0.11\textwidth]{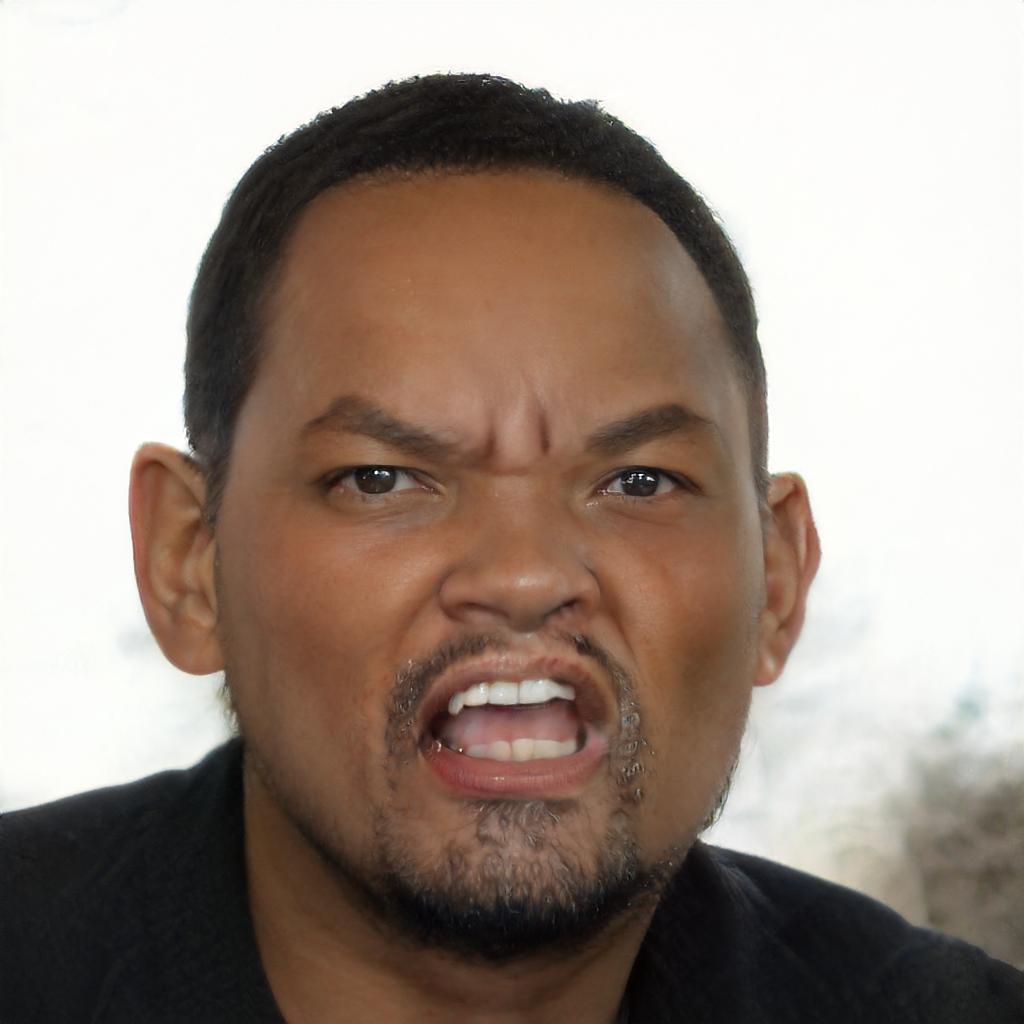} &
		\includegraphics[width=0.11\textwidth]{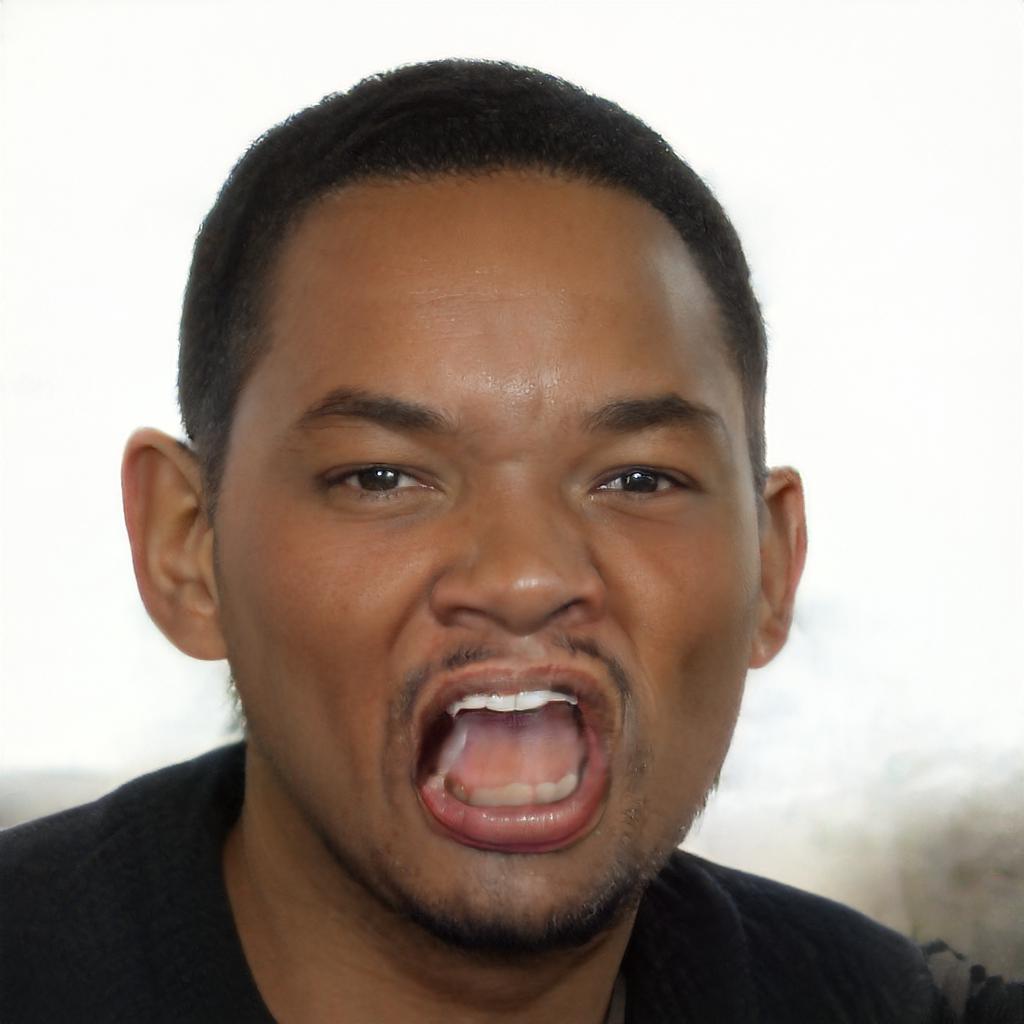} &

		\includegraphics[width=0.11\textwidth]{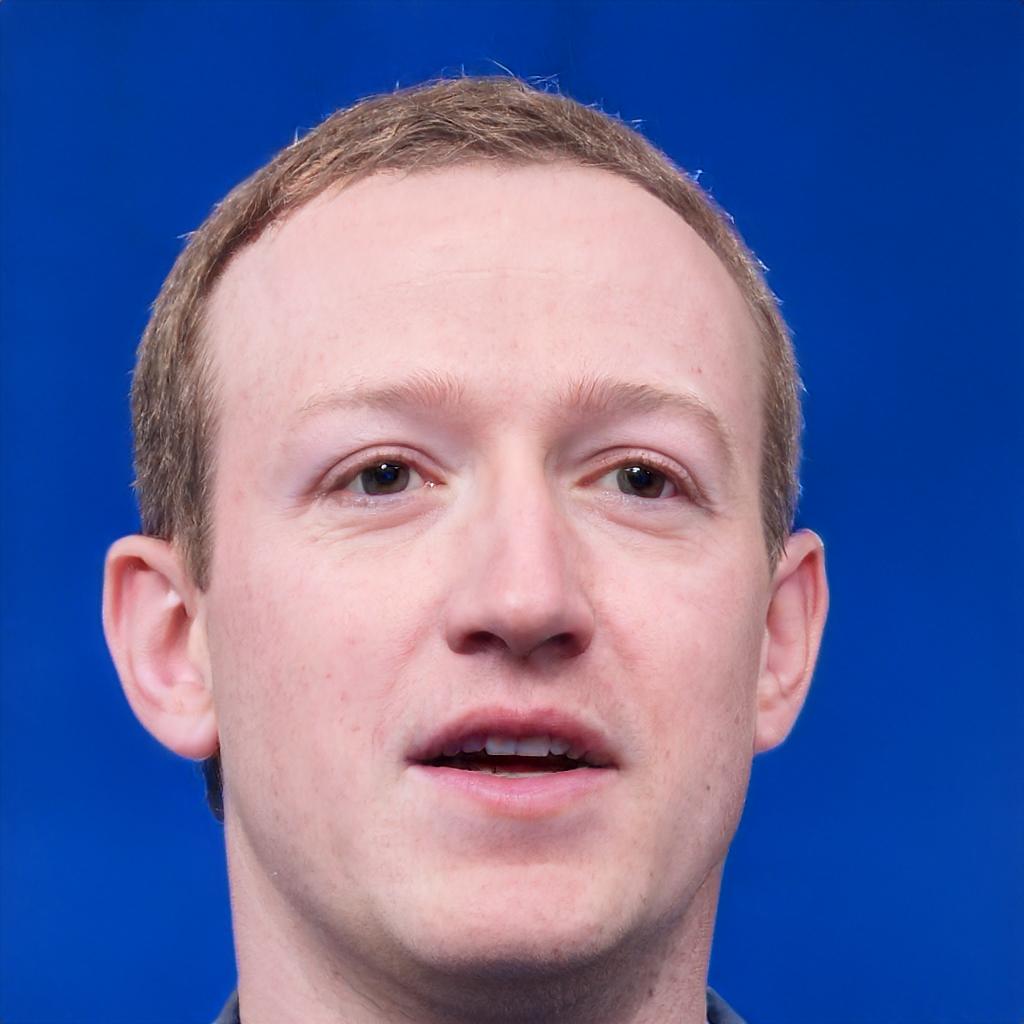} &
		\includegraphics[width=0.11\textwidth]{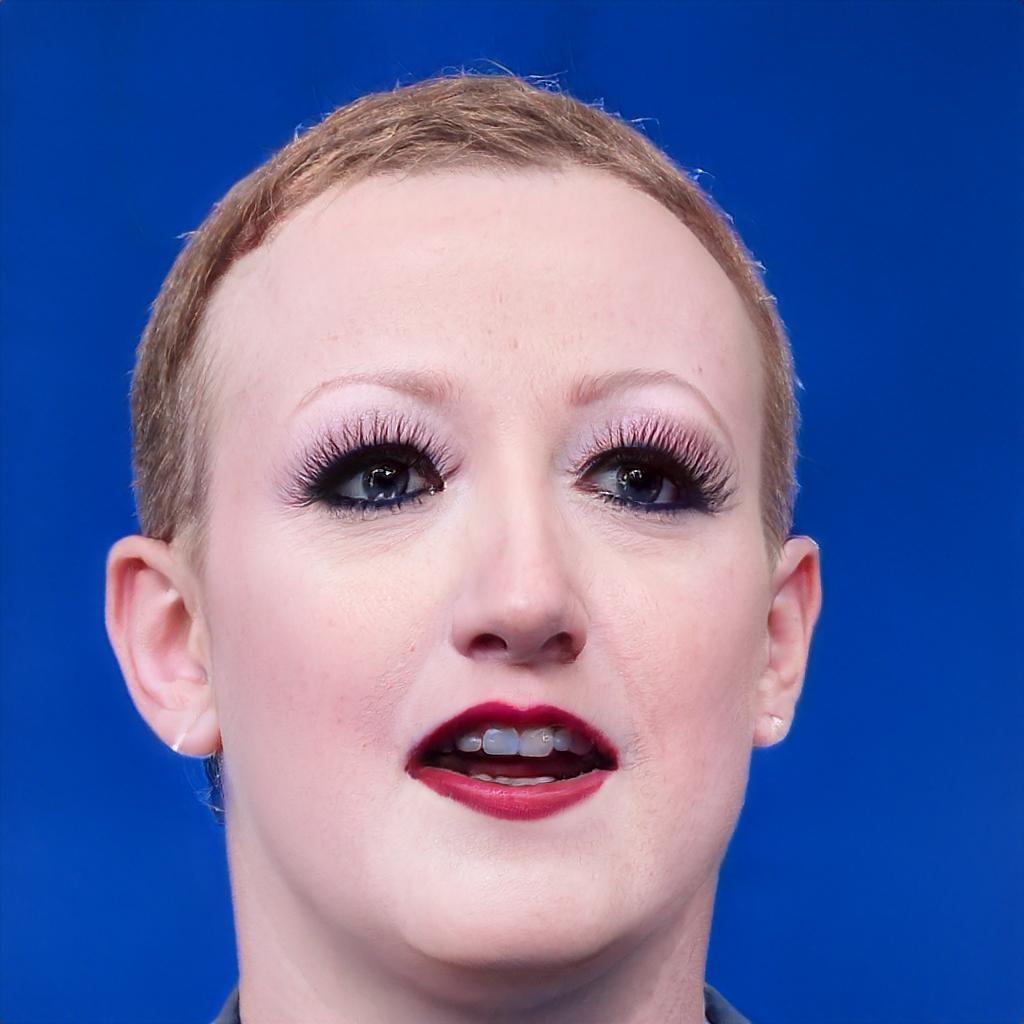} &
		\includegraphics[width=0.11\textwidth]{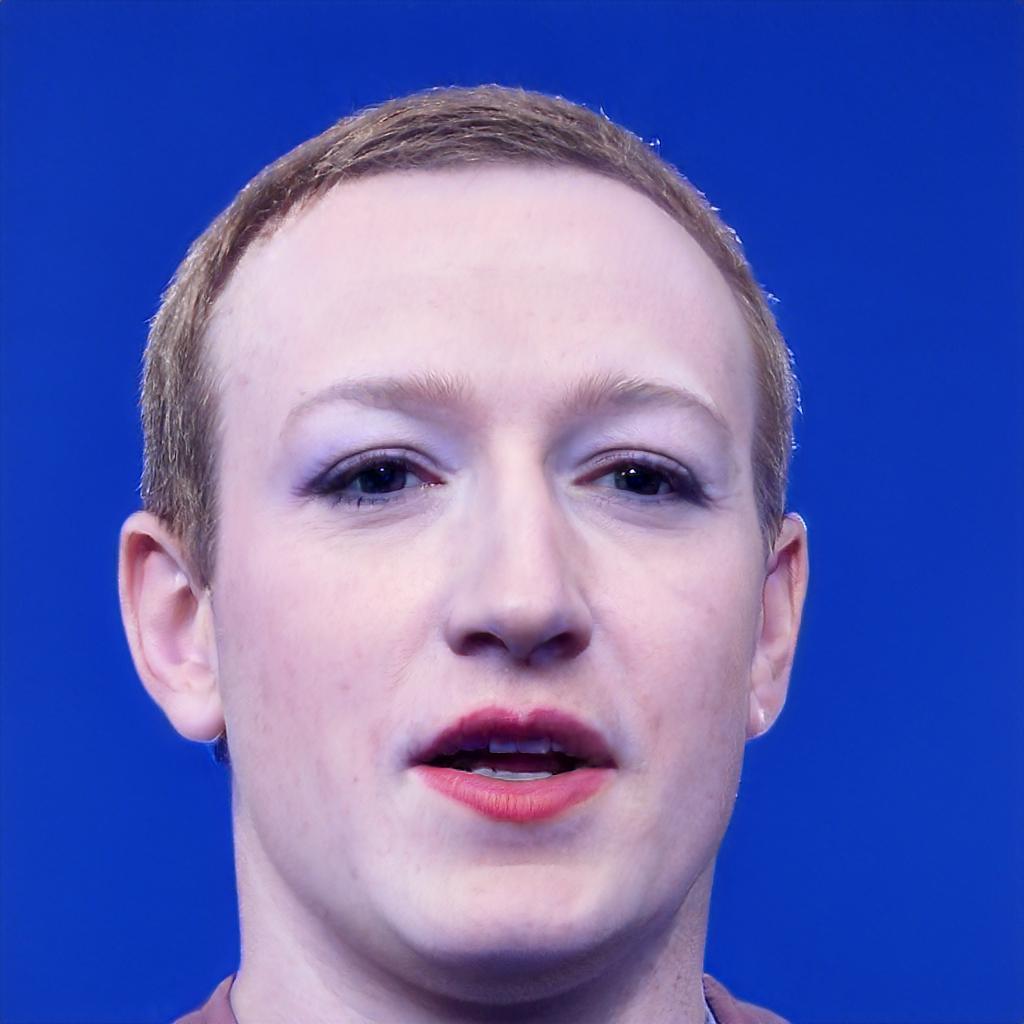} \\
		
	\multicolumn{3}{c}{\ruleline{0.30\textwidth}{Taylor Swift}} &\multicolumn{3}{c}{\ruleline{0.30\textwidth}{Angry}} &\multicolumn{3}{c}{\ruleline{0.30\textwidth}{Makeup}} 
	\end{tabular}}
	\end{center}
	\caption{\label{fig:global-vs-mapper-supp} We compare our global directions with our latent mapper using three different kinds of attributes.}
\end{figure*}

Figure~\ref{fig:global-vs-mapper-supp} shows an additional comparison between text-driven manipulation using our global directions method and our latent mapper. Our observations are similar to the ones we made regarding Figure 10 in the main paper.

\begin{figure*}[p]
	\setlength{\tabcolsep}{1pt}	
	\centering
	\begin{tabular}{ccccc}
		{\footnotesize Original} & {\footnotesize Lion+} & {\footnotesize Lion++} & {\footnotesize Wolf+} & {\footnotesize Wolf++}   \\
		
		\includegraphics[width=0.38  \columnwidth]{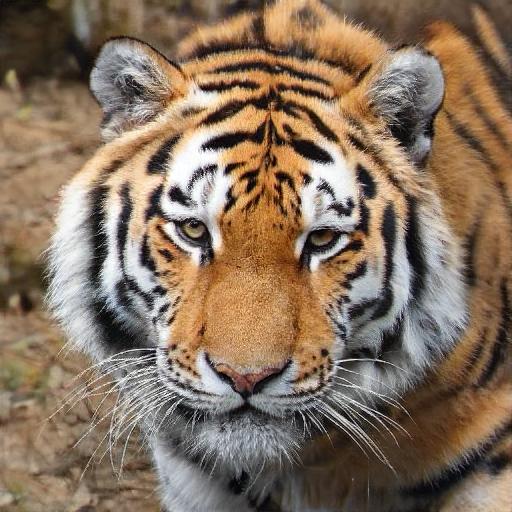} &
		\includegraphics[width=0.38  \columnwidth]{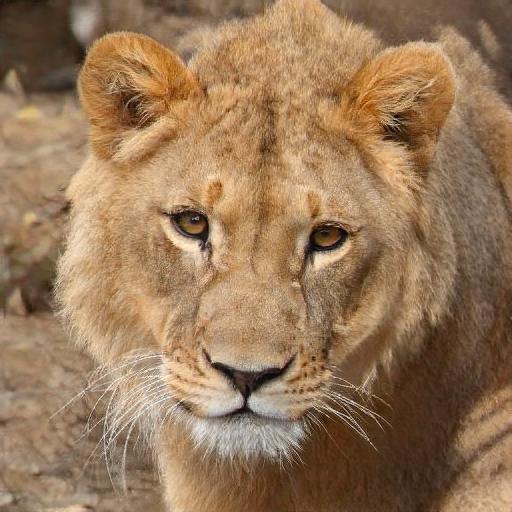} &
		\includegraphics[width=0.38  \columnwidth]{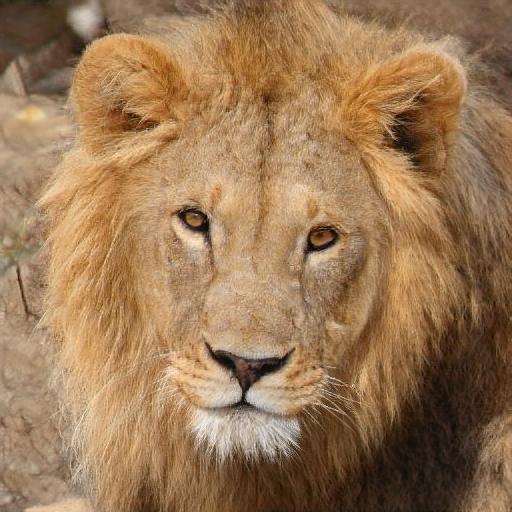} &
		\includegraphics[width=0.38  \columnwidth]{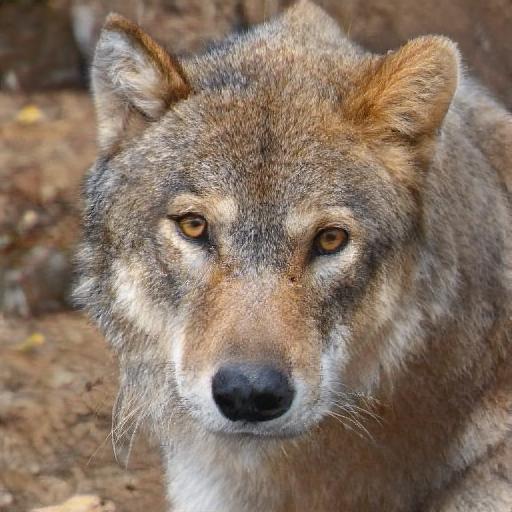} &
		\includegraphics[width=0.38  \columnwidth]{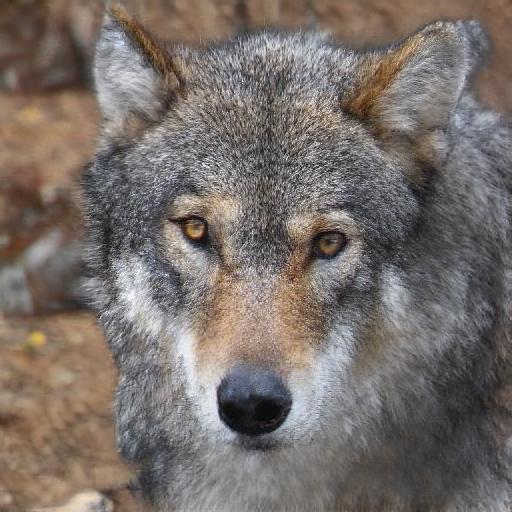} \\
		
		\includegraphics[width=0.38  \columnwidth]{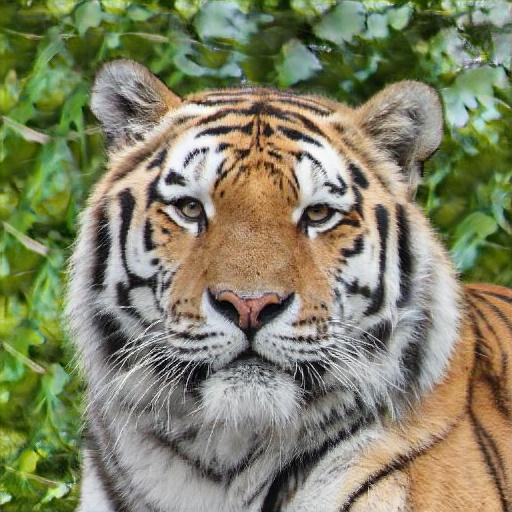} &
		\includegraphics[width=0.38  \columnwidth]{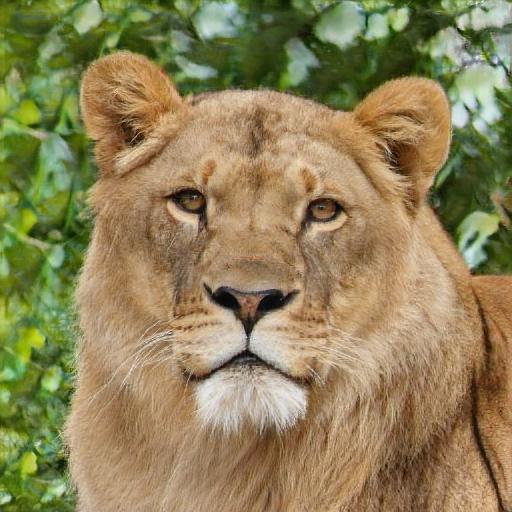} &
		\includegraphics[width=0.38  \columnwidth]{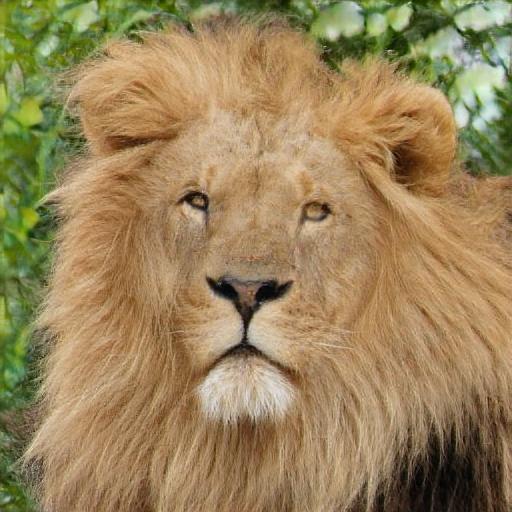} &
		\includegraphics[width=0.38  \columnwidth]{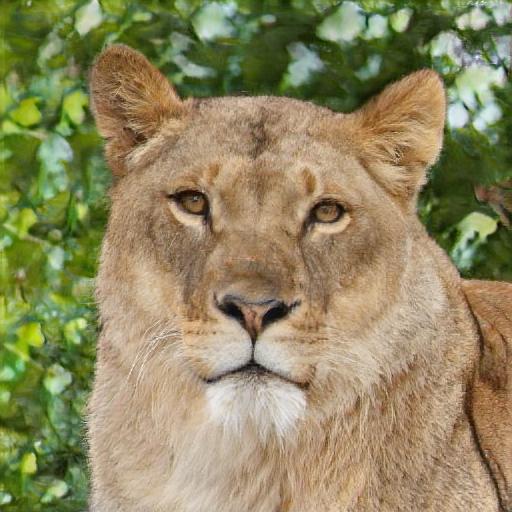} &
		\includegraphics[width=0.38  \columnwidth]{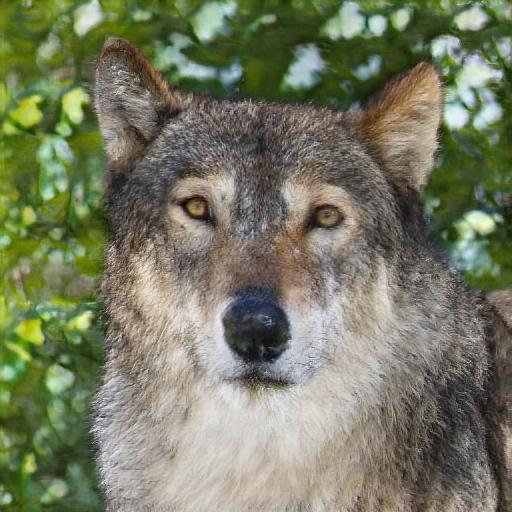} \\
		
		\includegraphics[width=0.38  \columnwidth]{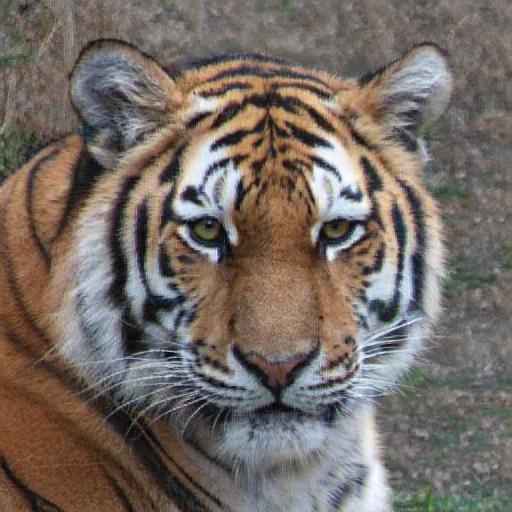} &
		\includegraphics[width=0.38  \columnwidth]{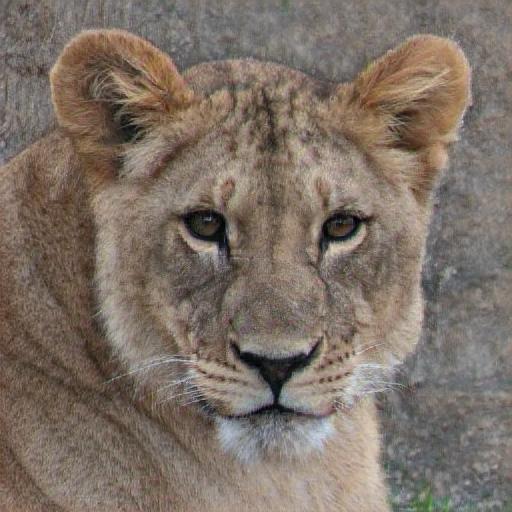} &
		\includegraphics[width=0.38  \columnwidth]{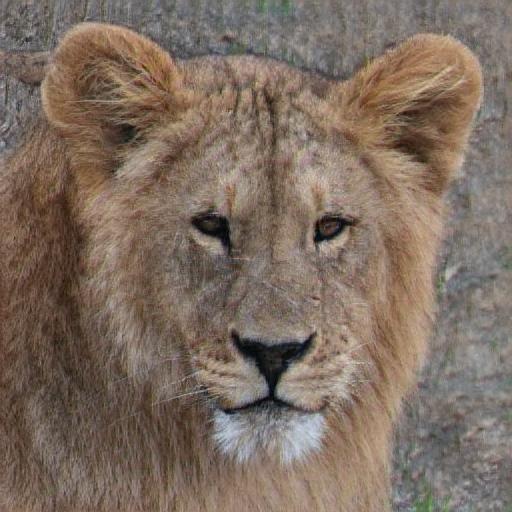} &
		\includegraphics[width=0.38  \columnwidth]{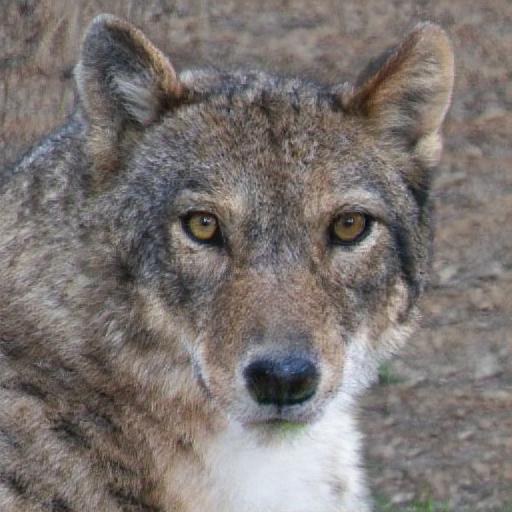} &
		\includegraphics[width=0.38  \columnwidth]{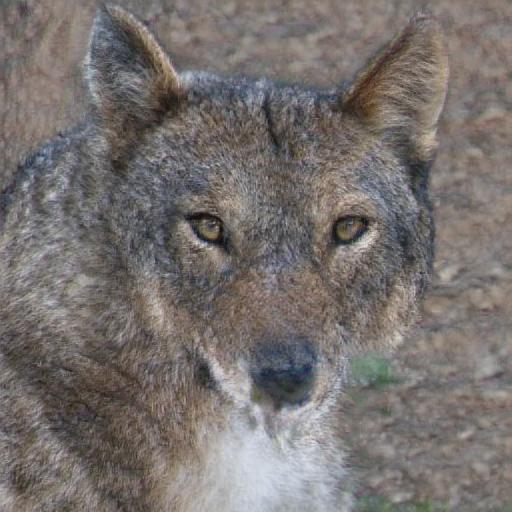} \\
		
		\includegraphics[width=0.38  \columnwidth]{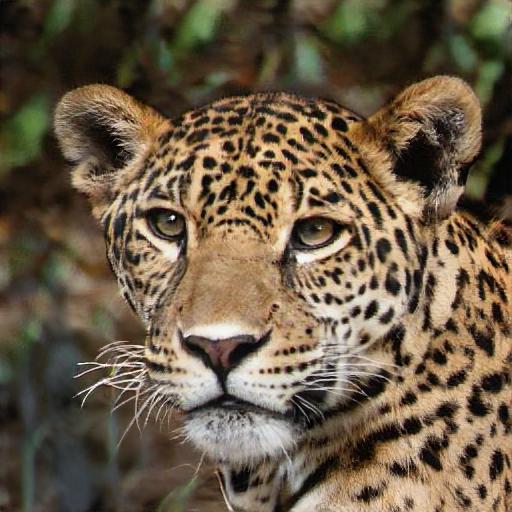} &
		\includegraphics[width=0.38  \columnwidth]{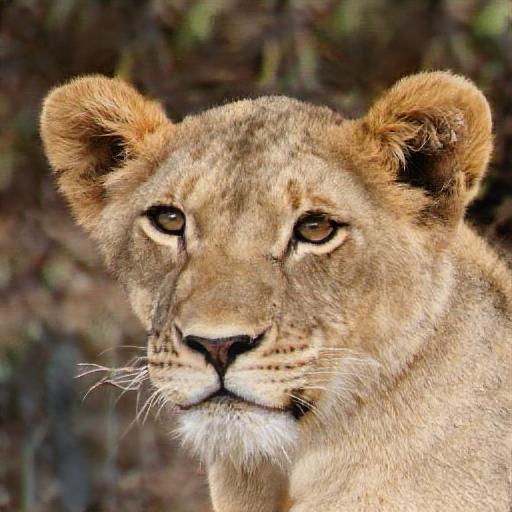} &
		\includegraphics[width=0.38  \columnwidth]{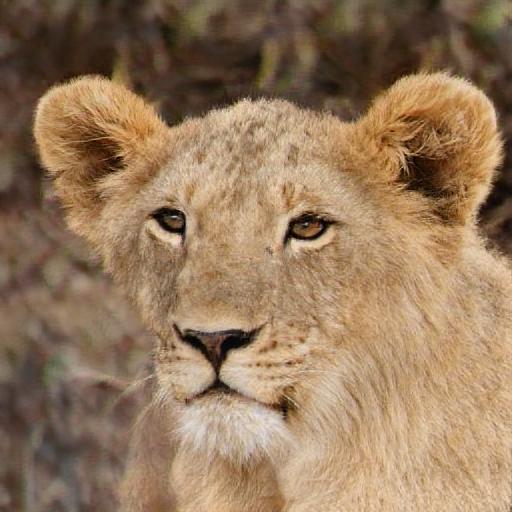} &
		\includegraphics[width=0.38  \columnwidth]{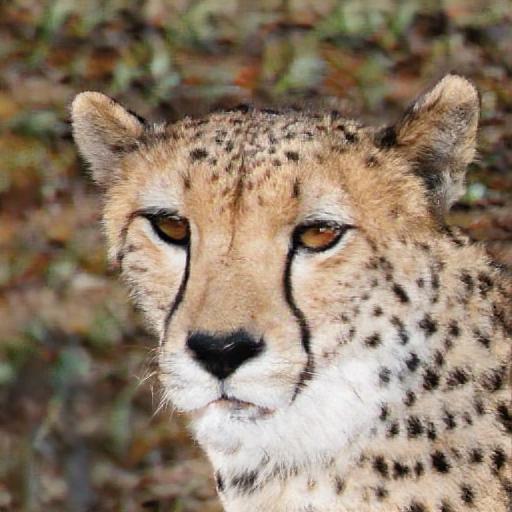} &
		\includegraphics[width=0.38  \columnwidth]{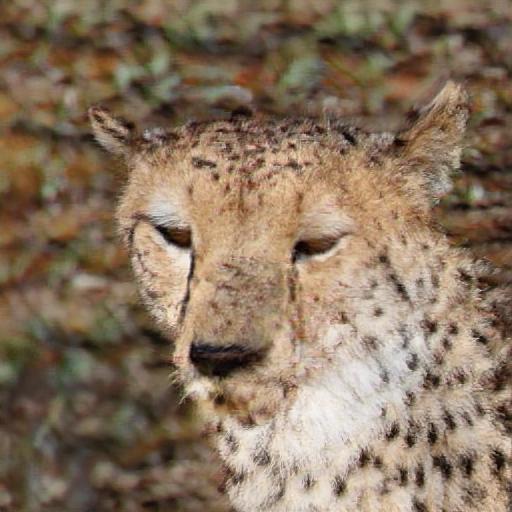} \\
		
		\includegraphics[width=0.38  \columnwidth]{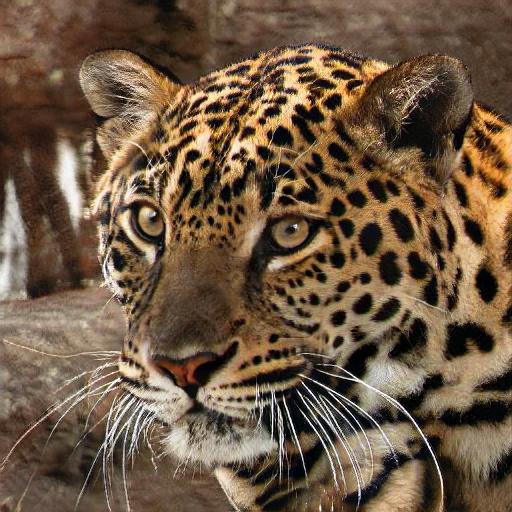} &
		\includegraphics[width=0.38  \columnwidth]{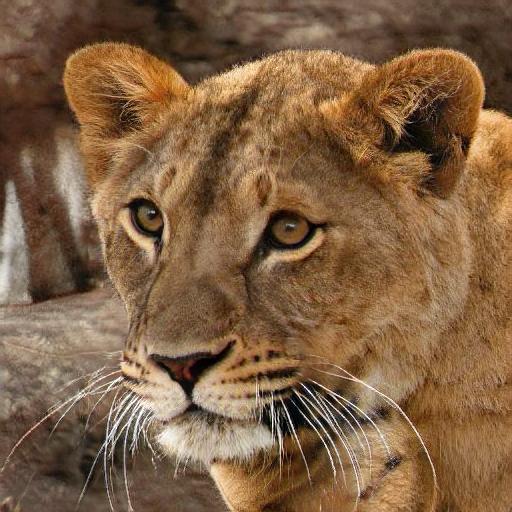} &
		\includegraphics[width=0.38  \columnwidth]{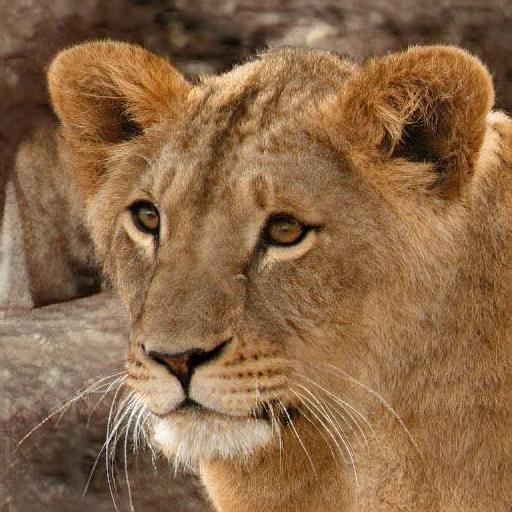} &
		\includegraphics[width=0.38  \columnwidth]{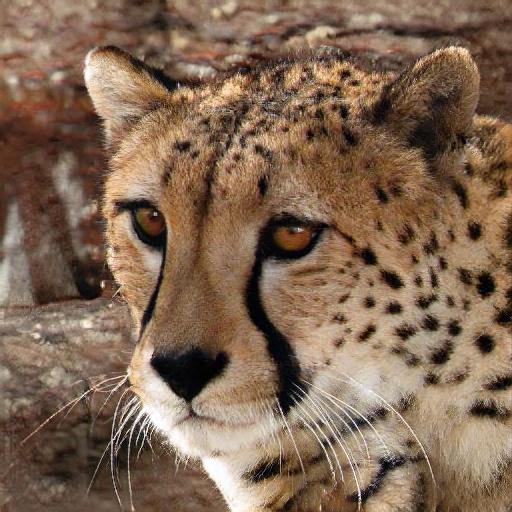} &
		\includegraphics[width=0.38  \columnwidth]{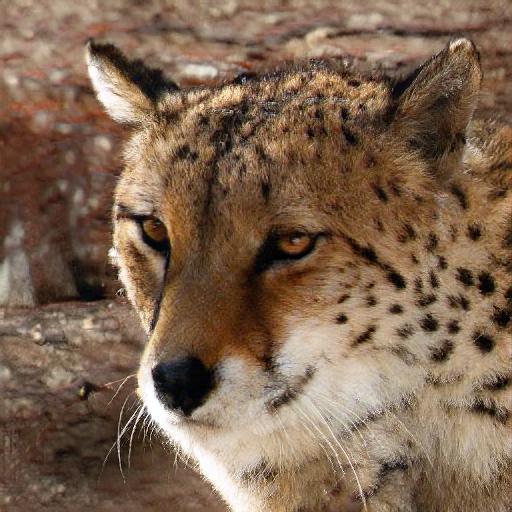} \\

	\end{tabular}
	
	\caption{Drastic manipulations in visually diverse datasets are sometimes difficult to achieve using our global directions. Here we use StyleGAN-ada \cite{karras2020training} pretrained on AFHQ wild \cite{choi2020stargan}, which contains wolves, lions, tigers and foxes. There is a smaller domain gap between tigers and lions, which mainly involves color and texture transformations. However, there is a larger domain gap between tigers and wolves, which, in addition to color and texture transformations, also involves more drastic shape deformations. This figure demonstrates that our global directions method is more successful in transforming tigers into lions, while failing in some cases to transform tigers to wolves.
		The ``+'' and ``++'' indicate medium and strong manipulation strength, respectively.
	}
	\label{fig:tiger}
\end{figure*}

Finally, Figure~\ref{fig:tiger} demonstrates that drastic manipulations in visually diverse datasets are sometimes difficult to achieve using our global directions. Here we use StyleGAN-ada \cite{karras2020training} pretrained on AFHQ wild \cite{choi2020stargan}, which contains wolves, lions, tigers and foxes. There is a smaller domain gap between tigers and lions, which mainly involves color and texture transformations. However, there is a larger domain gap between tigers and wolves, which, in addition to color and texture transformations, also involves more drastic shape deformations. This figure demonstrates that our global directions method is more successful in transforming tigers into lions, while failing in some cases to transform tigers to wolves.

\section{Video}

We show examples of interactive text-driven image manipulation in our supplementary video. We use a simple heuristic method to determine the initial disentanglement threshold ($\beta$). The threshold is chosen such that $k$ channels will be active. For real face manipulation, we set the initial strength to $\alpha=3$ and the disentanglement threshold so that $k=20$. For real car manipulation, we set the initial values to $\alpha=3$ and $k=100$. For generated cat manipulation, we set the initial values to $\alpha=7$ and $k=100$.

\end{document}